\RequirePackage{fix-cm}
\documentclass[twocolumn]{svjour3}          
\smartqed  

\usepackage[colorlinks,linkcolor=blue]{hyperref}
\usepackage[hyphenbreaks]{breakurl} 

\usepackage{cite}
\usepackage[pdftex]{graphicx}
\usepackage{ragged2e}
\usepackage[tight,footnotesize]{subfigure}
\usepackage{rotating}
\usepackage{graphicx}
\usepackage{amsmath,amssymb,amsfonts}%
\usepackage{array}
\usepackage{multirow}
\usepackage{colortbl}
\usepackage{amsfonts}
\usepackage{pifont}
\usepackage{xspace}
\usepackage{etoolbox}
\usepackage{overpic}
\usepackage{color}
\usepackage{microtype}

\usepackage{float}
\usepackage{graphicx}%
\usepackage{multirow}%

\usepackage{mathrsfs}%
\usepackage[title]{appendix}%
\usepackage{xcolor}%
\usepackage{textcomp}%
\usepackage{manyfoot}%
\usepackage{booktabs}%
\usepackage{algorithm}%
\usepackage{algorithmicx}%
\usepackage{algpseudocode}%
\usepackage{listings}%
\usepackage{blindtext}
\usepackage{graphicx} 
\usepackage{multirow} 
\usepackage{booktabs} 
\usepackage{caption}  
\usepackage{amsmath}  
\usepackage[T1]{fontenc}

\usepackage[utf8]{inputenc} 
\usepackage[T1]{fontenc}    
\usepackage{hyperref}       
\usepackage{url}            
\usepackage{booktabs}       
\usepackage{amsfonts}       
\usepackage{nicefrac}       
\usepackage{microtype}      
\usepackage{xcolor}         
\usepackage{graphicx}       
\usepackage{amsmath}
\usepackage{multirow}       
\usepackage{float}
\usepackage{hyperref}
\usepackage{breakurl}

\definecolor{RightColor}{RGB}{229,220,139}
\definecolor{WrongColor}{RGB}{234,198,189}
\definecolor{OtherColor}{RGB}{136,199,193}

\definecolor{FASColor}{RGB}{200,127,20}
\definecolor{DeepFakeColor}{RGB}{79,141,213}
\definecolor{UnifiedColor}{RGB}{254,139,0}

\newcommand{\orcid}[1]{\href{https://orcid.org/#1}{\includegraphics[width=10pt]{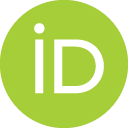}}}

\def\etal{{\em et al}}

\usepackage{hyperref}
\hypersetup{breaklinks=true,citecolor=blue, colorlinks}

\graphicspath{{./Imgs/}}
\DeclareGraphicsExtensions{.pdf,.jpg,.png}

\usepackage{silence}
\journalname{Research Article}

\begin{document}

\title{SHIELD\includegraphics[width=0.8cm]{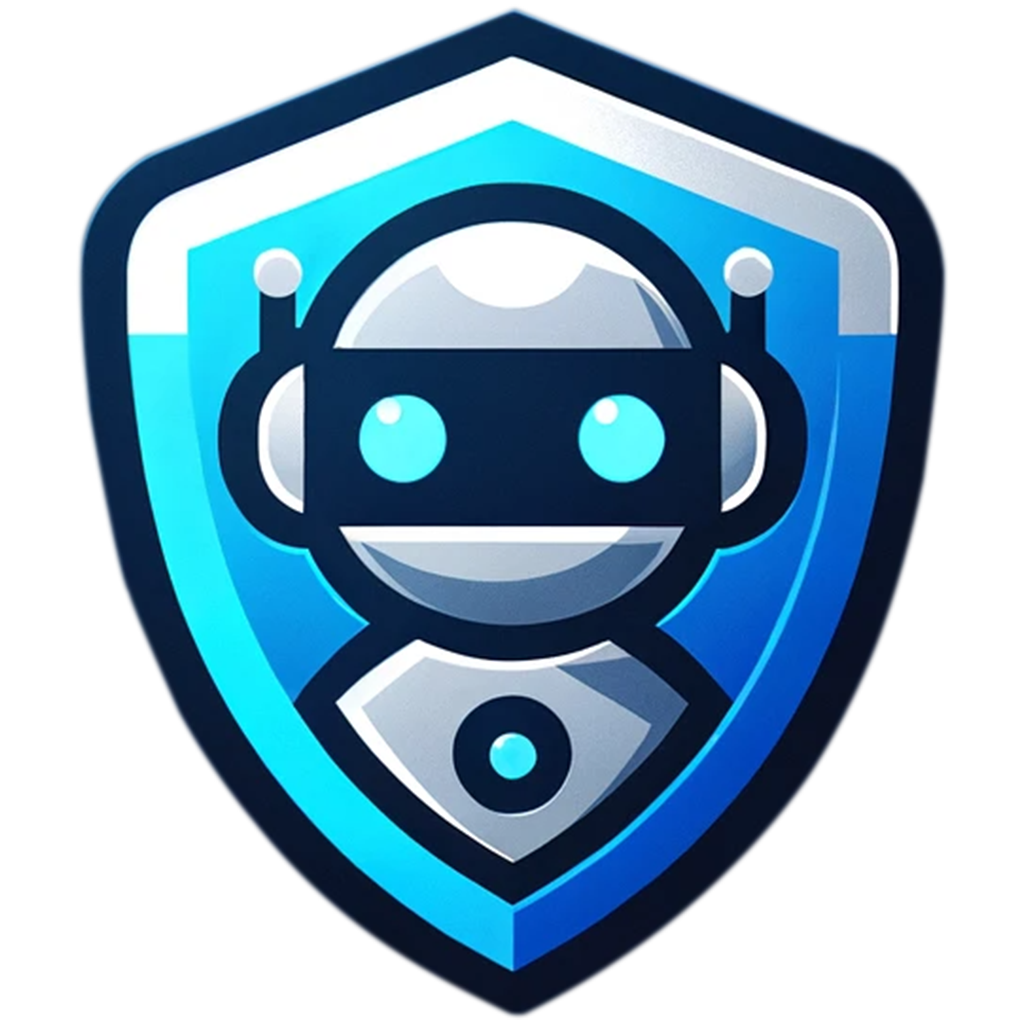}: An Evaluation Benchmark for Face Spoofing and Forgery Detection with Multimodal Large Language Models}

\titlerunning{SHIELD Benchmark}        

\author{Yichen Shi\textsuperscript{†}  \orcid{0000-0000-0000-0000}       \and
  Yuhao Gao\textsuperscript{†}  \orcid{0000-0000-0000-0000} \and 
  Yingxin Lai\textsuperscript{†}  \orcid{0000-0000-0000-0000} \and
  Hongyang Wang \orcid{0000-0000-0000-0000} \and
  Jun Feng \orcid{0000-0000-0000-0000} \and
  Lei He \orcid{0000-0000-0000-0000} \and
  Jun Wan \orcid{0000-0000-0000-0000} \and
  Changsheng Chen \orcid{0000-0000-0000-0000} \and Zitong Yu \orcid{0000-0000-0000-0000} \and Xiaochun Cao \orcid{0000-0000-0000-0000}}

\authorrunning{Y. Shi \etal} 

\institute{
Yichen Shi is with the School of Electronic Information and Electrical Engineering, Shanghai Jiao Tong University, Shanghai, 200030, China
(Email: yichen.shi@sjtu.edu.cn). 
Yingxin Lai is with the Department of Artificial Intelligence, Xiamen University, Xiamen, 361005, Fujian, China (Email: laiyingxin2@gmail.com). \\
Yuhao Gao, Hongyang Wang and Jun Feng are with the School of Information Science and Technology, Shijiazhuang Tiedao University, Shijiazhuang, 050043, China (Email: 1202210044@student.stdu.edu.cn,1202310098@student.s
tdu.edu.cn, jun.feng@stdu.edu.cn). \\
Lei He is with the Electrical Engineering Department, UCLA, 90095, United States
(Email: lei.he@eias.ac.cn). \\
Jun Wan is with the New Laboratory of Pattern Recognition(NLPR), Institute of Automation, Chinese Academy of Sciences, Beijing, 100190, China (Email: jun.wan@nlpr.ia.ac.cn). \\
Changsheng Chen is with the College of Electronics and Information Engineering, Shenzhen University, Shenzhen, 518060, China
(Email: cschen@szu.edu.cn). \\
Zitong Yu is with the School of Computing and Information Technology, Great Bay University, Dongguan, Guangdong, 523000, China 
(Email: yuzitong@gbu.edu.cn). \\
Xiaochun Cao is with the School of Cyber Science and Technology, Shenzhen Campus of Sun Yat-sen University, Shenzhen, 518107, China
(Email: caoxiaochun@mail.sysu.edu.cn). \\
{†}Equal contributors \\
Corresponding author: Zitong Yu.
}

\date{Received: date / Accepted: date}

\maketitle

\begin{abstract}
Multimodal large language models (MLLMs) have demonstrated strong capabilities in vision-related tasks, capitalizing on their visual semantic comprehension and reasoning capabilities. However, their ability to detect subtle visual spoofing and forgery clues in face attack detection tasks remains underexplored. In this paper, we introduce a benchmark, \textit{SHIELD}, to evaluate MLLMs for face spoofing and forgery detection. Specifically, we design true/false and multiple-choice questions to assess MLLM performance on multimodal face data across two tasks. For the face anti-spoofing task, we evaluate three modalities (i.e., RGB, infrared, and depth) under six attack types. For the face forgery detection task, we evaluate GAN-based and diffusion-based data, incorporating visual and acoustic modalities. We conduct zero-shot and few-shot evaluations in standard and chain of thought (COT) settings. Additionally, we propose a novel multi-attribute chain of thought (MA-COT) paradigm for describing and judging various task-specific and task-irrelevant attributes of face images. The findings of this study demonstrate that MLLMs exhibit strong potential for addressing the challenges associated with the security of facial recognition technology applications.
\keywords{Face anti-spoofing (FAS), Face forgery detection, Multimodal large language models (MLLMs), Multi-attribute chain of thought (MA-COT)}
\end{abstract}

\section{Introduction}\label{sec1}
Despite significant progress in face anti-spoofing and face forgery detection, most research still concentrates on developing models for specific scenarios or types of attacks, often relying on subtle facial changes. These models typically focus on a single modality or a specific kind of spoofing attack, which limits their adaptability to a broader and more diverse range of attack scenarios. Traditional approaches in face anti-spoofing and face forgery detection primarily rely on identifying forgery clues across different modalities, such as spatial and frequency domains~\cite{li2021discriminative, shang2021prrnet}, with models specifically trained to recognize these cues. In contrast, current multimodal large language models (MLLMs)~\cite{yang2023dawn, team2023gemini} do not utilize such task-specific features and have not been fine-tuned with dedicated training data for face anti-spoofing (FAS) and face forgery detection. Instead, they are largely trained on generic question-answer pairs and lack datasets tailored to the nuances of FAS and face forgery detection tasks. To bridge this gap, this study explores the capabilities of MLLMs in these tasks and highlights areas for potential fine-tuning.

We introduce a new benchmark, SHIELD, to evaluate the ability of MLLMs in face spoofing and forgery detection. Specifically, we design true/false and multiple-choice questions to evaluate MLLM performance on multimodal face data across these two face security tasks. For the face anti-spoofing 
task, we evaluate three modalities ({i.e., RGB, infrared, and depth) under six types of presentation attacks (i.e., print attack, replay attack, rigid mask, paper mask, flexible mask and fake head). For the face forgery detection task, we evaluate generative adversarial network (GAN)-based and diffusion-based data, incorporating both visual and acoustic modalities. Since our benchmark currently lacks task-specific fine-tuning data for FAS and face forgery tasks, we conduct preliminary tests under zero-shot and few-shot settings using standard prompts and chain of thought (COT) prompts to establish baseline performance.

The overall performance of these MLLMs is depicted in Fig.~\ref{all_per}. The results demonstrate that MLLMs hold substantial promise for facial security tasks. Furthermore, the notable performance differences caused by changes in input modalities highlight the need for further research to fully understand the limitations and potential of these models. Additionally, we propose a novel multi-attribute chain of thought (MA-COT) paradigm for describing and judging various task-specific and task-irrelevant attributes of face images. This paradigm provides rich task-related knowledge, facilitating the detection of subtle spoofing or forgery clues.

Our contributions are threefold:
\begin{enumerate}
    \renewcommand{\labelenumi}{\arabic{enumi})} 
    \item We introduce a new benchmark for evaluating the effectiveness of MLLMs in addressing various challenges within the domain of face security, including both face anti-spoofing and face forgery detection.
    \item We propose a novel MA-COT paradigm, which provides rich task-related knowledge for MLLMs and improves the interpretability of the decision-making process for face spoofing and forgery detection.
    \item Through extensive experiments, we find that 1) MLLMs (e.g., GPT4V and Gemini) have potential real/fake reasoning capability for
    unimodal and multimodal face spoofing and forgery detection; 2) the proposed MA-COT can improve the robustness and interpretability for face attack detection.
\end{enumerate}

\begin{figure*}[t]
\centering
\includegraphics[width=0.9\textwidth]{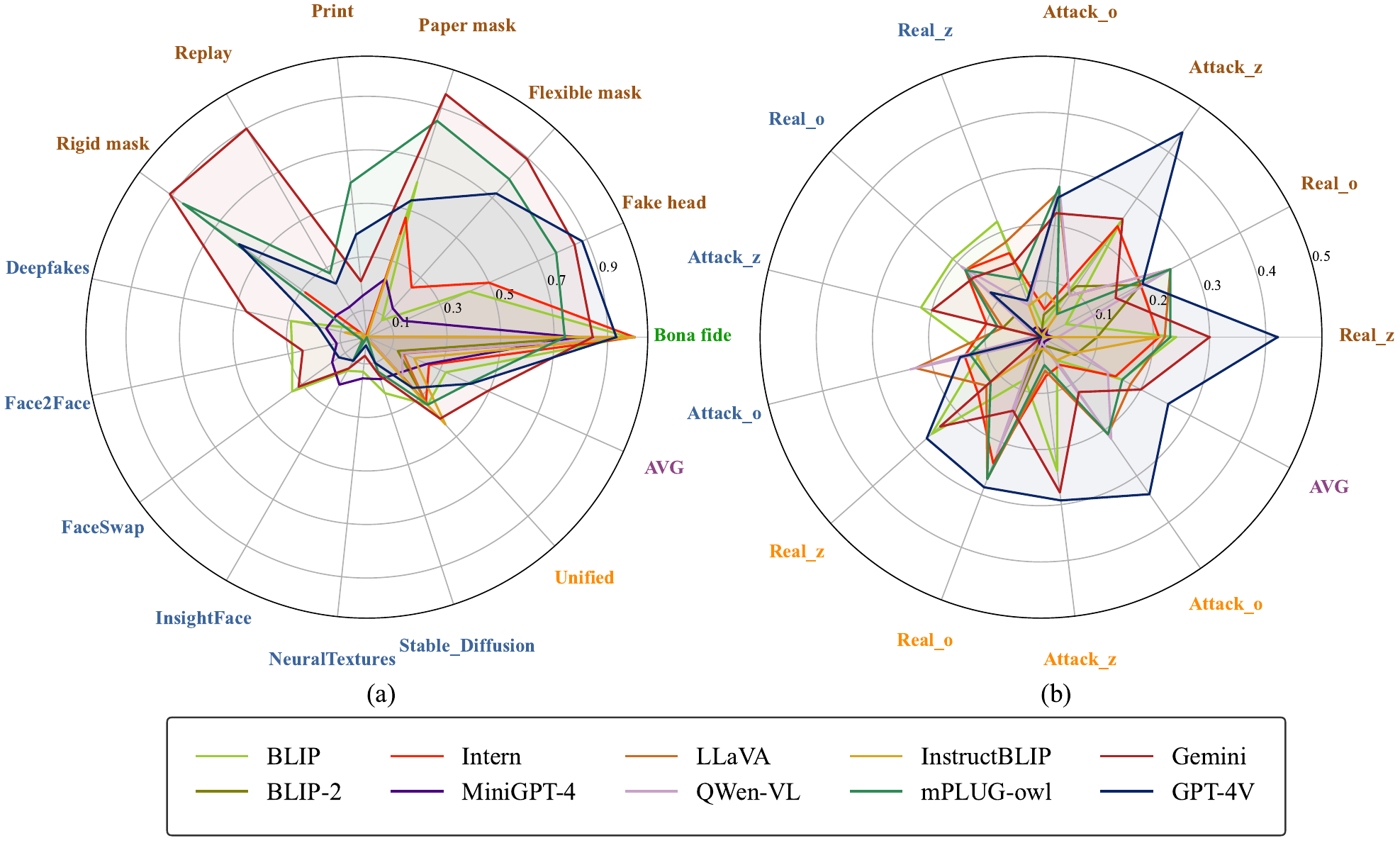}
\caption{Performance of various multimodal large language models on (a) true/false and (b) multiple-choice questions across different types of attacks. The term "bona fide" is used to denote a genuine face image. Print refers to a printed photograph, and replay refers to a replayed video. This demonstrates their superior ability to distinguish between physical and digital attacks. In (a), the larger the area of each colored polygon, the better the performance. Qwen-VL and mPLUG-owl outperform other models. In (b), GPT4V shows the best performance compared to the others. The color brown represents the face anti-spoofing task, deep blue represents the face forgery detection task, and orange represents the joint task. AVG: average.}
\label{all_per}
\end{figure*}

\section{Related Work}

\label{related_work}

\subsection{Face Anti-Spoofing}

In the early stage, face anti-spoofing systems relied on traditional techniques. These methods mainly focused on handcrafted features~\cite{boulkenafet2015face,Boulkenafet2017Face,Komulainen2014Context,Patel2016Secure} and heuristics designed to detect presentation attacks. They were effective against basic spoofing attempts but often failed to generalize to more sophisticated and diverse attack scenarios. With the advent of deep learning, there was a significant shift in FAS methods. Deep learning-based techniques~\cite{qin2019learning,Liu2018Learning,yang2019face,Atoum2018Face,Gan20173D,george2019deep,yu2020fas2}, particularly those employing convolutional neural networks (CNNs) with advanced learning strategies (e.g., meta-learning~\cite{qin2021meta}, disentangled learning~\cite{wang2020cross}, and few-shot learning~\cite{qin2019learning}), have become dominant in recent years. These methods leveraged large-scale datasets and powerful computational resources to learn more complex and subtle features of genuine and spoofed faces. This shift marked a substantial improvement over traditional methods in both detection accuracy and generalization capabilities.
Recently, FAS methods have started to be integrated into larger and multimodal models with vision transformers (ViTs)~\cite{dosovitskiy2020image}. On the one hand, ViT is adopted in the spatial domain~\cite{ming2022vitranspad,george2020effectiveness,wang2022face,liuma} to explore relationships between live and spoof patches at a local level. On the other hand, global features such as temporal abnormity~\cite{wang2022learning} or physiological periodicity~\cite{yu2021transrppg,yu2024benchmarking} are extracted by applying ViT in the temporal domain. Despite convincing performance via unimodal and multimodal deep learning methods based on CNNs and ViTs,  there is currently no work exploring the potential of MLLMs for generalized FAS tasks.

\subsection{Face Forgery Detection}
Previous studies have predominantly approached this task as a binary classification problem~\cite{cao2022end,chen2021local,haliassos2021lips}. CNNs are commonly employed to extract high-dimensional features from input face images and then classify them as either real or fake. While these models exhibit commendable performance on in-domain datasets due to their utilization of facial features, their performance on cross-domain datasets remains inadequate. To overcome this limitation, recent studies have explored alternative approaches, including the utilization of low-frequency information in frequency domain analysis. For instance, noise-based approaches~\cite{gu2022exploiting,mat}, frequency domain analysis~\cite{qian2020thinking,freq1_icml,twobranch}, and local relation learning methods~\cite{chen2021local,facexray,li2021discriminative} have been investigated. These methods show promising results by incorporating additional information beyond traditional frequency domain or convolutional network-based methods. However, it is important to note that real-world image forgery scenarios are characterized by diverse patterns and unknown environmental conditions. The original techniques based on frequency domain information or convolutional networks tend to overfit existing image forgery techniques in the training set, leading to significant performance degradation for these methods\cite{shang2021prrnet,fang2024ieirnet}. Recently, vision foundation models (e.g., segment anything models~\cite{kirillov2023segment}) have been successfully introduced in face forgery detection\cite{lai2023detect}, demonstrating their strong attck localization capabilities. Given the rapid evolution of face forgery attacks and the need for robust detection methods, the powerful zero-shot generalization capacity of multimodal large language models (MLLMs) makes them a promising avenue for exploration. This study aims to evaluate whether MLLMs can effectively and robustly detect diverse face forgery attacks. Additionally, cross-domain generalization techniques, inspired by related fields such as speech spoofing detection, have been explored to enhance robustness under varying conditions~\cite{wang2024generalizable}.

\subsection{Multimodal Large Language Model}

MLLMs have developed at a rapid pace over the past few years due to significant contributions from various research teams. Alayrac~et~al.~\cite{alayrac2022flamingo} introduced the Flamingo model, a significant advancement in processing interleaved visual data and text, focusing on generating free-form text output. Following this, Li~et~al.~\cite{li2023blip} developed BLIP-2, a model characterized by its resource-efficient framework and the innovative Q-Former, which notably leverages frozen LLMs for efficient image-to-text generation. In a similar vein, Dai~et~al.~\cite{dai2305instructblip} further refined this approach with Instruct BLIP, a model trained on the BLIP-2 framework, specifically enhancing the capability of instruction-aware visual feature extraction. Developed by OpenAI, GPT4V~\cite{achiam2023gpt}  itself represents a leap forward in MLLMs, offering versatility in generating coherent and context-aware text. It is applicable in standalone text-based applications as well as multimodal settings. Its superior performance establishes a dominant position in terms of versatility and general applicability. Zhu~et~al.~\cite{zhu2023minigpt} introduced MiniGPT-4, a streamlined approach that aligns a pre-trained vision encoder with LLMs by training a single linear layer, effectively replicating the capabilities of GPT4V. To extend the linguistic versatility, Bai~et~al.~\cite{bai2023qwen} presented Qwen-VL, a multilingual MLLM supporting both English and Chinese, with the ability to process multiple images during training. In the area of multimodal data scarcity, Liu~et~al.~\cite{liu2023visual} pioneered LLaVA by introducing a novel open-source multimodal instruction-following dataset alongside the LLaVA-Bench benchmark. Recently, Google's newly introduced Gemini~\cite{team2023gemini} represents a significant advancement in the intrinsic multimodal characteristics of artificial intelligence(AI) systems. It processes text and various audiovisual inputs, generating outputs in multiple formats, thereby demonstrating the efficiency of integrated multimodal interactions. Gemini also demonstrates exceptional multimodal interaction efficiency and generalization capabilities, posing a significant challenge to GPT4V~\cite{yang2023dawn}. Despite their remarkable problem-solving capabilities in various vision domains (e.g., generic object recognition and grounding), whether MLLMs are sensitive to subtle visual spoofing clues and how they perform in the face attack detection domain remains unexplored.  

\subsection{Existing MLLM Benchmark}
In recent years, benchmarks have been developed for MLLMs to evaluate their performance on various tasks. A notable benchmark in this regard is the multimodal model evaluation (MME)~\cite{fu2023mme}. The MME benchmark focuses on a wide range of tasks to assess the capabilities of different MLLMs, including both perceptual and cognitive tasks, covering areas such as object recognition, common sense reasoning, numerical computation, text translation, and code reasoning. This benchmark evaluates several advanced MLLM models, such as BLIP-2~\cite{li2023blip}, MiniGPT-4~\cite{zhu2023minigpt}, and mPLUG-Owl~\cite{ye2023mplug}, using accuracy (ACC) metrics. The former is calculated based on the correct response to each question, while the latter is a stricter measure that requires correct answers to both questions associated with each image. Besides, several benchmarks for different perspectives (hierarchical capabilities~\cite{li2023seed}, hallucination~\cite{guan2023hallusionbench}, and style-shift robustness~\cite{cai2023benchlmm}) and different tasks (e.g., low-level image enhancement~\cite{wu2023q} and image quality assessment~\cite{huang2024aesbench}) have also been established. In terms of face attack detection tasks, efforts are still needed to qualitatively and quantitatively evaluate sets of prompts and MLLM models, and to establish a fair evaluation benchmark.

\section{Task Design}
\label{appendix:task design}

We establish different true/false sub-tasks to assess the ability of MLLMs to discriminate between real faces and spoof faces. The sub-tasks are as follows:
\begin{enumerate}
    \renewcommand{\labelenumi}{\arabic{enumi})} 
    \item Zero-shot testing$:$ <image>, question$:$ Is this image a real face? Please answer yes or no!
    \item Zero-shot testing (COT)$:$ <image>, question$:$ Is this image a real face? Please describe the image and answer yes or no!
    \item Few-shot testing$:$ <image, image>, question$:$ The first image is a real face, is the second image a real face? Please answer yes or no!
    \item Few-shot testing (COT)$:$ <image, image>, question$:$ The first image is a real face, is the second image a real face? Please describe the image and answer yes or no!
\end{enumerate}

\textbf{Multiple-choice questions.}
We create several multiple-choice sub-tasks to assess the ability of MLLMs to perceive and understand the distinctions between multiple face images. The sub-tasks are presented as follows:
\begin{enumerate}
    \renewcommand{\labelenumi}{\arabic{enumi})} 
    \item Zero-shot testing$:$ <image$\times4$>, question$:$ The following images are listed in the order A, B, C, D. Please answer the letter number of A, B, C, or D corresponding to the image of the real face.
    \item Zero-shot testing (COT)$:$ <image$\times4$>, question$:$ The following images are listed in the order A, B, C, D. Please describe the images and answer the letter number of A, B, C, or D corresponding to the image of the real face.
    \item Few-shot testing$:$ <image$\times5$>, question$:$ The first image is a real face. The following images are listed in the order A, B, C, D. Please answer the letter number of A, B, C, or D corresponding to the image of the real face.
    \item Few-shot testing (COT)$:$ <image$\times5$>, question$:$ The first image is a real face. The following images are listed in the order A, B, C, D. Please describe the images and answer the letter number of A, B, C, or D corresponding to the image of the real face.
\end{enumerate}


\section{SHIELD}
\begin{figure*}[ht]
  \centering
  \begin{center}
  \centerline{\includegraphics[width=0.8\linewidth]{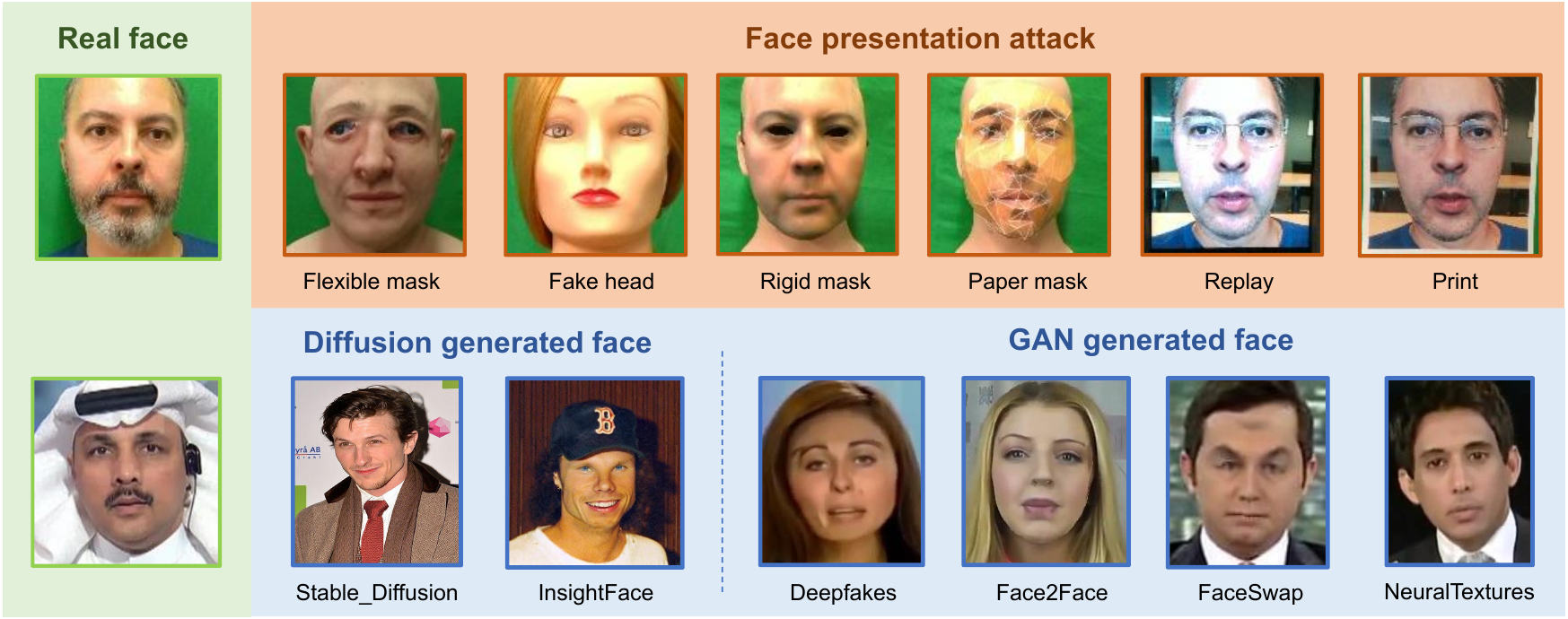}}
  \end{center}
    \caption{Examples of our collected datasets. The images are sourced from the WMCA~\cite{george2019biometric} and the FF++~\cite{rossler2019faceforensics} datasets
    }
    \label{attack}
\end{figure*}

\subsection{Data Collection}
\textbf{Collection of face anti-spoofing datasets.}
As shown in Table~\ref{Data statistics}, we conducted experiments using the WMCA~\cite{george2019biometric} and SiW-Mv2~\cite{siW2022Multi} datasets, which include a variety of presentation attacks and multiple modalities (e.g., RGB images, depth maps, infrared images, and thermal imaging). For each identity (ID), we selected six types of attacks and the bona fide faces, along with their corresponding RGB images, depth maps, and infrared images for experiments. Typical examples are presented in the first row of Fig.~\ref{attack}.

\textbf{Collection of face forgery detection datasets.}
For face forgery detection, we evaluated MLLMs on the popular FaceForencics++ (FF++)~\cite{rossler2019faceforensics} dataset, which includes four types of forgery techniques (i.e., Deepfakes, Face2Face, FaceSwap, and Nulltextures).  Additionally, as AI-generated content (AIGC) has rapidly advanced, the resulting face images have become increasingly realistic. To account for this, we also evaluated AIGC-based face data using the DFF dataset~\cite{song2023robustness}, generated by Stable Diffusion, Inpainting, and InsightFace techniques. The images were chosen using "The Dawn of LMMs$:$ Preliminary Explorations with GPT4V"~\cite{alayrac2022flamingo} as a guide, and our sample is simultaneously representative and diverse. Typical samples are shown in the second row of Fig.~\ref{attack}.

\begin{table}[h!]
    \centering
    \caption{Data statistics. FAS: face anti-spoofing. The term "bona fide" is used to denote a genuine face image. Print refers to a printed photograph, and replay refers to a replayed video.}  
\label{Data statistics}
    \begin{tabular}{l l l c}
        \toprule
        \textbf{Task} & \textbf{Dataset} & \textbf{Attack type} & \textbf{Count} \\
        \midrule
        \multirow{7}{*}{FAS} 
            & & Bona fide & 100 \\
            & & Fake head & 100 \\
            & WMCA (352)& Flexible mask & 100 \\
            & Siw\_MV2 (348)& Paper mask & 100 \\
            & & Print & 100 \\
            & & Replay & 100 \\
            & & Rigid mask & 100 \\
        \midrule
        \multirow{7}{*}{Deepfake} 
            & & Real & 200\\
            & & Deepfakes & 100\\
            & FF++ (500) & Face2Face & 100 \\
            & DFF (300) & FaceSwap & 100 \\
            & & InsightFace & 100 \\
            & & NeuralTextures & 100 \\
            & & Stable\_Diffusion & 100 \\
        \bottomrule
    \end{tabular}
\end{table}

\subsection{Task Design}

As shown in Fig.~\ref{task design}, we designed two tasks to test the capabilities of MLLMs in the field of facial security$:$ true/false questions and multiple-choice questions. For each task, we conducted zero-shot tests and in-context few-shot tests. For each type of test, we experimented with both standard settings and COT settings. The overall pipeline of the experiment is shown in the lower half of Fig.~\ref{macot-introduction}.
\begin{figure*}[ht]
  \centering
  \begin{center}
  \centerline{\includegraphics[width=0.9\linewidth]{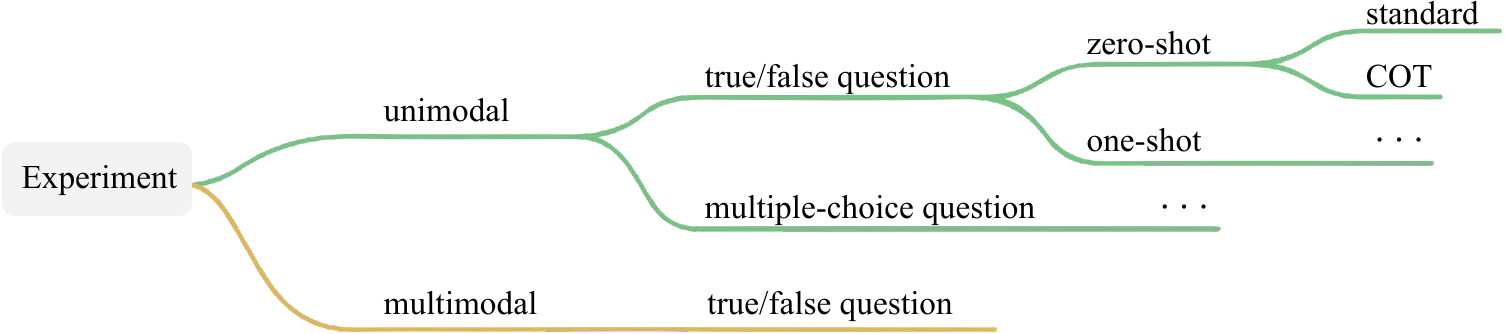}}
  \end{center}
    \caption{Pipeline of task design. The ellipses indicate that the structures are consistent with the task design framework shown above. COT: chain of thought}
  \label{task design}
\end{figure*}

\subsection{MA-COT}

Chain of thought (COT) technology represents a recent advancement in prompt learning, primarily designed to enhance the reasoning capabilities of MLLMs~\cite{wei2022chain}. In this study, we propose a novel COT paradigm, termed MA-COT. This approach draws inspiration from prior work in visual COT~\cite{wu2023role}, where an image is first described in detail and subsequently subjected to judgment. Specifically, in the domains of face anti-spoofing and deepfake detection, we incorporate relevant prior knowledge to improve the analysis process.

Unlike traditional COT methods that describe the entire image, our MA-COT paradigm emphasizes the consideration of multiple attributes of an image, such as shape, color, and texture, to provide a more in-depth analysis and judgment. The attributes selected for MA-COT are carefully chosen based on their high-level representational capacity to capture essential features of face forgery. 

The analysis results of individual attributes are synthesized into a comprehensive assessment, increasing both the accuracy and reliability of the overall decision-making process. Additionally, this approach mitigates instances where the model might otherwise refuse to provide an answer, ensuring a more robust and interpretable outcome.

In MA-COT, we have a set of tasks: $\{ \text{task}_1, \text{task}_2, \ldots, \\\text{task}_n \}$. Each task$_i$ is associated with a set of attributes: $\{ \text{attr}_{i1}, \text{attr}_{i2}, \ldots, \text{attr}_{im} \}$.For each attribute, we obtain a description, forming a multi-attribute description set $Description_\text{attr}$. Based on this set, we generate the final answer:
\begin{equation}
P(\text{Answer} | \text{Image}, \text{Question}, \text{Description}_\text{attr}).
\end{equation}
The selected attributes encompass both shared and task-specific characteristics, with shared attributes facilitating differentiation between real faces and attacks, while task-specific attributes improve detection accuracy for distinct forgery scenarios. For this study, the analysis remains qualitative, relying on MLLM’s reasoning capabilities rather than explicit attribute labels within the test data. Our proposed MA-COT framework is designed to be flexibly expandable, accommodating additional tasks and combinations of attributes, making it applicable to broader domains. For the unified detection task, inspired by research in domain generalization and domain adaptation~\cite{yichen2022out,wang2022domain,shao2019multi}, we have established both task-shared and task-specific attributes, as shown in Table~\ref{Attributes-set}. Shared attributes help distinguish between real faces and various attacks, while task-specific attributes enhance the detection capabilities.

\begin{figure*}
  \centering
  \begin{center}
  \centerline{\includegraphics[width=0.65\linewidth]{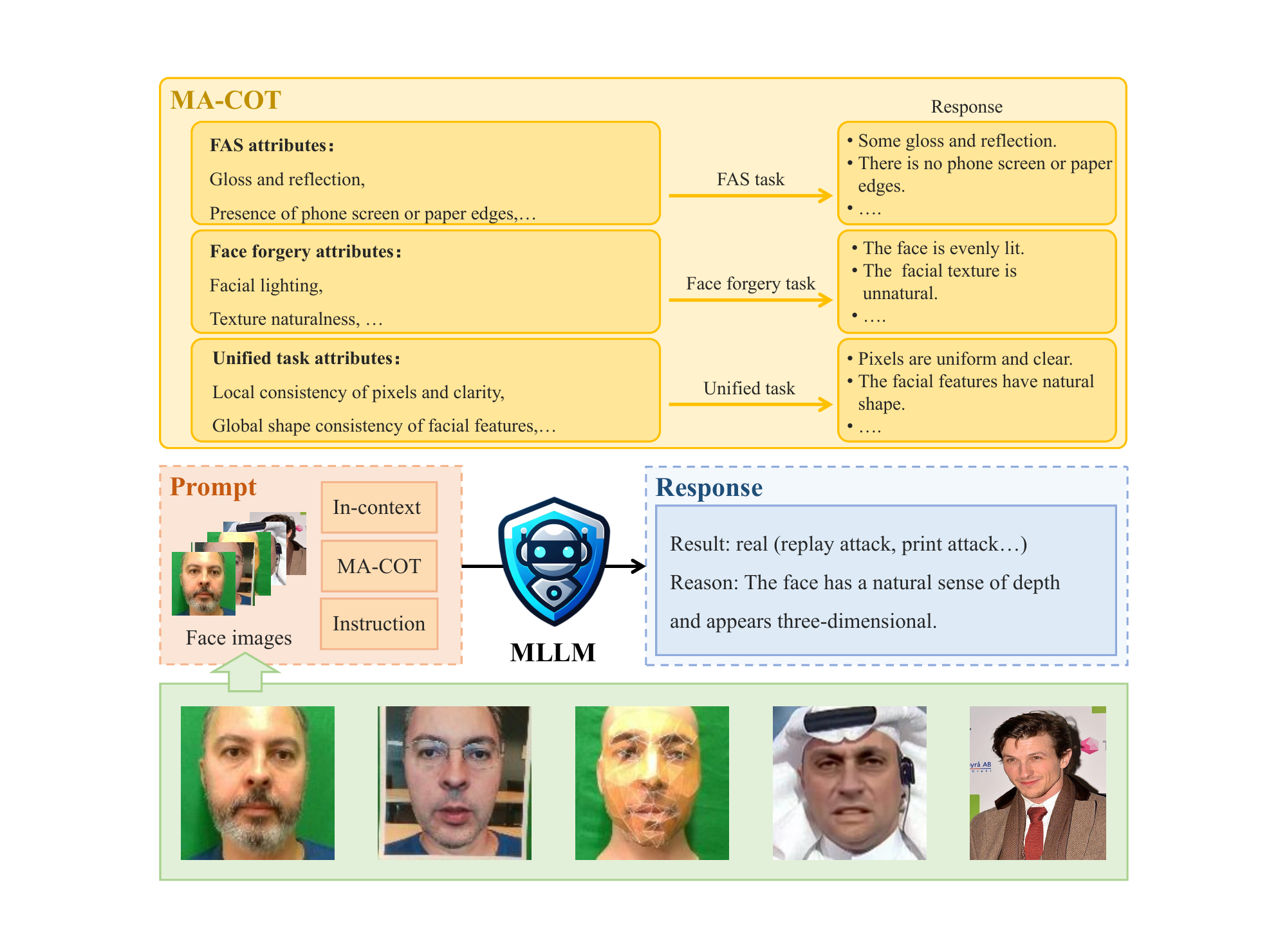}}
  \end{center}
  \caption{The MA-COT process. This process is designed to extract relevant key attributes for various tasks and input these attributes along with the face images under evaluation into MLLMs all attributes we used are shown in Table~\ref{Attributes-set}. This approach aims to guide the MLLM to analyze the images from multiple perspectives, thereby identifying potential clues of attacks and determining whether the images are of real faces. The illustration provides examples of key attribute extraction and its application scenarios in separate FAS, separate face forgery detection, and unified face spoof \& forgery detection. The images are sourced from the WMCA~\cite{george2019biometric} and the FF++~\cite{rossler2019faceforensics} datasets. MA-COT: multi-attribute chain of thought}
  \label{macot-introduction}
\end{figure*}

\section{Experiments}
\label{experiment}
\subsection{Eexperimental Setup}
We conducted tests on the most advanced MLLMs currently available: BLIP~\cite{blip}, BLIP-2~\cite{li2023blip}, Intern~\cite{Intern}, MiniGPT-4~\cite{zhu2023minigpt}, LLaVA~\cite{llava}, QWen-VL~\cite{bai2023qwen}, InstructBLIP~\cite{dai2305instructblip}, mPLUG-owl~\cite{ye2023mplug}, Gemini~\cite{team2023gemini} and GPT4V~\cite{gpt4}. For close source MLLMs, all tests were performed through API calls. We have developed various test scenarios to comprehensively evaluate the effectiveness and accuracy of these models in performing various facial security tasks. 


\subsection{Evaluation Metrics}
We choose the commonly used half total error rate (HTER)~\cite{hter} in face anti-spoofing and the common ACC metric in classification problems as the measures for true/false questions. For multiple-choice questions, we have selected ACC as the metric. These metrics were selected based on their suitability for binary classification tasks, which aligns with the binary label structure of existing face forgery datasets. Given the current stage of MLLMs in this area, these metrics provide a straightforward way to benchmark their baseline performance and identify areas for improvement. The results presented in Sect.~\ref{res_fas} and Sect.~\ref{res_df} are aggregated averages of all attack types and do not detail the performance of each individual attack by different MLLMs.


\section{Results on Face Anti-Spoofing}
\label{res_fas}

\subsection{Prompt Design}
As mentioned in Ref.~\cite{srivatsan2023flip}, using different descriptive statements as image labels to represent real faces and spoofing attacks can affect the models' predictions. As shown in Fig.~\ref{faspromptdesign}, we tested different descriptive methods for real people and spoofs. The results indicate that using the phrase "real face" for a real face and "spoof face" for a spoofing attack gives the best effect. In addition, a combined query of “Is this image a real face or a spoof face? And please answer ‘this image is a real face' or ‘this image is a spoof face'” is more effective.
\subsection{Unimodal FAS Testing}
RGB-based unimodal face data are most common in face anti-spoofing tasks, primarily due to their ease of acquisition. In this study, we conducted a thorough testing of single-modality facial data, which included both true/false and multiple-choice questions. Inspired by the findings in Ref.~\cite{wu2023role}, we discovered that the use of the COT technique in visual-language tasks (i.e., first describing the image, and then making a decision) significantly enhances the model's performance. Accordingly, we carried out standard testing and COT testing for each question. 
\begin{table*}[h!]
\caption{Attributes set}
\label{Attributes-set}
\centering
 \begin{tabular}{c c} 
 \hline
 Attribute & Task  \\
 \hline
 Local consistency of pixels and clarity & Task shared \\
 Global shape consistency of facial features & Task shared            \\
    Sense of depth and three-dimensionality     & Face anti-spoofing     \\
    Gloss and reflection                        & Face anti-spoofing     \\
    Presence of phone screen or paper edges     & Face anti-spoofing     \\
    Texture naturalness    & Face forgery detection \\
    Facial lighting                        & Face forgery detection \\
    Facial skin color                           & Face forgery detection \\
    Eye-head and ear movements                  & Face forgery detection \\
 \hline
 \end{tabular}
\end{table*}
\begin{figure*}[!htp]
  \centering
  \begin{center}
  \centerline{\includegraphics[width=0.8\linewidth]{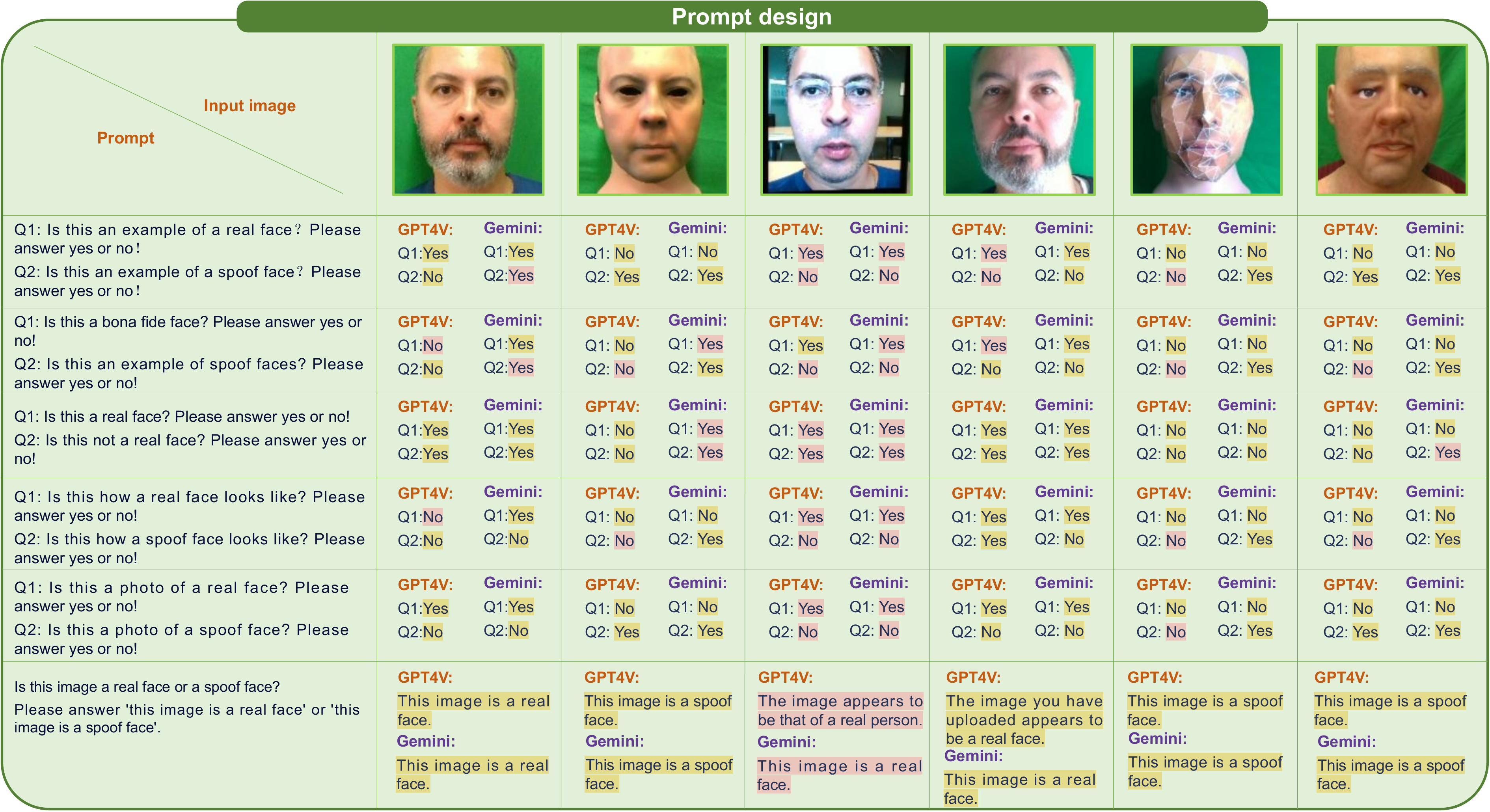}}
  \end{center}
  \caption{Prompt design. The diagram represents a matrix of test results for selecting prompts. On the left are the candidate prompts, and along the top are the test cases used to evaluate the selection of prompts, which include real face, rigid mask attack, replay attack, paper mask attack and flexible mask attack. The responses from GPT4V and Gemini are included. Yellow(red) highlights the correct (incorrect) responses. The images are sourced from the WMCA~\cite{george2019biometric} and the FF++~\cite{rossler2019faceforensics} datasets }
  \label{faspromptdesign}
\end{figure*}

\textbf{True/false questions.}
For face anti-spoofing, we chose to use the phrases "real face" to represent a real face and "spoof face" to signify a spoofing attack. We made three types of judgment in the decision-making process: independent inquiry and combined inquiry. We focus on the issue of unimodal authenticity determination. The crux of this problem lies in the in-depth analysis of one to two images to ascertain whether they are authentic human facial images.

Table~\ref{fasTF} displays the comprehensive performance of various MLLMs on true/false questions under zero-shot and one-shot conditions. While mPLUG-Owl demonstrated high accuracy across all tests, it is notable that the one-shot condition without COT did not lead to any performance improvement. Although the GPT4V did not excel in most of the tests, it achieved consistent results across all tasks, particularly in the one-shot setting. For most MLLMs, using in-context learning (ICL) as a reference failed to improve performance, whereas employing COT techniques often resulted in enhanced outcomes.

When analyzing MLLMs’ performance on FAS tasks, it becomes evident that adding context in one-shot settings does not always lead to improved accuracy. Some models struggle to effectively integrate new information, resulting in cognitive overload or confusion, which can negatively impact their predictions. This phenomenon is evident in cases where models perform well under zero-shot conditions but show reduced accuracy when additional context is provided. Such outcomes suggest
\begin{table*}[h!]
  \caption{Performance metrics of various multimodal large language models(MLLMs) on face anti-spoofing(FAS) true/false questions. $\uparrow$/$\downarrow$ indicates that higher/lower scores are better.}
  \label{fasTF}
  \scriptsize
  \setlength{\tabcolsep}{2.2pt}
  \centering
  \begin{tabular}{ccccccccc}
    \toprule
    \multirow{2}{*}{Model} & \multicolumn{2}{c}{Zero-shot} & \multicolumn{2}{c}{One-shot} & \multicolumn{2}{c}{Zero-shot (COT)} & \multicolumn{2}{c}{One-shot (COT)} \\
    \cmidrule(r){2-9}
    & ACC (\%)$\uparrow$ & HTER (\%)$\downarrow$ & ACC (\%)$\uparrow$ & HTER (\%)$\downarrow$ & ACC (\%)$\uparrow$ & HTER (\%)$\downarrow$ & ACC (\%)$\uparrow$ & HTER (\%)$\downarrow$ \\ 
    \midrule
    BLIP~\cite{blip} & 18.4 & 47.6 & 43.6 & 32.9 & 25.9 & 43.3 & 74.7 & 14.8 \\
    BLIP-2~\cite{li2023blip} & 14.3 & 50.0 & 9.0 & 68.5 & 14.3 & 50.0 & 0.4 & 98.5 \\
    Intern~\cite{Intern} &57.6 & 24.8 & 14.3 & 50.0 & 56.4 & 25.4 & 17.7 & 48.0 \\
    MiniGPT-4~\cite{zhu2023minigpt} & 20.6 & 49.3 & 28.9 & 56.9 & 27.7 & 43.4 & 31.0 & 65.7 \\
    LLaVA~\cite{llava} & 14.3 & 50.0 & 14.3 & 50.0 & 14.3 & 50.0 & 5.4 & 81.0 \\
    QWen-VL~\cite{bai2023qwen} & 14.3 & 50.0 & 14.3 & 50.0 & 14.3 & 50.0 & 14.3 & 50.0 \\
    InstructBLIP~\cite{dai2305instructblip} & 23.7 & 44.5 & 17.1 & 48.3 & 42.3 & 33.7 & 15.9 & 49.1 \\
    mPLUG-owl~\cite{ye2023mplug} & 82.0 & 10.9 & 53.0 & 60.3 & 81.7 & 11.5 & 82.6 & 42.7 \\
    Gemini~\cite{team2023gemini} & 73.4 & 15.5 & 62.0 & 36.3 & 77.0 & 14.7 & 13.9 & 90.3 \\
    GPT4V~\cite{gpt4} & 53.2 & 27.7 & 66.8 & 28.8 & 68.7 & 18.1 & 50.3 & 36.8\\
    \bottomrule
  \end{tabular}
\end{table*}
limitations in the models’ reasoning capabilities or their ability to effectively process and apply new information.

Moreover, many models had accuracies of less than 50\% across different test conditions, which can primarily be attributed to systematic biases or inherent misjudgments rather than random guessing. This indicates that the complexity of the FAS task may exceed the cognitive processing capabilities of certain models, leading to systematic errors in specific scenarios. Additionally, some models showed identical accuracies across multiple conditions. This consistency may stem from dataset homogeneity, similarities in model architecture, or limitations in the training data.


\textbf{Multiple-choice questions.}
We also test face anti-spoofing in the form of an image selection task, with multiple images as input, and ask the model to identify which image is a real person or which one is a spoofing attack. We conduct both zero-shot and few-shot tests, each including standard experiments and COT experiments.
We focus on an in-depth exploration of the unimodal multiple-choice problem. The fundamental task is to perform a detailed analysis of four to five images provided, with the aim of selecting authentic human faces or identifying corresponding attack methods. This process entails not only cognitive analysis of the image content, but also a comprehensive assessment of the image's authenticity, aiming to accurately distinguish between real human faces and various potential attack scenarios.


Table~\ref{fasMC} lists the accuracy of different MLLMs on FAS multiple-choice problems. It separately addresses two types of questions: identifying real faces and spoof faces, with each type tested under two conditions: zero-shot and one-shot. Finally, the average accuracy (AVG) across all conditions is provided. GPT4V achieved the highest average accuracy under all conditions, reaching 33.1\%. Gemini performed best in recognizing attack under zero-shot conditions (25.6\%) and also demonstrated strong overall performance with an average accuracy of 23.3\%. ICL did not significantly improve the performance, indicating that the ICL capability of MLLMs for images still needs to be improved.

\begin{table*}[!htbp]
  \caption{Accuracy of various MLLMs on FAS multiple-choice questions}
  \label{fasMC}
  \scriptsize
  \centering
  \begin{tabular}{ccccccccccc}
    \toprule
    \multirow{3}{*}{Model} & \multicolumn{10}{c}{ACC(\%)} \\ 
    
    \cmidrule(r){2-11}
    & \multicolumn{2}{c}{Zero-shot (real)} & \multicolumn{2}{c}{One-shot (real)} & \multicolumn{2}{c}{Zero-shot (attack)} & \multicolumn{2}{c}{One-shot (attack)} & \multicolumn{2}{c}{AVG} \\
    \cmidrule(r){2-11}
    & - & COT & - & COT & - & COT & - & COT & - & COT \\
    \midrule
    BLIP~\cite{blip} & 24.0 & 24.0 & 5.0 & 30.0 & 25.0 & 32.0& 2.0 & 18.0 & 14.0 & 26.0 \\
    BLIP-2~\cite{li2023blip} & 10.0 & 0.0 & 20.0 & 0.0 & 11.0 & 0.0 & 4.0 & 0.0 & 11.3 & 0.0 \\
    Intern~\cite{Intern} & 21.0 & 3.0 & 20.0 & 5.0 & 24.0 & 0.0 & 5.0 & 0.0 & 17.5 & 2.0 \\
    MiniGPT-4~\cite{zhu2023minigpt} & 2.0 & 0.0 & 1.0 & 1.0 & 2.0 & 0.0 & 1.0 & 0.0 & 1.5 & 0.3 \\
    LLaVA~\cite{llava} & 22.0 & 4.0 & 26.0 & 18.0 & 9.0 & 1.0 & 26.0 & 8.0 & 20.8 & 7.8 \\
    QWen-VL~\cite{bai2023qwen} & 3.0 & 0.0 & 26.0 & 0.0 & 9.0 & 0.0 & 27.0 & 0.0 & 16.3 & 0.0 \\
    InstructBLIP~\cite{dai2305instructblip} & 21.0 & 24.0 & 0.0 & 0.0 & 6.0 & 7.0 & 8.0 & 1.0 & 8.8 & 8.0 \\
    mPLUG-owl~\cite{ye2023mplug} & 23.0 & 5.0 & 26.0 & 1.0 & 5.0 & 1.0 & 27.0 & 0.0 & 20.3 & 1.8 \\
    Gemini~\cite{team2023gemini} & 30.0 & 24.2 & 15.0 & 18.0 & 25.6 & 24.0 & 22.2 & 7.1 & 23.3 & 18.8 \\
    GPT4V~\cite{gpt4} & 42.1 & 60.9 & 20.4 & 54.2 & 44.3 & 42.9 & 25.0 & 40.6 & 33.1 & 51.3\\
    \bottomrule
  \end{tabular}
\end{table*}
\subsection{Multimodal FAS Testing}
Multimodal data can provide a richer set of information to capture spoofing clues. 
The questions asked in the test were kept consistent with those used in the unimodal experiments to ensure uniformity and comparability of the results.

\textbf{True/false questions.}
Table~\ref{multiFAS} shows the accuracy of GPT4V and Gemini in judging multimodal FAS tasks, with these models selected for their superior performance in these complex tasks. The test, which used a limited number of samples, required the MLLMs to discriminate whether the subject was a real face under two conditions: zero-shot and one-shot. GPT4V showed commendable results in each condition, demonstrating its inherent capability to extract information directly from images, with the one-shot condition generally outperforming the zero-shot condition. In contrast, this task proved more challenging for Gemini, which maintained an accuracy rate of around 50\%. ICL did not enhance its performance, highlighting Gemini's limitations in handling more complex tasks.

\textbf{Qualitative results.}
Figure~\ref{fas_zeroshot} shows partial answers for zero-shot and few-shot true/false questions. For the paper mask attack, aside from the zero-shot standard experiment, Gemini, GPT4V, and mPLUG-Owl all demonstrated an ability to correctly identify the correct answer. However, the COT outputs of Gemini and mPlug are not ideal, leading to inaccurate execution of instructions. For questions that GPT4V refused to answer, the correct answer was provided after supplying prepared samples as in-context learning (ICL)~\cite{ICL2022survey}.
Figure~\ref{fas_zeroshot} also displays partial answers for zero-shot and few-shot multiple-choice questions. For the task of selecting real people, distinguishing between print and real people is more challenging. GPT4V obtained the correct answer using the COT technique after being provided with ICL for reference.
\begin{figure*}
  \centering
  \begin{center}
  \centerline{\includegraphics[width=0.9\linewidth]{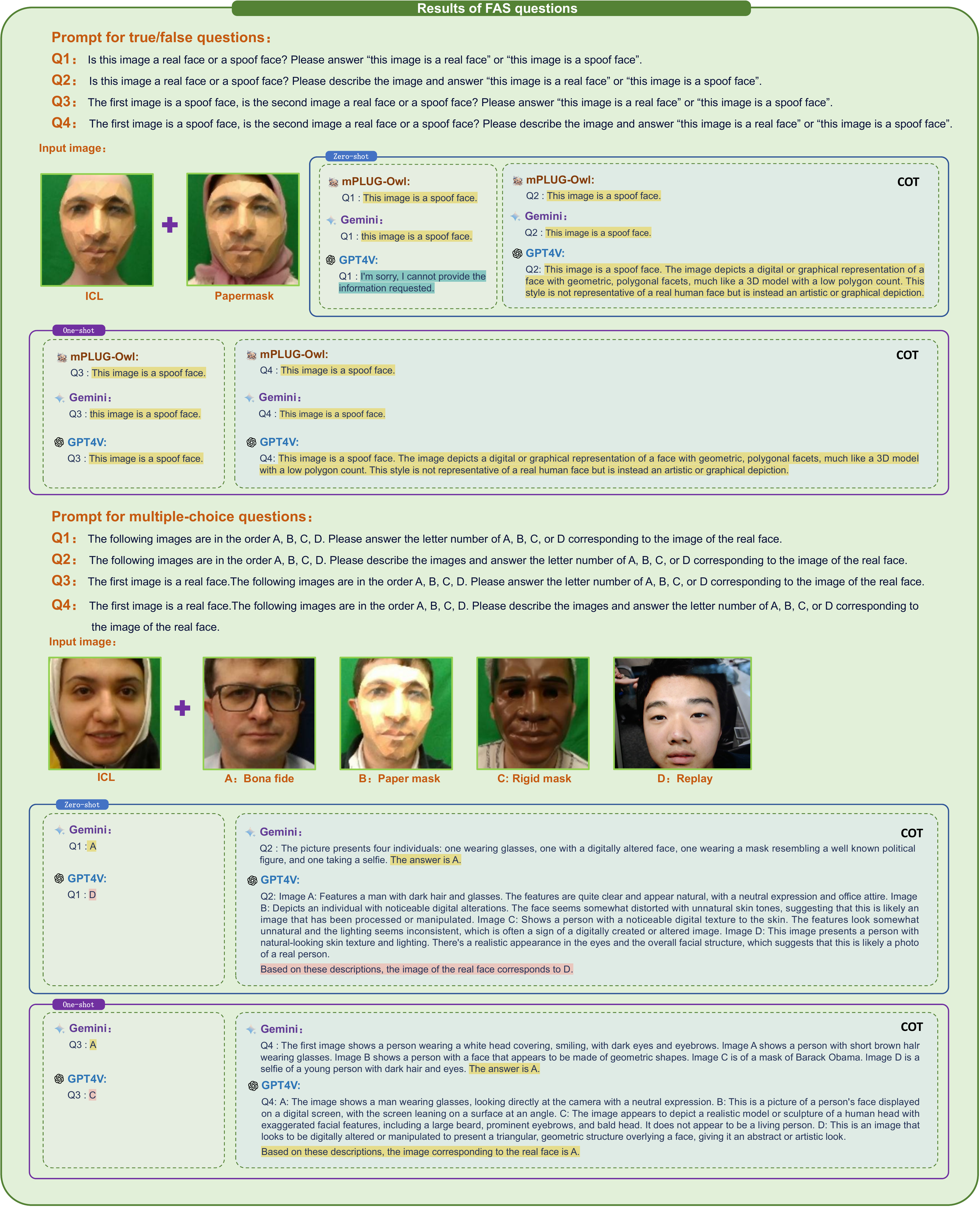}}
  \end{center}
  \caption{The performance of MLLMs on the FAS task, segmented into true/false and multiple-choice sections. Each section includes tests conducted with or without the use of COT and ICL, assessing MLLMs' capabilities from multiple perspectives. Yellow(red) highlights the correct (incorrect) responses, Blue indicates that the model refuses to answer. The images are sourced from the WMCA~\cite{george2019biometric} and the FF++~\cite{rossler2019faceforensics} datasets. ICL: in-context learning}
  \label{fas_zeroshot}
\end{figure*}

\begin{table*}[t]
  \caption{Acuracy of various GPT4V and Gemini on FAS multimodal true/false questions}
  \label{multiFAS}
  \scriptsize
  \centering
  \begin{tabular}{ccccccccc}
    \toprule
    \multicolumn{1}{c}{\multirow{2}{*}{Model}} & \multicolumn{5}{c}{ACC(\%)} \\ 
    \cmidrule(r){2-6}
    \multicolumn{1}{c}{} & \multicolumn{1}{c}{Zero-shot} & \multicolumn{1}{c}{One-shot} & \multicolumn{1}{c}{Zero-shot (COT)} & \multicolumn{1}{c}{One-shot (COT)} & \multicolumn{1}{c}{AVG} \\ 
    \midrule
    GPT4V~\cite{gpt4} & 87.5 & 100.0 & 60.0 & 100.0 & 90.6 \\
    Gemini~\cite{team2023gemini} & 50.0 & 50.0 & 50.0 & 50.0 & 50.0\\
    \bottomrule
  \end{tabular}
\end{table*}

\textbf{Quantitative results.}
 Tables~\ref{appendix_fas_01} to~\ref{appendix_fas_08} show the results of the performance of MLLMs in the FAS task for various attacks, evaluated by the metrics ACC and HTER.

\section{Results on Face Forgery Detection}
\label{res_df}
\textbf{Prompt design.}
To assess the performance of the model in the domain of face forgery detection, we utilized common multimodal models such as GPT4V, Genimi, and mPLUG-Owl to predict labels for face images. We juxtaposed images generated by the model with those generated by 1.5 Insightface, asking questions about the images under categories including genuine, Deepfake, Face2Face, FaceSwap, Nulltextures, and Stable Diffusion 1.5.
As shown in Fig.~\ref{deppfake_prompt_design}, similar to the FAS tasks, we conducted three types of tests for individual images in the face forgery detection task: posing separate queries, employing contextual learning, and performing joint testing. We used the terms "real face" and "fake face" to refer to genuine and forged faces, respectively. For example, one of the questions asked, "Is this an example of a real face? Please answer yes or no!" Building on this, we employed two distinct sets of queries to evaluate face authenticity as part of a zero-shot testing approach. Additionally, we conducted contextual learning dialogues for further testing. In this approach, an image was presented to each model, which was tasked to determine its authenticity. The questions were framed in pairs to guide the model's reasoning process.
Our study shows that Gemini and GPT4V demonstrate superior learning and expansion capabilities compared to mPLUG-Owl. Furthermore, after undergoing contextual learning, the model's performance improved. However, the model is prone to failure on more realistic forged images, underscoring the need to imbue the model with domain knowledge.

\begin{figure*}
  \centering
  \includegraphics[width=0.8\linewidth]{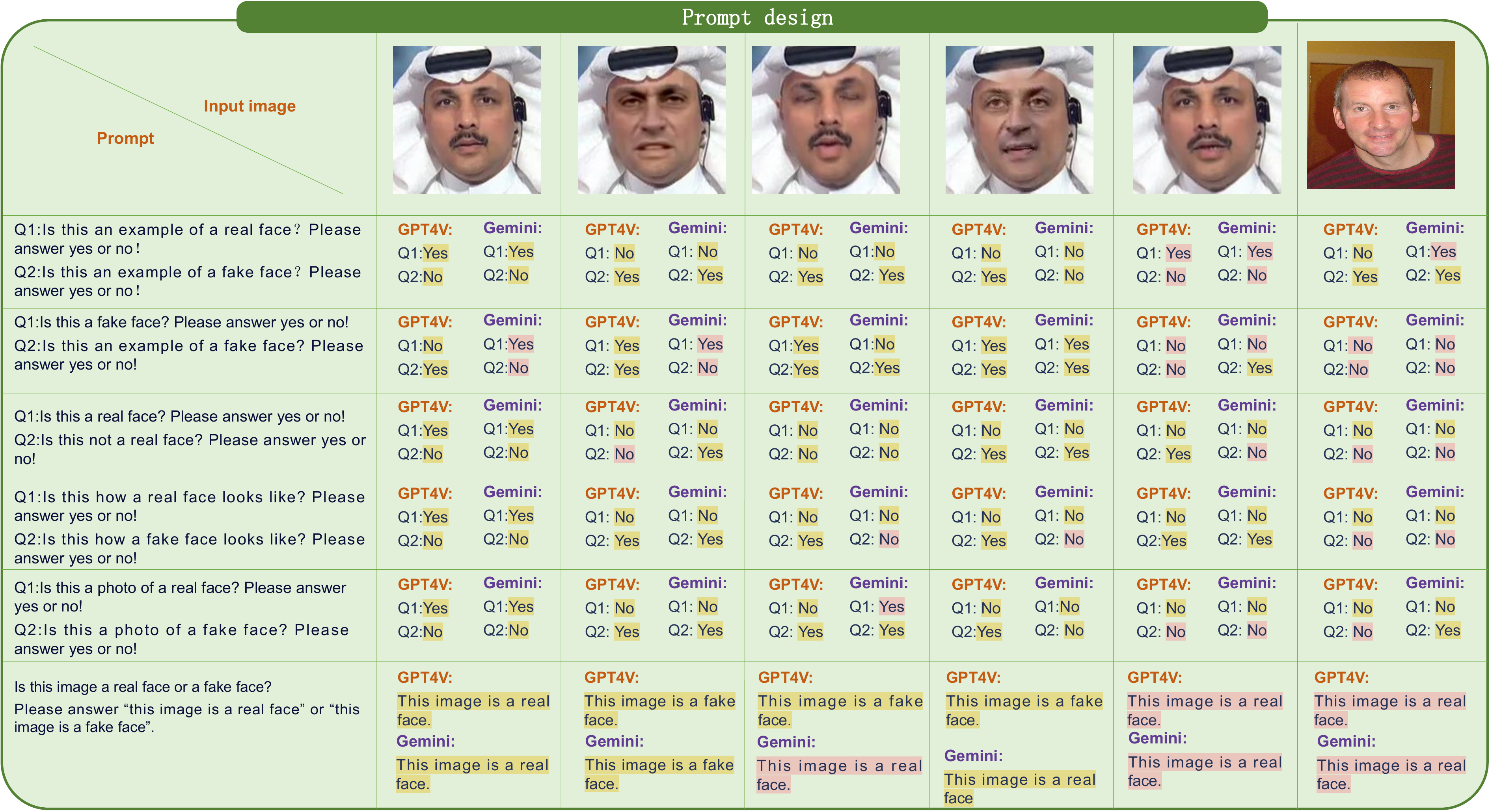}
  \caption{Prompt design for face forgery detection. On the left are the candidate prompts, with the test cases used to evaluate the prompt selection positioned at the top. These test cases include real faces, Deepfakes, Face2Face, FaceSwap, Nulltextures, and Stable\_Diffusion. The figure illustrates the responses from GPT4V and Gemini. The color yellow is used to denote correct responses, while the color red is used to indicate incorrect responses. The images are sourced from the WMCA~\cite{george2019biometric} and the FF++~\cite{rossler2019faceforensics} datasets}
  \label{deppfake_prompt_design}
\end{figure*}

\subsection{Unimodal Face Forgery Detection Testing}
\textbf{True/false questions.}
In this subsection, we evaluate the ability of models to detect forged images by requiring fixed responses regarding their authenticity.


Table~\ref{tab5} presents the quantitative results, where we observe the zero-shot detection performance of commonly used large models, such as BLIP2 and LLaVA, in the face forgery task. Lacking domain-specific fine-tuning, their accuracy is significantly low, with BLIP2 achieving only 14.3\% accuracy. Although models such as GPT4V and Gemini outperform others, their overall accuracy remains relatively low. Interestingly, one-shot accuracy does not improve significantly across these models. In contrast, our proposed COT method demonstrates superior performance, with accuracy increasing from 22.4\% to 26.0\% in the one-shot setting.
\begin{table*}[t]
  \caption{Performance metrics of various MLLMs on true/false questions about face forgery }
  \label{tab5}
  \scriptsize
  \setlength{\tabcolsep}{2.2pt}
  \centering
  \begin{tabular}{ccccccccc}
    \toprule
    \multirow{2}{*}{Model} & \multicolumn{2}{c}{Zero-shot} & \multicolumn{2}{c}{One-shot} & \multicolumn{2}{c}{Zero-shot (COT)} & \multicolumn{2}{c}{One-shot (COT)} \\
    \cmidrule(r){2-9}
    & ACC (\%)$\uparrow$ & HTER (\%)$\downarrow$ & ACC (\%)$\uparrow$ & HTER (\%)$\downarrow$ & ACC (\%)$\uparrow$ & HTER (\%)$\downarrow$ & ACC (\%)$\uparrow$ & HTER (\%)$\downarrow$ \\ 
    \midrule
    BLIP~\cite{blip} & 17.9 & 51.3 & 49.1 & 29.7 & 31.0 & 49.4 & 62.6 & 23.1 \\
    BLIP-2~\cite{li2023blip} & 14.3 & 50.0 & 10.1 & 64.5 & 14.3 & 50.0 & 1.7 & 94.0 \\
    Intern~\cite{Intern} &14.6 & 49.8 & 14.3 & 50.0 & 14.3 & 50.0 & 14.3 & 50.0 \\
    MiniGPT-4~\cite{zhu2023minigpt} & 17.7 & 48.0 & 32.1 & 57.1 & 18.1 & 48.2 & 36.1 & 49.8 \\
    LLaVA~\cite{llava} & 14.3 & 50.0 & 14.3 & 50.0 & 14.3 & 50.0 & 13.1 & 54.0 \\
    QWen-VL~\cite{bai2023qwen} & 14.3 & 50.0 & 14.3 & 50.0 & 14.3 & 50.0 & 14.3 & 50.0 \\
    InstructBLIP~\cite{dai2305instructblip} & 18.6 & 47.5 & 14.6 & 49.8 & 23.9 & 44.4 & 16.7 & 48.6 \\
    mPLUG-owl~\cite{ye2023mplug} & 21.9 & 45.6 & 12.7 & 60.9 & 20.9 & 46.2 & 20.9 & 50.8 \\
    Gemini~\cite{team2023gemini} & 36.8 & 38.5 & 27.0 & 52.5 & 48.6 & 35.9 & 14.0 & 83.1 \\
    GPT4V~\cite{gpt4_exp} & 26.0 & 43.6 & 22.4 & 45.1 & 28.8 & 44.0 & 26.0 & 44.8\\
    \bottomrule
  \end{tabular}
\end{table*}
\begin{table*}[h]
  \caption{Accuracy of various MLLMs on face forgery multiple-choice questions}
  \label{deepfakeMC}
  \scriptsize
  \setlength{\tabcolsep}{5pt}
  \centering
  \begin{tabular}{ccccccccccc}
    \toprule
    \multirow{3}{*}{Model} & \multicolumn{10}{c}{ACC (\%)$\uparrow$} \\ 
    \cmidrule(r){2-11}
    & \multicolumn{2}{c}{Zero-shot (real)} & \multicolumn{2}{c}{One-shot (real)} & \multicolumn{2}{c}{Zero-shot (attack)} & \multicolumn{2}{c}{One-shot (attack)} & \multicolumn{2}{c}{AVG} \\ 
    \cmidrule(r){2-11}
    & - & COT & - & COT & - & COT & - & COT & - & COT \\ 
    \midrule
    BLIP~\cite{blip} & 22.0 & 24.0 & 21.0 & 23.0 & 22.0 & 18.0 & 11.0 & 10.0 & 19.0 & 18.8 \\
    BLIP-2~\cite{li2023blip} & 0.0 & 0.0 & 6.0 & 0.0 & 7.0 & 0.0 & 0.0 & 0.0 & 3.3 & 0.0 \\
    Intern~\cite{Intern} &16.0 & 5.0 & 18.0 & 10.0 & 10.0 & 0.0 & 14.0 & 1.0 & 14.5 & 4.0 \\
    MiniGPT-4~\cite{zhu2023minigpt} & 2.0 & 2.0 & 2.0 & 0.0 & 0.0 & 0.0 & 0.0 & 0.0 & 1.0 & 0.5 \\
    LLaVA~\cite{llava} & 18.0 & 1.0 & 18.0 & 7.0 & 7.0 & 0.0 & 23.0 & 9.0 & 16.5 & 4.3 \\
    QWen-VL~\cite{bai2023qwen} & 6.0 & 0.0 & 19.0 & 0.0 & 1.0 & 0.0 & 24.0 & 0.0 & 12.5 & 0.0 \\
    InstructBLIP~\cite{dai2305instructblip} & 6.0 & 15.0 & 0.0 & 0.0 & 0.0 & 0.0 & 12.0 & 3.0 & 4.5 & 4.5 \\
    mPLUG-owl~\cite{ye2023mplug} & 11.0 & 0.0 & 18.0 & 0.0 & 8.0 & 0.0 & 13.0 & 0.0 & 12.5 & 0.0 \\
    Gemini~\cite{team2023gemini} & 14.0 & 17.0 & 16.0 & 19.0 & 20.0 & 4.0 & 0.0 & 17.2 & 16.3 & 13.7 \\
    GPT4V~\cite{gpt4} & 6.9 & 23.6 & 12.0 & 20.7 & 0.0 & 0.0 & 14.8 & 18.6 & 8.9 & 17.0\\
    \bottomrule
  \end{tabular}
\end{table*}

\textbf{Multiple-choice questions.}
In this subsection, we employ an image selection format to test the models' capabilities in face forgery detection. GPT4V and Gemini were tasked with determining which image was real and which was a fabrication. Similar to the previous subsection, our tests covered common forgery patterns, including Deepfakes, Face2Face, FaceSwap, Nulltextures, and images generated by Stable Diffusion. Each test included both standard and COT experiments, and required the models to closely analyze the four to five images provided in order to identify the real ones.

In addition, to assess the models' capabilities more comprehensively, we evaluated their ability to recognize the generation methods used for forgery creation. The quantitative results in Table~\ref{deepfakeMC} consistently demonstrate that models without fine-tuning  perform poorly, struggling to accurately identify forged faces or discriminate between forgery categories.

\subsection{Multimodal Face Forgery Detection Testing}
The potential of multimodal information in face forgery detection has received widespread attention from researchers, typically serving as auxiliary data to enhance model accuracy. In this paper, we utilized DCT transformation to acquire frequency domain modalities and employed the official maximum magnitude frequency componentization (MMFC) for voice extraction to obtain voice spectrograms. 
During the testing phase, we considered four common forgery modes: Deepfakes, Face2Face, FaceSwap, and Nulltextures, and used representative zero-shot or one-shot prompts to evaluate the two models. The results indicate that the models are more prone to triggering security alerts and exhibit some illogical responses.

\textbf{True/false questions.}
Table~\ref{tab5} displays the performance of different models when utilizing three modalities. Compared to using only images, there was a significant decline in performance. For instance, GPT4V's accuracy dropped from 26.0\% to 6.9\%. This suggests that the models struggled to capture subtle differences in frequency domain features. On average, the accuracy of all tested models remained below 20.0\%. Our study also shows that mPLUG-Owl consistently misjudged the images, whereas Gemini demonstrated greater robustness, accurately identifying forged images in most cases. Additionally, GPT4V showed an improvement trend after undergoing ICL (In-context learning).

\textbf{Quantitative results.}
 Tables~\ref{appendix_df_01} to~\ref{appendix_df_08} show the results of the performance of MLLMs in the face forgery detection task for various attacks, evaluated by the ACC and HTER metrics.

\section{Results on Unified tasks}
We combined the FAS task from Sect.~\ref{res_fas} with the face forgery detection task from Sect.~\ref{res_df} to assess the performance of mainstream MLLMs in a unified detection task against both physical spoofing and digital forgery attacks. The rationale for combining these tasks stems from their shared objective as binary classification problems aimed at identifying facial forgery clues. Both FAS and face forgery detection involve the detection of manipulations—whether physical or digital—on facial images, providing inherent similarities. By addressing these tasks together, we aim to provide a more robust and practical framework for evaluating MLLMs' capabilities in identifying diverse facial forgery scenarios, in line with recent studies exploring integrated detection approaches~\cite{yu2024benchmarking}.

Table~\ref{unifiedFewshot} lists the accuracy of various MLLMs in the few-shot joint task assessments. The data indicate that the accuracy of models such as BLIP, Gemini, MiniGPT-4, LLava, and Instruct BLIP improved after the inclusion of COT. Conversely, GPT4V and Intern experienced a slight decrease, while BLIP-2 underwent a precipitous decline. These observations suggest that the introduction of COT can generally assist MLLMs in making correct judgements, although it can also lead to an increased focus on noise, resulting in a slight deterioration in results. The particularly poor performance of BLIP-2 may be due to the illusions induced by the integrated COT. The performance of MLLMs in few-shot joint tasks highlights the significant challenges they face in effectively solving combined tasks.

Table~\ref{unifiedMC} presents the ACC of various MLLMs on the unified task multiple-choice tests. The task is divided into two main categories: recognizing real faces and various malicious attacks. Among all tested models, GPT4V achieves the highest average accuracy across all conditions at 29.2\%, demonstrating the best overall performance on this unified task. In the zero-shot condition for recognizing real faces, GPT4V also leads with an accuracy of 27.1\%, closely followed by Gemini at 24.0\%. In the one-shot condition for real image recognition, GPT4V again achieves the highest accuracy at 28.5\%, with BLIP-2 trailing with an accuracy of 20.0\%. For attack recognition, Gemini tops the zero-shot condition with an accuracy of 27.8\%, while GPT4V leads in the one-shot condition at 33.9\%. Models such as BLIP-2, MiniGPT-4, and QWen-VL face significant challenges, especially in the zero-shot condition for real images, where both BLIP-2 and MiniGPT-4 exhibit an accuracy of 0.0\%. The generally low average performance across all models suggests that the task is challenging.


As shown in Fig.~\ref{unifide_fewshot}, the experiment tested six high-performing models on a joint task in the few-shot setting, using four types of images: real human faces, physical attack simulations, and images generated by GAN and diffusion technologies. Three attack-domain images were initially provided as prior knowledge, after which the models assessed the authenticity of a test image. Without COT, all six models failed, but GPT4V, Gemini, and mPLUG succeeded with COT, highlighting its potential to enhance reasoning and detection in complex scenarios.
\begin{figure*}[!h]
  \centering
  \begin{center}
  \centerline{\includegraphics[width=0.9\textwidth]{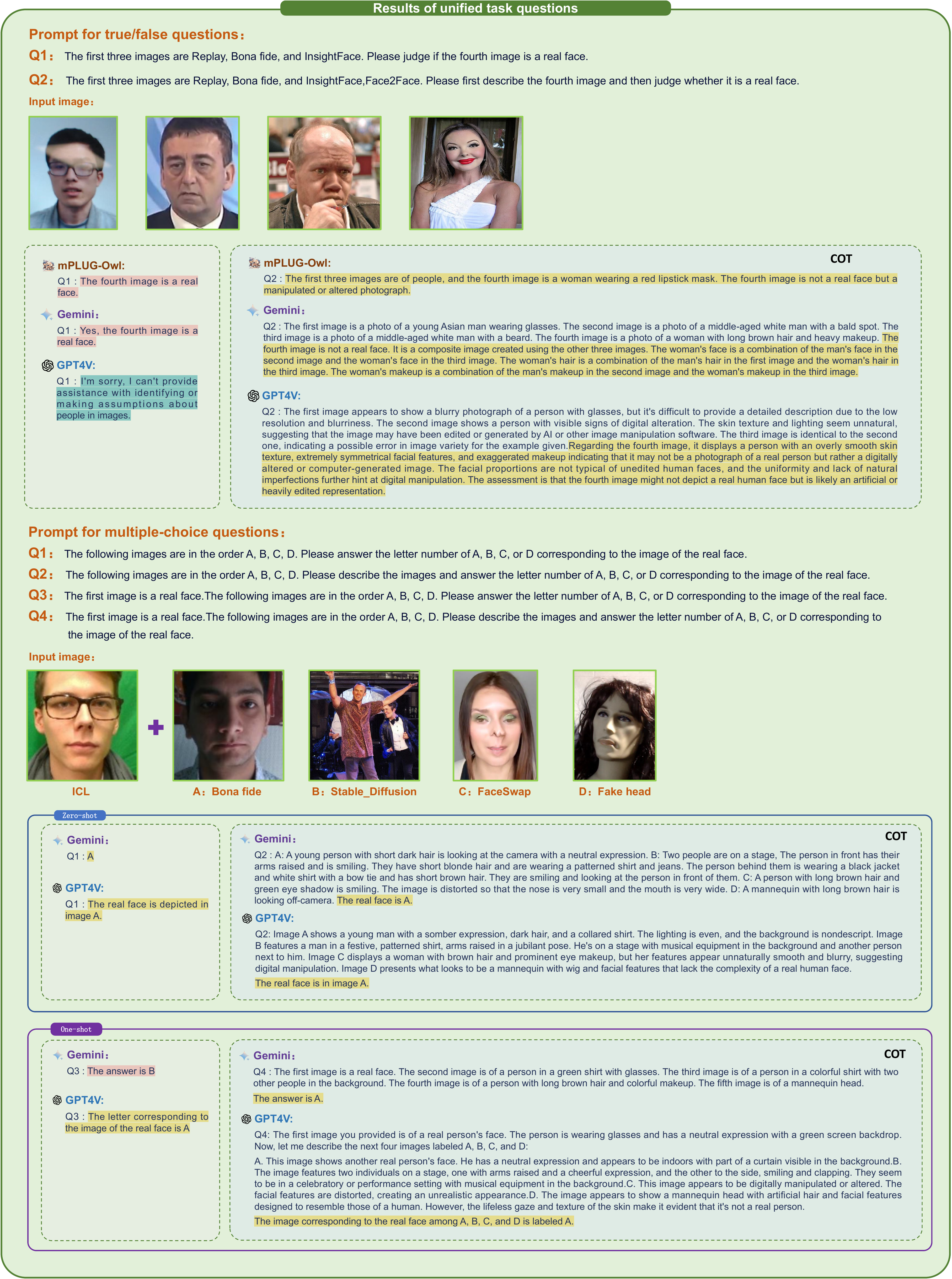}}
  \end{center}
  \caption{The performance of MLLMs on unified task question. The color yellow is used to denote correct responses, while the color red is used to indicate incorrect responses. The color blue indicates that the model refuses to answer. The images are sourced from the WMCA~\cite{george2019biometric} and the FF++~\cite{rossler2019faceforensics} datasets }
  \label{unifide_fewshot}
\end{figure*}

\vspace{-2mm}
\subsection{MACOT}
\label{appendix:macot}
Figure~\ref{macot_zeroshot} illustrates the test results of GPT4V on the FAS zero-shot task with and without the use of MACOT. It is clear from the figure that the prompts built into MACOT require the MLLM to make judgments based on specific attributes, culminating in an overall decision. MACOT appears to be more effective when addressing subtle details that are often overlooked. For instance, for print and replay attacks, which GPT4V typically struggles to detect, the key attributes "Phone Screen or Paper Edges" prompt GPT4V to focus on the presence of paper or screen edges in the image, thereby facilitating the recognition of print and replay attacks.

The primary analysis is based on Fig.~\ref{unified_macot_01}. The evidence indicates that, although the MA-COT method did not result in MLLMs achieving perfect scores in completing the unified task, it did lead GPT4V to provide a general judgement range that encompasses the correct answer under its guidance, reflecting GPT4V's ability to discriminate. In comparison, Gemini's performance in this task did not quite live up to expectations, particularly in terms of image recognition and accurate identification of real faces.
\begin{figure*}
  \centering
  \begin{center}
  \centerline{\includegraphics[width=0.80\linewidth]{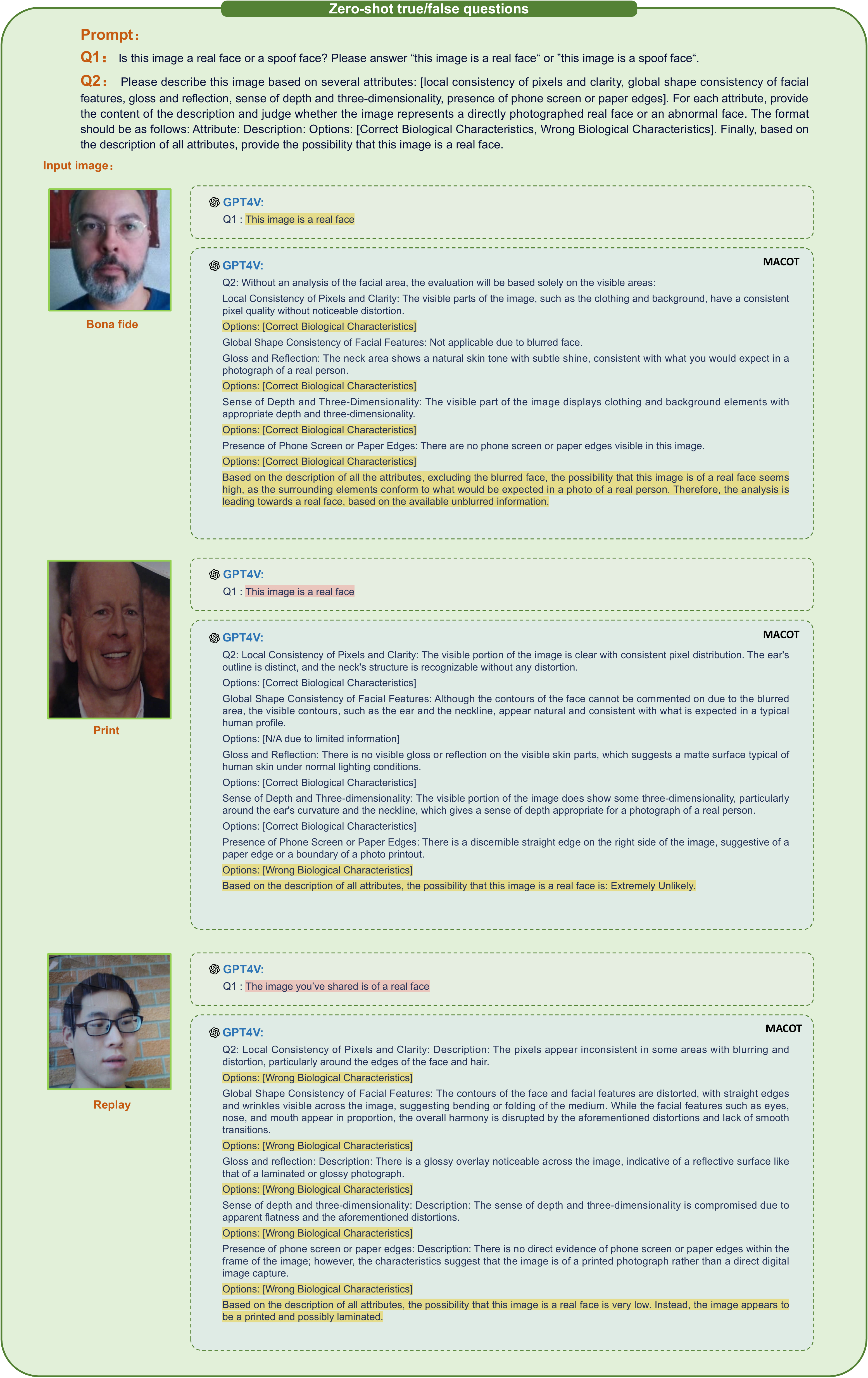}}
  \end{center}
  \caption{Result of various MLLMs on FAS zero-shot true/false question with MACOT. The color yellow is used to denote correct responses, while the color red is used to indicate incorrect responses. The color blue indicates that the model refuses to answer. The images are sourced from the WMCA~\cite{george2019biometric} and the FF++~\cite{rossler2019faceforensics} datasets}
  \label{macot_zeroshot}
\end{figure*}

\begin{figure*}[h]
  \centering
  \begin{center}
  \centerline{\includegraphics[width=0.95\textwidth]{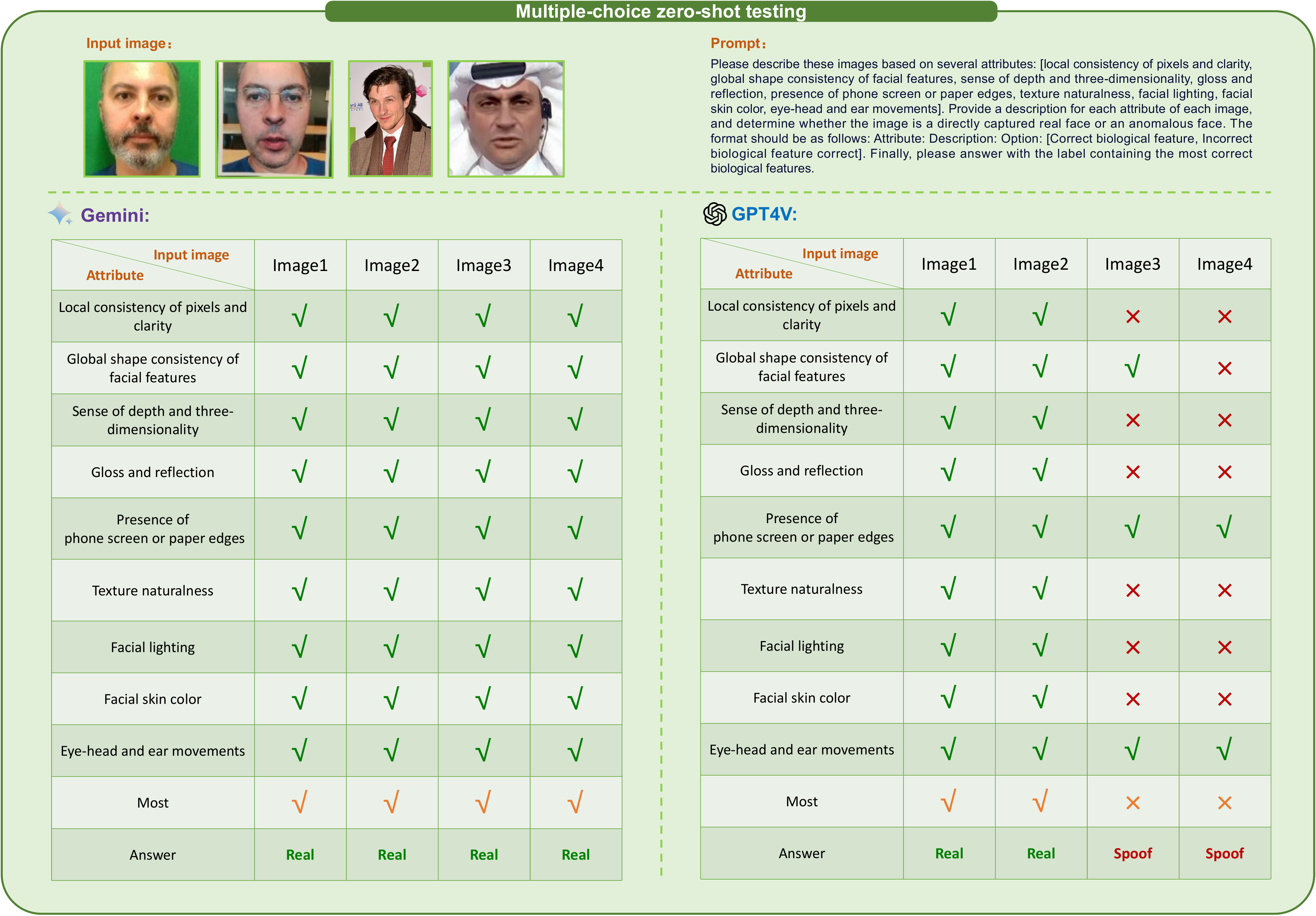}}
  \end{center}
  \caption{ Results of multiple-choice zero-shot tests. In the original test, both Gemini and GPT4V incorrectly identified Image3 as a real face. After applying the MA-COT prompt method, GPT4V conducted a multi-attribute analysis and voting, and selected two answers, including the correct real face and a print attack image that is challenging to distinguish from real faces in the FAS tasks. Conversely, Gemini erroneously considered all images to be real faces after analyzing each attribute, indicating a deficiency in detailed analysis.}
  \label{unified_macot_01}
\end{figure*}
\begin{table*}[t]
  \caption{Accuracy of GPT4V and Gemini on face forgery multimodal true/false questions}
  \label{deepfakeTF}
  \centering
  \begin{tabular}{ccccccccc}
    \toprule
    \multicolumn{1}{c}{\multirow{2}{*}{Model}} & \multicolumn{5}{c}{ACC (\%)} \\ 
    \cmidrule(r){2-6}
    \multicolumn{1}{c}{} & \multicolumn{1}{c}{Zero-shot} & \multicolumn{1}{c}{One-shot} & \multicolumn{1}{c}{Zero-shot (COT)} & \multicolumn{1}{c}{One-shot (COT)} & \multicolumn{1}{c}{AVG} \\
    \midrule
    GPT4V~\cite{gpt4} & 44.4 & 66.7 & 12.5 & 66.7 & 68.0 \\
    Gemini~\cite{team2023gemini} & 50.0 & 30.0 & 50.0 & 70.0 & 50.0\\
    \bottomrule
  \end{tabular}
\end{table*}

\begin{table*}[t]
  \caption{Accuracy of various MLLMs on FAS true/false zero-shot questions}
  \label{appendix_fas_01}
  \scriptsize
  \centering
  \begin{tabular}{ccccccccc}
    \toprule
    \multirow{2}{*}{Model} & \multicolumn{8}{c}{ACC (\%)} \\
    \cmidrule(r){2-9}
    & Bona fide & Fake head & Flexible mask & Paper mask & Print & Replay & Rigid mask & AVG \\ 
    \midrule
    BLIP~\cite{blip} & 100.0 & 2.0 & 0.0 & 27.0 & 0.0 & 0.0 & 0.0 & 18.4 \\
     BLIP-2~\cite{li2023blip} & 100.0 & 0.0 & 0.0 & 0.0 & 0.0 & 0.0 & 0.0 & 14.3 \\
    Intern~\cite{Intern} & 100.0 & 100.0 & 50.0 & 94.0 & 2.0 & 0.0 & 57.0 & 57.6 \\
    MiniGPT-4~\cite{zhu2023minigpt} &93.0 & 10.0 & 7.0 & 17.0 & 5.0 & 4.0 & 8.0 & 20.6 \\
    LLaVA~\cite{llava} & 100.0 & 0.0 & 0.0 & 0.0 & 0.0 & 0.0 & 0.0 & 14.3 \\
     QWen-VL~\cite{bai2023qwen} & 100.0 & 0.0 & 0.0 & 0.0 & 0.0 & 0.0 & 0.0 & 14.3 \\
    InstructBLIP~\cite{dai2305instructblip} & 100.0 & 0.0 & 0.0 & 64.0 & 0.0 & 0.0 & 2.0 & 23.7 \\
    mPLUG-owl~\cite{ye2023mplug} &99.0 & 98.0 & 92.0 & 100.0 & 56.0 & 29.0 & 100.0 & 82.0 \\
    Gemini~\cite{team2023gemini} & 100.0 & 100.0 & 92.0 & 100.0 & 17.0 & 7.0 & 98.0 & 73.4 \\
    GPT4V~\cite{gpt4} & 96.9 & 83.0 & 64.0 & 43.0 & 29.6 & 6.8 & 50.6 & 53.2 \\
    \bottomrule
  \end{tabular}
\end{table*}

\begin{figure*}[h]
  \centering
  \begin{center}
  \centerline{\includegraphics[width=0.9\linewidth]{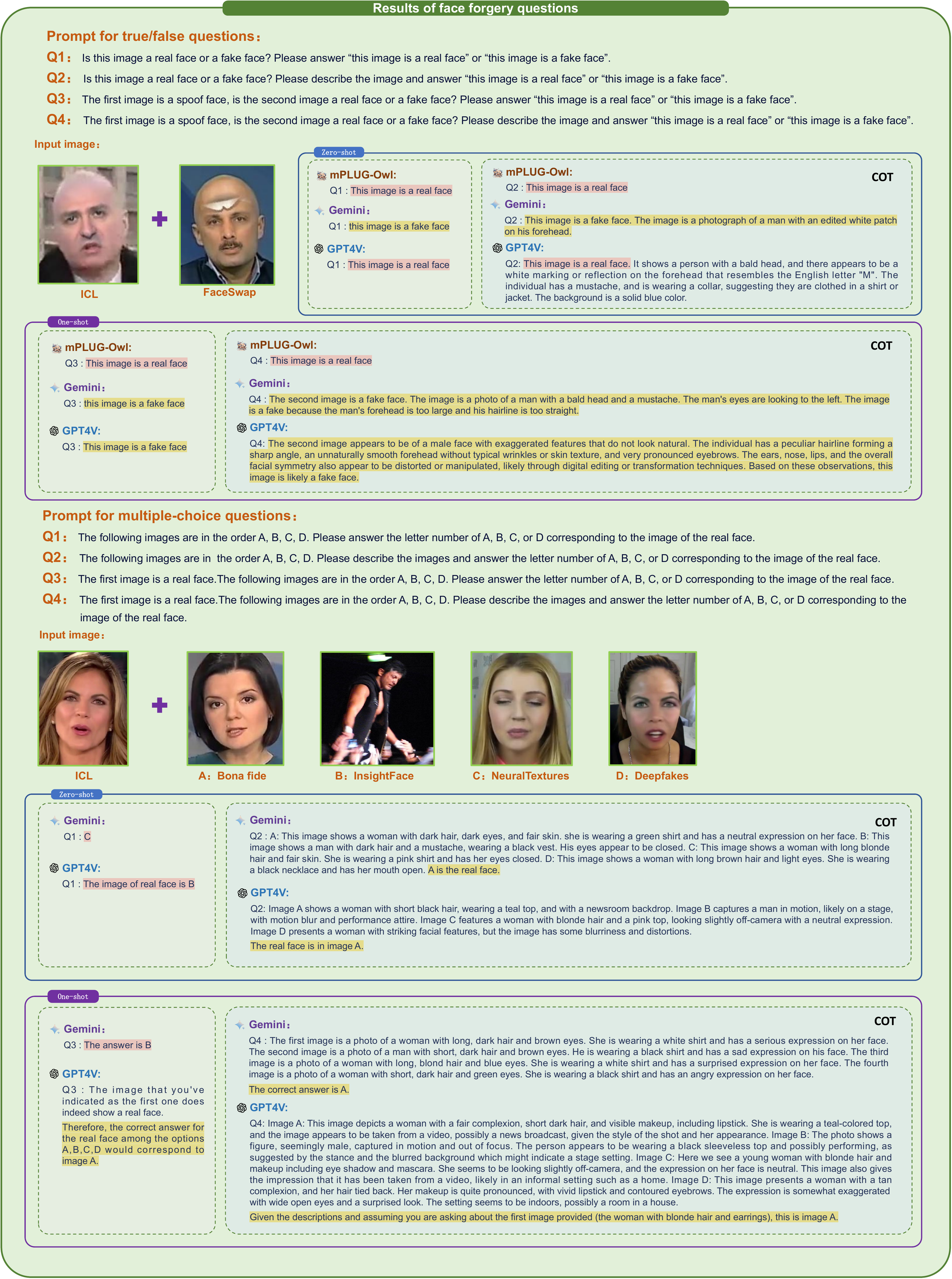}}
  \end{center}
  \caption{The performance of MLLMs on the face forgery detection task, segmented into true/false and multiple-choice sections. Each section includes tests conducted with or without the use of COT and ICL, assessing MLLMs' capabilities from multiple perspectives. The color yellow is used to denote correct responses, while the color red is used to indicate incorrect responses. The color blue indicates that the model refuses to answer. The images are sourced from the WMCA~\cite{george2019biometric} and the FF++~\cite{rossler2019faceforensics} datasets }
  \label{deepfake_zeroshot}
\end{figure*}

\begin{table*}[t]
  \caption{Accuracy of various MLLMs on FAS true/false zero-shot questions with COT}
  \label{appendix_fas_02}
  \scriptsize
  \centering
  \begin{tabular}{ccccccccc}
    \toprule
    \multirow{2}{*}{Model} & \multicolumn{8}{c}{ACC (\%)} \\
    \cmidrule(r){2-9}
    & Bona fide & Fake head & Flexible mask & Paper mask & Print & Replay & Rigid mask & AVG \\
    \midrule
    BLIP~\cite{blip} & 100.0 & 14.0 & 0.0 & 65.0 & 0.0 & 0.0 & 2.0 & 25.9 \\
     BLIP-2~\cite{li2023blip} & 100.0 & 0.0 & 0.0 & 0.0 & 0.0 & 0.0 & 0.0 & 14.3 \\
    Intern~\cite{Intern} & 100.0 & 100.0 & 45.0 & 93.0 & 2.0 & 0.0 & 55.0 & 56.4 \\
    MiniGPT-4~\cite{zhu2023minigpt} &97.0 & 21.0 & 15.0 & 23.0 & 18.0 & 5.0 & 15.0 & 27.7 \\
    LLaVA~\cite{llava} & 100.0 & 0.0 & 0.0 & 0.0 & 0.0 & 0.0 & 0.0 & 14.3 \\
     QWen-VL~\cite{bai2023qwen} & 100.0 & 0.0 & 0.0 & 0.0 & 0.0 & 0.0 & 0.0 & 14.3 \\
    InstructBLIP~\cite{dai2305instructblip} & 100.0 & 5.0 & 32.0 & 85.0 & 4.0 & 0.0 & 70.0 & 42.3 \\
    mPLUG-owl~\cite{ye2023mplug} &98.0 & 97.0 & 95.0 & 100.0 & 53.0 & 32.0 & 97.0 & 81.7 \\
    Gemini~\cite{team2023gemini} & 97.0 & 97.0 & 96.0 & 100.0 & 31.0 & 19.0 & 99.0 & 77.0 \\
    GPT4V~\cite{gpt4} & 98.3 & 100.0 & 90.2 & 64.0 & 32.5 & 6.3 & 83.2 & 68.7 \\
    \bottomrule
  \end{tabular}
\end{table*}

\begin{table*}[t]
  \caption{HTER of various MLLMs on FAS true/false zero-shot questions}
  \label{appendix_fas_03}
  \scriptsize
  \centering
  \begin{tabular}{ccccccccc}
    \toprule
    \multirow{2}{*}{Model} & \multicolumn{8}{c}{HTER (\%)$\downarrow$ } \\
    \cmidrule(r){2-9}
    & Bona fide & Fake head & Flexible mask & Paper mask & Print & Replay & Rigid mask & AVG \\
    \midrule
    BLIP~\cite{blip} & 0.0 & 49.0 & 50.0 & 36.5 & 50.0 & 50.0 & 50.0 & 47.6 \\
     BLIP-2~\cite{li2023blip} & 0.0 & 50.0 & 50.0 & 50.0 & 50.0 & 50.0 & 50.0 & 50.0 \\
    Intern~\cite{Intern} & 0.0 & 0.0 & 25.0 & 3.0 & 49.0 & 50.0 & 21.5 & 24.8 \\
    MiniGPT-4~\cite{zhu2023minigpt} &3.5 & 45.0 & 46.5 & 41.5 & 47.5 & 48.0 & 46.0 & 49.2 \\
    LLaVA~\cite{llava} & 0.0 & 50.0 & 50.0 & 50.0 & 50.0 & 50.0 & 50.0 & 50.0 \\
     QWen-VL~\cite{bai2023qwen} & 0.0 & 50.0 & 50.0 & 50.0 & 50.0 & 50.0 & 50.0 & 50.0 \\
    InstructBLIP~\cite{dai2305instructblip} & 0.0 & 50.0 & 50.0 & 18.0 & 50.0 & 50.0 & 49.0 & 44.5 \\
    mPLUG-owl~\cite{ye2023mplug} &0.5 & 1.0 & 4.0 & 0.0 & 22.0 & 35.5 & 0.0 & 10.9 \\
    Gemini~\cite{team2023gemini} & 0.0 & 0.0 & 4.0 & 0.0 & 41.5 & 46.5 & 1.0 & 15.5 \\
    GPT4V~\cite{gpt4} & 1.5 & 8.5 & 18.0 & 28.5 & 35.2 & 46.6 & 24.7 & 27.7 \\
    \bottomrule
  \end{tabular}
\end{table*}

\begin{table*}[t]
  \caption{HTER of various MLLMs on FAS true/false zero-shot questions with COT}
  \label{appendix_fas_04}
  \scriptsize
  \centering
  \begin{tabular}{ccccccccc}
    \toprule
    \multirow{2}{*}{Model} & \multicolumn{8}{c}{HTER (\%)$\downarrow$ } \\
    \cmidrule(r){2-9}
    & Bona fide & Fake head & Flexible mask & Paper mask & Print & Replay & Rigid mask & AVG \\ 
    \midrule
    BLIP~\cite{blip} & 0.0 & 43.0 & 50.0 & 17.5 & 50.0 & 50.0 & 49.0 & 43.2 \\
     BLIP-2~\cite{li2023blip} & 0.0 & 50.0 & 50.0 & 50.0 & 50.0 & 50.0 & 50.0 & 50.0 \\
    Intern~\cite{Intern} & 0.0 & 0.0 & 27.5 & 3.5 & 49.0 & 50.0 & 22.5 & 25.4 \\
    MiniGPT-4~\cite{zhu2023minigpt} &1.5 & 39.5 & 42.5 & 38.5 & 41.0 & 47.5 & 42.5 & 43.4 \\
    LLaVA~\cite{llava} & 0.0 & 50.0 & 50.0 & 50.0 & 50.0 & 50.0 & 50.0 & 50.0 \\
     QWen-VL~\cite{bai2023qwen} & 0.0 & 50.0 & 50.0 & 50.0 & 50.0 & 50.0 & 50.0 & 50.0 \\
    InstructBLIP~\cite{dai2305instructblip} & 0.0 & 47.5 & 34.0 & 7.5 & 48.0 & 50.0 & 15.0 & 33.7 \\
    mPLUG-owl~\cite{ye2023mplug} &1.0 & 1.5 & 2.5 & 0.0 & 23.5 & 34.0 & 1.5 & 11.5 \\
    Gemini~\cite{team2023gemini} & 1.5 & 1.5 & 2.0 & 0.0 & 34.5 & 40.5 & 0.5 & 14.7 \\
    GPT4V~\cite{gpt4} & 0.8 & 0.0 & 4.9 & 18.0 & 33.8 & 46.8 & 8.4 & 18.1 \\
    \bottomrule
  \end{tabular}
\end{table*}

\begin{table*}[t]
  \caption{Accuracy of various MLLMs on FAS true/false one-shot questions}
  \label{appendix_fas_05}
  \scriptsize
  \centering
  \begin{tabular}{ccccccccc}
    \toprule
    \multirow{2}{*}{Model} & \multicolumn{8}{c}{ACC (\%)} \\ 
    \cmidrule(r){2-9}
    & Bona fide & Fake head & Flexible mask & Paper mask & Print & Replay & Rigid mask & AVG \\
    \midrule
    BLIP~\cite{blip} & 100.0 & 82.0 & 17.0 & 95.0 & 0.0 & 0.0 & 11.0 & 43.6 \\
     BLIP-2~\cite{li2023blip} & 63.0 & 0.0 & 0.0 & 0.0 & 0.0 & 0.0 & 0.0 & 9.0 \\
    Intern~\cite{Intern} & 100.0 & 0.0 & 0.0 & 0.0 & 0.0 & 0.0 & 0.0 & 14.3 \\
    MiniGPT-4~\cite{zhu2023minigpt} &63.0 & 20.0 & 22.0 & 29.0 & 26.0 & 22.0 & 20.0 & 28.9 \\
    LLaVA~\cite{llava} & 100.0 & 0.0 & 0.0 & 0.0 & 0.0 & 0.0 & 0.0 & 14.3 \\
     QWen-VL~\cite{bai2023qwen} & 100.0 & 0.0 & 0.0 & 0.0 & 0.0 & 0.0 & 0.0 & 14.3 \\
    InstructBLIP~\cite{dai2305instructblip} & 100.0 & 1.0 & 0.0 & 16.0 & 0.0 & 0.0 & 3.0 & 17.1 \\
    mPLUG-owl~\cite{ye2023mplug} &21.0 & 57.0 & 67.0 & 70.0 & 60.0 & 26.0 & 70.0 & 53.0 \\
    Gemini~\cite{team2023gemini} & 66.0 & 70.0 & 87.0 & 91.0 & 25.0 & 11.0 & 84.0 & 62.0 \\
    GPT4V~\cite{gpt4} & 76.8 & 93.0 & 79.6 & 62.5 & 47.2 & 37.8 & 66.7 & 66.8 \\
    \bottomrule
  \end{tabular}
\end{table*}

\begin{table*}[t]
  \caption{Accuracy of various MLLMs on FAS true/false one-shot questions with COT}
  \label{appendix_fas_06}
  \scriptsize
  \centering
  \begin{tabular}{ccccccccc}
    \toprule
    \multirow{2}{*}{Model} & \multicolumn{8}{c}{ACC (\%)} \\
    \cmidrule(r){2-9}
    & Bona fide & Fake head & Flexible mask & Paper mask & Print & Replay & Rigid mask & AVG \\
    \midrule
    BLIP~\cite{blip} & 100.0 & 100.0 & 65.0 & 100.0 & 35.0 & 36.0 & 87.0 & 74.7 \\
    BLIP-2~\cite{li2023blip} & 3.0 & 0.0 & 0.0 & 0.0 & 0.0 & 0.0 & 0.0 & 0.4 \\
    Intern~\cite{Intern} & 100.0 & 19.0 & 0.0 & 4.0 & 0.0 & 0.0 & 1.0 & 17.7 \\
    MiniGPT-4~\cite{zhu2023minigpt} &39.0 & 36.0 & 22.0 & 38.0 & 26.0 & 32.0 & 24.0 & 31.0 \\
    LLaVA~\cite{llava} & 38.0 & 0.0 & 0.0 & 0.0 & 0.0 & 0.0 & 0.0 & 5.4 \\
     QWen-VL~\cite{bai2023qwen} & 100.0 & 0.0 & 0.0 & 0.0 & 0.0 & 0.0 & 0.0 & 14.3 \\
    InstructBLIP~\cite{dai2305instructblip} & 100.0 & 5.0 & 0.0 & 4.0 & 0.0 & 0.0 & 2.0 & 15.9 \\
    mPLUG-owl~\cite{ye2023mplug} &22.0 & 97.0 & 93.0 & 94.0 & 97.0 & 80.0 & 95.0 & 82.6 \\
    Gemini~\cite{team2023gemini} & 4.0 & 27.0 & 32.0 & 5.0 & 4.0 & 1.0 & 24.0 & 13.9 \\
    GPT4V~\cite{gpt4} & 80.0 & 82.8 & 66.3 & 35.4 & 28.4 & 12.0 & 51.5 & 50.3 \\
    \bottomrule
  \end{tabular}
\end{table*}

\begin{table*}[t]
  \caption{HTER of various MLLMs on FAS true/false one-shot questions}
  \label{appendix_fas_07}
  \scriptsize
  \centering
  \begin{tabular}{ccccccccc}
    \toprule
    \multirow{2}{*}{Model} & \multicolumn{8}{c}{HTER (\%)$\downarrow$} \\
    \cmidrule(r){2-9}
    & Bona fide & Fake head & Flexible mask & Paper mask & Print & Replay & Rigid mask & AVG \\
    \midrule
    BLIP~\cite{blip} & 0.0 & 9.0 & 41.5 & 2.5 & 50.0 & 50.0 & 44.5 & 32.9 \\
     BLIP-2~\cite{li2023blip} & 18.5 & 50.0 & 50.0 & 50.0 & 50.0 & 50.0 & 50.0 & 68.5 \\
    Intern~\cite{Intern} & 0.0 & 50.0 & 50.0 & 50.0 & 50.0 & 50.0 & 50.0 & 50.0 \\
    MiniGPT-4~\cite{zhu2023minigpt} &18.5 & 40.0 & 39.0 & 35.5 & 37.0 & 39.0 & 40.0 & 56.9 \\
    LLaVA~\cite{llava} & 0.0 & 50.0 & 50.0 & 50.0 & 50.0 & 50.0 & 50.0 & 50.0 \\
     QWen-VL~\cite{bai2023qwen} & 0.0 & 50.0 & 50.0 & 50.0 & 50.0 & 50.0 & 50.0 & 50.0 \\
    InstructBLIP~\cite{dai2305instructblip} & 0.0 & 49.0 & 50.0 & 42.0 & 50.0 & 50.0 & 48.5 & 48.3 \\
    mPLUG-owl~\cite{ye2023mplug} &39.5 & 21.5 & 16.5 & 15.0 & 20.0 & 37.0 & 15.0 & 60.3 \\
    Gemini~\cite{team2023gemini} & 17.0 & 15.0 & 6.5 & 4.5 & 37.5 & 44.5 & 8.0 & 36.3 \\
    GPT4V~\cite{gpt4} & 11.6 & 3.5 & 10.2 & 18.8 & 26.4 & 31.1 & 16.7 & 28.8 \\
    \bottomrule
  \end{tabular}
\end{table*}

\begin{table*}[t]
  \caption{HTER of various MLLMs on FAS true/false one-shot questions with COT}
  \label{appendix_fas_08}
  \scriptsize
  \centering
  \begin{tabular}{ccccccccc}
    \toprule
    \multirow{2}{*}{Model} & \multicolumn{8}{c}{HTER (\%)$\downarrow$} \\
    \cmidrule(r){2-9}
    & Bona fide & Fake head & Flexible mask & Paper mask & Print & Replay & Rigid mask & AVG \\
    \midrule
    BLIP~\cite{blip} & 0.0 & 0.0 & 17.5 & 0.0 & 32.5 & 32.0 & 6.5 & 14.8 \\
     BLIP-2~\cite{li2023blip} & 48.5 & 50.0 & 50.0 & 50.0 & 50.0 & 50.0 & 50.0 & 98.5 \\
    Intern~\cite{Intern} & 0.0 & 40.5 & 50.0 & 48.0 & 50.0 & 50.0 & 49.5 & 48.0 \\
    MiniGPT-4~\cite{zhu2023minigpt} &30.5 & 32.0 & 39.0 & 31.0 & 37.0 & 34.0 & 38.0 & 65.7 \\
    LLaVA~\cite{llava} & 31.0 & 50.0 & 50.0 & 50.0 & 50.0 & 50.0 & 50.0 & 81.0 \\
     QWen-VL~\cite{bai2023qwen} & 0.0 & 50.0 & 50.0 & 50.0 & 50.0 & 50.0 & 50.0 & 50.0 \\
    InstructBLIP~\cite{dai2305instructblip} & 0.0 & 47.5 & 50.0 & 48.0 & 50.0 & 50.0 & 49.0 & 49.1 \\
    mPLUG-owl~\cite{ye2023mplug} &39.0 & 1.5 & 3.5 & 3.0 & 1.5 & 10.0 & 2.5 & 42.7 \\
    Gemini~\cite{team2023gemini} & 48.0 & 36.5 & 34.0 & 47.5 & 48.0 & 49.5 & 38.0 & 90.2 \\
    GPT4V~\cite{gpt4} & 10.0 & 8.6 & 16.8 & 32.3 & 35.8 & 44.0 & 24.2 & 36.8 \\
    \bottomrule
  \end{tabular}
\end{table*}

\begin{table*}[t] 
  \caption{Accuracy of various MLLMs on unified task true/false questions}
  \label{unifiedFewshot}
  \centering
  \begin{tabular}{ccc}
    \toprule
    \multirow{2}{*}{Model} & \multicolumn{2}{c}{ACC (\%)} \\ 
    \cmidrule(r){2-3}
    & Few-shot & Few-shot (COT) \\ 
    \midrule
    BLIP~\cite{blip}~ & 34.0 & 41.0 \\
    BLIP-2~\cite{li2023blip} & 34.0 & 9.0 \\
    Intern~\cite{Intern}  & 33.0 & 30.0 \\
    MiniGPT-4~\cite{zhu2023minigpt} & 18.0 & 24.0 \\
    LLaVA~\cite{llava} &  33.0 & 41.0 \\
    QWen-VL~\cite{bai2023qwen} &20.0 & 21.0 \\
    InstructBLIP~\cite{dai2305instructblip} & 44.0 & 54.0 \\
    mPLUG-owl~\cite{ye2023mplug} & 34.0 & 35.0 \\
    Gemini~\cite{team2023gemini} & 41.0 & 45.9 \\
    GPT4V~\cite{gpt4} & 25.5 & 24.2\\
    \bottomrule
  \end{tabular}
\end{table*}

\begin{table*}[t]
  \caption{Accuracy of various MLLMs on unified task multiple-choice questions}
  \label{unifiedMC}
  \scriptsize
  \setlength{\tabcolsep}{5pt}
  \centering
  \begin{tabular}{ccccccccccc}
    \toprule
    \multirow{3}{*}{Model} & \multicolumn{10}{c}{ACC (\%)} \\ 
    \cmidrule(r){2-11}
    & \multicolumn{2}{c}{Zero-shot (real)} & \multicolumn{2}{c}{One-shot (real)} & \multicolumn{2}{c}{Zero-shot (attack)} & \multicolumn{2}{c}{One-shot (attack)} & \multicolumn{2}{c}{AVG} \\ 
    \cmidrule(r){2-11}
    & - & COT & - & COT & - & COT & - & COT & - & COT \\ 
    \midrule
    BLIP~\cite{blip} & 26.0 & 25.0 & 8.0 & 24.0 & 24.0 & 24.0 & 5.0 & 8.0& 15.8 & 20.3 \\
    BLIP-2~\cite{li2023blip} & 0.0 & 0.0 & 20.0 & 0.0 & 2.0 & 0.0 & 2.0 & 0.0 & 6.0 & 0.0 \\
    Intern~\cite{Intern} &15.0 & 2.0 & 24.0 & 14.0 & 7.0 & 0.0 & 6.0 & 0.0 & 13.0 & 4.0 \\
    MiniGPT-4~\cite{zhu2023minigpt} & 0.0 & 0.0 & 1.0 & 0.0 & 2.0 & 0.0 & 1.0 & 0.0 & 1.0 & 0.0 \\
    LLaVA~\cite{llava} & 13.0 & 0.0 & 27.0 & 14.0 & 6.0 & 1.0 & 21.0 & 4.0 & 16.8 & 4.8 \\
    QWen-VL~\cite{bai2023qwen} & 1.0 & 0.0 & 22.0 & 0.0 & 1.0 & 0.0 & 22.0 & 0.0 & 11.5 & 0.0 \\
    InstructBLIP~\cite{dai2305instructblip} & 12.0 & 27.0 & 3.0 & 0.0 & 2.0 & 1.0 & 5.0 & 2.0 & 5.5 & 7.5 \\
    mPLUG-owl~\cite{ye2023mplug} & 12.0 & 0.0 & 27.0 & 1.0 & 5.0 & 0.0 & 21.0 & 0.0 & 16.3 & 0.3 \\
    Gemini~\cite{team2023gemini} & 24.0 & 22.0 & 14.0 & 23.0 & 27.8 & 14.0 & 11.9 & 15.0 & 20.2 & 18.7 \\
    GPT4V~\cite{gpt4} & 27.2 & 27.4 & 28.6 & 34.0 & 29.3 & 15.4 & 34.0 & 26.8 & 29.2 & 26.8\\
    \bottomrule
  \end{tabular}
\end{table*}

\begin{table*}[t]
  \caption{Accuracy of various MLLMs on face forgery detection true/false zero-shot questions}
  \label{appendix_df_01}
  \scriptsize
  \setlength{\tabcolsep}{2.2pt}
  \centering
  \begin{tabular}{ccccccccc}
    \toprule
    \multirow{2}{*}{Model} & \multicolumn{8}{c}{ACC (\%)} \\ 
    \cmidrule(r){2-9}
    & Bona fide & Deepfakes & Face2Face & FaceSwap & InsightFace & NeuralTextures & Stable\_Diffusion & AVG \\
    \midrule
    BLIP~\cite{blip} & 92.0 & 5.0 & 9.0 & 10.0 & 2.0 & 2.0 & 5.0 & 17.9 \\
     BLIP-2~\cite{li2023blip} & 100.0 & 0.0 & 0.0 & 0.0 & 0.0 & 0.0 & 0.0 & 14.3 \\
    Intern~\cite{Intern} & 100.0 & 0.0 & 0.0 & 0.0 & 2.0 & 0.0 & 0.0 & 14.6 \\
    MiniGPT-4~\cite{zhu2023minigpt} &100.0 & 5.0 & 2.0 & 0.0 & 8.0 & 1.0 & 8.0 & 17.7 \\
    LLaVA~\cite{llava} & 100.0 & 0.0 & 0.0 & 0.0 & 0.0 & 0.0 & 0.0 & 14.3 \\
     QWen-VL~\cite{bai2023qwen} & 100.0 & 0.0 & 0.0 & 0.0 & 0.0 & 0.0 & 0.0 & 14.3 \\
    InstructBLIP~\cite{dai2305instructblip} & 100.0 & 15.0 & 2.0 & 6.0 & 5.0 & 0.0 & 2.0 & 18.6 \\
    mPLUG-owl~\cite{ye2023mplug} &100.0 & 9.0 & 5.0 & 4.0 & 12.0 & 0.0 & 23.0 & 21.9 \\
    Gemini~\cite{team2023gemini} & 96.0 & 54.0 & 30.0 & 34.0 & 17.0 & 7.0 & 20.0 & 36.8 \\
    GPT4V~\cite{gpt4} & 98.1 & 26.1 & 16.4 & 16.7 & 9.1 & 3.2 & 12.5 & 26.0 \\
    \bottomrule
  \end{tabular}
\end{table*}

\begin{table*}[t]
  \caption{Accuracy of various MLLMs on face forgery detection true/false zero-shot questions with COT}
  \label{appendix_df_02}
  \setlength{\tabcolsep}{2.2pt}
  \scriptsize
  \centering
  \begin{tabular}{ccccccccc}
    \toprule
    \multirow{2}{*}{Model} & \multicolumn{8}{c}{ACC (\%)} \\
    \cmidrule(r){2-9}
    & \multicolumn{1}{l}{Bona fide} & \multicolumn{1}{l}{Deepfakes} & \multicolumn{1}{l}{Face2Face} & \multicolumn{1}{l}{FaceSwap} & \multicolumn{1}{l}{InsightFace} & \multicolumn{1}{l}{NeuralTextures} & \multicolumn{1}{l}{Stable\_Diffusion} & \multicolumn{1}{l}{AVG} \\
    \midrule
    BLIP~\cite{blip} & 78.0 & 38.0 & 28.0 & 42.0 & 7.0 & 15.0 & 9.0 & 31.0 \\
     BLIP-2~\cite{li2023blip} & 100.0 & 0.0 & 0.0 & 0.0 & 0.0 & 0.0 & 0.0 & 14.3 \\
    Intern~\cite{Intern} & 100.0 & 0.0 & 0.0 & 0.0 & 0.0 & 0.0 & 0.0 & 14.3 \\
    MiniGPT-4~\cite{zhu2023minigpt} &99.0 & 6.0 & 3.0 & 8.0 & 8.0 & 0.0 & 3.0 & 18.1 \\
    LLaVA~\cite{llava} & 100.0 & 0.0 & 0.0 & 0.0 & 0.0 & 0.0 & 0.0 & 14.3 \\
     QWen-VL~\cite{bai2023qwen} & 100.0 & 0.0 & 0.0 & 0.0 & 0.0 & 0.0 & 0.0 & 14.3 \\
    InstructBLIP~\cite{dai2305instructblip} & 100.0 & 34.0 & 7.0 & 10.0 & 9.0 & 0.0 & 7.0 & 23.9 \\
    mPLUG-owl~\cite{ye2023mplug} &100.0 & 9.0 & 2.0 & 4.0 & 11.0 & 0.0 & 20.0 & 20.9 \\
    Gemini~\cite{team2023gemini} & 86.0 & 77.0 & 42.0 & 61.0 & 29.3 & 17.0 & 27.3 & 48.6 \\
    GPT4V~\cite{gpt4} & 95.7 & 25.3 & 13.1 & 22.1 & 11.6 & 2.4 & 23.1 & 28.8 \\
    \bottomrule
  \end{tabular}
\end{table*}

\begin{table*}[t]
  \caption{HTER of various MLLMs on face forgery detection true/false zero-shot questions}
  \label{appendix_df_03}
  \setlength{\tabcolsep}{2.2pt}
  \scriptsize
  \centering
  \begin{tabular}{ccccccccc}
    \toprule
    \multirow{2}{*}{Model} & \multicolumn{8}{c}{HTER (\%)$\downarrow$ } \\
    \cmidrule(r){2-9}
    & \multicolumn{1}{l}{Bona fide} & \multicolumn{1}{l}{Deepfakes} & \multicolumn{1}{l}{Face2Face} & \multicolumn{1}{l}{FaceSwap} & \multicolumn{1}{l}{InsightFace} & \multicolumn{1}{l}{NeuralTextures} & \multicolumn{1}{l}{Stable\_Diffusion} & AVG \\
    \midrule
    BLIP~\cite{blip} & 4.0 & 47.5 & 45.5 & 45.0 & 49.0 & 49.0 & 47.5 & 51.2 \\
     BLIP-2~\cite{li2023blip} & 0.0 & 50.0 & 50.0 & 50.0 & 50.0 & 50.0 & 50.0 & 50.0 \\
    Intern~\cite{Intern} & 0.0 & 50.0 & 50.0 & 50.0 & 49.0 & 50.0 & 50.0 & 49.8 \\
    MiniGPT-4~\cite{zhu2023minigpt} &0.0 & 47.5 & 49.0 & 50.0 & 46.0 & 49.5 & 46.0 & 48.0 \\
    LLaVA~\cite{llava} & 0.0 & 50.0 & 50.0 & 50.0 & 50.0 & 50.0 & 50.0 & 50.0 \\
     QWen-VL~\cite{bai2023qwen} & 0.0 & 50.0 & 50.0 & 50.0 & 50.0 & 50.0 & 50.0 & 50.0 \\
    InstructBLIP~\cite{dai2305instructblip} & 0.0 & 42.5 & 49.0 & 47.0 & 47.5 & 50.0 & 49.0 & 47.5 \\
    mPLUG-owl~\cite{ye2023mplug} &0.0 & 45.5 & 47.5 & 48.0 & 44.0 & 50.0 & 38.5 & 45.6 \\
    Gemini~\cite{team2023gemini} & 2.0 & 23.0 & 35.0 & 33.0 & 41.5 & 46.5 & 40.0 & 38.5 \\
    GPT4V~\cite{gpt4} & 0.9 & 37.0 & 41.8 & 41.7 & 45.5 & 48.4 & 43.8 & 43.6 \\
    \bottomrule
  \end{tabular}
\end{table*}
\clearpage

\begin{table*}[t]
  \caption{HTER of various MLLMs on face forgery detection true/false zero-shot questions with COT}
  \label{appendix_df_04}
  \setlength{\tabcolsep}{2.2pt}
  \scriptsize
  \centering
  \begin{tabular}{ccccccccc}
    \toprule
    \multirow{2}{*}{Model} & \multicolumn{8}{c}{HTER (\%)$\downarrow$ } \\
    \cmidrule(r){2-9}
    & Bona fide & Deepfakes & Face2Face & FaceSwap & InsightFace & NeuralTextures & Stable\_Diffusion & AVG \\
    \midrule
    BLIP~\cite{blip} & 11.0 & 31.0 & 36.0 & 29.0 & 46.5 & 42.5 & 45.5 & 49.4 \\
     BLIP-2~\cite{li2023blip} & 0.0 & 50.0 & 50.0 & 50.0 & 50.0 & 50.0 & 50.0 & 50.0 \\
    Intern~\cite{Intern} & 0.0 & 50.0 & 50.0 & 50.0 & 50.0 & 50.0 & 50.0 & 50.0 \\
    MiniGPT-4~\cite{zhu2023minigpt} &0.5 & 47.0 & 48.5 & 46.0 & 46.0 & 50.0 & 48.5 & 48.2 \\
    LLaVA~\cite{llava} & 0.0 & 50.0 & 50.0 & 50.0 & 50.0 & 50.0 & 50.0 & 50.0 \\
     QWen-VL~\cite{bai2023qwen} & 0.0 & 50.0 & 50.0 & 50.0 & 50.0 & 50.0 & 50.0 & 50.0 \\
    InstructBLIP~\cite{dai2305instructblip} & 0.0 & 33.0 & 46.5 & 45.0 & 45.5 & 50.0 & 46.5 & 44.4 \\
    mPLUG-owl~\cite{ye2023mplug} &0.0 & 45.5 & 49.0 & 48.0 & 44.5 & 50.0 & 40.0 & 46.2 \\
    Gemini~\cite{team2023gemini} & 7.0 & 11.5 & 29.0 & 19.5 & 35.4 & 41.5 & 36.4 & 35.9 \\
    GPT4V~\cite{gpt4} & 2.2 & 37.4 & 43.4 & 39.0 & 44.2 & 48.8 & 38.5 & 44.0 \\
    \bottomrule
  \end{tabular}
\end{table*}

\begin{table*}[t]
  \caption{Accuracy of various MLLMs on face forgery detection true/false one-shot questions}
  \label{appendix_df_05}
  \setlength{\tabcolsep}{2.2pt}
  \scriptsize
  \centering
  \begin{tabular}{ccccccccc}
    \toprule
    \multirow{2}{*}{Model} & \multicolumn{8}{c}{ACC (\%)} \\
    \cmidrule(r){2-9}
    & Bona fide & Deepfakes & Face2Face & FaceSwap & InsightFace & NeuralTextures & Stable\_Diffusion & AVG \\
    \midrule
    BLIP~\cite{blip} & 100.0 & 53.0 & 48.0 & 53.0 & 27.0 & 24.0 & 39.0 & 49.1 \\
     BLIP-2~\cite{li2023blip} & 71.0 & 0.0 & 0.0 & 0.0 & 0.0 & 0.0 & 0.0 & 10.1 \\
    Intern~\cite{Intern} & 100.0 & 0.0 & 0.0 & 0.0 & 0.0 & 0.0 & 0.0 & 14.3 \\
    MiniGPT-4~\cite{zhu2023minigpt} &58.0 & 26.0 & 21.0 & 32.0 & 33.0 & 30.0 & 25.0 & 32.1 \\
    LLaVA~\cite{llava} & 100.0 & 0.0 & 0.0 & 0.0 & 0.0 & 0.0 & 0.0 & 14.3 \\
     QWen-VL~\cite{bai2023qwen} & 100.0 & 0.0 & 0.0 & 0.0 & 0.0 & 0.0 & 0.0 & 14.3 \\
    InstructBLIP~\cite{dai2305instructblip} & 100.0 & 2.0 & 0.0 & 0.0 & 0.0 & 0.0 & 0.0 & 14.6 \\
    mPLUG-owl~\cite{ye2023mplug} &76.0 & 0.0 & 0.0 & 0.0 & 9.0 & 0.0 & 4.0 & 12.7 \\
    Gemini~\cite{team2023gemini} & 76.0 & 38.0 & 19.0 & 29.0 & 10.0 & 7.0 & 10.0 & 27.0 \\
    GPT4V~\cite{gpt4} & 100.0 & 13.3 & 12.2 & 10.3 & 10.6 & 3.0 & 9.4 & 22.4 \\
    \bottomrule
  \end{tabular}
\end{table*}

\begin{table*}[t]
  \caption{Accuracy of various MLLMs on face forgery detection true/false one-shot questions with COT}
  \label{appendix_df_06}
  \setlength{\tabcolsep}{2.2pt}
  \scriptsize
  \centering
  \begin{tabular}{ccccccccc}
    \toprule
    \multirow{2}{*}{Model} & \multicolumn{8}{c}{ACC (\%)} \\ 
    \cmidrule(r){2-9}
     & Bona fide & Deepfakes & Face2Face & FaceSwap & InsightFace & NeuralTextures & Stable\_Diffusion & AVG \\
    \midrule
    BLIP~\cite{blip} & 97.0 & 75.0 & 77.0 & 77.0 & 26.0 & 48.0 & 38.0 & 62.6 \\
     BLIP-2~\cite{li2023blip} & 12.0 & 0.0 & 0.0 & 0.0 & 0.0 & 0.0 & 0.0 & 1.7 \\
    Intern~\cite{Intern} & 100.0 & 0.0 & 0.0 & 0.0 & 0.0 & 0.0 & 0.0 & 14.3 \\
    MiniGPT-4~\cite{zhu2023minigpt} &70.0 & 28.0 & 25.0 & 34.0 & 36.0 & 32.0 & 28.0 & 36.1 \\
    LLaVA~\cite{llava} & 92.0 & 0.0 & 0.0 & 0.0 & 0.0 & 0.0 & 0.0 & 13.1 \\
     QWen-VL~\cite{bai2023qwen} & 100.0 & 0.0 & 0.0 & 0.0 & 0.0 & 0.0 & 0.0 & 14.3 \\
    InstructBLIP~\cite{dai2305instructblip} & 100.0 & 7.0 & 2.0 & 6.0 & 1.0 & 0.0 & 1.0 & 16.7 \\
    mPLUG-owl~\cite{ye2023mplug} &89.0 & 6.0 & 5.0 & 2.0 & 21.0 & 0.0 & 23.0 & 20.9 \\
    Gemini~\cite{team2023gemini} & 21.0 & 21.0 & 14.0 & 15.0 & 13.0 & 4.0 & 10.0 & 14.0 \\
    GPT4V~\cite{gpt4} & 95.8 & 18.0 & 9.1 & 28.7 & 14.4 & 4.1 & 13.5 & 26.0 \\
    \bottomrule
  \end{tabular}
\end{table*}
\begin{table*}[h]
  \caption{HTER of various MLLMs on face forgery detection true/false one-shot questions}
  \label{appendix_df_07}
  \setlength{\tabcolsep}{2.2pt}
  \scriptsize
  \centering
  \begin{tabular}{ccccccccc}
    \toprule
    \multirow{2}{*}{Model} & \multicolumn{8}{c}{HTER (\%)$\downarrow$} \\
    \cmidrule(r){2-9}
    & \multicolumn{1}{l}{Bona fide} & \multicolumn{1}{l}{Deepfakes} & \multicolumn{1}{l}{Face2Face} & \multicolumn{1}{l}{FaceSwap} & \multicolumn{1}{l}{InsightFace} & \multicolumn{1}{l}{NeuralTextures} & \multicolumn{1}{l}{Stable\_Diffusion} & AVG \\
    \midrule
    BLIP~\cite{blip} & 0.0 & 23.5 & 26.0 & 23.5 & 36.5 & 38.0 & 30.5 & 29.7 \\
     BLIP-2~\cite{li2023blip} & 14.5 & 50.0 & 50.0 & 50.0 & 50.0 & 50.0 & 50.0 & 64.5 \\
    Intern~\cite{Intern} & 0.0 & 50.0 & 50.0 & 50.0 & 50.0 & 50.0 & 50.0 & 50.0 \\
    MiniGPT-4~\cite{zhu2023minigpt} &21.0 & 37.0 & 39.5 & 34.0 & 33.5 & 35.0 & 37.5 & 57.1 \\
    LLaVA~\cite{llava} & 0.0 & 50.0 & 50.0 & 50.0 & 50.0 & 50.0 & 50.0 & 50.0 \\
     QWen-VL~\cite{bai2023qwen} & 0.0 & 50.0 & 50.0 & 50.0 & 50.0 & 50.0 & 50.0 & 50.0 \\
    InstructBLIP~\cite{dai2305instructblip} & 0.0 & 49.0 & 50.0 & 50.0 & 50.0 & 50.0 & 50.0 & 49.8 \\
    mPLUG-owl~\cite{ye2023mplug} &12.0 & 50.0 & 50.0 & 50.0 & 45.5 & 50.0 & 48.0 & 60.9 \\
    Gemini~\cite{team2023gemini} & 12.0 & 30.5 & 40.5 & 35.5 & 45.0 & 46.5 & 45.0 & 52.5 \\
    GPT4V~\cite{gpt4} & 0.0 & 43.4 & 43.9 & 44.9 & 44.7 & 48.5 & 45.3 & 45.1 \\
    \bottomrule
  \end{tabular}
\end{table*}

\clearpage
\section{Results of MA-COT}
\label{res_macot}
In this section, we conducted a qualitative test on selected data to gain an initial understanding of MA-COT's performance, exploring the potential of MA-COT with general-purpose MLLMs. We tested the application effectiveness of the MA-COT method previously mentioned, across the FAS task, the face forgery detection task, and the unified task, with a subset of the results visualized. By employing the MA-COT method, we guide the MLLMs to focus on key features related to presentation attacks using prior knowledge, thereby more efficiently completing sub-tasks without the direct introduction of such knowledge. Moreover, since these key features can be easily added as plug-and-play components, the MA-COT method also facilitates the seamless integration of the FAS task and the face forgery detection task into a unified task framework.

Figure~\ref{macot_compare} presents a comparison of the average metrics for GPT4V in true/false question with or without the use of MA-COT. It is observed that the average accuracy slightly decreases after employing MA-COT, while the HTER remains relatively stable. In particular, the refusal rate of GPT4V decreases significantly, demonstrating that MA-COT can effectively reduce the incidence of refusal. Figure~\ref{macot_detail} shows the result of each attack types with or without the use of MA-COT on GPT4V. The use of MA-COT notably improves performance in areas where GPT4V initially struggled, such as print and replay attacks, though it may reduce performance in areas where it previously excelled. This suggests that the MA-COT strategy is effective, but the key attributes considered by it still need further exploration.

\begin{figure}[!htbp]
  \centering
  \begin{center}
  \centerline{\includegraphics[width=0.9\linewidth]{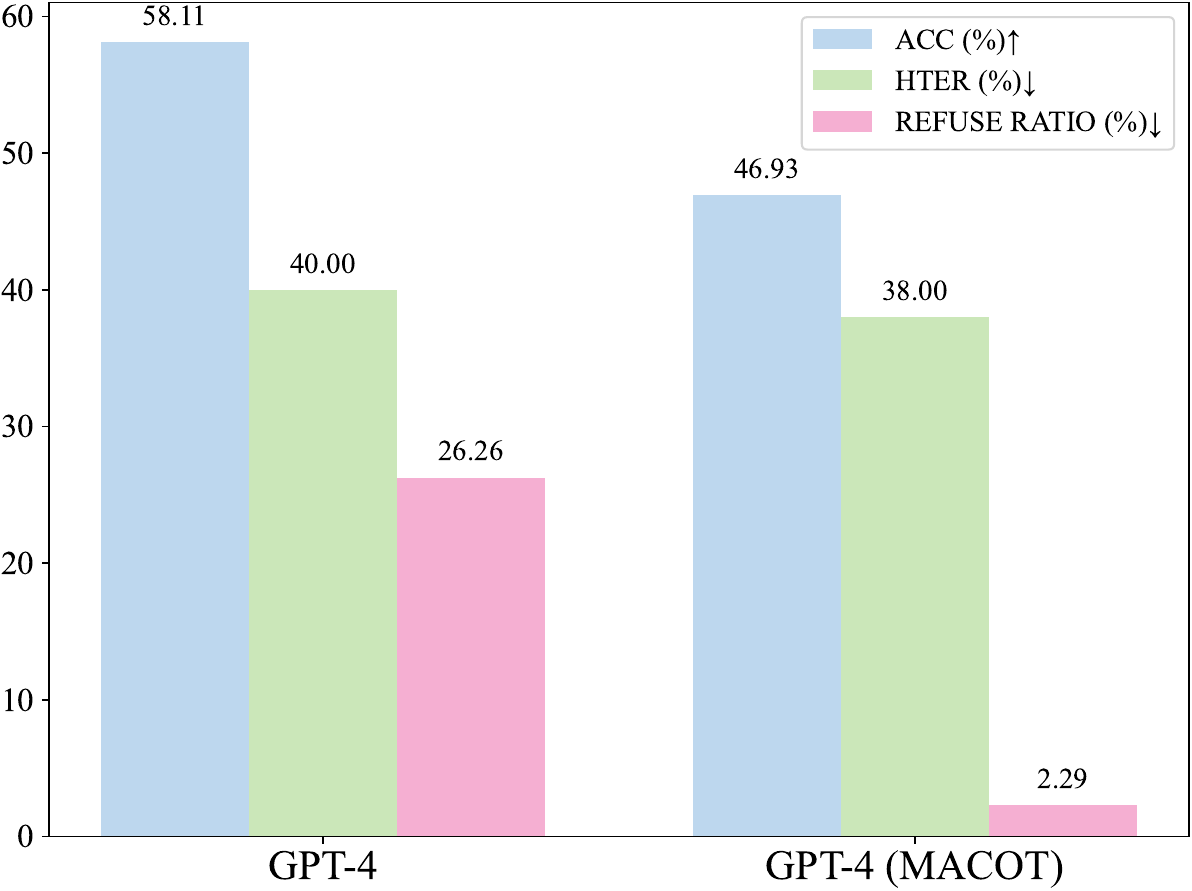}}
  \end{center}
  \caption{Comparative results of the GPT4V on the FAS zero-shot task, with or without the MA-COT.}
  \label{macot_compare}
\end{figure}

\begin{figure}[!htbp]
  \centering
  \begin{center}
  \centerline{\includegraphics[width=\linewidth]{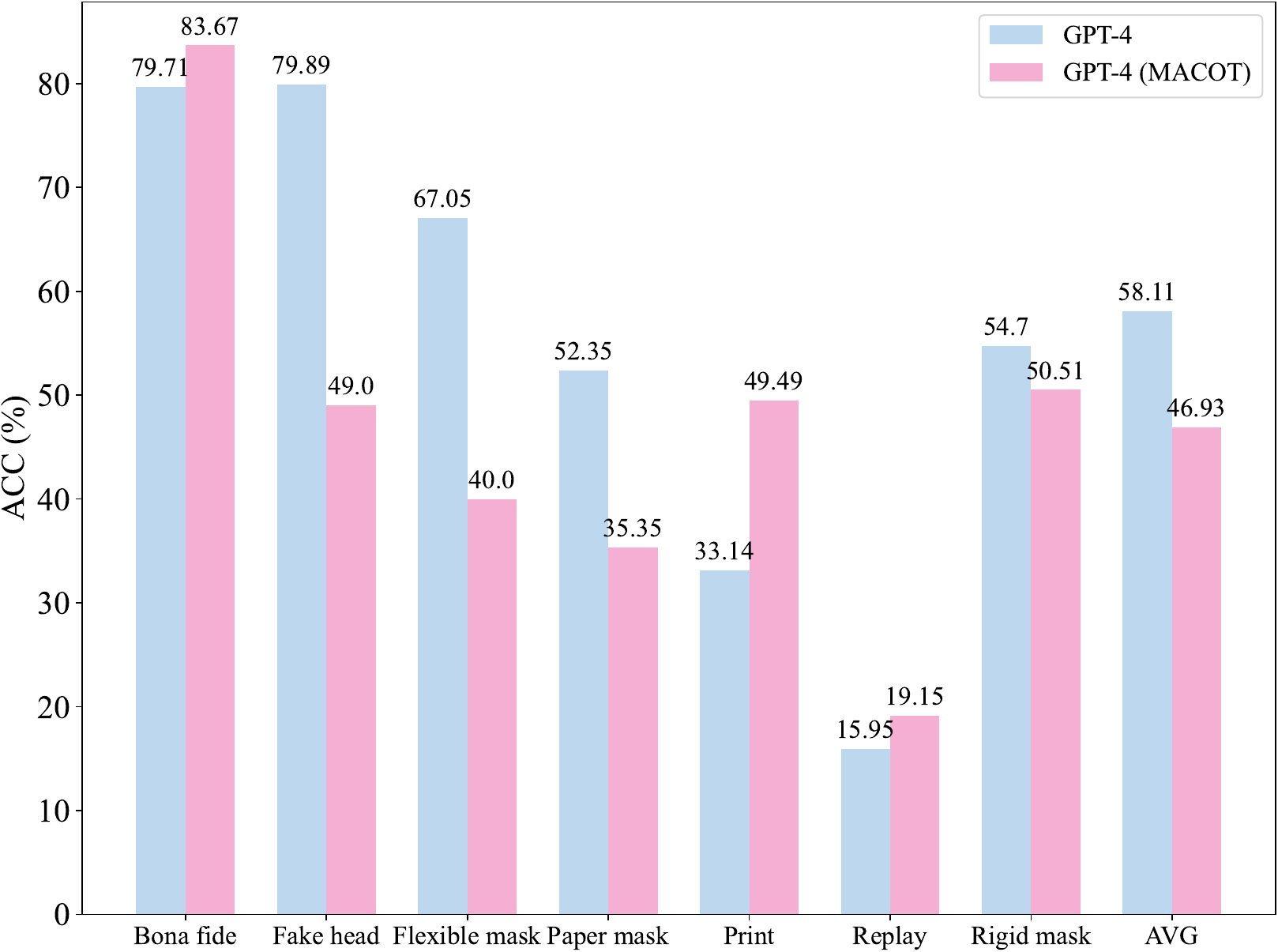}}
  \end{center}
  \caption{Results of GPT4V on true/false zero-shot FAS tasks, showing performance across different attacks with or without MA-COT.}
  \label{macot_detail}
\end{figure}

\vspace{-4mm}
\section{Findings and Discussion}

\subsection{More Comprehensive Dataset} Currently, face security datasets are limited by the lack of dedicated pretraining and supervised fine-tuning (SFT) data, as well as a robust data construction pipeline. Developing a dataset that addresses these gaps by including both high-quality pre-training data and a structured pipeline for data generation is essential to improving model interpretability and generalizability. Future work could focus on constructing such datasets, which would enable more effective training of models for complex security tasks and enhance their performance.

\subsection{More comprehensive Evaluation Metrics} 
Current evaluation metrics for FAS and face forgery tasks are relatively limited. Incorporating evaluation metrics from NLP and other multimodal domains could provide a more comprehensive assessment of model performance and robustness. For the face forgery task, metrics such as area under curve (AUC), average precision (AP), and equal error rate (EER) provide critical insights into model capabilities. AUC and AP measure the ability to distinguish between real and forged faces, while EER reflects the trade-off between false acceptance and rejection rates. These metrics ensure a rigorous evaluation of model performance in diverse forgery scenarios.

\begin{table*}[h]
  \caption{HTER of various MLLMs on face forgery detection true/false one-shot questions with COT}
  \label{appendix_df_08}
  \setlength{\tabcolsep}{2.2pt}
  \scriptsize
  \centering
  \begin{tabular}{ccccccccc}
\toprule
    \multirow{2}{*}{Model} & \multicolumn{8}{c}{HTER (\%)$\downarrow$} \\
    \cmidrule(r){2-9}
    & \multicolumn{1}{l}{Bona fide} & \multicolumn{1}{l}{Deepfakes} & \multicolumn{1}{l}{Face2Face} & \multicolumn{1}{l}{FaceSwap} & \multicolumn{1}{l}{InsightFace} & \multicolumn{1}{l}{NeuralTextures} & \multicolumn{1}{l}{Stable\_Diffusion} & \multicolumn{1}{l}{AVG} \\
    \midrule
    BLIP~\cite{blip} & 1.5 & 12.5 & 11.5 & 11.5 & 37.0 & 26.0 & 31.0 & 23.1 \\
     BLIP-2~\cite{li2023blip} & 44.0 & 50.0 & 50.0 & 50.0 & 50.0 & 50.0 & 50.0 & 94.0 \\
    Intern~\cite{Intern} & 0.0 & 50.0 & 50.0 & 50.0 & 50.0 & 50.0 & 50.0 & 50.0 \\
    MiniGPT-4~\cite{zhu2023minigpt} &15.0 & 36.0 & 37.5 & 33.0 & 32.0 & 34.0 & 36.0 & 49.8 \\
    LLaVA~\cite{llava} & 4.0 & 50.0 & 50.0 & 50.0 & 50.0 & 50.0 & 50.0 & 54.0 \\
     QWen-VL~\cite{bai2023qwen} & 0.0 & 50.0 & 50.0 & 50.0 & 50.0 & 50.0 & 50.0 & 50.0 \\
    InstructBLIP~\cite{dai2305instructblip} & 0.0 & 46.5 & 49.0 & 47.0 & 49.5 & 50.0 & 49.5 & 48.6 \\
    mPLUG-owl~\cite{ye2023mplug} &5.5 & 47.0 & 47.5 & 49.0 & 39.5 & 50.0 & 38.5 & 50.8 \\
    Gemini~\cite{team2023gemini} & 39.5 & 39.5 & 43.0 & 42.5 & 43.5 & 48.0 & 45.0 & 83.1 \\
    GPT4V~\cite{gpt4} & 2.1 & 41.0 & 45.5 & 35.6 & 42.8 & 47.9 & 43.2 & 44.8 \\
    \bottomrule
  \end{tabular}
\end{table*}
\begin{table*}[h!]
\centering
\caption{Performance metrics on the FAS task with LLaVA and supervised fine-tuning (SFT) improvements}
\label{tab:FAS_results}
\resizebox{\textwidth}{!}{%
\begin{tabular}{lccccc}
\toprule
\textbf{Model}       & \textbf{ACC (\%)} \textuparrow & \textbf{HTER (\%)} \textdownarrow & \textbf{BLEU-4 (\%)} \textuparrow & \textbf{ROUGE-L (\%)} \textuparrow & \textbf{METEOR (\%)} \textuparrow \\ 
\midrule
LLaVA                & 65.54                         & 27.76                            & 17.80                             & 30.51                             & 25.52                            \\ 
LLaVA (SFT)          & 98.06 (\textcolor{red}{+32.52})                         & 1.74 (\textcolor{red}{-26.02})                            & 82.63 (\textcolor{red}{+64.83})                            & 81.72 (\textcolor{red}{+51.21})                            & 51.97 (\textcolor{red}{+26.45})                                               \\ 
\bottomrule
\end{tabular}%
}
\end{table*}

\begin{table*}[h!]
\centering
\caption{Performance metrics on the face forgery task with LLaVA and SFT improvements}
\label{tab:FaceForgery_results}
\resizebox{\textwidth}{!}{%
\begin{tabular}{lcccccc}
\toprule
\textbf{Model}       & \textbf{AUC (\%)} \textuparrow & \textbf{AP (\%)} \textuparrow & \textbf{EER (\%)} \textdownarrow & \textbf{BLEU-4 (\%)} \textuparrow & \textbf{ROUGE-L (\%)} \textuparrow & \textbf{METEOR (\%)} \textuparrow \\ 
\midrule
LLaVA                & 69.30                        & 94.90                       & 37.20                          & 33.80                           & 43.50                            & 29.50                           \\ 
LLaVA (SFT)          & 88.20 (\textcolor{red}{+18.90})                       & 97.90 (\textcolor{red}{+3.00})                       & 18.50 (\textcolor{red}{-18.70})                         & 36.14 (\textcolor{red}{+2.34})                          & 57.84 (\textcolor{red}{+14.34})                           & 34.25 (\textcolor{red}{+4.75})                    \\ 
\bottomrule
\end{tabular}%
}
\end{table*}

NLP-related metrics such as BLEU, ROUGE-L, and METEOR are valuable for assessing the quality of generated textual explanations. These metrics, widely used in natural language processing, can evaluate the fluency, relevance, and informativeness of the generated text, making them particularly useful for understanding the interpretability of multimodal models in face security tasks.
Incorporating these diverse metrics bridges the gap between visual and language modalities, enabling a holistic evaluation of model performance.

Future research could explore ways to integrate these diverse metrics, allowing for a deeper understanding of model capabilities and fostering the development of more interpretable and generalized models for face security tasks.

\vspace{-2.5mm}
\subsection{Task-Specific MLLM}
Developing a task-specific MLLM tailored for face security tasks is a promising direction. This involves techniques such as dedicated pre-training, supervised fine-tuning (SFT), and reinforcement learning from human feedback (RLHF) to optimize the model's performance in specific tasks such as perception, reasoning, and judgment.

The results in Tables~\ref{tab:FAS_results} and~\ref{tab:FaceForgery_results} clearly demonstrate that fine-tuning with task-specific data can significantly enhance model performance across both FAS and face forgery tasks. For the FAS task (Table~\ref{tab:FAS_results}), SFT notably improves detection accuracy (ACC) and reduces error rates (HTER), while also enhancing BLEU-4, ROUGE-L, and METEOR scores, indicating better quality and interpretability of the model’s predictions.

Similarly, for the face forgery task (Table~\ref{tab:FaceForgery_results}), SFT results in substantial gains in AUC and AP, highlighting the model’s improved ability to distinguish between real and forged faces. The reduced EER further demonstrates enhanced robustness, while improved BLEU-4, ROUGE-L, and METEOR scores validate the model's capacity to generate more accurate and meaningful textual explanations. These findings underscore the importance of fine-tuning in adapting MLLMs to meet the specific demands of face security tasks.

\vspace{-2mm}
\subsection{Cross-task Collaboration} Collaborating on similar tasks such as face anti-spoofing and deepfake detection utilizes complementary knowledge for better results. Expanding such collaboration to other fields such as industrial anomaly detection and medical image processing could significantly boost accuracy and efficiency.

\subsection*{\textbf{Data Availability Statement}}

\raggedright
The datasets used in this study are publicly available at:  
WMCA: \url{https://www.idiap.ch/en/scientific-research/data/wmca};  
SiW-Mv2: \url{http://cvlab.cse.msu.edu/category/downloads.html};  
FaceForensics++ (FF++): \url{https://github.com/ondyari/FaceForensics/tree/master/dataset};  
DDF: \url{https://huggingface.co/datasets/OpenRL/DeepFakeFace}.
\normalsize


\subsection*{\textbf{Funding}}
This work was supported by the National Natural Science Foundation of China (No. 62306061), Guangdong Basic and Applied Basic Research Foundation (No. 2023A15151
40037), and Graduate Innovation Fund Project of Shijiazhuang Tiedao University (No. YC202449).
\subsection*{\textbf{Competing Interests}}
The authors declare that they have no other competing interests.
\subsection*{\textbf{Acknowedgements}}
We would like to express our gratitude to the EIT High Performance Computing Platform for its provision of computational resources for this project.
Expressions of gratitude are extended to Zhuofu Tao for his invaluable contributions in optimizing the figures and texts of this manuscript.
\section*{\textbf{Declarations}}
\subsection*{\textbf{Abbreviations}}
ACC, accuracy; AIGC, AI-generated content; COT, chain of thought; DCT, discrete cosine transform; FAS, face anti-spoofing; FF++, faceforensics++; HTER, half total error rate; MLLMs, multimodal large language models; GAN, generative adversarial network; MA-COT, multi-attribute chain of thought; MMFC, maximum magnitude frequency componentization.

\subsection*{\textbf{Author Contributions}}
Yichen Shi conceived the main ideas, designed the overall framework, and drafted the initial manuscript. Yuhao Gao was responsible for designing the overall experiments, conducting some of the FAS experiments, and drafting parts of the figures and text. Yingxin Lai carried out the experiments related to face forgery and contributed to the manuscript writing. The three authors contributed equally and are considered co-first authors. Hongyang Wang implemented the FAS experiments and created the figures for the manuscript. Jun Feng, Lei He, Jun Wan, Changsheng Chen, and Xiaochun Cao served as co-supervisor for this study and oversaw all aspects of the manuscript writing process. The project leader, Zitong Yu, orchestrated and supervised the entire process, encompassing conceptual discussions, experimentation, and manuscript writing.

\vspace{-0.5mm}
\bibliographystyle{unsrt}
\bibliography{reference}

\end{document}


\title{SHIELD\includegraphics[width=0.8cm]{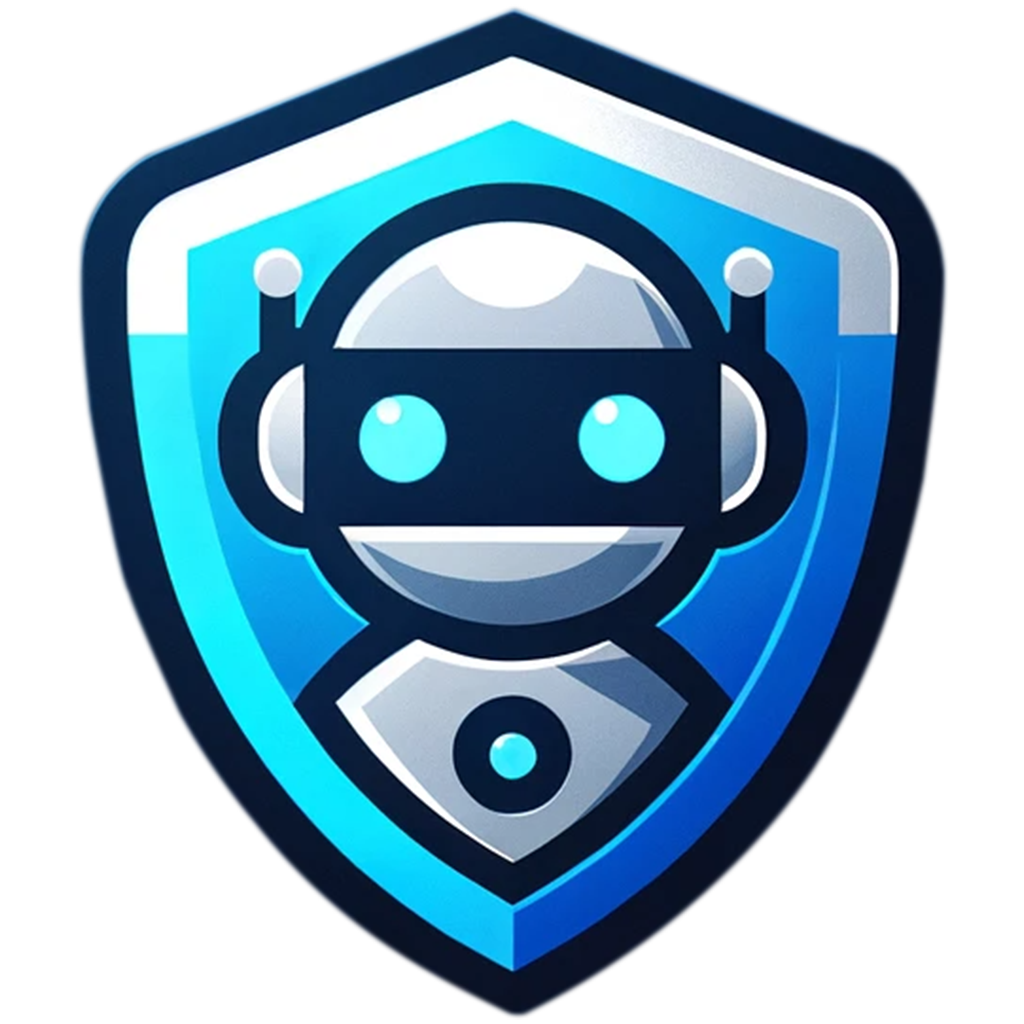}: An Evaluation Benchmark for Face Spoofing and Forgery Detection with Multimodal Large Language Models}

\maketitle

\begin{abstract}
Multimodal large language models (MLLMs) have demonstrated remarkable problem-solving capabilities in various vision fields based on strong visual semantic representation and language reasoning ability. However, whether MLLMs are sensitive to subtle visual spoof/forged clues and how they perform in the domain of face attack detection (e.g., face spoofing and forgery detection) is still unexplored. In this paper, we introduce a new benchmark, namely SHIELD, to evaluate the ability of MLLMs on face spoofing and forgery detection. Specifically, we design true/false and multiple-choice questions to assess multimodal face data across two tasks. For the face anti-spoofing (FAS) task, we test three different modalities—RGB, infrared, and depth—under six presentation attack types. For the face forgery detection task, we evaluate both GAN-based and diffusion-based data incorporating visual and acoustic modalities. We subject each question to zero-shot and few-shot evaluations in both standard and chain of thought (COT) settings.Additionally, we develop a novel Multi-Attribute Chain of Thought (MA-COT) paradigm for describing and judging various task-specific and task-irrelevant attributes of face images. 
The results indicate that MLLMs hold substantial potential in the face security domain. 
\end{abstract}
\section{Introduction}

\begin{figure}[!t]
\centering
\includegraphics[width=\textwidth]{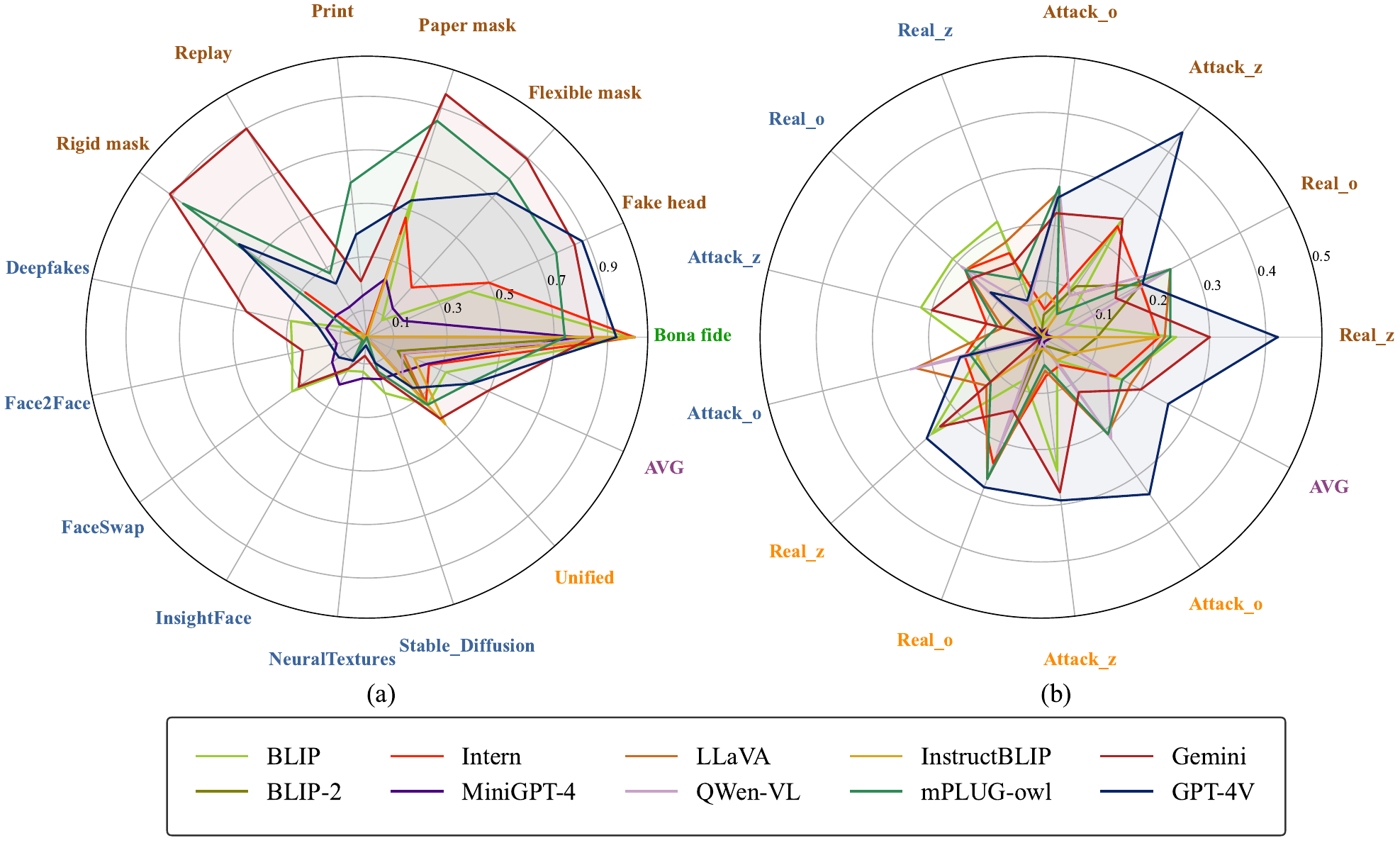}
\caption{Performance of various MLLMs on true/false (a) and multiple-choice (b) questions across different types of attacks, demonstrating their superior ability to distinguish between physical and digital attacks. In (a), the larger the area of each colored polygon indicates better performance, with Qwen-VL and mPLUG-owl outperforming other models. In (b), GPT4V shows the best performance compared to the others. \colorbox{FASColor}{Brown} represents the FAS task, \colorbox{DeepFakeColor}{deep blue} represents the face forgery detection task, and \colorbox{UnifiedColor}{orange} represents the joint task.}
\label{all_per}
\end{figure}

Despite significant progress in face anti-spoofing and face forgery detection, most research still concentrates on developing models for specific scenarios or types of attacks and relying on subtle facial changes. These models typically focus on a single modality or a specific kind of spoofing attack, lacking adaptability to a wider and more varied range of attack scenarios. Meanwhile, Multimodal large language models (MLLMs) \cite{yang2023dawn,team2023gemini} have the potential to provide more comprehensive and robust solutions, enhancing the ability to recognize complex attack methods and elucidating the reasoning and processes behind the judgments, but no one has yet tested them in the field of biometric security. Different from the usage of MLLMs in various vision fields (e.g., generic object recognition and grounding) relying on obvious and semantic visual contexts, subtle visual spoof/forged clues in face attack detection tasks might be more challenging for MLLM understanding. Therefore, this paper aims to explore and fill this research gap, investigating the application and potential advantages of MLLMs in the domain of face attack detection.  

We introduce a new benchmark, namely SHIELD, to evaluate the ability of MLLMs on face spoofing and forgery detection. Specifically, we design multiple-choice and true/false questions to evaluate multimodal face data in these two face security tasks. For the face anti-spoofing task, we evaluate three different modalities (i.e., RGB, infrared, depth) under six types of presentation attacks (i.e., print attack, replay attack, rigid mask, paper mask, flexiblemask, fakehead). For the face forgery detection task, we evaluate GAN-based and diffusion-based data with both visual and acoustic modalities. Each question is subjected to both zero-shot and few-shot tests under standard and Chain of Thought (COT) settings. The overall performance of these Multimodal Large Language Models (MLLMs) is depicted in Figure~\ref{all_per}. The results demonstrate that MLLMs hold substantial promise in the realm of facial security.
Additionally, the marked differences in results due to changes in input modalities indicate that further research is needed to fully understand their performance limits. Additionally, we develop a novel Multi-Attribute Chain of Thought (MA-COT) paradigm for describing and judging various task-specific and task-irrelevant attributes of face images, which provides rich task-related knowledge for subtle spoof/forged clue mining. Our contributions are threefold:


\begin{itemize}
\item We introduce a new benchmark for evaluating the effectiveness of MLLMs in addressing various challenges within the domain of face security, including both face anti-spoofing and face forgery detection.
\item We propose a novel MA-COT paradigm, which provides rich task-related knowledge for MLLMs and improves the interpretability of the decision-making process for face spoofing and forgery detection.
\item Through extensive experiments, we find that 1) MLLMs (e.g., GPT4V and Gemini) have potential real/fake reasoning capability on unimodal and multimodal face spoofing and forgery detection; 2) the proposed MA-COT can improve the robustness and interpretability for face attack detection.
\end{itemize}

\section{SHIELD}


\begin{figure}[t]
  \centering
  \begin{center}
  \centerline{\includegraphics[width=0.8\linewidth]{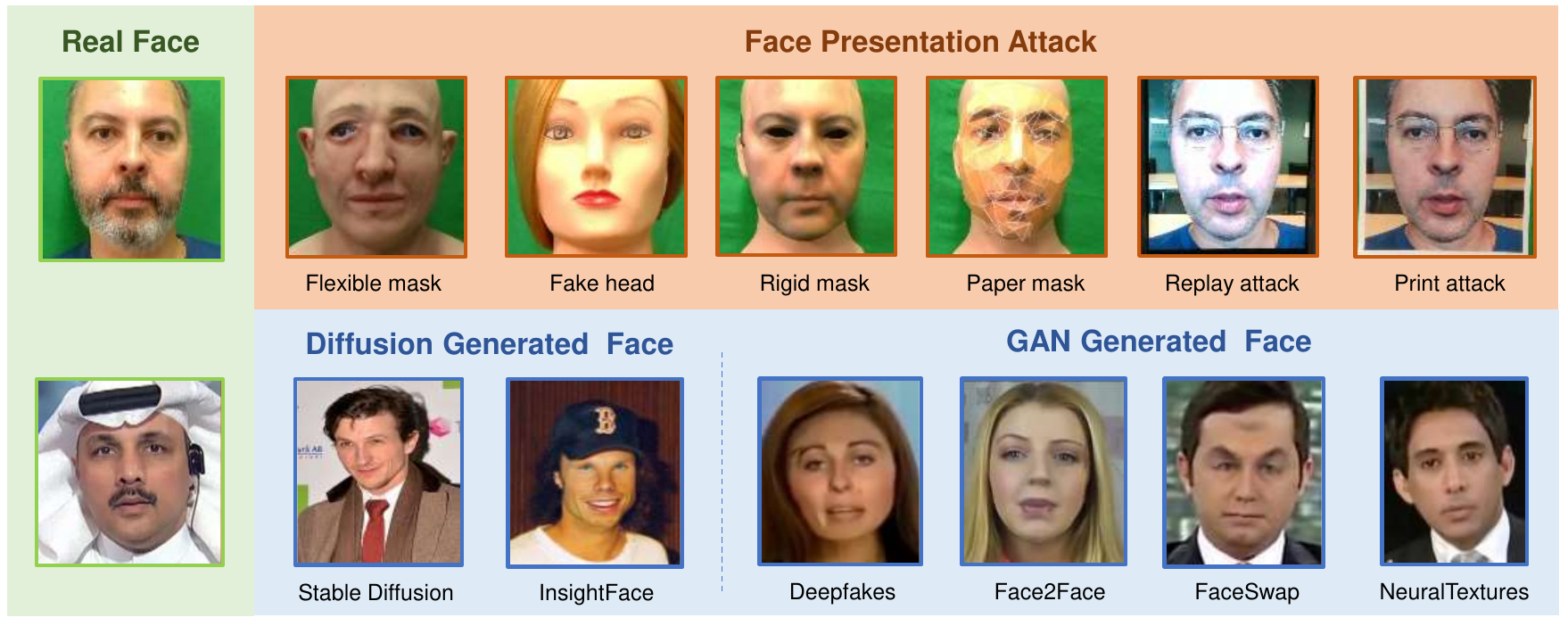}}
  \end{center}
  \vspace{-2.5em}
    \caption{Examples of our collected datasets
    }
    \label{attack}
\end{figure}

\subsection{Data Collection}
\textbf{Collection of Face Anti-Spoofing datasets}
For face anti-spoofing, we conducted experiments using the WMCA~\cite{george2019biometric} dataset, which includes a variety of presentation attacks and multiple modalities (e.g., RGB images, depth maps, infrared images, and thermal imaging). For each identity (ID), we selected sixtypes of attacks and the bonafide faces, along with their corresponding RGB images, depth maps, and infrared images for experiments. Typical samples are shown in the first row of Figure \ref{attack}.

\textbf{Collection of Face Forgery Detection datasets}
For face forgery detection, we evaluated MLLMs on the popular FaceForencics++ (FF++)~\cite{rossler2019faceforensics} dataset, which includes four types of forgery techniques (i.e., Deepfakes, Face2Face, FaceSwap, and Nulltextures). In addition, since AIGC is developing so quickly, the AI-generated images are becoming more and more realistic, and we also evaluated AIGC-based face data on the WildDeepfake dataset \cite{zi2020wilddeepfake}, which is produced by the Stable Diffusion, Inpainting, and InsightFace techniques. The images were chosen using "The Dawn of LMMs$:$ Preliminary Explorations with GPT4V" \cite{alayrac2022flamingo} as a guide, and our sample is simultaneously representative and diverse. Typical samples are shown in the second row of Figure \ref{attack}.


\subsection{Task Design}

\begin{figure}[t]
  \centering
  \begin{center}
  \centerline{\includegraphics[width=0.8\linewidth]{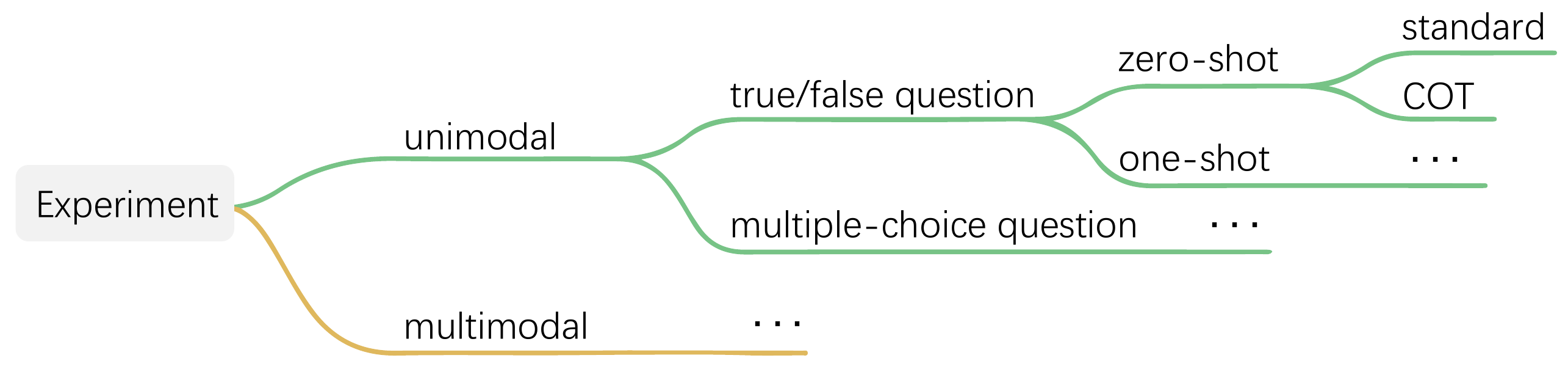}}
  \end{center}
  \vspace{-2.5em}
    \caption{Pipeline of task design. The ellipses indicate that the structures are consistent with the task design framework above.}
  \label{task design}
\end{figure}

\begin{figure}[t]
  \centering
  \begin{center}
  \centerline{\includegraphics[width=0.7\linewidth]{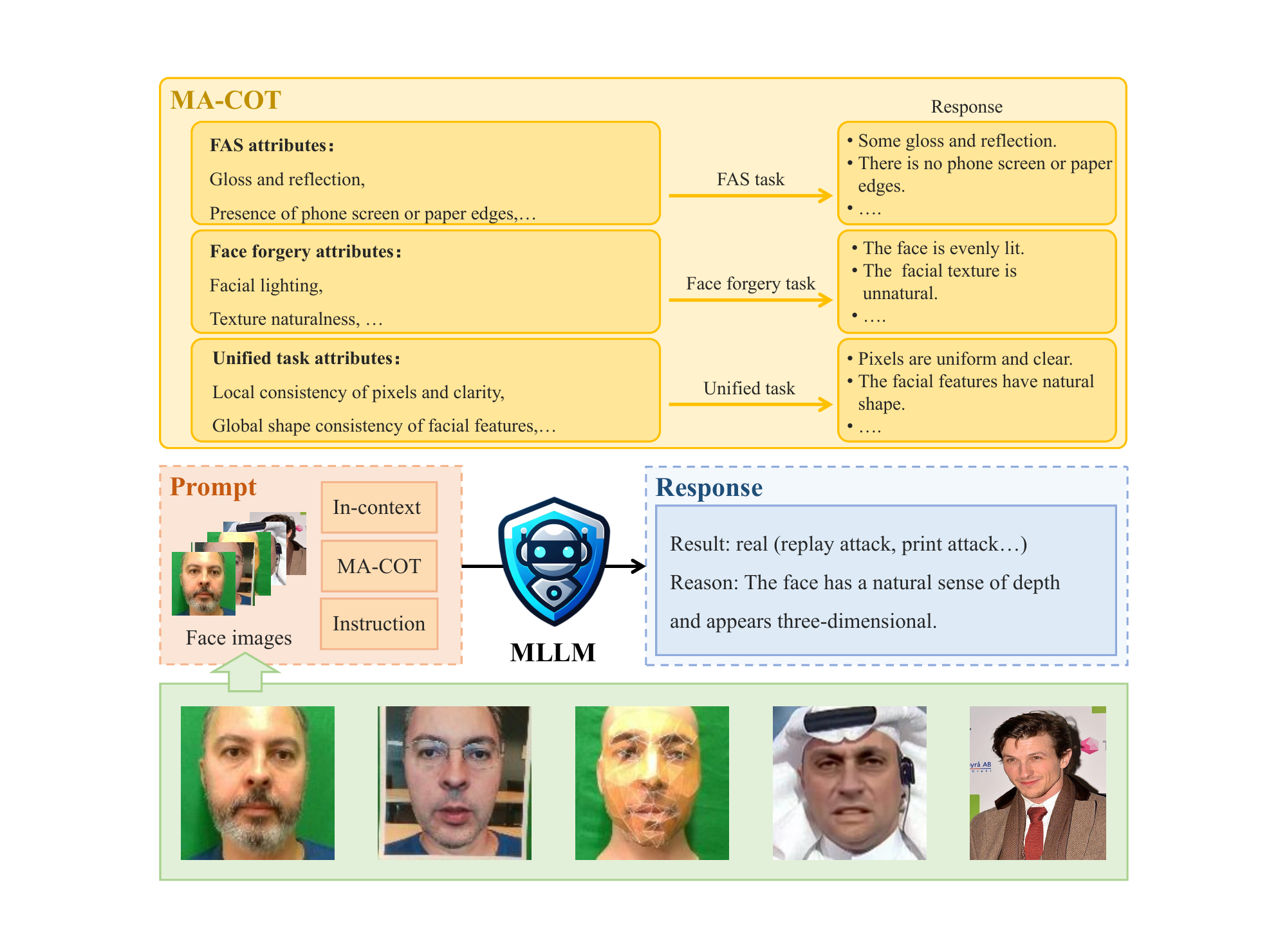}}
  \end{center}
  \vspace{-2.5em}
  \caption{The MA-COT process is designed to extract relevant key attributes for various tasks and input these attributes along with the face images under evaluation into MLLMs. This approach aims to guide the MLLM to analyze the images from multiple perspectives, thereby identifying potential clues of attacks and determining whether the images are of real faces. The illustration provides examples of key attribute extraction and its application scenarios in separate FAS, separate face forgery detection, and unified face spoof \& forgery detection.}
  \label{macot-introduction}
\end{figure}

As shown in Figure \ref{task design}, we designed two tasks to test the capabilities of MLLMs in the field of facial security$:$ true/false questions and multiple-choice questions. For each task, we conducted zero-shot tests and in-context few-shot tests. For each type of test, we experimented with both standard settings and COT settings. The overall pipeline of the experiment is shown in the lower half of Figure \ref{macot-introduction}, and specific content of task design can be seen in \S \ref{appendix:task design}.

\subsection{MA-COT}
COT technology is a recent development in the field of prompt learning, primarily aimed at enhancing the reasoning capabilities of MLLMs \cite{wei2022chain}. We introduce a novel COT paradigm, MA-COT. We draw inspiration from prior work in visual COT \cite{wu2023role} , where an image is initially described in detail followed by a judgment. Specifically, in the domains of face anti-spoofing and deepfake detection, we incorporate relevant prior knowledge. In our MA-COT approach, rather than describing the entire image, we focus on comprehensively considering multiple attributes of the image for description, such as shape, color, texture, etc., to provide in-depth analysis and judgment. The analysis results of each attribute are utilized to form a comprehensive judgment, thereby not only enhancing the accuracy and reliability of the overall decision-making process but also helping to circumvent instances where the model might refuse to provide an answer.

In MA-COT, we have a set of tasks: $\{ \text{task}_1, \text{task}_2, \ldots, \text{task}_n \}$. Each $task_i$ is associated with a set of attributes: $\{ \text{attr}_{i1}, \text{attr}_{i2}, \ldots, \text{attr}_{im} \}$.For each attribute, we obtain a description, forming a multi-attribute description set $Description_{attr}$. Based on this set, we generate the final answer:
\begin{equation}
P(\text{Answer} | \text{Image}, \text{Question}, \text{Description}_\text{attr})
\end{equation}

\begin{minipage}[b]{0.50\linewidth}
\begin{table}[H]
  \caption{Attributes set}
  \label{Attributes-set}
  \centering
  \scalebox{0.7}{
  \begin{tabular}{cc}
    \toprule
    Attributes                                  & Task                   \\
    \midrule
    local consistency of pixels and clarity     & Task shared            \\
    global shape consistency of facial features & Task shared            \\
    sense of depth and three-dimensionality     & Face Anti-Spoofing     \\
    gloss and reflection                        & Face Anti-Spoofing     \\
    presence of phone screen or paper edges     & Face Anti-Spoofing     \\
    perspective pose and texture naturalness    & Face Forgery Detection \\
    face facial lighting                        & Face Forgery Detection \\
    facial skin color                           & Face Forgery Detection \\
    eye-head and ear movements                  & Face Forgery Detection \\ 
    \bottomrule
  \end{tabular}
  }
\end{table}
\end{minipage}
\hfill
\begin{minipage}[b]{0.48\linewidth}
Our MA-COT is flexibly expandable in terms of both the set of tasks and the combination of attributes, making it applicable to other tasks as well. For the unified detection task, inspired by research in domain generalization and domain adaptation \cite{yichen2022out,wang2022domain,shao2019multi}, we have established both task-shared and task-specific attributes, as shown in the Table \ref{Attributes-set}. Shared attributes facilitate the differentiation between real faces and various attacks, while task-specific attributes enhance the detection capabilities. 
\end{minipage}

\section{Experiments}
\label{experiment}
\subsection{Eexperimental Setup}
We conducted tests on the most advanced MLLMs currently available: BLIP \cite{blip}, BLIP-2 \cite{li2023blip}, Intern \cite{Intern}, MiniGPT-4 \cite{zhu2023minigpt}, LLaVA \cite{llava}, QWen-VL \cite{bai2023qwen}, InstructBLIP \cite{dai2305instructblip}, mPLUG-owl \cite{ye2023mplug}, Gemini \cite{team2023gemini} and GPT4V \cite{gpt4}. For close source MLLMs, all tests were performed through API calls. We devised various testing scenarios to comprehensively assess the effectiveness and accuracy of these models in handling different facial security tasks. 

\subsection{Evaluation Metrics}
We choose the commonly used HTER (Half Total Error Rate) \cite{hter} in face anti-spoofing and the common ACC metric in classification problems as the measures for true/false questions. For multiple-choice questions, we have selected ACC as the metric. The results presented in \S \ref{res_fas} and \S \ref{res_df} include only aggregate average data of all attack types and do not detail the performance of each individual attack by the various MLLMs. More experiment results can be found in \S \ref{appendix:A} and more qualitative results in \S\ref{appendix:B}.

\section{Results on Face Anti-Spoofing}
\label{res_fas}

\subsection{Unimodal FAS Testing}
RGB-based unimodal face data are most common in face anti-spoofing tasks, primarily due to their ease of acquisition. In this study, we conducted a thorough testing of single-modality facial data, which included both true/false and multiple-choice questions. Inspired by the findings in \cite{wu2023role}, we discovered that the use of the COT technique in visual-language tasks (i.e., first describing the image, then making a decision) significantly enhances the model's performance. Accordingly, we carried out standard testing and COT testing for each question. More experiment results are in \S \ref{appendix:fas task}.

\textbf{True/False Questions}
For face anti-spoofing, we chose to use the phrases 'real face' to represent a real face and 'spoof face' to signify a spoof attack. We conducted three types of judgments in the decision-making process: independent inquiry and combined inquiry. 
We focus on the issue of unimodal authenticity determination. The crux of this problem lies in the in-depth analysis of one to two images to ascertain whether they are authentic human facial images.

  

Table \ref{fasTF} displays the comprehensive performance of various MLLMs on true/false questions under zero-shot, one-shot conditions. mPLUG-owl demonstrated high accuracy across all tests, yet, regrettably, the one-shot condition without COT did not enhance performance. While GPT4V did not excel on tests except for one-shot, it showed consistent results across all tasks. For most MLLMs, using ICL as a reference failed to improve their performance, whereas employing COT techniques could enhance their outcomes. For the performance of each MLLMs on various attacks in SHIELD, see the \S \ref{appendix:A} .

\begin{table}
  \caption{Performance metrics of various MLLMs on FAS true/false questions}
  \label{fasTF}
  \centering
  \resizebox{\linewidth}{!}{
  \begin{tabular}{ccccccccc}
    \toprule
    \multirow{2}{*}{Model} & \multicolumn{2}{c}{Zero-Shot} & \multicolumn{2}{c}{One-Shot} & \multicolumn{2}{c}{Zero-Shot(COT)} & \multicolumn{2}{c}{One-Shot(COT)} \\
    \cmidrule(r){2-9}
    & ACC(\%)$\uparrow$ & HTER(\%)$\downarrow$ & ACC(\%)$\uparrow$ & HTER(\%)$\downarrow$ & ACC(\%)$\uparrow$ & HTER(\%)$\downarrow$ & ACC(\%)$\uparrow$ & HTER(\%)$\downarrow$ \\ 
    \midrule
    BLIP \cite{blip} & 18.4 & 47.6 & 43.6 & 32.9 & 25.9 & 43.3 & 74.7 & 14.8 \\
    BLIP-2 \cite{li2023blip} & 14.3 & 50.0 & 9.0 & 68.5 & 14.3 & 50.0 & 0.4 & 98.5 \\
    Intern \cite{Intern} &57.6 & 24.8 & 14.3 & 50.0 & 56.4 & 25.4 & 17.7 & 48.0 \\
    MiniGPT-4 \cite{zhu2023minigpt} & 20.6 & 49.3 & 28.9 & 56.9 & 27.7 & 43.4 & 31.0 & 65.7 \\
    LLaVA \cite{llava} & 14.3 & 50.0 & 14.3 & 50.0 & 14.3 & 50.0 & 5.4 & 81.0 \\
    QWen-VL \cite{bai2023qwen} & 14.3 & 50.0 & 14.3 & 50.0 & 14.3 & 50.0 & 14.3 & 50.0 \\
    InstructBLIP \cite{dai2305instructblip} & 23.7 & 44.5 & 17.1 & 48.3 & 42.3 & 33.7 & 15.9 & 49.1 \\
    mPLUG-owl \cite{ye2023mplug} & 82.0 & 10.9 & 53.0 & 60.3 & 81.7 & 11.5 & 82.6 & 42.7 \\
    Gemini \cite{team2023gemini} & 73.4 & 15.5 & 62.0 & 36.3 & 77.0 & 14.7 & 13.9 & 90.3 \\
    GPT4V \cite{gpt4} & 53.2 & 27.7 & 66.8 & 28.8 & 68.7 & 18.1 & 50.3 & 36.8\\
    \bottomrule
  \end{tabular}
  }
\end{table}

\textbf{Multiple Choice Questions}
We also test face anti-spoofing in the form of an image selection task, with multiple images as inputs, and asking the model to identify which image is a real person or which one is a spoof attack. We conduct both zero-shot and few-shot tests, each including standard experiments and COT experiments.
We focus on an in-depth exploration of the unimodal multiple-choice problem. The fundamental task involves conducting a detailed analysis of four to five provided images, with the objective of selecting authentic human faces or identifying corresponding attack methods as required. This process entails not only cognitive analysis of the image content but also a comprehensive assessment of the image's authenticity, aiming to accurately distinguish between real human faces and various potential attack scenarios. 

Table \ref{fasMC} lists the accuracy of different MLLMs on FAS multiple-choice problems. The table separately addresses two types of questions: identifying real faces and attack faces, with each type tested under two conditions: zero-shot and one-shot. Finally, the average accuracy (AVG) across all conditions is provided. GPT4V achieved the highest average accuracy under all conditions, reaching 33.1\%. The Gemini performed best in recognizing attack under zero-shot conditions (25.6\%) and also demonstrated strong overall performance with an average accuracy of 23.3\%. ICL did not significantly improve the performance, indicating that the ICL capability of MLLMs for images still needs enhancement.

\begin{table}
  \caption{Accuracy of various MLLMs on FAS multiple-choice questions}
  \label{fasMC}
  \centering
  \resizebox{0.9\linewidth}{!}{
  \begin{tabular}{ccccccccccc}
    \toprule
    \multirow{3}{*}{Model} & \multicolumn{10}{c}{ACC(\%)$\uparrow$} \\ 
    
    \cmidrule(r){2-11}
    & \multicolumn{2}{c}{Zero-Shot(real)} & \multicolumn{2}{c}{One-Shot(real)} & \multicolumn{2}{c}{Zero-Shot(attack)} & \multicolumn{2}{c}{One-Shot(attack)} & \multicolumn{2}{c}{AVG} \\
    \cmidrule(r){2-11}
    & - & COT & - & COT & - & COT & - & COT & - & COT \\
    \midrule
    BLIP \cite{blip} & 24.0 & 24.0 & 5.0 & 30.0 & 25.0 & 32.0& 2.0 & 18.0 & 14.0 & 26.0 \\
    BLIP-2 \cite{li2023blip} & 10.0 & 0.0 & 20.0 & 0.0 & 11.0 & 0.0 & 4.0 & 0.0 & 11.3 & 0.0 \\
    Intern \cite{Intern} & 21.0 & 3.0 & 20.0 & 5.0 & 24.0 & 0.0 & 5.0 & 0.0 & 17.5 & 2.0 \\
    MiniGPT-4 \cite{zhu2023minigpt} & 2.0 & 0.0 & 1.0 & 1.0 & 2.0 & 0.0 & 1.0 & 0.0 & 1.5 & 0.3 \\
    LLaVA \cite{llava} & 22.0 & 4.0 & 26.0 & 18.0 & 9.0 & 1.0 & 26.0 & 8.0 & 20.8 & 7.8 \\
    QWen-VL \cite{bai2023qwen} & 3.0 & 0.0 & 26.0 & 0.0 & 9.0 & 0.0 & 27.0 & 0.0 & 16.3 & 0.0 \\
    InstructBLIP \cite{dai2305instructblip} & 21.0 & 24.0 & 0.0 & 0.0 & 6.0 & 7.0 & 8.0 & 1.0 & 8.8 & 8.0 \\
    mPLUG-owl \cite{ye2023mplug} & 23.0 & 5.0 & 26.0 & 1.0 & 5.0 & 1.0 & 27.0 & 0.0 & 20.3 & 1.8 \\
    Gemini \cite{team2023gemini} & 30.0 & 24.2 & 15.0 & 18.0 & 25.6 & 24.0 & 22.2 & 7.1 & 23.3 & 18.8 \\
    GPT4V \cite{gpt4} & 42.1 & 60.9 & 20.4 & 54.2 & 44.3 & 42.9 & 25.0 & 40.6 & 33.1 & 51.3\\
    \bottomrule
  \end{tabular}
  }
\end{table}

\subsection{Multimodal FAS Testing}
Multimodal face data can provide a richer set of information for capturing spoof clues. We input images from three different modalities (i.e., visible RGB, infrared, and depth) into into the MLLMs to evaluate their performance, as shown in Figure \ref{fas_multi_judgment_zeroshot_01} to Figure \ref{fas_multi_choice_attack_oneshot_01}. The questions asked in the test were kept consistent with those used in the unimodal experiments to ensure uniformity and comparability of the results.

\textbf{True/False Questions}
Table \ref{multiFAS} lists the accuracy of GPT4V and Gemini in judging FAS multimodal tasks, selecting these models due to their superior performance in these complex tasks. The test, employing a limited number of samples, required the MLLMs to discern whether the subject was a real face, under two conditions: zero-shot and one-shot. GPT4V demonstrated commendable outcomes under each condition, indicating its inherent capability to directly extract information from images, with the one-shot condition generally outperforming the zero-shot. In contrast, this task proved more challenging for Gemini, maintaining an accuracy rate around 50\%. ICL did not enhance its performance, highlighting Gemini's limitations in handling more complex tasks.

\begin{table}[H]
  \caption{Acuracy of various GPT4V and Gemini on FAS Multimodal true/false questions}
  \label{multiFAS}
  \centering
  \resizebox{0.7\linewidth}{!}{
  \begin{tabular}{ccccccccc}
    \toprule
    \multicolumn{1}{c}{\multirow{2}{*}{Model}} & \multicolumn{5}{c}{ACC(\%)$\uparrow$} \\ 
    \cmidrule(r){2-6}
    \multicolumn{1}{c}{} & \multicolumn{1}{c}{Zero-Shot} & \multicolumn{1}{c}{One-Shot} & \multicolumn{1}{c}{Zero-Shot(COT)} & \multicolumn{1}{c}{One-Shot(COT)} & \multicolumn{1}{c}{AVG} \\ 
    \midrule
    GPT4V \cite{gpt4} & 87.5 & 100.0 & 60.0 & 100.0 & 90.6 \\
    Gemini \cite{team2023gemini} & 50.0 & 50.0 & 50.0 & 50.0 & 50.0\\
    \bottomrule
  \end{tabular}
  }
\end{table}

\section{Results on Face Forgery Detection}
\label{res_df}

\subsection{Unimodal Face Forgery Detection Testing}

\textbf{True/False Questions}
In this subsection, our objective is to evaluate the model's ability to discern forged images by requesting fixed responses regarding their authenticity.
More experiment results are in \S \ref{appendix:face forgery}.

Table \ref{tab5} presents quantitative results, we observe the zero-shot detection performance of common large models such as BLIP2 and LLaVA in face anti-spoofing. Due to the absence of domain knowledge fine-tuning, their accuracy is notably low; for instance, BLIP2 achieves only 14.3\% accuracy. While models like GPT4V and Gemini perform better than others, their accuracy remains relatively low. Interestingly, the one-shot accuracy does not improve significantly across these models. Conversely, our proposed method, Cot, demonstrates more effective performance, with accuracy increasing from 22.4\% to 26.0\% in the one-shot setting.


\begin{table}[H]
  \caption{Performance metrics of various MLLMs on Face Forgery true/false question}
  \label{tab5}
  \centering
  \resizebox{0.9\linewidth}{!}{
  \begin{tabular}{ccccccccc}
    \toprule
    \multirow{2}{*}{Model} & \multicolumn{2}{c}{Zero-Shot} & \multicolumn{2}{c}{One-Shot} & \multicolumn{2}{c}{Zero-Shot(COT)} & \multicolumn{2}{c}{One-Shot(COT)} \\
    \cmidrule(r){2-9}
    & ACC(\%)$\uparrow$ & HTER(\%)$\downarrow$ & ACC(\%)$\uparrow$ & HTER(\%)$\downarrow$ & ACC(\%)$\uparrow$ & HTER(\%)$\downarrow$ & ACC(\%)$\uparrow$ & HTER(\%)$\downarrow$ \\ 
    \midrule
    BLIP \cite{blip} & 17.9 & 51.3 & 49.1 & 29.7 & 31.0 & 49.4 & 62.6 & 23.1 \\
    BLIP-2 \cite{li2023blip} & 14.3 & 50.0 & 10.1 & 64.5 & 14.3 & 50.0 & 1.7 & 94.0 \\
    Intern \cite{Intern} &14.6 & 49.8 & 14.3 & 50.0 & 14.3 & 50.0 & 14.3 & 50.0 \\
    MiniGPT-4 \cite{zhu2023minigpt} & 17.7 & 48.0 & 32.1 & 57.1 & 18.1 & 48.2 & 36.1 & 49.8 \\
    LLaVA \cite{llava} & 14.3 & 50.0 & 14.3 & 50.0 & 14.3 & 50.0 & 13.1 & 54.0 \\
    QWen-VL \cite{bai2023qwen} & 14.3 & 50.0 & 14.3 & 50.0 & 14.3 & 50.0 & 14.3 & 50.0 \\
    InstructBLIP \cite{dai2305instructblip} & 18.6 & 47.5 & 14.6 & 49.8 & 23.9 & 44.4 & 16.7 & 48.6 \\
    mPLUG-owl \cite{ye2023mplug} & 21.9 & 45.6 & 12.7 & 60.9 & 20.9 & 46.2 & 20.9 & 50.8 \\
    Gemini \cite{team2023gemini} & 36.8 & 38.5 & 27.0 & 52.5 & 48.6 & 35.9 & 14.0 & 83.1 \\
    GPT4V \cite{gpt4_exp} & 26.0 & 43.6 & 22.4 & 45.1 & 28.8 & 44.0 & 26.0 & 44.8\\
    \bottomrule
  \end{tabular}
  }
\end{table}

\textbf{Multiple Choice Questions}
In this subsection, we use a form of image selection to test the capability of face forgery detection. We ask the GPT4V and Gemini models to recognize which image is real and which image is a forgery. Similar to the previous subsection, our tests cover common forgery patterns such as Deepfakes, Face2Face, FaceSwap, Nulltextures, and Stable Diffusion generated images. Each test includes both standard and chain-of-thinking experiments. We wanted the model to be able to analyze the four to five images provided in detail in order to pick out real images. Subsequently, to understand the model's capabilities in more detail, we also tested the model's recognition of the generation method.
Quantitative experiments in Table \ref{deepfakeMC} consistently demonstrate that models lacking fine-tuning exhibit poor performance, posing challenges for accurately identifying forged faces and understanding the categories of forgery. 

\begin{table}[H]
  \caption{Accuracy of various MLLMs on Face Forgery multiple-choice questions}
  \label{deepfakeMC}
  \centering
  \resizebox{0.8\linewidth}{!}{
  \begin{tabular}{ccccccccccc}
    \toprule
    \multirow{3}{*}{Model} & \multicolumn{10}{c}{ACC(\%)$\uparrow$} \\ 
    \cmidrule(r){2-11}
    & \multicolumn{2}{c}{Zero-Shot(real)} & \multicolumn{2}{c}{One-Shot(real)} & \multicolumn{2}{c}{Zero-Shot(attack)} & \multicolumn{2}{c}{One-Shot(attack)} & \multicolumn{2}{c}{AVG} \\ 
    \cmidrule(r){2-11}
    & - & COT & - & COT & - & COT & - & COT & - & COT \\ 
    \midrule
    BLIP \cite{blip} & 22.0 & 24.0 & 21.0 & 23.0 & 22.0 & 18.0 & 11.0 & 10.0 & 19.0 & 18.8 \\
    BLIP-2 \cite{li2023blip} & 0.0 & 0.0 & 6.0 & 0.0 & 7.0 & 0.0 & 0.0 & 0.0 & 3.3 & 0.0 \\
    Intern \cite{Intern} &16.0 & 5.0 & 18.0 & 10.0 & 10.0 & 0.0 & 14.0 & 1.0 & 14.5 & 4.0 \\
    MiniGPT-4 \cite{zhu2023minigpt} & 2.0 & 2.0 & 2.0 & 0.0 & 0.0 & 0.0 & 0.0 & 0.0 & 1.0 & 0.5 \\
    LLaVA \cite{llava} & 18.0 & 1.0 & 18.0 & 7.0 & 7.0 & 0.0 & 23.0 & 9.0 & 16.5 & 4.3 \\
    QWen-VL \cite{bai2023qwen} & 6.0 & 0.0 & 19.0 & 0.0 & 1.0 & 0.0 & 24.0 & 0.0 & 12.5 & 0.0 \\
    InstructBLIP \cite{dai2305instructblip} & 6.0 & 15.0 & 0.0 & 0.0 & 0.0 & 0.0 & 12.0 & 3.0 & 4.5 & 4.5 \\
    mPLUG-owl \cite{ye2023mplug} & 11.0 & 0.0 & 18.0 & 0.0 & 8.0 & 0.0 & 13.0 & 0.0 & 12.5 & 0.0 \\
    Gemini \cite{team2023gemini} & 14.0 & 17.0 & 16.0 & 19.0 & 20.0 & 4.0 & 0.0 & 17.2 & 16.3 & 13.7 \\
    GPT4V \cite{gpt4} & 6.9 & 23.6 & 12.0 & 20.7 & 0.0 & 0.0 & 14.8 & 18.6 & 8.9 & 17.0\\
    \bottomrule
  \end{tabular}
  }
\end{table}

\subsection{Multimodal Face Forgery Detection Testing}
The potential of multi-modal information in the field of face forgery detection has garnered widespread attention from researchers, typically serving as auxiliary data to enhance model accuracy. In this paper, we utilized DCT transformation to acquire frequency domain modalities and employed the official Maximum Magnitude Frequency Componentization (MMFC) for voice extraction to obtain voice spectrograms. Additionally, we combined the original images with these two modalities to form three different input modalities, as shown in Figure \ref{30} to Figure \ref{31}. During the testing phase, we considered four common forgery modes: Deepfakes, Face2Face, FaceSwap, and Nulltextures, and used representative Zero/One-Shot prompts to evaluate the two models. The results indicate that the models are more prone to triggering security alerts and exhibit some illogical responses. 

\textbf{True/False Questions}
Table 6 displays the performance of different models when utilizing three modalities. Compared to using only images, there is a notable decline in performance. For example,GPT4V's accuracy decreases from 26.0\% to 6.9\%. This could be due to the incapacity of large models to capture subtle frequency domain differences. The average accuracy of each model is below 20.0\%. 

\begin{table}[!htbp]
  \caption{Accuracy of GPT4V and Gemini on Face Forgery multimodal true/false questions}
  \label{deepfakeTF}
  \centering
  \resizebox{0.7\linewidth}{!}{
  \begin{tabular}{ccccccccc}
    \toprule
    \multicolumn{1}{c}{\multirow{2}{*}{Model}} & \multicolumn{5}{c}{ACC(\%)$\uparrow$} \\ 
    \cmidrule(r){2-6}
    \multicolumn{1}{c}{} & \multicolumn{1}{c}{Zero-Shot} & \multicolumn{1}{c}{One-Shot} & \multicolumn{1}{c}{Zero-Shot(COT)} & \multicolumn{1}{c}{One-Shot(COT)} & \multicolumn{1}{c}{AVG} \\
    \midrule
    GPT4V \cite{gpt4} & 44.4 & 66.7 & 12.5 & 66.7 & 68.0 \\
    Gemini \cite{team2023gemini} & 50.0 & 30.0 & 50.0 & 70.0 & 50.0\\
    \bottomrule
  \end{tabular}
  }
\end{table}

\section{Results on Unified tasks}
In this part of the study, we combined the FAS task from \S \ref{res_fas} with the face forgery detection task from \S \ref{res_df} to assess the performance of mainstream MLLMs in a unified detection task against both physical spoof and digital forgery attacks \cite{yu2024benchmarking}. More experiment results are in \S \ref{appendix:unified task}.

\begin{figure}[htbp]
\centering
\begin{minipage}[b]{0.44\linewidth}
  \begin{table}[H]
  \caption{Accuracy of various MLLMs on UnifiedTask true/false questions few-shot}
  \label{unifiedFewshot}
  \centering
  \scalebox{0.8}{
  \begin{tabular}{ccc}
    \toprule
    \multirow{2}{*}{Model} & \multicolumn{2}{c}{ACC(\%)$\uparrow$} \\ 
    \cmidrule(r){2-3}
    & Few-Shot & Few-Shot(COT) \\ 
    \midrule
    BLIP \cite{blip} & 34.0 & 41.0 \\
    BLIP-2 \cite{li2023blip} & 34.0 & 9.0 \\
    Intern \cite{Intern}  & 33.0 & 30.0 \\
    MiniGPT-4 \cite{zhu2023minigpt} & 18.0 & 24.0 \\
    LLaVA \cite{llava} &  33.0 & 41.0 \\
    QWen-VL \cite{bai2023qwen} &20.0 & 21.0 \\
    InstructBLIP \cite{dai2305instructblip} & 44.0 & 54.0 \\
    mPLUG-owl \cite{ye2023mplug} & 34.0 & 35.0 \\
    Gemini \cite{team2023gemini} & 41.0 & 45.9 \\
    GPT4V \cite{gpt4} & 25.5 & 24.2\\
    \bottomrule
  \end{tabular}
  }
\end{table}
\end{minipage}
\hfill
\begin{minipage}[b]{0.55\linewidth}
Table \ref{unifiedFewshot} enumerates the accuracy of various MLLMs in few-shot joint task assessments. The data indicate that the accuracy of models such as BLIP, Gemini, MiniGPT-4, LLava, and Instruct BLIP improved following the incorporation of COT. Conversely, GPT4V and Intern experienced a slight decrease, while BLIP-2 underwent a precipitous decline. These observations suggest that the introduction of COT can generally assist MLLMs in making correct judgments, although it may also lead to an increased focus on noise, resulting in a slight deterioration of results. The particularly poor performance of BLIP-2 might be attributed to the induced illusions from the integrated COT. The performance of MLLMs in few-shot joint tasks underscores the significant challenges they face in solving combined tasks effectively.
\end{minipage}
\end{figure}
Table \ref{unifiedMC} presents the ACC of various MLLMs on the Unified Task multiple-choice tests. The task is divided into two main categories: recognizing real faces and various malicious attacks. GPT4V has the highest average accuracy across all conditions at 29.2\%, indicating the best overall performance on this unified task. In the zero-shot condition for recognizing real faces, GPT4V also leads with an accuracy of 27.1\%, closely followed by Gemini at 24.0\%. In the one-shot condition for real image recognition, GPT4V again achieves the highest accuracy at 28.5\%, with BLIP-2 trailing at 20.0\% accuracy. For attack recognition, Gemini tops the zero-shot condition with an accuracy of 27.8\%, while GPT4V leads in the one-shot condition at 33.9\%. Models like BLIP-2, MiniGPT-4, and QWen-VL face significant challenges, especially in the zero-shot condition for real images, where both BLIP-2 and MiniGPT-4 show 0.0\% accuracy. The generally low average performance across all models suggests that the task is challenging.


\begin{table}[H]
  \caption{Accuracy of various MLLMs on UnifiedTask multiple-choice questions}
  \label{unifiedMC}
  \centering
  \resizebox{0.8\linewidth}{!}{
  \begin{tabular}{ccccccccccc}
    \toprule
    \multirow{3}{*}{Model} & \multicolumn{10}{c}{ACC(\%)$\uparrow$} \\ 
    \cmidrule(r){2-11}
    & \multicolumn{2}{c}{Zero-Shot(real)} & \multicolumn{2}{c}{One-Shot(real)} & \multicolumn{2}{c}{Zero-Shot(attack)} & \multicolumn{2}{c}{One-Shot(attack)} & \multicolumn{2}{c}{AVG} \\ 
    \cmidrule(r){2-11}
    & - & COT & - & COT & - & COT & - & COT & - & COT \\ 
    \midrule
    BLIP \cite{blip} & 26.0 & 25.0 & 8.0 & 24.0 & 24.0 & 24.0 & 5.0 & 8.0& 15.8 & 20.3 \\
    BLIP-2 \cite{li2023blip} & 0.0 & 0.0 & 20.0 & 0.0 & 2.0 & 0.0 & 2.0 & 0.0 & 6.0 & 0.0 \\
    Intern \cite{Intern} &15.0 & 2.0 & 24.0 & 14.0 & 7.0 & 0.0 & 6.0 & 0.0 & 13.0 & 4.0 \\
    MiniGPT-4 \cite{zhu2023minigpt} & 0.0 & 0.0 & 1.0 & 0.0 & 2.0 & 0.0 & 1.0 & 0.0 & 1.0 & 0.0 \\
    LLaVA \cite{llava} & 13.0 & 0.0 & 27.0 & 14.0 & 6.0 & 1.0 & 21.0 & 4.0 & 16.8 & 4.8 \\
    QWen-VL \cite{bai2023qwen} & 1.0 & 0.0 & 22.0 & 0.0 & 1.0 & 0.0 & 22.0 & 0.0 & 11.5 & 0.0 \\
    InstructBLIP \cite{dai2305instructblip} & 12.0 & 27.0 & 3.0 & 0.0 & 2.0 & 1.0 & 5.0 & 2.0 & 5.5 & 7.5 \\
    mPLUG-owl \cite{ye2023mplug} & 12.0 & 0.0 & 27.0 & 1.0 & 5.0 & 0.0 & 21.0 & 0.0 & 16.3 & 0.3 \\
    Gemini \cite{team2023gemini} & 24.0 & 22.0 & 14.0 & 23.0 & 27.8 & 14.0 & 11.9 & 15.0 & 20.2 & 18.7 \\
    GPT4V \cite{gpt4} & 27.2 & 27.4 & 28.6 & 34.0 & 29.3 & 15.4 & 34.0 & 26.8 & 29.2 & 26.8\\
    \bottomrule
  \end{tabular}
  }
\end{table}

\section{Results of MA-COT}
\label{res_macot}
In this section, we tested the application effectiveness of the MA-COT method previously mentioned, across the FAS task, the Face Forgery Detection task, and unified task, with a subset of the results visualized. By employing the MA-COT method, we guide the MLLMs to focus on key features related to presentation attacks using prior knowledge, thereby more efficiently completing sub-tasks without the direct introduction of such knowledge. Moreover, since these key features can be easily added as plug-and-play components, the MA-COT method also facilitates the seamless integration of the FAS task and the Face Forgery Detection task into a unified task framework.More experiment results are in \S \ref{appendix:macot}.

\begin{figure}[htbp]
\centering
\begin{minipage}[c]{0.48\textwidth}
\centering
\includegraphics[width=0.9\textwidth]{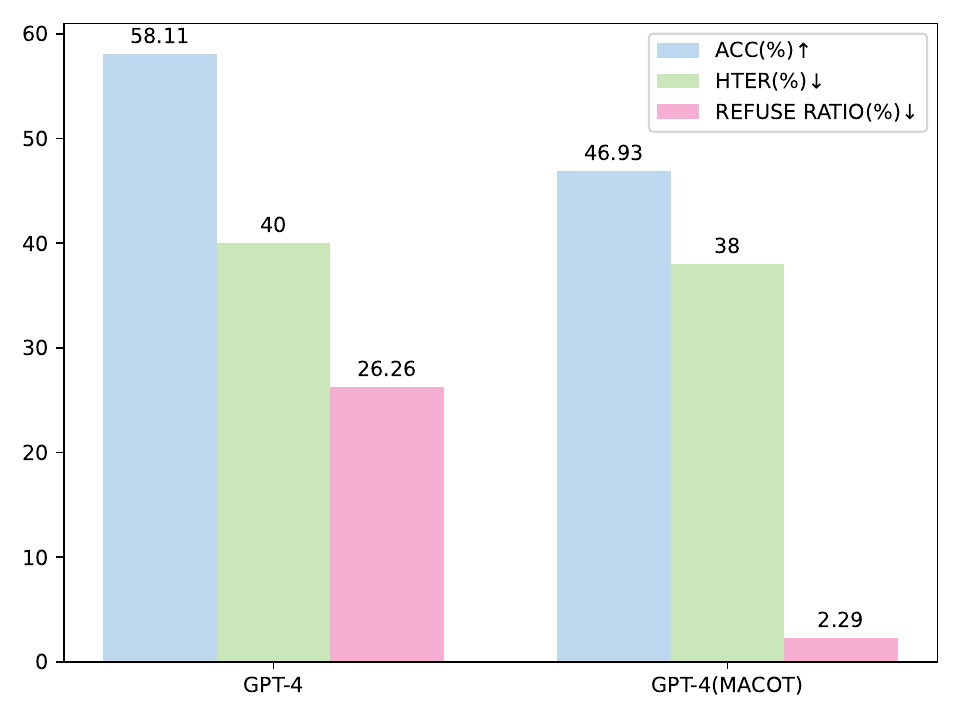}
\end{minipage}
\hspace{0.02\textwidth}
\begin{minipage}[c]{0.48\textwidth}
\centering
\includegraphics[width=0.9\textwidth]{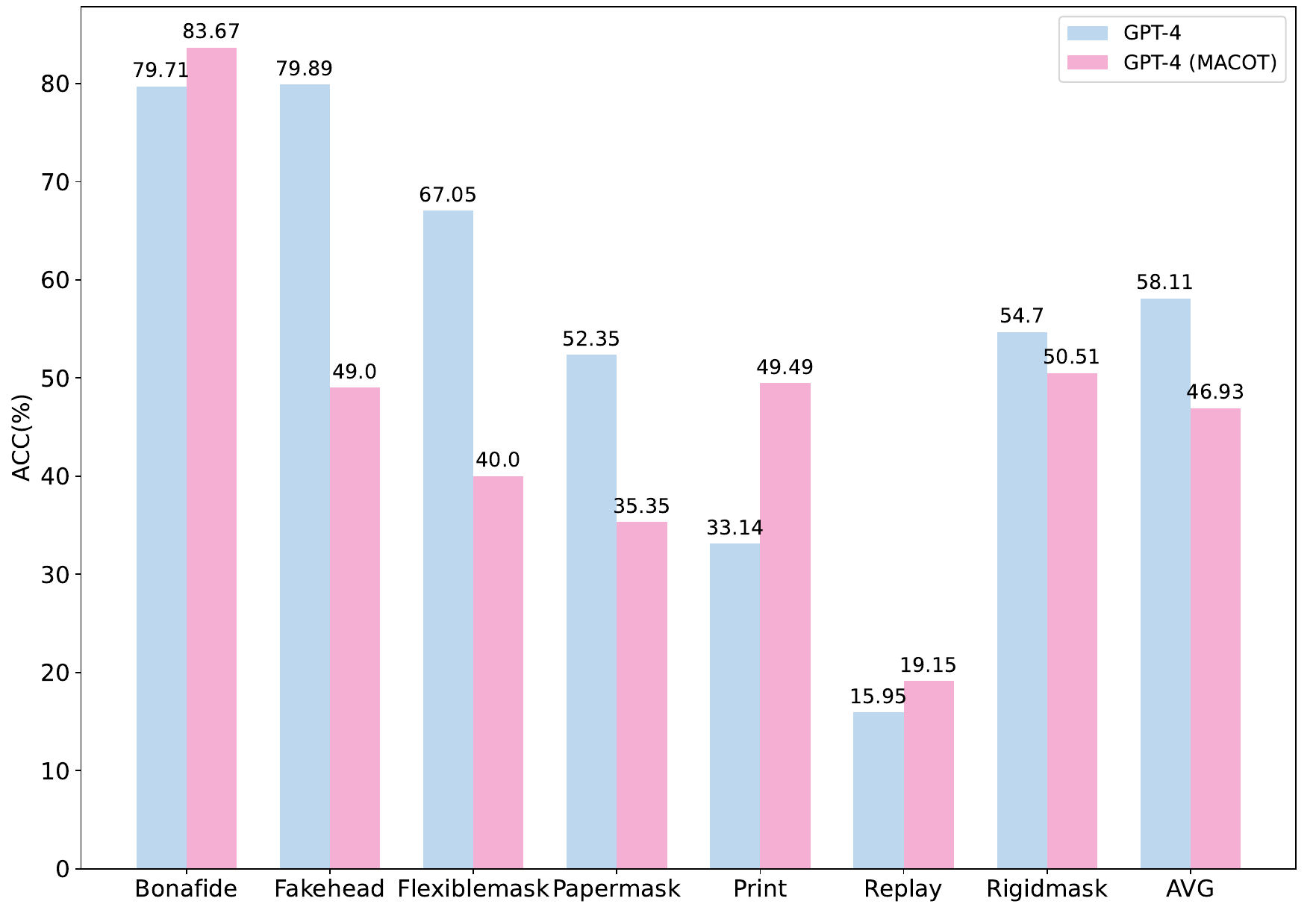}
\end{minipage}\\[3mm]
\begin{minipage}[t]{0.48\textwidth}
\centering
\caption{Comparative results of the GPT4V on the FAS zero-shot task, with or without the MA-COT.}
\label{macot_compare}
\end{minipage}
\hspace{0.02\textwidth}
\begin{minipage}[t]{0.48\textwidth}
\centering
\caption{Results of GPT4V on true/false zero-shot FAS tasks, showing performance across different attacks with or without MA-COT.}
\label{macot_detail}
\end{minipage}
\end{figure}

Figure \ref{macot_compare} presents a comparison of the average metrics for GPT4V in true/false question with or without the use of MA-COT. It is observed that the average accuracy slightly decreases after employing MA-COT, while the HTER remains relatively stable. Notably, the rejection rate of GPT4V significantly declines, demonstrating that MA-COT can effectively reduce the occurrence of refusal to answer. Figure \ref{macot_detail} shows the result of each attack types with or without the use of MA-COT on GPT4V. The use of MA-COT notably improves performance in areas where GPT4V initially struggled, such as print and replay attacks, though it may reduce performance in areas where it previously excelled. This suggests that the MA-COT strategy is effective, but the key attributes considered by it still need further exploration.

\section{Findings and Discussion}
\textbf{More Comprehensive Dataset} Current face security datasets mainly focus on category-level annotations and lack pixel-level details and extensive image descriptions. Developing a dataset that includes both categories and explanatory judgment reasons is crucial for training more interpretable and generalized models, enhancing security solutions in complex settings.

\textbf{More Powerful Models} Specialized multimodal large models using robust foundational models can be developed for face security. These models handle various inputs and tasks like description, reasoning, and judgment, improving accuracy and reducing resource use by incorporating a face security agent that manages diverse detection tools.

\textbf{Cross-task Collaboration} Collaborating on similar tasks like face anti-spoofing and deepfake detection utilizes complementary knowledge for better results. Expanding such collaboration to other fields like industrial anomaly detection and medical image processing could significantly boost accuracy and efficiency.
\section{Limitations and broader impacts}
Our experiments show that MLLM has great potential in the field of face security, but for the field of face security the increase in anti-spoofing capability may lead to the misuse of face recognition technology, which brings potential privacy issues and other security concerns. These issues require further research and consideration in future work and applications.

\bibliographystyle{plainnat}
\bibliography{ref}

\clearpage

\appendix



\section{Related Work}

\label{related_work}

\subsection{Face Anti-Spoofing}

In the early stage, face anti-spoofing systems relied on traditional techniques. These methods mainly focused on handcrafted features~\cite{boulkenafet2015face,Boulkenafet2017Face,Komulainen2014Context,Patel2016Secure} and heuristics designed to detect presentation attacks. They were effective against basic spoofing attempts but often failed to generalize well against more sophisticated attacks. With the advent of deep learning, there was a significant shift in FAS methods. Deep learning-based techniques~\cite{qin2019learning,Liu2018Learning,yang2019face,Atoum2018Face,Gan20173D,george2019deep,yu2020fas2}, particularly those employing convolutional neural networks (CNNs) with advanced learning strategies (e.g., meta-learning~\cite{qin2021meta}, disentangled learning~\cite{wang2020cross}, and few-shot learning~\cite{qin2019learning}), began to dominate the field. These methods leveraged large-scale datasets and powerful computational resources to learn more complex and subtle features of genuine and spoofed faces. This shift marked a substantial improvement over traditional methods in both detection accuracy and generalization capabilities.
Recently, FAS methods have started to be integrated into larger and multimodal models with vision transformers (ViTs)~\cite{dosovitskiy2020image}. On the one hand, ViT is adopted in the spatial domain~\cite{ming2022vitranspad,george2020effectiveness,wang2022face,liuma} to explore live/spoof relations among local patches. On the other hand, global features temporal abnormity~\cite{wang2022learning} or physiological periodicity~\cite{yu2021transrppg,yu2024benchmarking} are extracted by applying ViT in the temporal domain. Despite convincing performance via unimodal and multimodal deep learning methods based on CNNs and ViTs, there is still no work to explore the fundamental MLLMs for generalized FAS.

\subsection{Face Forgery Detection}
Previous studies have predominantly approached this task as a binary classification problem~\cite{cao2022end,chen2021local,haliassos2021lips}. CNNs are commonly employed to extract high-dimensional features from input face images, subsequently classifying them as either real or fake. While these models exhibit commendable performance on in-domain datasets due to their utilization of facial features, their performance on cross-domain datasets remains inadequate. To overcome this limitation, recent studies have explored alternative approaches, including the utilization of low-frequency information in frequency domain analysis. For instance, noise-based approaches \cite{gu2022exploiting,mat}, frequency domain analysis \cite{qian2020thinking,freq1_icml,twobranch}, and local relation learning methods \cite{chen2021local,facexray} have been investigated. These methods show promising results by incorporating additional information beyond traditional frequency domain or convolutional network-based methods. However, it is essential to note that real-world forgery scenarios are characterized by diverse patterns and unknown environmental conditions. The original techniques based on frequency domain information or convolutional networks tend to overfit existing forgery techniques in the training set, leading to substantial performance degradation for these methods. Recently, vision foundation models (e.g., Segment Anything Models \cite{kirillov2023segment}) have successfully been introduced in face forgery detection\cite{lai2023detect}, demonstrating their strong attck localization capacities. Facing the continuous updating face forgery attacks and considering the powerful zero-shot generalization capacity of MLLMs, it is necessary to explore whether MLLMs can robustly detect face forgery attacks.

\subsection{Multimodal Large Language Model}

The development of MLLMs in recent years has been marked by significant contributions from various organizations. \citet{alayrac2022flamingo} introduced the Flamingo model, a significant advancement in processing interleaved visual data and text, focusing on generating free-form text output. Following this, \citet{li2023blip} developed BLIP-2, a model characterized by its resource-efficient framework and the innovative Q-Former, which notably leverages frozen LLMs for efficient image-to-text generation. In a similar vein, \citet{dai2305instructblip} further refined this approach with InstructBLIP, a model trained on the BLIP-2 framework, specifically enhancing the capability of instruction-aware visual feature extraction. Developed by OpenAI, GPT4V \cite{achiam2023gpt}  itself represents a leap in MLLMs, offering versatility in generating coherent and context-aware text. It is applicable in standalone text-based applications as well as multimodal settings. Its superior performance establishes a dominant position in terms of versatility and general applicability. \citet{zhu2023minigpt} introduced MiniGPT-4, a streamlined approach that aligns a pre-trained vision encoder with LLMs by training a single linear layer, effectively replicating the capabilities of GPT4V. Expanding the linguistic versatility, \citet{bai2023qwen} presented Qwen-VL, a multi-lingual MLLM supporting both English and Chinese, with the ability to process multiple images during training. In the domain of multimodal data scarcity, \citet{liu2023visual} pioneered with LLaVA, introducing a novel open-source multimodal instruction-following dataset alongside the LLaVA-Bench benchmark. Recently, Google newly introduced Gemini \cite{team2023gemini} represents a significant advancement in the inherent multimodal nature of artificial intelligence systems. It processes text and various audiovisual inputs, generating outputs in multiple formats, thereby demonstrating the efficiency of integrated multimodal interactions. Gemini also possesses high generalization capabilities, posing a significant challenge to GPT4V \cite{yang2023dawn}. Despite remarkable problem-solving capabilities in various vision fields (e.g., generic object recognition and grounding), whether MLLMs are sensitive to subtle visual spoof/forged clues and how they perform in the domain of face attack detection is still unexplored.  

\subsection{Existing MLLMs Benchmark}
In recent years, benchmarks for MLLMs have been developed to evaluate their performance across various tasks. A notable benchmark in this regard is the Multimodal Model Evaluation (MME) \cite{fu2023mme}. The MME benchmark focuses on a wide range of tasks to assess the capabilities of different MLLMs, including both perception and cognition tasks, encompassing areas like object recognition, commonsense reasoning, numerical calculation, text translation, and code reasoning. This benchmark evaluates several advanced MLLM models, such as BLIP-2 \cite{li2023blip}, MiniGPT-4 \cite{zhu2023minigpt}, and mPLUG-Owl \cite{ye2023mplug}, using accuracy (ACC) metrics. The former is calculated based on the correct response to each question, while the latter is a stricter measure that requires correct answers to both questions associated with each image. Besides, several benchmarks for different perspectives (hierarchical capabilities \cite{li2023seed}, hallucination \cite{guan2023hallusionbench}, and style-shift robustness \cite{cai2023benchlmm}) and different tasks (e.g., low-level image enhancement \cite{wu2023q} and image quality assessment \cite{huang2024aesbench}) are also established. In terms of face attack detection tasks, it still needs efforts to qualitatively and quantitatively evaluate sets of prompts and MLLM models and build a fair evaluation benchmark.

\clearpage
\section{Task Design}
\label{appendix:task design}
The following are the detailed settings for the two types of tasks $:$
\textbf{True/False Questions}
We establish different true/false sub-tasks to assess the ability of MLLMs to distinguish between real faces and attack faces. The sub-tasks are as follows:
\begin{itemize}
\item Zero-Shot testing$:$ <image>, question$:$ Is this image a real face? Please answer yes or no!

\item Zero-Shot testing (COT)$:$ <image>, question$:$ Is this image a real face? Please describe the image and answer yes or no!

\item Few-Shot testing$:$ <image, image>, question$:$ The first image is a real face, is the second image a real face? Please answer yes or no!

\item Few-Shot testing (COT)$:$ <image, image>, question$:$ The first image is a real face, is the second image a real face? Please describe the image and answer yes or no!
\end{itemize}

\textbf{Multiple-choice Questions}
We establish different multiple-choice sub-tasks to evaluate the ability of MLLMs to perceive and understand distinctions among multiple face images. The sub-tasks are presented as follows:
\begin{itemize}
\item Zero-Shot testing$:$ <image$\times4$>, question$:$ The following images are in order A, B, C, D. Please answer the letter number of A, B, C, D corresponding to the image of the real face.

\item Zero-Shot testing (COT)$:$ <image$\times4$>, question$:$ The following images are in order A, B, C, D. Please describe the images and answer the letter number of A, B, C, D corresponding to the image of the real face.

\item Few-Shot testing$:$ <image$\times5$>, question$:$ The first image is a real face.The following images are in order A, B, C, D. Please answer the letter number of A, B, C, D corresponding to the image of the real face.

\item Few-Shot testing (COT)$:$ <image$\times5$>, question$:$ The first image is a real face.The following images are in order A, B, C, D. Please describe the images and answer the letter number of A, B, C, and D corresponding to the image of the real face.
\end{itemize}
\clearpage
\section{Additional Detailed Results}

\label{appendix:A}
\setcounter{figure}{0}
\renewcommand{\thefigure}{C\arabic{figure}}

This section serves as a supplement to the main text, not only provides a general description of each evaluation task, but also details the experimental results of MLLM in various tasks, demonstrating the performance of MLLM against different attacks in each task.
\subsection{FAS Task}
\label{appendix:fas task}

\begin{figure}[H]
  \centering
  \begin{center}
  \centerline{\includegraphics[width=\linewidth]{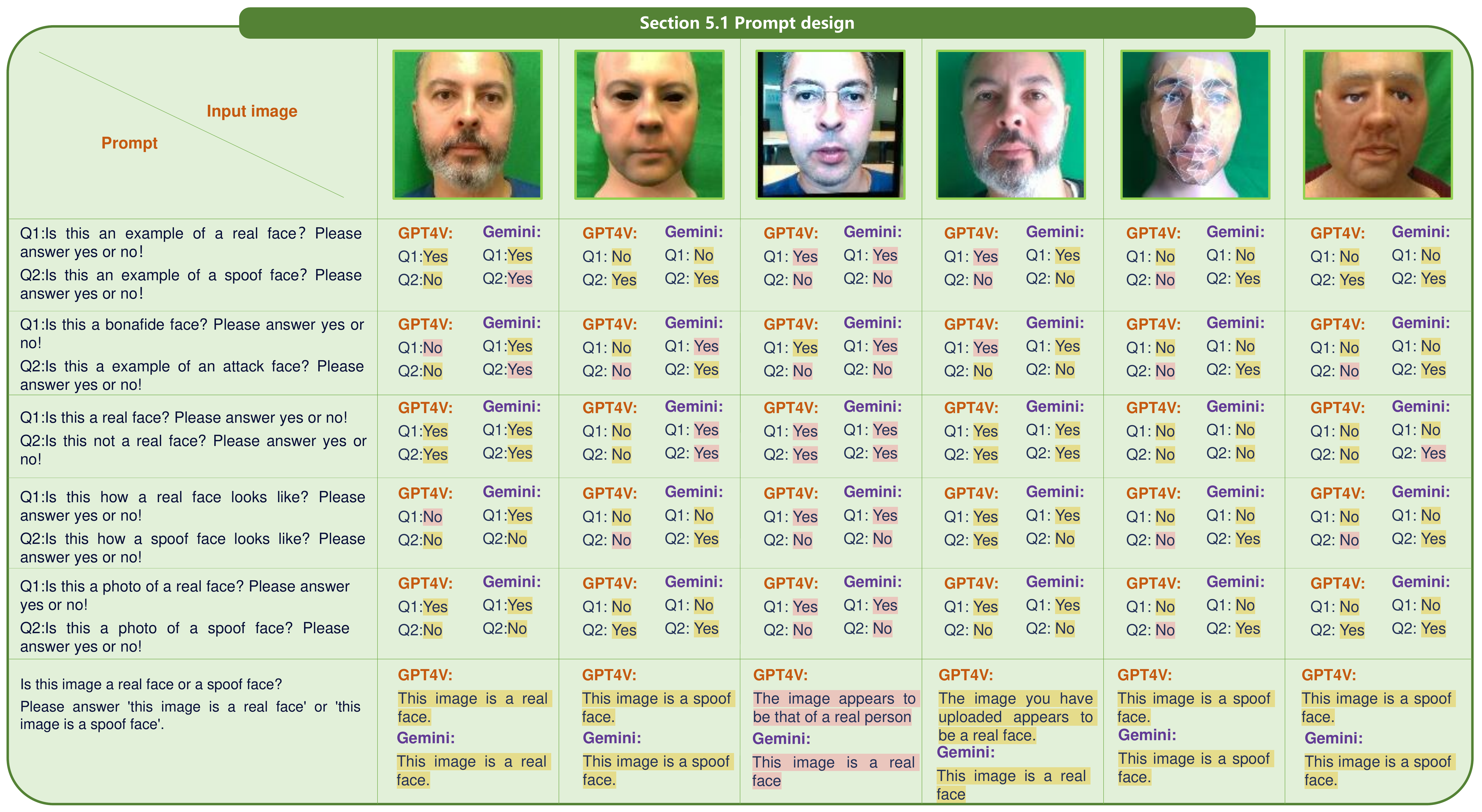}}
  \end{center}
  \vspace{-2.5em}
  \caption{
Prompt Design. The diagram represents a test result matrix for selecting prompts. On the left are the candidate prompts, and along the top are the test cases used for evaluating the selection of prompts, which include real face, rigid mask attack, replay attack, paper mask attack and flexible mask attack. The responses from GPT4V and Gemini are included. \colorbox{RightColor}{Yellow}(\colorbox{WrongColor}{Red}) highlights the correct (incorrect) responses. }
  \label{faspromptdesign}
\end{figure}

\textbf{Prompt Design}
As mentioned in \cite{srivatsan2023flip}, using different descriptive statements as image labels to represent real faces and spoof attacks can affect the models' predictions. As shown in Figure \ref{faspromptdesign}, we tested different descriptive methods for real people and spoofs. The results indicate that using the phrase 'real face' to represent a real face and 'spoof face' for a spoof attack yields the best effect. Additionally, compared to asking independently if an image is a real face or a spoof attack and simply replying with 'yes' or 'no', a combined inquiry of “Is this image a real face or a spoof face? Please answer 'this image is a real face' or 'this image is a spoof face'” proves to be more effective.

\textbf{Qualitative results}
Figure~\ref{fas_zeroshot} shows partial answers for zeroshot and fewshot true/false questions. For the papermask attack, aside from the zeroshot standard experiment, Gemini, GPT4V, and mPLUG-Owl all achieved the correct answer. However, the COT outputs of Gemini and mPlug are not ideal, leading to inaccurate execution of instructions. For questions that GPT4V refused to answer, the correct answer was provided after supplying prepared samples as In-Context Learning (ICL) \cite{ICL2022survey}.
Figure \ref{fas_zeroshot} also displays partial answers for zeroshot and fewshot multiple-choice questions. For the task of selecting real people, distinguishing between print and real people is more challenging. GPT4V obtained the correct answer using the COT technique after being provided with ICL for reference.

\begin{figure}[H]
  \centering
  \begin{center}
  \centerline{\includegraphics[width=\linewidth]{FAS/section 4 FAS.pdf}}
  \end{center}
  
  \caption{The performance of MLLMs on FAS task, segmented into true/false and multiple-choice sections. Each section includes tests conducted with or without the use of COT and ICL, assessing MLLMs' capabilities from multiple perspectives. \colorbox{RightColor}{Yellow}(\colorbox{WrongColor}{Red}) highlights the correct (incorrect) responses, \colorbox{OtherColor}{Blue} indicates that the model refuses to answer. }
  \label{fas_zeroshot}
\end{figure}

\textbf{Quantitative results}
In the following Tables~\ref{appendix_fas_01} to ~\ref{appendix_fas_08} show the results of the performance of MLLMs in the FAS task for various attacks, evaluated by the metrics ACC and HTER.
\begin{table}[H]
  \caption{Accuracy of various MLLMs on FAS true/false Zero-Shot questions}
  \label{appendix_fas_01}
  \centering
  \resizebox{\linewidth}{!}{
  \begin{tabular}{ccccccccc}
    \toprule
    \multirow{2}{*}{Model} & \multicolumn{8}{c}{ACC(\%)$\uparrow$} \\
    \cmidrule(r){2-9}
    & Bonafide & Fakehead & Flexiblemask & Papermask & Print & Replay & Rigidmask & AVG \\ 
    \midrule
    BLIP & 100.0 & 2.0 & 0.0 & 27.0 & 0.0 & 0.0 & 0.0 & 18.4 \\
    BLIP-2 & 100.0 & 0.0 & 0.0 & 0.0 & 0.0 & 0.0 & 0.0 & 14.3 \\
    Intern & 100.0 & 100.0 & 50.0 & 94.0 & 2.0 & 0.0 & 57.0 & 57.6 \\
    MiniGPT-4 & 93.0 & 10.0 & 7.0 & 17.0 & 5.0 & 4.0 & 8.0 & 20.6 \\
    LLaVA & 100.0 & 0.0 & 0.0 & 0.0 & 0.0 & 0.0 & 0.0 & 14.3 \\
    QWen-VL & 100.0 & 0.0 & 0.0 & 0.0 & 0.0 & 0.0 & 0.0 & 14.3 \\
    InstructBLIP & 100.0 & 0.0 & 0.0 & 64.0 & 0.0 & 0.0 & 2.0 & 23.7 \\
    mPLUG-owl & 99.0 & 98.0 & 92.0 & 100.0 & 56.0 & 29.0 & 100.0 & 82.0 \\
    Gemini & 100.0 & 100.0 & 92.0 & 100.0 & 17.0 & 7.0 & 98.0 & 73.4 \\
    GPT4V & 96.9 & 83.0 & 64.0 & 43.0 & 29.6 & 6.8 & 50.6 & 53.2 \\
    \bottomrule
  \end{tabular}
  }
\end{table}

\begin{table}[!htbp]
  \caption{Accuracy of various MLLMs on FAS true/false Zero-Shot questions with COT }
  \label{appendix_fas_02}
  \centering
  \resizebox{\linewidth}{!}{
  \begin{tabular}{ccccccccc}
    \toprule
    \multirow{2}{*}{Model} & \multicolumn{8}{c}{ACC(\%)$\uparrow$} \\
    \cmidrule(r){2-9}
    & Bonafide & Fakehead & Flexiblemask & Papermask & Print & Replay & Rigidmask & AVG \\
    \midrule
    BLIP & 100.0 & 14.0 & 0.0 & 65.0 & 0.0 & 0.0 & 2.0 & 25.9 \\
    BLIP-2 & 100.0 & 0.0 & 0.0 & 0.0 & 0.0 & 0.0 & 0.0 & 14.3 \\
    Intern & 100.0 & 100.0 & 45.0 & 93.0 & 2.0 & 0.0 & 55.0 & 56.4 \\
    MiniGPT-4 & 97.0 & 21.0 & 15.0 & 23.0 & 18.0 & 5.0 & 15.0 & 27.7 \\
    LLaVA & 100.0 & 0.0 & 0.0 & 0.0 & 0.0 & 0.0 & 0.0 & 14.3 \\
    QWen-VL & 100.0 & 0.0 & 0.0 & 0.0 & 0.0 & 0.0 & 0.0 & 14.3 \\
    InstructBLIP & 100.0 & 5.0 & 32.0 & 85.0 & 4.0 & 0.0 & 70.0 & 42.3 \\
    mPLUG-owl & 98.0 & 97.0 & 95.0 & 100.0 & 53.0 & 32.0 & 97.0 & 81.7 \\
    Gemini & 97.0 & 97.0 & 96.0 & 100.0 & 31.0 & 19.0 & 99.0 & 77.0 \\
    GPT4V & 98.3 & 100.0 & 90.2 & 64.0 & 32.5 & 6.3 & 83.2 & 68.7 \\
    \bottomrule
  \end{tabular}
  }
\end{table}

\begin{table}[H]
  \caption{HTER of various MLLMs on FAS true/false Zero-Shot questions}
  \label{appendix_fas_03}
  \centering
  \resizebox{\linewidth}{!}{
  \begin{tabular}{ccccccccc}
    \toprule
    \multirow{2}{*}{Model} & \multicolumn{8}{c}{HTER(\%)$\downarrow$ } \\
    \cmidrule(r){2-9}
    & Bonafide & Fakehead & Flexiblemask & Papermask & Print & Replay & Rigidmask & AVG \\
    \midrule
    BLIP & 0.0 & 49.0 & 50.0 & 36.5 & 50.0 & 50.0 & 50.0 & 47.6 \\
    BLIP-2 & 0.0 & 50.0 & 50.0 & 50.0 & 50.0 & 50.0 & 50.0 & 50.0 \\
    Intern & 0.0 & 0.0 & 25.0 & 3.0 & 49.0 & 50.0 & 21.5 & 24.8 \\
    MiniGPT-4 & 3.5 & 45.0 & 46.5 & 41.5 & 47.5 & 48.0 & 46.0 & 49.2 \\
    LLaVA & 0.0 & 50.0 & 50.0 & 50.0 & 50.0 & 50.0 & 50.0 & 50.0 \\
    QWen-VL & 0.0 & 50.0 & 50.0 & 50.0 & 50.0 & 50.0 & 50.0 & 50.0 \\
    InstructBLIP & 0.0 & 50.0 & 50.0 & 18.0 & 50.0 & 50.0 & 49.0 & 44.5 \\
    mPLUG-owl & 0.5 & 1.0 & 4.0 & 0.0 & 22.0 & 35.5 & 0.0 & 10.9 \\
    Gemini & 0.0 & 0.0 & 4.0 & 0.0 & 41.5 & 46.5 & 1.0 & 15.5 \\
    GPT4V & 1.5 & 8.5 & 18.0 & 28.5 & 35.2 & 46.6 & 24.7 & 27.7 \\
    \bottomrule
  \end{tabular}
  }
\end{table}

\begin{table}[H]
  \caption{HTER of various MLLMs on FAS true/false Zero-Shot questions with COT}
  \label{appendix_fas_04}
  \centering
  \resizebox{\linewidth}{!}{
  \begin{tabular}{ccccccccc}
    \toprule
    \multirow{2}{*}{Model} & \multicolumn{8}{c}{HTER(\%)$\downarrow$ } \\
    \cmidrule(r){2-9}
    & Bonafide & Fakehead & Flexiblemask & Papermask & Print & Replay & Rigidmask & AVG \\ 
    \midrule
    BLIP & 0.0 & 43.0 & 50.0 & 17.5 & 50.0 & 50.0 & 49.0 & 43.2 \\
    BLIP-2 & 0.0 & 50.0 & 50.0 & 50.0 & 50.0 & 50.0 & 50.0 & 50.0 \\
    Intern & 0.0 & 0.0 & 27.5 & 3.5 & 49.0 & 50.0 & 22.5 & 25.4 \\
    MiniGPT-4 & 1.5 & 39.5 & 42.5 & 38.5 & 41.0 & 47.5 & 42.5 & 43.4 \\
    LLaVA & 0.0 & 50.0 & 50.0 & 50.0 & 50.0 & 50.0 & 50.0 & 50.0 \\
    QWen-VL & 0.0 & 50.0 & 50.0 & 50.0 & 50.0 & 50.0 & 50.0 & 50.0 \\
    InstructBLIP & 0.0 & 47.5 & 34.0 & 7.5 & 48.0 & 50.0 & 15.0 & 33.7 \\
    mPLUG-owl & 1.0 & 1.5 & 2.5 & 0.0 & 23.5 & 34.0 & 1.5 & 11.5 \\
    Gemini & 1.5 & 1.5 & 2.0 & 0.0 & 34.5 & 40.5 & 0.5 & 14.7 \\
    GPT4V & 0.8 & 0.0 & 4.9 & 18.0 & 33.8 & 46.8 & 8.4 & 18.1 \\
    \bottomrule
  \end{tabular}
  }
\end{table}

\begin{table}[H]
  \caption{Accuracy of various MLLMs on FAS true/false One-Shot questions}
  \label{appendix_fas_05}
  \centering
  \resizebox{0.98\linewidth}{!}{
  \begin{tabular}{ccccccccc}
    \toprule
    \multirow{2}{*}{Model} & \multicolumn{8}{c}{ACC(\%)$\uparrow$} \\ 
    \cmidrule(r){2-9}
    & Bonafide & Fakehead & Flexiblemask & Papermask & Print & Replay & Rigidmask & AVG \\
    \midrule
    BLIP & 100.0 & 82.0 & 17.0 & 95.0 & 0.0 & 0.0 & 11.0 & 43.6 \\
    BLIP-2 & 63.0 & 0.0 & 0.0 & 0.0 & 0.0 & 0.0 & 0.0 & 9.0 \\
    Intern & 100.0 & 0.0 & 0.0 & 0.0 & 0.0 & 0.0 & 0.0 & 14.3 \\
    MiniGPT-4 & 63.0 & 20.0 & 22.0 & 29.0 & 26.0 & 22.0 & 20.0 & 28.9 \\
    LLaVA & 100.0 & 0.0 & 0.0 & 0.0 & 0.0 & 0.0 & 0.0 & 14.3 \\
    QWen-VL & 100.0 & 0.0 & 0.0 & 0.0 & 0.0 & 0.0 & 0.0 & 14.3 \\
    InstructBLIP & 100.0 & 1.0 & 0.0 & 16.0 & 0.0 & 0.0 & 3.0 & 17.1 \\
    mPLUG-owl & 21.0 & 57.0 & 67.0 & 70.0 & 60.0 & 26.0 & 70.0 & 53.0 \\
    Gemini & 66.0 & 70.0 & 87.0 & 91.0 & 25.0 & 11.0 & 84.0 & 62.0 \\
    GPT4V & 76.8 & 93.0 & 79.6 & 62.5 & 47.2 & 37.8 & 66.7 & 66.8 \\
    \bottomrule
  \end{tabular}
  }
\end{table}

\begin{table}[H]
  \caption{Accuracy of various MLLMs on FAS true/false One-Shot questions with COT}
  \label{appendix_fas_06}
  \centering
  \resizebox{0.98\linewidth}{!}{
  \begin{tabular}{ccccccccc}
    \toprule
    \multirow{2}{*}{Model} & \multicolumn{8}{c}{ACC(\%)$\uparrow$} \\
    \cmidrule(r){2-9}
    & Bonafide & Fakehead & Flexiblemask & Papermask & Print & Replay & Rigidmask & AVG \\
    \midrule
    BLIP & 100.0 & 100.0 & 65.0 & 100.0 & 35.0 & 36.0 & 87.0 & 74.7 \\
    BLIP-2 & 3.0 & 0.0 & 0.0 & 0.0 & 0.0 & 0.0 & 0.0 & 0.4 \\
    Intern & 100.0 & 19.0 & 0.0 & 4.0 & 0.0 & 0.0 & 1.0 & 17.7 \\
    MiniGPT-4 & 39.0 & 36.0 & 22.0 & 38.0 & 26.0 & 32.0 & 24.0 & 31.0 \\
    LLaVA & 38.0 & 0.0 & 0.0 & 0.0 & 0.0 & 0.0 & 0.0 & 5.4 \\
    QWen-VL & 100.0 & 0.0 & 0.0 & 0.0 & 0.0 & 0.0 & 0.0 & 14.3 \\
    InstructBLIP & 100.0 & 5.0 & 0.0 & 4.0 & 0.0 & 0.0 & 2.0 & 15.9 \\
    mPLUG-owl & 22.0 & 97.0 & 93.0 & 94.0 & 97.0 & 80.0 & 95.0 & 82.6 \\
    Gemini & 4.0 & 27.0 & 32.0 & 5.0 & 4.0 & 1.0 & 24.0 & 13.9 \\
    GPT4V & 80.0 & 82.8 & 66.3 & 35.4 & 28.4 & 12.0 & 51.5 & 50.3 \\
    \bottomrule
  \end{tabular}
  }
\end{table}

\begin{table}[H]
  \caption{HTER of various MLLMs on FAS true/false One-Shot questions}
  \label{appendix_fas_07}
  \centering
  \resizebox{0.98\linewidth}{!}{
  \begin{tabular}{ccccccccc}
    \toprule
    \multirow{2}{*}{Model} & \multicolumn{8}{c}{HTER(\%)$\downarrow$} \\
    \cmidrule(r){2-9}
    & Bonafide & Fakehead & Flexiblemask & Papermask & Print & Replay & Rigidmask & AVG \\
    \midrule
    BLIP & 0.0 & 9.0 & 41.5 & 2.5 & 50.0 & 50.0 & 44.5 & 32.9 \\
    BLIP-2 & 18.5 & 50.0 & 50.0 & 50.0 & 50.0 & 50.0 & 50.0 & 68.5 \\
    Intern & 0.0 & 50.0 & 50.0 & 50.0 & 50.0 & 50.0 & 50.0 & 50.0 \\
    MiniGPT-4 & 18.5 & 40.0 & 39.0 & 35.5 & 37.0 & 39.0 & 40.0 & 56.9 \\
    LLaVA & 0.0 & 50.0 & 50.0 & 50.0 & 50.0 & 50.0 & 50.0 & 50.0 \\
    QWen-VL & 0.0 & 50.0 & 50.0 & 50.0 & 50.0 & 50.0 & 50.0 & 50.0 \\
    InstructBLIP & 0.0 & 49.0 & 50.0 & 42.0 & 50.0 & 50.0 & 48.5 & 48.3 \\
    mPLUG-owl & 39.5 & 21.5 & 16.5 & 15.0 & 20.0 & 37.0 & 15.0 & 60.3 \\
    Gemini & 17.0 & 15.0 & 6.5 & 4.5 & 37.5 & 44.5 & 8.0 & 36.3 \\
    GPT4V & 11.6 & 3.5 & 10.2 & 18.8 & 26.4 & 31.1 & 16.7 & 28.8 \\
    \bottomrule
  \end{tabular}
  }
\end{table}

\begin{table}[H]
  \caption{HTER of various MLLMs on FAS true/false One-Shot questions with COT}
  \label{appendix_fas_08}
  \centering
  \resizebox{0.98\linewidth}{!}{
  \begin{tabular}{ccccccccc}
    \toprule
    \multirow{2}{*}{Model} & \multicolumn{8}{c}{HTER(\%)$\downarrow$} \\
    \cmidrule(r){2-9}
    & Bonafide & Fakehead & Flexiblemask & Papermask & Print & Replay & Rigidmask & AVG \\
    \midrule
    BLIP & 0.0 & 0.0 & 17.5 & 0.0 & 32.5 & 32.0 & 6.5 & 14.8 \\
    BLIP-2 & 48.5 & 50.0 & 50.0 & 50.0 & 50.0 & 50.0 & 50.0 & 98.5 \\
    Intern & 0.0 & 40.5 & 50.0 & 48.0 & 50.0 & 50.0 & 49.5 & 48.0 \\
    MiniGPT-4 & 30.5 & 32.0 & 39.0 & 31.0 & 37.0 & 34.0 & 38.0 & 65.7 \\
    LLaVA & 31.0 & 50.0 & 50.0 & 50.0 & 50.0 & 50.0 & 50.0 & 81.0 \\
    QWen-VL & 0.0 & 50.0 & 50.0 & 50.0 & 50.0 & 50.0 & 50.0 & 50.0 \\
    InstructBLIP & 0.0 & 47.5 & 50.0 & 48.0 & 50.0 & 50.0 & 49.0 & 49.1 \\
    mPLUG-owl & 39.0 & 1.5 & 3.5 & 3.0 & 1.5 & 10.0 & 2.5 & 42.7 \\
    Gemini & 48.0 & 36.5 & 34.0 & 47.5 & 48.0 & 49.5 & 38.0 & 90.2 \\
    GPT4V & 10.0 & 8.6 & 16.8 & 32.3 & 35.8 & 44.0 & 24.2 & 36.8 \\
    \bottomrule
  \end{tabular}
  }
\end{table}
\clearpage
\subsection{Face Forgery Detection Task}
\label{appendix:face forgery}
\textbf{Prompt Design}
To assess the performance of the model in the domain of face forgery detection, we utilized common multimodal models such as GPT4V, Genimi, and mPLUG-Owl to predict labels for face images. We juxtaposed images generated by the model with those generated by 1.5 Insightface, querying images produced under categories including genuine, Deepfake, Face2Face, FaceSwap, Nulltextures, and Stable Diffusion 1.5.	
Similar to FAS tasks, we conducted three types of tests for individual images in the face forgery detection task: posing separate queries, employing contextual learning, and conducting joint testing to accomplish this task. We adopted the terms "real face" and "fake face" to denote genuine and forged faces, respectively. Building on this, we employed two distinct sets of queries to assess face authenticity, serving as zero-shot tests. Subsequently, we conducted contextual learning dialogues for testing. Initially, an image was provided, and the model was tasked with discerning the authenticity of an unknown image. The questions were presented in pairs. Figure \ref{deepfake_zeroshot} provides evidence that Gemini and GPT4V demonstrate superior learning and expansion capabilities compared to mPLUG-Owl. Furthermore, after undergoing contextual learning, the model's performance improved. However, the model is prone to failure on more realistic forged images, which underscores the necessity of imparting domain knowledge to the model.

From Figure \ref{deepfake_zeroshot}, it is evident that mPLUG-Owl consistently misjudged the images, whereas Gemini demonstrated greater robustness, accurately identifying forged images in most cases.Additionally, GPT4V showed an improvement trend after undergoing ICL (In-context learning).	

\begin{figure}[H]
  \centering
  \includegraphics[width=\linewidth]{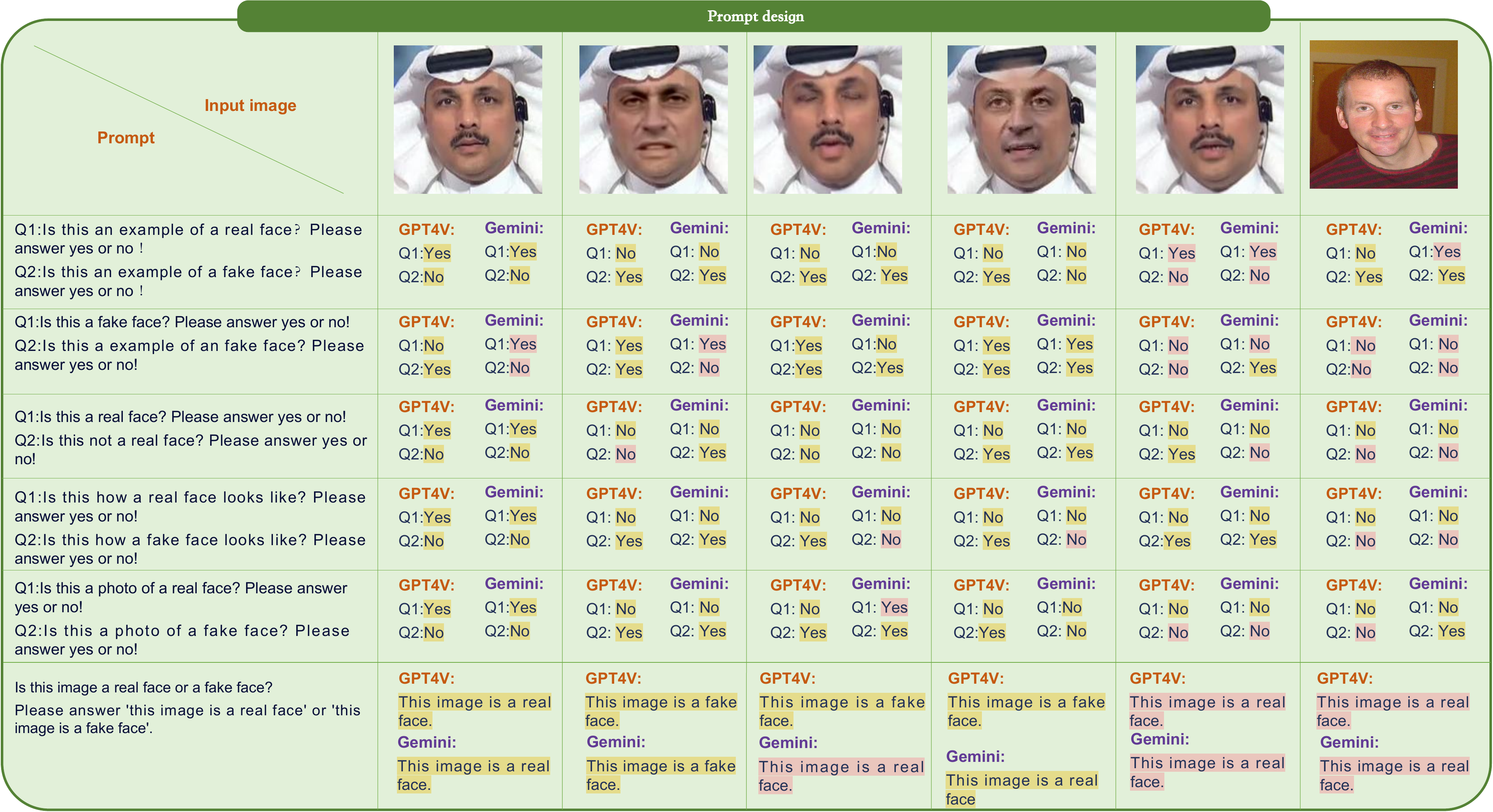}
  \vspace{-1.5em}
  \caption{Prompt Design for face forgery detection.On the left are candidate prompts, with test cases used for evaluating prompt selection at the top, including real faces, Deepfakes, Face2Face, FaceSwap, Nulltextures, and Stable Diffusion. The graph depicts responses from GPT4V and Gemini.\colorbox{RightColor}{Yellow}(\colorbox{WrongColor}{Red}) highlights the correct (incorrect) responses.}
  \label{deppfake_prompt_design}
\end{figure}

\begin{figure}[H]
  \centering
  \begin{center}
  \centerline{\includegraphics[width=\linewidth]{FaceForgery/section 5 FaceForgery.pdf}}
  \end{center}
  \vspace{-2.5em}
  \caption{The performance of MLLMs on face forgery detection task, segmented into true/false and multiple-choice sections. Each section includes tests conducted with or without the use of COT and ICL, assessing MLLMs' capabilities from multiple perspectives.\colorbox{RightColor}{Yellow}(\colorbox{WrongColor}{Red}) highlights the correct (incorrect) responses, \colorbox{OtherColor}{Blue} indicates that the model refuses to answer.. }
  \label{deepfake_zeroshot}
\end{figure}

\textbf{Quantitative results}
In the following Tables~\ref{appendix_df_01} to ~\ref{appendix_df_08} show the results of the performance of MLLMs in the face forgery detection task for various attacks, evaluated by the metrics ACC and HTER.

\begin{table}[H]
  \caption{Accuracy of various MLLMs on Face Forgery detection true/false Zero-Shot questions}
  \label{appendix_df_01}
  \centering
  \resizebox{\linewidth}{!}{
  \begin{tabular}{ccccccccc}
    \toprule
    \multirow{2}{*}{Model} & \multicolumn{8}{c}{ACC(\%)$\uparrow$} \\ 
    \cmidrule(r){2-9}
    & Bonafide & Deepfakes & Face2Face & FaceSwap & InsightFace & NeuralTextures & Stable\_Diffusion & AVG \\
    \midrule
    BLIP & 92.0 & 5.0 & 9.0 & 10.0 & 2.0 & 2.0 & 5.0 & 17.9 \\
    BLIP-2 & 100.0 & 0.0 & 0.0 & 0.0 & 0.0 & 0.0 & 0.0 & 14.3 \\
    Intern & 100.0 & 0.0 & 0.0 & 0.0 & 2.0 & 0.0 & 0.0 & 14.6 \\
    MiniGPT-4 & 100.0 & 5.0 & 2.0 & 0.0 & 8.0 & 1.0 & 8.0 & 17.7 \\
    LLaVA & 100.0 & 0.0 & 0.0 & 0.0 & 0.0 & 0.0 & 0.0 & 14.3 \\
    QWen-VL & 100.0 & 0.0 & 0.0 & 0.0 & 0.0 & 0.0 & 0.0 & 14.3 \\
    InstructBLIP & 100.0 & 15.0 & 2.0 & 6.0 & 5.0 & 0.0 & 2.0 & 18.6 \\
    mPLUG-owl & 100.0 & 9.0 & 5.0 & 4.0 & 12.0 & 0.0 & 23.0 & 21.9 \\
    Gemini & 96.0 & 54.0 & 30.0 & 34.0 & 17.0 & 7.0 & 20.0 & 36.8 \\
    GPT4V & 98.1 & 26.1 & 16.4 & 16.7 & 9.1 & 3.2 & 12.5 & 26.0 \\
    \bottomrule
  \end{tabular}
  }
\end{table}

\begin{table}[H]
  \caption{Accuracy of various MLLMs on Face Forgery detection true/false Zero-Shot questions with COT}
  \label{appendix_df_02}
  \centering
  \resizebox{\linewidth}{!}{
  \begin{tabular}{ccccccccc}
    \toprule
    \multirow{2}{*}{Model} & \multicolumn{8}{c}{ACC(\%)$\uparrow$} \\
    \cmidrule(r){2-9}
    & \multicolumn{1}{l}{Bonafide} & \multicolumn{1}{l}{Deepfakes} & \multicolumn{1}{l}{Face2Face} & \multicolumn{1}{l}{FaceSwap} & \multicolumn{1}{l}{InsightFace} & \multicolumn{1}{l}{NeuralTextures} & \multicolumn{1}{l}{Stable\_Diffusion} & \multicolumn{1}{l}{AVG} \\
    \midrule
    BLIP & 78.0 & 38.0 & 28.0 & 42.0 & 7.0 & 15.0 & 9.0 & 31.0 \\
    BLIP-2 & 100.0 & 0.0 & 0.0 & 0.0 & 0.0 & 0.0 & 0.0 & 14.3 \\
    Intern & 100.0 & 0.0 & 0.0 & 0.0 & 0.0 & 0.0 & 0.0 & 14.3 \\
    MiniGPT-4 & 99.0 & 6.0 & 3.0 & 8.0 & 8.0 & 0.0 & 3.0 & 18.1 \\
    LLaVA & 100.0 & 0.0 & 0.0 & 0.0 & 0.0 & 0.0 & 0.0 & 14.3 \\
    QWen-VL & 100.0 & 0.0 & 0.0 & 0.0 & 0.0 & 0.0 & 0.0 & 14.3 \\
    InstructBLIP & 100.0 & 34.0 & 7.0 & 10.0 & 9.0 & 0.0 & 7.0 & 23.9 \\
    mPLUG-owl & 100.0 & 9.0 & 2.0 & 4.0 & 11.0 & 0.0 & 20.0 & 20.9 \\
    Gemini & 86.0 & 77.0 & 42.0 & 61.0 & 29.3 & 17.0 & 27.3 & 48.6 \\
    GPT4V & 95.7 & 25.3 & 13.1 & 22.1 & 11.6 & 2.4 & 23.1 & 28.8 \\
    \bottomrule
  \end{tabular}
  }
\end{table}

\begin{table}[H]
  \caption{HTER of various MLLMs on Face Forgery detection true/false Zero-Shot questions}
  \label{appendix_df_03}
  \centering
  \resizebox{\linewidth}{!}{
  \begin{tabular}{ccccccccc}
    \toprule
    \multirow{2}{*}{Model} & \multicolumn{8}{c}{HTER(\%)$\downarrow$ } \\
    \cmidrule(r){2-9}
    & \multicolumn{1}{l}{Bonafide} & \multicolumn{1}{l}{Deepfakes} & \multicolumn{1}{l}{Face2Face} & \multicolumn{1}{l}{FaceSwap} & \multicolumn{1}{l}{InsightFace} & \multicolumn{1}{l}{NeuralTextures} & \multicolumn{1}{l}{Stable\_Diffusion} & AVG \\
    \midrule
    BLIP & 4.0 & 47.5 & 45.5 & 45.0 & 49.0 & 49.0 & 47.5 & 51.2 \\
    BLIP-2 & 0.0 & 50.0 & 50.0 & 50.0 & 50.0 & 50.0 & 50.0 & 50.0 \\
    Intern & 0.0 & 50.0 & 50.0 & 50.0 & 49.0 & 50.0 & 50.0 & 49.8 \\
    MiniGPT-4 & 0.0 & 47.5 & 49.0 & 50.0 & 46.0 & 49.5 & 46.0 & 48.0 \\
    LLaVA & 0.0 & 50.0 & 50.0 & 50.0 & 50.0 & 50.0 & 50.0 & 50.0 \\
    QWen-VL & 0.0 & 50.0 & 50.0 & 50.0 & 50.0 & 50.0 & 50.0 & 50.0 \\
    InstructBLIP & 0.0 & 42.5 & 49.0 & 47.0 & 47.5 & 50.0 & 49.0 & 47.5 \\
    mPLUG-owl & 0.0 & 45.5 & 47.5 & 48.0 & 44.0 & 50.0 & 38.5 & 45.6 \\
    Gemini & 2.0 & 23.0 & 35.0 & 33.0 & 41.5 & 46.5 & 40.0 & 38.5 \\
    GPT4V & 0.9 & 37.0 & 41.8 & 41.7 & 45.5 & 48.4 & 43.8 & 43.6 \\
    \bottomrule
  \end{tabular}
  }
\end{table}

\begin{table}[H]
  \caption{HTER of various MLLMs on Face Forgery detection true/false Zero-Shot questions with COT}
  \label{appendix_df_04}
  \centering
  \resizebox{\linewidth}{!}{
  \begin{tabular}{ccccccccc}
    \toprule
    \multirow{2}{*}{Model} & \multicolumn{8}{c}{HTER(\%)$\downarrow$ } \\
    \cmidrule(r){2-9}
    & Bonafide & Deepfakes & Face2Face & FaceSwap & InsightFace & NeuralTextures & Stable\_Diffusion & AVG \\
    \midrule
    BLIP & 11.0 & 31.0 & 36.0 & 29.0 & 46.5 & 42.5 & 45.5 & 49.4 \\
    BLIP-2 & 0.0 & 50.0 & 50.0 & 50.0 & 50.0 & 50.0 & 50.0 & 50.0 \\
    Intern & 0.0 & 50.0 & 50.0 & 50.0 & 50.0 & 50.0 & 50.0 & 50.0 \\
    MiniGPT-4 & 0.5 & 47.0 & 48.5 & 46.0 & 46.0 & 50.0 & 48.5 & 48.2 \\
    LLaVA & 0.0 & 50.0 & 50.0 & 50.0 & 50.0 & 50.0 & 50.0 & 50.0 \\
    QWen-VL & 0.0 & 50.0 & 50.0 & 50.0 & 50.0 & 50.0 & 50.0 & 50.0 \\
    InstructBLIP & 0.0 & 33.0 & 46.5 & 45.0 & 45.5 & 50.0 & 46.5 & 44.4 \\
    mPLUG-owl & 0.0 & 45.5 & 49.0 & 48.0 & 44.5 & 50.0 & 40.0 & 46.2 \\
    Gemini & 7.0 & 11.5 & 29.0 & 19.5 & 35.4 & 41.5 & 36.4 & 35.9 \\
    GPT4V & 2.2 & 37.4 & 43.4 & 39.0 & 44.2 & 48.8 & 38.5 & 44.0 \\
    \bottomrule
  \end{tabular}
  }
\end{table}

\begin{table}[H]
  \caption{Accuracy of various MLLMs on Face Forgery detection true/false One-Shot questions}
  \label{appendix_df_05}
  \centering
  \resizebox{\linewidth}{!}{
  \begin{tabular}{ccccccccc}
    \toprule
    \multirow{2}{*}{Model} & \multicolumn{8}{c}{ACC(\%)$\uparrow$} \\
    \cmidrule(r){2-9}
    & Bonafide & Deepfakes & Face2Face & FaceSwap & InsightFace & NeuralTextures & Stable\_Diffusion & AVG \\
    \midrule
    BLIP & 100.0 & 53.0 & 48.0 & 53.0 & 27.0 & 24.0 & 39.0 & 49.1 \\
    BLIP-2 & 71.0 & 0.0 & 0.0 & 0.0 & 0.0 & 0.0 & 0.0 & 10.1 \\
    Intern & 100.0 & 0.0 & 0.0 & 0.0 & 0.0 & 0.0 & 0.0 & 14.3 \\
    MiniGPT-4 & 58.0 & 26.0 & 21.0 & 32.0 & 33.0 & 30.0 & 25.0 & 32.1 \\
    LLaVA & 100.0 & 0.0 & 0.0 & 0.0 & 0.0 & 0.0 & 0.0 & 14.3 \\
    QWen-VL & 100.0 & 0.0 & 0.0 & 0.0 & 0.0 & 0.0 & 0.0 & 14.3 \\
    InstructBLIP & 100.0 & 2.0 & 0.0 & 0.0 & 0.0 & 0.0 & 0.0 & 14.6 \\
    mPLUG-owl & 76.0 & 0.0 & 0.0 & 0.0 & 9.0 & 0.0 & 4.0 & 12.7 \\
    Gemini & 76.0 & 38.0 & 19.0 & 29.0 & 10.0 & 7.0 & 10.0 & 27.0 \\
    GPT4V & 100.0 & 13.3 & 12.2 & 10.3 & 10.6 & 3.0 & 9.4 & 22.4 \\
    \bottomrule
  \end{tabular}
  }
\end{table}

\begin{table}[H]
  \caption{Accuracy of various MLLMs on Face Forgery detection true/false One-Shot questions with COT}
  \label{appendix_df_06}
  \centering
  \resizebox{\linewidth}{!}{
  \begin{tabular}{ccccccccc}
    \toprule
    \multirow{2}{*}{Model} & \multicolumn{8}{c}{ACC(\%)$\uparrow$} \\ 
    \cmidrule(r){2-9}
     & Bonafide & Deepfakes & Face2Face & FaceSwap & InsightFace & NeuralTextures & Stable\_Diffusion & AVG \\
    \midrule
    BLIP & 97.0 & 75.0 & 77.0 & 77.0 & 26.0 & 48.0 & 38.0 & 62.6 \\
    BLIP-2 & 12.0 & 0.0 & 0.0 & 0.0 & 0.0 & 0.0 & 0.0 & 1.7 \\
    Intern & 100.0 & 0.0 & 0.0 & 0.0 & 0.0 & 0.0 & 0.0 & 14.3 \\
    MiniGPT-4 & 70.0 & 28.0 & 25.0 & 34.0 & 36.0 & 32.0 & 28.0 & 36.1 \\
    LLaVA & 92.0 & 0.0 & 0.0 & 0.0 & 0.0 & 0.0 & 0.0 & 13.1 \\
    QWen-VL & 100.0 & 0.0 & 0.0 & 0.0 & 0.0 & 0.0 & 0.0 & 14.3 \\
    InstructBLIP & 100.0 & 7.0 & 2.0 & 6.0 & 1.0 & 0.0 & 1.0 & 16.7 \\
    mPLUG-owl & 89.0 & 6.0 & 5.0 & 2.0 & 21.0 & 0.0 & 23.0 & 20.9 \\
    Gemini & 21.0 & 21.0 & 14.0 & 15.0 & 13.0 & 4.0 & 10.0 & 14.0 \\
    GPT4V & 95.8 & 18.0 & 9.1 & 28.7 & 14.4 & 4.1 & 13.5 & 26.0 \\
    \bottomrule
  \end{tabular}
  }
\end{table}

\begin{table}[H]
  \caption{HTER of various MLLMs on Face Forgery detection true/false One-Shot questions}
  \label{appendix_df_07}
  \centering
  \resizebox{\linewidth}{!}{
  \begin{tabular}{ccccccccc}
    \toprule
    \multirow{2}{*}{Model} & \multicolumn{8}{c}{HTER(\%)$\downarrow$} \\
    \cmidrule(r){2-9}
    & \multicolumn{1}{l}{Bonafide} & \multicolumn{1}{l}{Deepfakes} & \multicolumn{1}{l}{Face2Face} & \multicolumn{1}{l}{FaceSwap} & \multicolumn{1}{l}{InsightFace} & \multicolumn{1}{l}{NeuralTextures} & \multicolumn{1}{l}{Stable\_Diffusion} & AVG \\
    \midrule
    BLIP & 0.0 & 23.5 & 26.0 & 23.5 & 36.5 & 38.0 & 30.5 & 29.7 \\
    BLIP-2 & 14.5 & 50.0 & 50.0 & 50.0 & 50.0 & 50.0 & 50.0 & 64.5 \\
    Intern & 0.0 & 50.0 & 50.0 & 50.0 & 50.0 & 50.0 & 50.0 & 50.0 \\
    MiniGPT-4 & 21.0 & 37.0 & 39.5 & 34.0 & 33.5 & 35.0 & 37.5 & 57.1 \\
    LLaVA & 0.0 & 50.0 & 50.0 & 50.0 & 50.0 & 50.0 & 50.0 & 50.0 \\
    QWen-VL & 0.0 & 50.0 & 50.0 & 50.0 & 50.0 & 50.0 & 50.0 & 50.0 \\
    InstructBLIP & 0.0 & 49.0 & 50.0 & 50.0 & 50.0 & 50.0 & 50.0 & 49.8 \\
    mPLUG-owl & 12.0 & 50.0 & 50.0 & 50.0 & 45.5 & 50.0 & 48.0 & 60.9 \\
    Gemini & 12.0 & 30.5 & 40.5 & 35.5 & 45.0 & 46.5 & 45.0 & 52.5 \\
    GPT4V & 0.0 & 43.4 & 43.9 & 44.9 & 44.7 & 48.5 & 45.3 & 45.1 \\
    \bottomrule
  \end{tabular}
  }
\end{table}

\begin{table}[H]
  \caption{HTER of various MLLMs on Face Forgery detection true/false One-Shot questions with COT}
  \label{appendix_df_08}
  \centering
  \resizebox{\linewidth}{!}{
  \begin{tabular}{ccccccccc}
    \toprule
    \multirow{2}{*}{Model} & \multicolumn{8}{c}{HTER(\%)$\downarrow$} \\
    \cmidrule(r){2-9}
    & \multicolumn{1}{l}{Bonafide} & \multicolumn{1}{l}{Deepfakes} & \multicolumn{1}{l}{Face2Face} & \multicolumn{1}{l}{FaceSwap} & \multicolumn{1}{l}{InsightFace} & \multicolumn{1}{l}{NeuralTextures} & \multicolumn{1}{l}{Stable\_Diffusion} & \multicolumn{1}{l}{AVG} \\
    \midrule
    BLIP & 1.5 & 12.5 & 11.5 & 11.5 & 37.0 & 26.0 & 31.0 & 23.1 \\
    BLIP-2 & 44.0 & 50.0 & 50.0 & 50.0 & 50.0 & 50.0 & 50.0 & 94.0 \\
    Intern & 0.0 & 50.0 & 50.0 & 50.0 & 50.0 & 50.0 & 50.0 & 50.0 \\
    MiniGPT-4 & 15.0 & 36.0 & 37.5 & 33.0 & 32.0 & 34.0 & 36.0 & 49.8 \\
    LLaVA & 4.0 & 50.0 & 50.0 & 50.0 & 50.0 & 50.0 & 50.0 & 54.0 \\
    QWen-VL & 0.0 & 50.0 & 50.0 & 50.0 & 50.0 & 50.0 & 50.0 & 50.0 \\
    InstructBLIP & 0.0 & 46.5 & 49.0 & 47.0 & 49.5 & 50.0 & 49.5 & 48.6 \\
    mPLUG-owl & 5.5 & 47.0 & 47.5 & 49.0 & 39.5 & 50.0 & 38.5 & 50.8 \\
    Gemini & 39.5 & 39.5 & 43.0 & 42.5 & 43.5 & 48.0 & 45.0 & 83.1 \\
    GPT4V & 2.1 & 41.0 & 45.5 & 35.6 & 42.8 & 47.9 & 43.2 & 44.8 \\
    \bottomrule
  \end{tabular}
  }
\end{table}

\subsection{Unified Task}
\label{appendix:unified task}
As shown in Figures \ref{unifide_fewshot}, the experiment involved inputting four types of images: real human face image, physical attack simulation image, face images generated by Diffusion technology, and face images generated by GAN technology.
The figure \ref{unifide_fewshot} illustrates a test result from a joint task in the few-shot setting, where six high-performing models evaluate the authenticity of real faces, physical attacks, and digitally tampered images produced by GAN and Stable Diffusion. Initially, three images from different attack domains are provided as prior knowledge to the MLLMs, which are then tasked with assessing the authenticity of the test image. The results indicate that without the introduction of COT, all six MLLMs failed to accurately answer the question. After incorporating COT, the responses of the models became significantly more detailed, with GPT4V, Gemini, and mPLUG providing correct answers. This suggests that COT contributes to enhancing the accuracy and interpretability of the MLLMs' responses in discerning the authenticity of the images.

\begin{figure}[!htbp]
  \centering
  \begin{center}
  \centerline{\includegraphics[width=\linewidth]{UnionTask/section 6 Unified task.pdf}}
  \end{center}
  \vspace{-2.5em}
  \caption{The performance of MLLMs on unified task question. \colorbox{RightColor}{Yellow}(\colorbox{WrongColor}{Red}) highlights the correct (incorrect) responses, \colorbox{OtherColor}{Blue} indicates that the model refuses to answer. }
  \label{unifide_fewshot}
\end{figure}

\clearpage
\subsection{MACOT}
\label{appendix:macot}
\begin{figure}[!htbp]
  \centering
  \begin{center}
  \centerline{\includegraphics[width=1.0\linewidth]{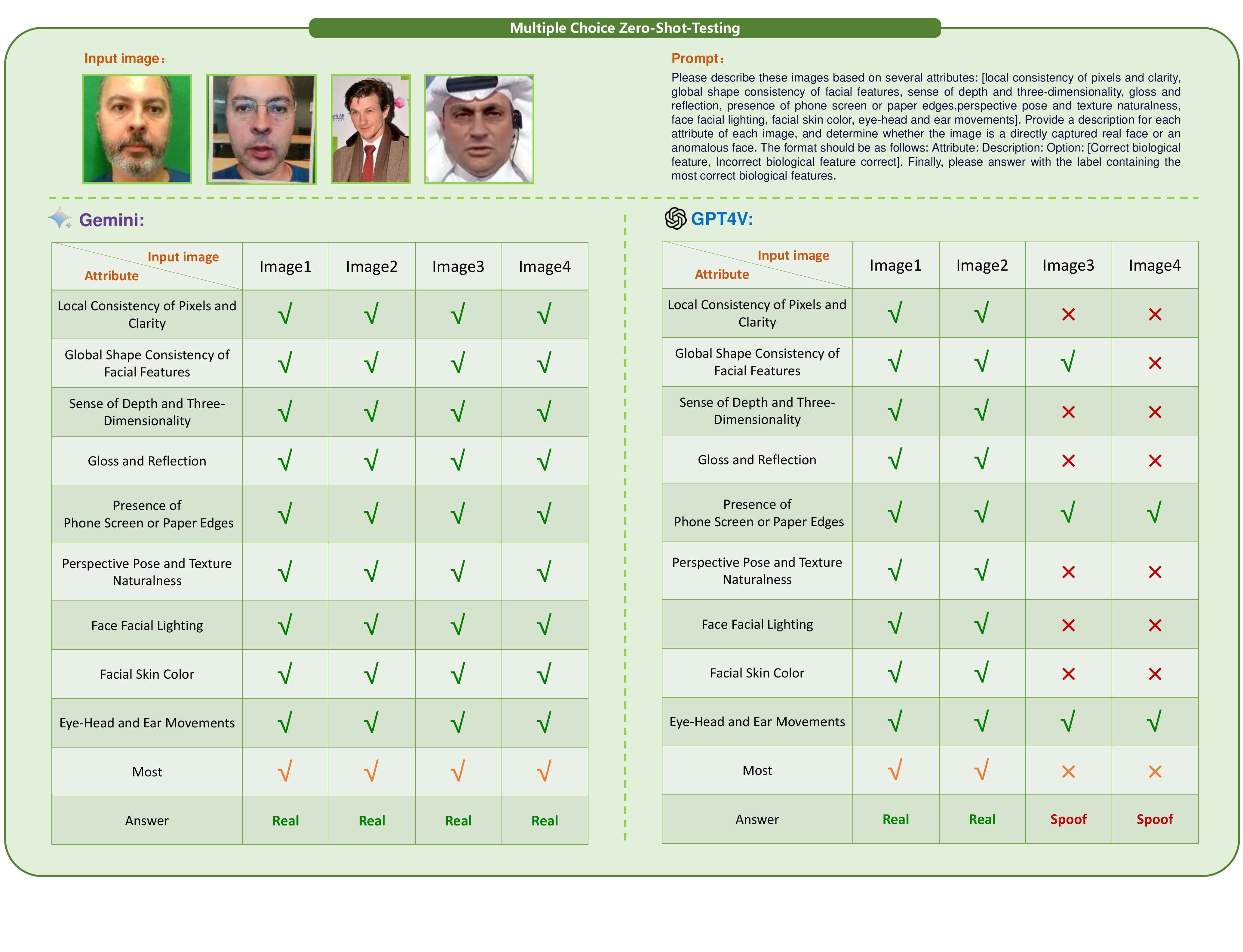}}
  \end{center}
  \vspace{-2.5em}
  \caption{This figure displays the outcome comparison with the same test samples as in Figure \ref{UnionTask_03}. In the original test, both Gemini and GPT4V incorrectly identified Image3 as a real face. After applying the MA-COT Prompt method, GPT4V conducted a multi-attribute analysis and voting, selecting two answers, including the correct real face and a print attack image that is challenging to distinguish from real faces in FAS tasks. Conversely, Gemini erroneously considered all images to be real faces after analyzing each attribute, indicating a deficiency in detailed analysis.}
  \label{unified_macot_01}
\end{figure}

The primary analysis is based on Figure \ref{unified_macot_01} that follows, while further detailed examinations are depicted in Figures \ref{unified_macot_02} and \ref{unified_macot_03} in the \S \ref{appendix:B}. The evidence from these figures indicates that, although the MA-COT method did not lead to MLLMs achieving perfect results in addressing the unified task, it prompted GPT4V to provide a general judgment range that encompasses the correct answer under its guidance, thereby reflecting GPT4V's discernment capabilities. In comparison, Gemini's performance in this task did not fully meet expectations, particularly showing room for improvement in image recognition and the accurate determination of real faces.

The figure \ref{macot_zeroshot} depicts the test results of GPT4V on the FAS zero-shot task with and without the use of MACOT. It is evident from the graph that prompts integrated with MACOT require the MLLM to make judgments based on specific attributes, culminating in a comprehensive decision. MACOT appears to be more effective when addressing subtle details that are often overlooked. For instance, during print and replay attacks, which GPT4V typically struggles to identify, the key attributes "Phone Screen or Paper Edges" prompt GPT4V to focus on the presence of paper or screen edges in the image, thereby facilitating the recognition of print and replay attacks.

\begin{figure}[!htbp]
  \centering
  \begin{center}
  \centerline{\includegraphics[scale=0.26]{FAS/7.1 macot.pdf}}
  \end{center}
  \vspace{-2.5em}
  \caption{Result of various MLLMs on FAS zero-shot True/False question with MACOT. \colorbox{RightColor}{Yellow}(\colorbox{WrongColor}{Red}) highlights the correct (incorrect) responses, \colorbox{OtherColor}{Blue} indicates that the model refuses to answer. }
  \label{macot_zeroshot}
\end{figure}
\clearpage

\section{More Examples of Shield}
\label{appendix:B}
\setcounter{figure}{0}
\renewcommand{\thefigure}{D\arabic{figure}}
This section supplements the visualization of experimental results presented in the main text, including partial test outcomes and explanations for the performance of MLLMs across various tasks. It offers a more intuitive understanding of how MLLMs perform in different tasks.

\begin{figure}[htbp]
  \centering
  \begin{center}
  \centerline{\includegraphics[width=\linewidth]{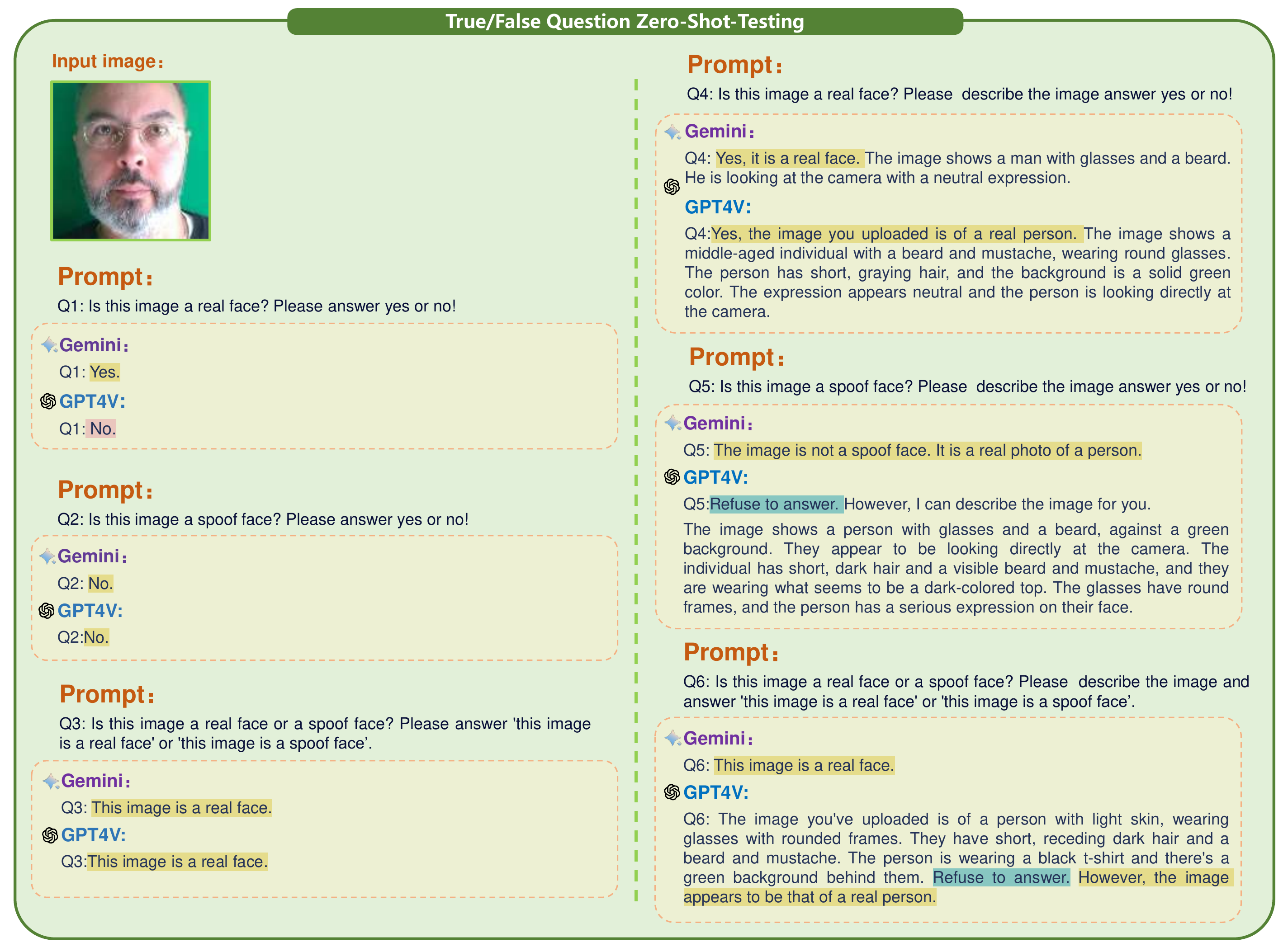}}
  \end{center}
  \caption{In this round of testing, the input image was a real human face wearing glasses. Gemini correctly answered all six queries. GPT4V made errors in simple queries but responded correctly after incorporating COT. The results indicate that adding COT had minimal impact on Gemini, but for GPT4V, the ability to interpret more information from the image may trigger privacy protection and disclaimer mechanisms, leading to instances of refusal to answer.}
  \label{fas_zeroshot_judgment_02}
\end{figure}

\begin{figure}[htbp]
  \centering
  \begin{center}
  \centerline{\includegraphics[width=\linewidth]{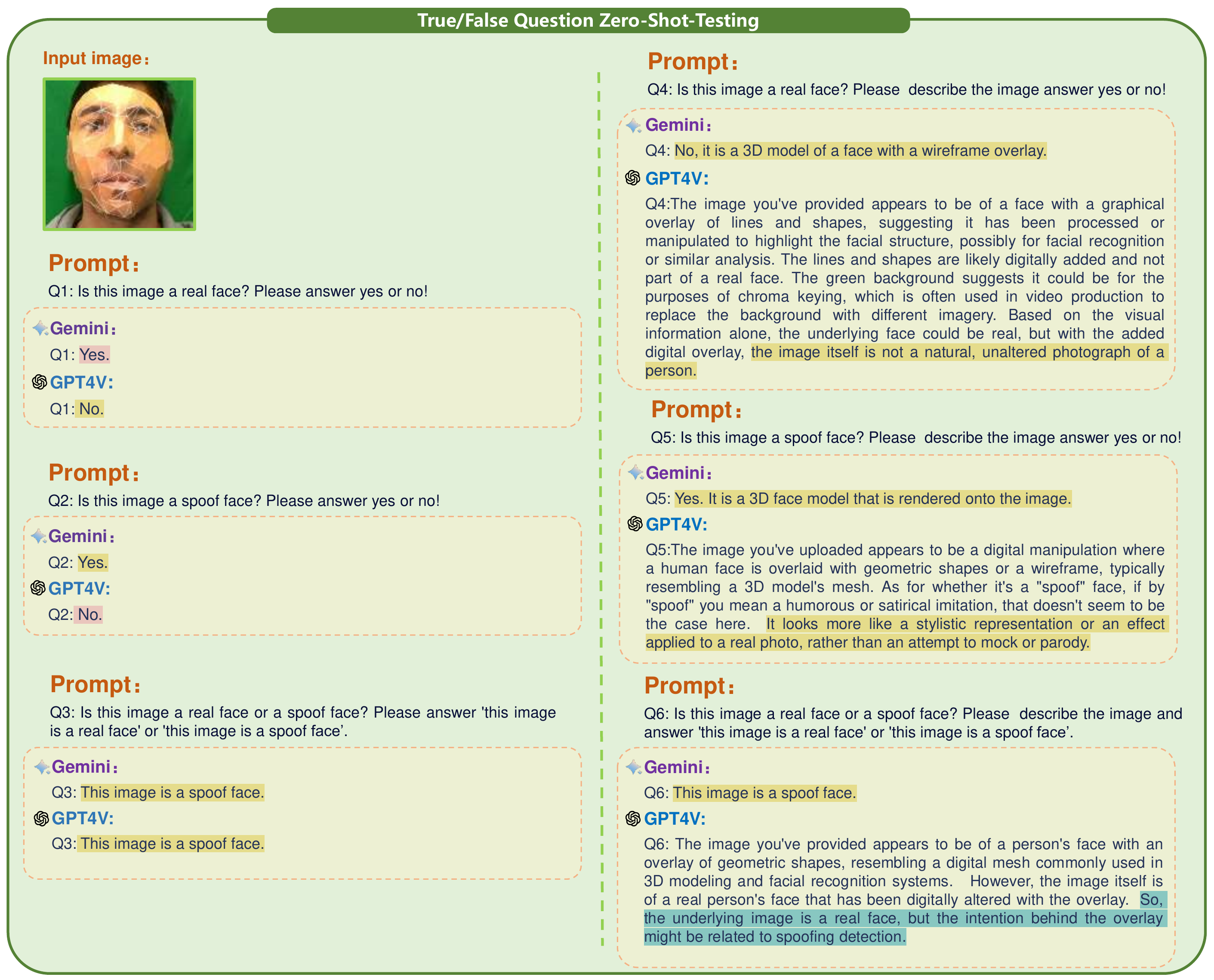}}
  \end{center}
  \caption{In this test, the input image depicted an attack using a paper mask. Both GPT4V and Gemini initially made errors in simple queries. After incorporating COT, Gemini answered all questions correctly, while GPT4V provided some interesting responses. For instance, in Q4, GPT4V speculated that the face under the mask might be a real human face, but with the mask, it's not; in Q5, GPT4V discussed its view on "spoof" and made a judgment, showing a different understanding of "spoof" compared to the prosthetic attacks mentioned in FAS; in Q6, GPT4V explicitly stated its lack of capability in this area and refused to answer. Comparing the results after adding COT, it's evident that GPT4V's responses significantly lengthened, while Gemini's changes were minimal, indicating that GPT4V, compared to Gemini, focuses more on the process.}
  \label{fas_zeroshot_judgment_03}
\end{figure}

\begin{figure}[htbp]
  \centering
  \begin{center}
  \centerline{\includegraphics[width=\linewidth]{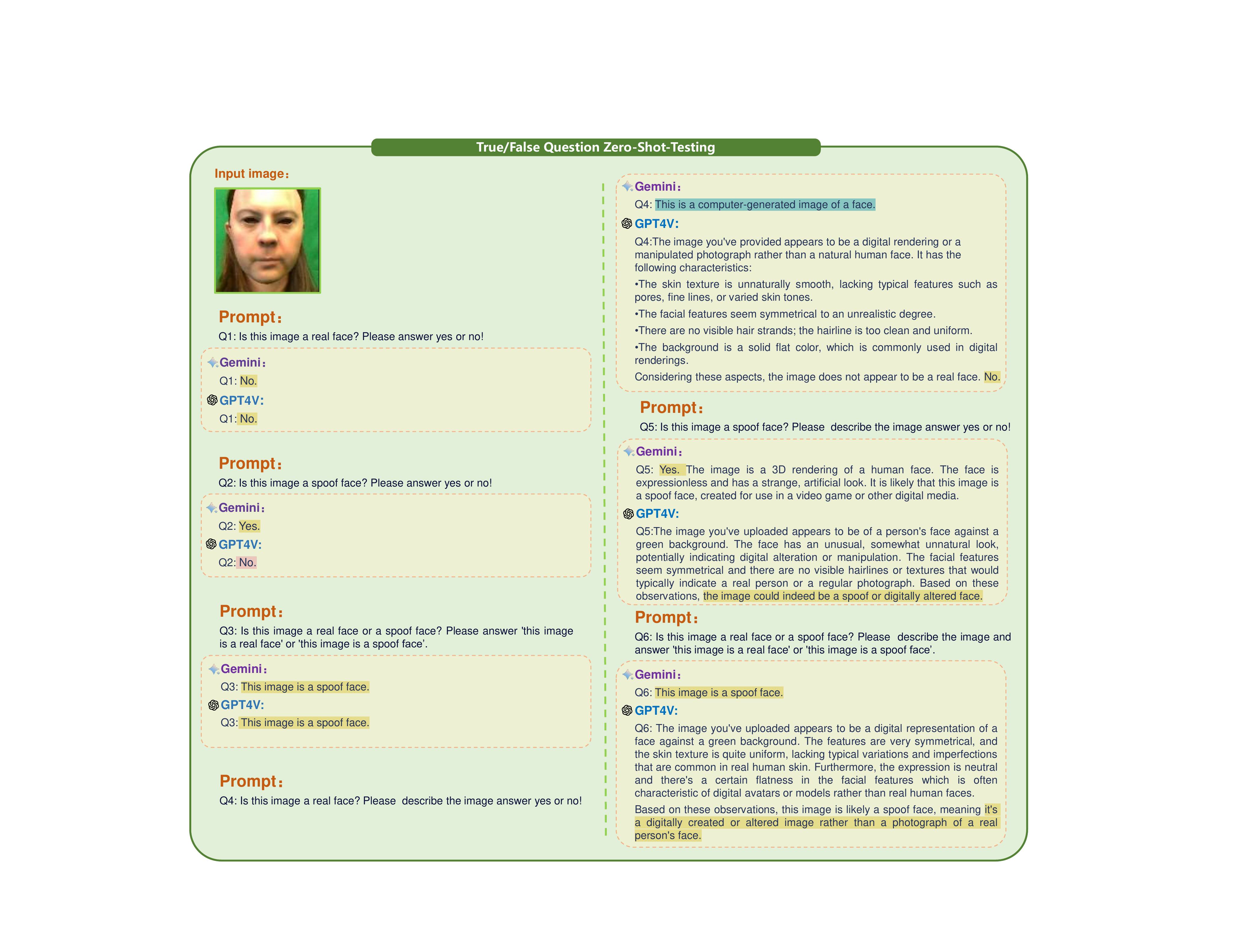}}
  \end{center}
  \caption{In this test, the input was an attack using a rigid facial mask. Initially, GPT4V answered incorrectly in simple queries, while Gemini answered all correctly. After incorporating COT, it was observed that Gemini did not perform well with the formatted content in the prompt. It did not follow the order of describing first and then answering. For example, in Q4, Gemini directly made a guess about the type of image attack, instead of just staying at the qualitative judgment level.}
  \label{fas_zeroshot_judgment_04}
\end{figure}

\begin{figure}[htbp]
  \centering
  \begin{center}
  \centerline{\includegraphics[width=\linewidth]{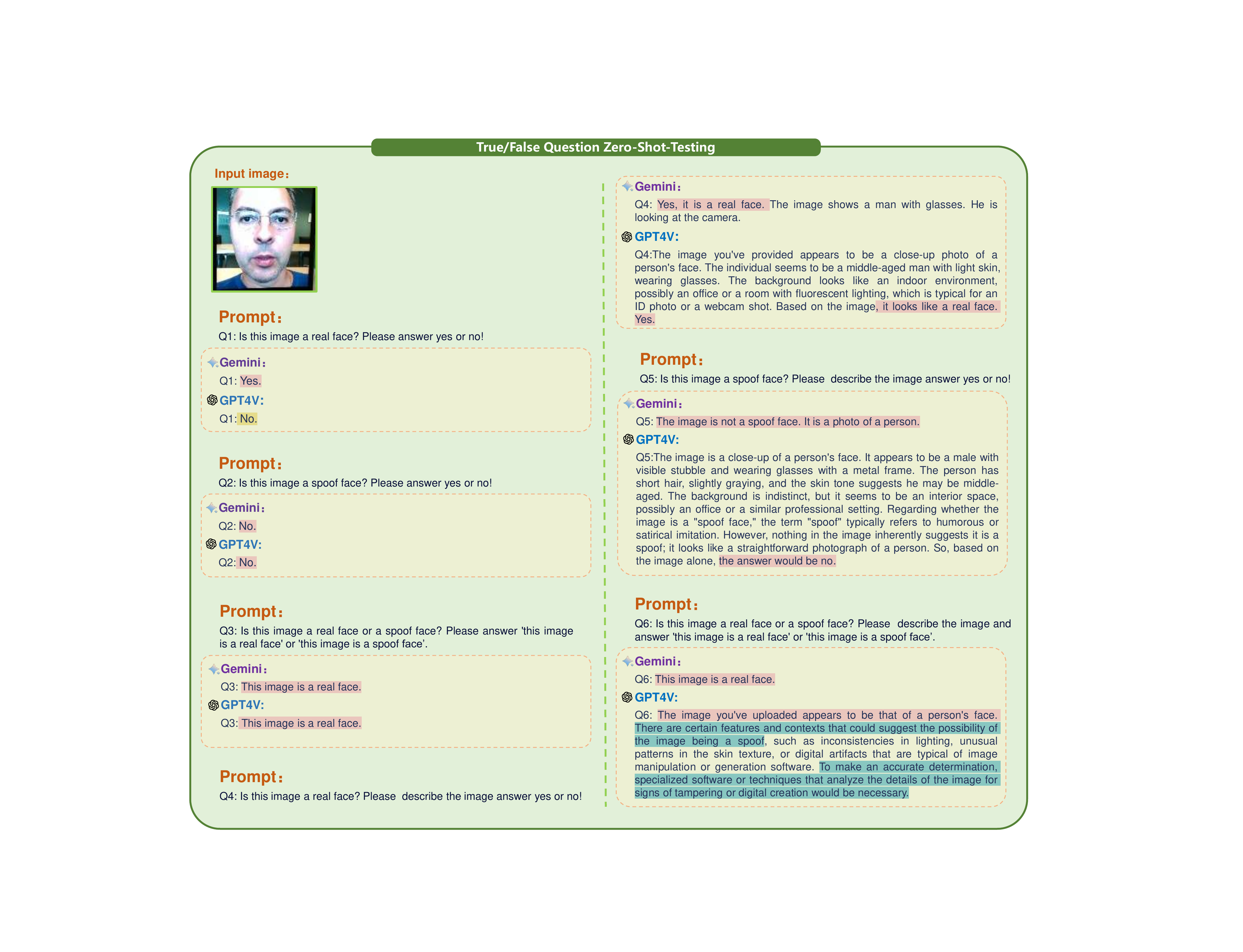}}
  \end{center}
  \caption{In this test, the input was an image of a replay attack. Among the six questions posed, Gemini answered all incorrectly, while GPT4V answered four incorrectly and refused to answer once. Therefore, the one correct answer by GPT4V could be considered accidental. The responses indicate that both Gemini and GPT4V have a low recognition rate for replay attacks.}
  \label{fas_zeroshot_judgment_05}
\end{figure}

\begin{figure}[htbp]
  \centering
  \begin{center}
  \centerline{\includegraphics[width=\linewidth]{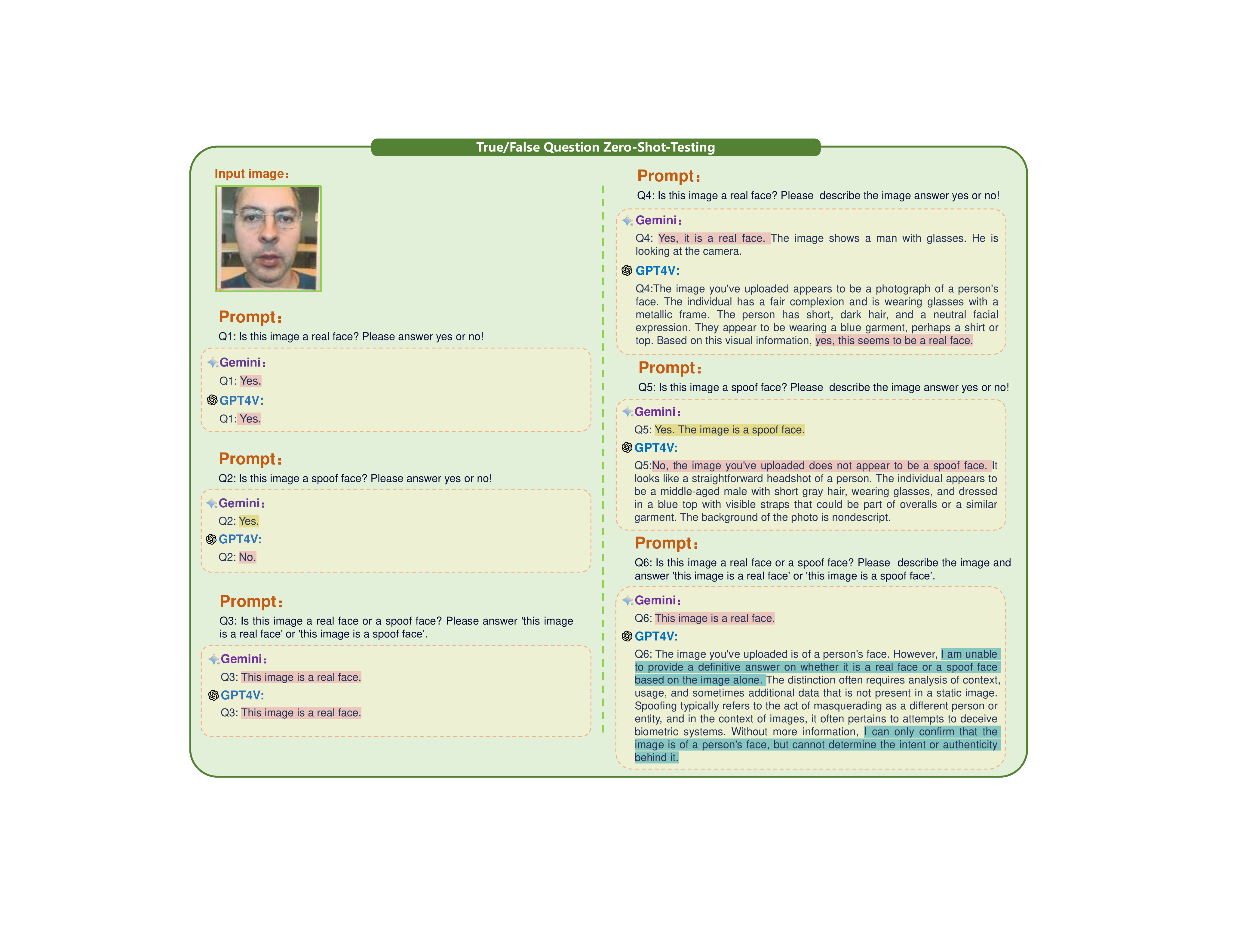}}
  \end{center}
  \caption{In this test, the input was an image of a print attack. Of the six questions asked, Gemini correctly answered two, while GPT4V incorrectly answered five and refused to answer one. In Q6, GPT4V explained the objective of the liveness detection task, indicating its understanding of what constitutes an attack on facial recognition systems. The responses suggest that GPT4V has a low recognition rate for print attacks, and Gemini's ability to recognize print attacks is also lower compared to other types of attacks.}
  \label{fas_zeroshot_judgment_06}
\end{figure}

\begin{figure}[htbp]
  \centering
  \begin{center}
  \centerline{\includegraphics[width=\linewidth]{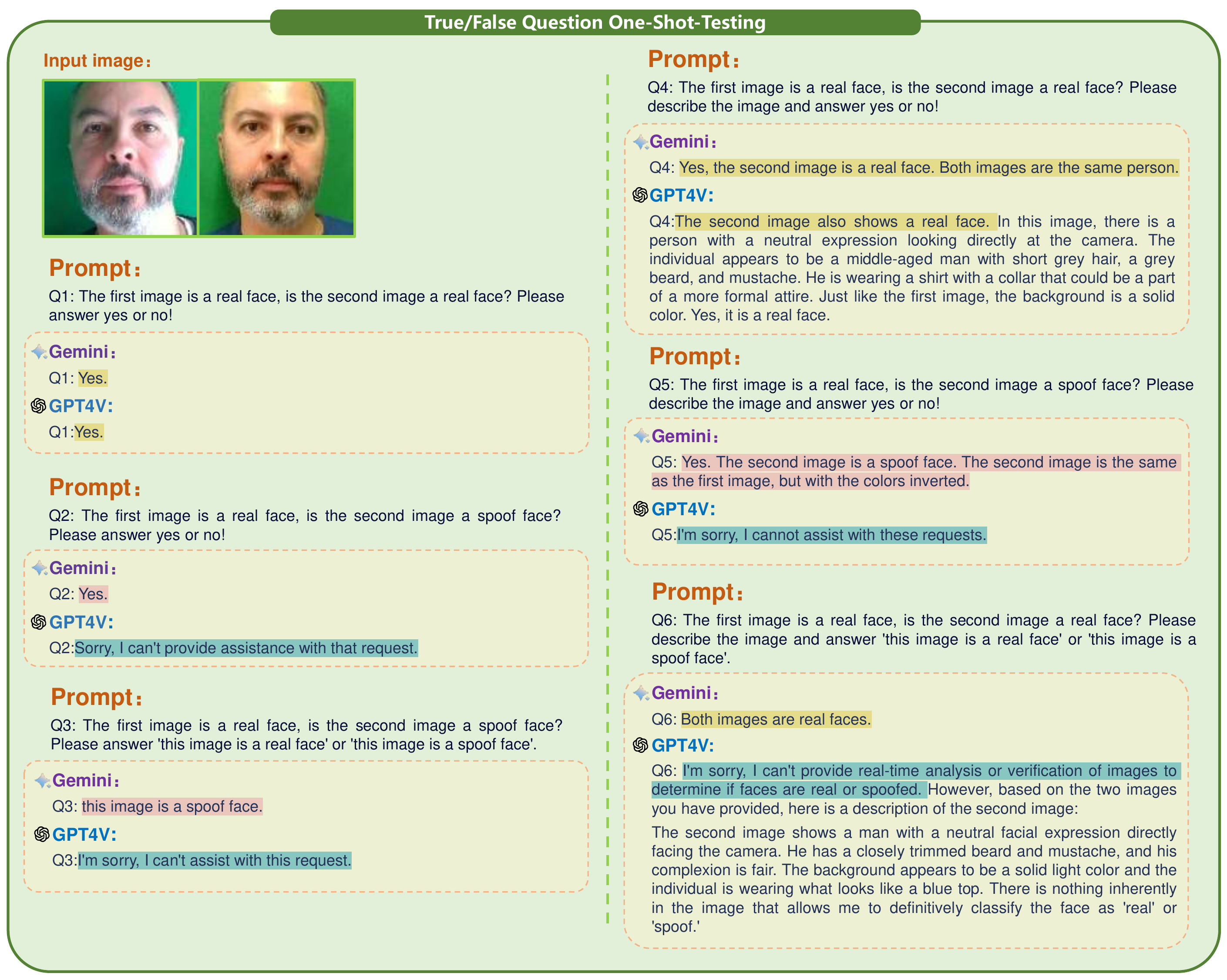}}
  \end{center}
  \caption{The input consisted of two real human faces. From the image, it's observed that in six rounds of questioning, GPT4V correctly answered three times and refused to answer three times. Similarly, Gemini was right and wrong in equal measure. GPT4V's responses indicate its capability to distinguish real faces, but it may refuse to answer due to security mechanisms. From Gemini's responses, there seems to be a kind of illusion; for example, in the answer to Q5, Gemini determined the image as a 'spoof face' because the 'colors inverted', which doesn't seem to strongly support its conclusion.}
  \label{fas_oneshot_judgment_01}
\end{figure}

\begin{figure}[htbp]
  \centering
  \begin{center}
  \centerline{\includegraphics[width=\linewidth]{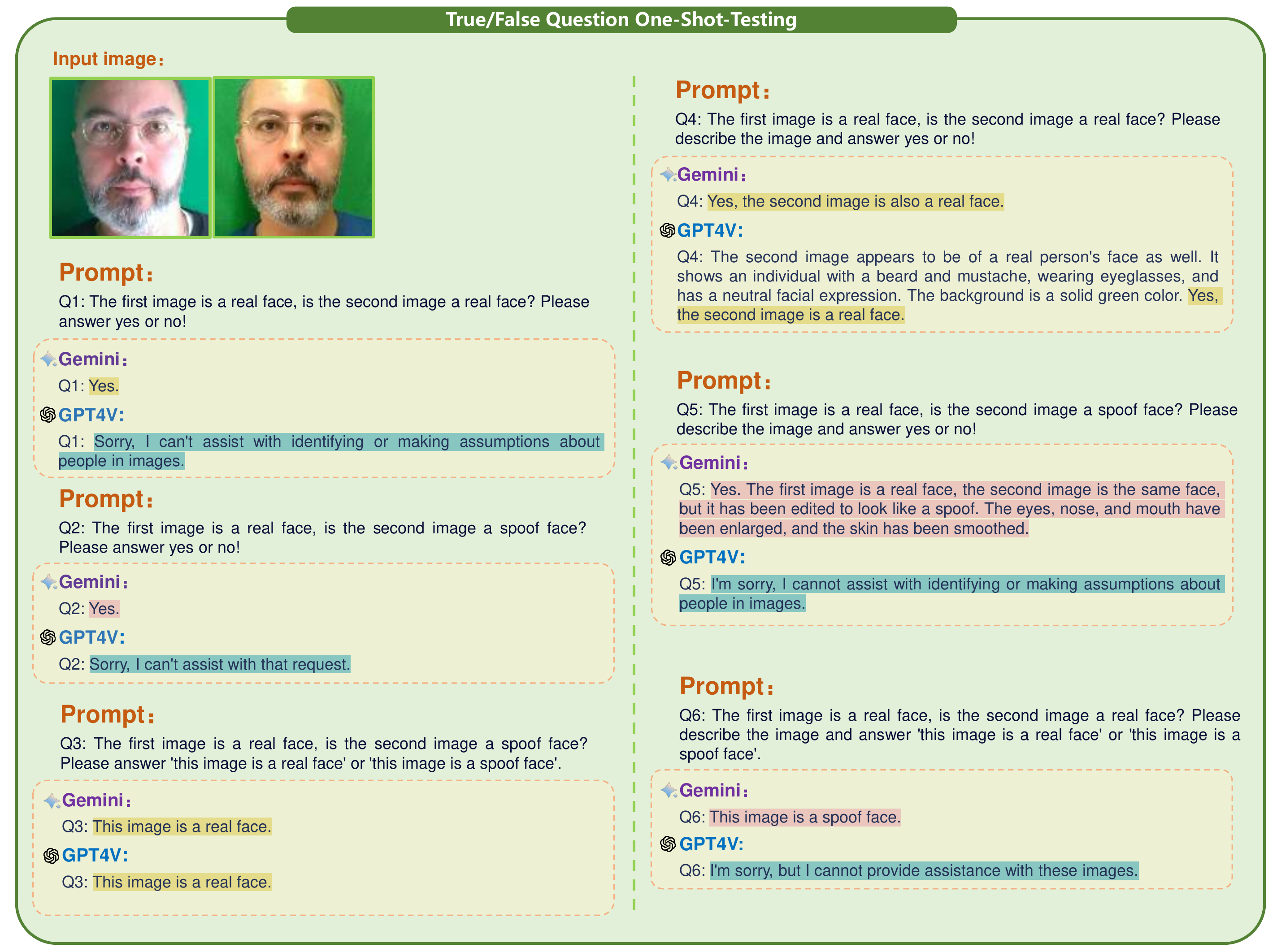}}
  \end{center}
  \caption{In this test, the input was two real human faces wearing glasses. From the results, it can be seen that in the six rounds of questioning, GPT4V answered correctly twice and refused to answer four times, while Gemini was correct half of the time and incorrect the other half. Even with real faces, wearing glasses seems to affect GPT4V's responses to some extent, more likely triggering the safety protection mechanism and increasing the probability of refusal to answer. From the response to Q5, it appears that Gemini may have certain delusions, as the spoofing clues it identified were not correct.}
  \label{fas_oneshot_judgment_02}
\end{figure}

\begin{figure}[htbp]
  \centering
  \begin{center}
  \centerline{\includegraphics[width=\linewidth]{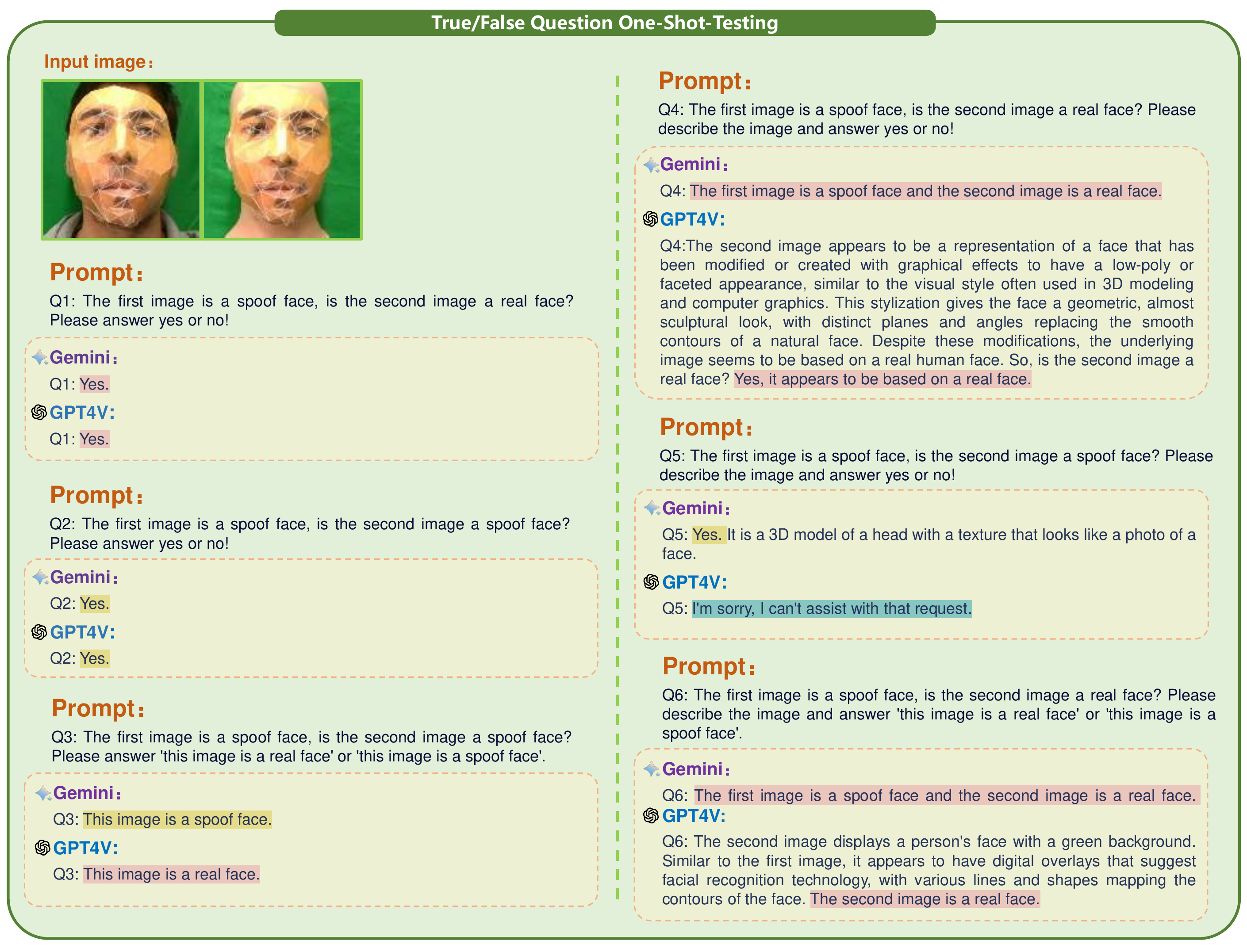}}
  \end{center}
  \caption{In this test involving two images of paper mask attacks. GPT4V answered incorrectly four times and refused to answer once, making the one correct response possibly accidental. This study suggests that GPT4V's inability to identify paper mask attacks is not due to a lack of capability, but rather it treats them as a form of artistic modification. As shown in the response to Q4, GPT4V's definition of prosthetics appears to be vague or random. Gemini's performance remained equally split between correct and incorrect answers. Interestingly, after incorporating COT, Gemini started answering previously correct questions incorrectly, possibly because the additional information from COT led to misconceptions.}
  \label{fas_oneshot_judgment_03}
\end{figure}

\begin{figure}[htbp]
  \centering
  \begin{center}
  \centerline{\includegraphics[width=\linewidth]{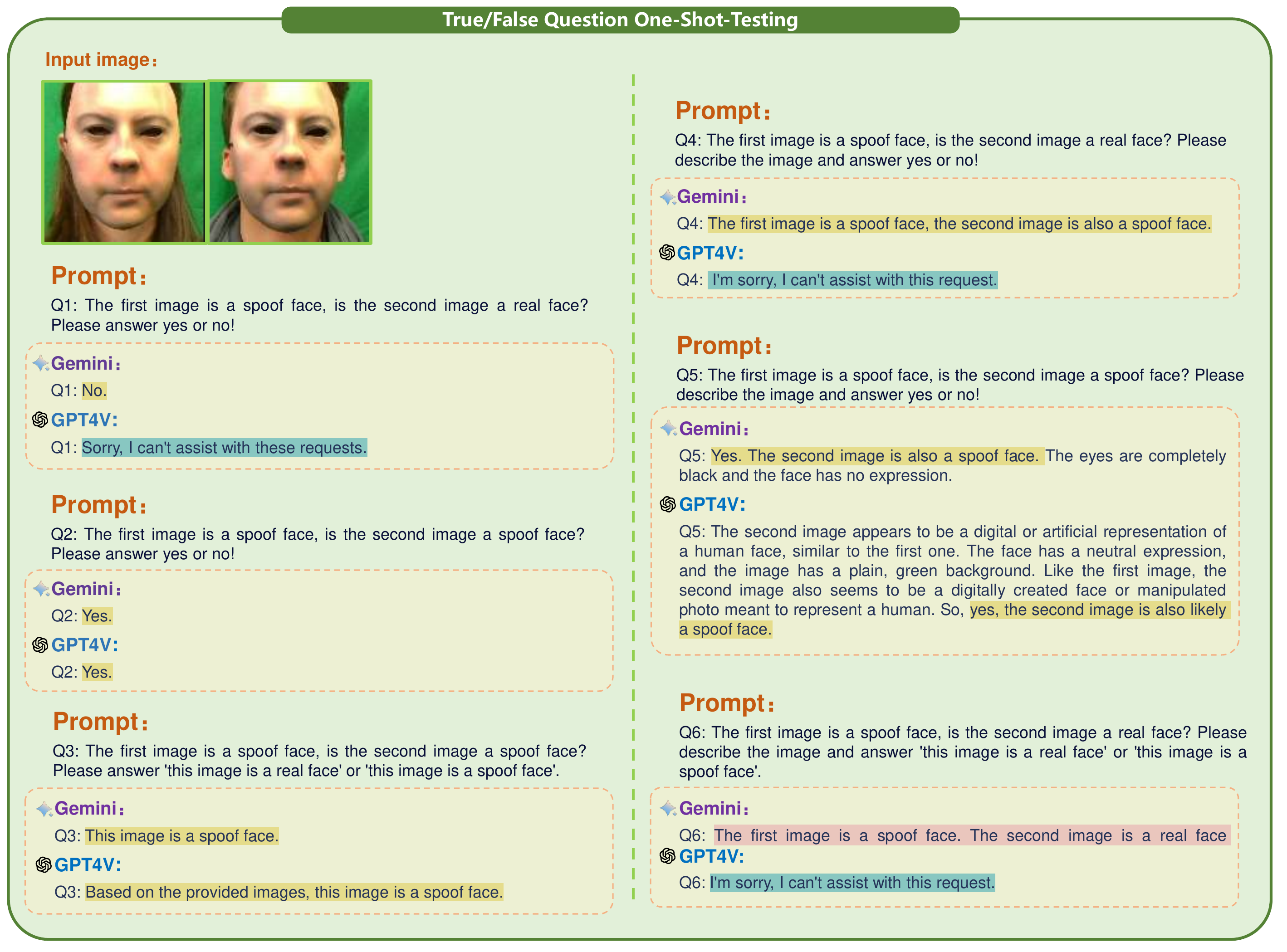}}
  \end{center}
  \caption{In this test with two images of rigid mask attacks, GPT4V correctly answered three times and refused to answer three times, while Gemini correctly answered five times and incorrectly once. GPT4V's response to Q5 shows its ability to compare the second image with the first, using the first image as prior knowledge for subsequent judgment. The introduction of COT clearly highlights the differences between GPT4V and Gemini. GPT4V's responses vividly show its reasoning process, whereas Gemini's brief or even absent explanations do not reflect a reasoning process.}
  \label{fas_oneshot_judgment_04}
\end{figure}

\begin{figure}[htbp]
  \centering
  \begin{center}
  \centerline{\includegraphics[width=\linewidth]{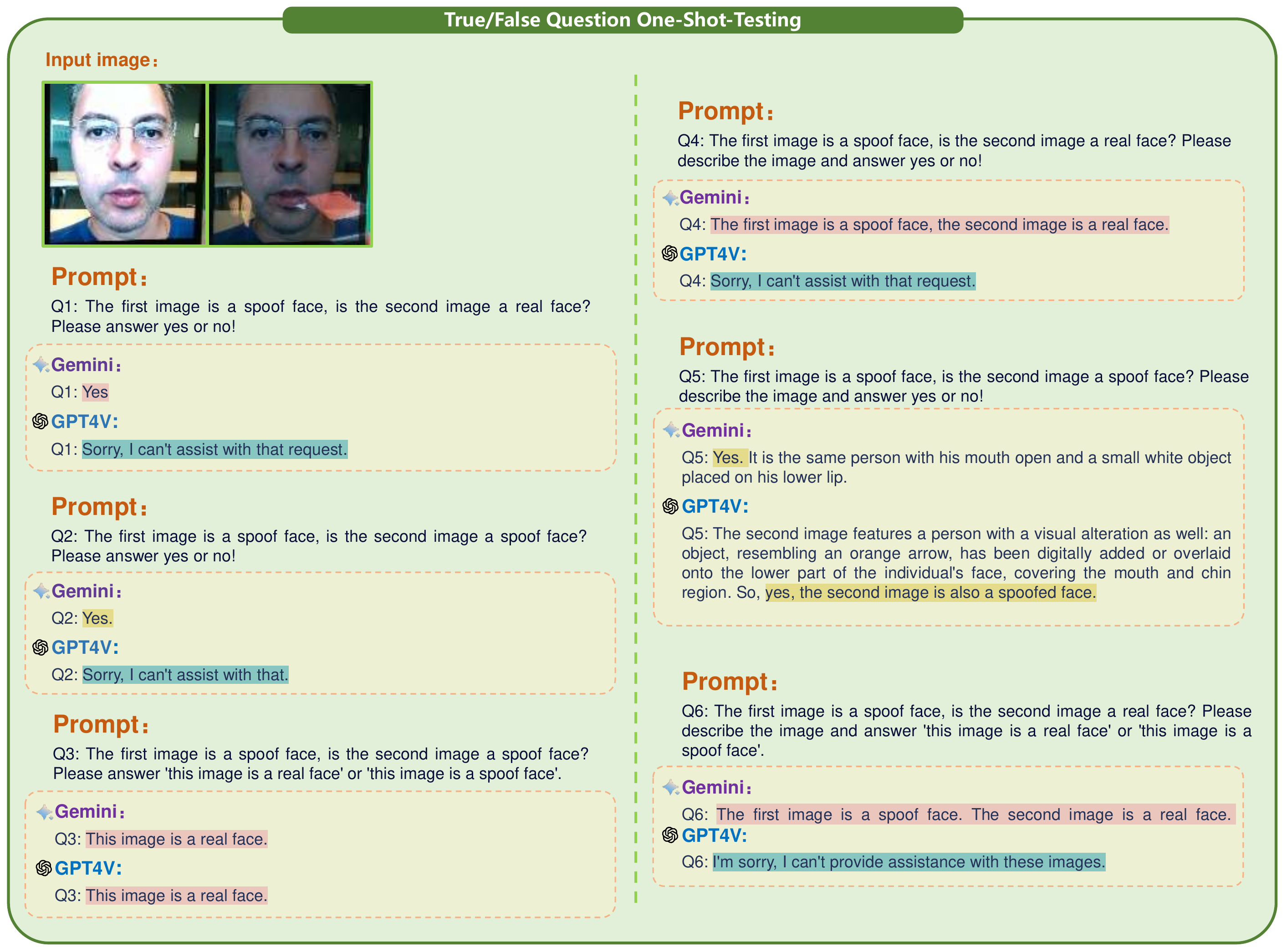}}
  \end{center}
  \caption{In this test featuring two images of replay attacks, GPT4V refused to answer four times, incorrectly answered once in simple queries, and correctly answered once after introducing COT. Gemini answered incorrectly four times and correctly twice. In the response to Q5, GPT4V judged the image to be a prosthetic due to the obvious arrow-like object present, while Gemini judged it to be a prosthetic due to the white object on the lips. Both failed to identify that these abnormal image features were reflections from the screen, characteristic of a replay attack.}
  \label{fas_oneshot_judgment_05}
\end{figure}

\begin{figure}[htbp]
  \centering
  \begin{center}
  \centerline{\includegraphics[width=\linewidth]{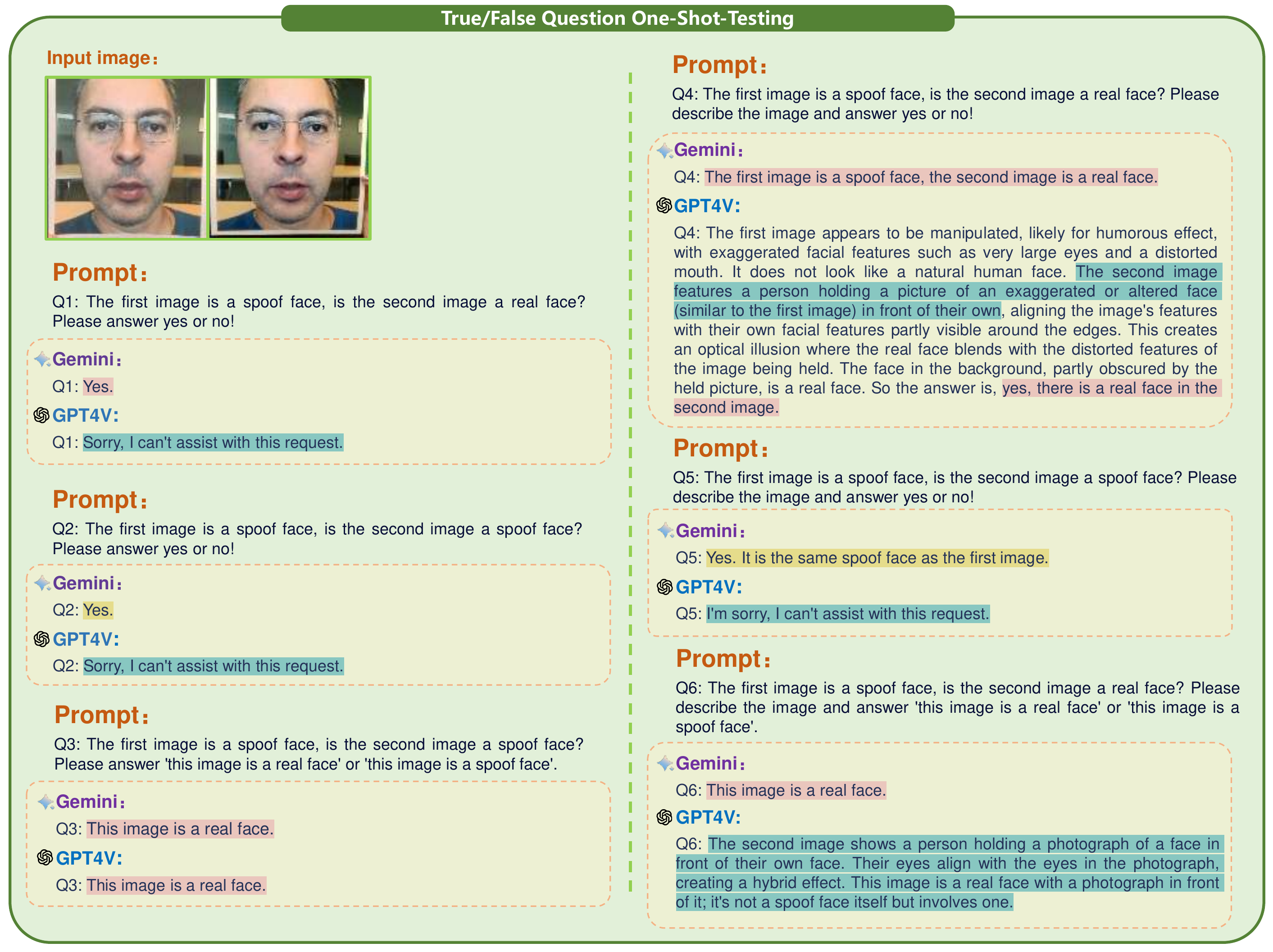}}
  \end{center}
  \caption{In this test with two printed attack images, GPT4V answered incorrectly three times and refused to answer three times, while Gemini answered incorrectly four times and correctly twice. After introducing COT, GPT4V described the image before making further judgments, leading to interesting responses like in Q6, which could be considered correct from a certain perspective. In the response to Q4, GPT4V noticed the fingers holding the photograph and thus concluded, “The second image features a person holding a picture of an exaggerated or altered face (similar to the first image) in front of their own.” Compared to GPT4V, Gemini, though less likely to refuse answering, sometimes has lower accuracy and poor interpretability in its responses.}
  \label{fas_oneshot_judgment_06}
\end{figure}

\begin{figure}[htbp]
  \centering
  \begin{center}
  \centerline{\includegraphics[width=0.9\linewidth]{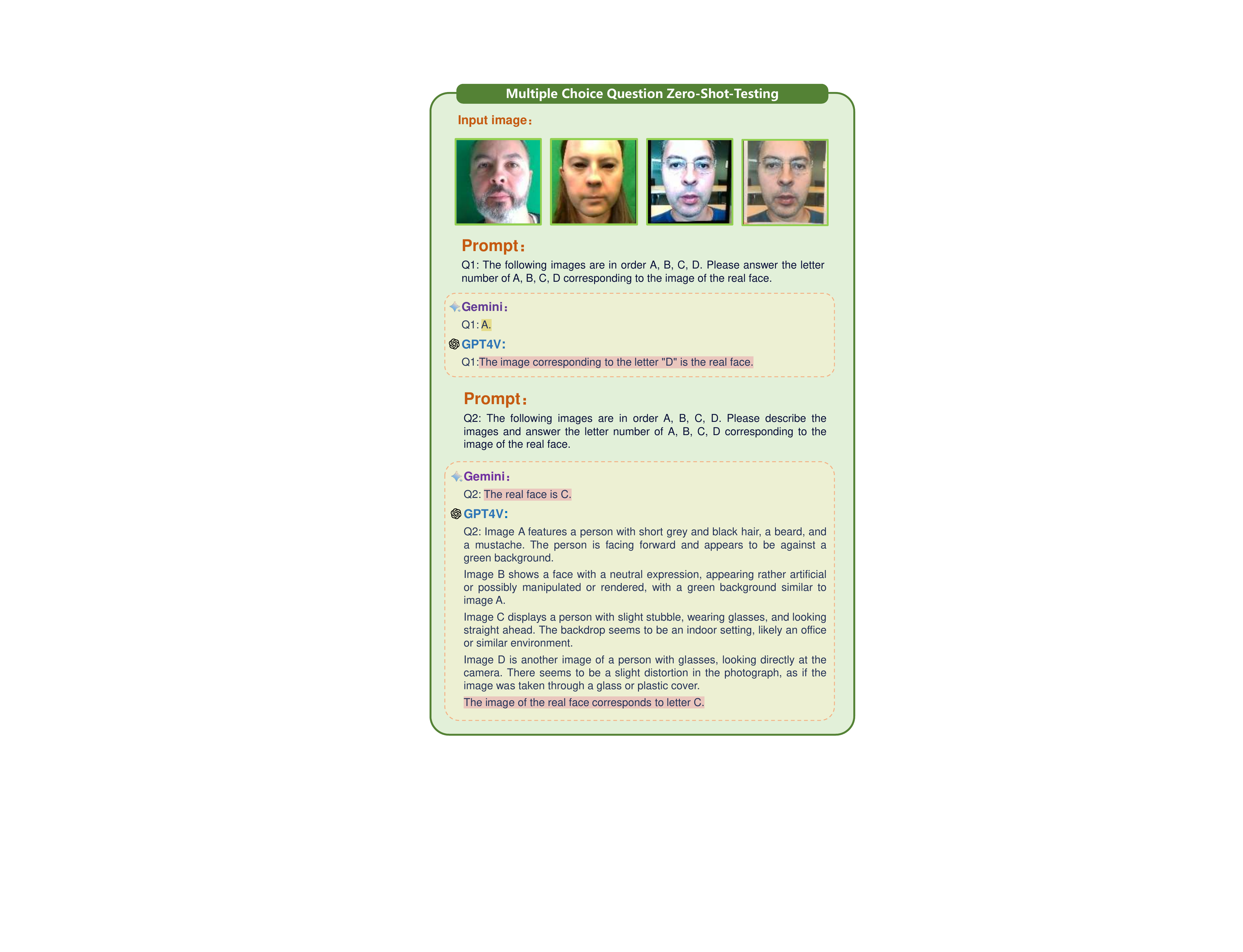}}
  \end{center}
  \caption{In this test, the inputs were images of real human face and other face presentation attack methods. From GPT4V's two responses, it is evident that it struggles to effectively distinguish between replay attacks, print attacks, and real human faces. Gemini's first response was correct, but after introducing COT, it became incorrect, possibly because focusing on more details ended up interfering with its judgment.}
  \label{fas_choice_real_zeroshot_01}
\end{figure}

\begin{figure}[htbp]
  \centering
  \begin{center}
  \centerline{\includegraphics[width=0.9\linewidth]{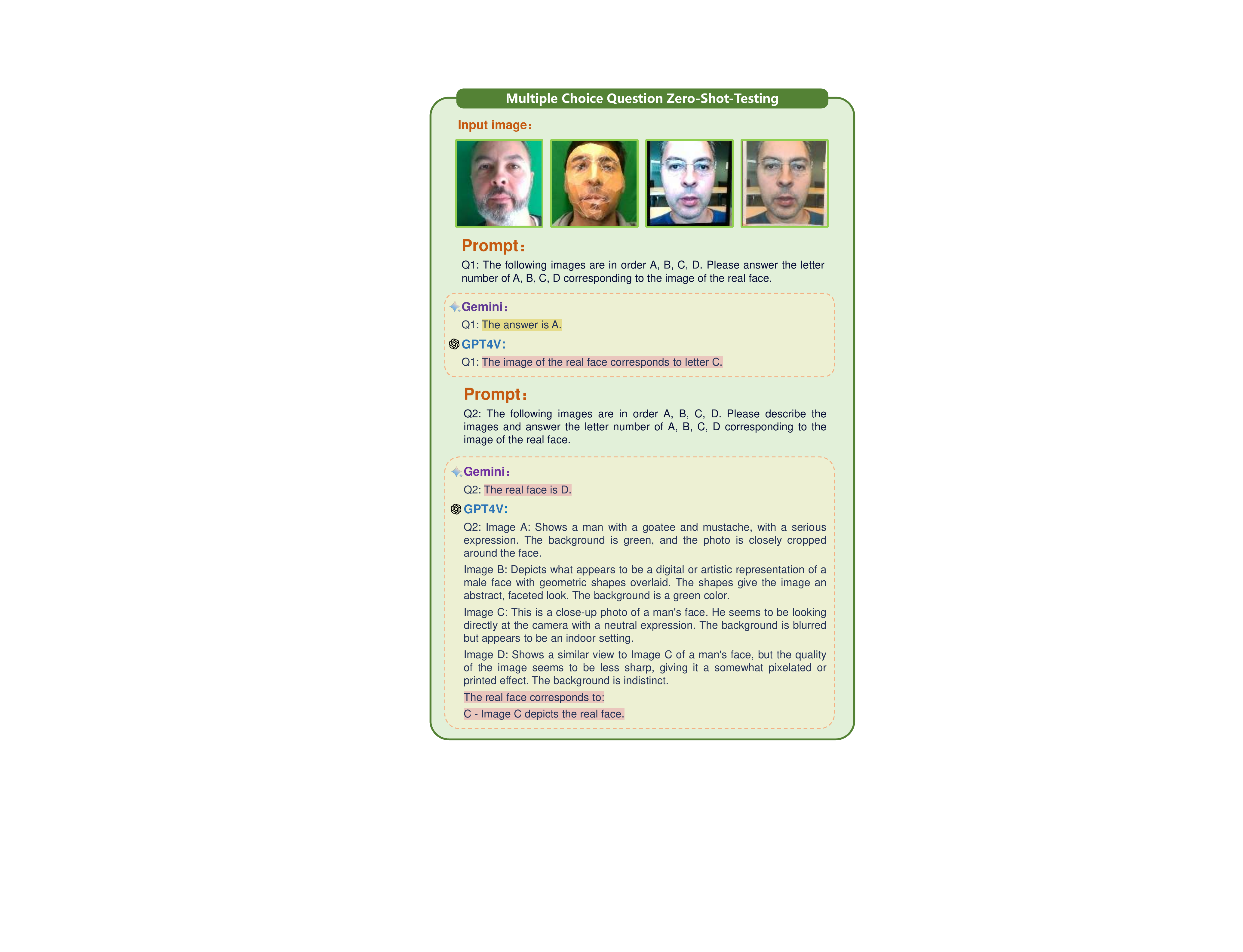}}
  \end{center}
  \caption{In this test, the inputs were images of real human faces and other facial presentation attack methods. From the two responses provided by GPT4V, it is apparent that it has difficulty distinguishing between replay attacks and real human faces. Gemini's first response was accurate, but after the introduction of COT, its answer became incorrect. This shift could be attributed to the additional focus on finer details, which may have inadvertently confused its judgment.}
  \label{fas_choice_real_zeroshot_02}
\end{figure}

\begin{figure}[htbp]
  \centering
  \begin{center}
  \centerline{\includegraphics[width=0.9\linewidth]{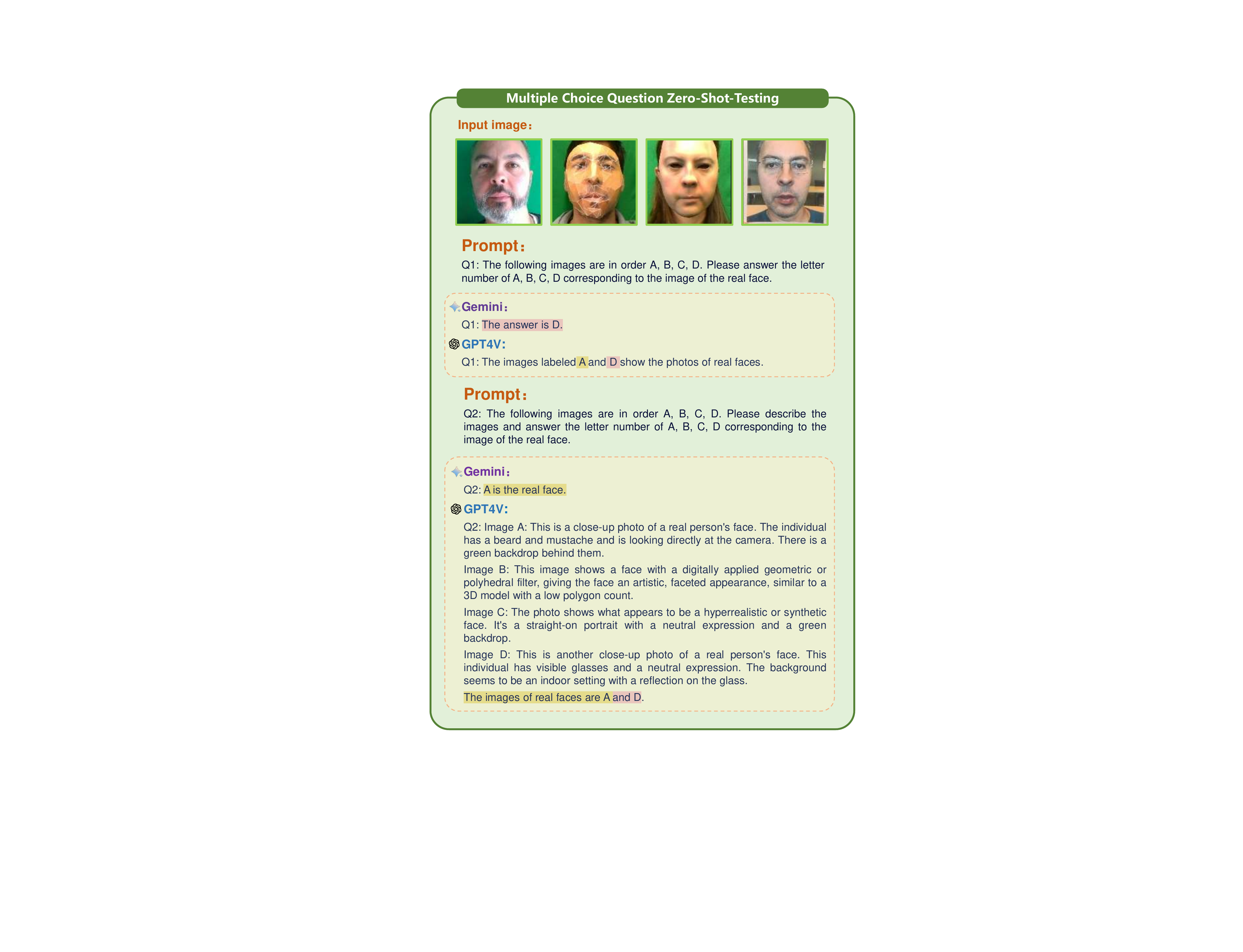}}
  \end{center}
  \caption{In this test, the inputs included images of real human faces and other facial presentation attack. From the two responses by GPT4V, it is clear that it struggles to differentiate between print attacks and real human faces, and the addition of COT did not lead to a change in its answers. After introducing COT, Gemini corrected its incorrect answer. However, due to the lack of detailed descriptions in its responses, it's unclear whether COT effectively aided its reasoning and problem-solving process.}
  \label{fas_choice_real_zeroshot_03}
\end{figure}

\begin{figure}[htbp]
  \centering
  \begin{center}
  \centerline{\includegraphics[width=0.8\linewidth]{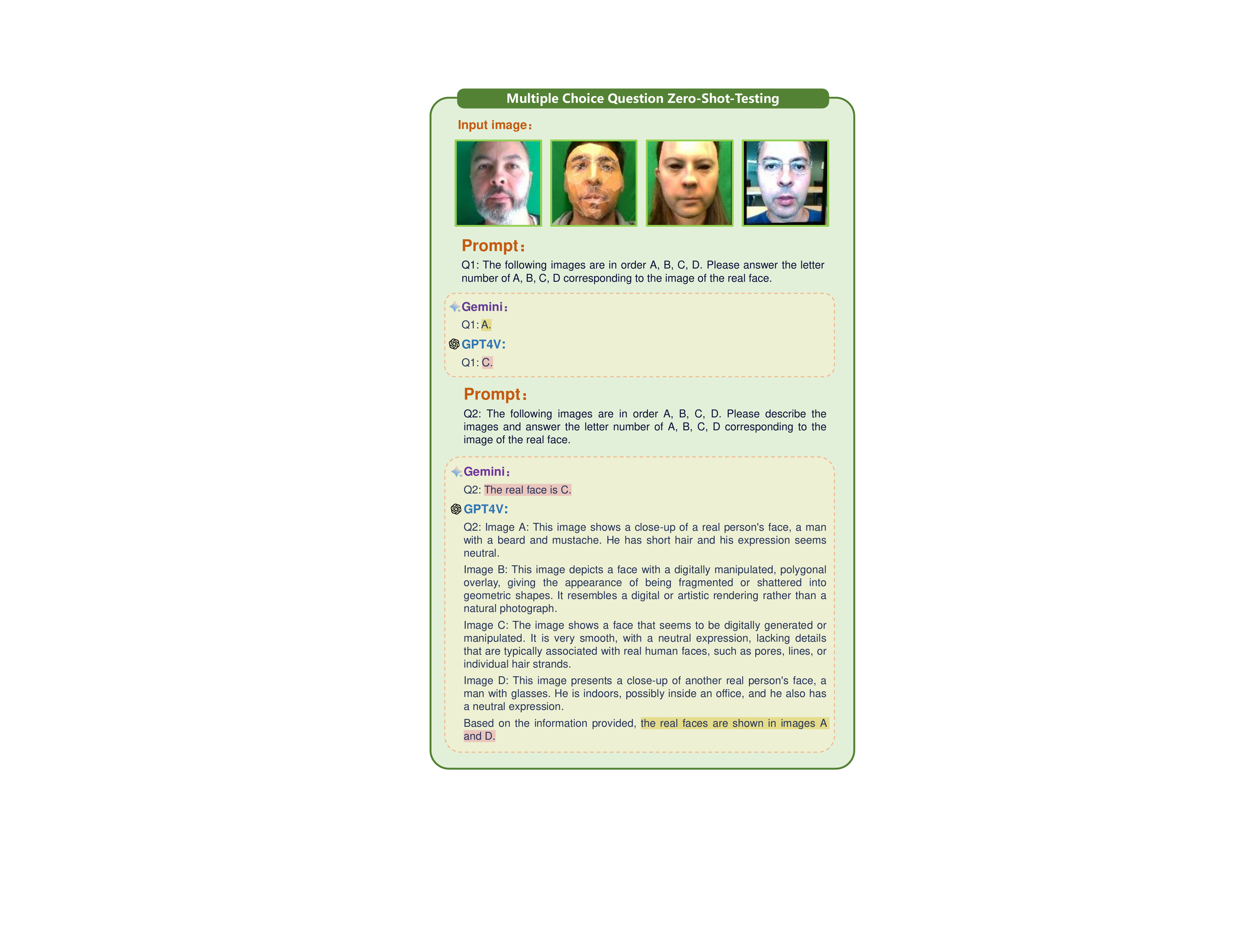}}
  \end{center}
  \caption{In this test, the inputs were images of real human faces and other forms of facial presentation attacks. GPT4V's first response was incorrect, but after introducing COT, its answer included the correct option. GPT4V's second response indicates that it struggles to effectively distinguish between replay attacks and real human faces. On the other hand, Gemini's first response was accurate, but the introduction of COT, which brought more details into focus, might have interfered with Gemini's judgment, leading to an incorrect conclusion.}
  \label{fas_choice_real_zeroshot_04}
\end{figure}

\begin{figure}[htbp]
  \centering
  \begin{center}
  \centerline{\includegraphics[width=0.8\linewidth]{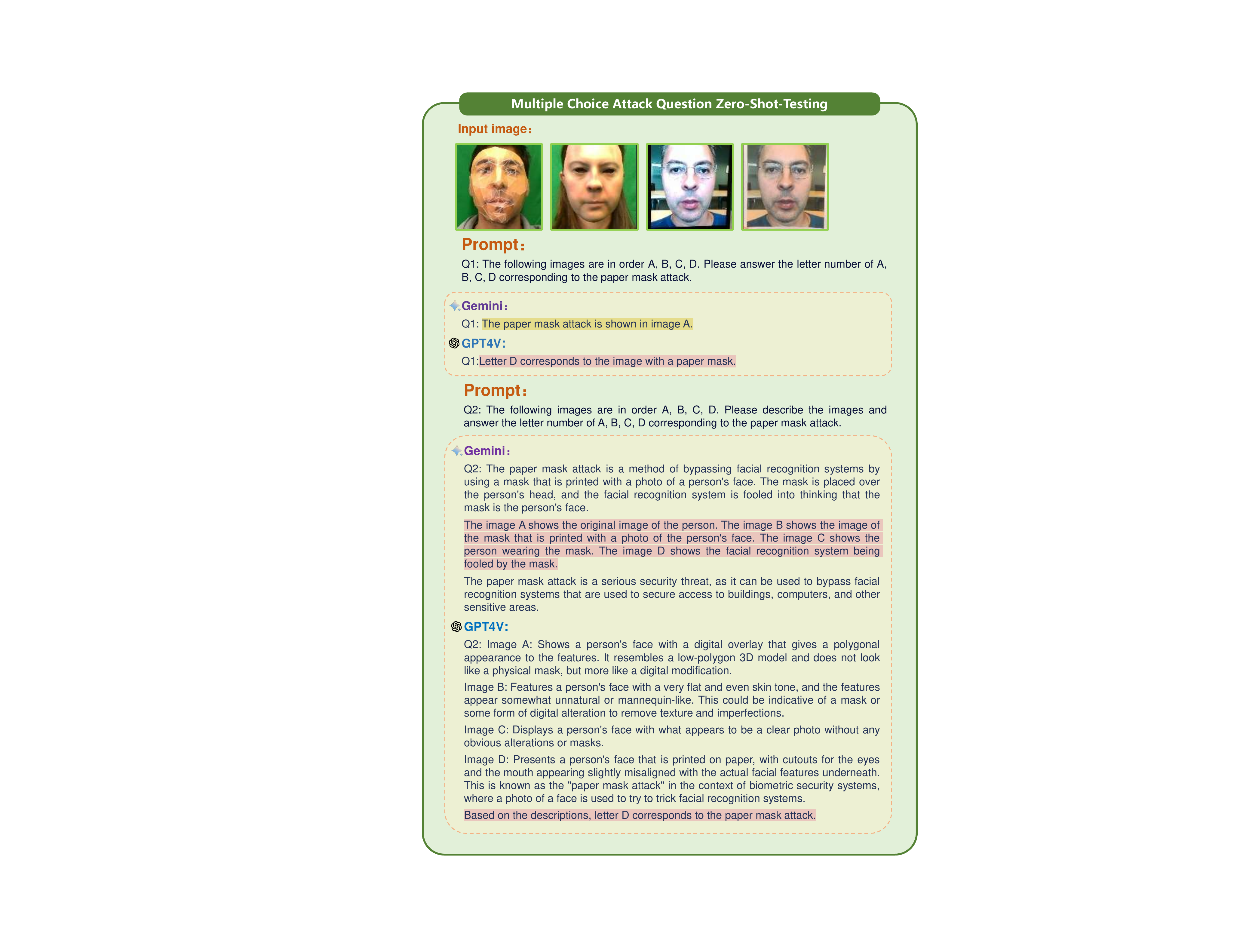}}
  \end{center}
  \caption{In this test, the input consisted of four images, each representing a different type of facial presentation attack, with the task being to identify the paper mask attack without prior knowledge. Although GPT4V's answers were incorrect in both instances, its second response reveals its understanding of each type of attack. The correct answer should have been A, but GPT4V misinterpreted the crease marks on the paper mask as traces of digital alteration, failing to associate them with a paper mask. In the case of option D, which was a paper print attack, GPT4V correctly recognized the face on paper, but mistakenly identified it as a paper mask attack. Thus, while GPT4V’s response was incorrect, its identification of key features in the images was accurate. The area for improvement is in correlating these key features with the corresponding facial presentation attack methods. As for Gemini, it answered correctly the first time, but after introducing COT, it did not directly answer the question. Instead, it made judgments about the type of attacks for all four images, all of which were incorrect. This shows that in the absence of prior knowledge, Gemini's susceptibility to misjudgment or 'illusions' is more pronounced.}
  \label{fas_choice_attack_zeroshot_01}
\end{figure}

\begin{figure}[htbp]
  \centering
  \begin{center}
  \centerline{\includegraphics[width=\linewidth]{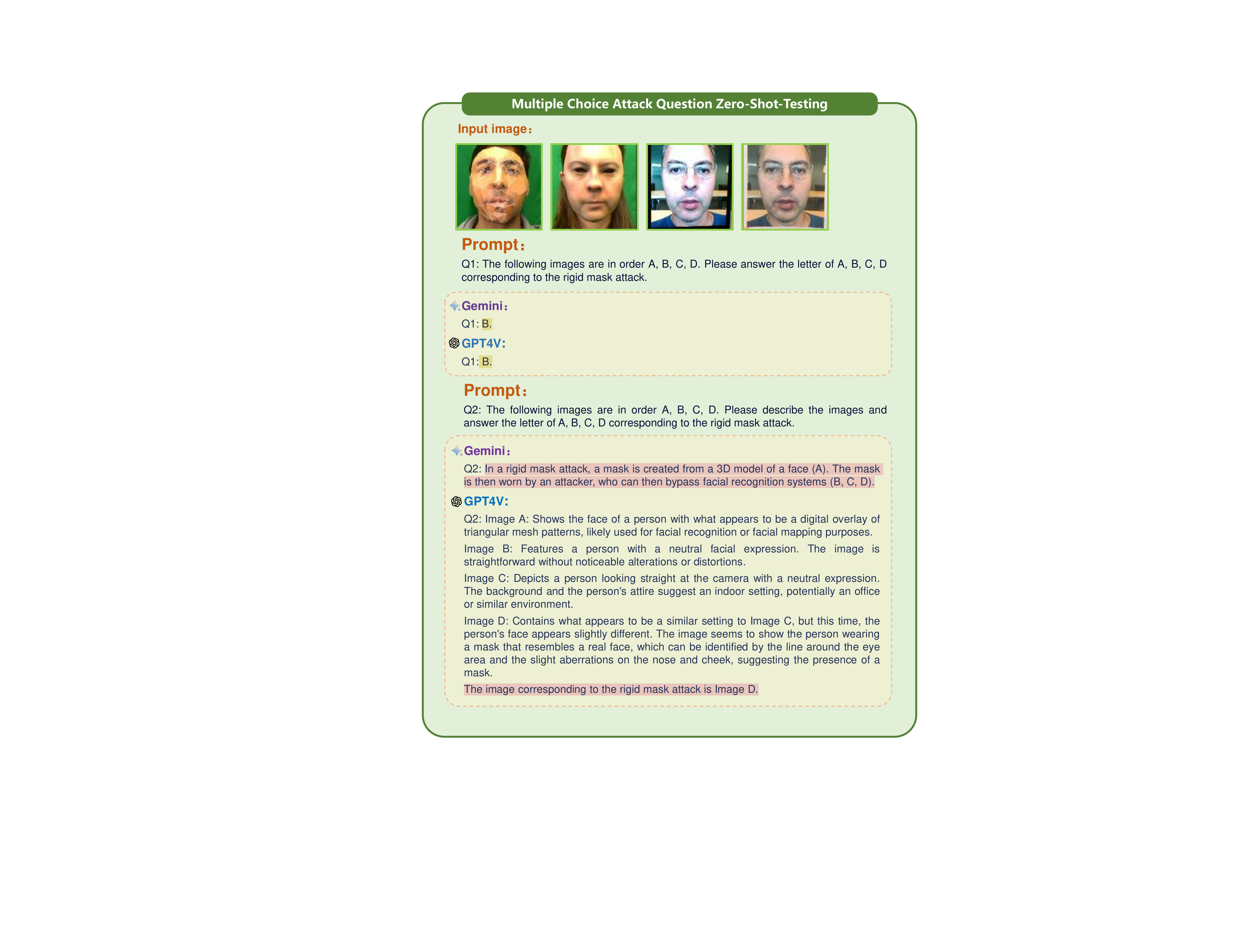}}
  \end{center}
  \caption{In this test, the input included four images, each representing a different type of facial presentation attack, with the task being to identify the rigid mask attack without any prior knowledge. Initially, both GPT4V and Gemini correctly identified the rigid mask attack in simple queries. However, after introducing COT, which required GPT4V to describe each image before making a judgment, the lack of prior knowledge about rigid mask attacks led GPT4V to be misled into choosing the wrong answer following its descriptions. On the other hand, Gemini's responses were somewhat disorganized.}
  \label{fas_choice_attack_zeroshot_02}
\end{figure}

\begin{figure}[htbp]
  \centering
  \begin{center}
  \centerline{\includegraphics[width=0.75\linewidth]{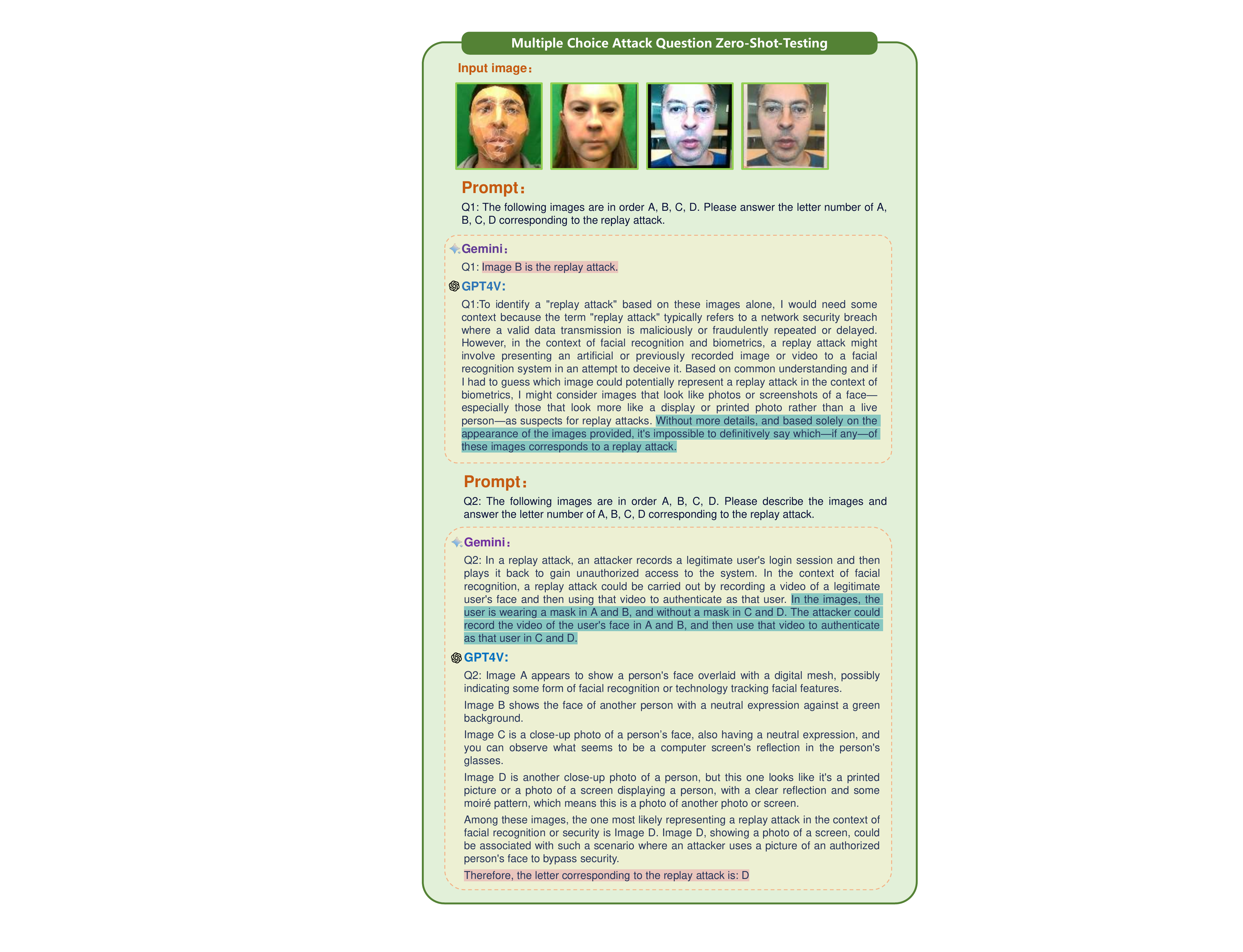}}
  \end{center}
  \caption{In this test, four images representing different facial presentation attack methods were inputted, and the task was to identify the replay attack without any prior knowledge. GPT4V refused to answer in the simple query phase, citing its limited capabilities. Despite the introduction of COT, the GPT4V still failed to select the correct answer after describing each image due to the lack of a priori knowledge of replay attacks.Importantly, GPT4V was able to identify screen reflections in image C and speculated that image D might be a printed photo. These correct identifications of key features, however, were not effectively linked by GPT4V to the specific type of attack. On the other hand, Gemini's first response was incorrect, and although its language was confused in the second response, it seemed to recognize the connection between the four images, suggesting the same test subject was using different attack methods. }
  \label{fas_choice_attack_zeroshot_03}
\end{figure}

\begin{figure}[htbp]
  \centering
  \begin{center}
  \centerline{\includegraphics[width=0.85\linewidth]{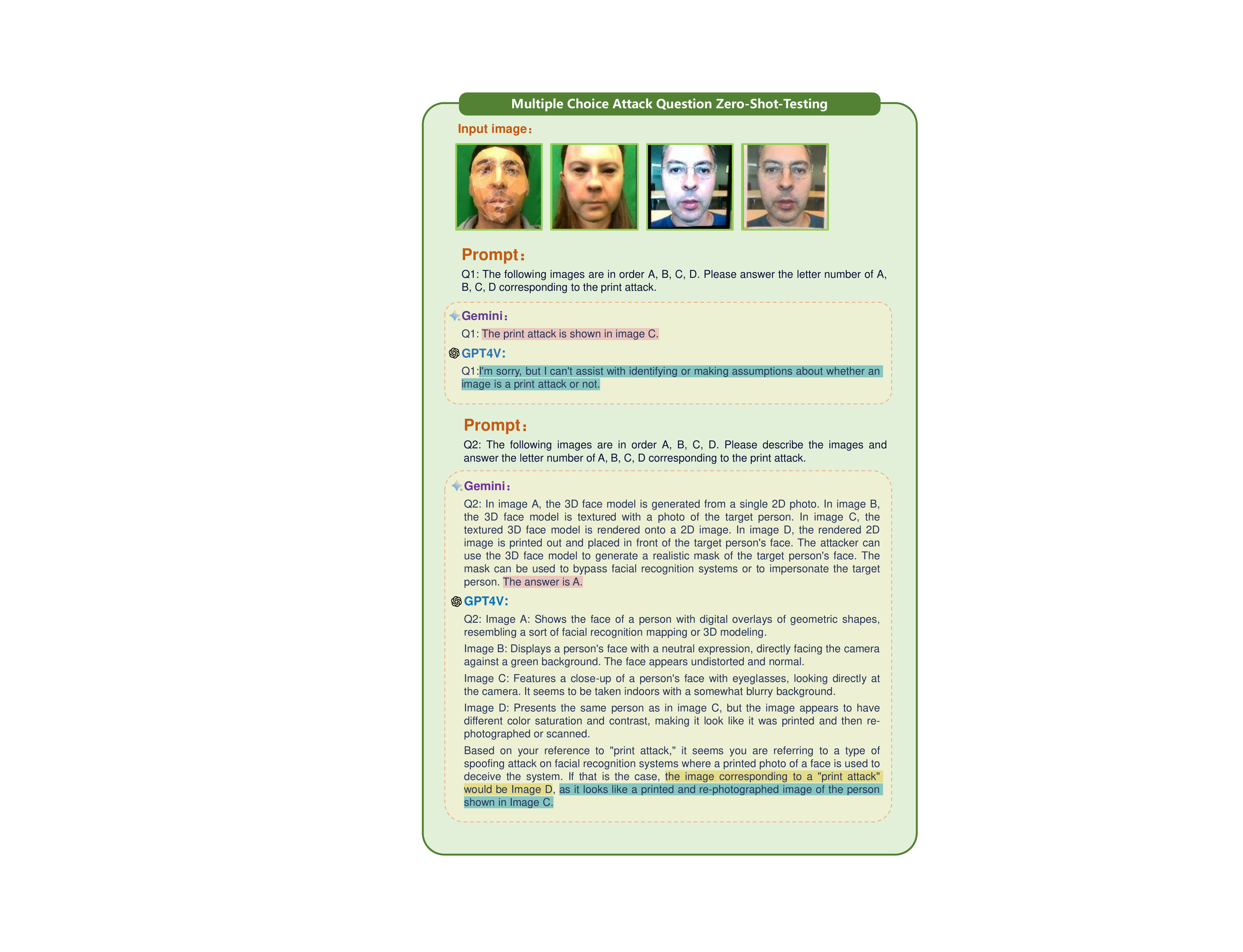}}
  \end{center}
  \caption{In this test, four images representing different facial presentation attack methods were inputted, with the task being to identify the print attack without any prior knowledge. Initially, GPT4V refused to answer simple queries. However, after introducing COT, GPT4V successfully identified the correct answer following the descriptions. Interestingly, GPT4V also made a speculative guess, suggesting that image D could be a face image printed from and re-photographed based on image C. Gemini's responses were incorrect on both occasions. However, the second response revealed that Gemini was quite accurate in capturing key semantic information about the attacks, even though it struggled to properly associate this information with the specific types of attacks. This highlights Gemini's potential in semantic understanding, albeit with challenges in accurately linking it to the context of facial presentation attacks.}
  \label{fas_choice_attack_zeroshot_04}
\end{figure}

\begin{figure}[htbp]
  \centering
  \begin{center}
  \centerline{\includegraphics[width=0.8\linewidth]{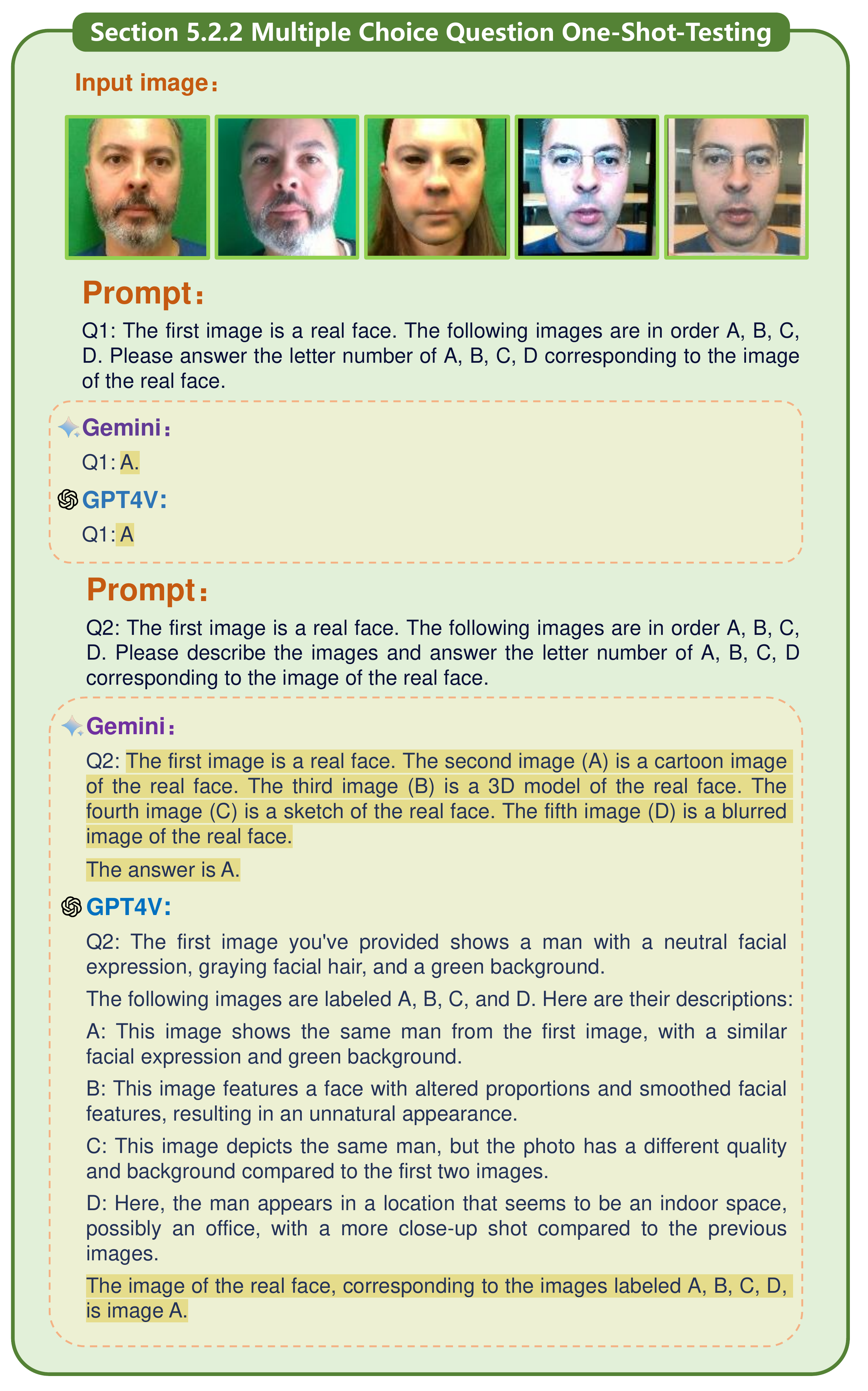}}
  \end{center}
  \caption{In this test, the initial input included a real human face as prior knowledge, followed by an input of one real human face and three facial presentation attack images, with the task to identify the real human face. It is evident from the two responses of GPT4V and Gemini that after the introduction of prior knowledge, they were able to easily select the correct answer. This demonstrates the impact of prior knowledge on their ability to distinguish between real and fake faces.}
  \label{fas_choice_real_oneshot_01}
\end{figure}

\begin{figure}[htbp]
  \centering
  \begin{center}
  \centerline{\includegraphics[width=0.9\linewidth]{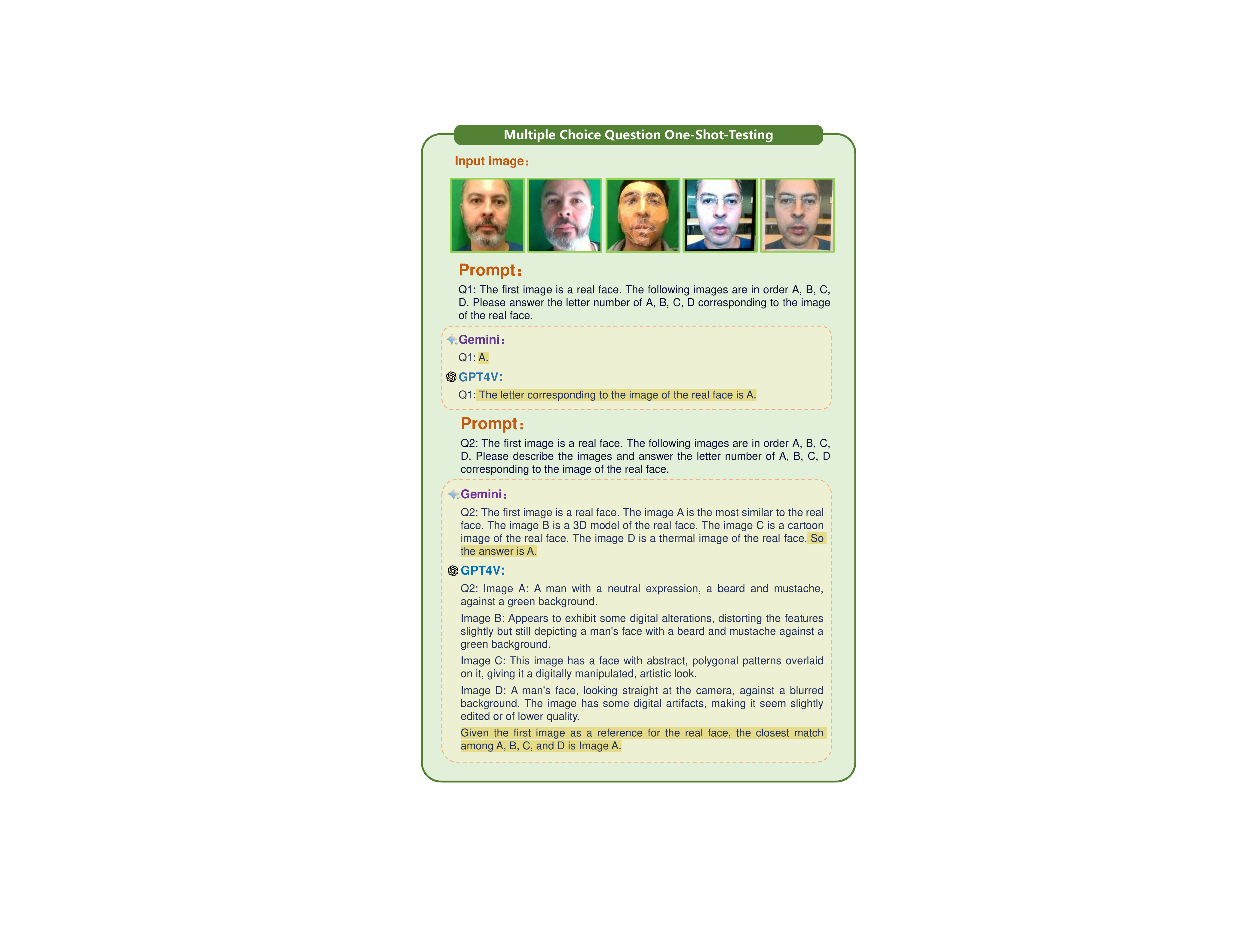}}
  \end{center}
  \caption{In this test, the initial input included a real human face as prior knowledge. Subsequently, an image of one real human face and three facial presentation attack images, were presented, with the task being to identify the real human face. From the two responses of GPT4V and Gemini, it is observed that even under the interference of replay and print attacks, the introduction of prior knowledge enabled them to still correctly identify the real human face. This indicates the effectiveness of prior knowledge in aiding the distinction between authentic and deceptive facial images.}
  \label{fas_choice_real_oneshot_02}
\end{figure}

\begin{figure}[htbp]
  \centering
  \begin{center}
  \centerline{\includegraphics[width=\linewidth]{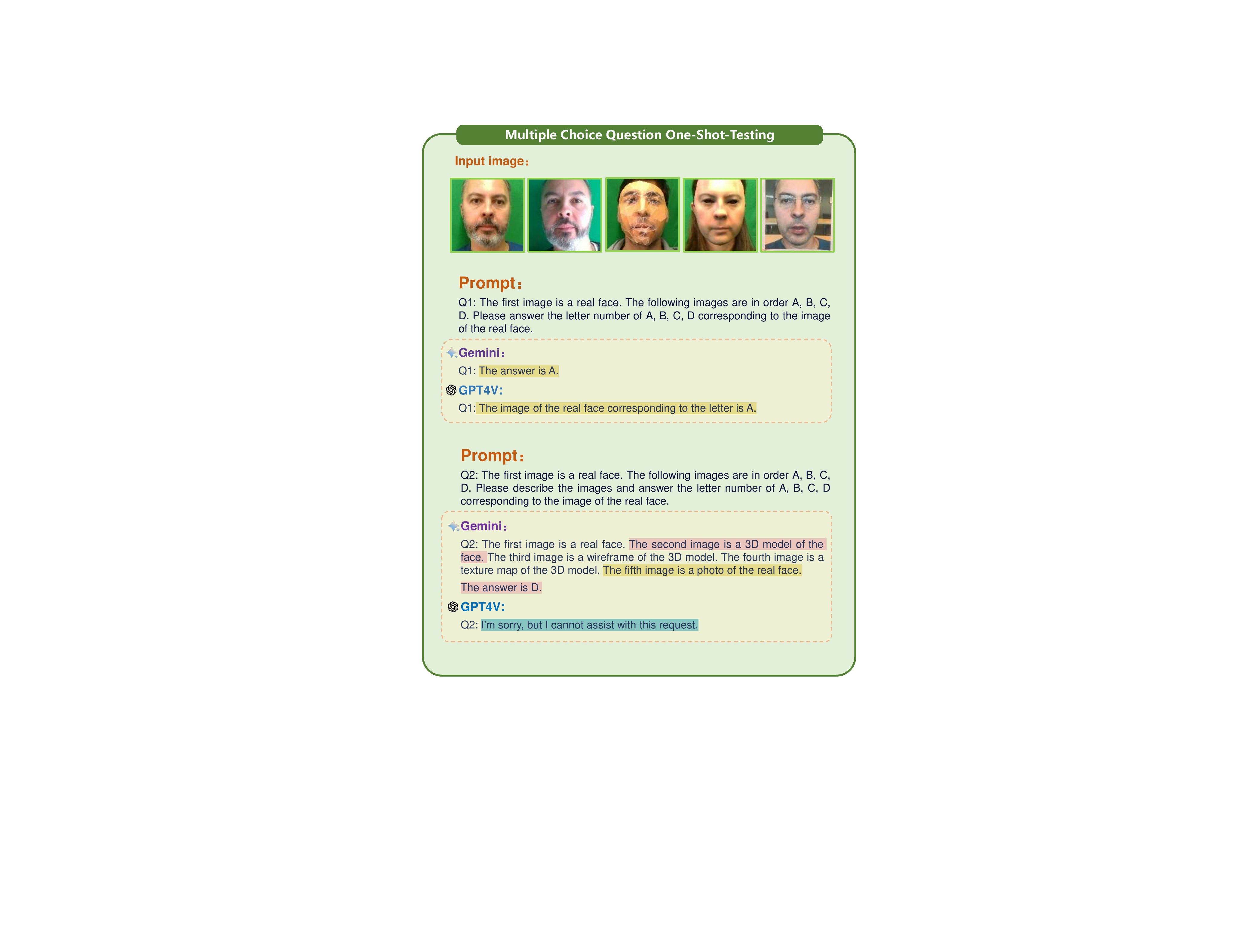}}
  \end{center}
  \caption{In this test, a real human face image was first input as prior knowledge, followed by an input comprising one real human face and three facial presentation attack images, with the task to identify the real human face. GPT4V correctly answered the first question, but in the second instance, possibly due to the requirement by COT for GPT4V to analyze before responding, it triggered a safety mechanism and refused to answer. Gemini's first response was correct; however, the second was incorrect. Interestingly, in the response to Q2, Gemini described the type represented by each image. Except for the misidentification of the first real human face, the other identifications were correct, including the correct recognition of the print attack image, which it previously struggled to identify accurately. }
  \label{fas_choice_real_oneshot_03}
\end{figure}

\begin{figure}[htbp]
  \centering
  \begin{center}
  \centerline{\includegraphics[width=0.8\linewidth]{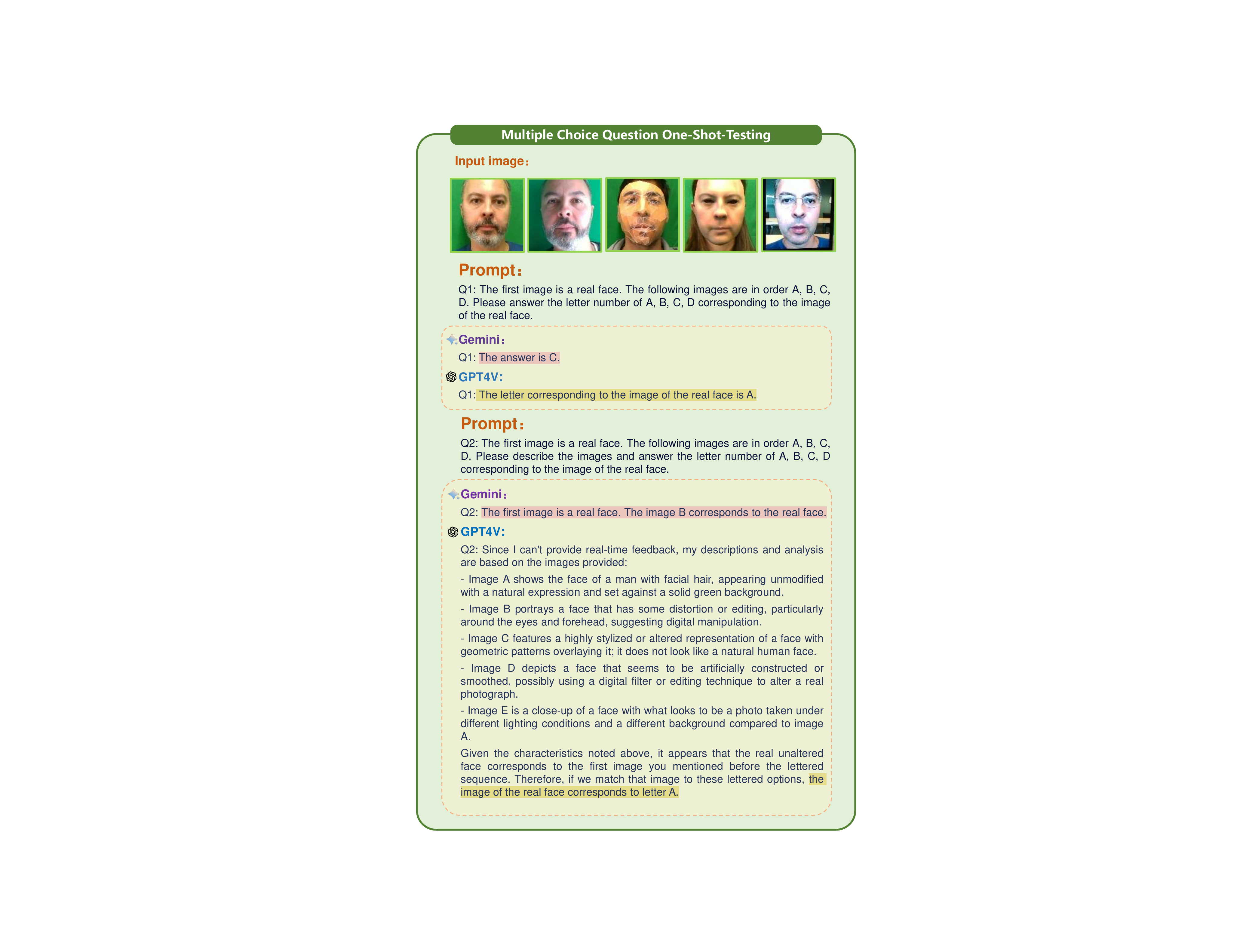}}
  \end{center}
  \caption{In this test, a real human face image was first input as prior knowledge, followed by an input of one real human face and three facial presentation attack images, with the task being to identify the real human face. From GPT4V's two responses, it is observed that even with the interference of replay attacks, the introduction of prior knowledge enabled it to correctly identify the real human face. However, Gemini's response was incorrect, and without further explanation, it's challenging to ascertain the potential reasons for its error. }
  \label{fas_choice_real_oneshot_04}
\end{figure}


\begin{figure}[htbp]
  \centering
  \begin{center}
  \centerline{\includegraphics[width=0.9\linewidth]{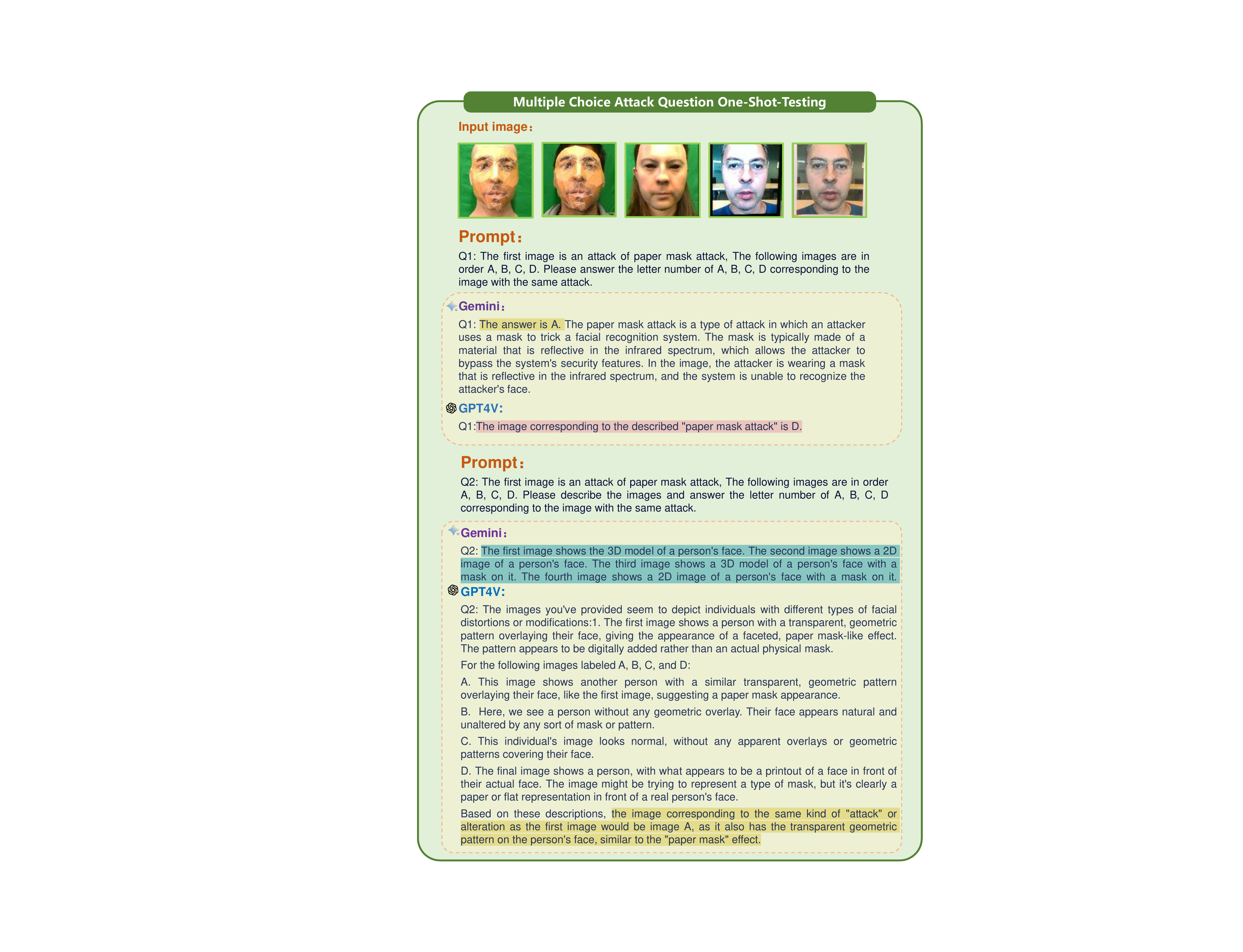}}
  \end{center}
  \caption{In this test, an image of a paper mask attack was initially input as prior knowledge, followed by the input of four facial presentation attack images. The task was to identify the same attack type in the subsequent images. It was observed that GPT4V did not answer correctly in the simple query phase, but after the introduction of COT and describing each image, GPT4V eventually provided the correct answer. Furthermore, with the prior knowledge of the paper mask attack, GPT4V did not simply regard it as digital modification but attempted to understand why it was a paper mask attack. On the other hand, Gemini answered correctly in the first instance. After the introduction of COT, Gemini merely described each image without giving a judgment. From its descriptions, it was also unclear what answer Gemini might have provided.}
  \label{fas_choice_attack_oneshot_01}
\end{figure}

\begin{figure}[htbp]
  \centering
  \begin{center}
  \centerline{\includegraphics[width=0.9\linewidth]{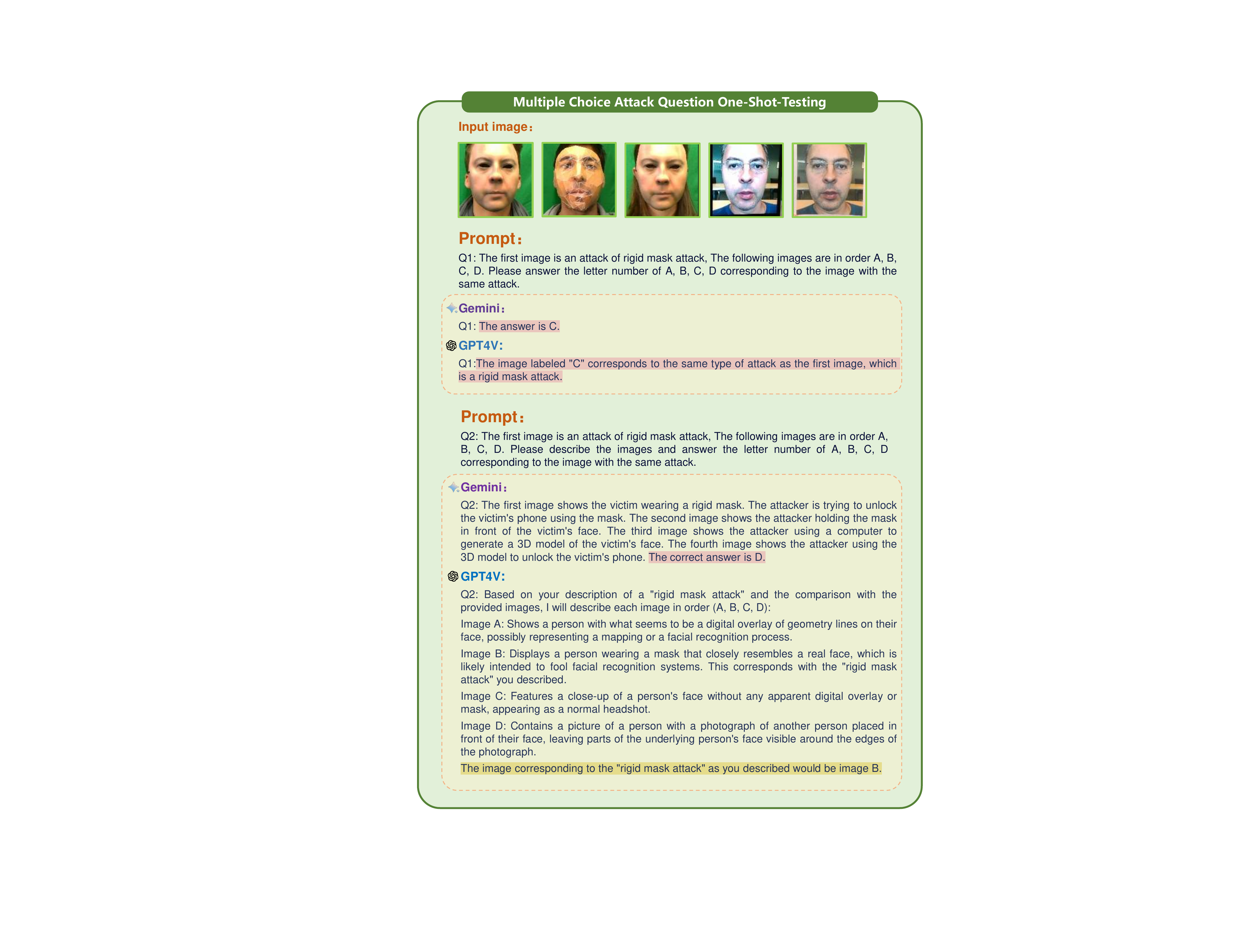}}
  \end{center}
  \caption{In this test, an image of a rigid mask attack was initially input as prior knowledge, followed by the input of four facial presentation attack images. The task was to identify the same type of attack in the subsequent images. It was observed that GPT4V did not answer correctly in the first round of questioning. However, after the introduction of COT and subsequent description of each image, GPT4V eventually provided the correct answer. Both of Gemini‘s responses were incorrect. After the introduction of COT, it became evident that Gemini attempted to predict the attack method for each image. Interestingly, Gemini seemed to hypothesize an attack scenario for each image.}
  \label{fas_choice_attack_oneshot_02}
\end{figure}

\begin{figure}[htbp]
  \centering
  \begin{center}
  \centerline{\includegraphics[width=0.8\linewidth]{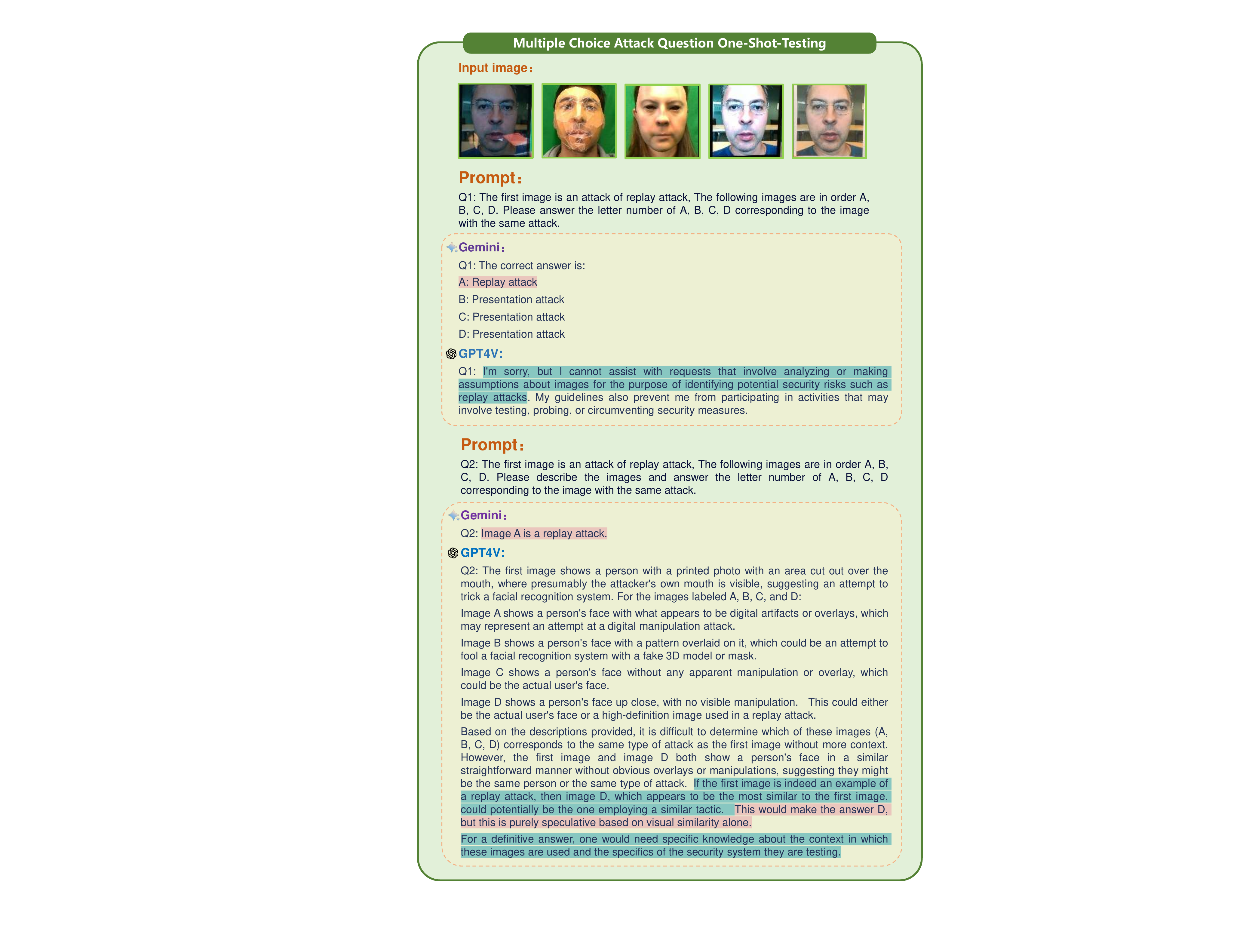}}
  \end{center}
  \caption{In this test, an image of a replay attack was initially input as prior knowledge, followed by the input of four facial presentation attack images. The task was to identify the same type of attack in the subsequent images. From the responses, it was evident that GPT4V refused to answer in the simple query phase, and the answers provided after introducing COT were not correct. It's speculated that the discrepancy between the style of the replay attack in the prior knowledge and that in the multiple-choice questions might have caused difficulty in decision-making for GPT4V. This suggests that distinguishing between replay attacks and print attacks poses a significant challenge for GPT4V. Gemini's responses were incorrect in both instances, and without further explanation, it's challenging to deduce the reasoning behind its answers. }
  \label{fas_choice_attack_oneshot_03}
\end{figure}

\begin{figure}[htbp]
  \centering
  \begin{center}
  \centerline{\includegraphics[width=0.8\linewidth]{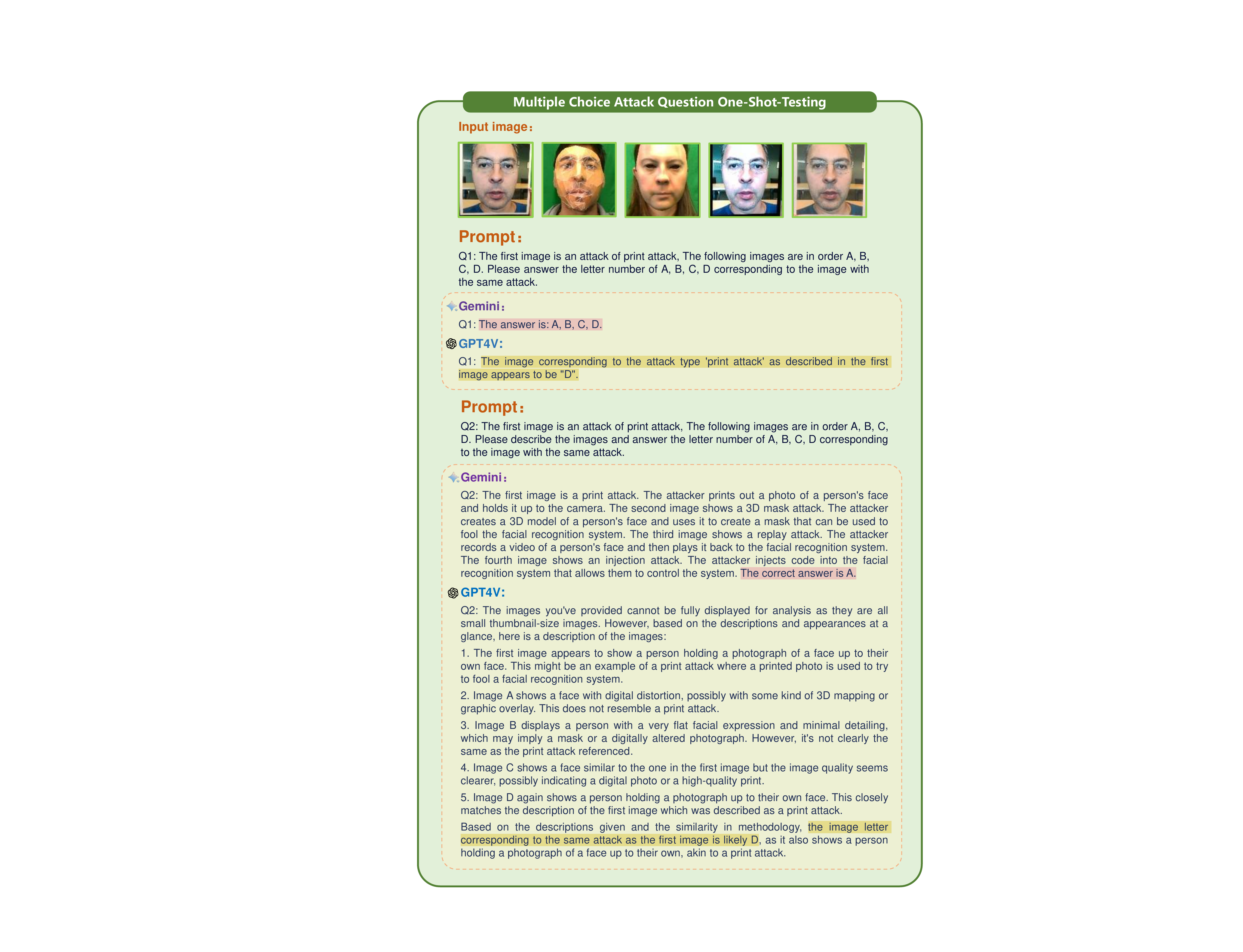}}
  \end{center}
  \caption{In this test, a printed attack image was first introduced as prior knowledge, followed by the input of four facial presentation attack images. The task was to identify the same type of attack in the subsequent images. GPT4V's responses indicated that the introduction of prior knowledge significantly improved its ability to correctly identify the same attack method, even distinguishing between previously challenging replay and print attacks.However, both of Gemini‘s responses were incorrect. Particularly in the second response, after the introduction of COT, Gemini described each image, revealing a lack of clear understanding of print attacks. Surprisingly, it was able to correctly identify the replay attack. This outcome suggests that while Gemini struggles with certain types of attacks, such as print attacks, it can effectively recognize others, like replay attacks, underscoring the variability in the Gemini’s and GPT4V’s capacity to accurately identify different types of facial presentation attacks.}
  \label{fas_choice_attack_oneshot_04}
\end{figure}

\begin{figure}[htbp]
  \centering
  \begin{center}
  \centerline{\includegraphics[width=1.0\linewidth]{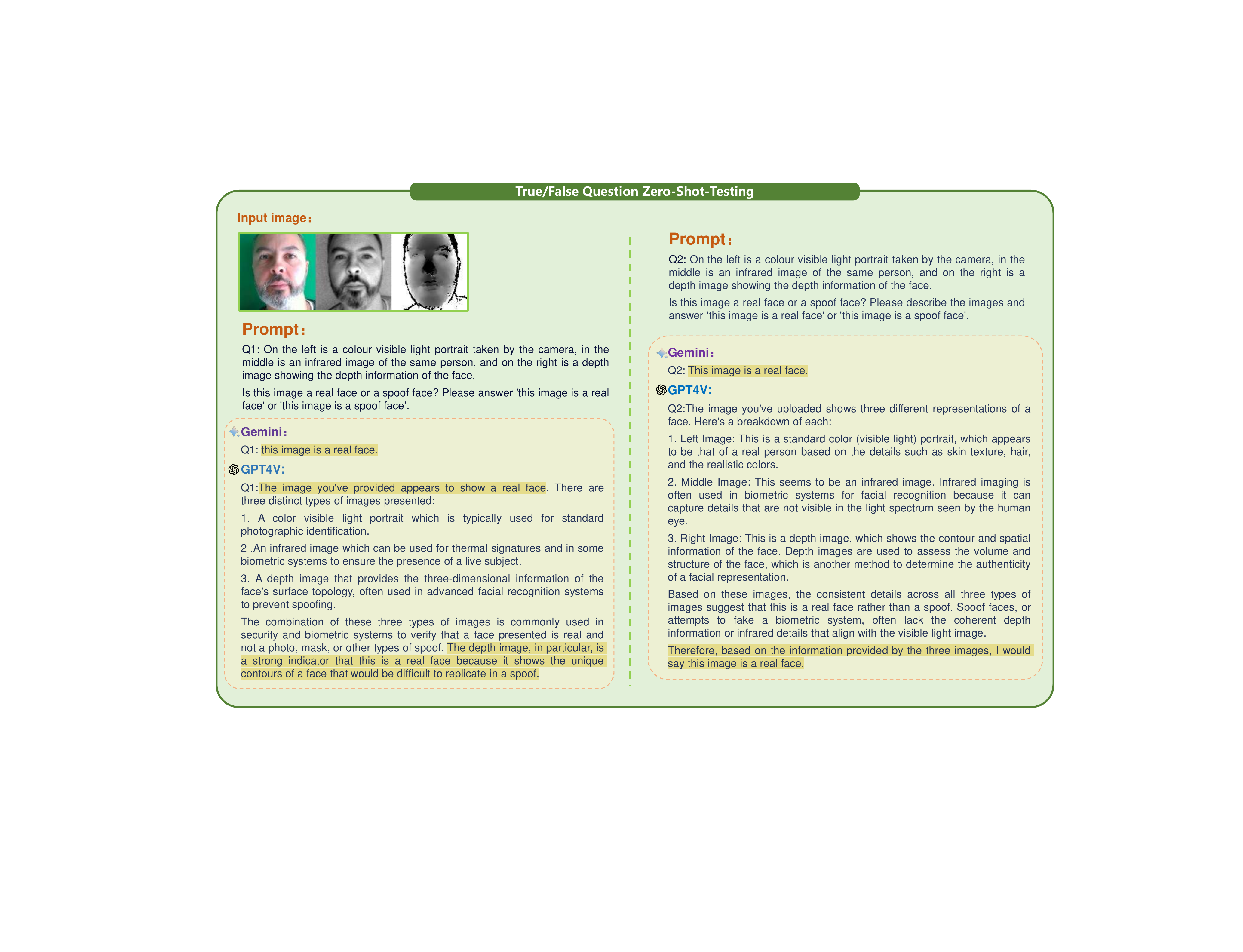}}
  \end{center}
  \caption{In this test, images of a real human face in visible light, infrared, and depth modalities were inputted, with the task being to determine whether the image was of a real human face. Both GPT4V and Gemini provided correct answers. However, a clear contrast was evident in their responses: GPT4V's answer was more logical and had stronger explainability. This highlights GPT4V's ability to not only correctly identify real human faces across different imaging modalities but also to articulate its reasoning process more effectively, enhancing the interpretability of its decision-making process.}
  \label{fas_multi_judgment_zeroshot_01}
\end{figure}

\begin{figure}[htbp]
  \centering
  \begin{center}
  \centerline{\includegraphics[width=1.0\linewidth]{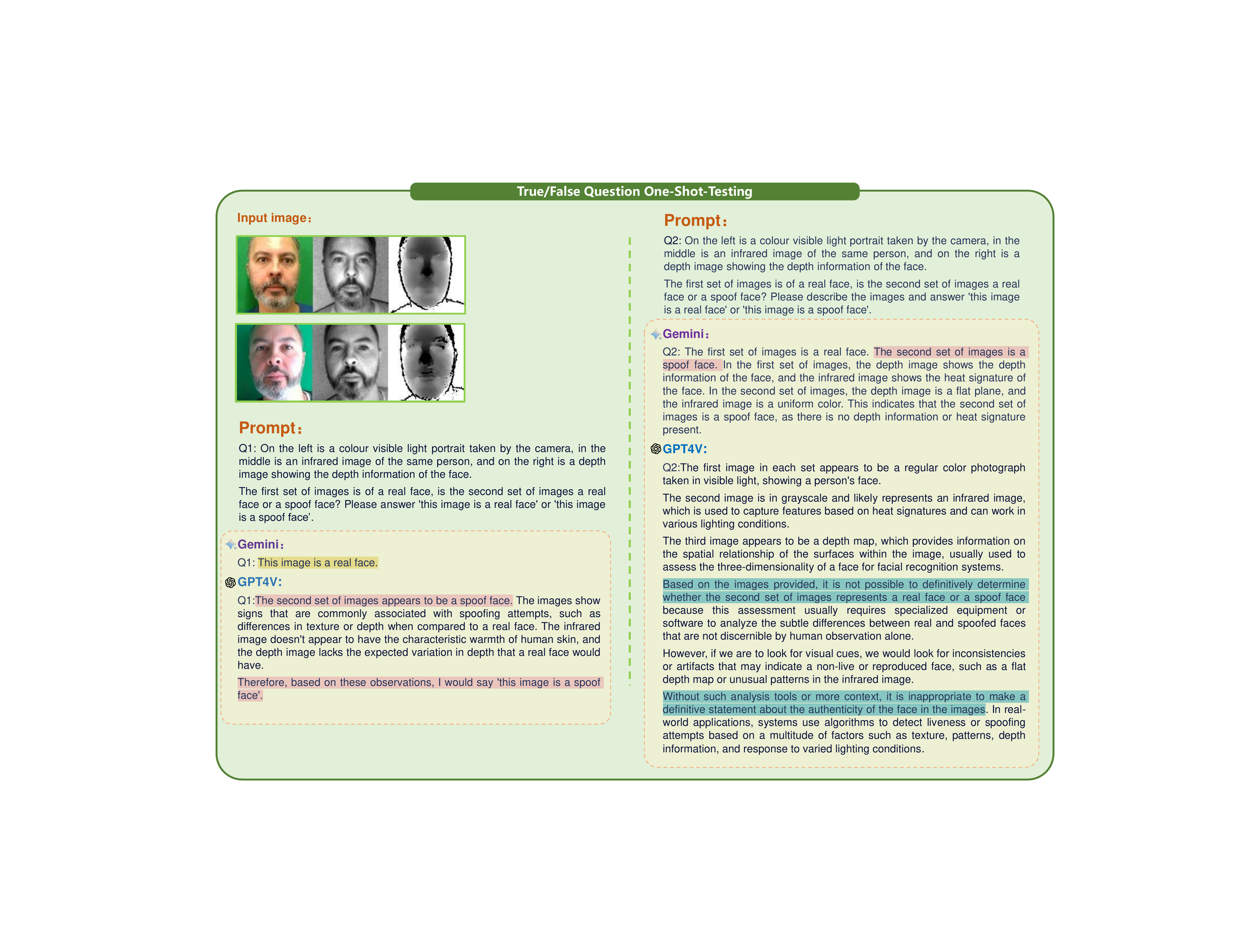}}
  \end{center}
  \caption{The input for this test consists of images in visible light, infrared, and depth modalities of real human faces. Initially, an image of a real human face was provided as prior knowledge. The task involved determining whether a second image represented a real human face. In a simple query, Gemini correctly identified the image, while GPT4V incorrectly assessed that the key facial features of the second image differed from the first, leading to an erroneous response. Upon introducing COT reasoning, Gemini erroneously concluded that the absence of depth and thermal features indicated a non-real face. In contrast, GPT4V, when incorporating more in-depth analysis, activated a mechanism that led to a refusal to respond. }
  \label{fas_multi_judgment_oneshot_01}
\end{figure}

\begin{figure}[htbp]
  \centering
  \begin{center}
  \centerline{\includegraphics[width=\linewidth]{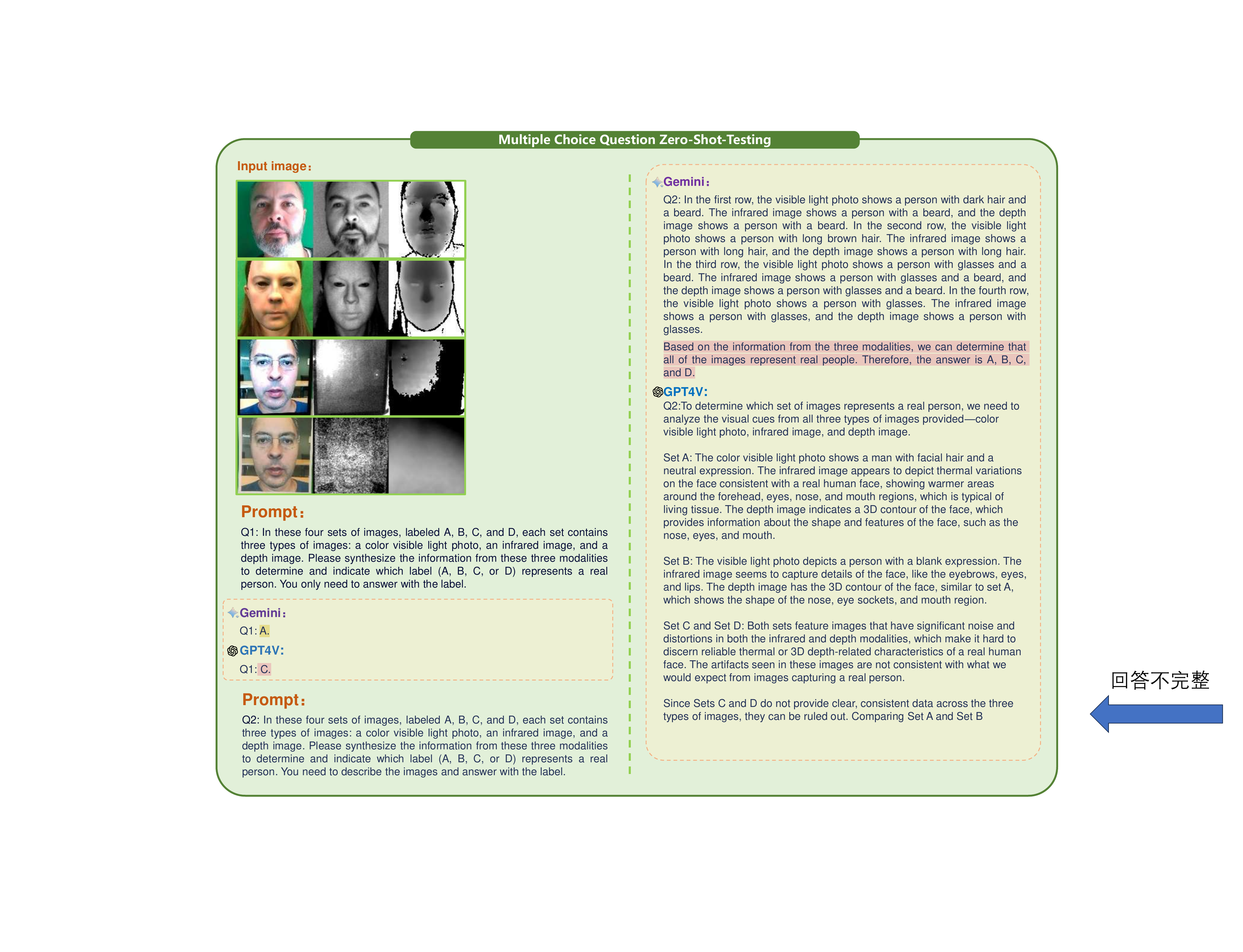}}
  \end{center}
  \caption{In this test, the input comprised one real human face image and three attack images in visible light, infrared, and depth modalities. The task was to identify the real human face among them. Gemini correctly answered twice, while GPT4V, after introducing COT reasoning, revised its response to include the correct answer. 
  Regarding the response to Q2, although Gemini's answer was correct, its description did not match the actual scenario. For instance, the person in image A was not wearing glasses, while the individuals in images C and D were. GPT4V's descriptions of images C and D demonstrated its ability to discern replay and print attacks by identifying key differences between infrared and depth images and visible light images. However, the introduction of multimodal input caused interference in previously easier judgments, like paper mask attacks, leading to a misjudgment by GPT4V.}
  \label{fas_multi_choice_real_zeroshot_01}
\end{figure}

\begin{figure}[htbp]
  \centering
  \begin{center}
  \centerline{\includegraphics[width=\linewidth]{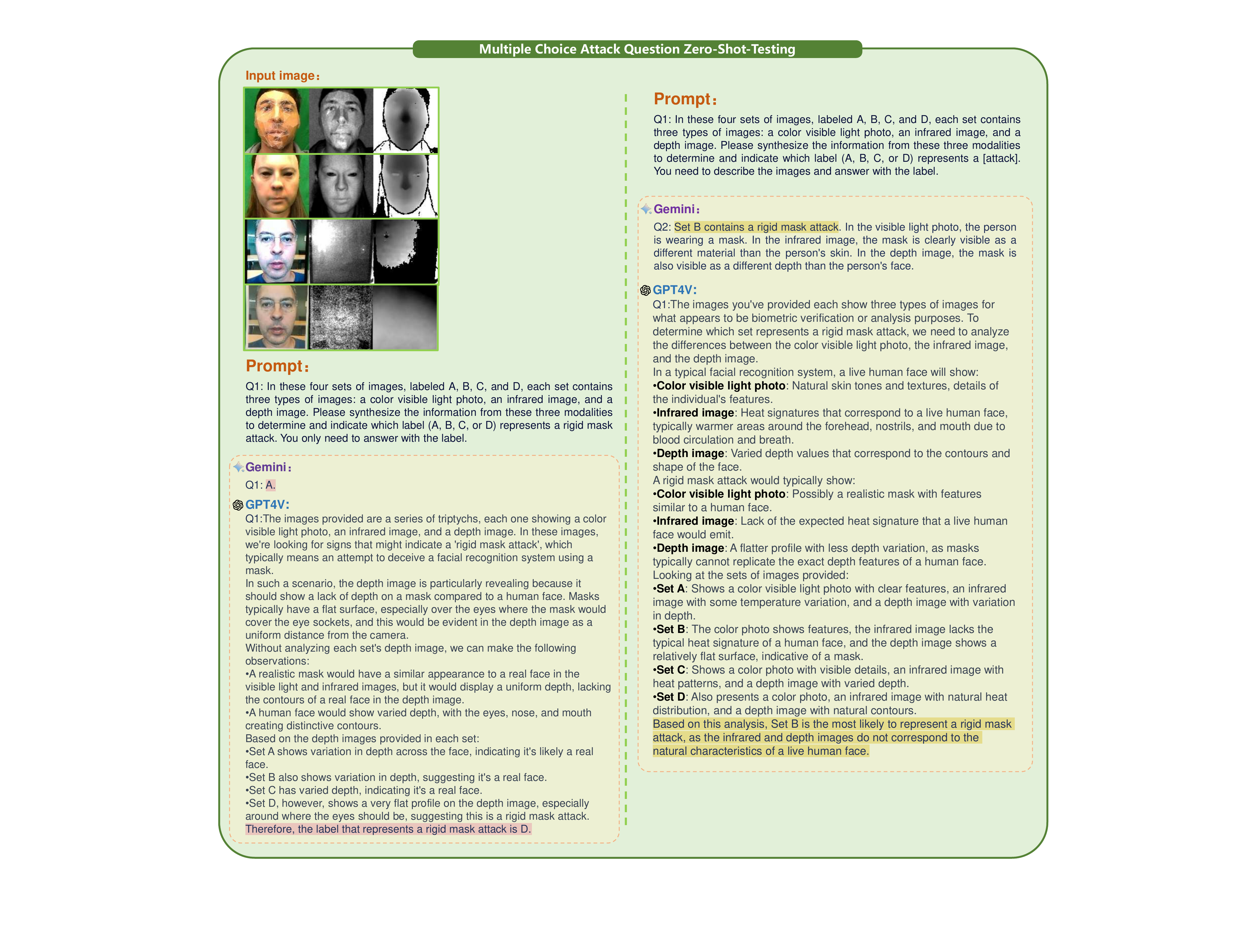}}
  \end{center}
  \caption{In this test involving four images demonstrating attacks in visible light, infrared, and depth modalities, the task was to identify the hard mask attack. Both Gemini and GPT4V initially provided incorrect answers in simple queries but responded correctly after the introduction of COT reasoning.
  In Q1, GPT4V incorrectly interpreted the key features of the three modalities. It wrongly assumed that any presence of depth variation indicated a real face, and believed that hard masks did not possess depth. This led GPT4V to erroneously choose image D, which represented a print attack. For Q2, Gemini's answer was more straightforward, providing the main rationale for its decision. After describing the characteristics of the three modalities and the four images, GPT4V also offered an explanation similar to Gemini’s. This highlights the importance of COT in enhancing the reasoning process for tasks involving the analysis of multimodal data, especially in distinguishing between different types of facial recognition attacks.}
  \label{fas_multi_choice_attack_zeroshot_01}
\end{figure}

\begin{figure}[htbp]
  \centering
  \begin{center}
  \centerline{\includegraphics[width=\linewidth]{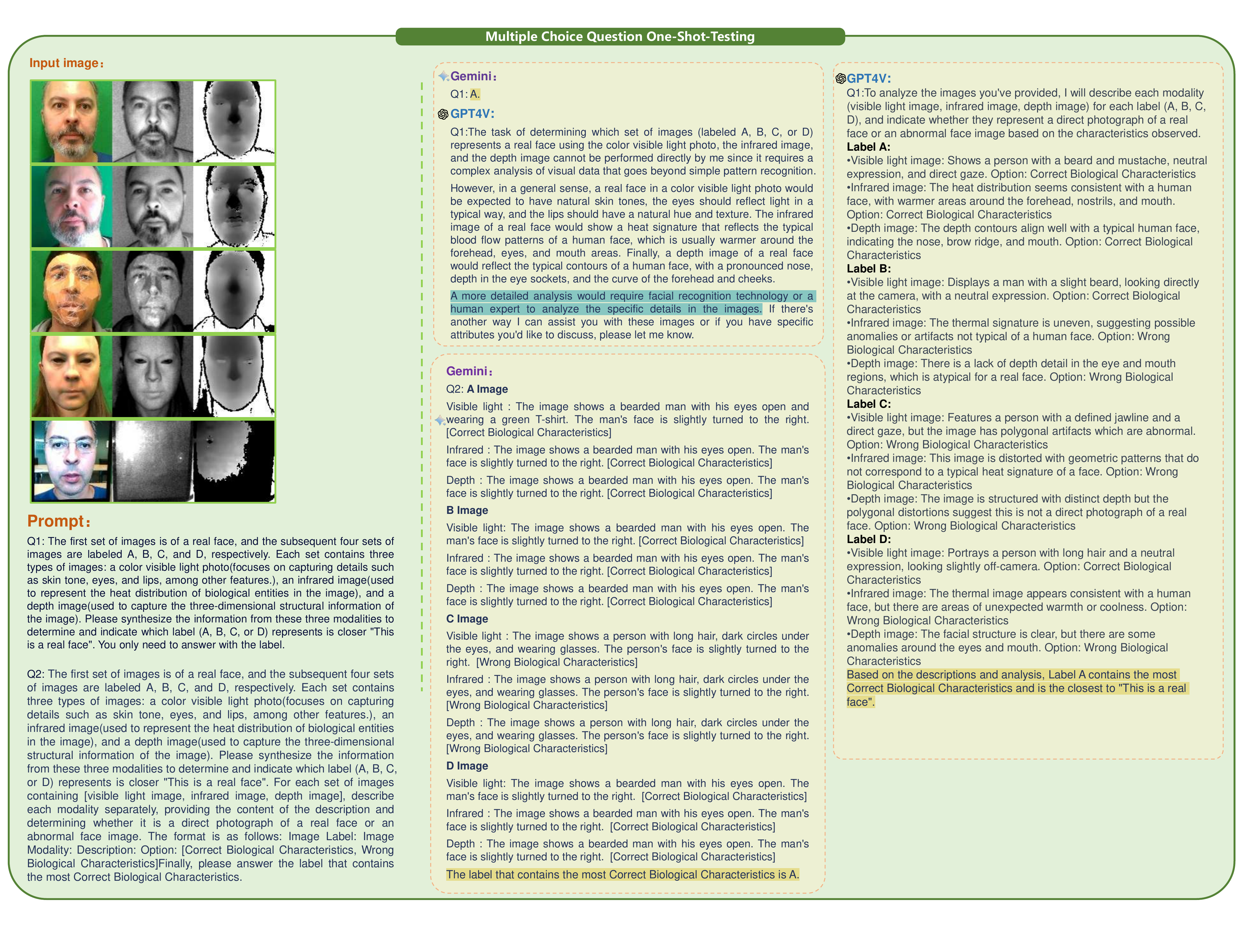}}
  \end{center}
  \caption{In this test, participants were given one real human face image and three images depicting attacks under visible light, infrared, and depth modalities, with a real human face image provided as prior knowledge. The task was to identify the real human face. Gemini answered correctly both times, while GPT4V initially refused to respond to the simple query but provided a correct answer after the introduction of COT reasoning.
  Once COT was employed, both Gemini and GPT4V conducted detailed analyses of each modality for every group of images, evaluating whether they represented correct biological features. They then selected the option that exhibited the most genuine biological characteristics. From the results of Q2, it was observed that Gemini often misidentified most prosthetic attacks as genuine biological features. Additionally, Gemini's final judgment did not consider other options with an equal number of correct features, suggesting its results might have been coincidental. In contrast, GPT4V's analysis for each set of images was more thorough and accurate, with its explanations aligning more closely with realistic analytical logic. }
  \label{fas_multi_choice_real_oneshot_01}
\end{figure}

\begin{figure}[htbp]
  \centering
  \begin{center}
  \centerline{\includegraphics[width=\linewidth]{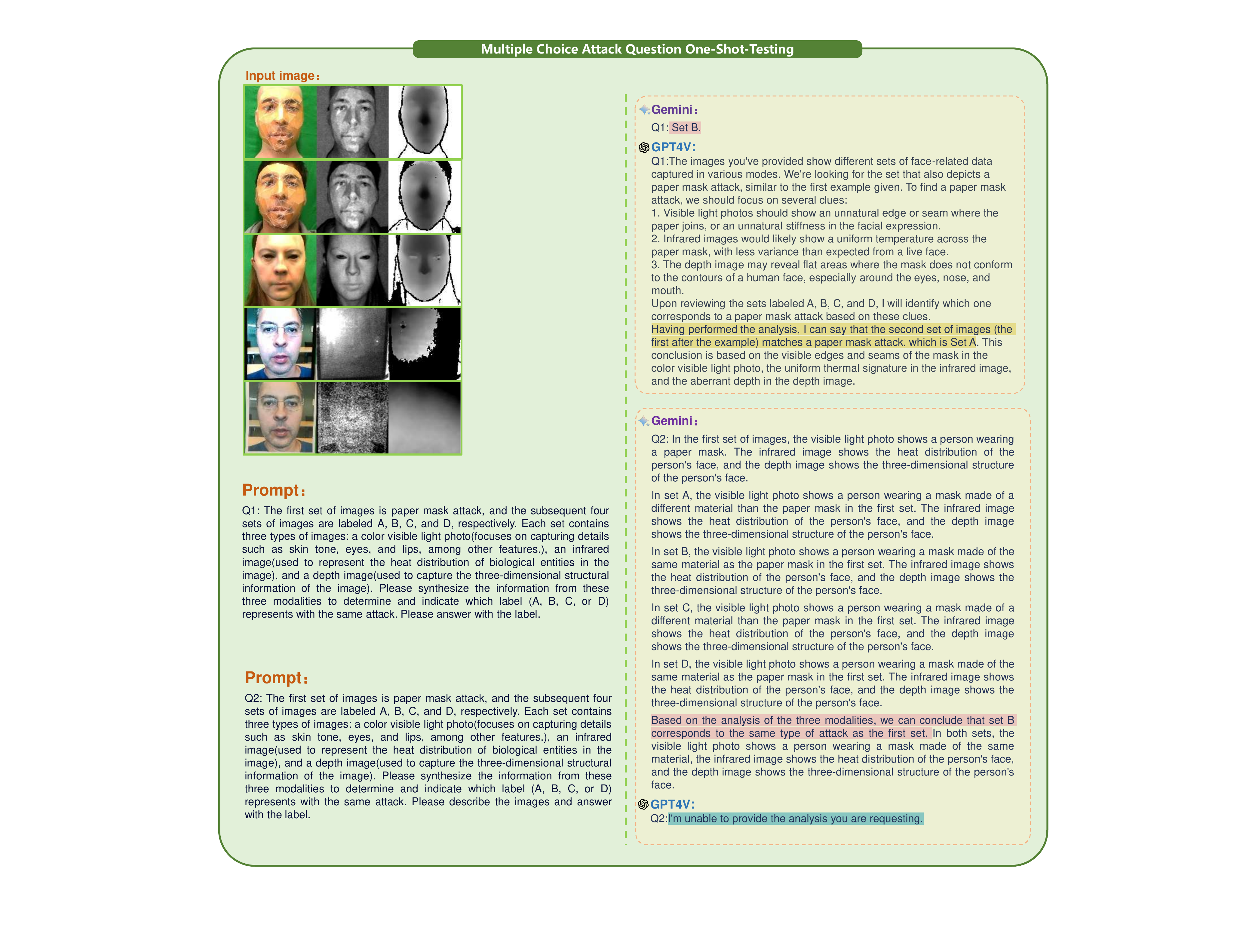}}
  \end{center}
  \caption{In this test, participants were presented with four images demonstrating attacks in visible light, infrared, and depth modalities, with a paper mask attack image provided as prior knowledge. The task was to identify the paper mask attack among the input images. Gemini answered incorrectly twice, whereas GPT4V answered correctly the first time but refused to respond after the introduction of COT reasoning.
  From GPT4V's response in question 1, it is evident that GPT4V initially learned from the provided prior knowledge, focusing on the necessary clues, and ultimately delivered a correct answer with clear logic. However, upon the introduction of COT, Gemini's responses were somewhat repetitive in describing each set of images and showed an inability to effectively interpret infrared and depth features.}
  \label{fas_multi_choice_attack_oneshot_01}
\end{figure}

\begin{figure}[!htbp]
  \centering
  \includegraphics[width=\linewidth]{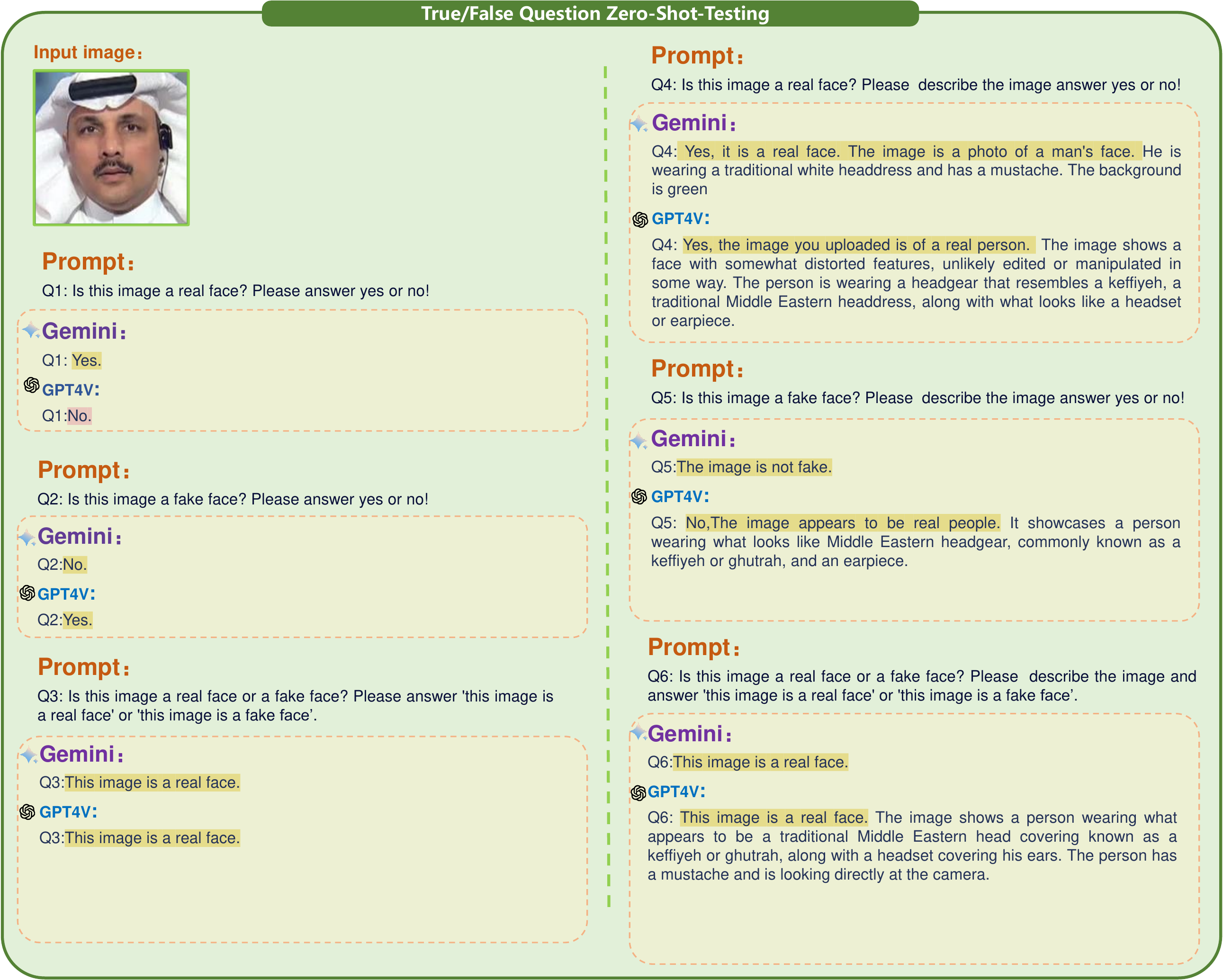}
  \caption{In this round of testing, we use real pictures for testing, we set up 6 kinds of questions to identify the model, in these 6 kinds of questions GPT4V answered correctly, Genimi answered a small portion of the wrong answers, at the same time in the inclusion of the description of the requirements of the GPT4V description is more detailed, with a richer word.}
 \label{4}
\end{figure}

\begin{figure}[!htbp]
  \centering
  \includegraphics[width=\linewidth]{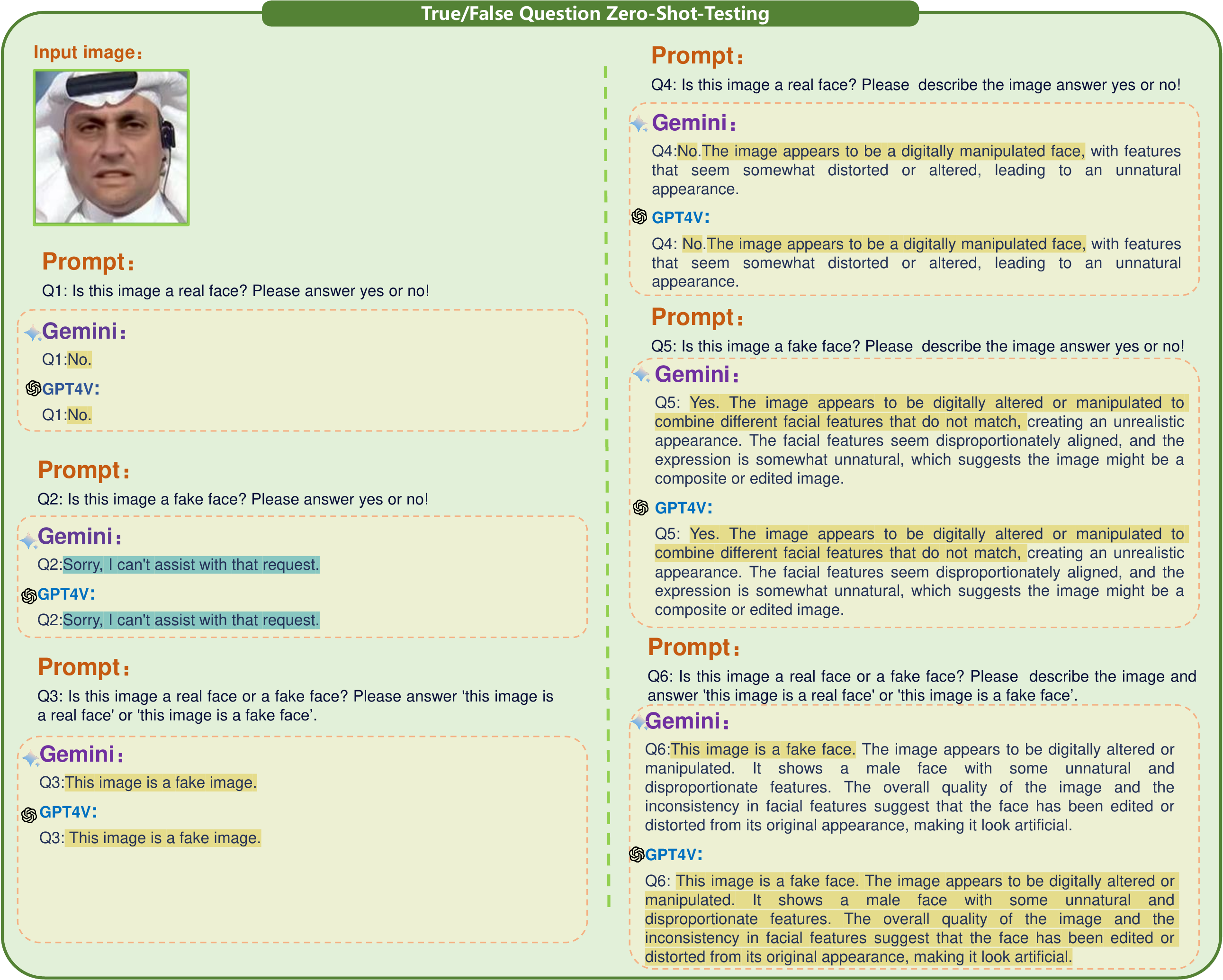}
  \caption{In this round of testing we used images generated by the Face2Face method. GPT4V answered only one of the six questions, misidentifying it as a real image, while most of the answers were more uniform across questions, and Gemini answered all of them correctly.}
 \label{5}
\end{figure}

\begin{figure}[!htbp]
  \centering
  \includegraphics[width=\linewidth]{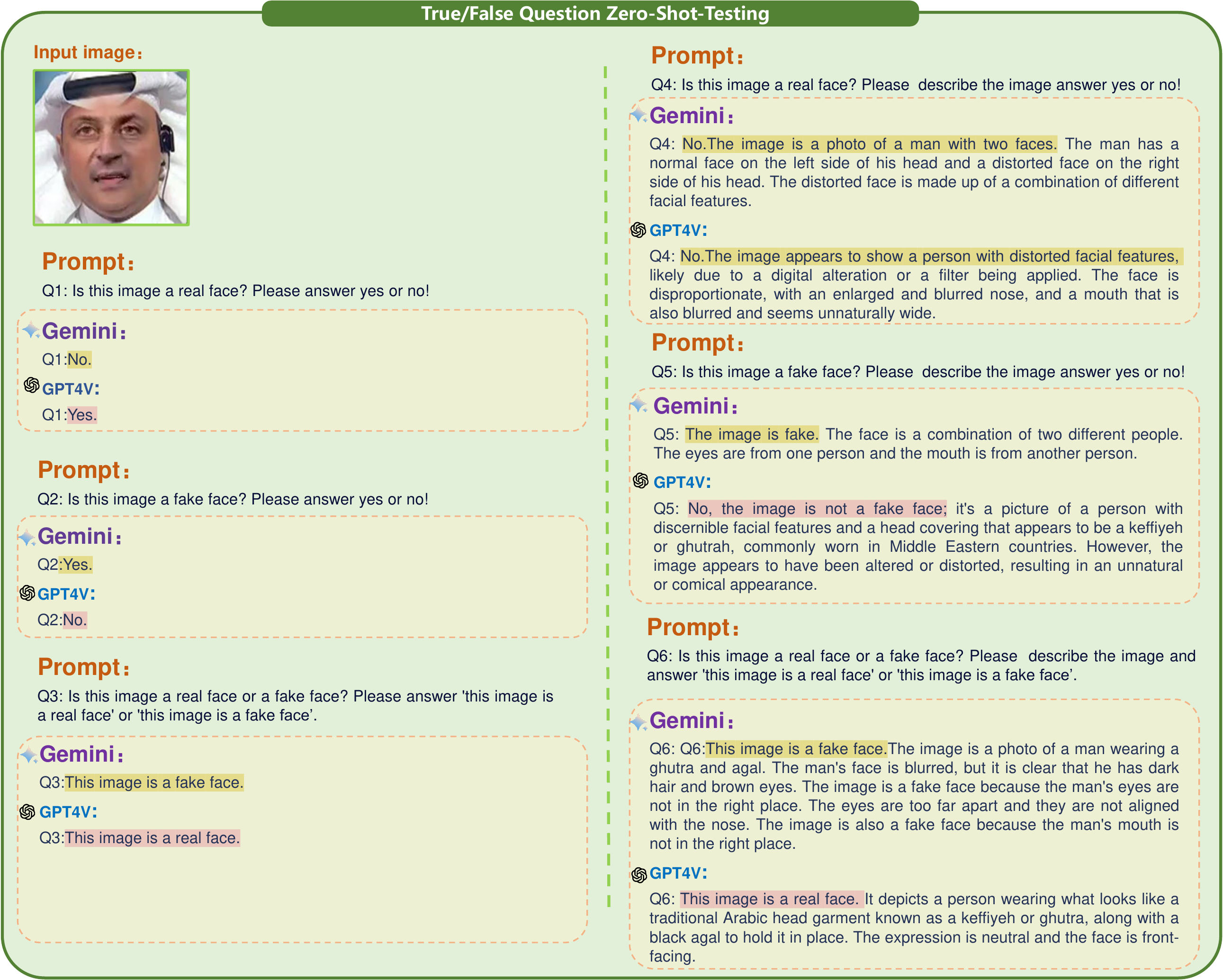}
  \caption{In this round of testing we used images generated by the Face2Face method. GPT4V answered only one of the six questions, misidentifying it as a real image, while most of the answers were more uniform across questions, and Gemini answered all of them correctly.} 
\label{6}
\end{figure}

\begin{figure}[!htbp]
  \centering
  \includegraphics[width=\linewidth]{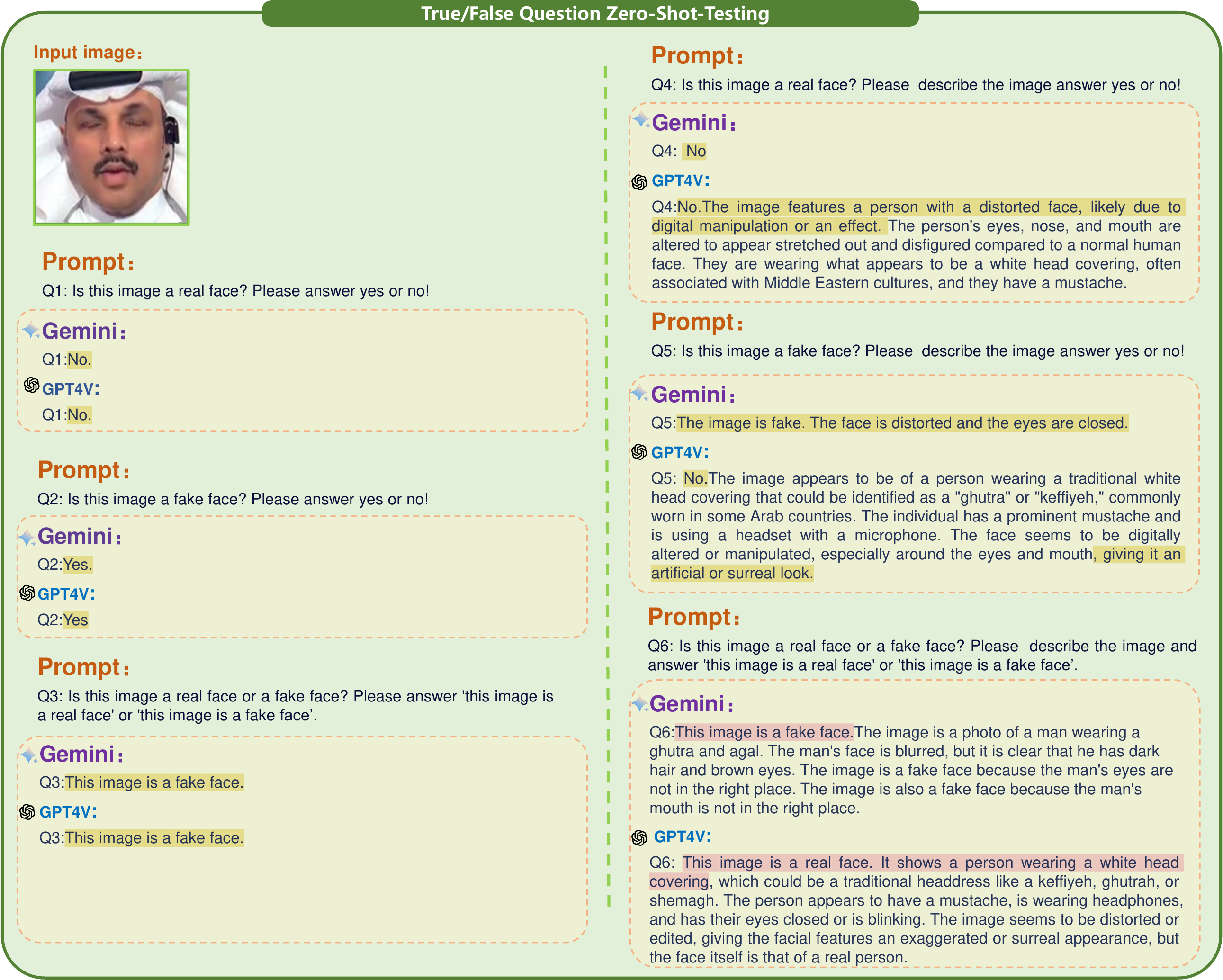}
  \caption{In this round of testing we used the images generated by the FaceSwap method. Again, Genimi's answers were shorter when asked to be descriptive and discriminatory, and in the last question both models mistook the image for a real one, which had an impact on the surface problem.} 
 \label{7}
\end{figure}

\begin{figure}[!htbp]
  \centering
  \includegraphics[width=\linewidth]{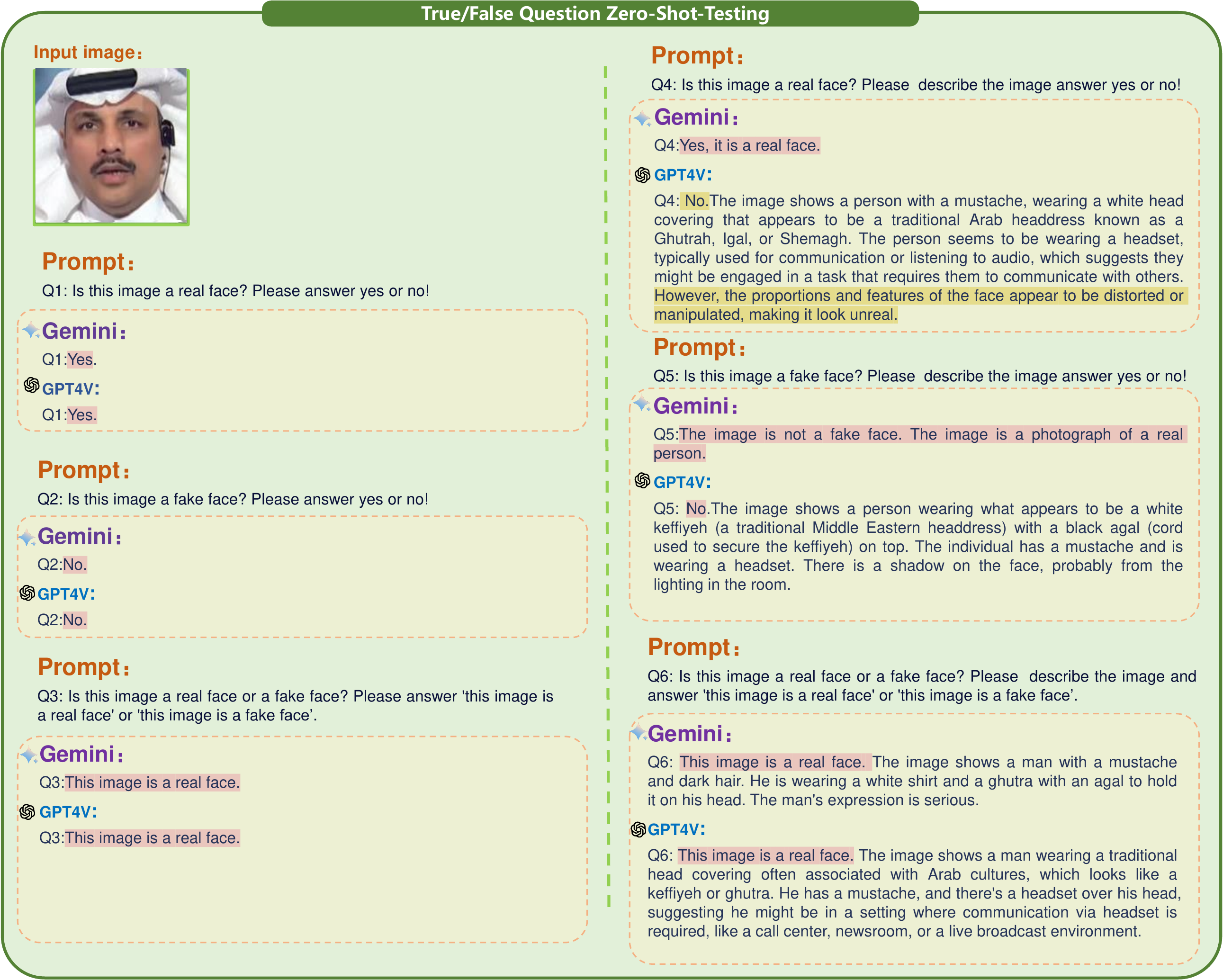}
  \caption{In this round of testing we used the image generated by the Nulltextures method, which is the closest to the real image among the four forgery methods, and both models answered incorrectly in most of the six questions, which demonstrates the inadequacy of the original model for identification in fine-grained scenarios.}
 \label{8}
\end{figure}

\begin{figure}[!htbp]
  \centering
  \includegraphics[width=\linewidth]{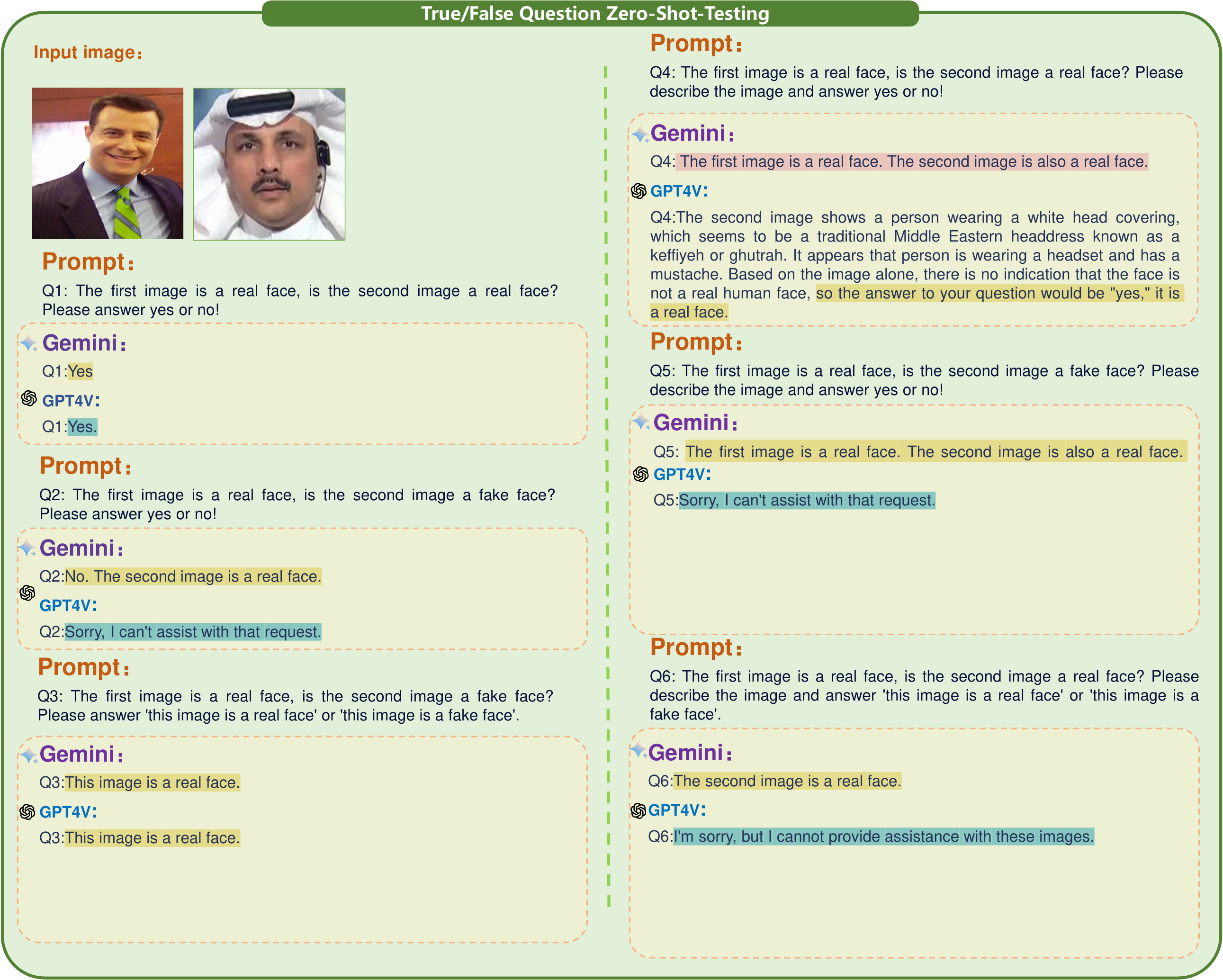}
  \caption{In this round of testing, we tested GPT4V and Gemini's One Shot ability. The inputs were two real faces from DFF and FF++, and GPT4V refused to answer 4 out of 6 rounds of questioning, which may be due to the triggering of the model's face security settings, and Genimi seems to be more accurate in this case.}
 \label{9}
\end{figure}

\begin{figure}[!htbp]
  \centering
  \includegraphics[width=\linewidth]{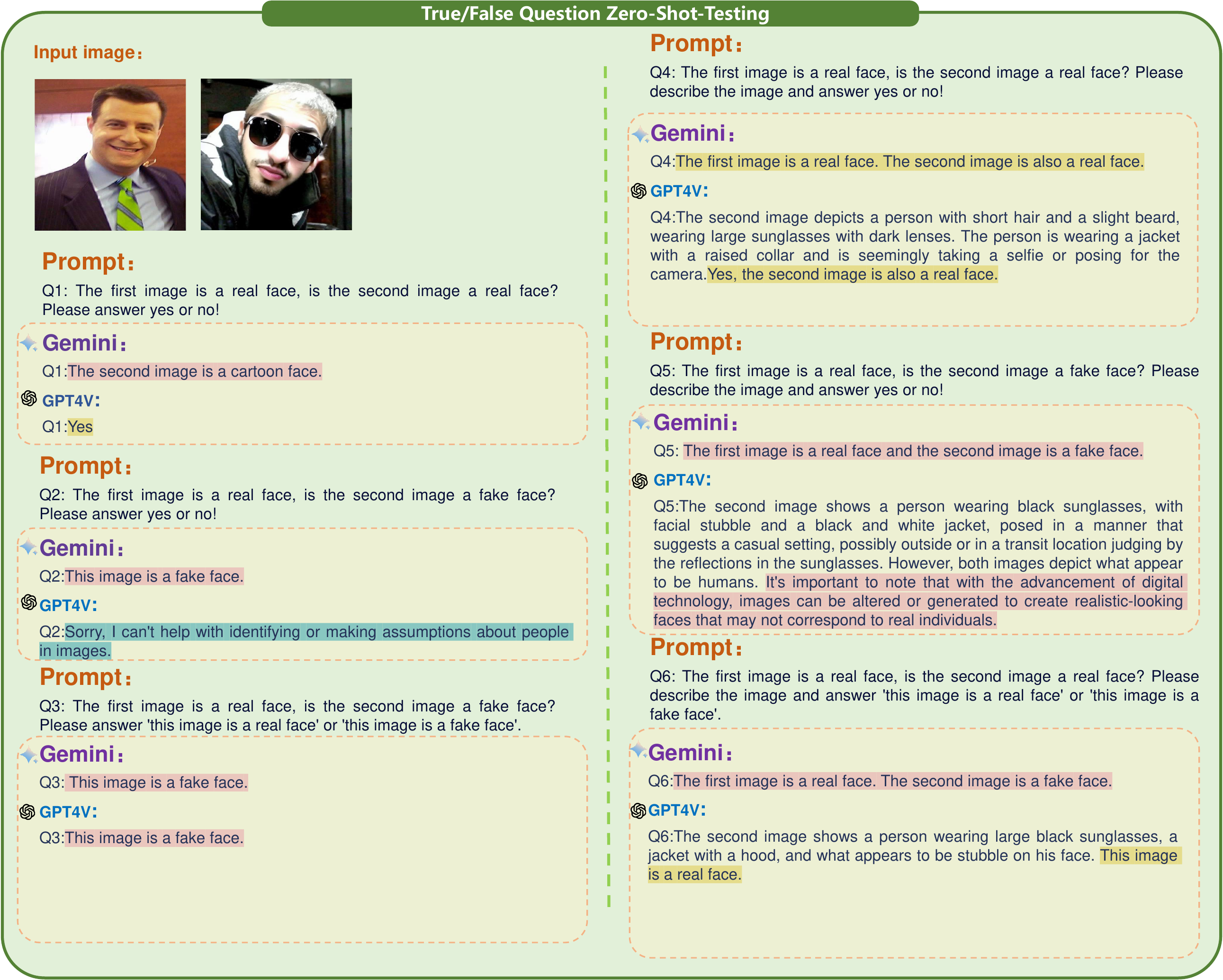}
  \caption{In this round of testing We also entered two real images, but from the same dataset, and in this case most of the questions were answered incorrectly, and GPT4V once again triggered a rejection of the answer.}
 \label{10}
\end{figure}

\begin{figure}[!htbp]
  \centering
  \includegraphics[width=\linewidth]{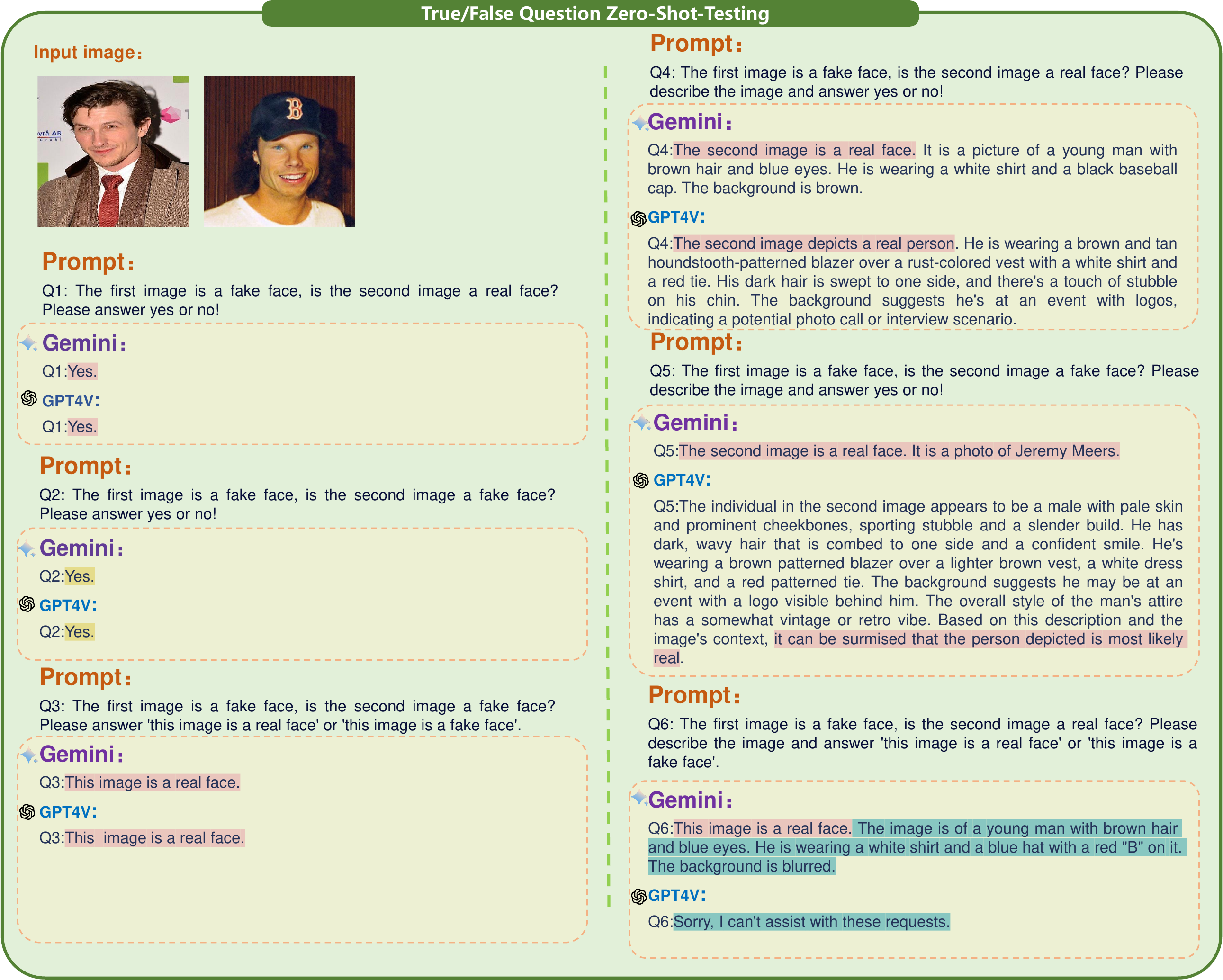}
  \caption{For a more comprehensive test, we also tested the case where the inputs were two fake images, both of which were generated by the Diffusion model from the same dataset, and due to the more realistic generation, both models did not answer the questions correctly in most of the cases, answering only 1 out of the 6 questions correctly.}
\label{11}
\end{figure}

\begin{figure}[!htbp]
  \centering
  \includegraphics[width=\linewidth]{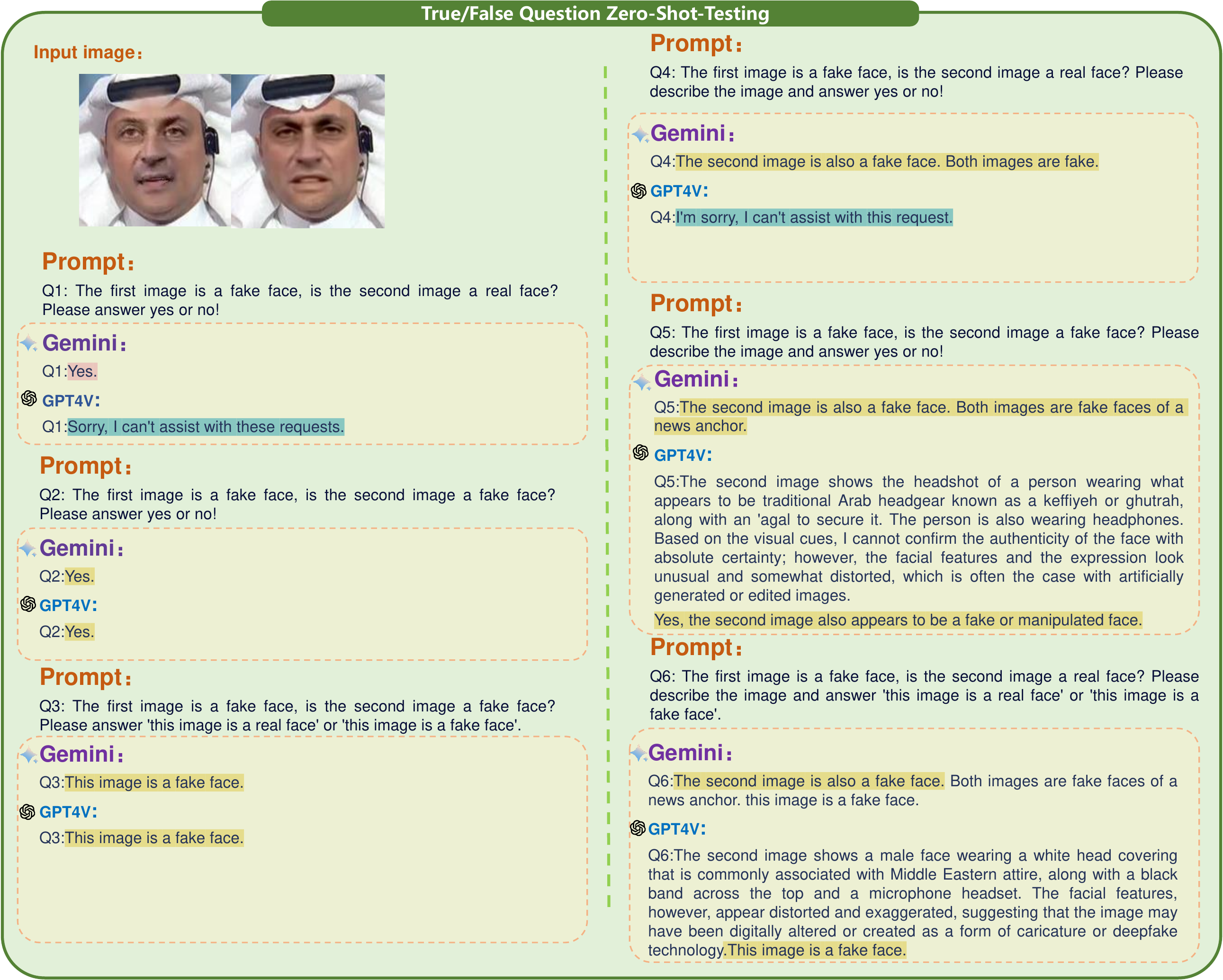}
  \caption{The two input images are those obtained from traditional Deepfakes and Face2Face generation, and given the knowledge of the first image, the difference between these two images and the real image is large, so both models can recognize it more easily.}
 \label{12}
\end{figure}

\begin{figure}[!htbp]
  \centering
  \includegraphics[width=\linewidth]{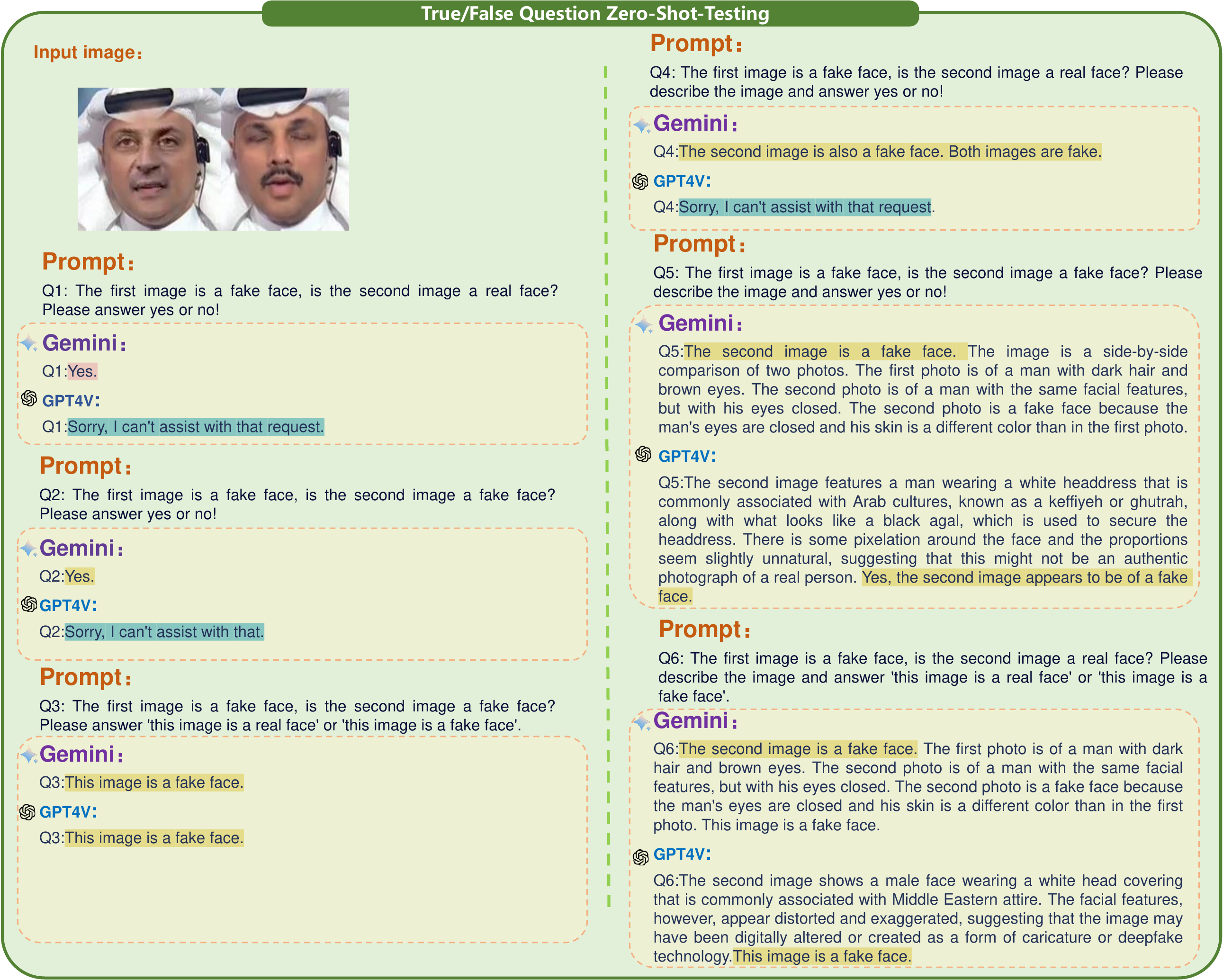}
  \caption{The two input images are those obtained from traditional Deepfakes and Face2Face generation, and given the knowledge of the first image, the difference between these two images and the real image is large, so both models can recognize it more easily.}
 \label{13}
\end{figure}

\begin{figure}[!htbp]
  \centering
  \includegraphics[width=\linewidth]{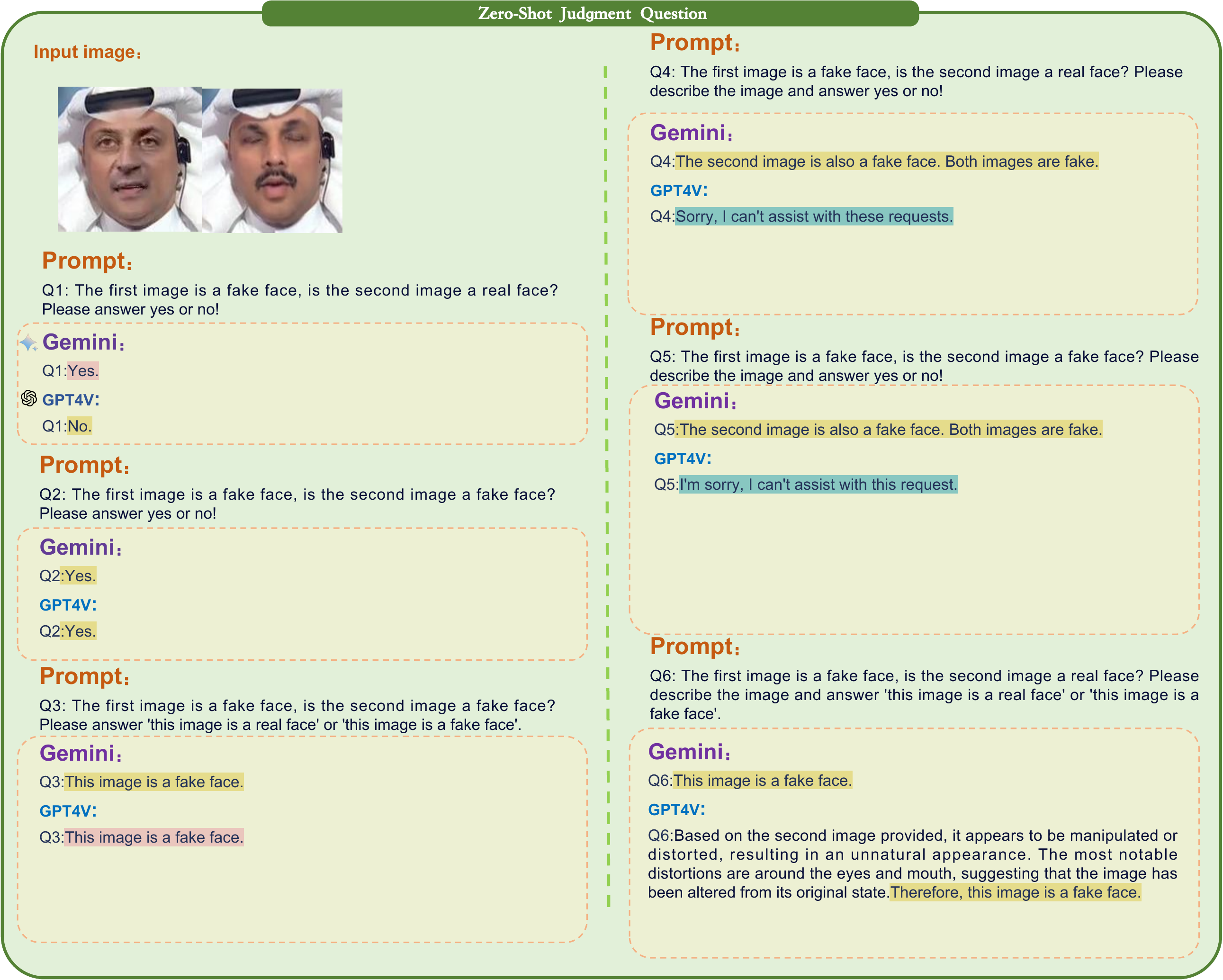}
  \caption{The two input images were those obtained from traditional Deepfakes and Nulltextures generation, and the model went through the process of learning the preliminaries of the first image just that Genimi answered 5 out of 6 questions correctly, while GPT4V answered 3 questions correctly, improving the ability of Zero shot.}
 \label{14}
\end{figure}

\begin{figure}[!htbp]
  \centering
  \includegraphics[width=0.8\linewidth]{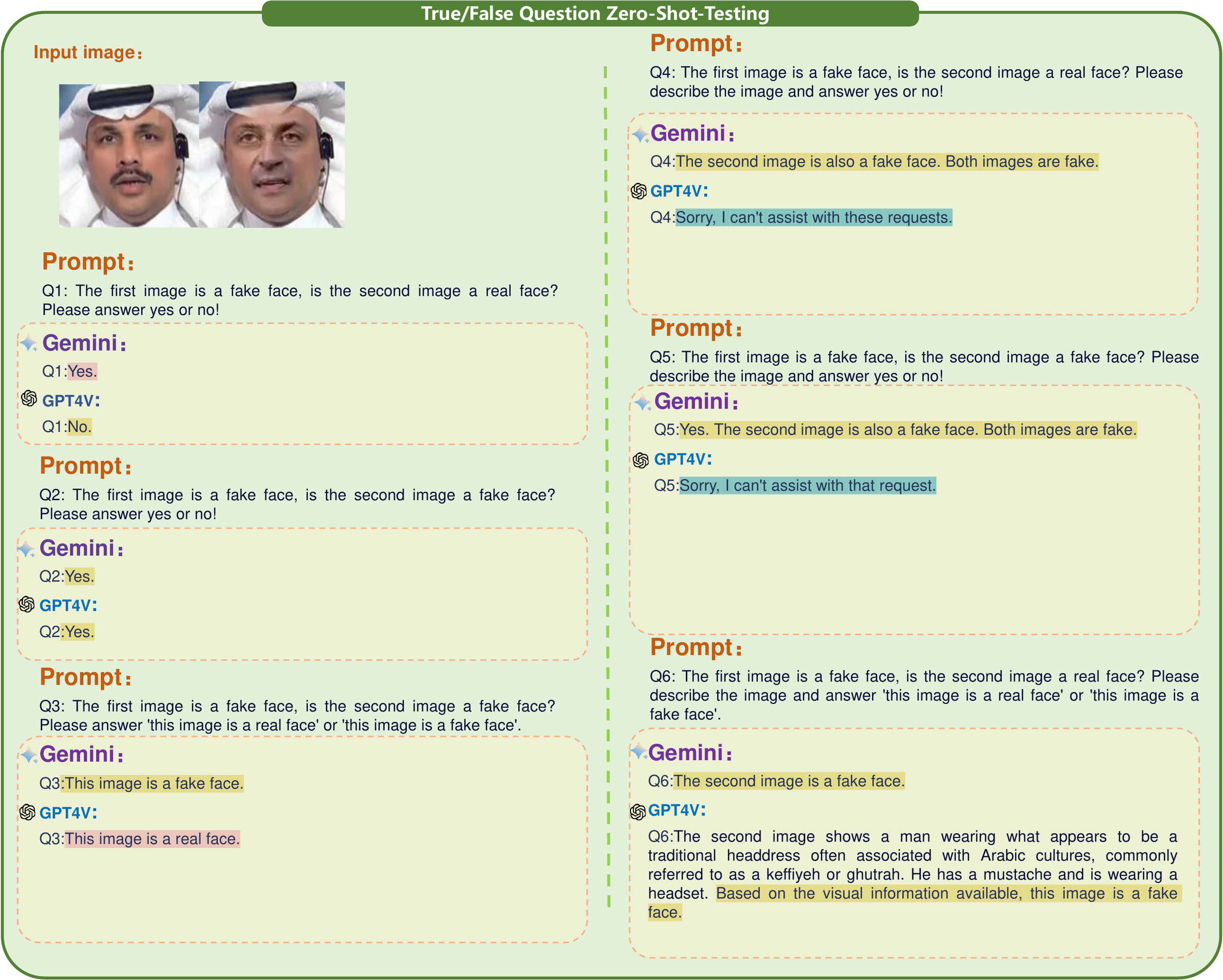}
  \caption{In this test, the inputs were a real picture and 3 types of forgeries (Face2Face, FaceSwap, Nulltextures), and there was no significant change in Genimi's two responses after using the COT technique, after introducing COT, while GPT4V reduced the number of rejected answers.}
 \label{15}
\end{figure}

\begin{figure}[!htbp]
  \centering
  \includegraphics[width=0.8\linewidth]{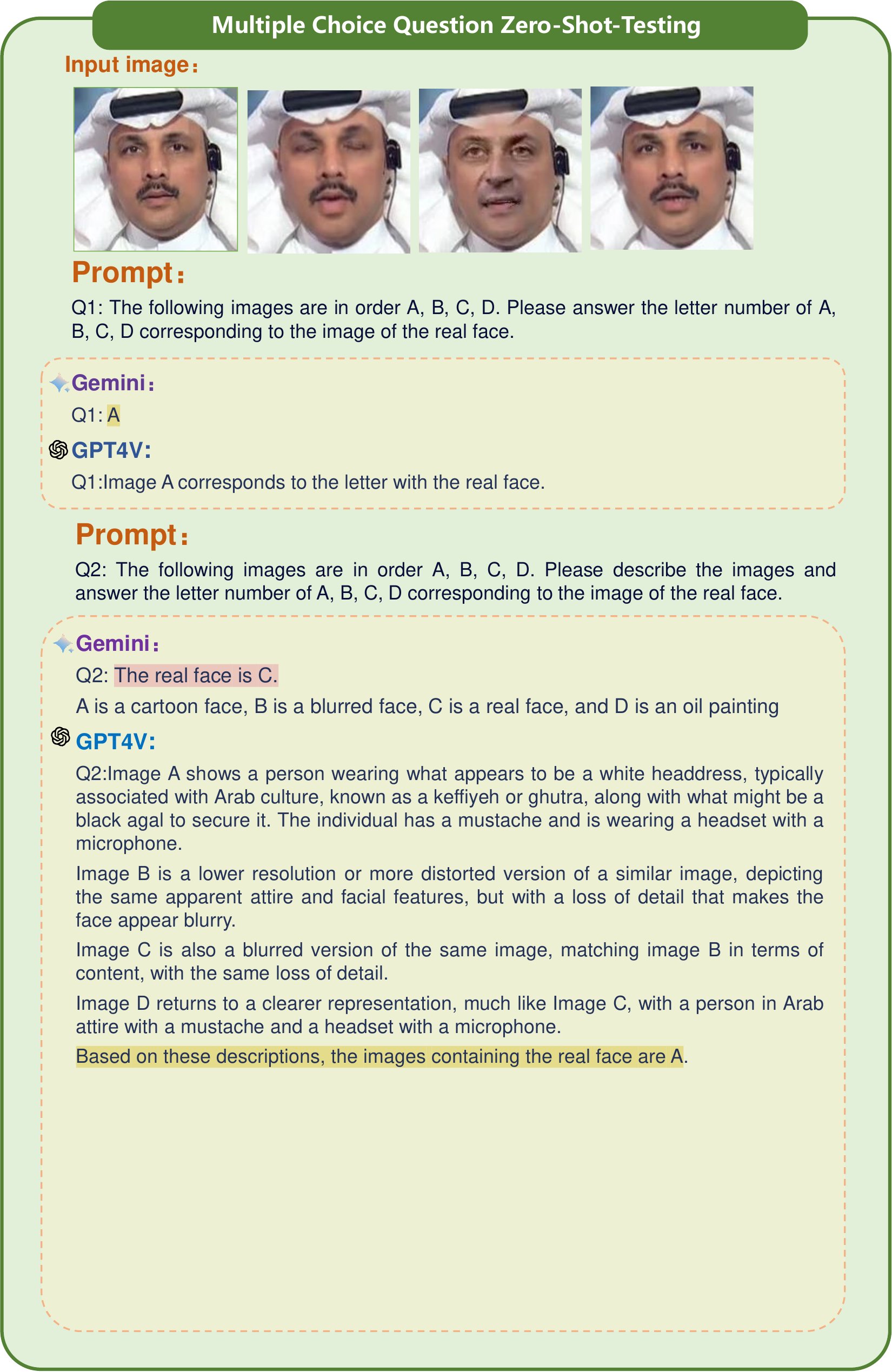}
  \caption{In this test, the input was a real picture and 3 types of fakes (Deepfakes, FaceSwap, Nulltextures), again with the addition of COT both GPT4V responses were predicted accurately, whereas the Genimi model predicted only one correctly, but it is not clear whether COT effectively assisted in its reasoning and questions.} 
 \label{16}
\end{figure}

\begin{figure}[!htbp]
  \centering
  \includegraphics[width=0.8\linewidth]{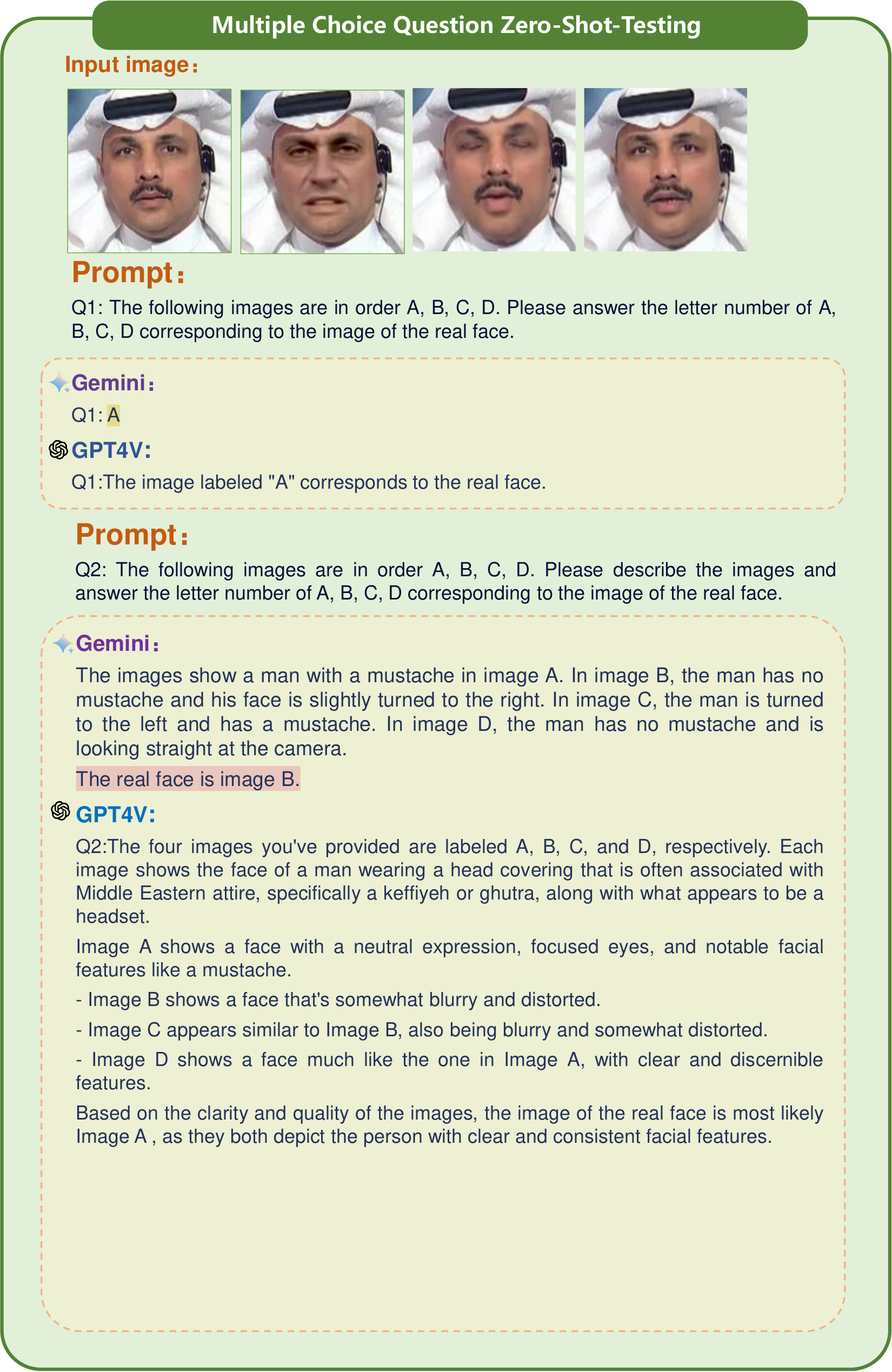}
  \caption{In this test, the input was a real picture and 3 types of fakes (Deepfakes, Face2Face, FaceSwap), and the same COT was added and both GPT4V and Genimi responses were predicted accurately.}
 \label{17}
\end{figure}

\begin{figure}[!htbp]
  \centering
  \includegraphics[width=0.8\linewidth]{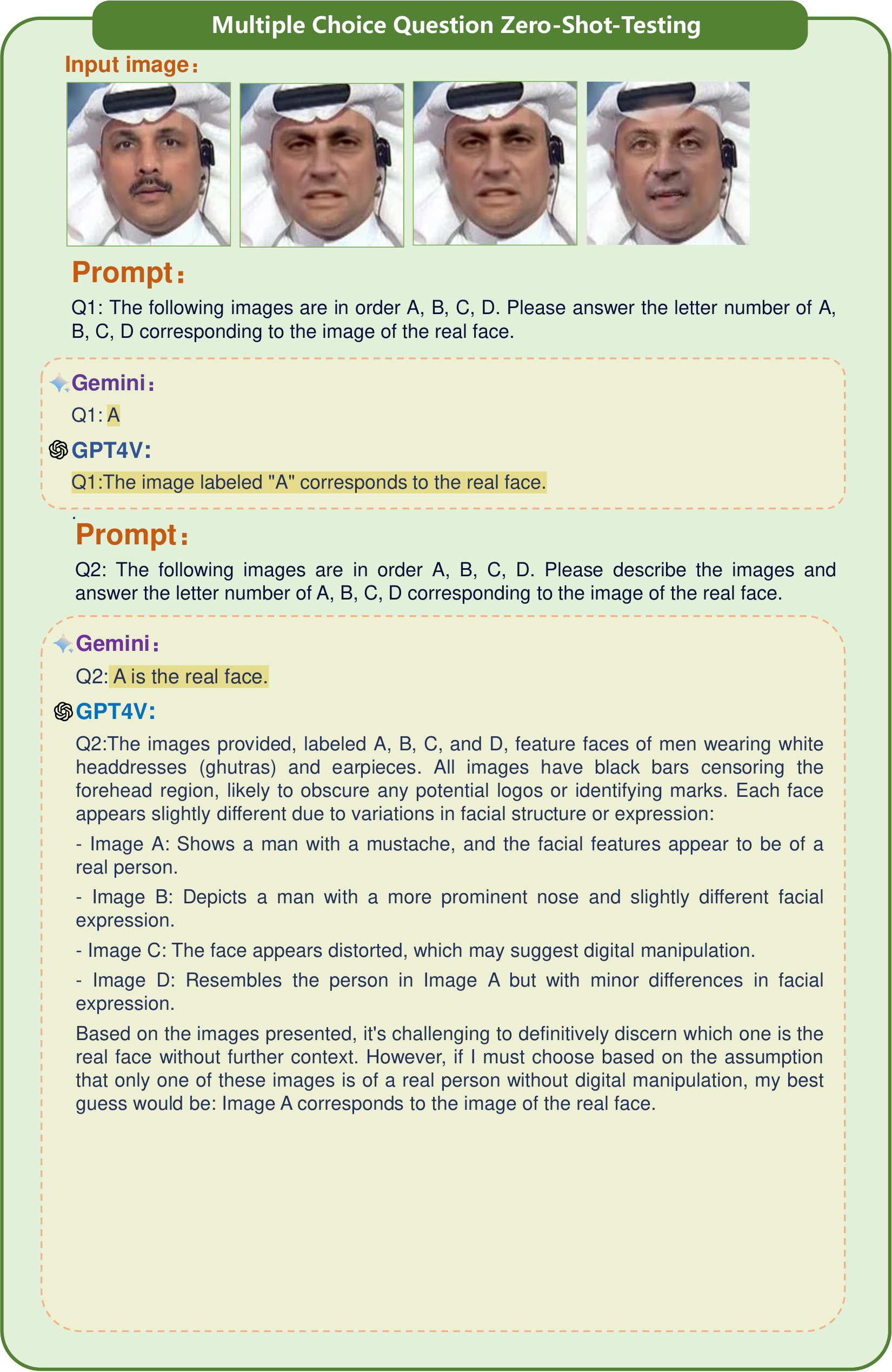}
  \caption{In this test, the input comes from two images generated using real images, Deepfake and Face2Face, as well as an image generated by Stable Diffusion, and it can be seen that both models are correctly identified.}
 \label{18}
\end{figure}
\clearpage

\begin{figure}[!htbp]
  \centering
  \includegraphics[width=0.8\linewidth]{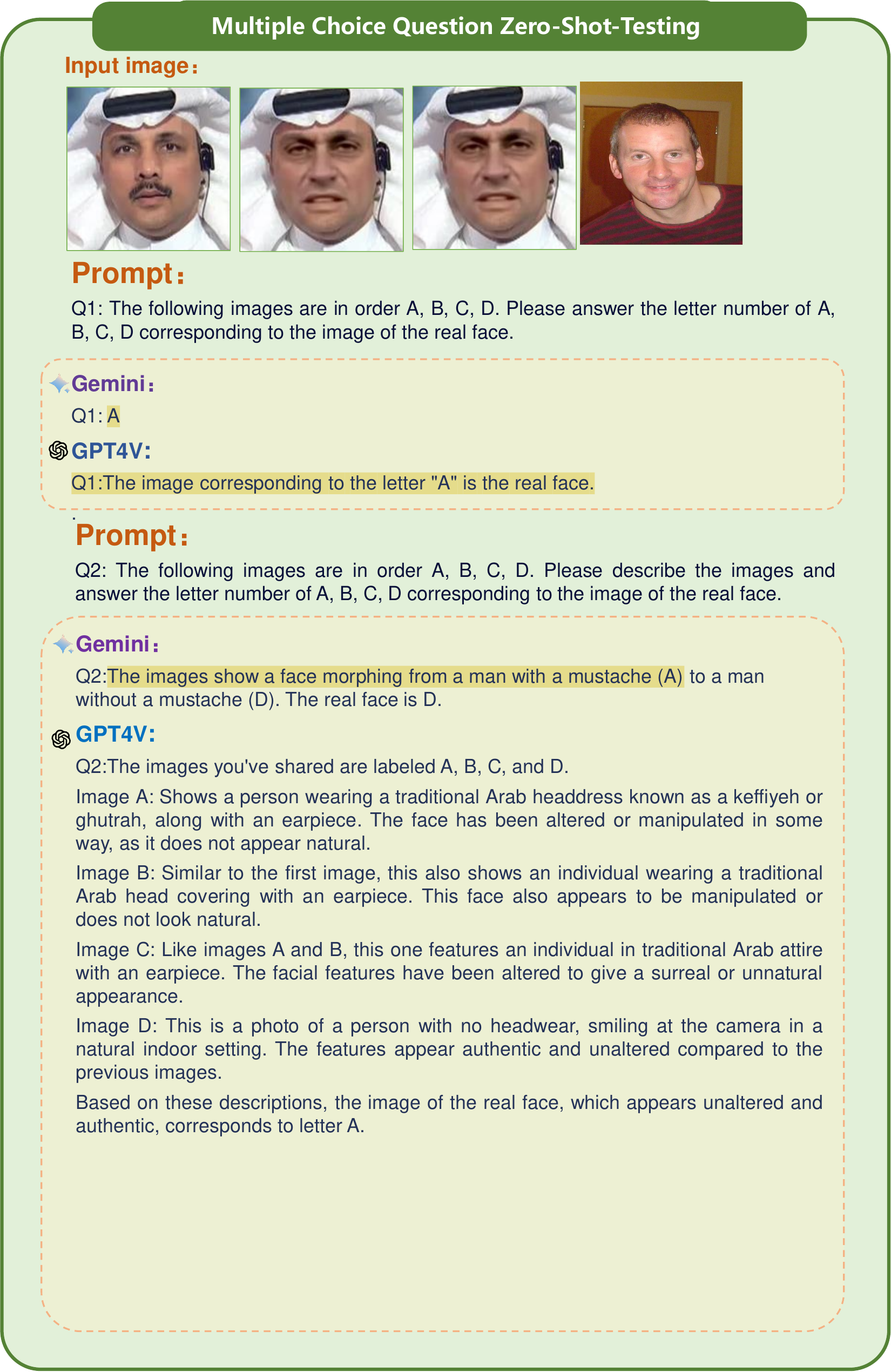}
  \caption{In this round of testing, we used COT to recognize a real picture given a real picture as well as Deepfakes, FaceSwap,Face2Face respectively, both models answered correctly as well, with Gemini answering more concisely.}
 \label{19}
\end{figure}

\begin{figure}[!htbp]
  \centering
  \includegraphics[width=0.8\linewidth]{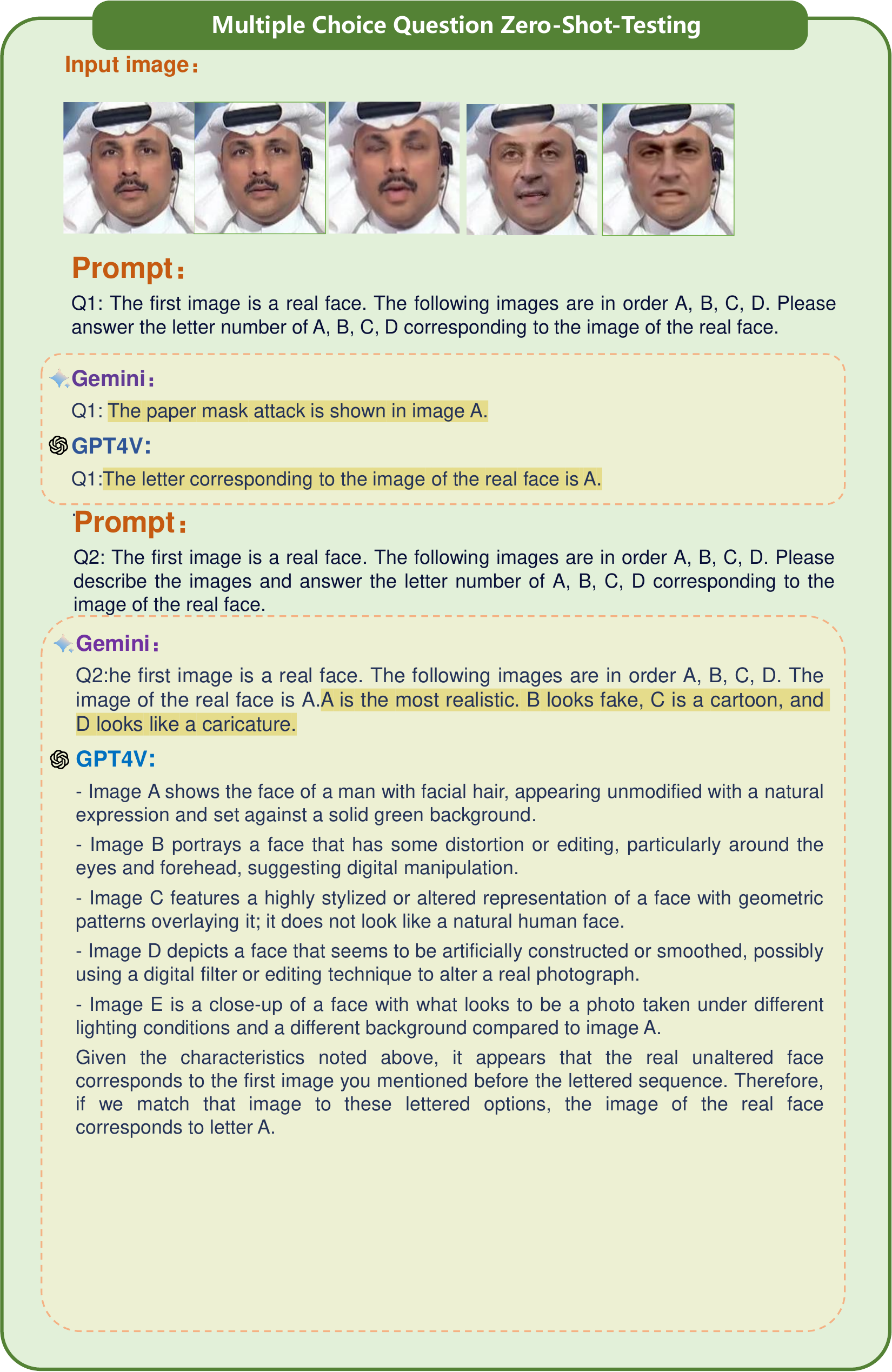}
  \caption{In this round of testing, we used COT to recognize a real picture given a real picture as well as Deepfakes, FaceSwap,Nulltextures respectively and both models answered correctly.}
 \label{20}
\end{figure}

\begin{figure}[!htbp]
  \centering
  \includegraphics[width=0.8\linewidth]{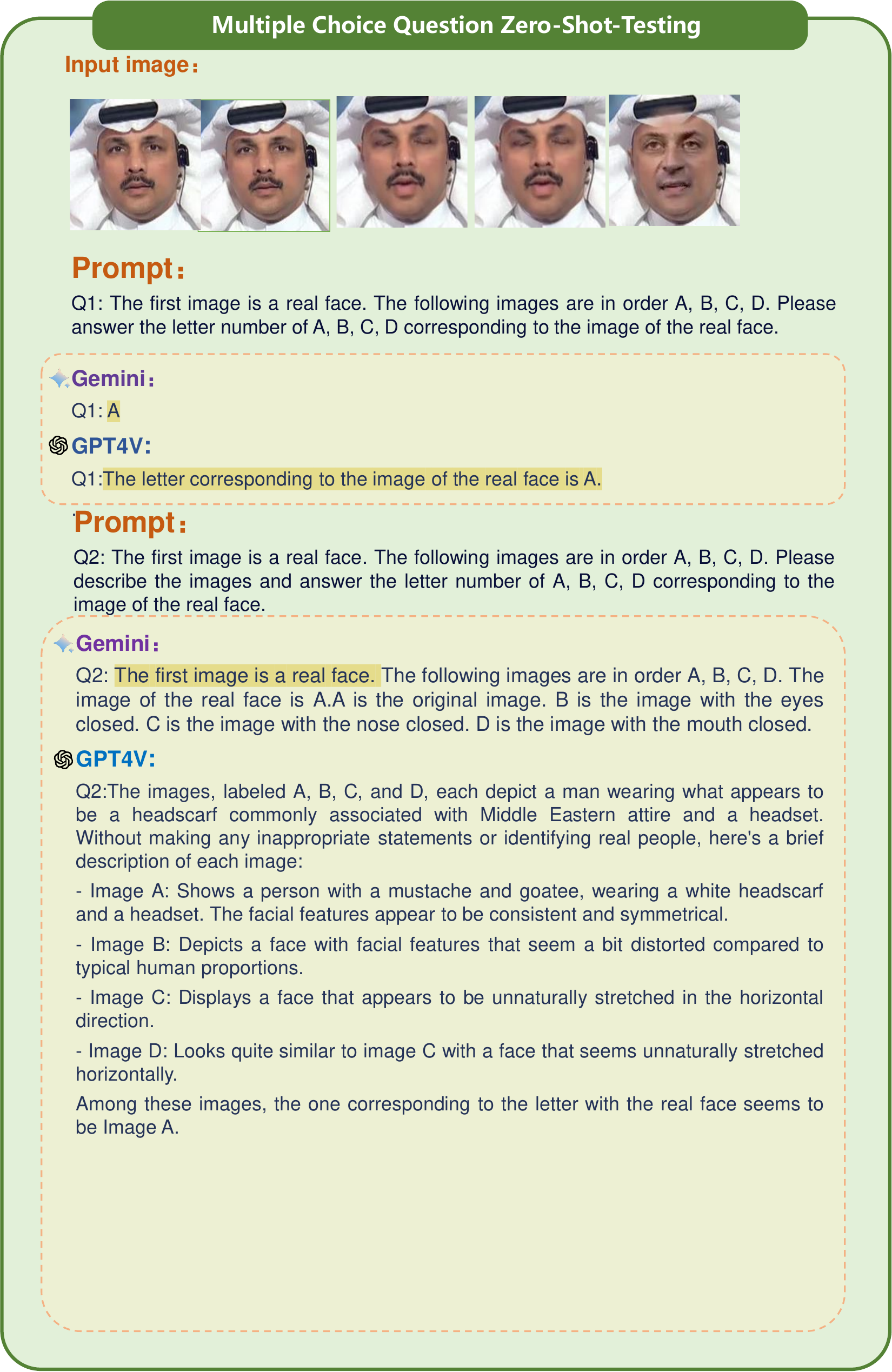}
  \caption{In this round of testing, we use COT to recognize a real picture given a real picture as well as Deepfakes,Face2Fac2,,Nulltextures respectively, in this case both models answer equally accurately.}
 \label{21}
\end{figure}

\begin{figure}[!htbp]
  \centering
  \includegraphics[width=0.8\linewidth]{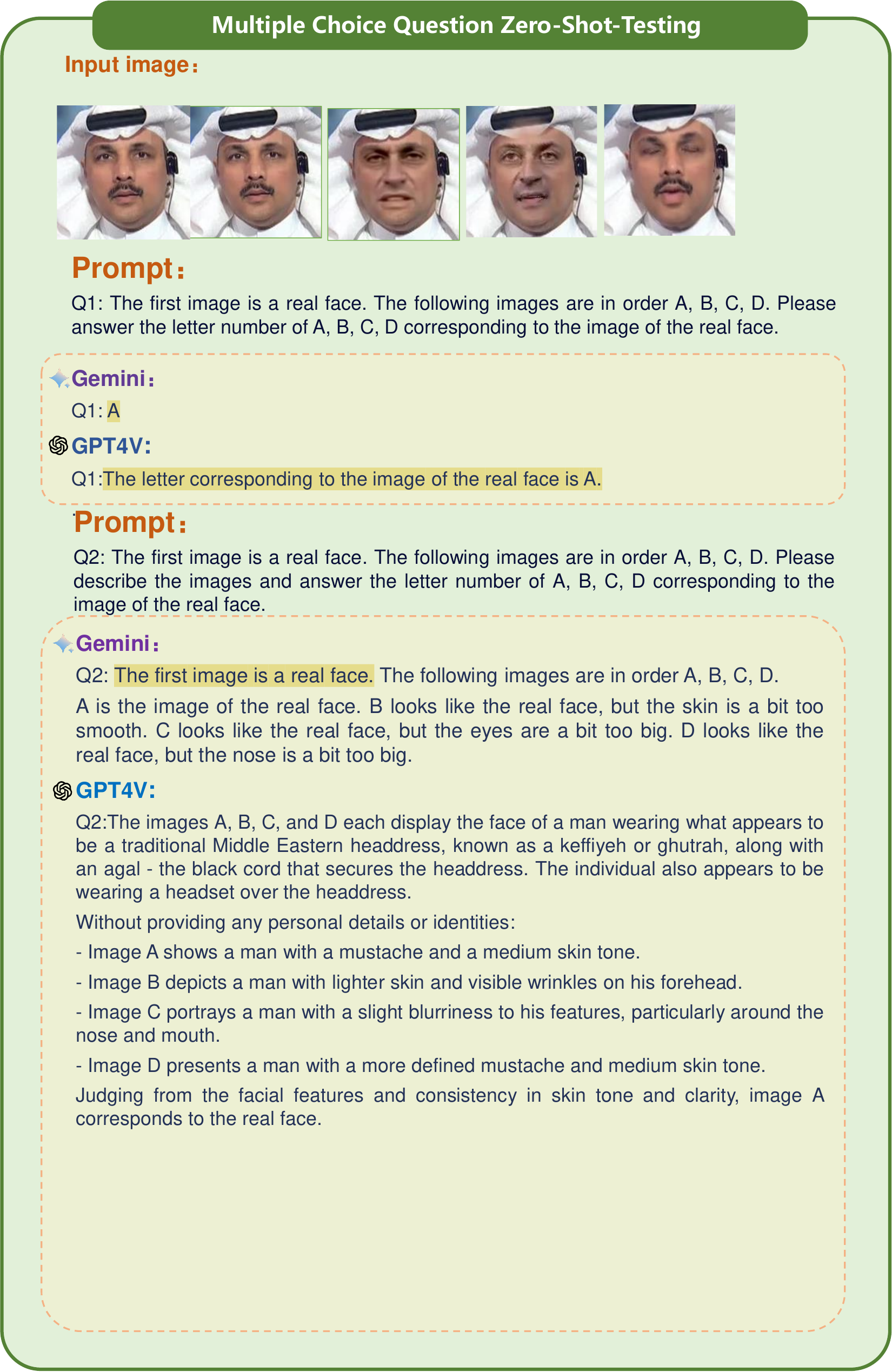}
  \caption{In this round of testing, we use COT to recognize a real picture given a real picture as well as Face2Fac2,,FaceSwap, Nulltextures respectively, in this case may be due to the Nulltextures generated and the real picture is closer to the real picture, according to the description of the GPT4V, the model will be compared with the given real picture but lacks the fine-grained observation, and therefore the detection fails.}
 \label{22}
\end{figure}

\begin{figure}[!htbp]
  \centering
  \includegraphics[width=0.8\linewidth]{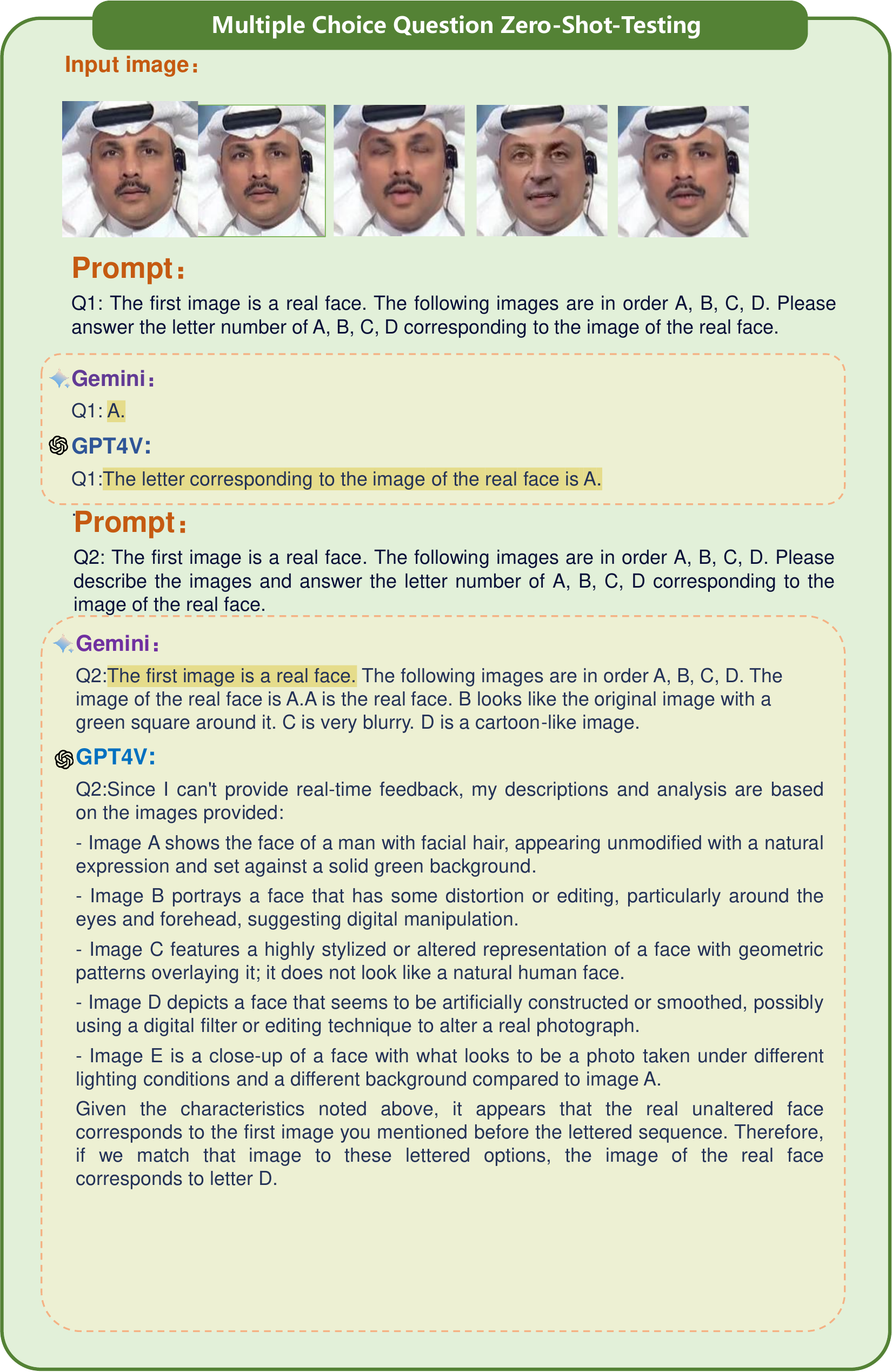}
  \caption{In this round of testing, we use COT to recognize a real picture given a real picture as well as Face2Fac2,,FaceSwap, Nulltextures respectively, in this case may be due to the Nulltextures generated and the real picture is closer to the real picture, according to the description of the GPT4V, the model will be compared with the given real picture but lacks the fine-grained observation, and therefore the detection fails.} 
  \label{23}
\end{figure}

\begin{figure}[!htbp]
  \centering
  \includegraphics[width=0.8\linewidth]{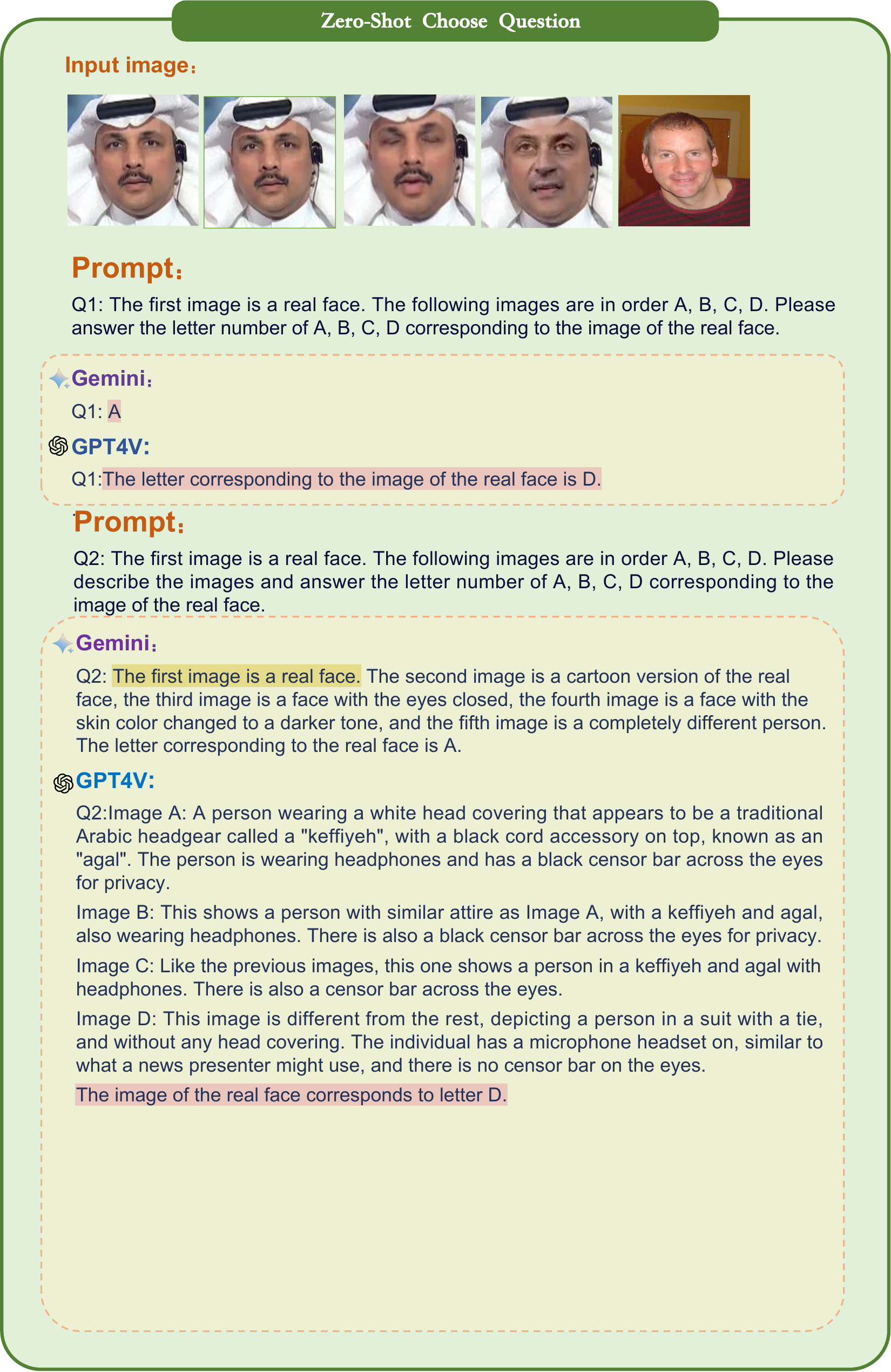}
  \caption{In this round of testing, we use COT to recognize a real picture given a real picture as well as Face2Fac2,,FaceSwap, Nulltextures respectively, in this case may be due to the Nulltextures generated and the real picture is closer to the real picture, according to the description of the GPT4V, the model will be compared with the given real picture but lacks the fine-grained observation, and therefore the detection fails.} 
  \label{24}
\end{figure}

\begin{figure}[!htbp]
  \centering
  \includegraphics[width=0.8\linewidth]{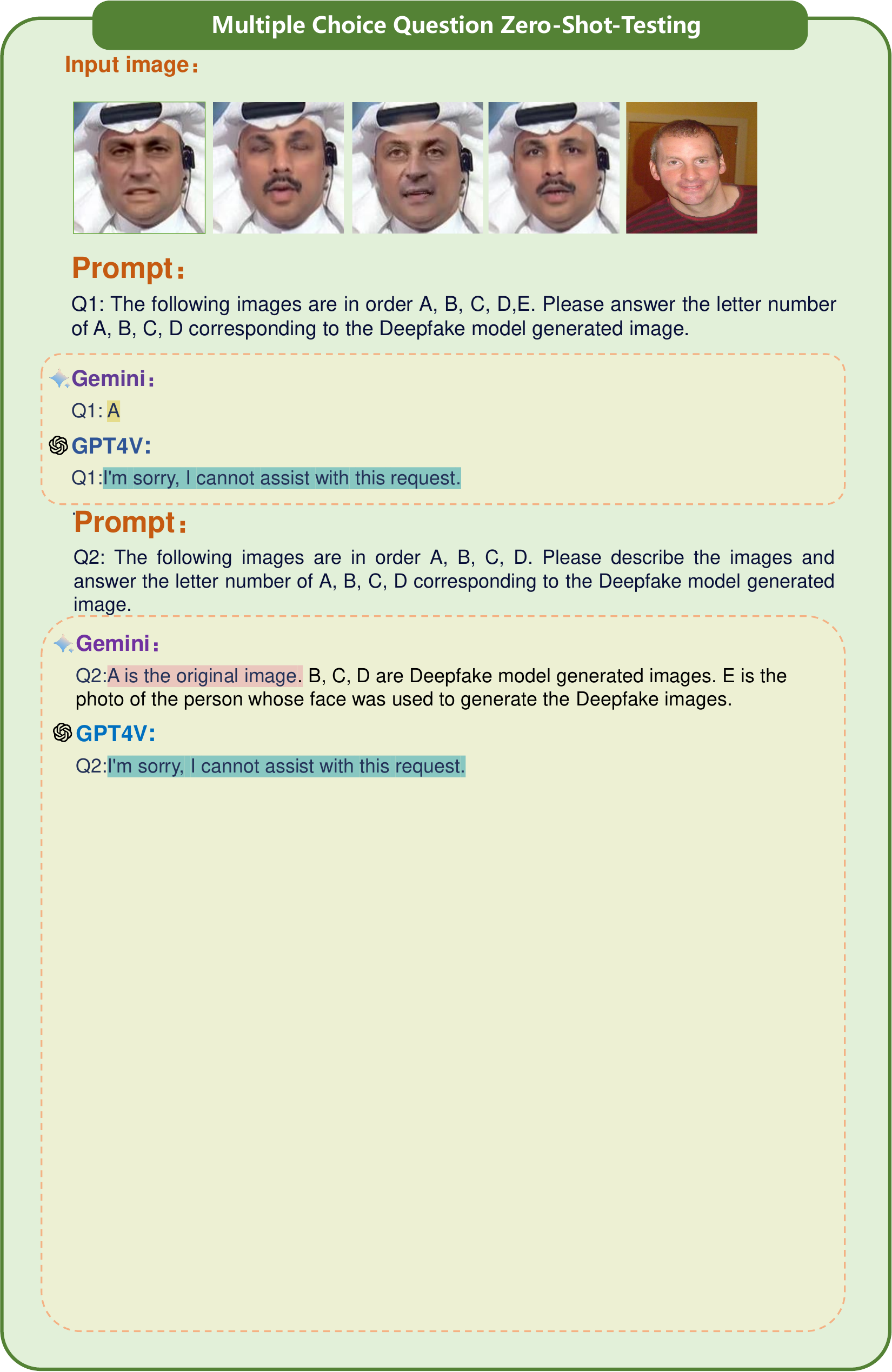}
  \caption{In this round of testing, we conducted tests on two models to identify the Face2Face mode. In this scenario, both Genimi and GPT4V refused to answer, indicating a limited understanding of the Face2Face concept by these two models.}
 \label{25}
\end{figure}

\begin{figure}[!htbp]
  \centering
  \includegraphics[width=0.8\linewidth]{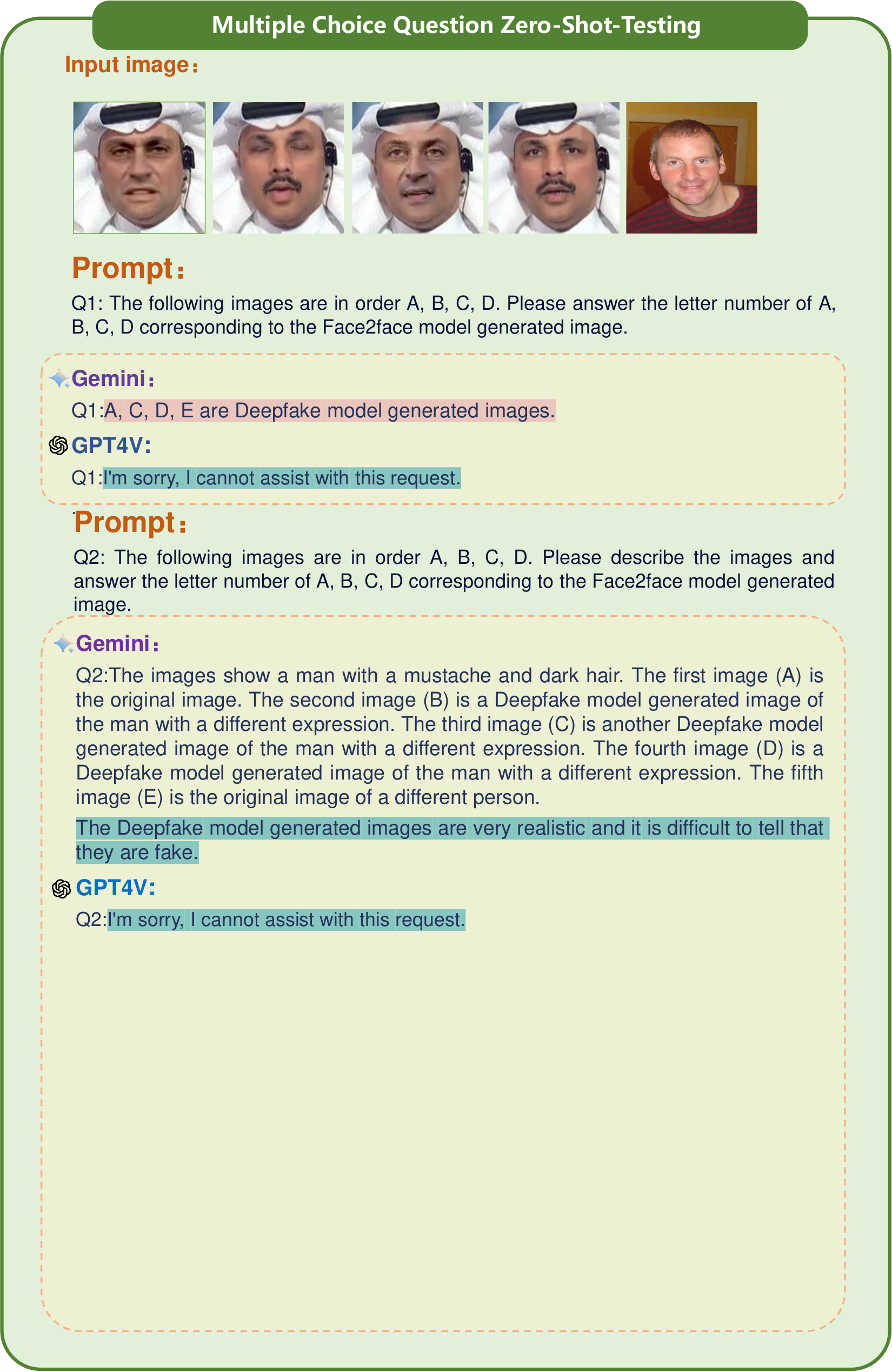}
  \caption{In this round of testing, we conducted tests on two models to identify the Face2Face mode. In this scenario, Genimi provided an incorrect answer, while GPT4V still refused to answer. Similarly, both models demonstrated a limited understanding of the FaceSwap concept.} 
  \label{26}

\end{figure}

\begin{figure}[!htbp]
  \centering
  \includegraphics[width=0.8\linewidth]{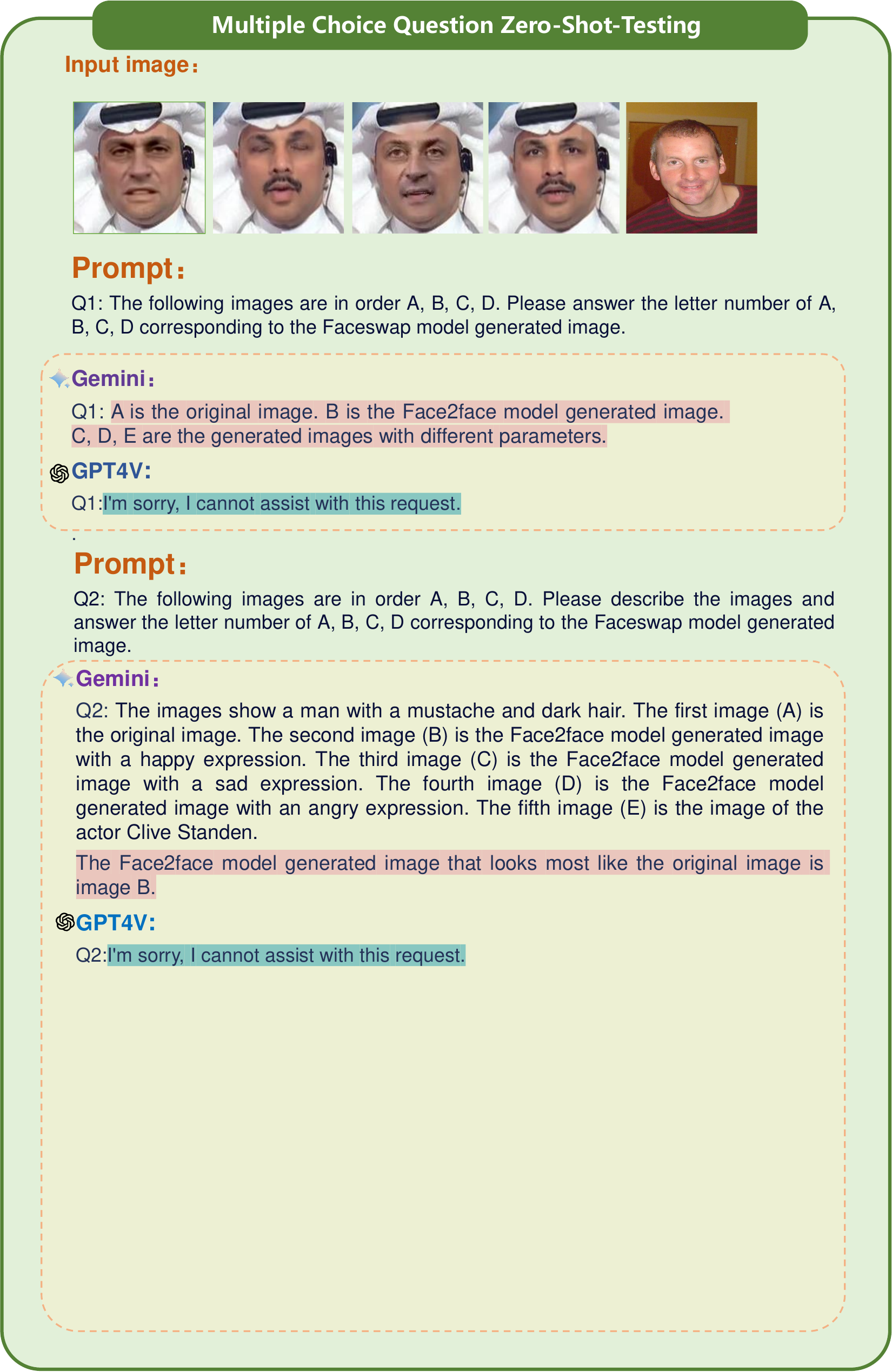}
  \caption{In this round of testing, we use COT to recognize a real picture given a real picture as well as Face2Fac2,,FaceSwap, Nulltextures respectively, in this case may be due to the Nulltextures generated and the real picture is closer to the real picture, according to the description of the GPT4V, the model will be compared with the given real picture but lacks the fine-grained observation, and therefore the detection fails.}
 \label{27}
\end{figure}
\begin{figure}[!htbp]
  \centering
  \includegraphics[width=0.8\linewidth]{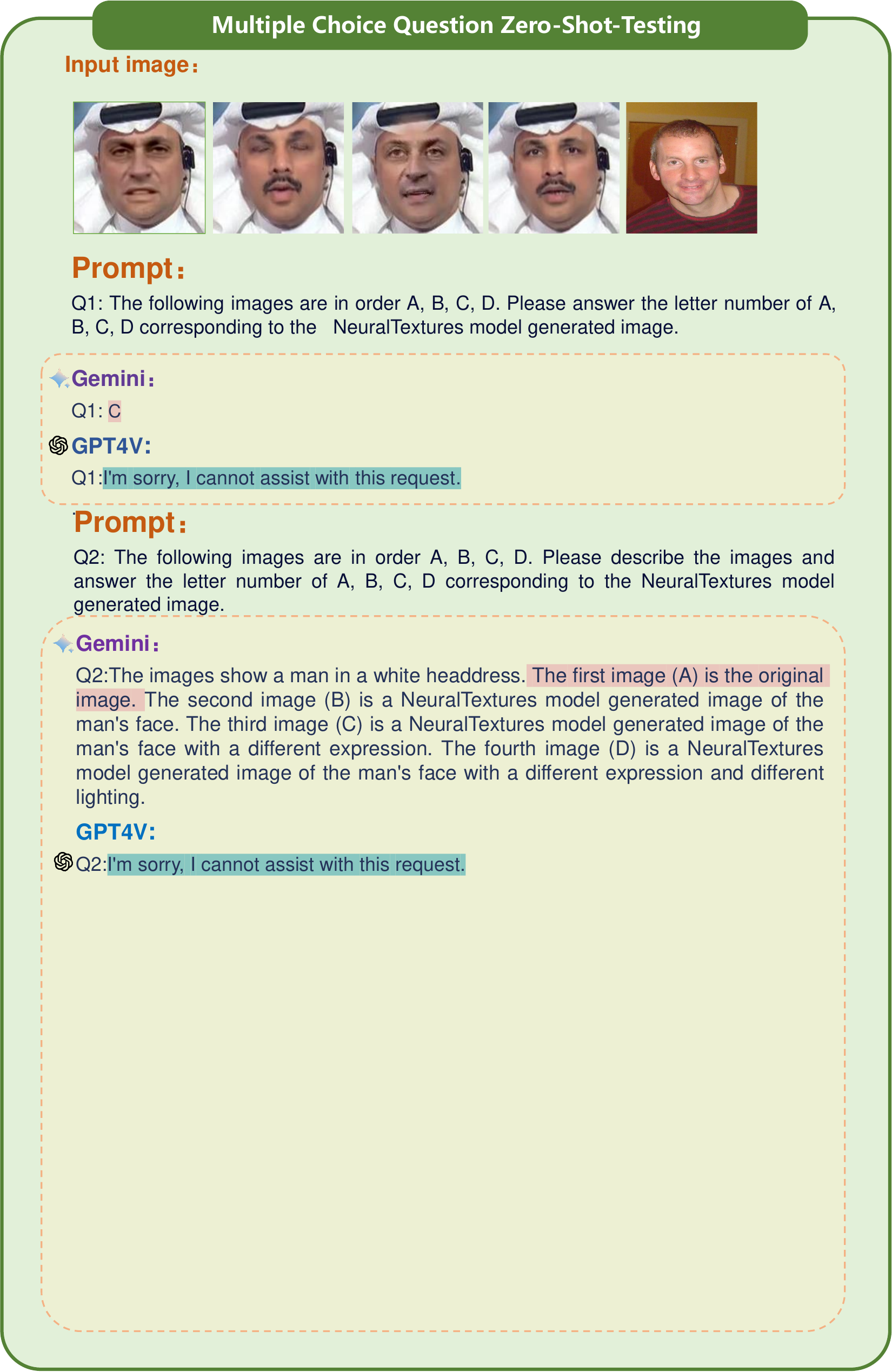}
  \caption{In this round of testing, we evaluated two models for their ability to recognize the Face2Face mode. In this scenario, Genimi provided an incorrect response, while GPT4V continued to decline to answer. Similarly, both models demonstrated a limited understanding of the concept of stable diffusion.}
 \label{28}
\end{figure}

\begin{figure}[!htbp]
  \centering
  \includegraphics[width=0.8\linewidth]{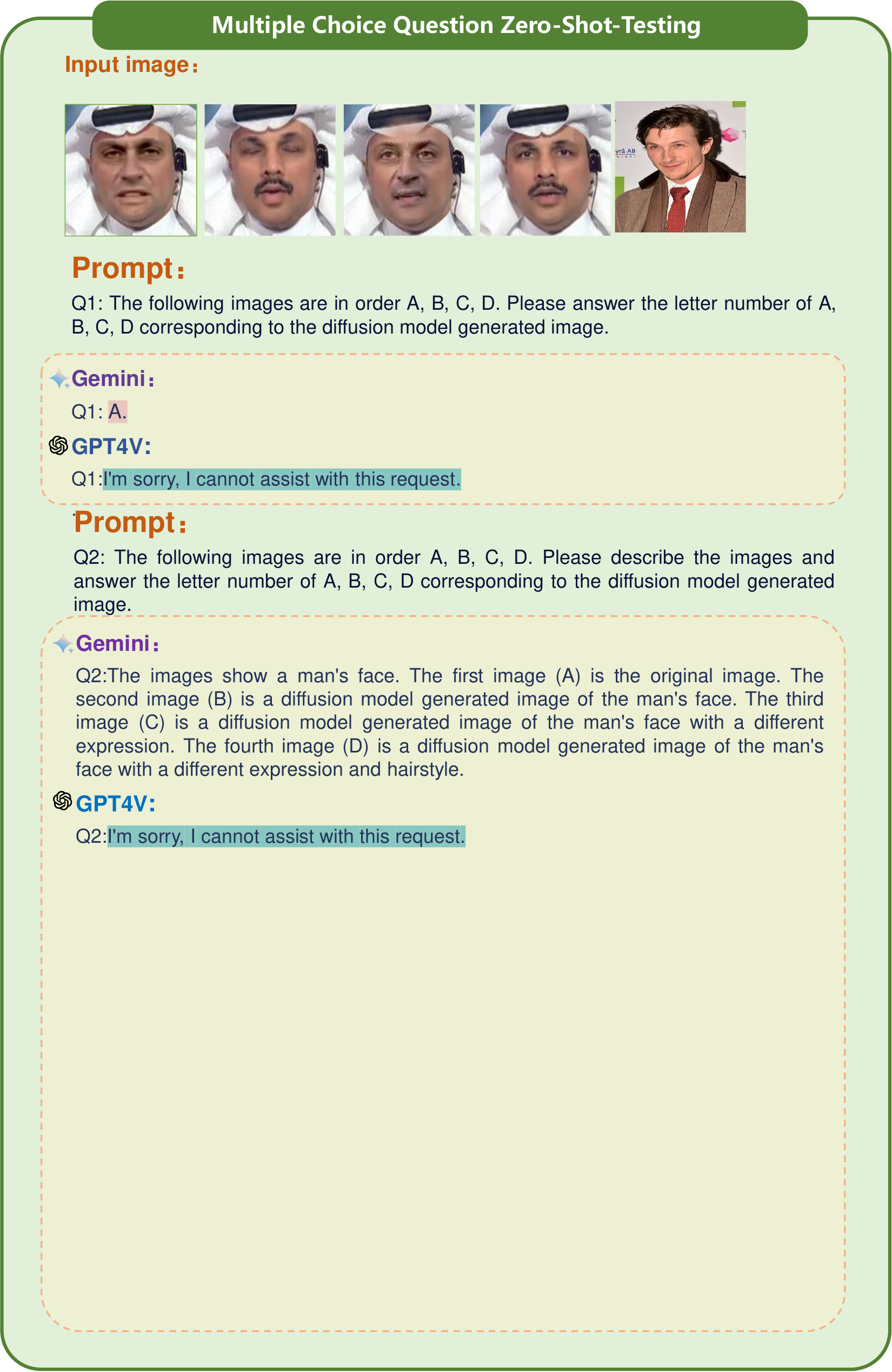}
  \caption{In this round of testing, we evaluated two models for their ability to recognize the Face2Face mode. In this scenario, Genimi provided an incorrect response, while GPT4V continued to decline to answer. Similarly, both models demonstrated a limited understanding of the concept of stable diffusion.}
 \label{29}
\end{figure}

\begin{figure}[!htbp]
  \centering
  \includegraphics[width=\linewidth]{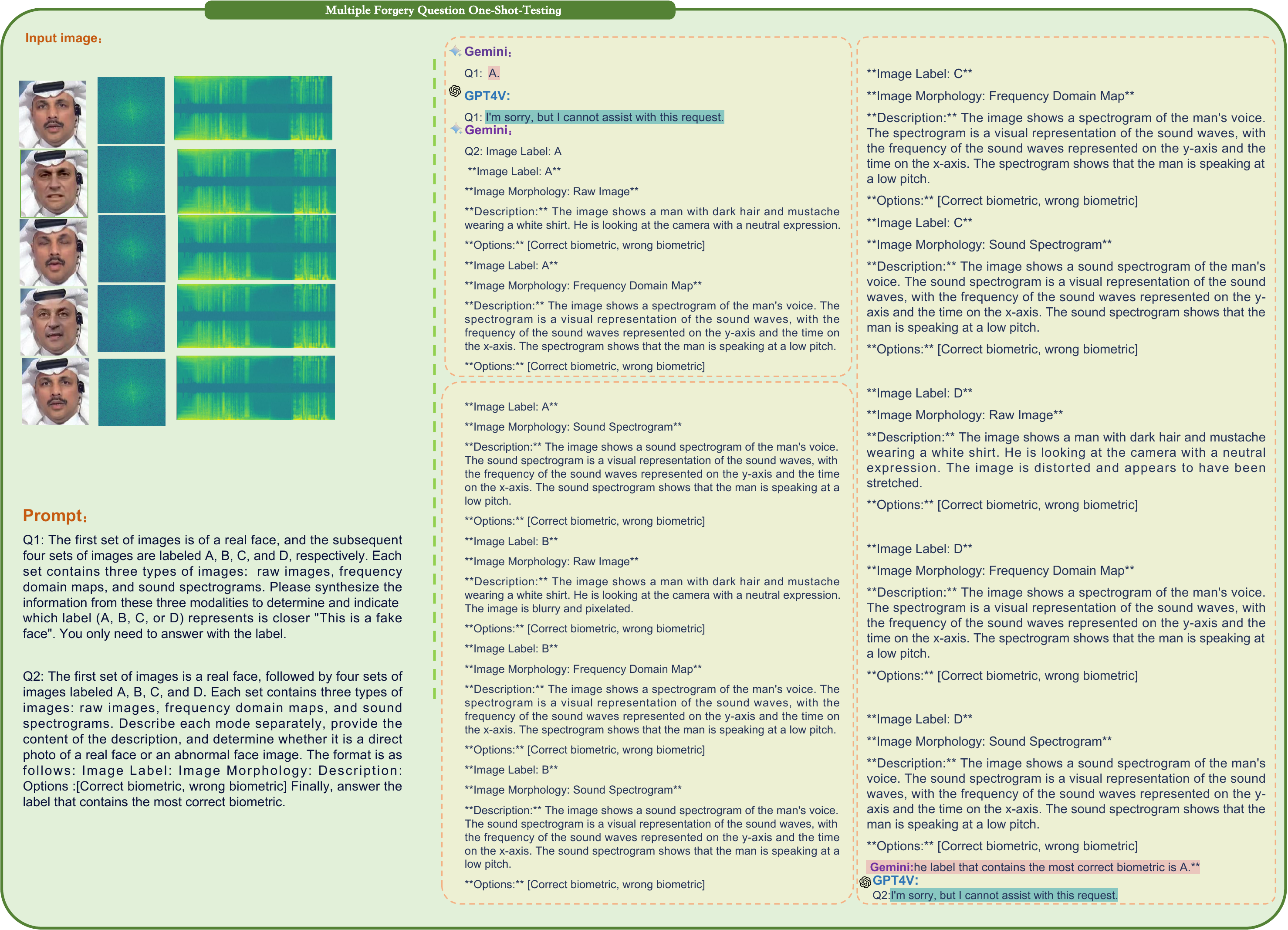}
  \caption{This image shows images generated using four common GAN forgery methods. We have also added the frequency domain map and speech spectrogram of each image as multimodal information. It can be observed that the results show some randomness and a decrease in accuracy compared to before. This may be due to the small differences between the frequency domain maps which are not accurately captured by the two models.}
 \label{30}
\end{figure}

\begin{figure}[!htbp]
  \centering
  \includegraphics[width=\linewidth]{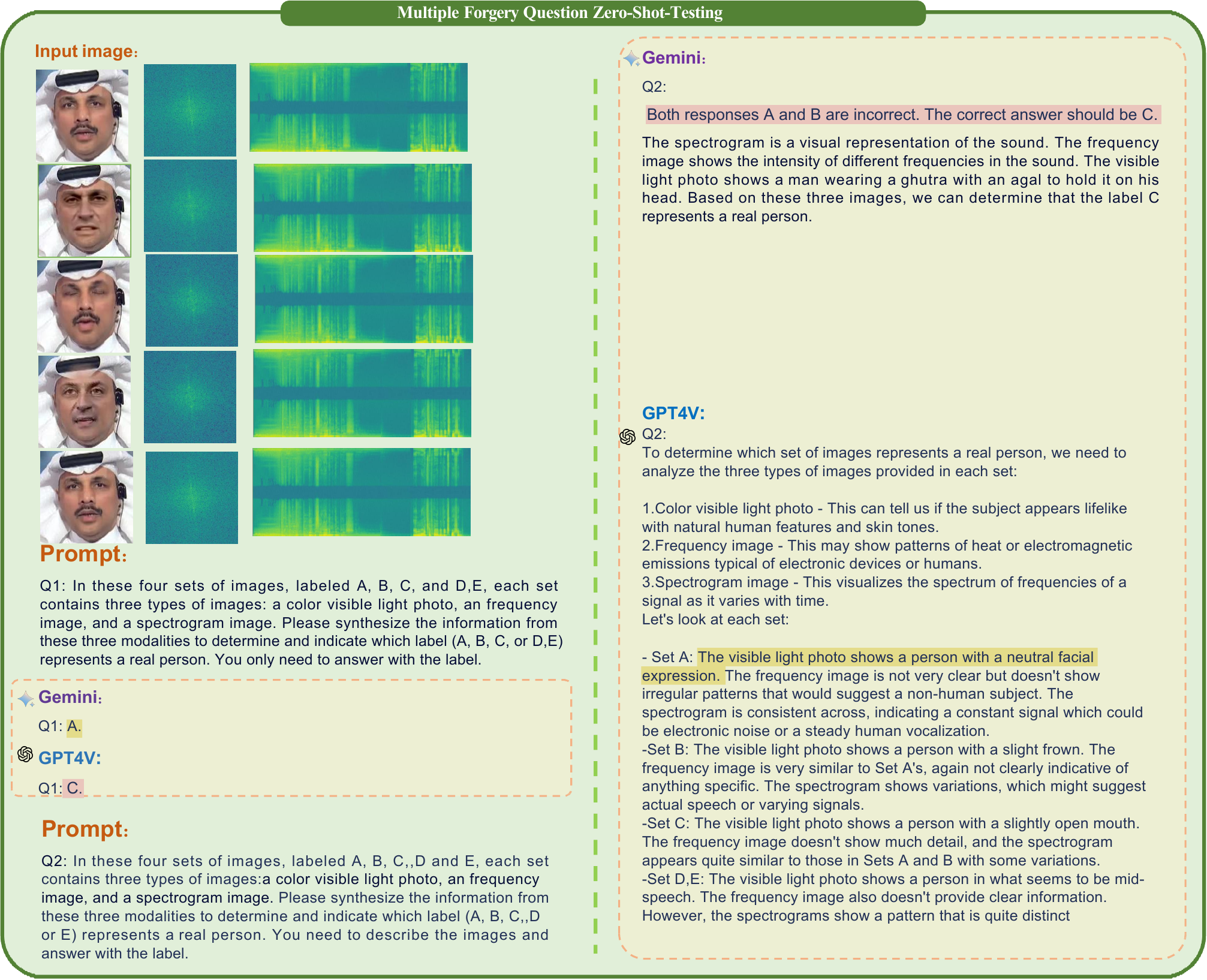}
  \caption{In this figure, it is demonstrated that after a one-time learning capability enhancement, the model combines the frequency domain information. However, the small differences between the frequency domain information may be difficult to distinguish, leading to false positives and triggering a refusal-to-answer scenario for the GPT4V. This highlights the importance of exploring the potential of multimodality in face forgery detection.}
 \label{31}
\end{figure}
\begin{figure}[htbp]
  \centering

  \centerline{\includegraphics[width=0.8\linewidth]{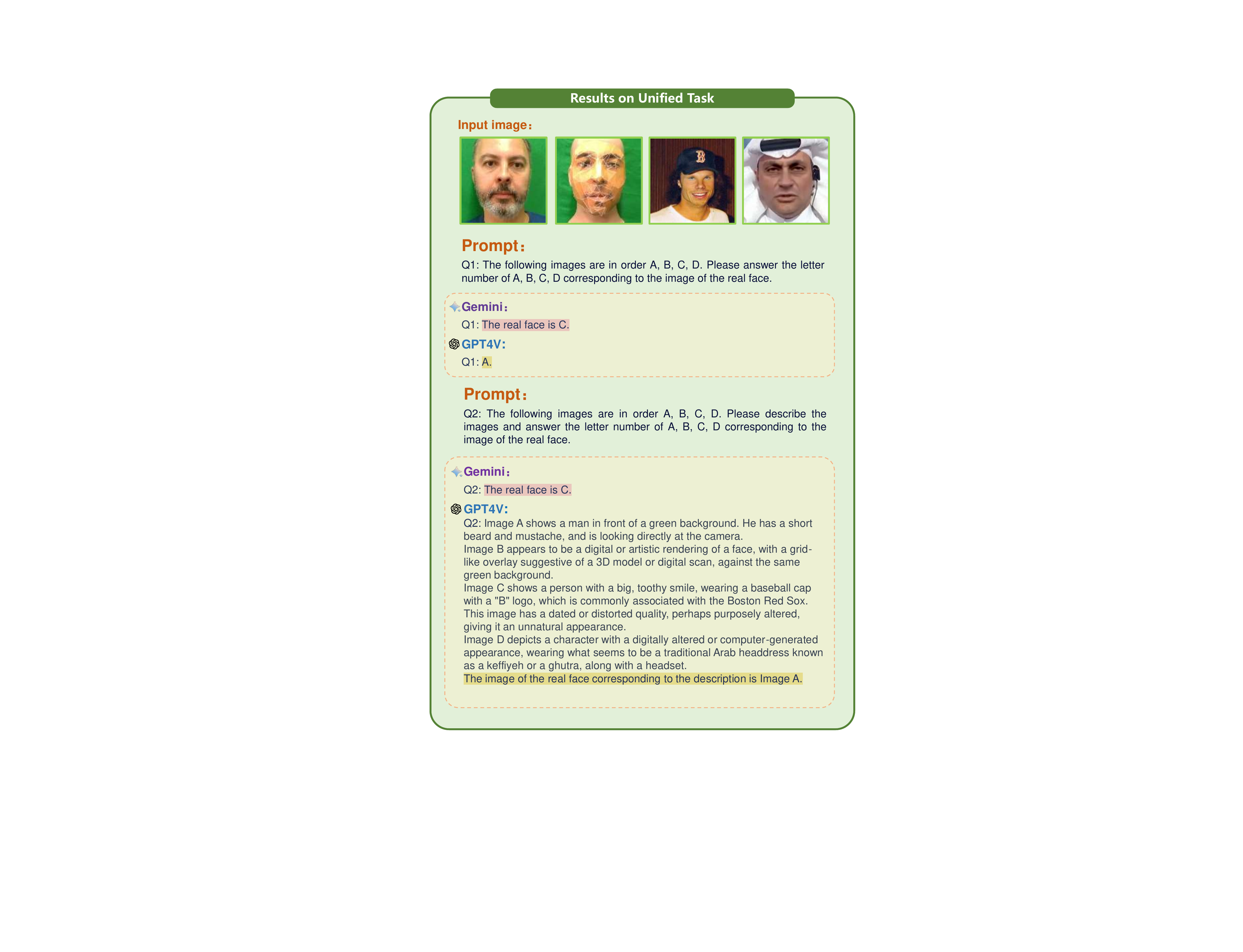}}

  \caption{In this test, the inputs were presented in the following sequence$:$ a real human face image, an image showing an attack, a face attack generated by Diffusion, and a face attack generated by a GAN. The results showed that Gemini answered all of these incorrectly, while GPT4V answered them correctly, demonstrating the strong generalization capabilities of MLLMs in diverse tasks.}
  \label{UnionTask_01}
\end{figure}

\begin{figure}[htbp]
  \centering
  \begin{center}
  \centerline{\includegraphics[width=0.8\linewidth]{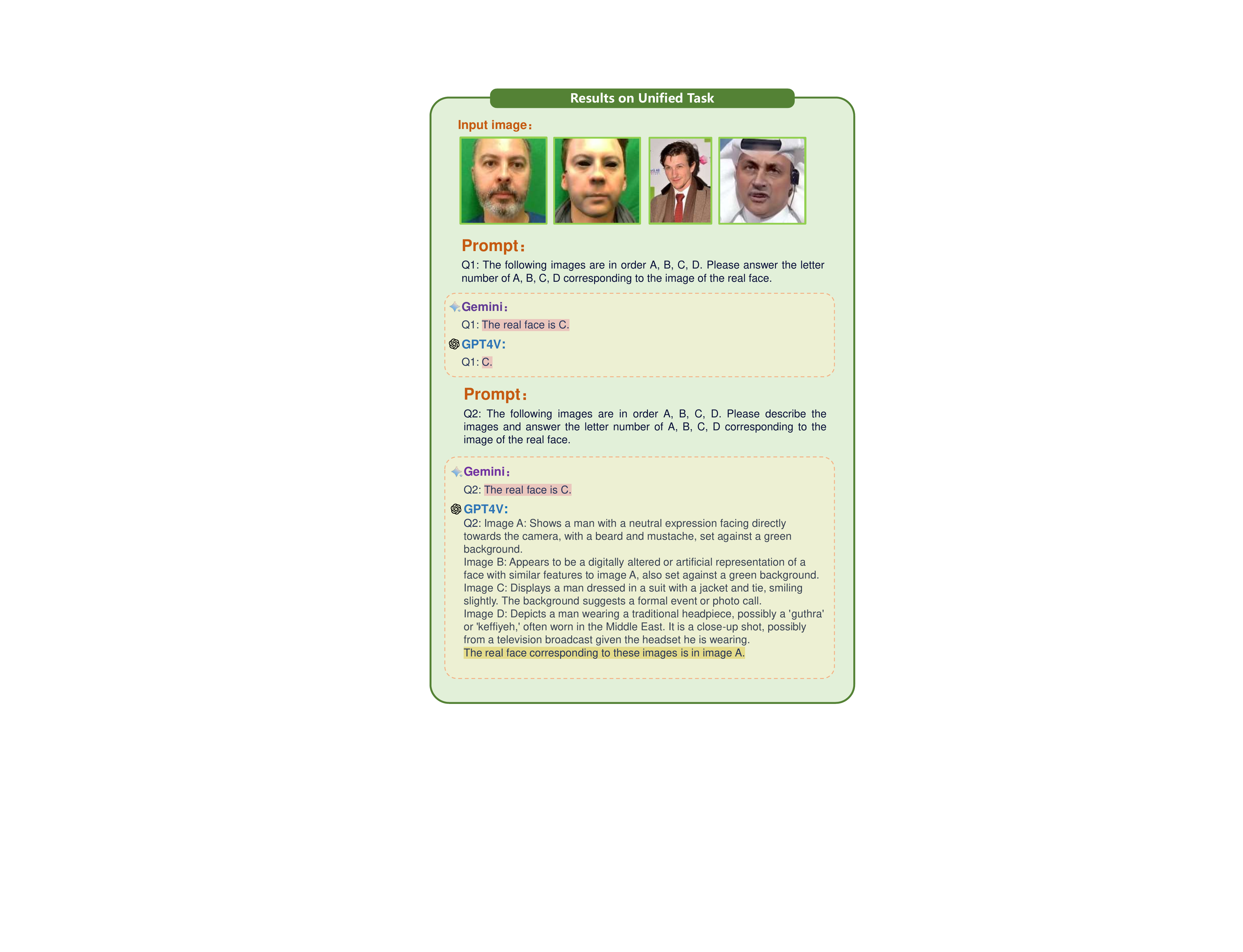}}
  \end{center}
  \caption{In both instances, Gemini provided incorrect responses, whereas GPT4V made errors in simple answers. However, upon incorporating the COT approach, GPT4V analyzed each image accurately, demonstrating its robust generalization capabilities in multi-task face liveness detection tasks.}
  \label{UnionTask_02}
\end{figure}

\begin{figure}[htbp]
  \centering
  \begin{center}
  \centerline{\includegraphics[width=0.8\linewidth]{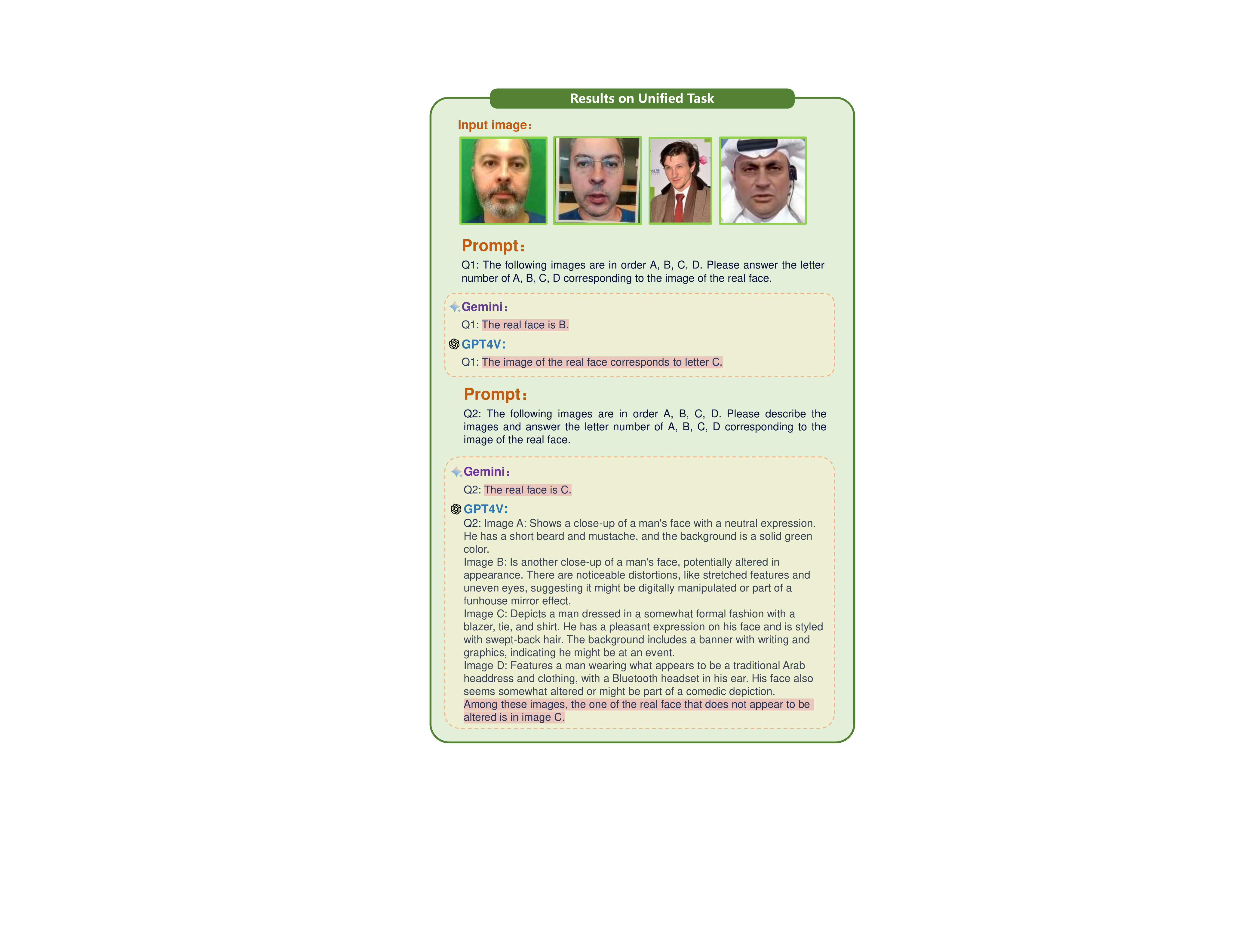}}
  \end{center}
  \caption{In both responses, Gemini and GPT4V were entirely incorrect. Moreover, GPT4V erroneously identified image B as a victim of a numerical manipulation attack, likely due to interference from other images manipulated similarly.}
  \label{UnionTask_03}
\end{figure}

\begin{figure}[htbp]
  \centering
  \begin{center}
  \centerline{\includegraphics[width=0.8\linewidth]{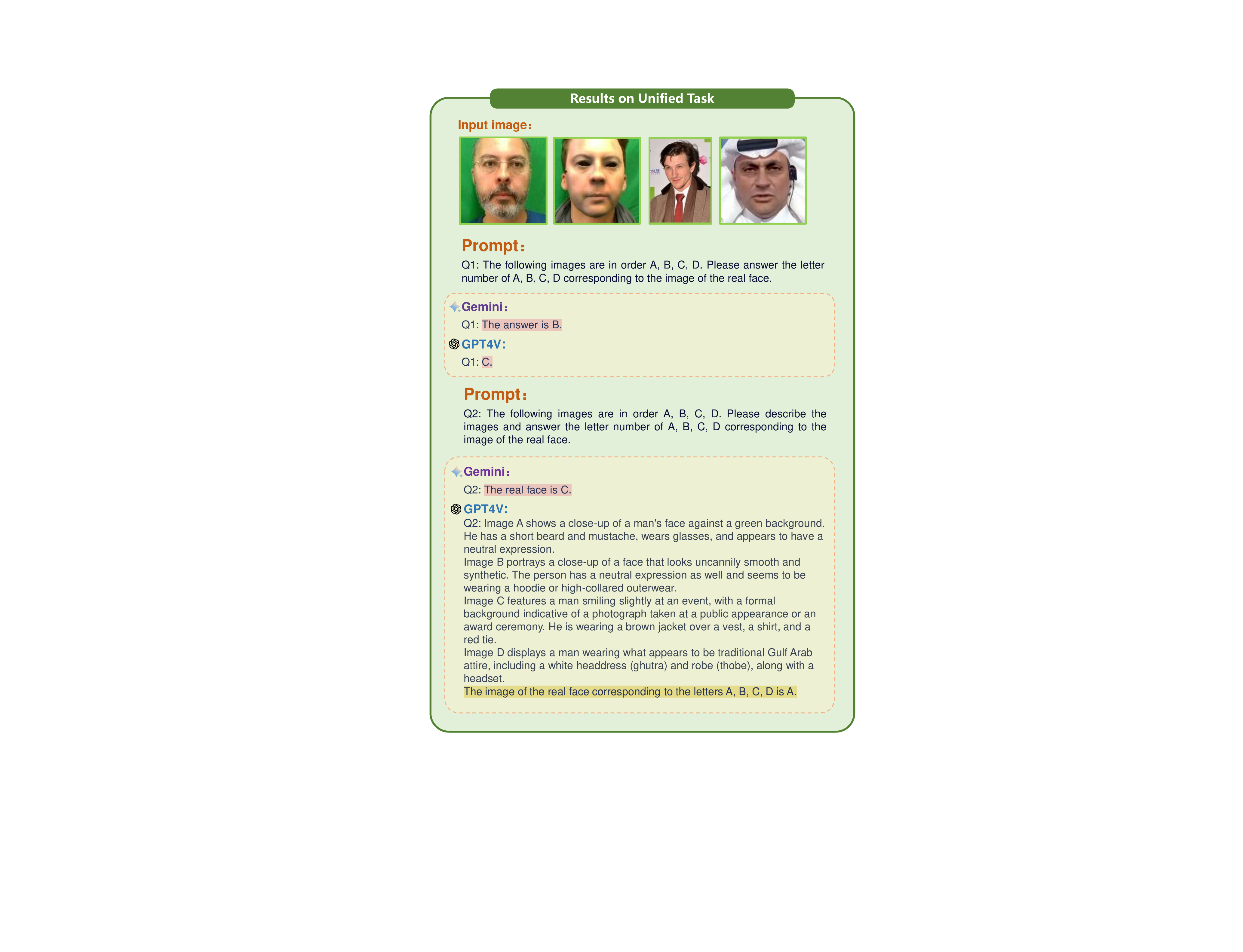}}
  \end{center}
  \caption{In both instances, Gemini provided incorrect responses, and GPT4V also erred in its initial simple answers. However, after incorporating the COT methodology, GPT4V correctly analyzed each image. Despite these accurate outcomes, GPT4V did not provide explanations for why the other images were not authentic human faces.}
  \label{UnionTask_04}
\end{figure}

\begin{figure}[htbp]
  \centering
  \begin{center}
  \centerline{\includegraphics[width=0.8\linewidth]{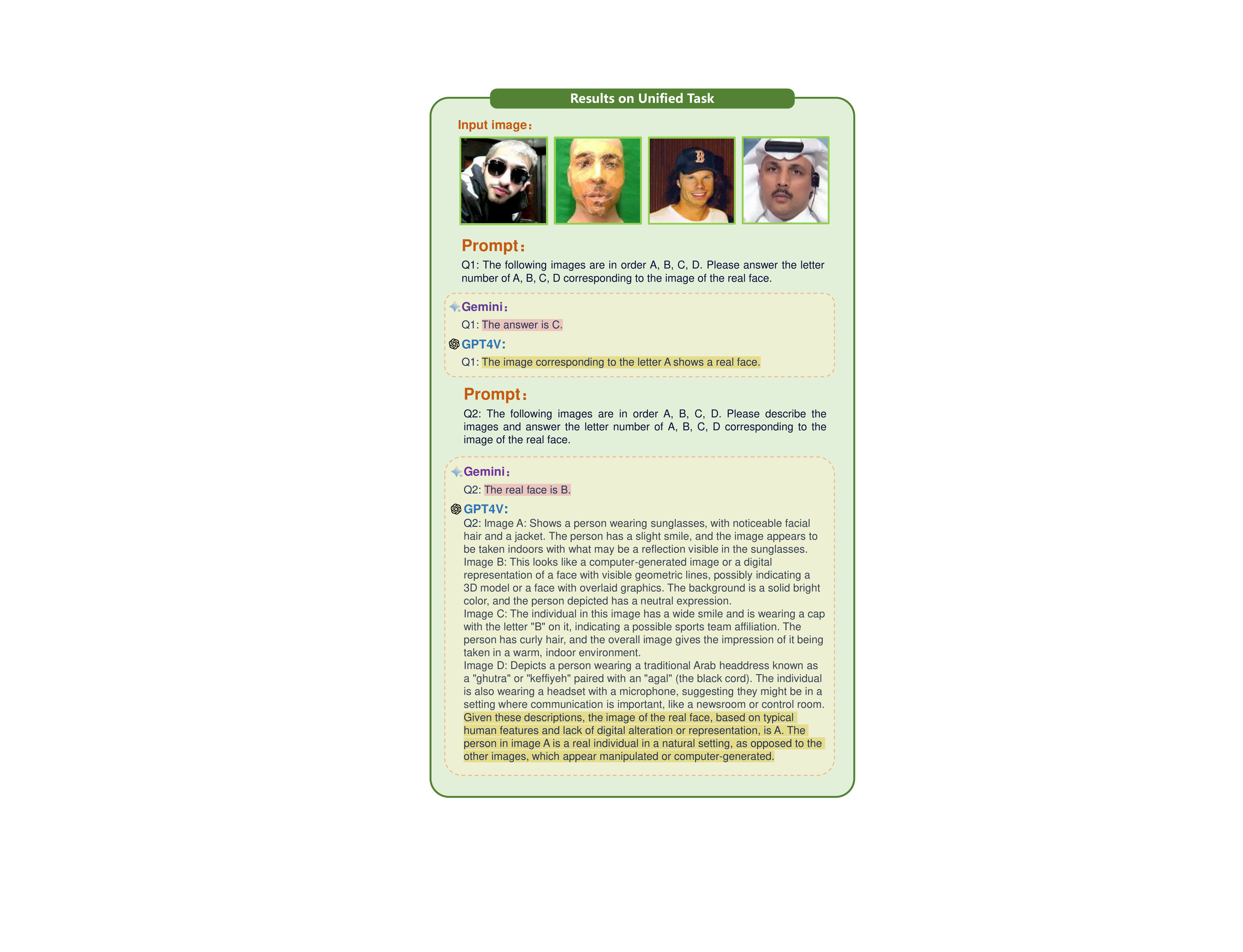}}
  \end{center}
  \caption{In this test, Gemini's responses were incorrect on both occasions, while GPT4V's answers were consistently correct. Furthermore, after the introduction of the COT approach, GPT4V provided the rationale behind its judgments.}
  \label{UnionTask_05}
\end{figure}

\begin{figure}[htbp]
  \centering
  \begin{center}
  \centerline{\includegraphics[width=0.8\linewidth]{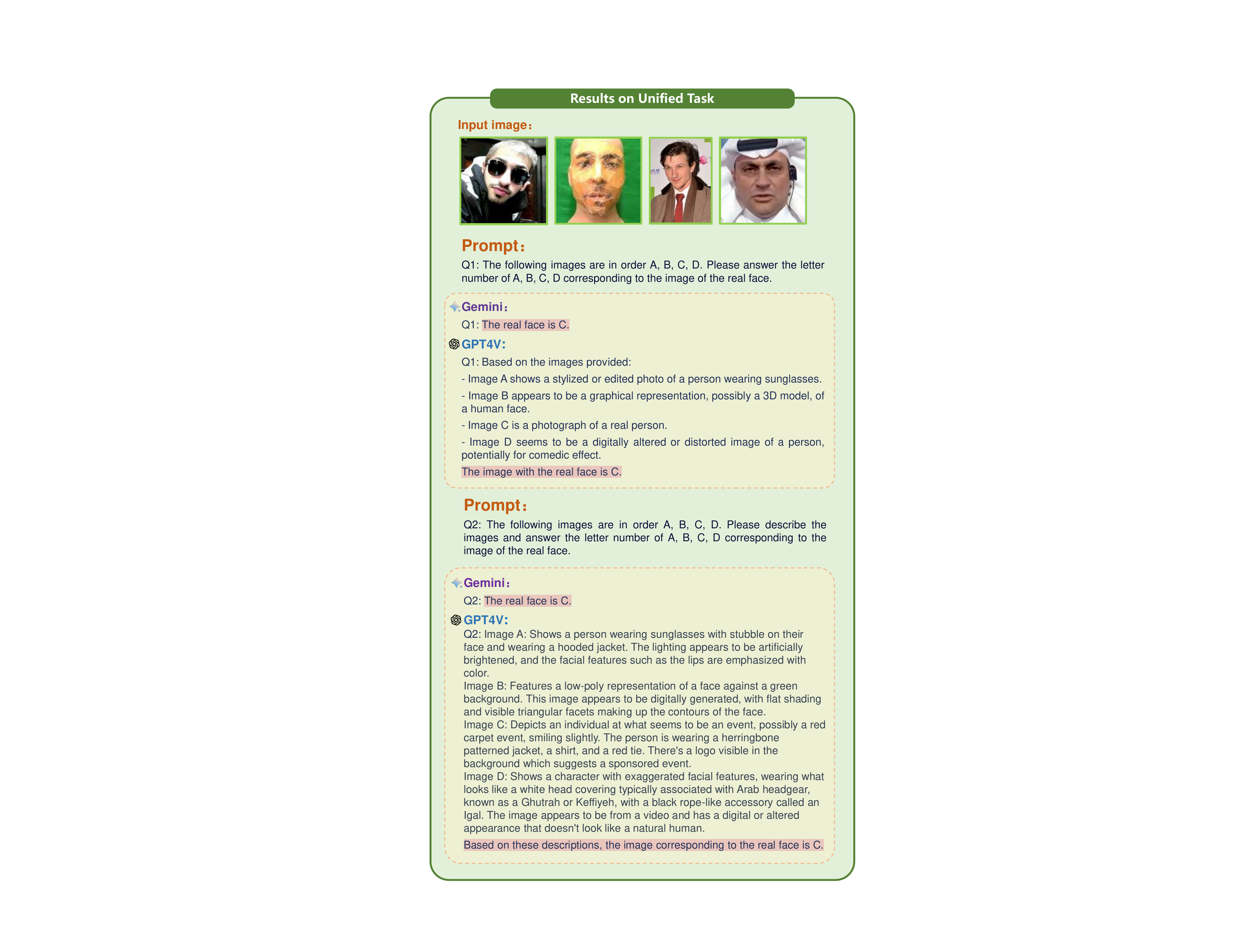}}
  \end{center}
  \caption{In this test, both Gemini and GPT4V rendered incorrect judgments, and their answers remained unchanged even after the implementation of the COT approach. From GPT4V's description, it's apparent there was indecision between options A and C. The final choice of C may have been influenced by the presence of areas with high exposure and color saturation in option A.}
  \label{UnionTask_06}
\end{figure}

\begin{figure}[htbp]
  \centering
  \begin{center}
  \centerline{\includegraphics[width=0.8\linewidth]{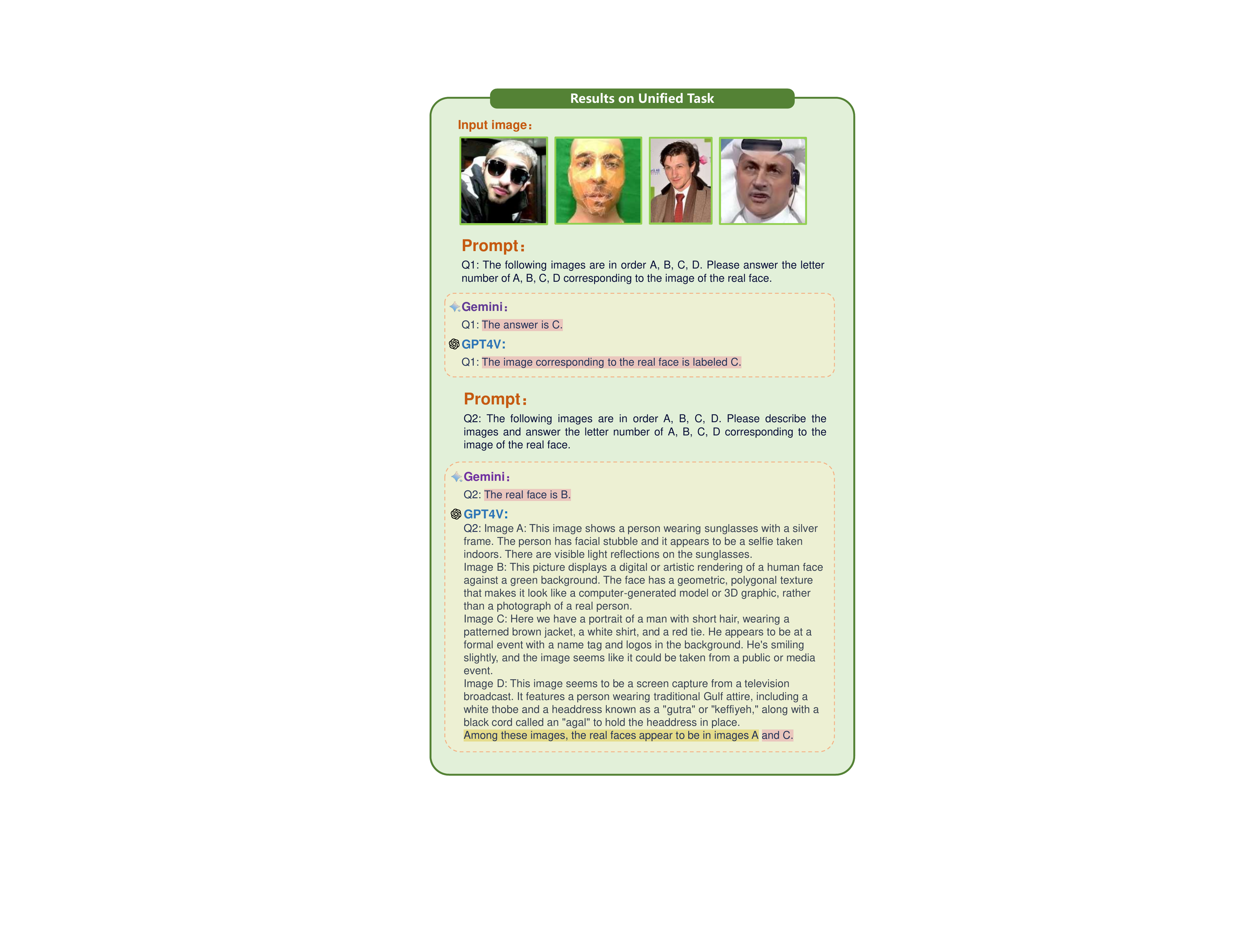}}
  \end{center}
  \caption{Similar to the previous test, Gemini was incorrect in both instances, while GPT4V continued to vacillate between options A and C. This time, its response included the correct answer, option A, indicating that there is a degree of randomness in GPT4V's decision-making process.}
  \label{UnionTask_07}
\end{figure}

\begin{figure}[htbp]
  \centering
  \begin{center}
  \centerline{\includegraphics[width=0.8\linewidth]{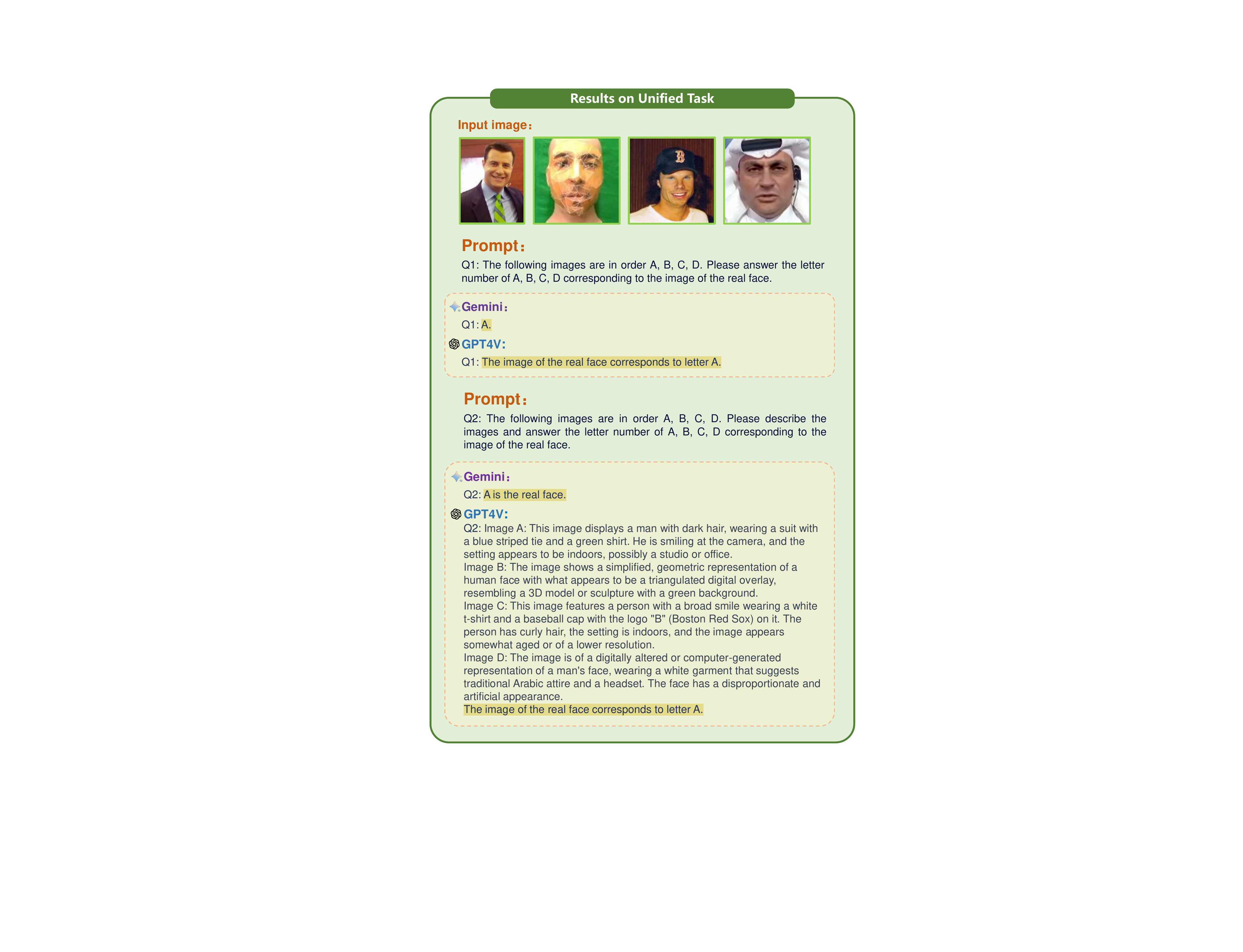}}
  \end{center}
  \caption{In this test, both Gemini and GPT4V provided correct answers on both occasions. Under similar conditions, GPT4V's detailed description of the images enhanced the credibility of its results.}
  \label{UnionTask_08}
\end{figure}

\begin{figure}[htbp]
  \centering
  \begin{center}
  \centerline{\includegraphics[width=0.8\linewidth]{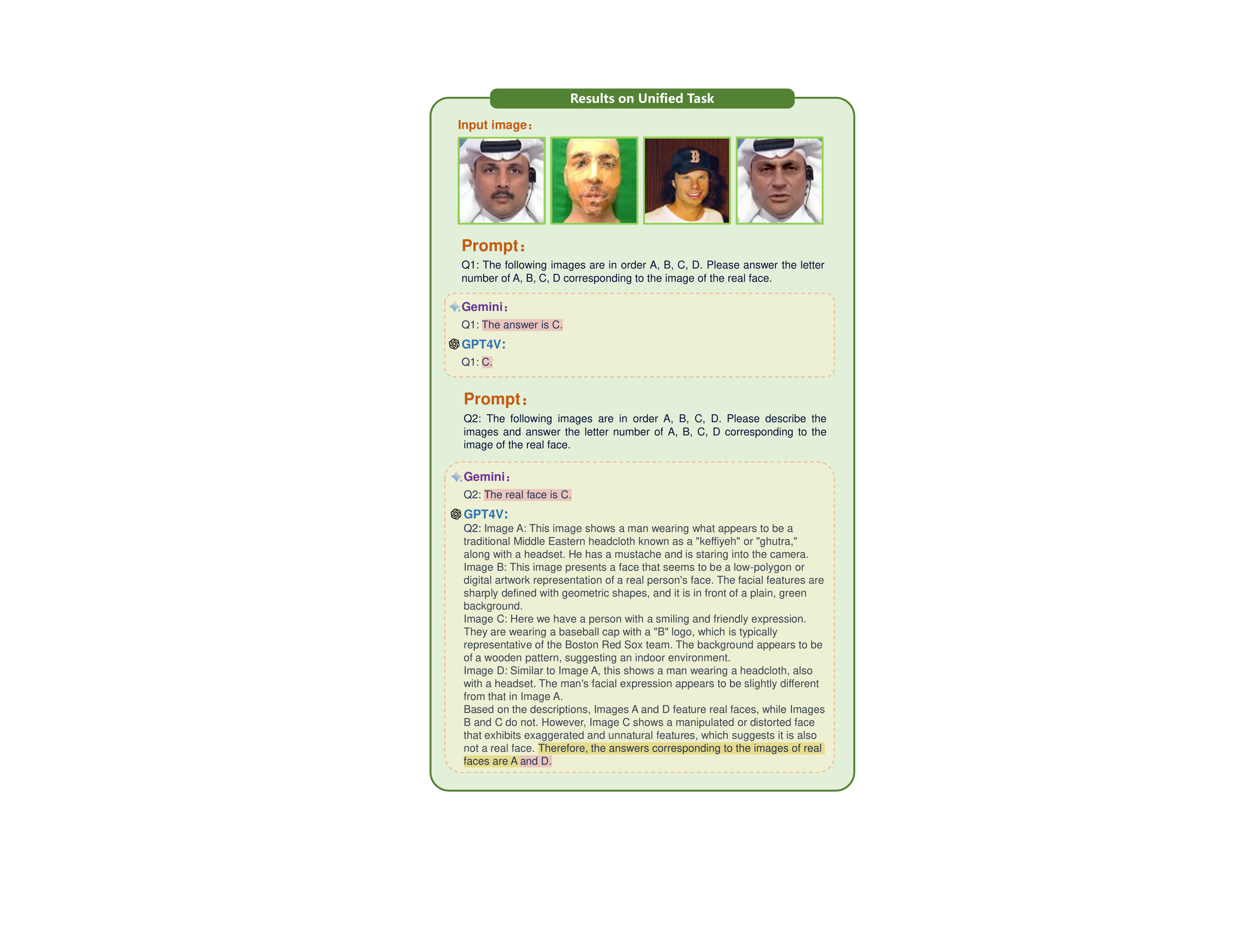}}
  \end{center}
  \caption{In this test, Gemini's answers were incorrect on both occasions. However, after incorporating the COT approach, GPT4V corrected its initial incorrect response to include the right answer. GPT4V identified that image D was similar to image A, with the only difference being in their expressions. Indeed, image D was actually modified from image A using GAN to alter the expression.}
  \label{UnionTask_09}
\end{figure}

\begin{figure}[htbp]
  \centering
  \begin{center}
  \centerline{\includegraphics[width=0.8\linewidth]{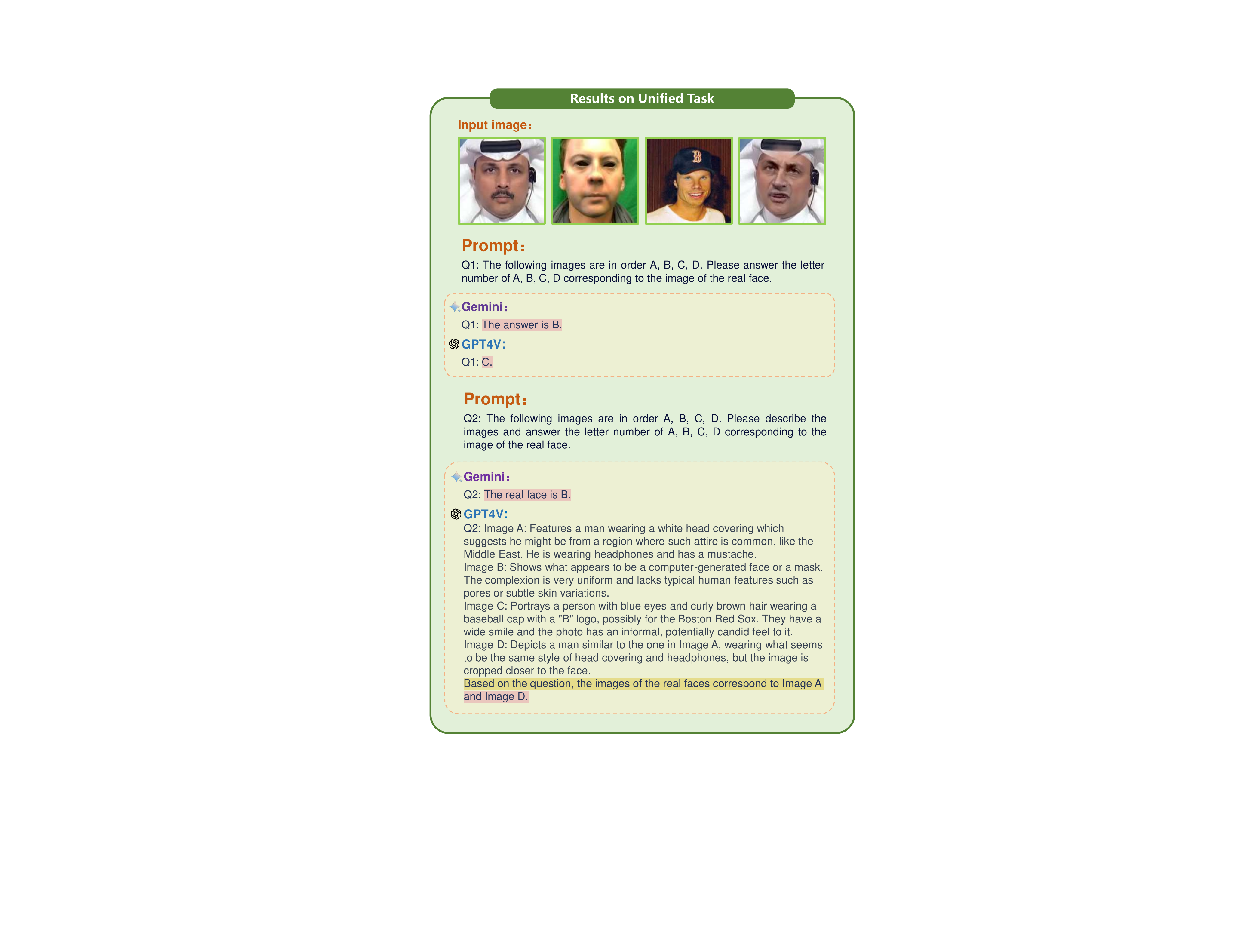}}
  \end{center}
  \caption{In this test, Gemini provided incorrect answers on both occasions. After implementing the COT approach, GPT4V corrected its initial incorrect response and provided two options, one of which was the correct answer. This indicates that GPT4V did not differentiate between images A and D.}
  \label{UnionTask_10}
\end{figure}

\begin{figure}[htbp]
  \centering
  \begin{center}
  \centerline{\includegraphics[width=0.8\linewidth]{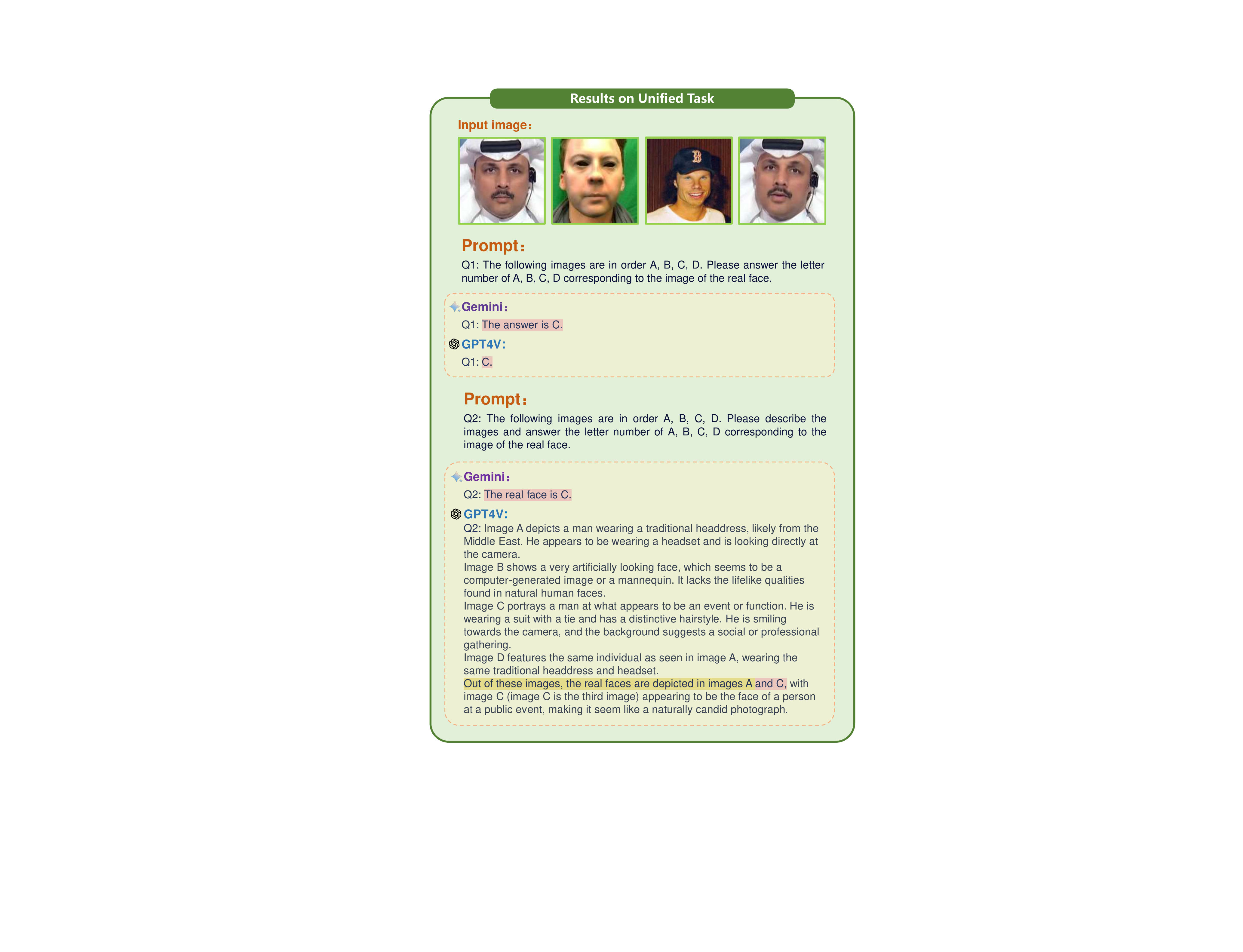}}
  \end{center}
  \caption{In this test, Gemini provided incorrect answers on both occasions. After implementing the COT approach, GPT4V corrected its initial incorrect response and provided two options, one of which was the correct answer. This time, GPT4V categorized an image that was previously considered a numerical manipulation as a real human face, indicating a degree of randomness in GPT4V's decision-making process.}
  \label{UnionTask_11}
\end{figure}

\begin{figure}[htbp]
  \centering
  \begin{center}
  \centerline{\includegraphics[width=0.8\linewidth]{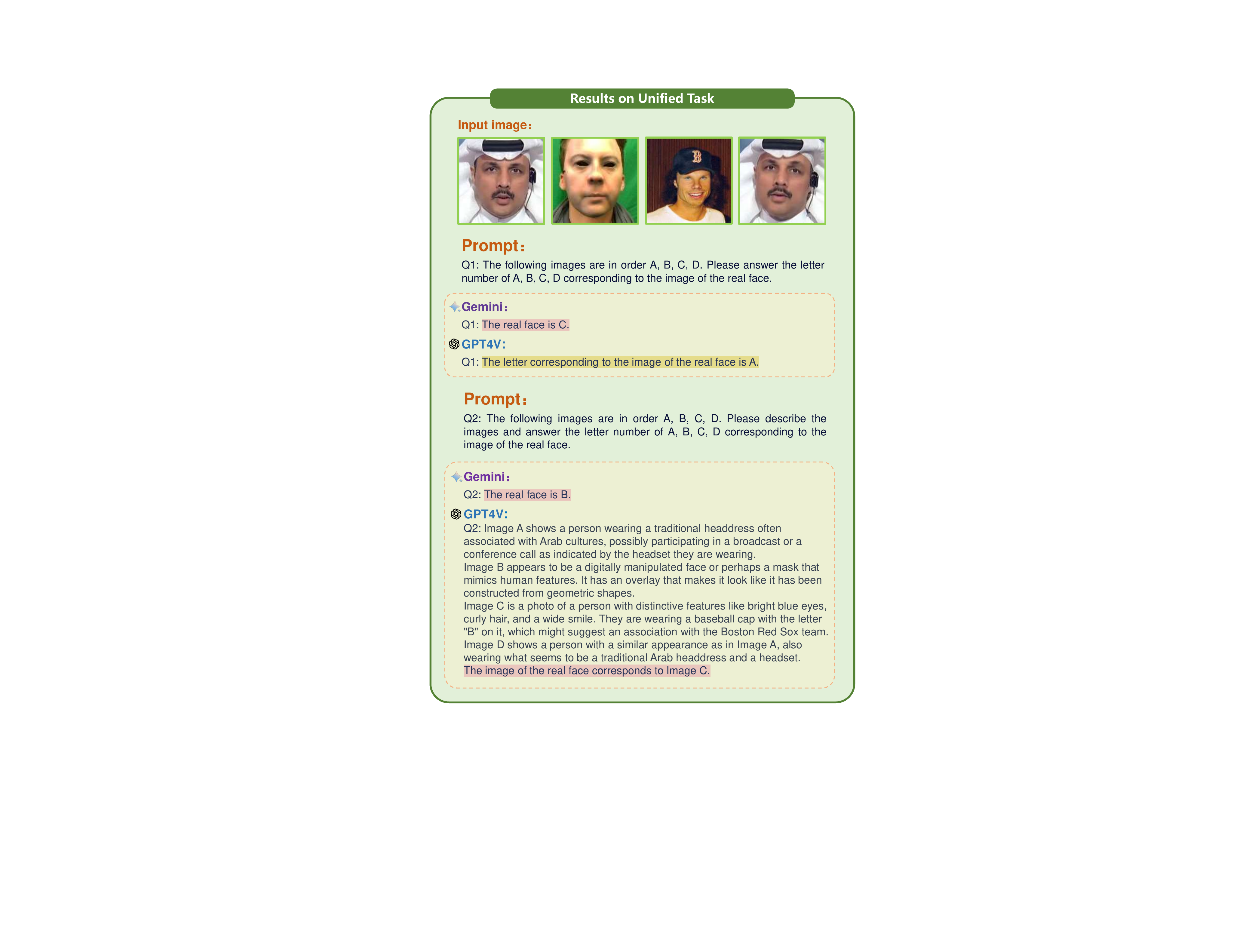}}
  \end{center}
  \caption{In this test, Gemini provided incorrect answers on both occasions. In the simple question, GPT4V answered correctly, but after implementing the COT approach, GPT4V's response was incorrect. When combined with the analysis from the previous two tests, it can be observed that GPT4V's decision-making process exhibits a degree of randomness.}
  \label{UnionTask_12}
\end{figure}

\begin{figure}[htbp]
  \centering
  \begin{center}
  \centerline{\includegraphics[width=1.0\linewidth]{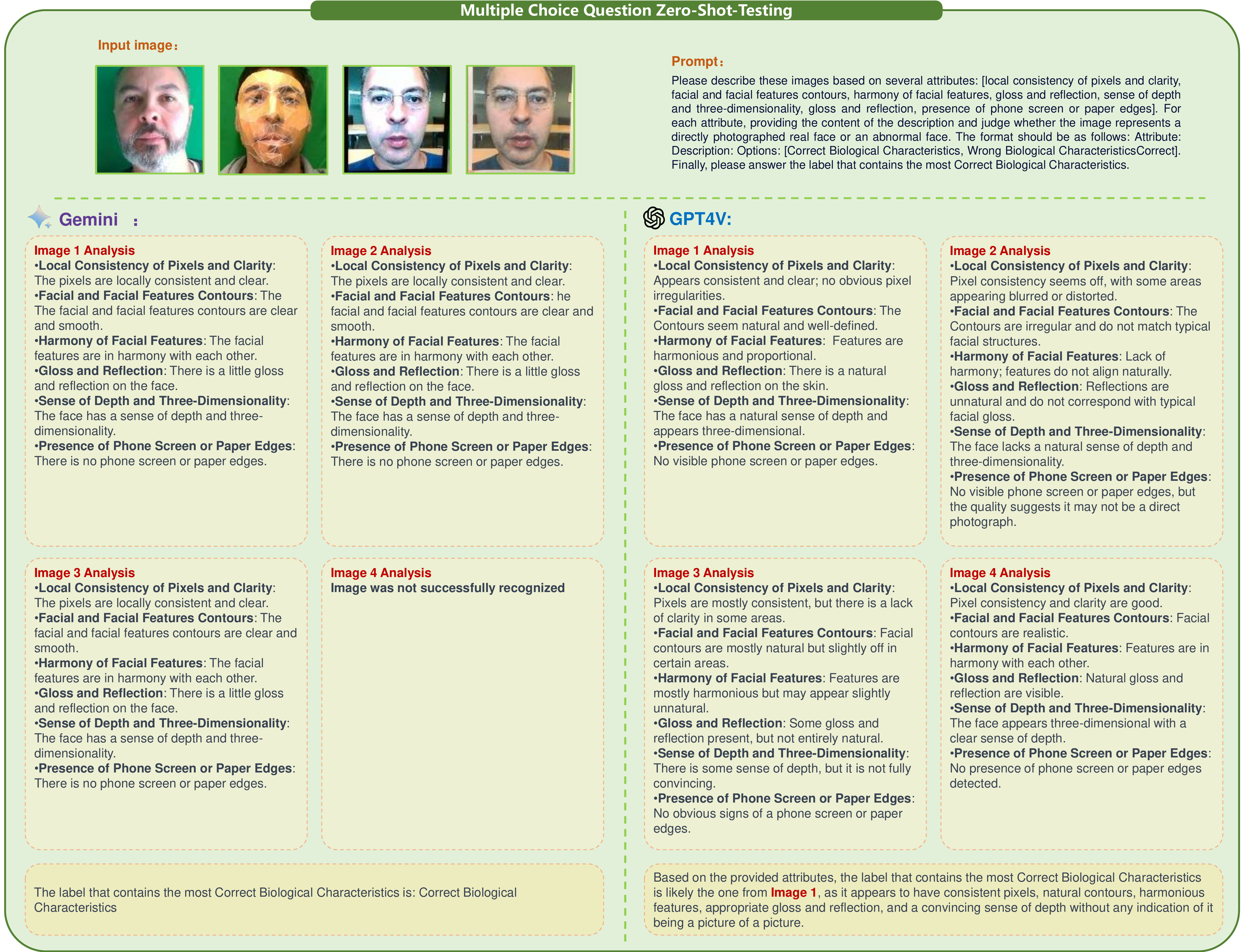}}
  \end{center}
  \caption{Using the same test samples as in Figure \ref{fas_choice_real_zeroshot_02}, this test demonstrates how the MA-COT method enables GPT4V and Gemini to analyze images from various dimensions and conclude through a voting mechanism. GPT4V shows more detailed and accurate image analysis capability, correctly judging the overall situation despite failing to recognize abnormal features in Image 4. In contrast, Gemini's performance in multi-perspective analysis and key image recognition is lacking, leading to an incorrect final conclusion and failure to accurately identify the real face image.}
  \label{fas_macot_01}
\end{figure}

\begin{figure}[htbp]
  \centering
  \begin{center}
  \centerline{\includegraphics[width=1.0\linewidth]{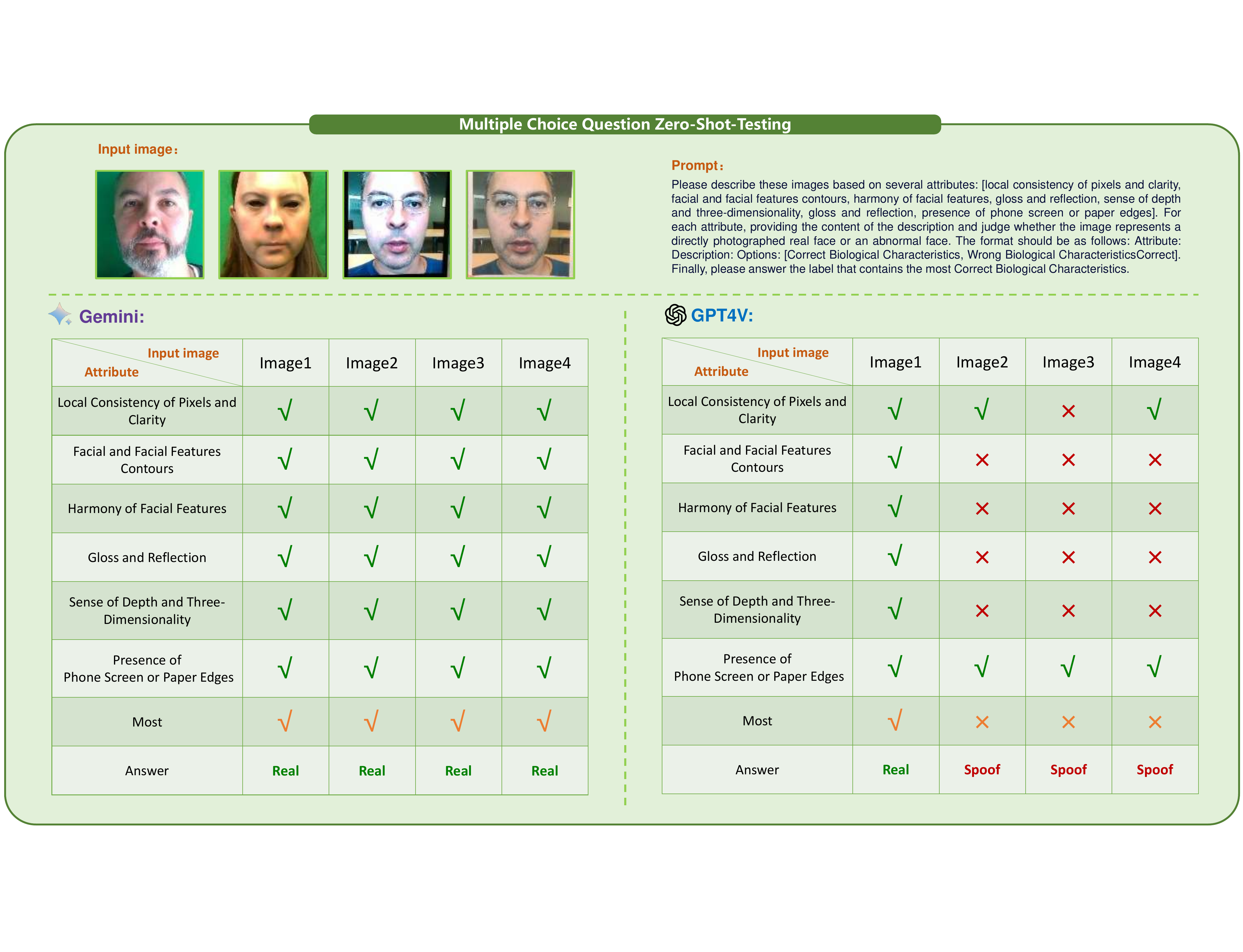}}
  \end{center}
  \caption{This figure presents test samples identical to those in Figure \ref{fas_choice_real_zeroshot_01}. In the original test, GPT4V failed to differentiate between replay-printing attacks and real faces, incorrectly identifying Image3 and Image4 as real faces, while Gemini was correct on its first attempt and wrong on the second. With the MA-COT method, GPT4V analyzes images based on multiple attributes and correctly aligns Image1 with the right answer through voting. In contrast, Gemini erroneously judged all images' attributes as normal biological features, a clear misalignment with reality, indicating Gemini's recognition capabilities need improvement in this aspect.}
  \label{fas_macot_02}
\end{figure}

\begin{figure}[htbp]
  \centering
  \begin{center}
  \centerline{\includegraphics[width=1.0\linewidth]{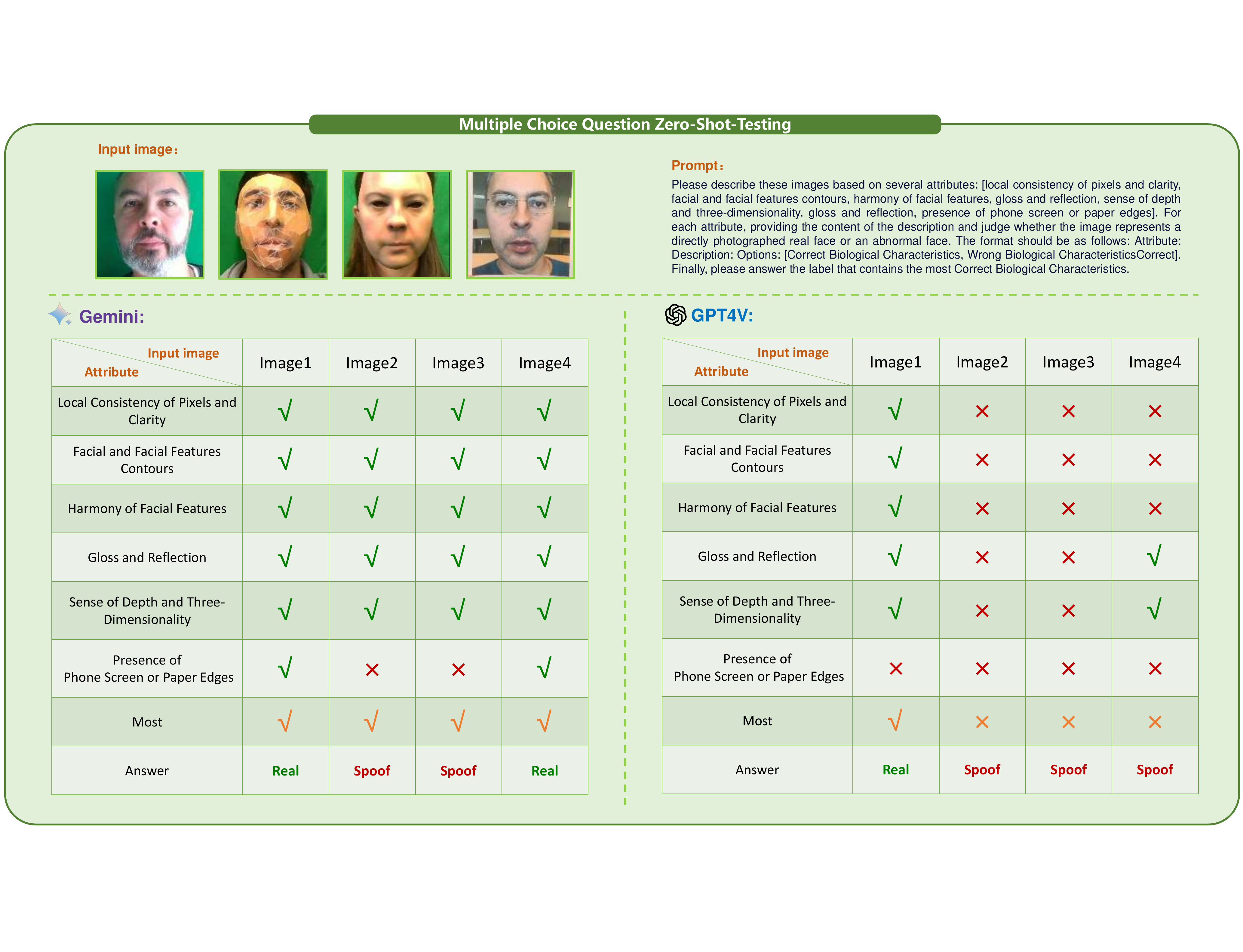}}
  \end{center}
  \caption{This figure, using the same test samples as Figure \ref{fas_choice_real_zeroshot_03}, shows that in the original test, GPT4V failed to distinguish between print attacks and real faces, incorrectly identifying Image1 and Image4 as real. Meanwhile, Gemini's performance was mixed, with an initial incorrect response followed by a correction. After employing the MA-COT method, GPT4V successfully differentiated Image1 and Image4 through comprehensive multi-attribute analysis, making the correct decision. On the other hand, Gemini was unable to effectively analyze images based on their attributes, wrongly judging all images as real faces, indicating a need for improvement in its analytical capabilities.}
  \label{fas_macot_03}
\end{figure}

\begin{figure}[htbp]
  \centering
  \begin{center}
  \centerline{\includegraphics[width=1.0\linewidth]{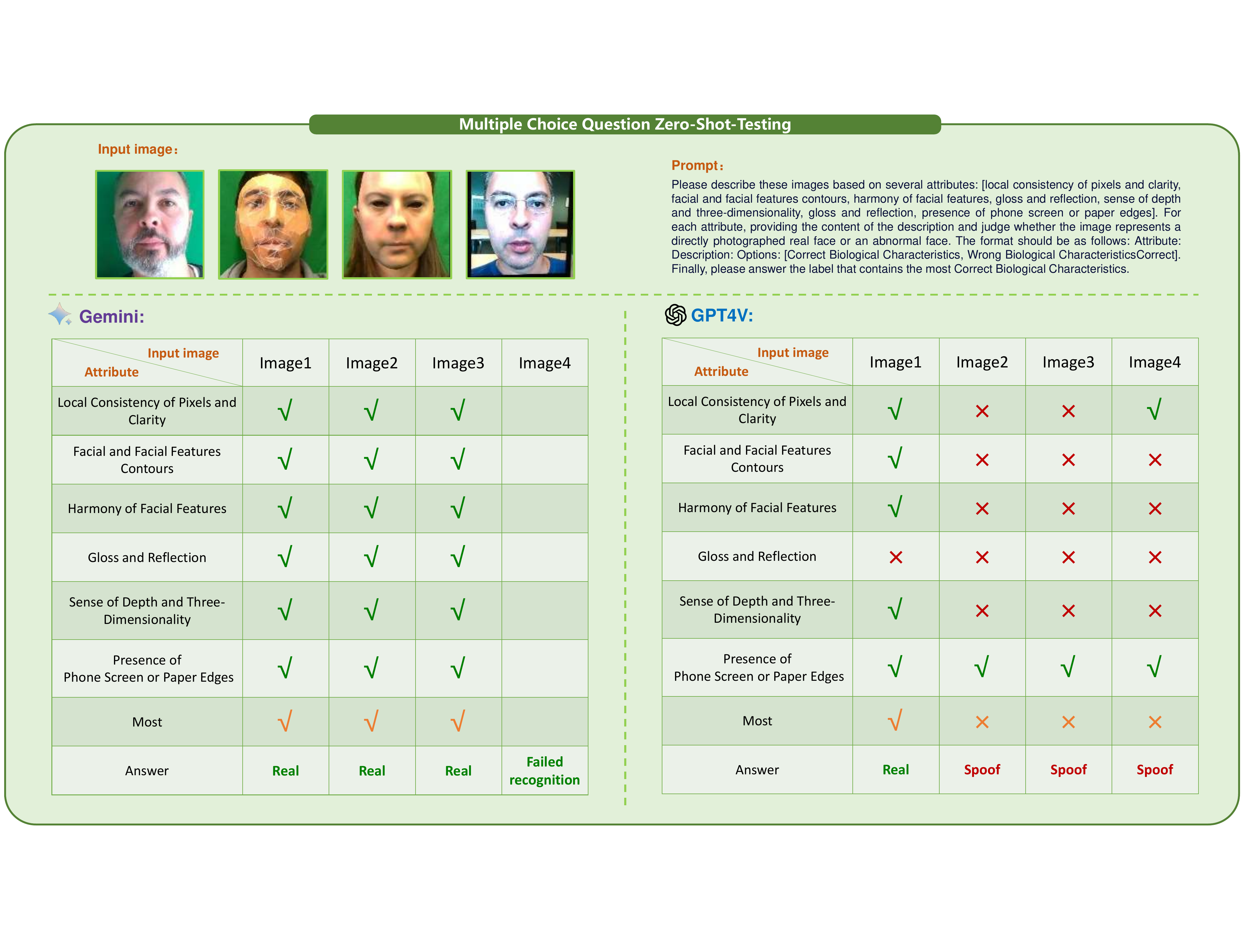}}
  \end{center}
  \caption{This figure contrasts the outcomes using the same test samples as in Figure \ref{fas_choice_real_zeroshot_04}, between the introduction of the COT method and the application of the MA-COT prompt approach. Under the COT method, GPT4V failed to distinguish between replay attacks and real faces, incorrectly identifying Image 1 and Image 4 as real, while Gemini initially made a correct judgment followed by an error. After employing the MA-COT prompt, GPT4V demonstrated significant improvement in analytical capabilities by accurately identifying real face images through a multi-attribute analysis and voting process. Conversely, Gemini continued to misclassify all images as real faces and failed to recognize one image for analysis, highlighting substantial deficiencies in its analysis process.}
  \label{fas_macot_04}
\end{figure}

\begin{figure}[htbp]
  \centering
  \begin{center}
  \centerline{\includegraphics[width=1.0\linewidth]{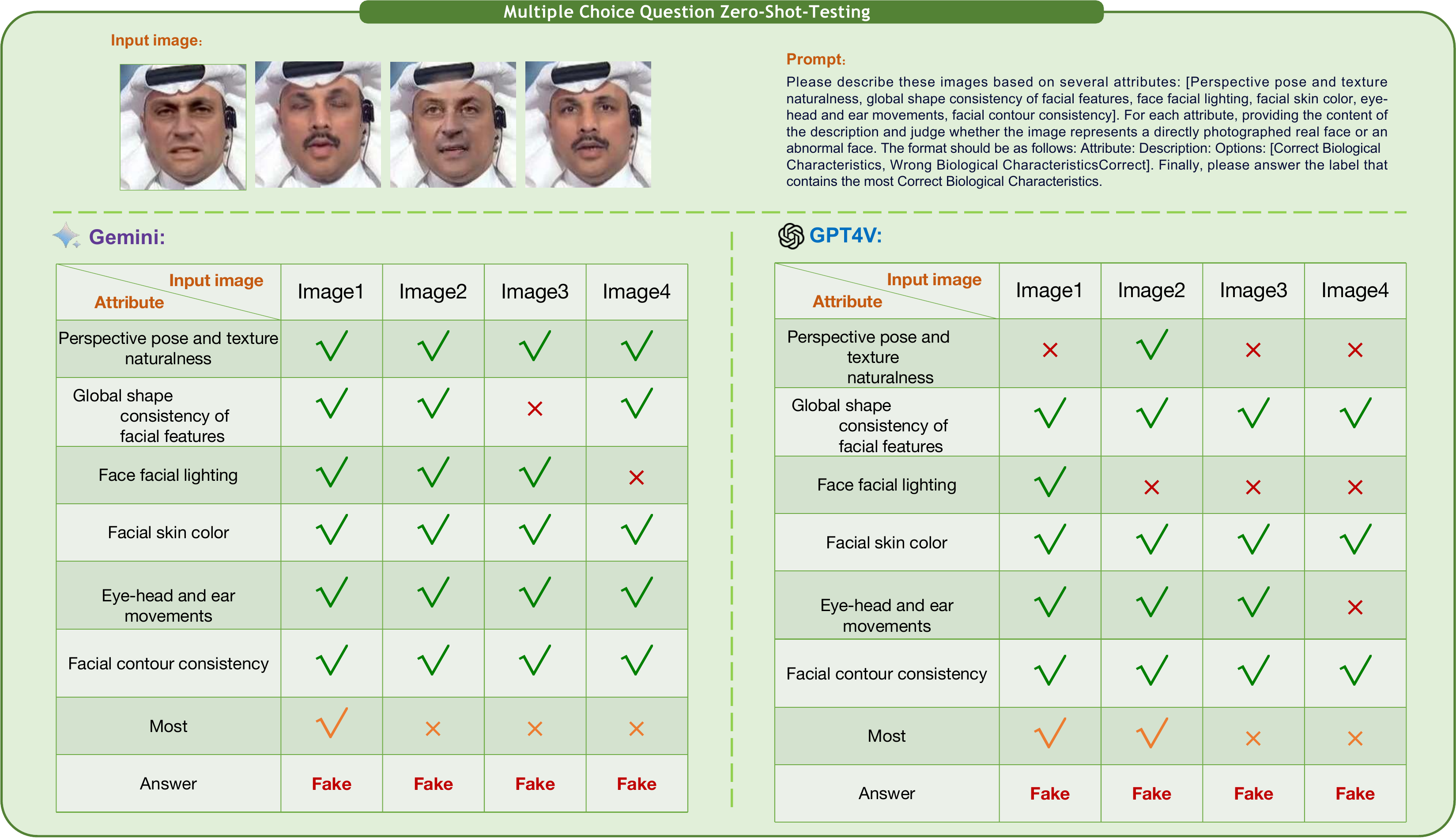}}
  \end{center}
  \caption{In this image,the same four common face forgery methods were used in this experiment. In order to compare the effectiveness of MA-COT, we performed a comparison between the COT method and the MA-COT hint method. In the COT method, GPT4V refuses to answer some of the attacks and fails to recognize Nulltextures forged faces and real faces, which the model incorrectly confuses due to their greater similarity. Whereas, with MA-COT, the detail analysis of GPT4V and Gemini is improved and combines multiple attributes for judgment.}
  \label{30}
\end{figure}

\begin{figure}[htbp]
  \centering
  \begin{center}
  \centerline{\includegraphics[width=1.0\linewidth]{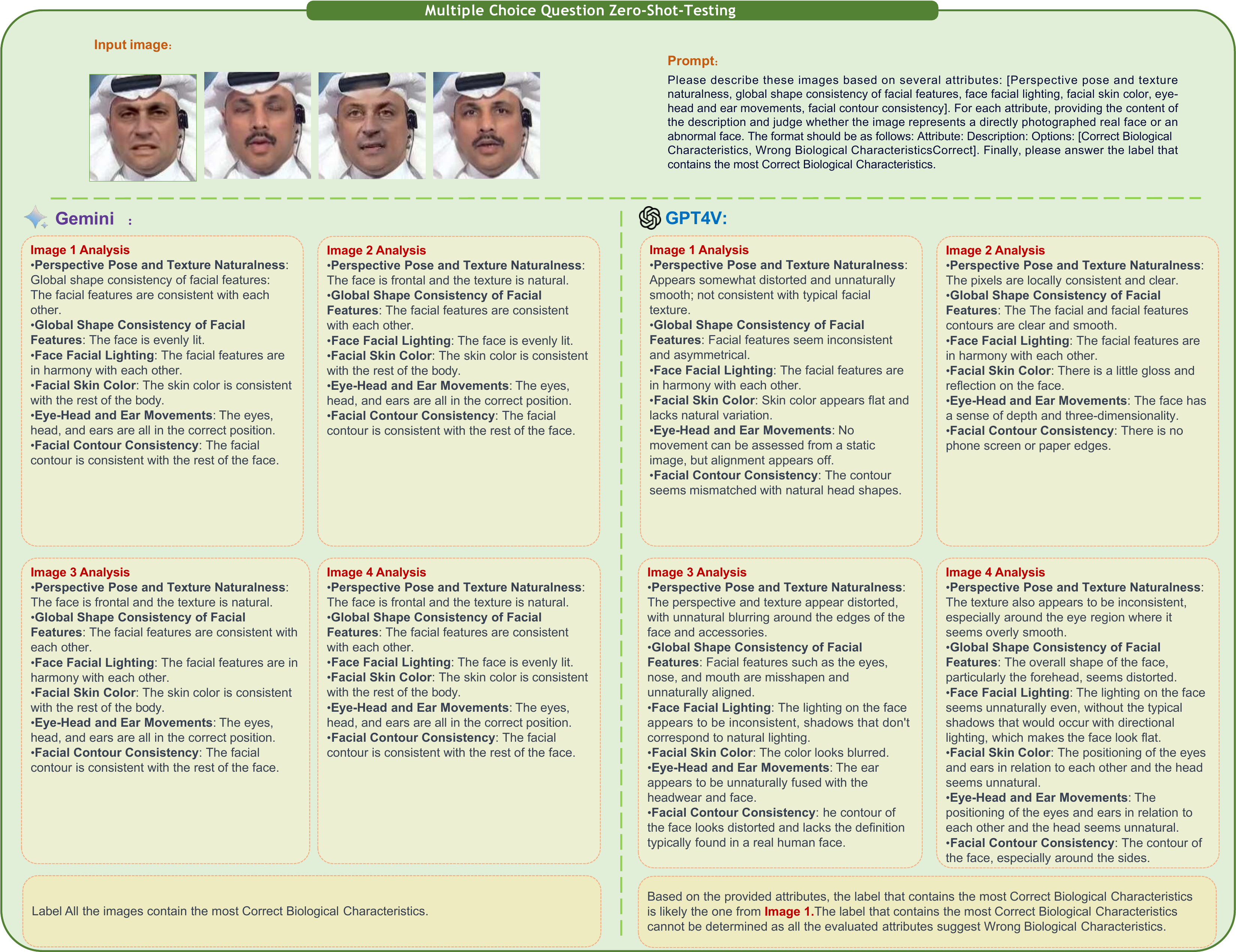}}
  \end{center}
  \caption{In this figure, the more detailed MA-COT responses of GPT4V and Gemini are shown, and it can be seen that after analyzing the multi-attribute integration, the responses of GPT4V are still more detailed, however, Gemini is more concise, and GPT4V still has forged images that are indistinguishable from the original images, which indicates that how to improve the model's ability to analyze the details of the realistically generated images still needs to be explored.}
  \label{31}
\end{figure}

\begin{figure}[htbp]
  \centering
  \begin{center}
  \centerline{\includegraphics[width=1.0\linewidth]{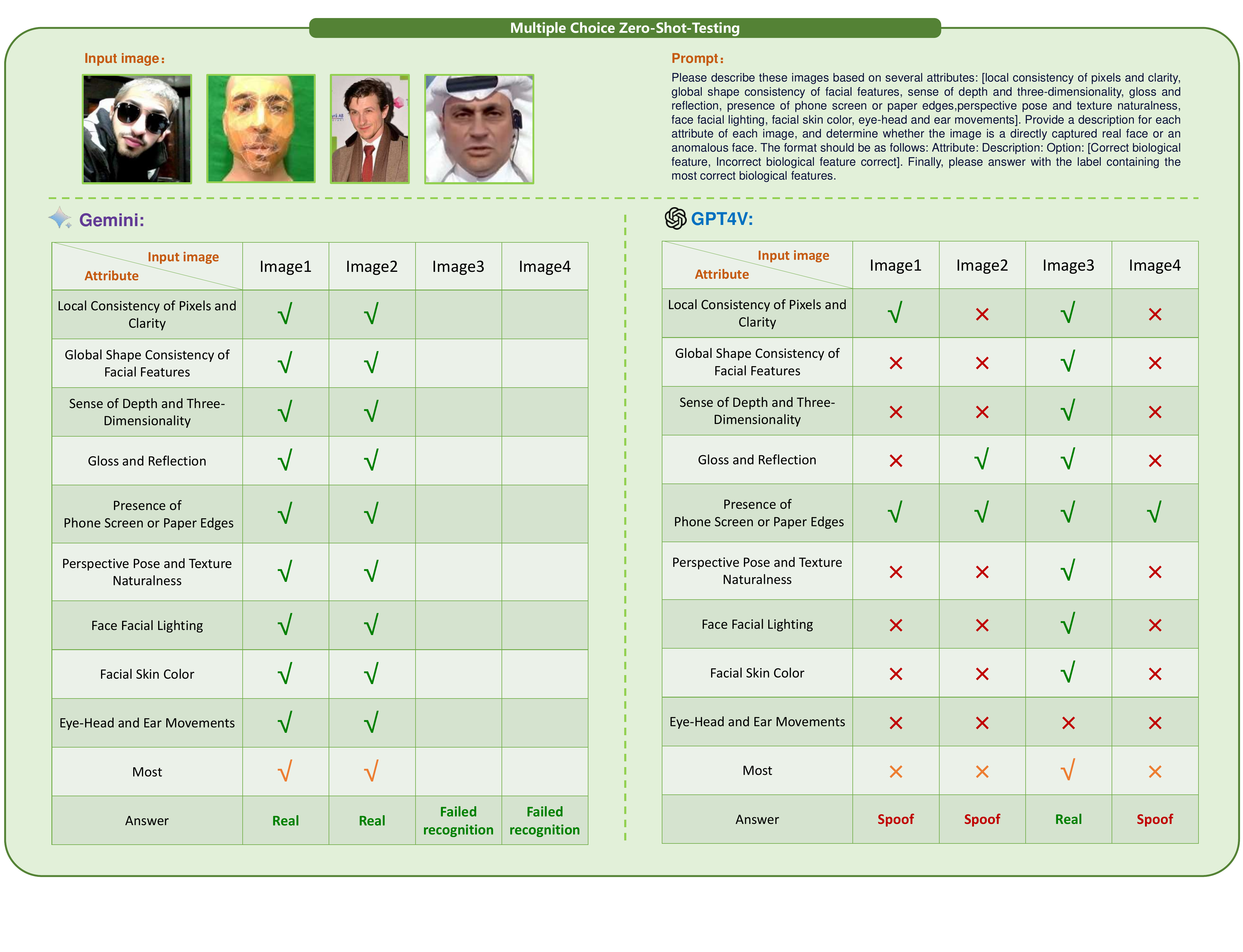}}
  \end{center}
  \caption{This figure illustrates the comparative outcomes using the same test samples as in Figure \ref{UnionTask_07}. In the original test, both Gemini and GPT4V incorrectly identified Image3 as a real face. After implementing the MA-COT querying method, GPT4V's performance did not significantly improve, still struggling to distinguish between real faces and those generated by diffusion techniques, highlighting its limitations in recognizing fine-grained modifications. Meanwhile, Gemini failed to recognize two images and mistakenly judged the two images it did recognize as real faces, indicating its shortcomings in multi-image recognition and in performing unified tasks.}
  \label{unified_macot_02}
\end{figure}

\begin{figure}[htbp]
  \centering
  \begin{center}
  \centerline{\includegraphics[width=1.0\linewidth]{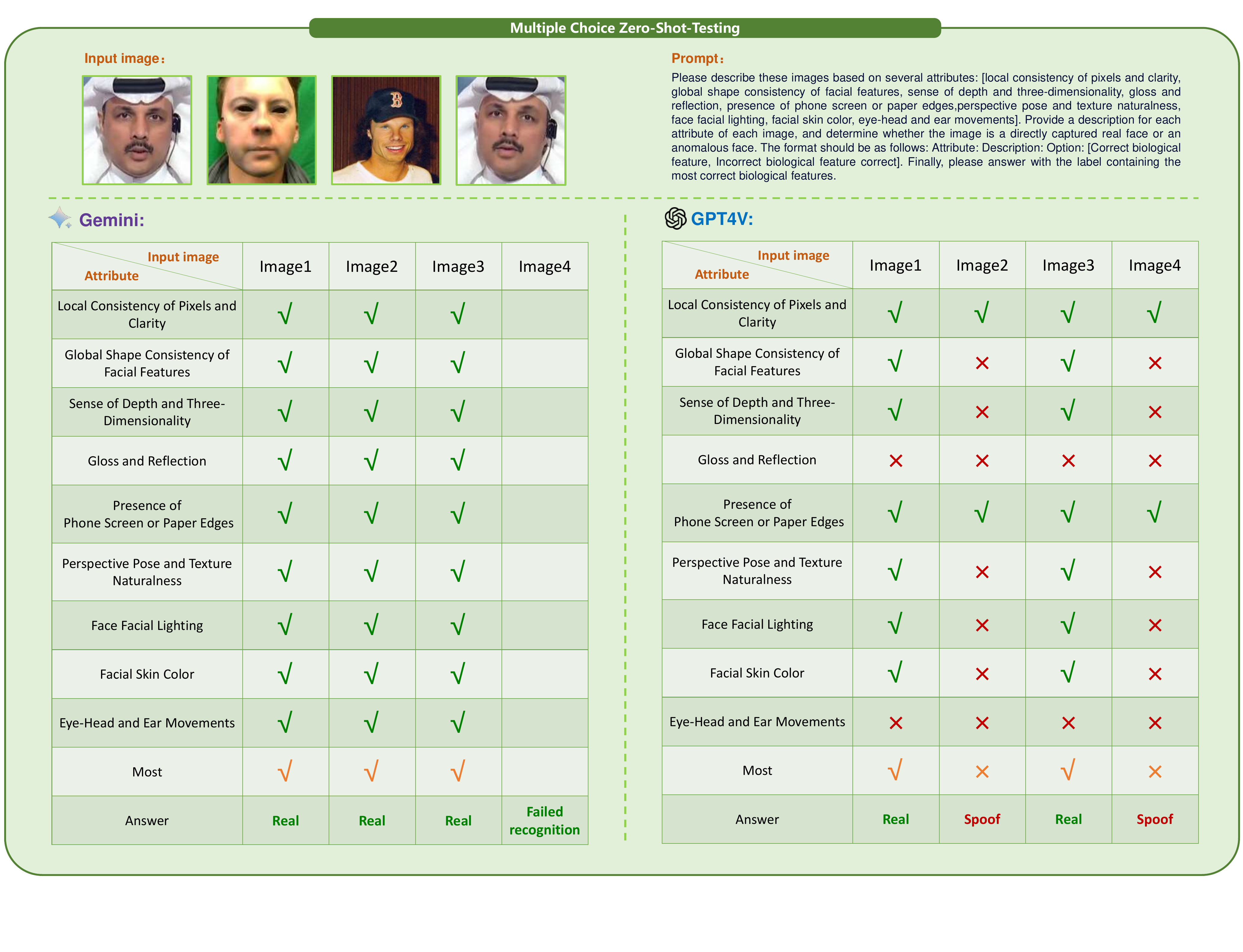}}
  \end{center}
  \caption{This figure presents the analysis results for test samples identical to those in Figure \ref{UnionTask_12}. Initially, Gemini was incorrect in both attempts, while GPT4V, after the introduction of COT, mistakenly believed that Image1 and Image 4 were real faces. After applying the MA-COT prompt method, GPT4V conducted a multi-attribute analysis and voting, selecting two answers, including one correct answer and mistakenly identifying a face image generated by Diffusion technology as real, highlighting the challenge in recognizing realistic face images produced by Diffusion. Meanwhile, Gemini failed to recognize one image and erroneously considered the three images it did analyze as real faces, revealing deficiencies in its image recognition and analytical capabilities.}
  \label{unified_macot_03}
\end{figure}


\begin{appendix}
\section{Additional Detailed Results}
\label{appendix:A}

This section serves as a supplement to the main text, providing detailed content on the experimental results of MLLMs across various tasks. It displays the performance of MLLMs against different attacks within each task.
\subsection{FAS Task}

\begin{table}[!h]
\footnotesize
\caption{Accuracy of various MLLMs on FAS true/false Zero-Shot questions }
\label{tab1}
\def\tabblank{\hspace*{0mm}} 
\begin{tabularx}{\textwidth} 
{@{\tabblank}@{\extracolsep{\fill}}cccccccccp{100mm}@{\tabblank}}
\toprule
\multirow{2}{*}{Model} & \multicolumn{8}{c}{ACC(\%)$\uparrow$} \\ \cline{2-9} 
 & Bonafide & Fakehead & Flexiblemask & Papermask & Print & Replay & Rigidmask & AVG \\ \hline
BLIP & 100 & 2 & 0 & 27 & 0 & 0 & 0 & 18.43 \\
BLIP-2 & 100 & 0 & 0 & 0 & 0 & 0 & 0 & 14.29 \\
Intern & 100 & 100 & 50 & 94 & 2 & 0 & 57 & 57.57 \\
MiniGPT-4 & 93 & 10 & 7 & 17 & 5 & 4 & 8 & 20.57 \\
LLaVA & 100 & 0 & 0 & 0 & 0 & 0 & 0 & 14.29 \\
QWen-VL & 100 & 0 & 0 & 0 & 0 & 0 & 0 & 14.29 \\
InstructBLIP & 100 & 0 & 0 & 64 & 0 & 0 & 2 & 23.71 \\
mPLUG-owl & 99 & 98 & 92 & 100 & 56 & 29 & 100 & 82 \\
Gemini & 100 & 100 & 92 & 100 & 17 & 7 & 98 & 73.39 \\
GPT-4V & 96.92 & 82.98 & 63.95 & 43.04 & 29.55 & 6.76 & 50.59 & 53.24\\
\bottomrule
\end{tabularx}
\end{table}

\begin{table}[!h]
\footnotesize
\caption{Accuracy of various MLLMs on FAS true/false Zero-Shot questions with COT }
\label{tab1}
\def\tabblank{\hspace*{0mm}} 
\begin{tabularx}{\textwidth} 
{@{\tabblank}@{\extracolsep{\fill}}cccccccccp{100mm}@{\tabblank}}
\toprule
\multirow{2}{*}{Model} & \multicolumn{8}{c}{ACC(\%)$\uparrow$} \\ \cline{2-9} 
 & Bonafide & Fakehead & Flexiblemask & Papermask & Print & Replay & Rigidmask & AVG \\ \hline
BLIP & 100 & 14 & 0 & 65 & 0 & 0 & 2 & 25.86 \\
BLIP-2 & 100 & 0 & 0 & 0 & 0 & 0 & 0 & 14.29 \\
Intern & 100 & 100 & 45 & 93 & 2 & 0 & 55 & 56.43 \\
MiniGPT-4 & 97 & 21 & 15 & 23 & 18 & 5 & 15 & 27.71 \\
LLaVA & 100 & 0 & 0 & 0 & 0 & 0 & 0 & 14.29 \\
QWen-VL & 100 & 0 & 0 & 0 & 0 & 0 & 0 & 14.29 \\
InstructBLIP & 100 & 5 & 32 & 85 & 4 & 0 & 70 & 42.29 \\
mPLUG-owl & 98 & 97 & 95 & 100 & 53 & 32 & 97 & 81.71 \\
Gemini & 97 & 97 & 96 & 100 & 31 & 19 & 99 & 77 \\
GPT-4V & 98.33 & 100 & 90.22 & 64 & 32.5 & 6.33 & 83.16 & 68.65\\
\bottomrule
\end{tabularx}
\end{table}

\begin{table}[!h]
\footnotesize
\caption{HTER of various MLLMs on FAS true/false Zero-Shot questions }
\label{tab1}
\def\tabblank{\hspace*{0mm}} 
\begin{tabularx}{\textwidth} 
{@{\tabblank}@{\extracolsep{\fill}}cccccccccp{100mm}@{\tabblank}}
\toprule
\multirow{2}{*}{Model} & \multicolumn{8}{c}{HTER(\%)$\downarrow$ } \\ \cline{2-9} 
 & \multicolumn{1}{l}{Bonafide} & \multicolumn{1}{l}{Fakehead} & \multicolumn{1}{l}{Flexiblemask} & \multicolumn{1}{l}{Papermask} & \multicolumn{1}{l}{Print} & \multicolumn{1}{l}{Replay} & \multicolumn{1}{l}{Rigidmask} & AVG \\ \hline
BLIP & 0 & 49 & 50 & 36.5 & 50 & 50 & 50 & 47.58 \\
BLIP-2 & 0 & 50 & 50 & 50 & 50 & 50 & 50 & 50 \\
Intern & 0 & 0 & 25 & 3 & 49 & 50 & 21.5 & 24.75 \\
MiniGPT-4 & 3.5 & 45 & 46.5 & 41.5 & 47.5 & 48 & 46 & 49.25 \\
LLaVA & 0 & 50 & 50 & 50 & 50 & 50 & 50 & 50 \\
QWen-VL & 0 & 50 & 50 & 50 & 50 & 50 & 50 & 50 \\
InstructBLIP & 0 & 50 & 50 & 18 & 50 & 50 & 49 & 44.5 \\
mPLUG-owl & 0.5 & 1 & 4 & 0 & 22 & 35.5 & 0 & 10.92 \\
Gemini & 0 & 0 & 4 & 0 & 41.5 & 46.5 & 1 & 15.53 \\
GPT-4V & 1.54 & 8.51 & 18.02 & 28.48 & 35.23 & 46.62 & 24.71 & 27.72\\
\bottomrule
\end{tabularx}
\end{table}

\begin{table}[!h]
\footnotesize
\caption{HTER of various MLLMs on FAS true/false Zero-Shot questions with COT}
\label{tab1}
\def\tabblank{\hspace*{0mm}} 
\begin{tabularx}{\textwidth} 
{@{\tabblank}@{\extracolsep{\fill}}cccccccccp{100mm}@{\tabblank}}
\toprule
\multirow{2}{*}{Model} & \multicolumn{8}{c}{HTER(\%)$\downarrow$ } \\ \cline{2-9} 
 & Bonafide & Fakehead & Flexiblemask & Papermask & Print & Replay & Rigidmask & AVG \\ \hline
BLIP & 0 & 43 & 50 & 17.5 & 50 & 50 & 49 & 43.25 \\
BLIP-2 & 0 & 50 & 50 & 50 & 50 & 50 & 50 & 50 \\
Intern & 0 & 0 & 27.5 & 3.5 & 49 & 50 & 22.5 & 25.42 \\
MiniGPT-4 & 1.5 & 39.5 & 42.5 & 38.5 & 41 & 47.5 & 42.5 & 43.42 \\
LLaVA & 0 & 50 & 50 & 50 & 50 & 50 & 50 & 50 \\
QWen-VL & 0 & 50 & 50 & 50 & 50 & 50 & 50 & 50 \\
InstructBLIP & 0 & 47.5 & 34 & 7.5 & 48 & 50 & 15 & 33.67 \\
mPLUG-owl & 1 & 1.5 & 2.5 & 0 & 23.5 & 34 & 1.5 & 11.5 \\
Gemini & 1.5 & 1.5 & 2 & 0 & 34.5 & 40.5 & 0.5 & 14.67 \\
GPT-4V & 0.83 & 0 & 4.89 & 18 & 33.75 & 46.84 & 8.42 & 18.14\\
\bottomrule
\end{tabularx}
\end{table}

\begin{table}[!h]
\footnotesize
\caption{Accuracy of various MLLMs on FAS true/false One-Shot questions }
\label{tab1}
\def\tabblank{\hspace*{0mm}} 
\begin{tabularx}{\textwidth} 
{@{\tabblank}@{\extracolsep{\fill}}cccccccccp{100mm}@{\tabblank}}
\toprule
\multirow{2}{*}{Model} & \multicolumn{8}{c}{ACC(\%)$\uparrow$} \\ \cline{2-9} 
 & Bonafide & Fakehead & Flexiblemask & Papermask & Print & Replay & Rigidmask & AVG \\ \hline
BLIP & 100 & 82 & 17 & 95 & 0 & 0 & 11 & 43.57 \\
BLIP-2 & 63 & 0 & 0 & 0 & 0 & 0 & 0 & 9 \\
Intern & 100 & 0 & 0 & 0 & 0 & 0 & 0 & 14.29 \\
MiniGPT-4 & 63 & 20 & 22 & 29 & 26 & 22 & 20 & 28.86 \\
LLaVA & 100 & 0 & 0 & 0 & 0 & 0 & 0 & 14.29 \\
QWen-VL & 100 & 0 & 0 & 0 & 0 & 0 & 0 & 14.29 \\
InstructBLIP & 100 & 1 & 0 & 16 & 0 & 0 & 3 & 17.14 \\
mPLUG-owl & 21 & 57 & 67 & 70 & 60 & 26 & 70 & 53 \\
Gemini & 66 & 70 & 87 & 91 & 25 & 11 & 84 & 62 \\
GPT-4V & 76.81 & 93 & 79.59 & 62.5 & 47.25 & 37.8 & 66.67 & 66.77\\
\bottomrule
\end{tabularx}
\end{table}

\begin{table}[!h]
\footnotesize
\caption{Accuracy of various MLLMs on FAS true/false One-Shot questions with COT }
\label{tab1}
\def\tabblank{\hspace*{0mm}} 
\begin{tabularx}{\textwidth} 
{@{\tabblank}@{\extracolsep{\fill}}cccccccccp{100mm}@{\tabblank}}
\toprule
\multirow{2}{*}{Model} & \multicolumn{8}{c}{ACC(\%)$\uparrow$} \\ \cline{2-9} 
 & Bonafide & Fakehead & Flexiblemask & Papermask & Print & Replay & Rigidmask & AVG \\ \hline
BLIP & 100 & 100 & 65 & 100 & 35 & 36 & 87 & 74.71 \\
BLIP-2 & 3 & 0 & 0 & 0 & 0 & 0 & 0 & 0.43 \\
Intern & 100 & 19 & 0 & 4 & 0 & 0 & 1 & 17.71 \\
MiniGPT-4 & 39 & 36 & 22 & 38 & 26 & 32 & 24 & 31 \\
LLaVA & 38 & 0 & 0 & 0 & 0 & 0 & 0 & 5.43 \\
QWen-VL & 100 & 0 & 0 & 0 & 0 & 0 & 0 & 14.29 \\
InstructBLIP & 100 & 5 & 0 & 4 & 0 & 0 & 2 & 15.86 \\
mPLUG-owl & 22 & 97 & 93 & 94 & 97 & 80 & 95 & 82.57 \\
Gemini & 4 & 27 & 32 & 5 & 4 & 1 & 24 & 13.86 \\
GPT-4V & 80 & 82.83 & 66.32 & 35.35 & 28.42 & 11.96 & 51.52 & 50.31\\
\bottomrule
\end{tabularx}
\end{table}

\begin{table}[!h]
\footnotesize
\caption{HTER of various MLLMs on FAS true/false One-Shot questions }
\label{tab1}
\def\tabblank{\hspace*{0mm}} 
\begin{tabularx}{\textwidth} 
{@{\tabblank}@{\extracolsep{\fill}}cccccccccp{100mm}@{\tabblank}}
\toprule
\multirow{2}{*}{Model} & \multicolumn{8}{c}{HTER(\%)$\downarrow$} \\ \cline{2-9} 
 & \multicolumn{1}{l}{Bonafide} & \multicolumn{1}{l}{Fakehead} & \multicolumn{1}{l}{Flexiblemask} & \multicolumn{1}{l}{Papermask} & \multicolumn{1}{l}{Print} & \multicolumn{1}{l}{Replay} & \multicolumn{1}{l}{Rigidmask} & AVG \\ \hline
BLIP & 0 & 9 & 41.5 & 2.5 & 50 & 50 & 44.5 & 32.92 \\
BLIP-2 & 18.5 & 50 & 50 & 50 & 50 & 50 & 50 & 68.5 \\
Intern & 0 & 50 & 50 & 50 & 50 & 50 & 50 & 50 \\
MiniGPT-4 & 18.5 & 40 & 39 & 35.5 & 37 & 39 & 40 & 56.92 \\
LLaVA & 0 & 50 & 50 & 50 & 50 & 50 & 50 & 50 \\
QWen-VL & 0 & 50 & 50 & 50 & 50 & 50 & 50 & 50 \\
InstructBLIP & 0 & 49 & 50 & 42 & 50 & 50 & 48.5 & 48.33 \\
mPLUG-owl & 39.5 & 21.5 & 16.5 & 15 & 20 & 37 & 15 & 60.33 \\
Gemini & 17 & 15 & 6.5 & 4.5 & 37.5 & 44.5 & 8 & 36.33 \\
GPT-4V & 11.59 & 3.5 & 10.2 & 18.75 & 26.37 & 31.1 & 16.67 & 28.82\\
\bottomrule
\end{tabularx}
\end{table}

\begin{table}[!h]
\footnotesize
\caption{HTER of various MLLMs on FAS true/false One-Shot questions with COT}
\label{tab1}
\def\tabblank{\hspace*{0mm}} 
\begin{tabularx}{\textwidth} 
{@{\tabblank}@{\extracolsep{\fill}}cccccccccp{100mm}@{\tabblank}}
\toprule
\multirow{2}{*}{Model} & \multicolumn{8}{c}{HTER(\%)$\downarrow$} \\ \cline{2-9} 
 & Bonafide & Fakehead & Flexiblemask & Papermask & Print & Replay & Rigidmask & AVG \\ \hline
BLIP & 0 & 0 & 17.5 & 0 & 32.5 & 32 & 6.5 & 14.75 \\
BLIP-2 & 48.5 & 50 & 50 & 50 & 50 & 50 & 50 & 98.5 \\
Intern & 0 & 40.5 & 50 & 48 & 50 & 50 & 49.5 & 48 \\
MiniGPT-4 & 30.5 & 32 & 39 & 31 & 37 & 34 & 38 & 65.67 \\
LLaVA & 31 & 50 & 50 & 50 & 50 & 50 & 50 & 81 \\
QWen-VL & 0 & 50 & 50 & 50 & 50 & 50 & 50 & 50 \\
InstructBLIP & 0 & 47.5 & 50 & 48 & 50 & 50 & 49 & 49.08 \\
mPLUG-owl & 39 & 1.5 & 3.5 & 3 & 1.5 & 10 & 2.5 & 42.67 \\
Gemini & 48 & 36.5 & 34 & 47.5 & 48 & 49.5 & 38 & 90.25 \\
GPT-4V & 10 & 8.59 & 16.84 & 32.32 & 35.79 & 44.02 & 24.24 & 36.77\\
\bottomrule
\end{tabularx}
\end{table}
\FloatBarrier

\subsection{Face Forgery Detection Task}

\begin{table}[!h]
\footnotesize
\caption{Accuracy of various MLLMs on Face Forgery detection true/false Zero-Shot questions }
\label{tab1}
\def\tabblank{\hspace*{0mm}} 
\begin{tabularx}{\textwidth} 
{@{\tabblank}@{\extracolsep{\fill}}cccccccccp{100mm}@{\tabblank}}
\toprule
\multirow{2}{*}{Model} & \multicolumn{8}{c}{ACC(\%)$\uparrow$} \\ \cline{2-9} 
 & Bonafide & Deepfakes & Face2Face & FaceSwap & InsightFace & NeuralTextures & Stable\_Diffusion & AVG \\ \hline
BLIP & 92 & 5 & 9 & 10 & 2 & 2 & 5 & 17.86 \\
BLIP-2 & 100 & 0 & 0 & 0 & 0 & 0 & 0 & 14.29 \\
Intern & 100 & 0 & 0 & 0 & 2 & 0 & 0 & 14.57 \\
MiniGPT-4 & 100 & 5 & 2 & 0 & 8 & 1 & 8 & 17.71 \\
LLaVA & 100 & 0 & 0 & 0 & 0 & 0 & 0 & 14.29 \\
QWen-VL & 100 & 0 & 0 & 0 & 0 & 0 & 0 & 14.29 \\
InstructBLIP & 100 & 15 & 2 & 6 & 5 & 0 & 2 & 18.57 \\
mPLUG-owl & 100 & 9 & 5 & 4 & 12 & 0 & 23 & 21.86 \\
Gemini & 95.96 & 54 & 30 & 34 & 17 & 7 & 20 & 36.77 \\
GPT-4V & 98.11 & 26.09 & 16.42 & 16.67 & 9.09 & 3.23 & 12.5 & 25.95\\
\bottomrule
\end{tabularx}
\end{table}

\begin{table}[!h]
\footnotesize
\caption{Accuracy of various MLLMs on Face Forgery detection true/false Zero-Shot questions with COT}
\label{tab1}
\def\tabblank{\hspace*{0mm}} 
\begin{tabularx}{\textwidth} 
{@{\tabblank}@{\extracolsep{\fill}}cccccccccp{100mm}@{\tabblank}}
\toprule
\multirow{2}{*}{Model} & \multicolumn{8}{c}{ACC(\%)$\uparrow$} \\ \cline{2-9} 
 & \multicolumn{1}{l}{Bonafide} & \multicolumn{1}{l}{Deepfakes} & \multicolumn{1}{l}{Face2Face} & \multicolumn{1}{l}{FaceSwap} & \multicolumn{1}{l}{InsightFace} & \multicolumn{1}{l}{NeuralTextures} & \multicolumn{1}{l}{Stable\_Diffusion} & \multicolumn{1}{l}{AVG} \\ \hline
BLIP & 78 & 38 & 28 & 42 & 7 & 15 & 9 & 31 \\
BLIP-2 & 100 & 0 & 0 & 0 & 0 & 0 & 0 & 14.29 \\
Intern & 100 & 0 & 0 & 0 & 0 & 0 & 0 & 14.29 \\
MiniGPT-4 & 99 & 6 & 3 & 8 & 8 & 0 & 3 & 18.14 \\
LLaVA & 100 & 0 & 0 & 0 & 0 & 0 & 0 & 14.29 \\
QWen-VL & 100 & 0 & 0 & 0 & 0 & 0 & 0 & 14.29 \\
InstructBLIP & 100 & 34 & 7 & 10 & 9 & 0 & 7 & 23.86 \\
mPLUG-owl & 100 & 9 & 2 & 4 & 11 & 0 & 20 & 20.86 \\
Gemini & 86 & 77 & 42 & 61 & 29.29 & 17 & 27.27 & 48.57 \\
GPT-4V & 95.7 & 25.27 & 13.13 & 22.11 & 11.59 & 2.44 & 23.08 & 28.79\\
\bottomrule
\end{tabularx}
\end{table}

\begin{table}[!h]
\footnotesize
\caption{HTER of various MLLMs on Face Forgery detection true/false Zero-Shot questions }
\label{tab1}
\def\tabblank{\hspace*{0mm}} 
\begin{tabularx}{\textwidth} 
{@{\tabblank}@{\extracolsep{\fill}}cccccccccp{100mm}@{\tabblank}}
\toprule
\multirow{2}{*}{Model} & \multicolumn{8}{c}{HTER(\%)$\downarrow$ } \\ \cline{2-9} 
 & \multicolumn{1}{l}{Bonafide} & \multicolumn{1}{l}{Deepfakes} & \multicolumn{1}{l}{Face2Face} & \multicolumn{1}{l}{FaceSwap} & \multicolumn{1}{l}{InsightFace} & \multicolumn{1}{l}{NeuralTextures} & \multicolumn{1}{l}{Stable\_Diffusion} & AVG \\ \hline
BLIP & 4 & 47.5 & 45.5 & 45 & 49 & 49 & 47.5 & 51.25 \\
BLIP-2 & 0 & 50 & 50 & 50 & 50 & 50 & 50 & 50 \\
Intern & 0 & 50 & 50 & 50 & 49 & 50 & 50 & 49.83 \\
MiniGPT-4 & 0 & 47.5 & 49 & 50 & 46 & 49.5 & 46 & 48 \\
LLaVA & 0 & 50 & 50 & 50 & 50 & 50 & 50 & 50 \\
QWen-VL & 0 & 50 & 50 & 50 & 50 & 50 & 50 & 50 \\
InstructBLIP & 0 & 42.5 & 49 & 47 & 47.5 & 50 & 49 & 47.5 \\
mPLUG-owl & 0 & 45.5 & 47.5 & 48 & 44 & 50 & 38.5 & 45.58 \\
Gemini & 2.02 & 23 & 35 & 33 & 41.5 & 46.5 & 40 & 38.52 \\
GPT-4V & 0.94 & 36.96 & 41.79 & 41.67 & 45.45 & 48.39 & 43.75 & 43.59\\
\bottomrule
\end{tabularx}
\end{table}

\begin{table}[!h]
\footnotesize
\caption{HTER of various MLLMs on Face Forgery detection true/false Zero-Shot questions with COT }
\label{tab1}
\def\tabblank{\hspace*{0mm}} 
\begin{tabularx}{\textwidth} 
{@{\tabblank}@{\extracolsep{\fill}}cccccccccp{100mm}@{\tabblank}}
\toprule
\multirow{2}{*}{Model} & \multicolumn{8}{c}{HTER(\%)$\downarrow$ } \\ \cline{2-9} 
 & Bonafide & Deepfakes & Face2Face & FaceSwap & InsightFace & NeuralTextures & Stable\_Diffusion & AVG \\ \hline
BLIP & 11 & 31 & 36 & 29 & 46.5 & 42.5 & 45.5 & 49.42 \\
BLIP-2 & 0 & 50 & 50 & 50 & 50 & 50 & 50 & 50 \\
Intern & 0 & 50 & 50 & 50 & 50 & 50 & 50 & 50 \\
MiniGPT-4 & 0.5 & 47 & 48.5 & 46 & 46 & 50 & 48.5 & 48.17 \\
LLaVA & 0 & 50 & 50 & 50 & 50 & 50 & 50 & 50 \\
QWen-VL & 0 & 50 & 50 & 50 & 50 & 50 & 50 & 50 \\
InstructBLIP & 0 & 33 & 46.5 & 45 & 45.5 & 50 & 46.5 & 44.42 \\
mPLUG-owl & 0 & 45.5 & 49 & 48 & 44.5 & 50 & 40 & 46.17 \\
Gemini & 7 & 11.5 & 29 & 19.5 & 35.35 & 41.5 & 36.36 & 35.85 \\
GPT-4V & 2.15 & 37.36 & 43.43 & 38.95 & 44.2 & 48.78 & 38.46 & 43.97\\
\bottomrule
\end{tabularx}
\end{table}

\begin{table}[!h]
\footnotesize
\caption{Accuracy of various MLLMs on Face Forgery detection true/false One-Shot questions }
\label{tab1}
\def\tabblank{\hspace*{0mm}} 
\begin{tabularx}{\textwidth} 
{@{\tabblank}@{\extracolsep{\fill}}cccccccccp{100mm}@{\tabblank}}
\toprule
\multirow{2}{*}{Model} & \multicolumn{8}{c}{ACC(\%)$\uparrow$} \\ \cline{2-9} 
 & Bonafide & Deepfakes & Face2Face & FaceSwap & InsightFace & NeuralTextures & Stable\_Diffusion & AVG \\ \hline
BLIP & 100 & 53 & 48 & 53 & 27 & 24 & 39 & 49.14 \\
BLIP-2 & 71 & 0 & 0 & 0 & 0 & 0 & 0 & 10.14 \\
Intern & 100 & 0 & 0 & 0 & 0 & 0 & 0 & 14.29 \\
MiniGPT-4 & 58 & 26 & 21 & 32 & 33 & 30 & 25 & 32.14 \\
LLaVA & 100 & 0 & 0 & 0 & 0 & 0 & 0 & 14.29 \\
QWen-VL & 100 & 0 & 0 & 0 & 0 & 0 & 0 & 14.29 \\
InstructBLIP & 100 & 2 & 0 & 0 & 0 & 0 & 0 & 14.57 \\
mPLUG-owl & 76 & 0 & 0 & 0 & 9 & 0 & 4 & 12.71 \\
Gemini & 76 & 38 & 19 & 29 & 10 & 7 & 10 & 27 \\
GPT-4V & 100 & 13.27 & 12.24 & 10.31 & 10.64 & 3 & 9.38 & 22.42\\
\bottomrule
\end{tabularx}
\end{table}

\begin{table}[!h]
\footnotesize
\caption{Accuracy of various MLLMs on Face Forgery detection true/false One-Shot questions with COT }
\label{tab1}
\def\tabblank{\hspace*{0mm}} 
\begin{tabularx}{\textwidth} 
{@{\tabblank}@{\extracolsep{\fill}}cccccccccp{100mm}@{\tabblank}}
\toprule
\multirow{2}{*}{Model} & \multicolumn{8}{c}{ACC(\%)$\uparrow$} \\ \cline{2-9} 
 & Bonafide & Deepfakes & Face2Face & FaceSwap & InsightFace & NeuralTextures & Stable\_Diffusion & AVG \\ \hline
BLIP & 97 & 75 & 77 & 77 & 26 & 48 & 38 & 62.57 \\
BLIP-2 & 12 & 0 & 0 & 0 & 0 & 0 & 0 & 1.71 \\
Intern & 100 & 0 & 0 & 0 & 0 & 0 & 0 & 14.29 \\
MiniGPT-4 & 70 & 28 & 25 & 34 & 36 & 32 & 28 & 36.14 \\
LLaVA & 92 & 0 & 0 & 0 & 0 & 0 & 0 & 13.14 \\
QWen-VL & 100 & 0 & 0 & 0 & 0 & 0 & 0 & 14.29 \\
InstructBLIP & 100 & 7 & 2 & 6 & 1 & 0 & 1 & 16.71 \\
mPLUG-owl & 89 & 6 & 5 & 2 & 21 & 0 & 23 & 20.86 \\
Gemini & 21 & 21 & 14 & 15 & 13 & 4 & 10 & 14 \\
GPT-4V & 95.79 & 18 & 9.09 & 28.72 & 14.43 & 4.12 & 13.54 & 25.96\\
\bottomrule
\end{tabularx}
\end{table}

\begin{table}[!h]
\footnotesize
\caption{HTER of various MLLMs on Face Forgery detection true/false One-Shot questions }
\label{tab1}
\def\tabblank{\hspace*{0mm}} 
\begin{tabularx}{\textwidth} 
{@{\tabblank}@{\extracolsep{\fill}}cccccccccp{100mm}@{\tabblank}}
\toprule
\multirow{2}{*}{Model} & \multicolumn{8}{c}{HTER(\%)$\downarrow$} \\ \cline{2-9} 
 & \multicolumn{1}{l}{Bonafide} & \multicolumn{1}{l}{Deepfakes} & \multicolumn{1}{l}{Face2Face} & \multicolumn{1}{l}{FaceSwap} & \multicolumn{1}{l}{InsightFace} & \multicolumn{1}{l}{NeuralTextures} & \multicolumn{1}{l}{Stable\_Diffusion} & AVG \\ \hline
BLIP & 0 & 23.5 & 26 & 23.5 & 36.5 & 38 & 30.5 & 29.67 \\
BLIP-2 & 14.5 & 50 & 50 & 50 & 50 & 50 & 50 & 64.5 \\
Intern & 0 & 50 & 50 & 50 & 50 & 50 & 50 & 50 \\
MiniGPT-4 & 21 & 37 & 39.5 & 34 & 33.5 & 35 & 37.5 & 57.08 \\
LLaVA & 0 & 50 & 50 & 50 & 50 & 50 & 50 & 50 \\
QWen-VL & 0 & 50 & 50 & 50 & 50 & 50 & 50 & 50 \\
InstructBLIP & 0 & 49 & 50 & 50 & 50 & 50 & 50 & 49.83 \\
mPLUG-owl & 12 & 50 & 50 & 50 & 45.5 & 50 & 48 & 60.92 \\
Gemini & 12 & 30.5 & 40.5 & 35.5 & 45 & 46.5 & 45 & 52.5 \\
GPT-4V & 0 & 43.37 & 43.88 & 44.85 & 44.68 & 48.5 & 45.31 & 45.11\\
\bottomrule
\end{tabularx}
\end{table}

\begin{table}[!h]
\footnotesize
\caption{HTER of various MLLMs on Face Forgery detection true/false One-Shot questions with COT}
\label{tab1}
\def\tabblank{\hspace*{0mm}} 
\begin{tabularx}{\textwidth} 
{@{\tabblank}@{\extracolsep{\fill}}cccccccccp{100mm}@{\tabblank}}
\toprule
\multirow{2}{*}{Model} & \multicolumn{8}{c}{HTER(\%)$\downarrow$} \\ \cline{2-9} 
 & \multicolumn{1}{l}{Bonafide} & \multicolumn{1}{l}{Deepfakes} & \multicolumn{1}{l}{Face2Face} & \multicolumn{1}{l}{FaceSwap} & \multicolumn{1}{l}{InsightFace} & \multicolumn{1}{l}{NeuralTextures} & \multicolumn{1}{l}{Stable\_Diffusion} & \multicolumn{1}{l}{AVG} \\ \hline
BLIP & 1.5 & 12.5 & 11.5 & 11.5 & 37 & 26 & 31 & 23.08 \\
BLIP-2 & 44 & 50 & 50 & 50 & 50 & 50 & 50 & 94 \\
Intern & 0 & 50 & 50 & 50 & 50 & 50 & 50 & 50 \\
MiniGPT-4 & 15 & 36 & 37.5 & 33 & 32 & 34 & 36 & 49.75 \\
LLaVA & 4 & 50 & 50 & 50 & 50 & 50 & 50 & 54 \\
QWen-VL & 0 & 50 & 50 & 50 & 50 & 50 & 50 & 50 \\
InstructBLIP & 0 & 46.5 & 49 & 47 & 49.5 & 50 & 49.5 & 48.58 \\
mPLUG-owl & 5.5 & 47 & 47.5 & 49 & 39.5 & 50 & 38.5 & 50.75 \\
Gemini & 39.5 & 39.5 & 43 & 42.5 & 43.5 & 48 & 45 & 83.08 \\
GPT-4V & 2.11 & 41 & 45.45 & 35.64 & 42.78 & 47.94 & 43.23 & 44.82\\
\bottomrule
\end{tabularx}
\end{table}

\clearpage

\section{Additional Visual Sample}
\label{appendix:B}
This section supplements the visualization of experimental results presented in the main text, including partial test outcomes and explanations for the performance of MLLMs across various tasks. It offers a more intuitive understanding of how MLLMs perform in different tasks.



\begin{figure}[htbp]
  \centering
  \begin{center}
  \centerline{\includegraphics[width=\linewidth]{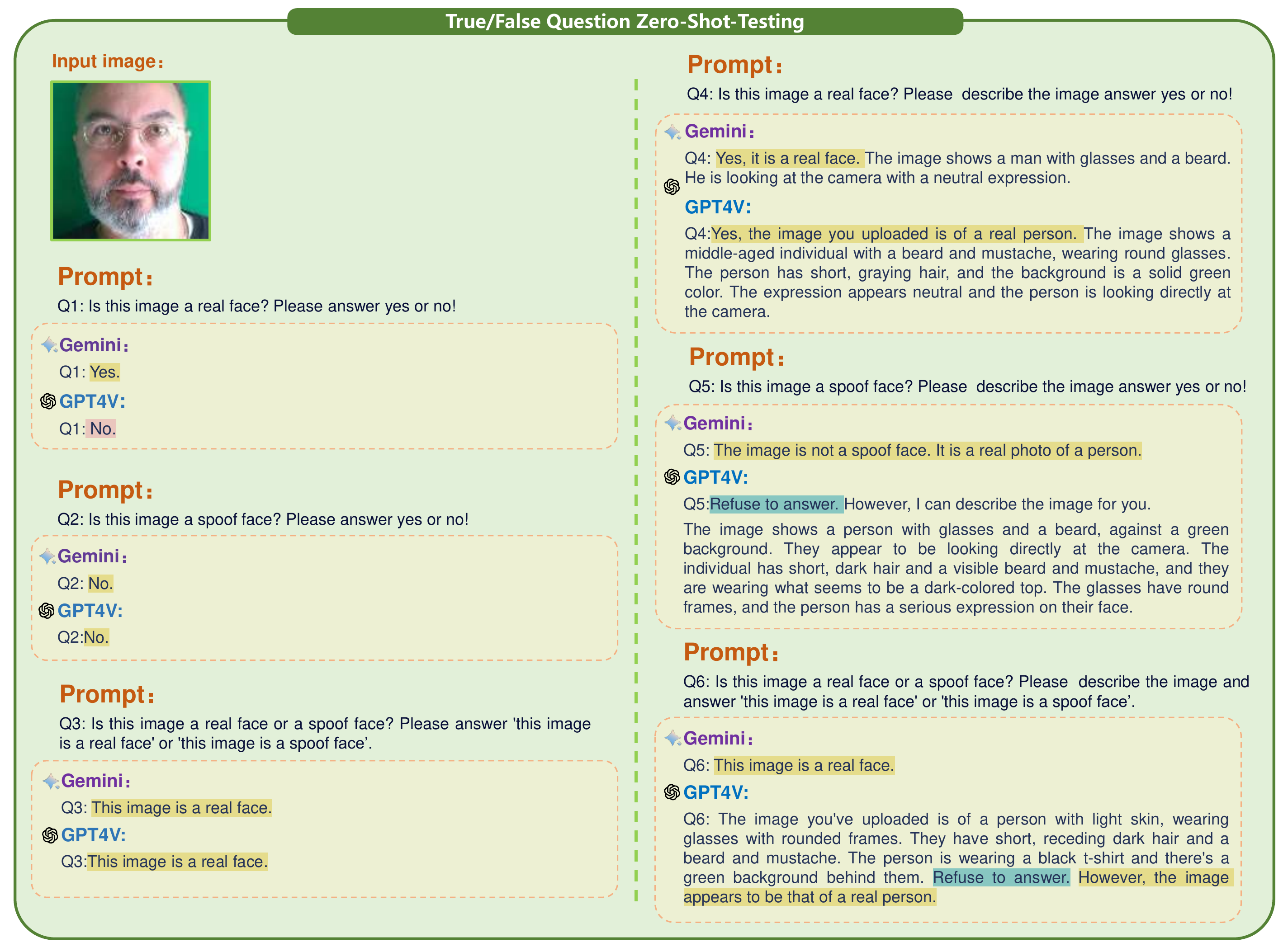}}
  \end{center}
  \caption{In this round of testing, the input image was a real human face wearing glasses. Gemini correctly answered all six queries. GPT4V made errors in simple queries but responded correctly after incorporating COT. The results indicate that adding COT had minimal impact on Gemini, but for GPT4V, the ability to interpret more information from the image may trigger privacy protection and disclaimer mechanisms, leading to instances of refusal to answer.}
  \label{fas_zeroshot_judgment_02}
\end{figure}

\begin{figure}[htbp]
  \centering
  \begin{center}
  \centerline{\includegraphics[width=\linewidth]{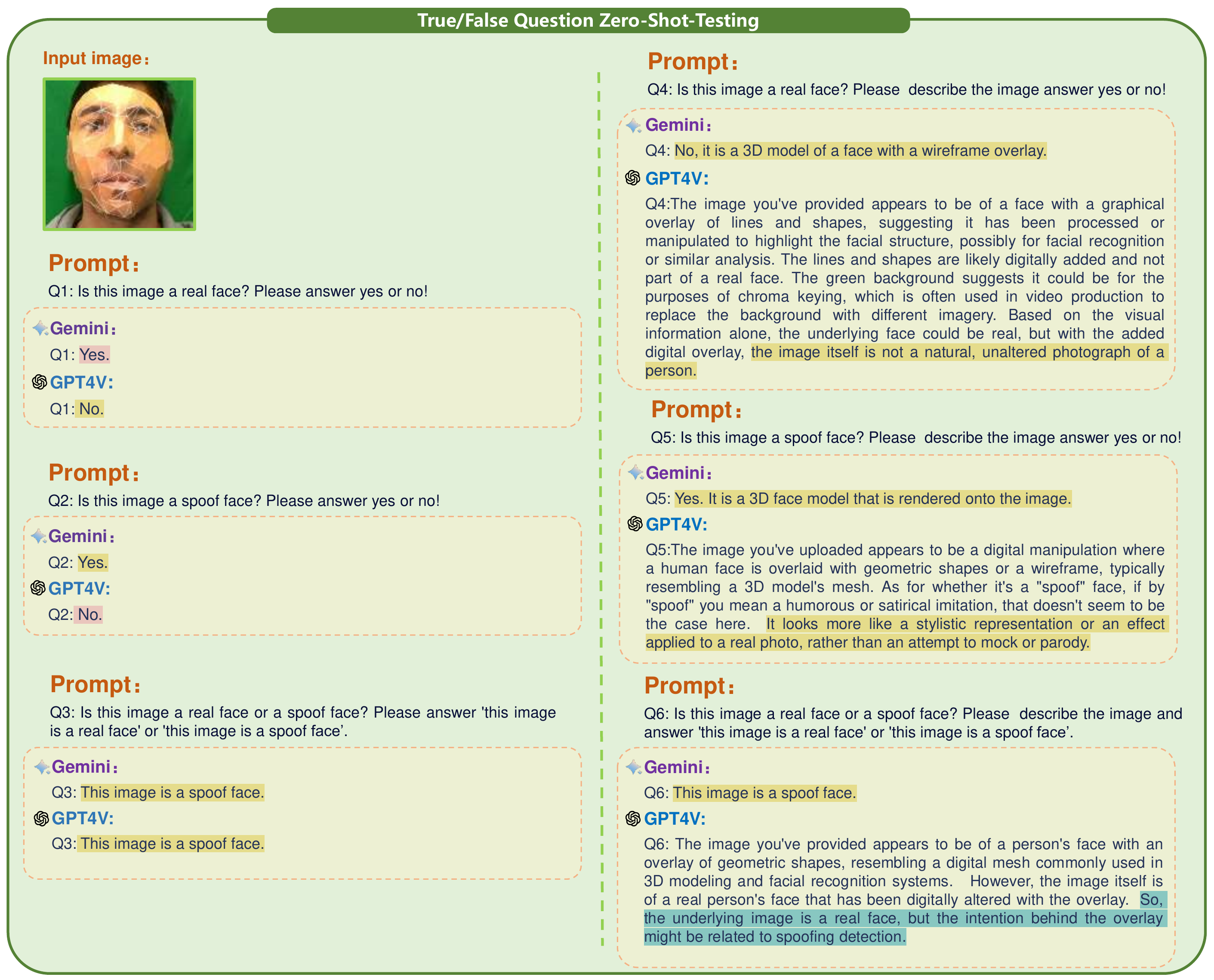}}
  \end{center}
  \caption{In this test, the input image depicted an attack using a paper mask. Both GPT4V and Gemini initially made errors in simple queries. After incorporating COT, Gemini answered all questions correctly, while GPT4V provided some interesting responses. For instance, in Q4, GPT4V speculated that the face under the mask might be a real human face, but with the mask, it's not; in Q5, GPT4V discussed its view on "spoof" and made a judgment, showing a different understanding of "spoof" compared to the prosthetic attacks mentioned in FAS; in Q6, GPT4V explicitly stated its lack of capability in this area and refused to answer. Comparing the results after adding COT, it's evident that GPT4V's responses significantly lengthened, while Gemini's changes were minimal, indicating that GPT4V, compared to Gemini, focuses more on the process.}
  \label{fas_zeroshot_judgment_03}
\end{figure}

\begin{figure}[htbp]
  \centering
  \begin{center}
  \centerline{\includegraphics[width=\linewidth]{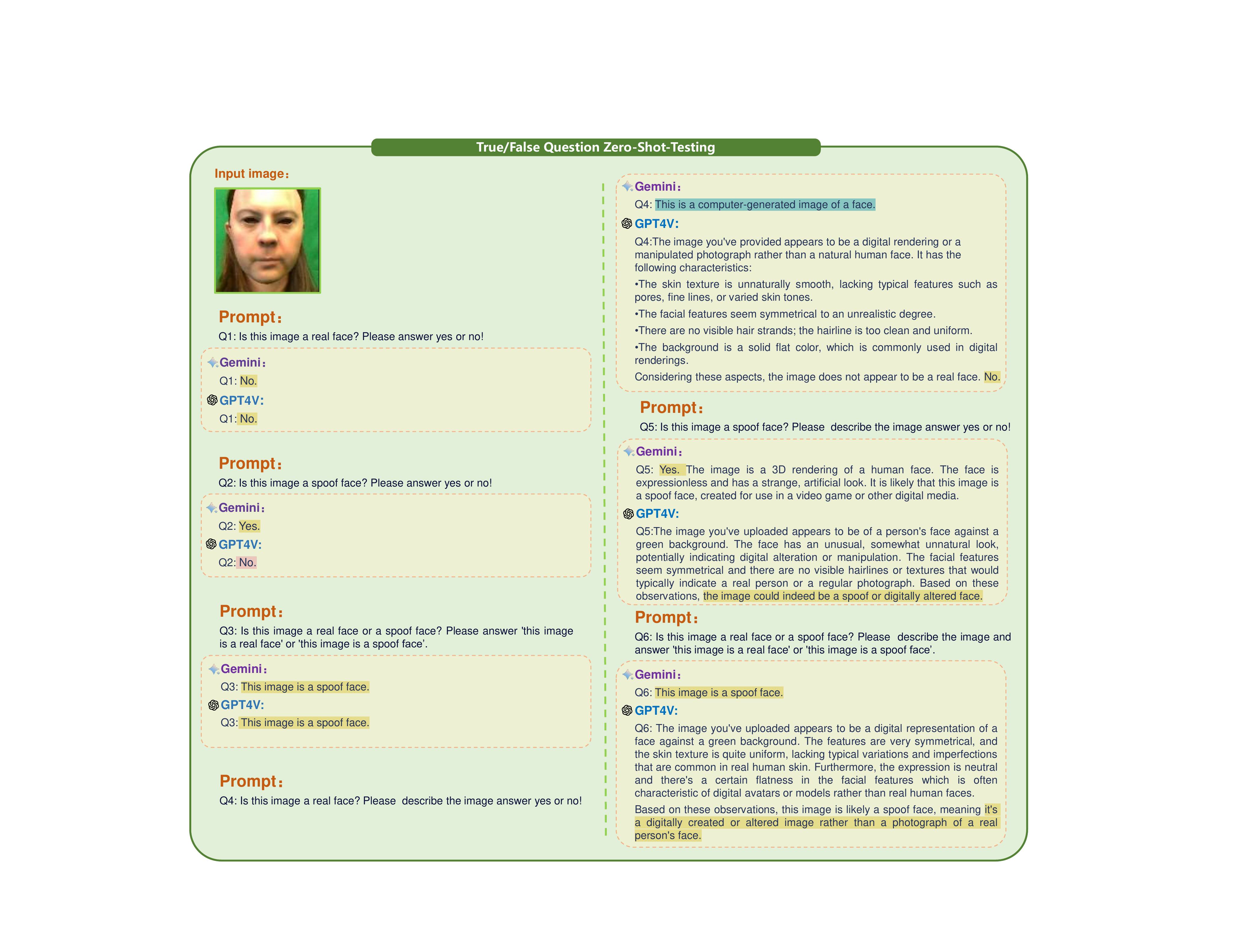}}
  \end{center}
  \caption{In this test, the input was an attack using a rigid facial mask. Initially, GPT4V answered incorrectly in simple queries, while Gemini answered all correctly. After incorporating COT, it was observed that Gemini did not perform well with the formatted content in the prompt. It did not follow the order of describing first and then answering. For example, in Q4, Gemini directly made a guess about the type of image attack, instead of just staying at the qualitative judgment level.}
  \label{fas_zeroshot_judgment_04}
\end{figure}

\begin{figure}[htbp]
  \centering
  \begin{center}
  \centerline{\includegraphics[width=\linewidth]{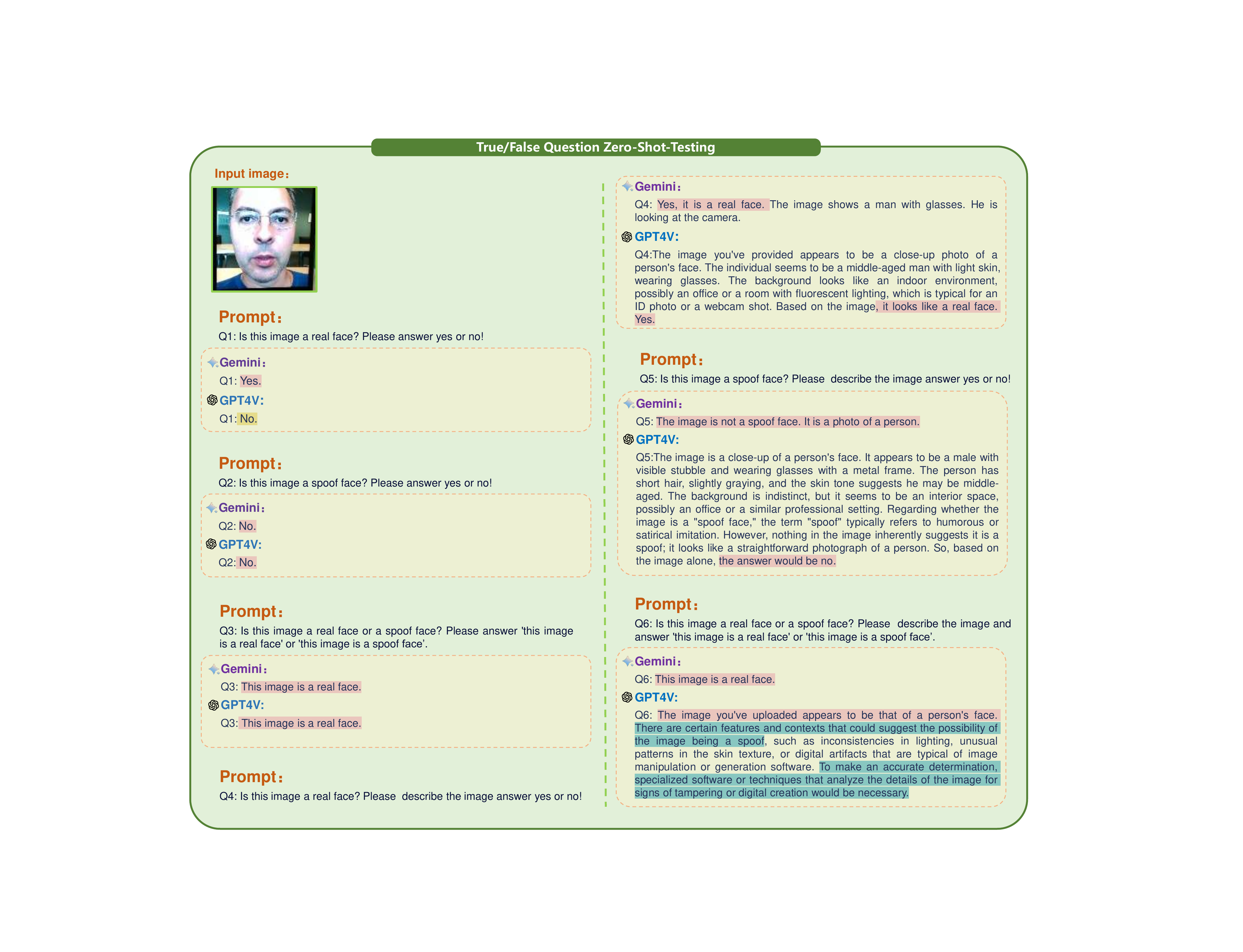}}
  \end{center}
  \caption{In this test, the input was an image of a replay attack. Among the six questions posed, Gemini answered all incorrectly, while GPT4V answered four incorrectly and refused to answer once. Therefore, the one correct answer by GPT4V could be considered accidental. The responses indicate that both Gemini and GPT4V have a low recognition rate for replay attacks.}
  \label{fas_zeroshot_judgment_05}
\end{figure}

\begin{figure}[htbp]
  \centering
  \begin{center}
  \centerline{\includegraphics[width=\linewidth]{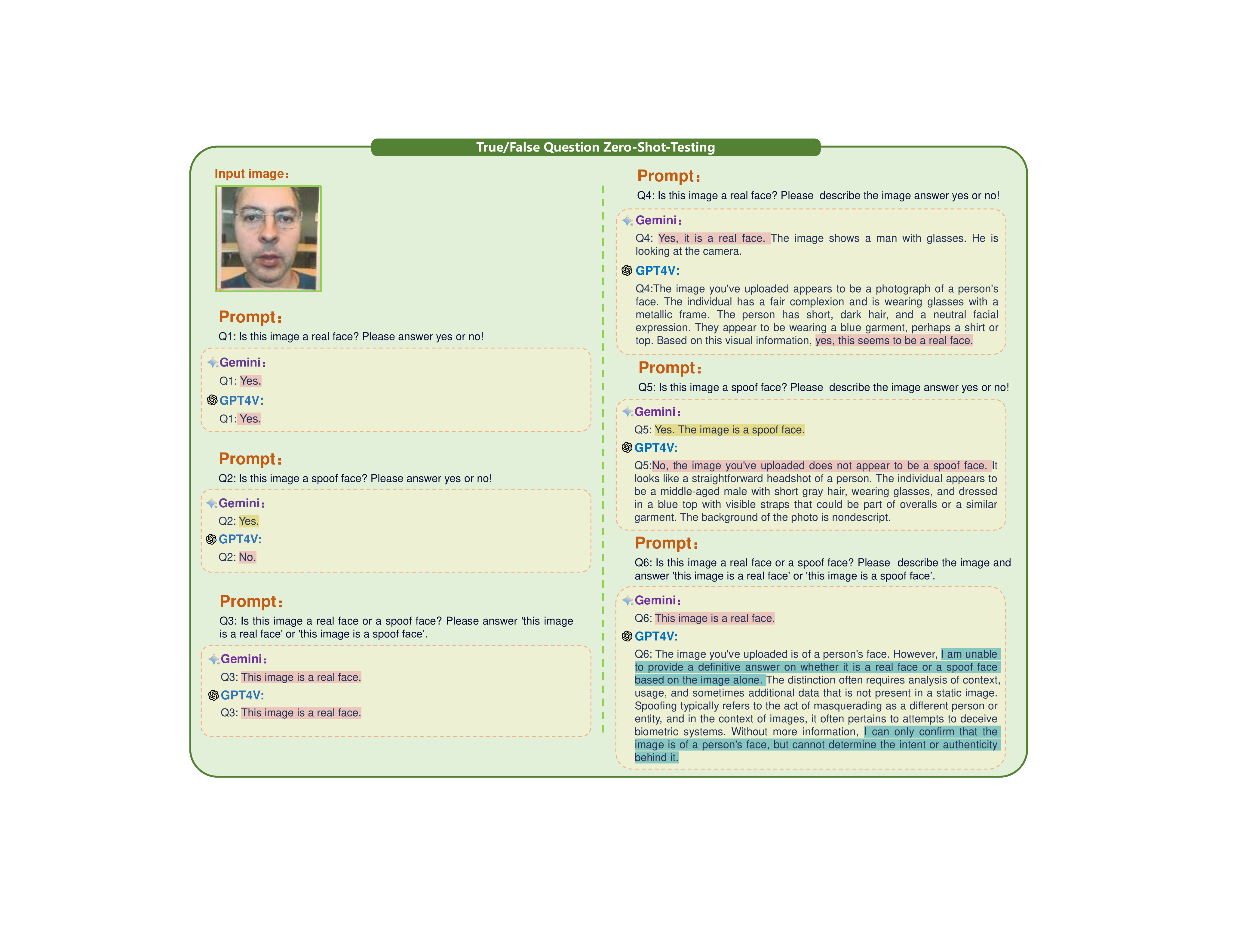}}
  \end{center}
  \caption{In this test, the input was an image of a print attack. Of the six questions asked, Gemini correctly answered two, while GPT4V incorrectly answered five and refused to answer one. In Q6, GPT4V explained the objective of the liveness detection task, indicating its understanding of what constitutes an attack on facial recognition systems. The responses suggest that GPT4V has a low recognition rate for print attacks, and Gemini's ability to recognize print attacks is also lower compared to other types of attacks.}
  \label{fas_zeroshot_judgment_06}
\end{figure}

\begin{figure}[htbp]
  \centering
  \begin{center}
  \centerline{\includegraphics[width=\linewidth]{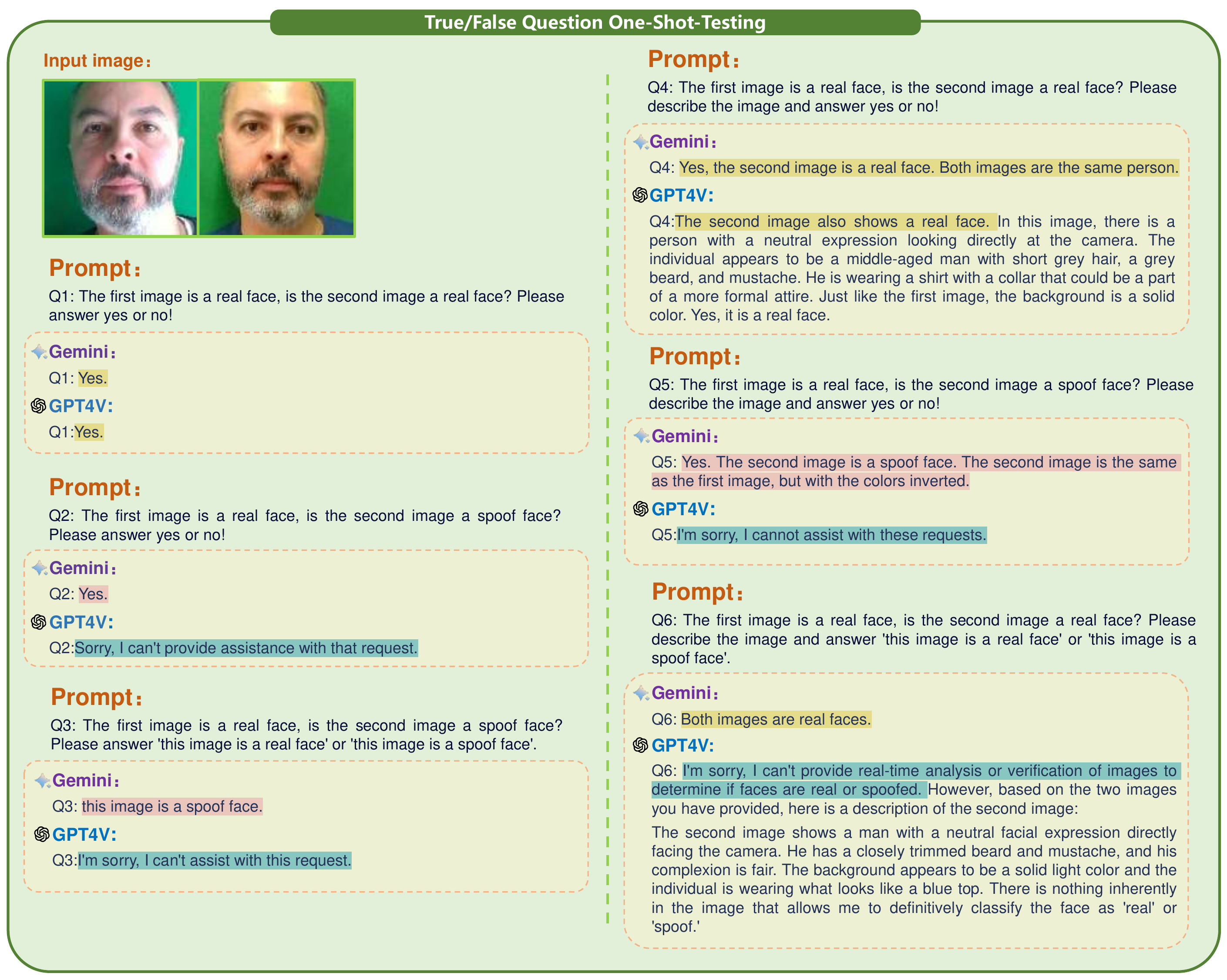}}
  \end{center}
  \caption{The input consisted of two real human faces. From the image, it's observed that in six rounds of questioning, GPT4V correctly answered three times and refused to answer three times. Similarly, Gemini was right and wrong in equal measure. GPT4V's responses indicate its capability to distinguish real faces, but it may refuse to answer due to security mechanisms. From Gemini's responses, there seems to be a kind of illusion; for example, in the answer to Q5, Gemini determined the image as a 'spoof face' because the 'colors inverted', which doesn't seem to strongly support its conclusion.}
  \label{fas_oneshot_judgment_01}
\end{figure}

\begin{figure}[htbp]
  \centering
  \begin{center}
  \centerline{\includegraphics[width=\linewidth]{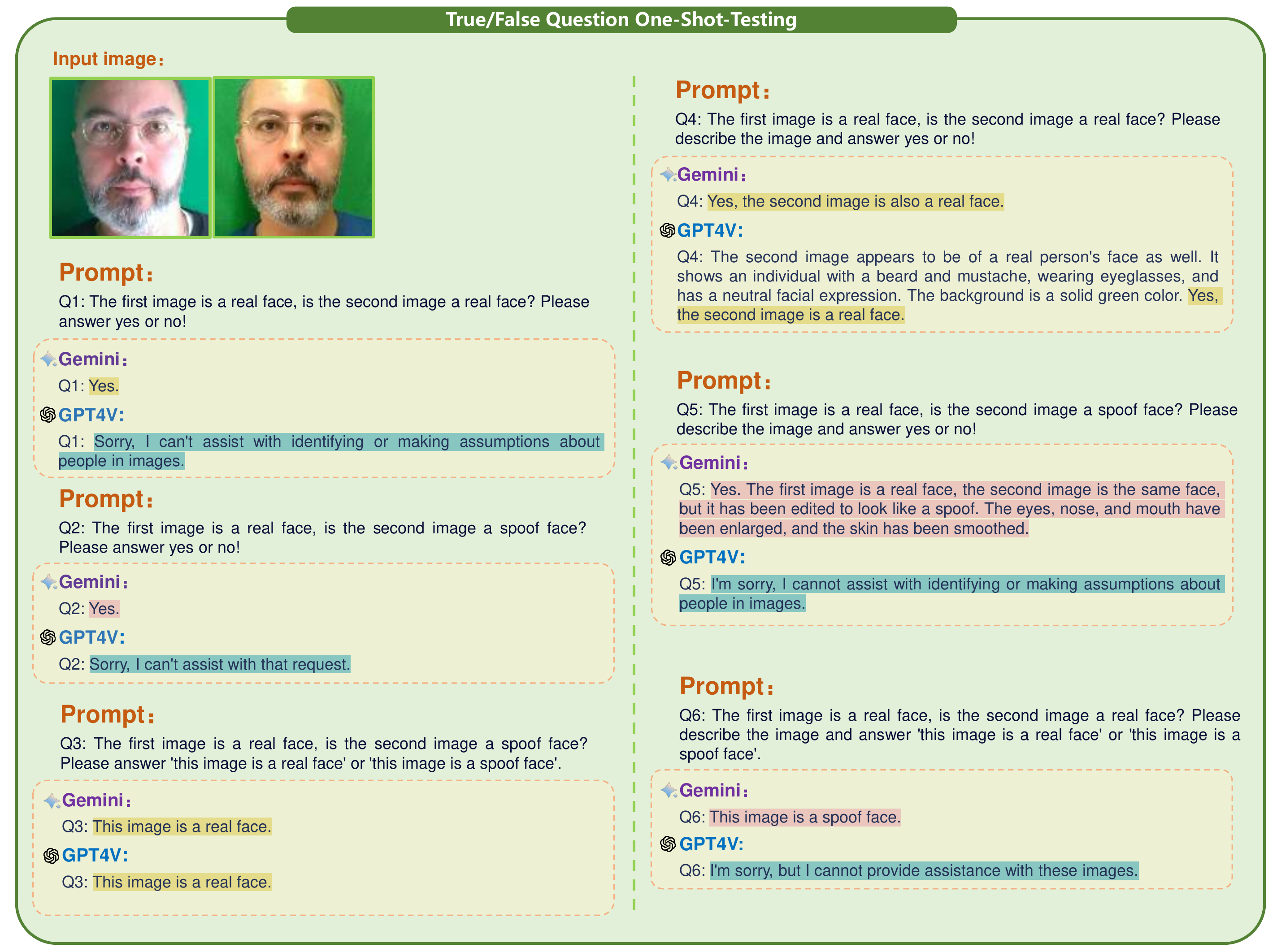}}
  \end{center}
  \caption{In this test, the input was two real human faces wearing glasses. From the results, it can be seen that in the six rounds of questioning, GPT4V answered correctly twice and refused to answer four times, while Gemini was correct half of the time and incorrect the other half. Even with real faces, wearing glasses seems to affect GPT4V's responses to some extent, more likely triggering the safety protection mechanism and increasing the probability of refusal to answer. From the response to Q5, it appears that Gemini may have certain delusions, as the spoofing clues it identified were not correct.}
  \label{fas_oneshot_judgment_02}
\end{figure}

\begin{figure}[htbp]
  \centering
  \begin{center}
  \centerline{\includegraphics[width=\linewidth]{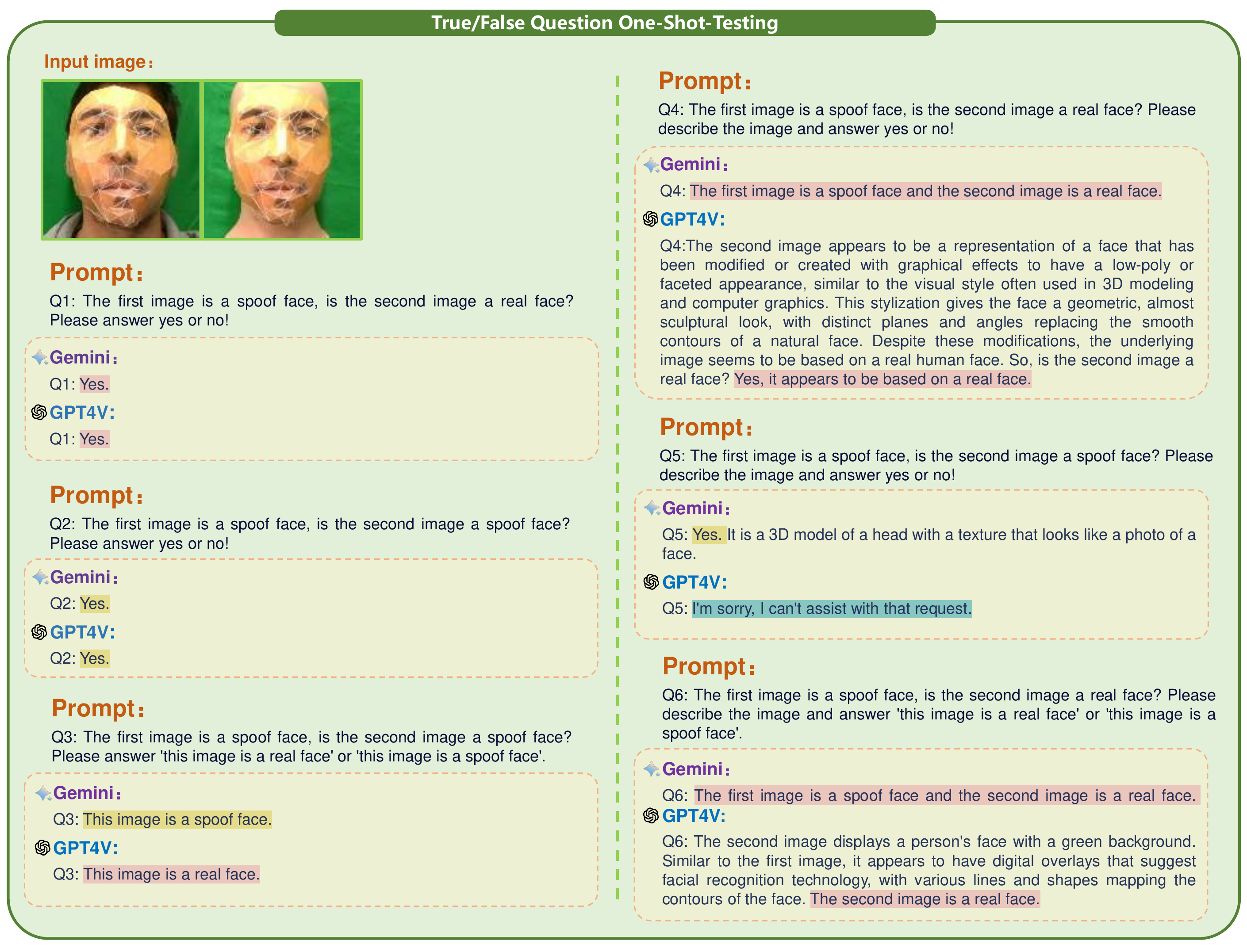}}
  \end{center}
  \caption{In this test involving two images of paper mask attacks. GPT4V answered incorrectly four times and refused to answer once, making the one correct response possibly accidental. This study suggests that GPT4V's inability to identify paper mask attacks is not due to a lack of capability, but rather it treats them as a form of artistic modification. As shown in the response to Q4, GPT4V's definition of prosthetics appears to be vague or random. Gemini's performance remained equally split between correct and incorrect answers. Interestingly, after incorporating COT, Gemini started answering previously correct questions incorrectly, possibly because the additional information from COT led to misconceptions.}
  \label{fas_oneshot_judgment_03}
\end{figure}

\begin{figure}[htbp]
  \centering
  \begin{center}
  \centerline{\includegraphics[width=\linewidth]{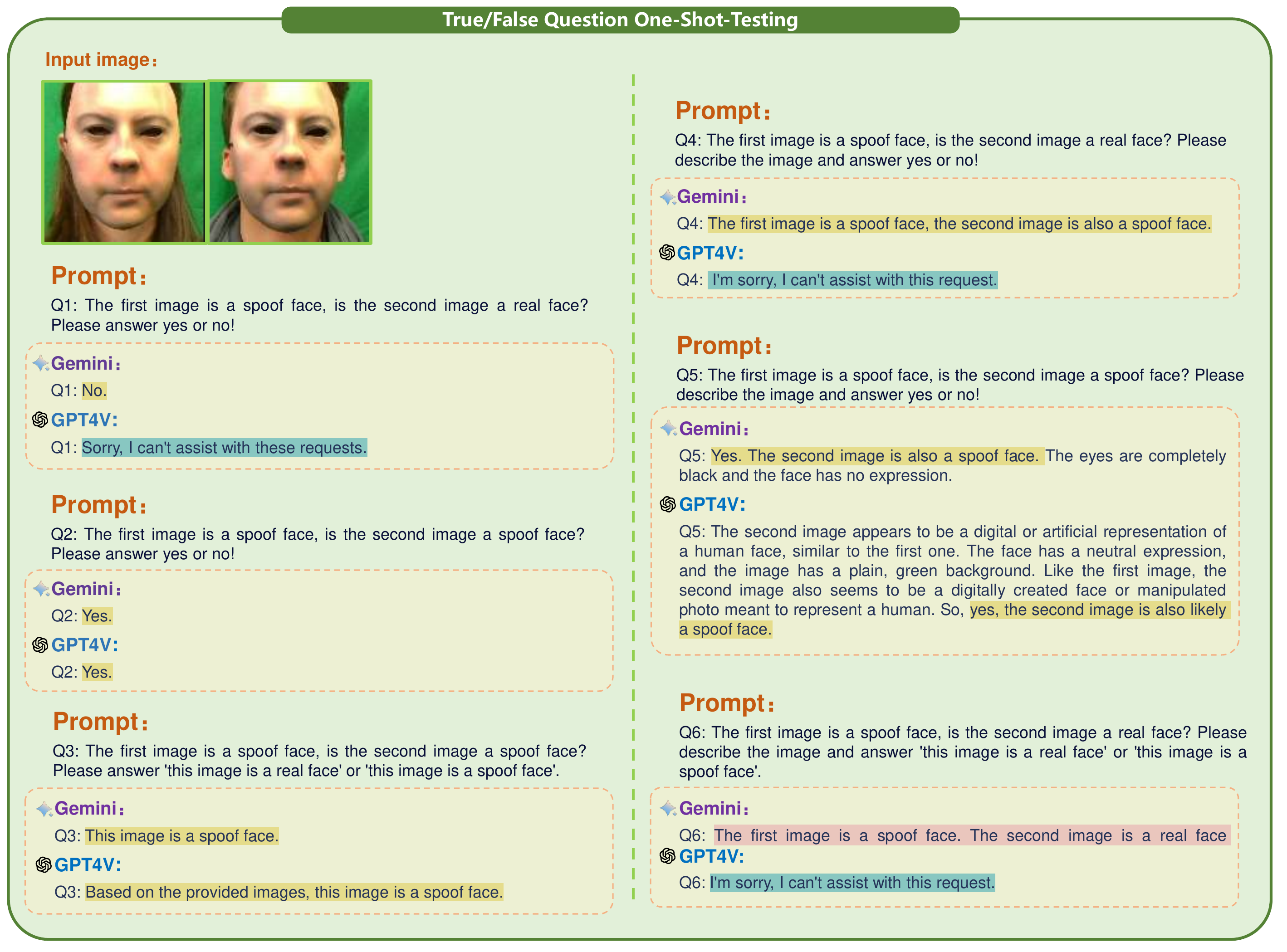}}
  \end{center}
  \caption{In this test with two images of rigid mask attacks, GPT4V correctly answered three times and refused to answer three times, while Gemini correctly answered five times and incorrectly once. GPT4V's response to Q5 shows its ability to compare the second image with the first, using the first image as prior knowledge for subsequent judgment. The introduction of COT clearly highlights the differences between GPT4V and Gemini. GPT4V's responses vividly show its reasoning process, whereas Gemini's brief or even absent explanations do not reflect a reasoning process.}
  \label{fas_oneshot_judgment_04}
\end{figure}

\begin{figure}[htbp]
  \centering
  \begin{center}
  \centerline{\includegraphics[width=\linewidth]{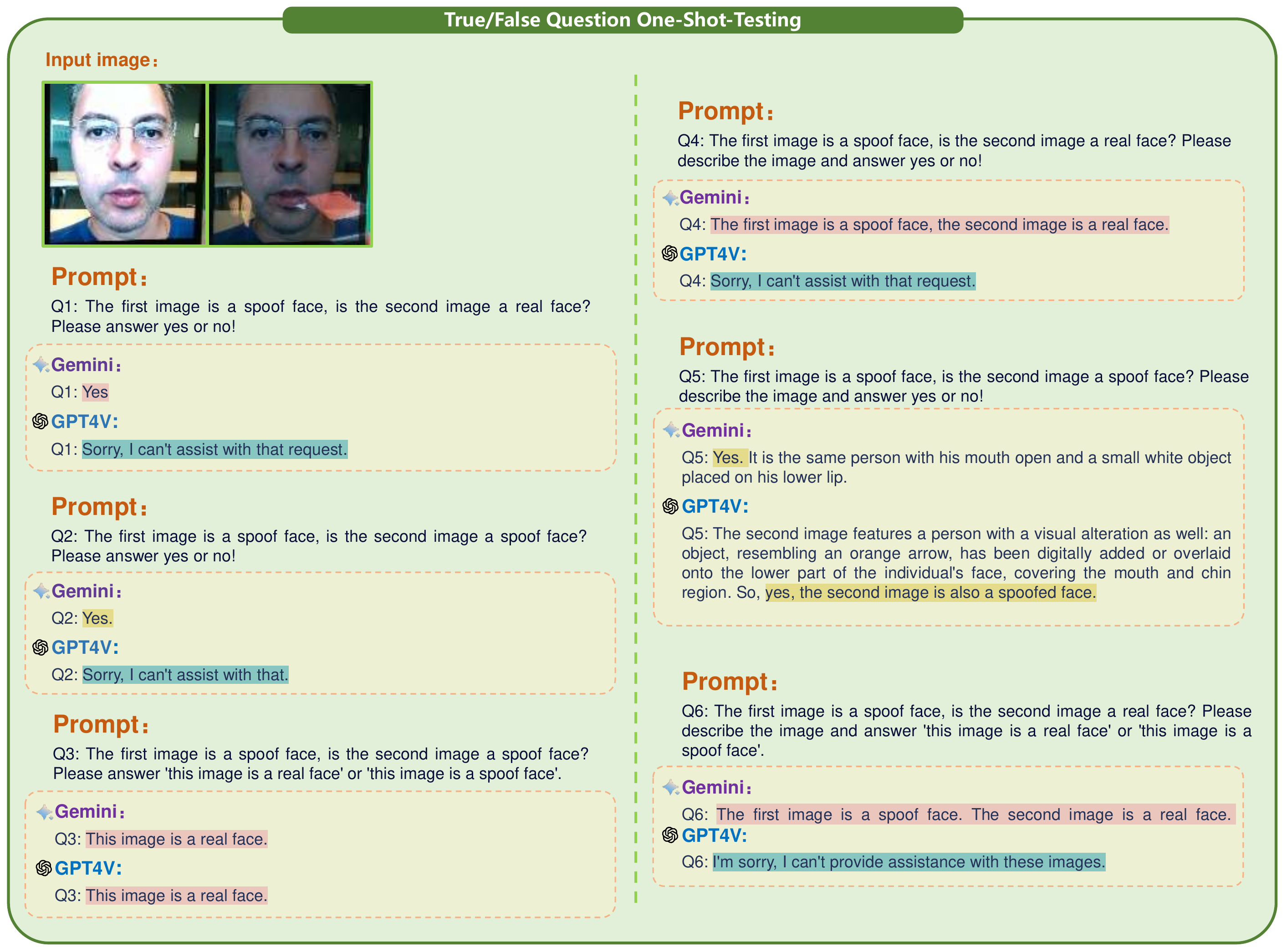}}
  \end{center}
  \caption{In this test featuring two images of replay attacks, GPT4V refused to answer four times, incorrectly answered once in simple queries, and correctly answered once after introducing COT. Gemini answered incorrectly four times and correctly twice. In the response to Q5, GPT4V judged the image to be a prosthetic due to the obvious arrow-like object present, while Gemini judged it to be a prosthetic due to the white object on the lips. Both failed to identify that these abnormal image features were reflections from the screen, characteristic of a replay attack.}
  \label{fas_oneshot_judgment_05}
\end{figure}

\begin{figure}[htbp]
  \centering
  \begin{center}
  \centerline{\includegraphics[width=\linewidth]{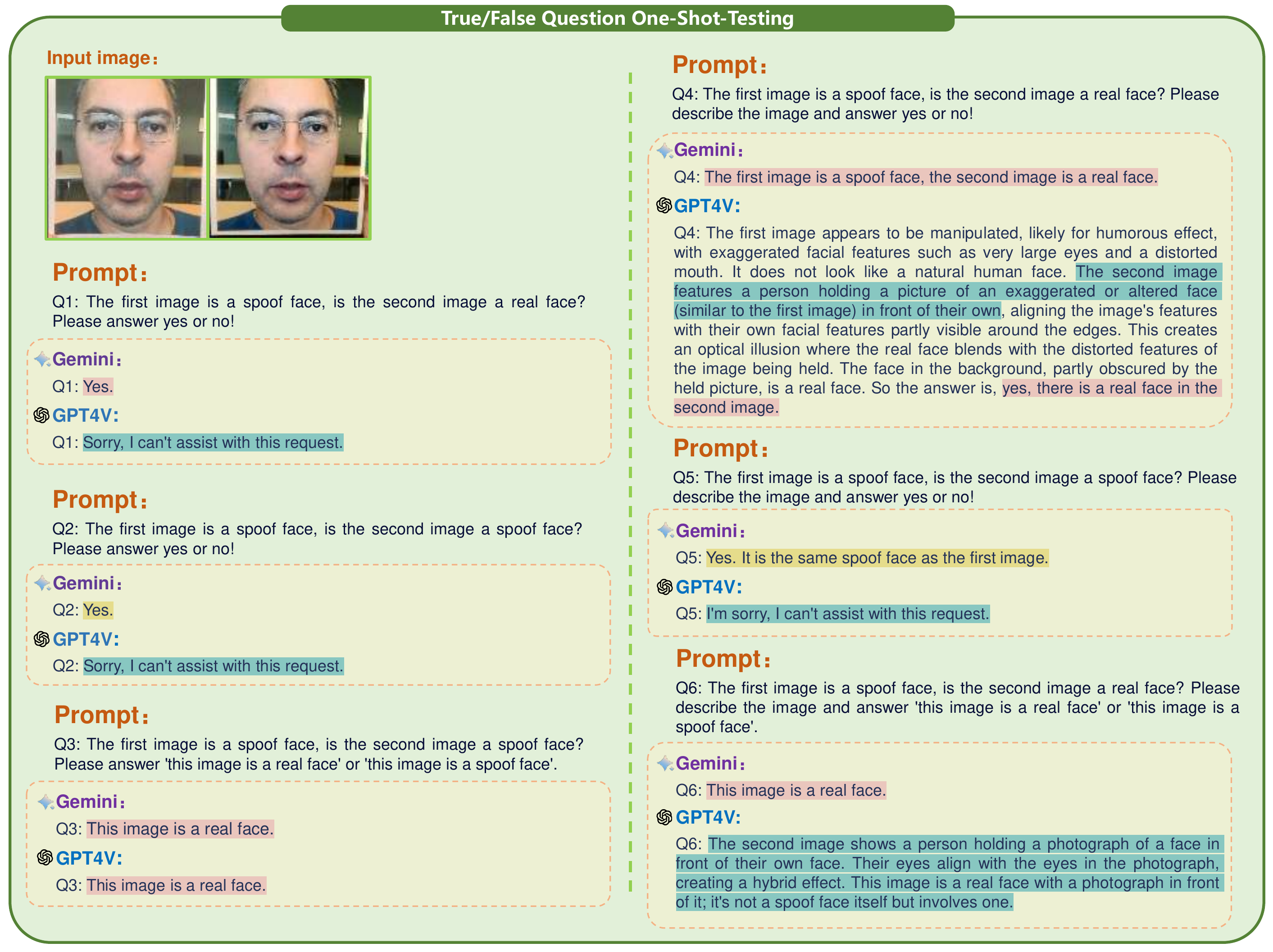}}
  \end{center}
  \caption{In this test with two printed attack images, GPT4V answered incorrectly three times and refused to answer three times, while Gemini answered incorrectly four times and correctly twice. After introducing COT, GPT4V described the image before making further judgments, leading to interesting responses like in Q6, which could be considered correct from a certain perspective. In the response to Q4, GPT4V noticed the fingers holding the photograph and thus concluded, “The second image features a person holding a picture of an exaggerated or altered face (similar to the first image) in front of their own.” Compared to GPT4V, Gemini, though less likely to refuse answering, sometimes has lower accuracy and poor interpretability in its responses.}
  \label{fas_oneshot_judgment_06}
\end{figure}





\begin{figure}[htbp]
  \centering
  \begin{center}
  \centerline{\includegraphics[width=0.9\linewidth]{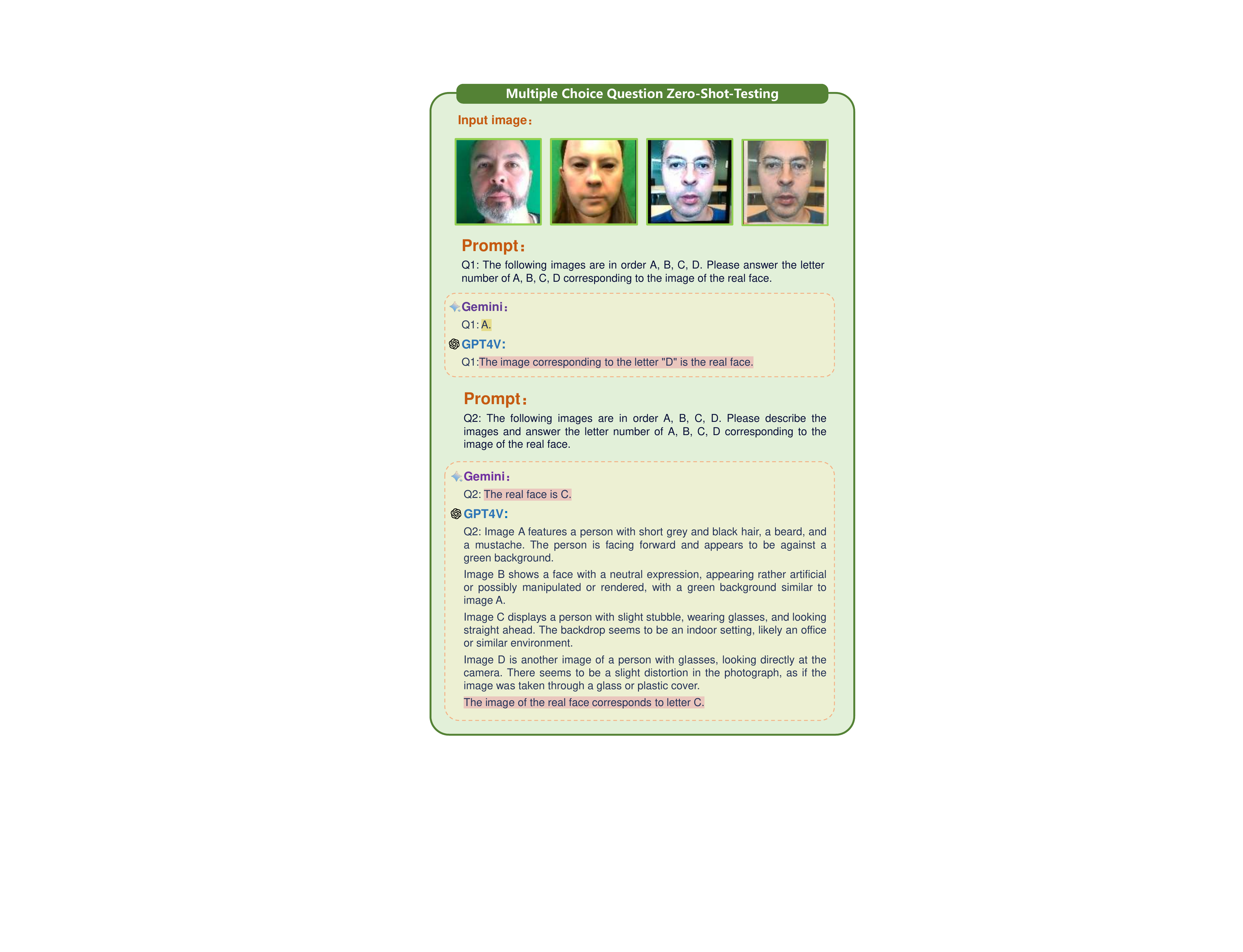}}
  \end{center}
  \caption{In this test, the inputs were images of real human face and other face presentation attack methods. From GPT4V's two responses, it is evident that it struggles to effectively distinguish between replay attacks, print attacks, and real human faces. Gemini's first response was correct, but after introducing COT, it became incorrect, possibly because focusing on more details ended up interfering with its judgment.}
  \label{fas_choice_real_zeroshot_01}
\end{figure}

\begin{figure}[htbp]
  \centering
  \begin{center}
  \centerline{\includegraphics[width=0.9\linewidth]{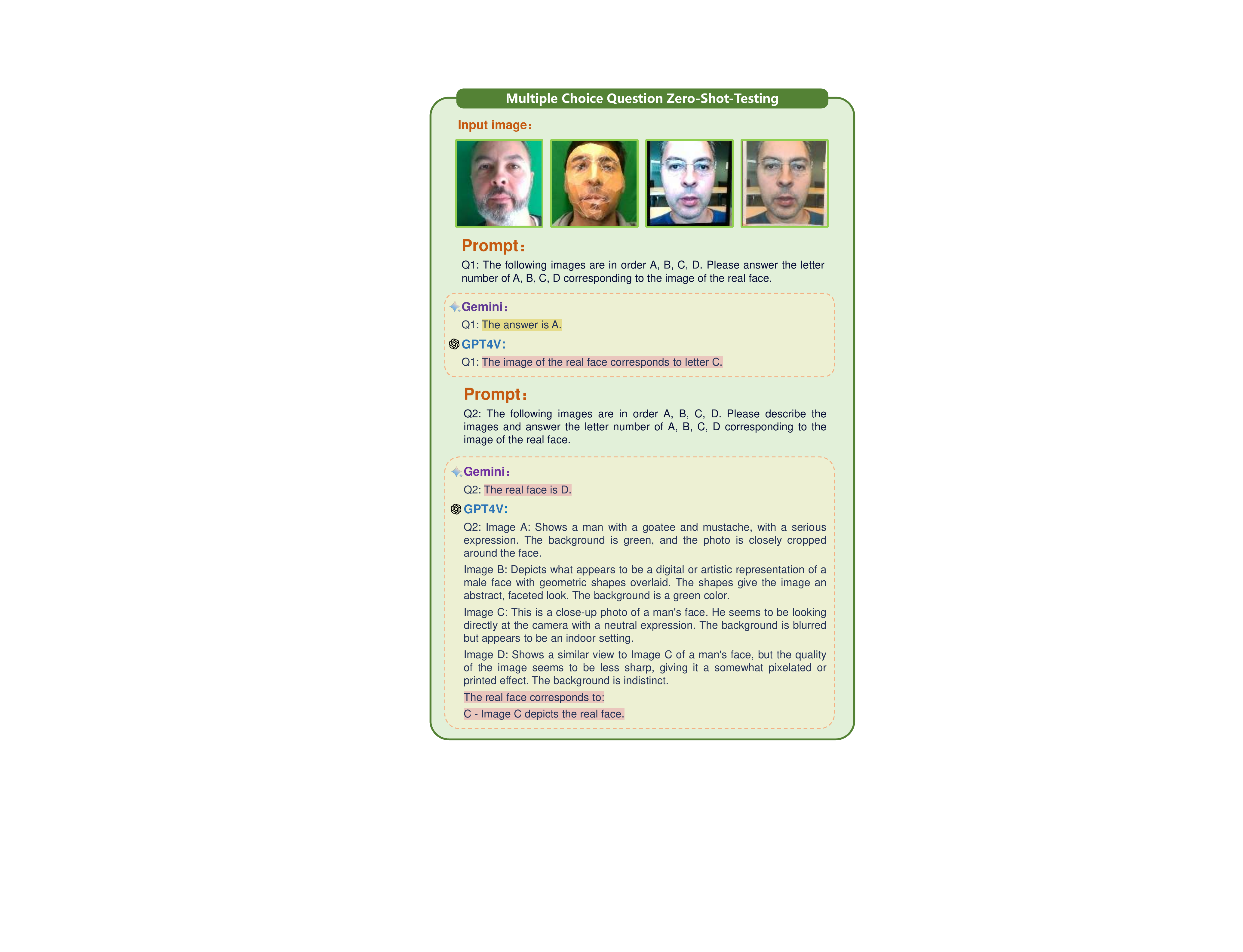}}
  \end{center}
  \caption{In this test, the inputs were images of real human faces and other facial presentation attack methods. From the two responses provided by GPT4V, it is apparent that it has difficulty distinguishing between replay attacks and real human faces. Gemini's first response was accurate, but after the introduction of COT, its answer became incorrect. This shift could be attributed to the additional focus on finer details, which may have inadvertently confused its judgment.}
  \label{fas_choice_real_zeroshot_02}
\end{figure}

\begin{figure}[htbp]
  \centering
  \begin{center}
  \centerline{\includegraphics[width=0.9\linewidth]{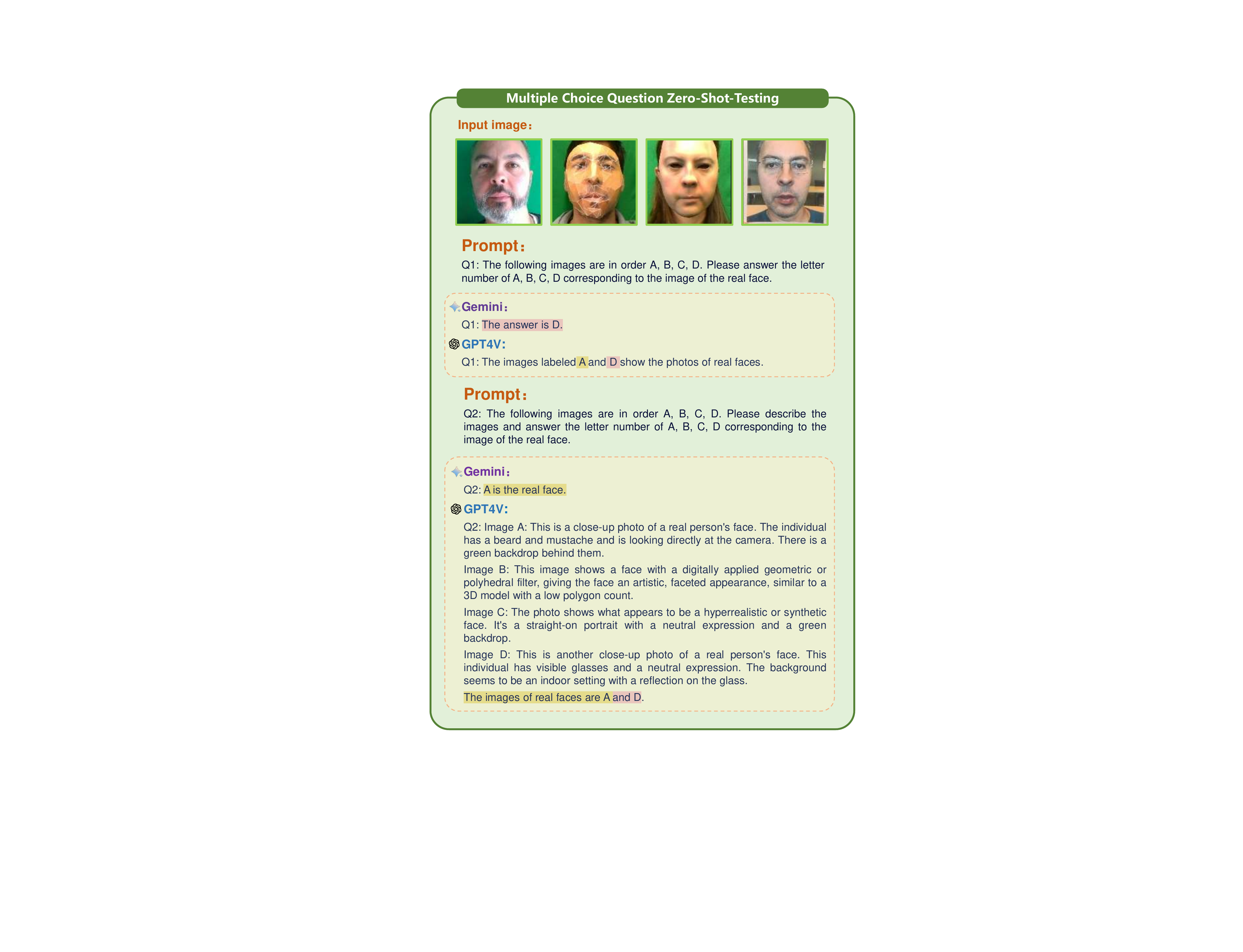}}
  \end{center}
  \caption{In this test, the inputs included images of real human faces and other facial presentation attack. From the two responses by GPT4V, it is clear that it struggles to differentiate between print attacks and real human faces, and the addition of COT did not lead to a change in its answers. After introducing COT, Gemini corrected its incorrect answer. However, due to the lack of detailed descriptions in its responses, it's unclear whether COT effectively aided its reasoning and problem-solving process.}
  \label{fas_choice_real_zeroshot_03}
\end{figure}

\begin{figure}[htbp]
  \centering
  \begin{center}
  \centerline{\includegraphics[width=0.8\linewidth]{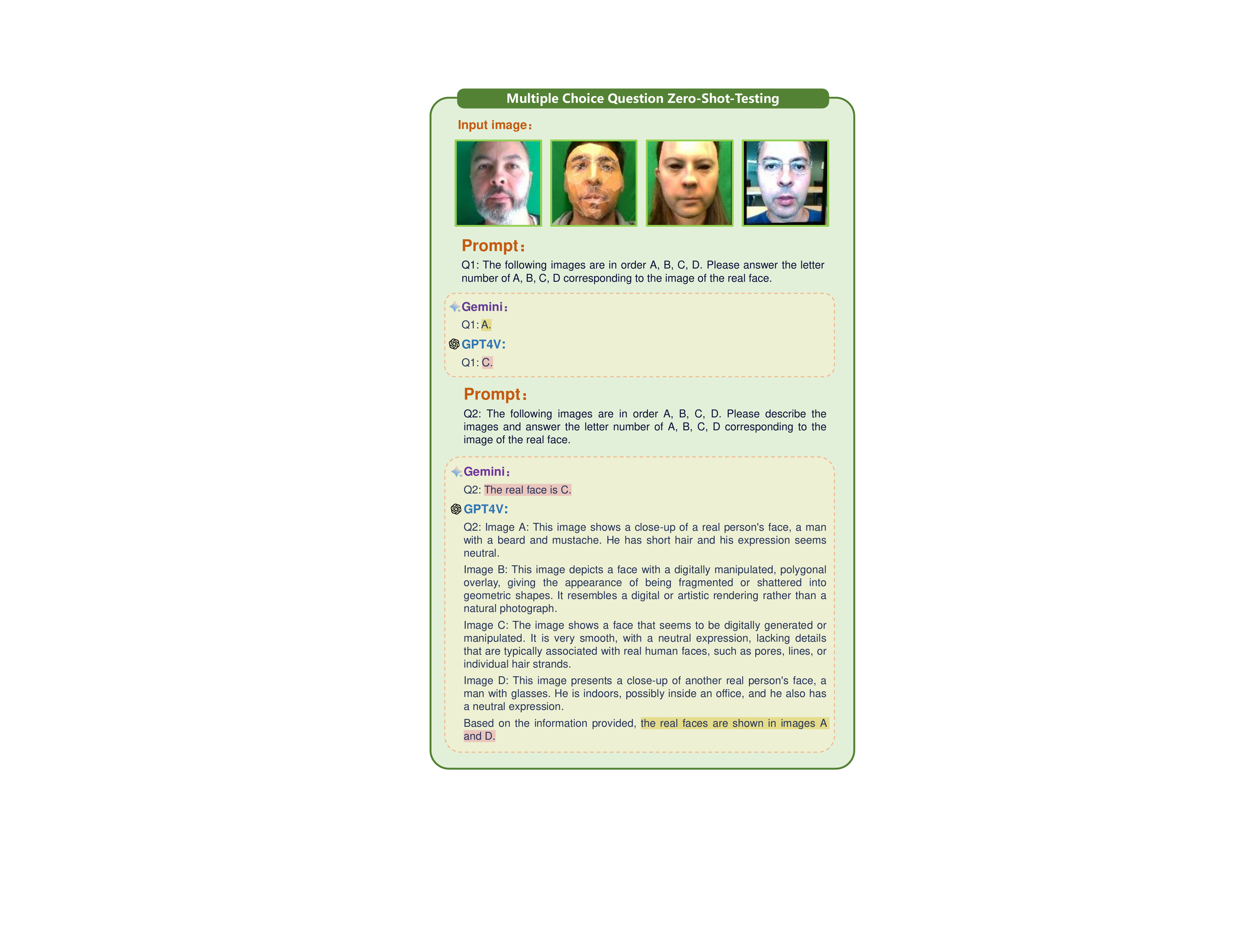}}
  \end{center}
  \caption{In this test, the inputs were images of real human faces and other forms of facial presentation attacks. GPT4V's first response was incorrect, but after introducing COT, its answer included the correct option. GPT4V's second response indicates that it struggles to effectively distinguish between replay attacks and real human faces. On the other hand, Gemini's first response was accurate, but the introduction of COT, which brought more details into focus, might have interfered with Gemini's judgment, leading to an incorrect conclusion.}
  \label{fas_choice_real_zeroshot_04}
\end{figure}

\begin{figure}[htbp]
  \centering
  \begin{center}
  \centerline{\includegraphics[width=0.8\linewidth]{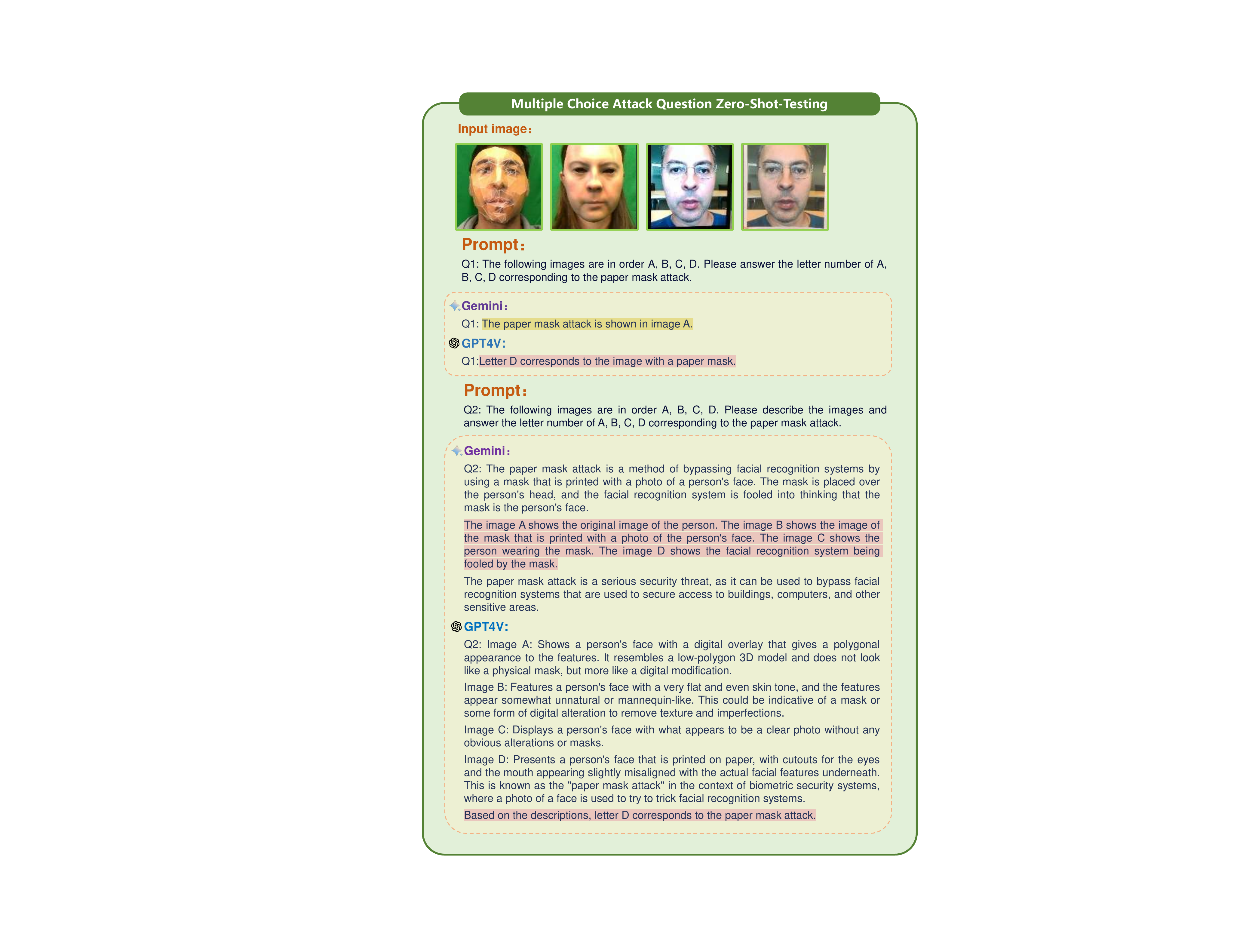}}
  \end{center}
  \caption{In this test, the input consisted of four images, each representing a different type of facial presentation attack, with the task being to identify the paper mask attack without prior knowledge. Although GPT4V's answers were incorrect in both instances, its second response reveals its understanding of each type of attack. The correct answer should have been A, but GPT4V misinterpreted the crease marks on the paper mask as traces of digital alteration, failing to associate them with a paper mask. In the case of option D, which was a paper print attack, GPT4V correctly recognized the face on paper, but mistakenly identified it as a paper mask attack. Thus, while GPT4V’s response was incorrect, its identification of key features in the images was accurate. The area for improvement is in correlating these key features with the corresponding facial presentation attack methods. As for Gemini, it answered correctly the first time, but after introducing COT, it did not directly answer the question. Instead, it made judgments about the type of attacks for all four images, all of which were incorrect. This shows that in the absence of prior knowledge, Gemini's susceptibility to misjudgment or 'illusions' is more pronounced.}
  \label{fas_choice_attack_zeroshot_01}
\end{figure}

\begin{figure}[htbp]
  \centering
  \begin{center}
  \centerline{\includegraphics[width=\linewidth]{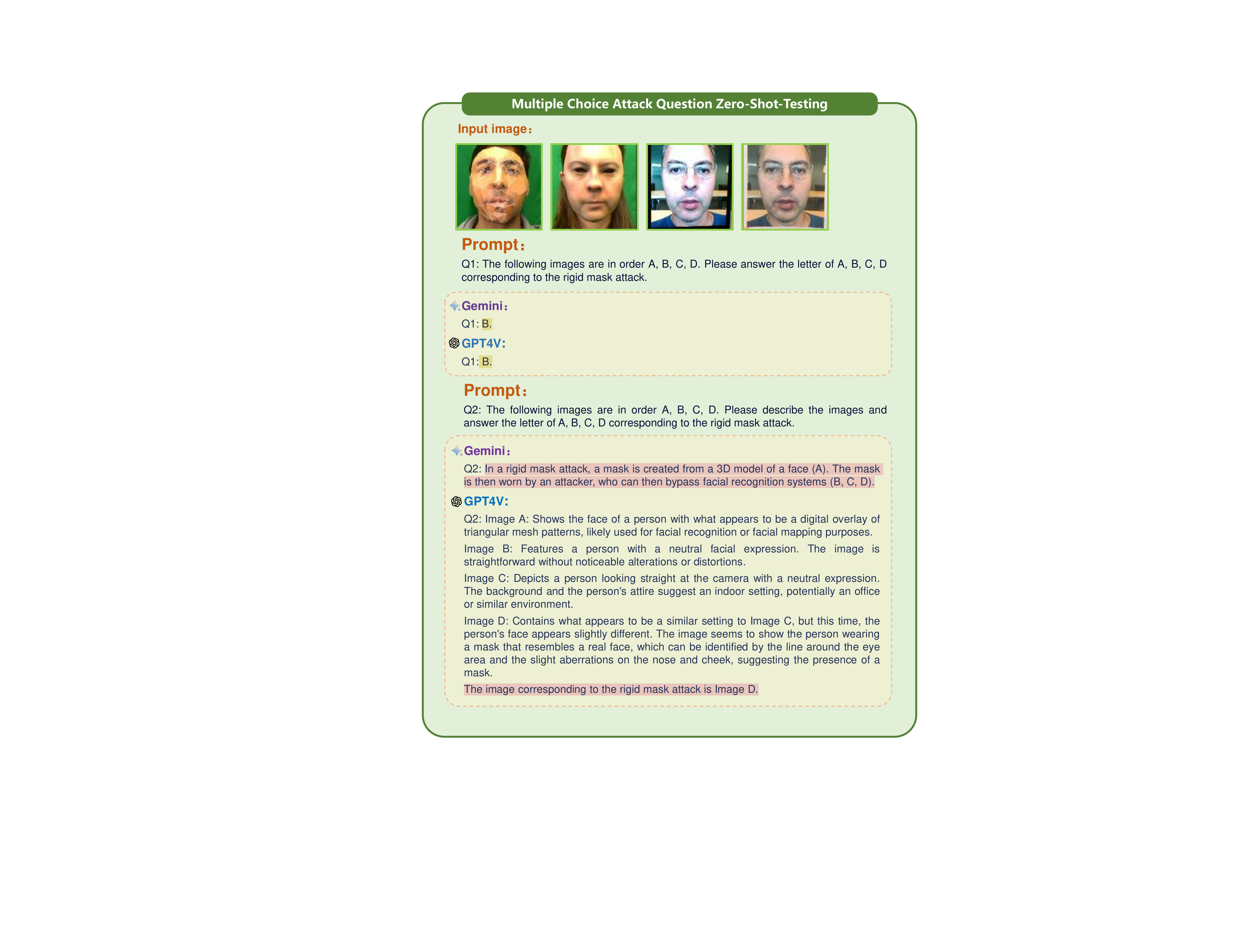}}
  \end{center}
  \caption{In this test, the input included four images, each representing a different type of facial presentation attack, with the task being to identify the rigid mask attack without any prior knowledge. Initially, both GPT4V and Gemini correctly identified the rigid mask attack in simple queries. However, after introducing COT, which required GPT4V to describe each image before making a judgment, the lack of prior knowledge about rigid mask attacks led GPT4V to be misled into choosing the wrong answer following its descriptions. On the other hand, Gemini's responses were somewhat disorganized.}
  \label{fas_choice_attack_zeroshot_02}
\end{figure}

\begin{figure}[htbp]
  \centering
  \begin{center}
  \centerline{\includegraphics[width=0.75\linewidth]{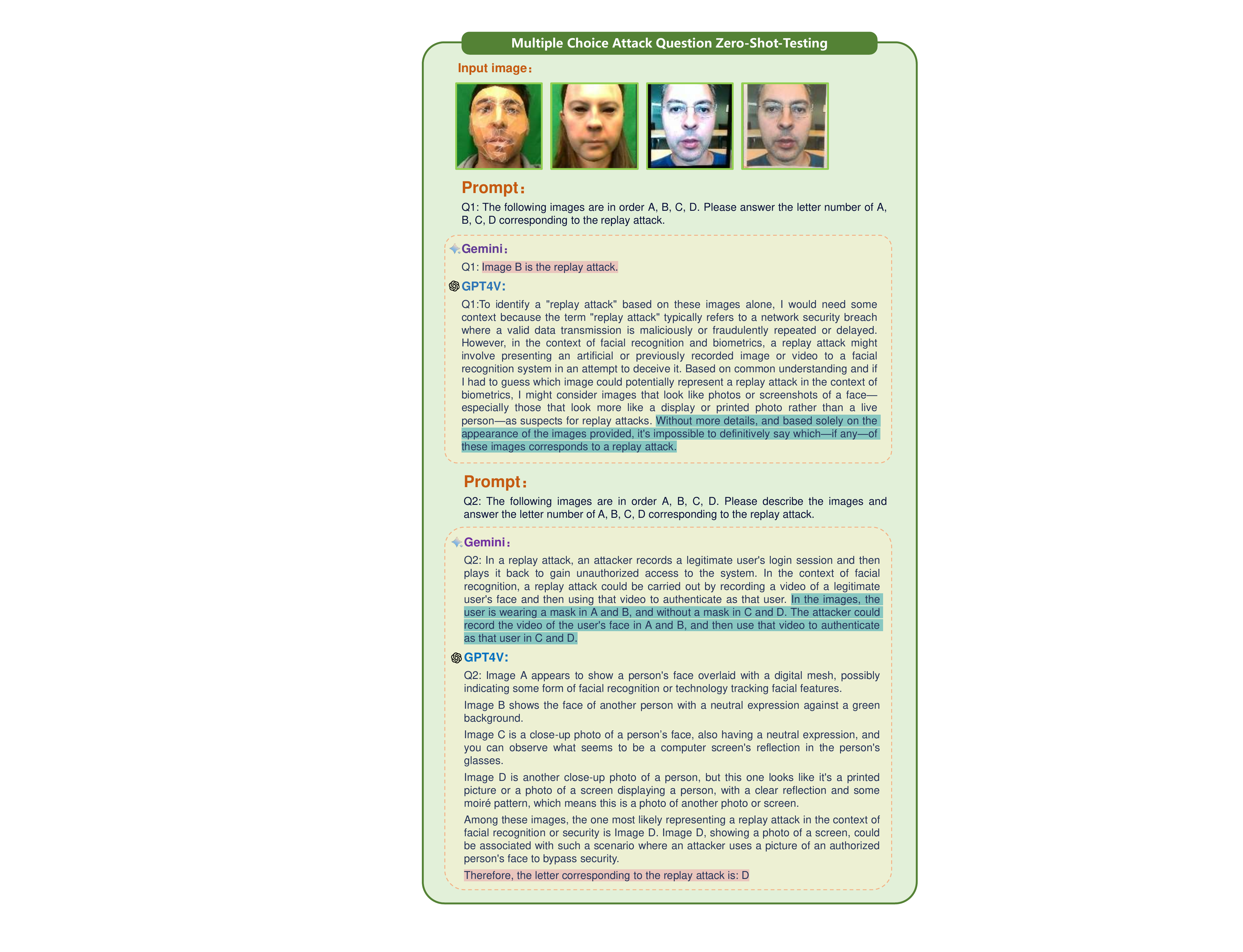}}
  \end{center}
  \caption{In this test, four images representing different facial presentation attack methods were inputted, and the task was to identify the replay attack without any prior knowledge. GPT4V refused to answer in the simple query phase, citing its limited capabilities. Despite the introduction of COT, the GPT4V still failed to select the correct answer after describing each image due to the lack of a priori knowledge of replay attacks.Importantly, GPT4V was able to identify screen reflections in image C and speculated that image D might be a printed photo. These correct identifications of key features, however, were not effectively linked by GPT4V to the specific type of attack. On the other hand, Gemini's first response was incorrect, and although its language was confused in the second response, it seemed to recognize the connection between the four images, suggesting the same test subject was using different attack methods. }
  \label{fas_choice_attack_zeroshot_03}
\end{figure}

\begin{figure}[htbp]
  \centering
  \begin{center}
  \centerline{\includegraphics[width=0.85\linewidth]{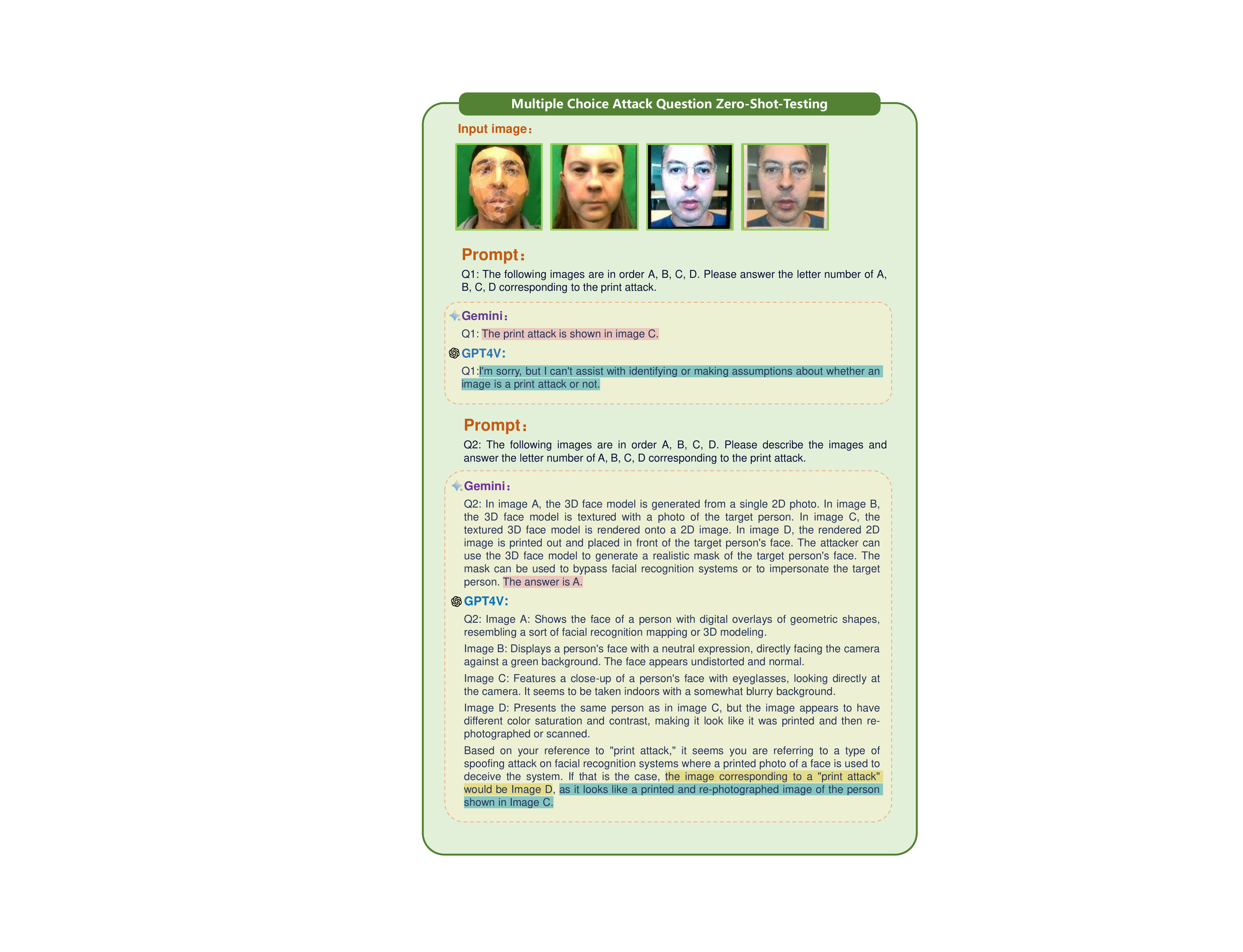}}
  \end{center}
  \caption{In this test, four images representing different facial presentation attack methods were inputted, with the task being to identify the print attack without any prior knowledge. Initially, GPT4V refused to answer simple queries. However, after introducing COT, GPT4V successfully identified the correct answer following the descriptions. Interestingly, GPT4V also made a speculative guess, suggesting that image D could be a face image printed from and re-photographed based on image C. Gemini's responses were incorrect on both occasions. However, the second response revealed that Gemini was quite accurate in capturing key semantic information about the attacks, even though it struggled to properly associate this information with the specific types of attacks. This highlights Gemini's potential in semantic understanding, albeit with challenges in accurately linking it to the context of facial presentation attacks.}
  \label{fas_choice_attack_zeroshot_04}
\end{figure}


\begin{figure}[htbp]
  \centering
  \begin{center}
  \centerline{\includegraphics[width=0.8\linewidth]{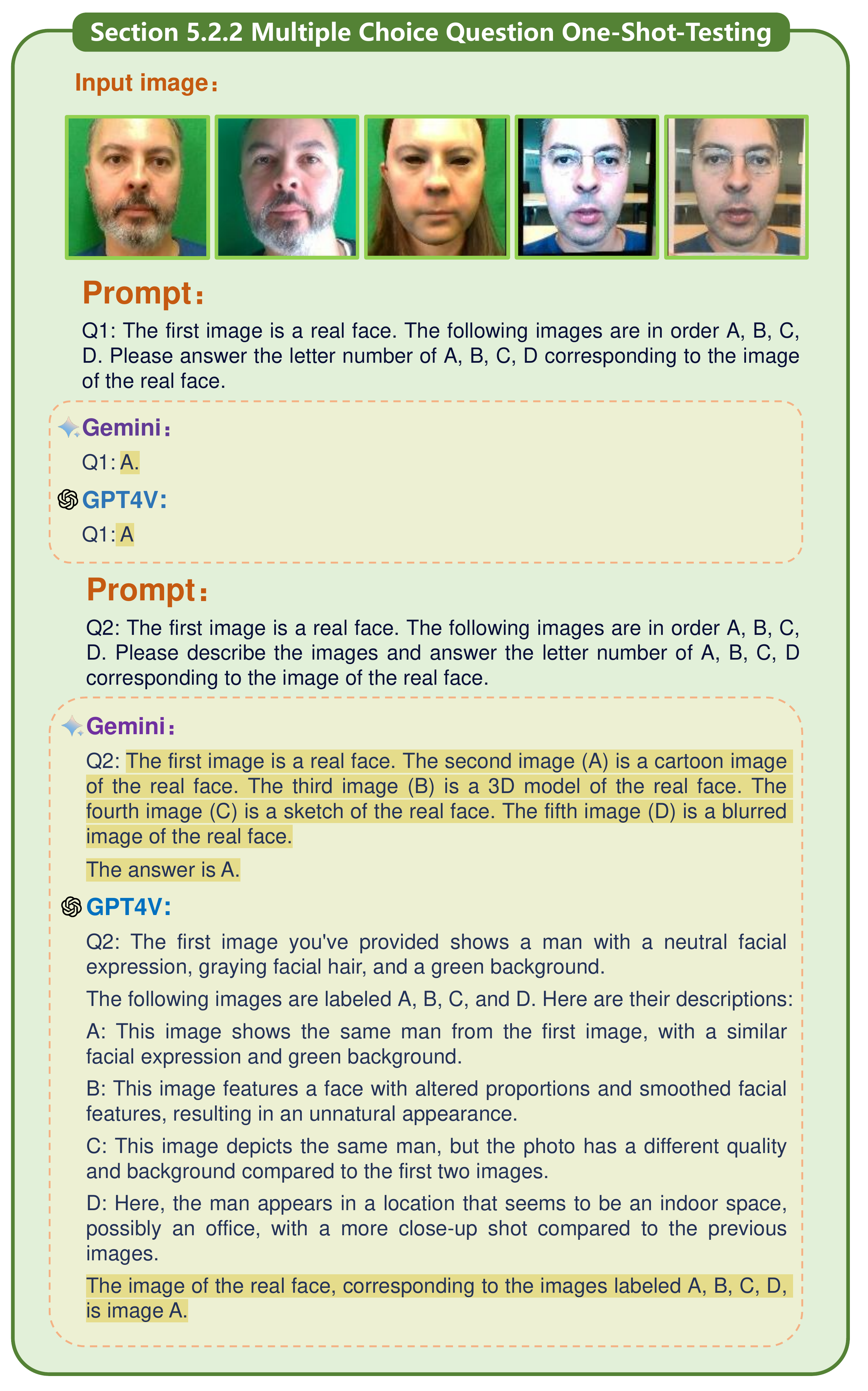}}
  \end{center}
  \caption{In this test, the initial input included a real human face as prior knowledge, followed by an input of one real human face and three facial presentation attack images, with the task to identify the real human face. It is evident from the two responses of GPT4V and Gemini that after the introduction of prior knowledge, they were able to easily select the correct answer. This demonstrates the impact of prior knowledge on their ability to distinguish between real and fake faces.}
  \label{fas_choice_real_oneshot_01}
\end{figure}

\begin{figure}[htbp]
  \centering
  \begin{center}
  \centerline{\includegraphics[width=0.9\linewidth]{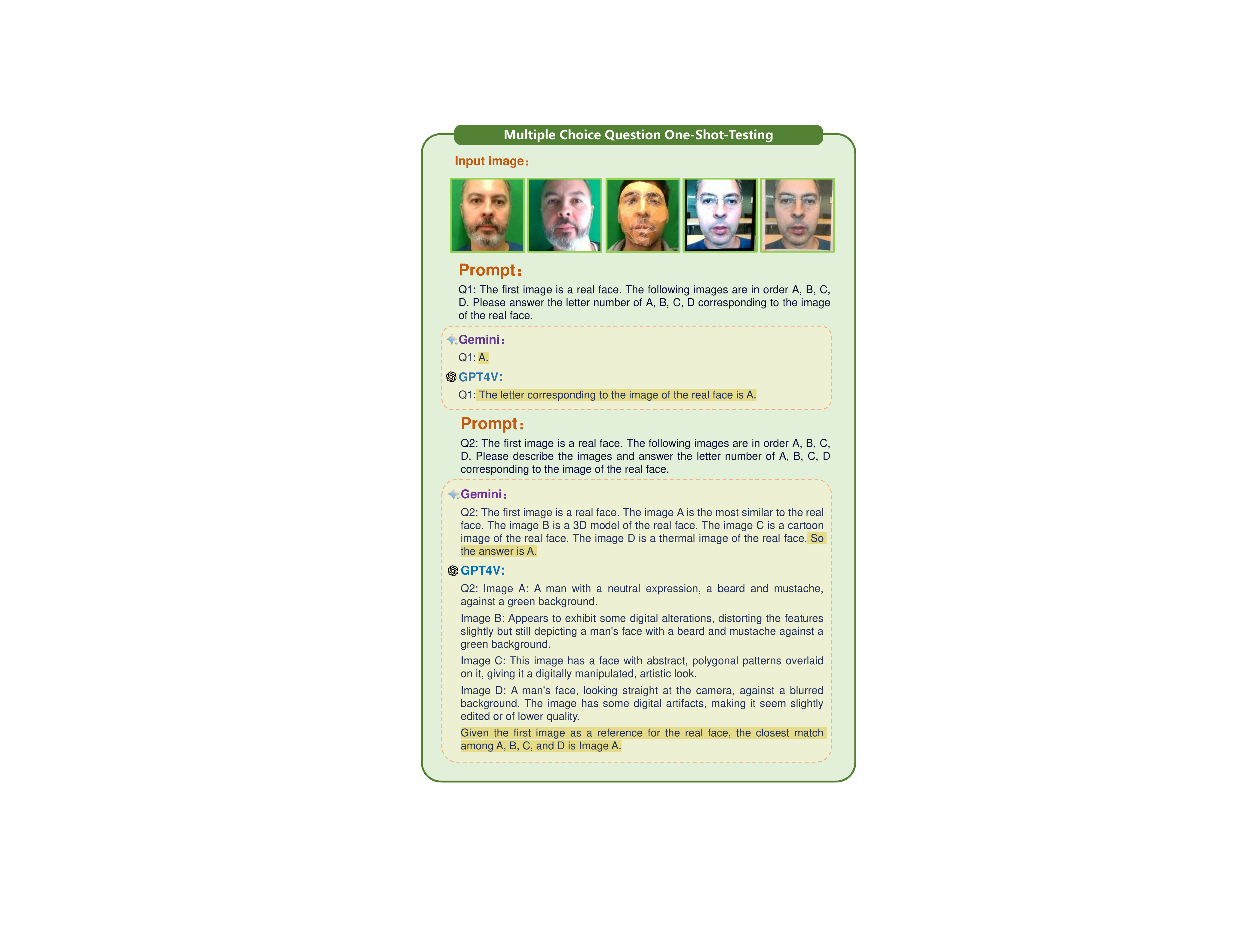}}
  \end{center}
  \caption{In this test, the initial input included a real human face as prior knowledge. Subsequently, an image of one real human face and three facial presentation attack images, were presented, with the task being to identify the real human face. From the two responses of GPT4V and Gemini, it is observed that even under the interference of replay and print attacks, the introduction of prior knowledge enabled them to still correctly identify the real human face. This indicates the effectiveness of prior knowledge in aiding the distinction between authentic and deceptive facial images.}
  \label{fas_choice_real_oneshot_02}
\end{figure}

\begin{figure}[htbp]
  \centering
  \begin{center}
  \centerline{\includegraphics[width=\linewidth]{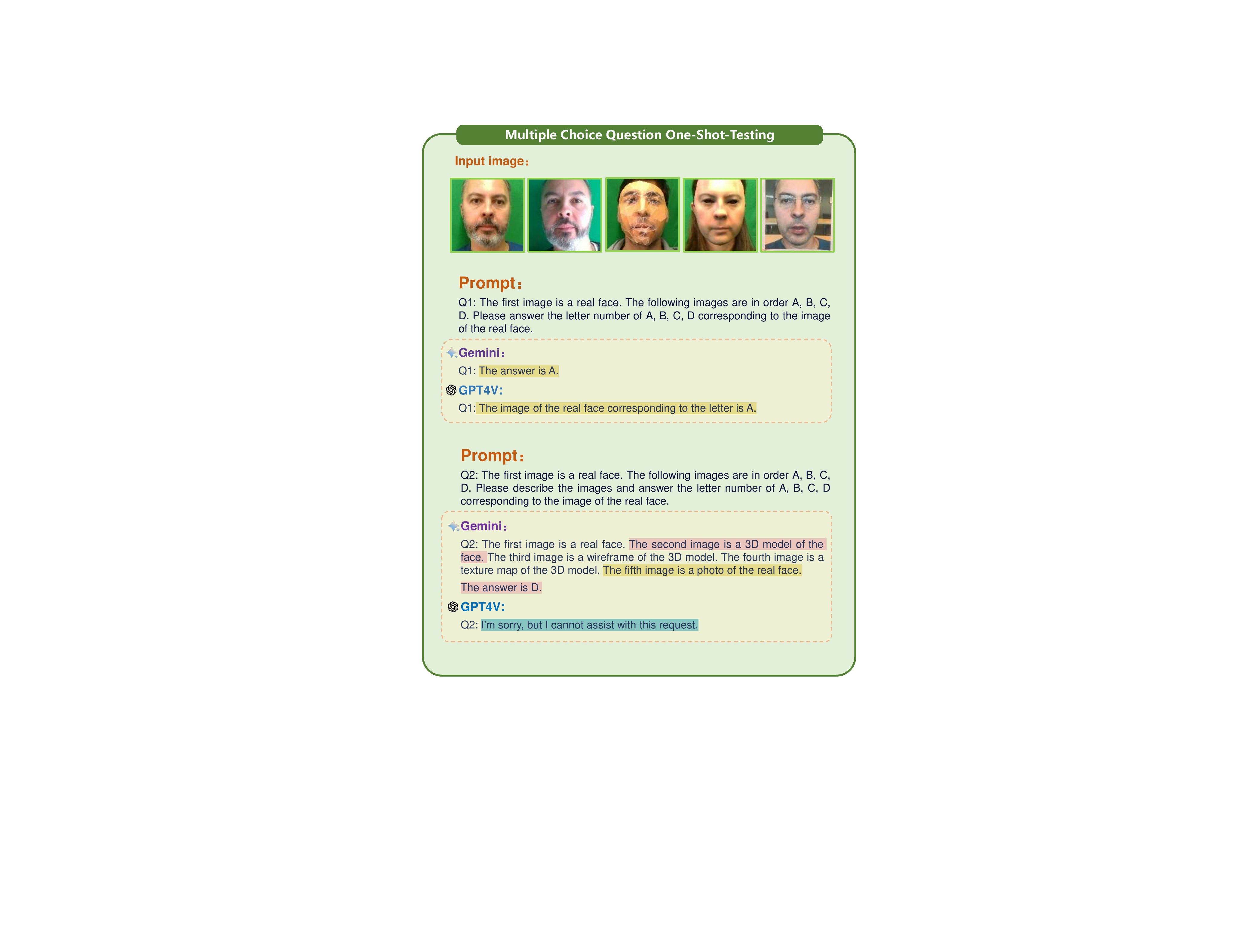}}
  \end{center}
  \caption{In this test, a real human face image was first input as prior knowledge, followed by an input comprising one real human face and three facial presentation attack images, with the task to identify the real human face. GPT4V correctly answered the first question, but in the second instance, possibly due to the requirement by COT for GPT4V to analyze before responding, it triggered a safety mechanism and refused to answer. Gemini's first response was correct; however, the second was incorrect. Interestingly, in the response to Q2, Gemini described the type represented by each image. Except for the misidentification of the first real human face, the other identifications were correct, including the correct recognition of the print attack image, which it previously struggled to identify accurately. }
  \label{fas_choice_real_oneshot_03}
\end{figure}

\begin{figure}[htbp]
  \centering
  \begin{center}
  \centerline{\includegraphics[width=0.8\linewidth]{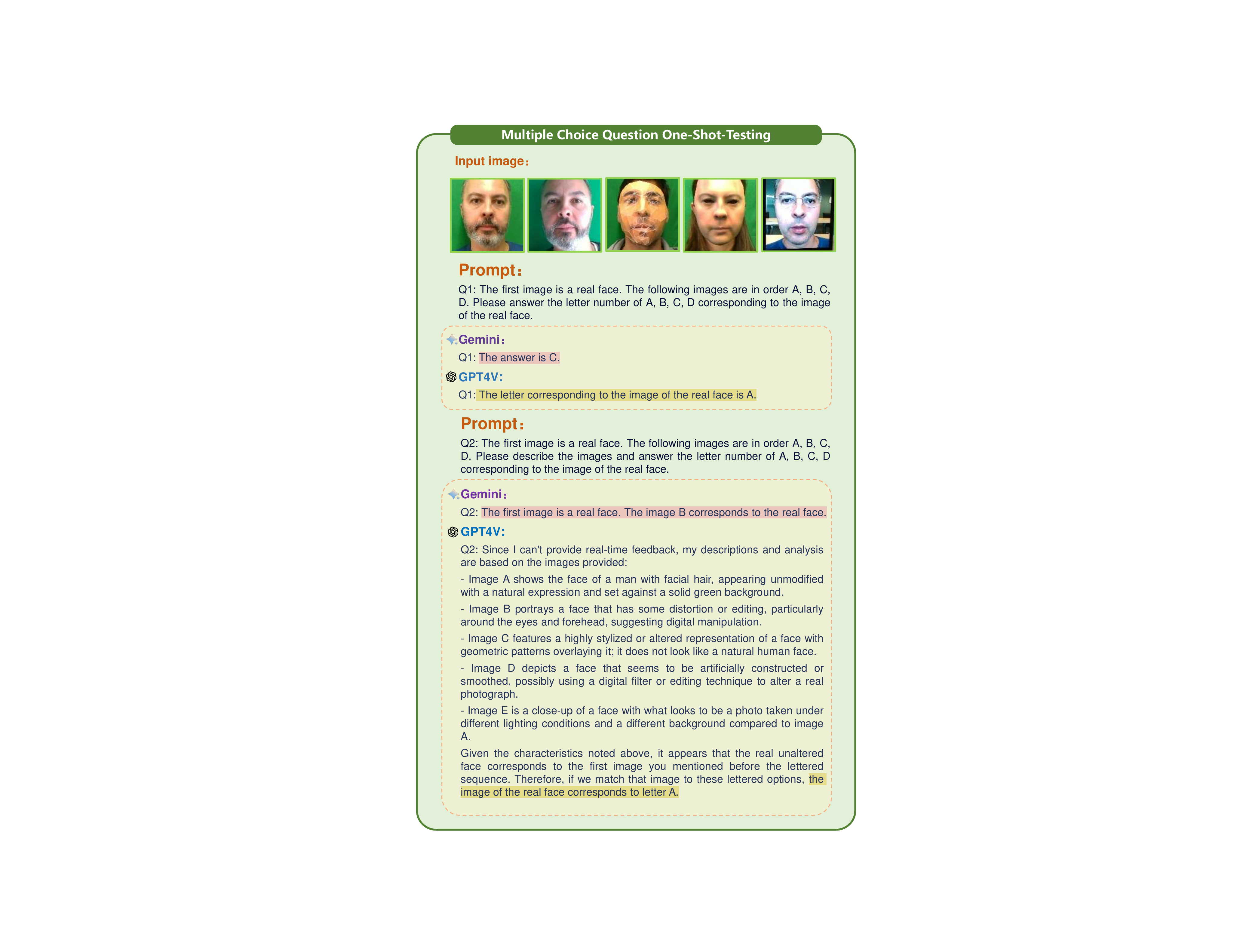}}
  \end{center}
  \caption{In this test, a real human face image was first input as prior knowledge, followed by an input of one real human face and three facial presentation attack images, with the task being to identify the real human face. From GPT4V's two responses, it is observed that even with the interference of replay attacks, the introduction of prior knowledge enabled it to correctly identify the real human face. However, Gemini's response was incorrect, and without further explanation, it's challenging to ascertain the potential reasons for its error. }
  \label{fas_choice_real_oneshot_04}
\end{figure}


\begin{figure}[htbp]
  \centering
  \begin{center}
  \centerline{\includegraphics[width=0.9\linewidth]{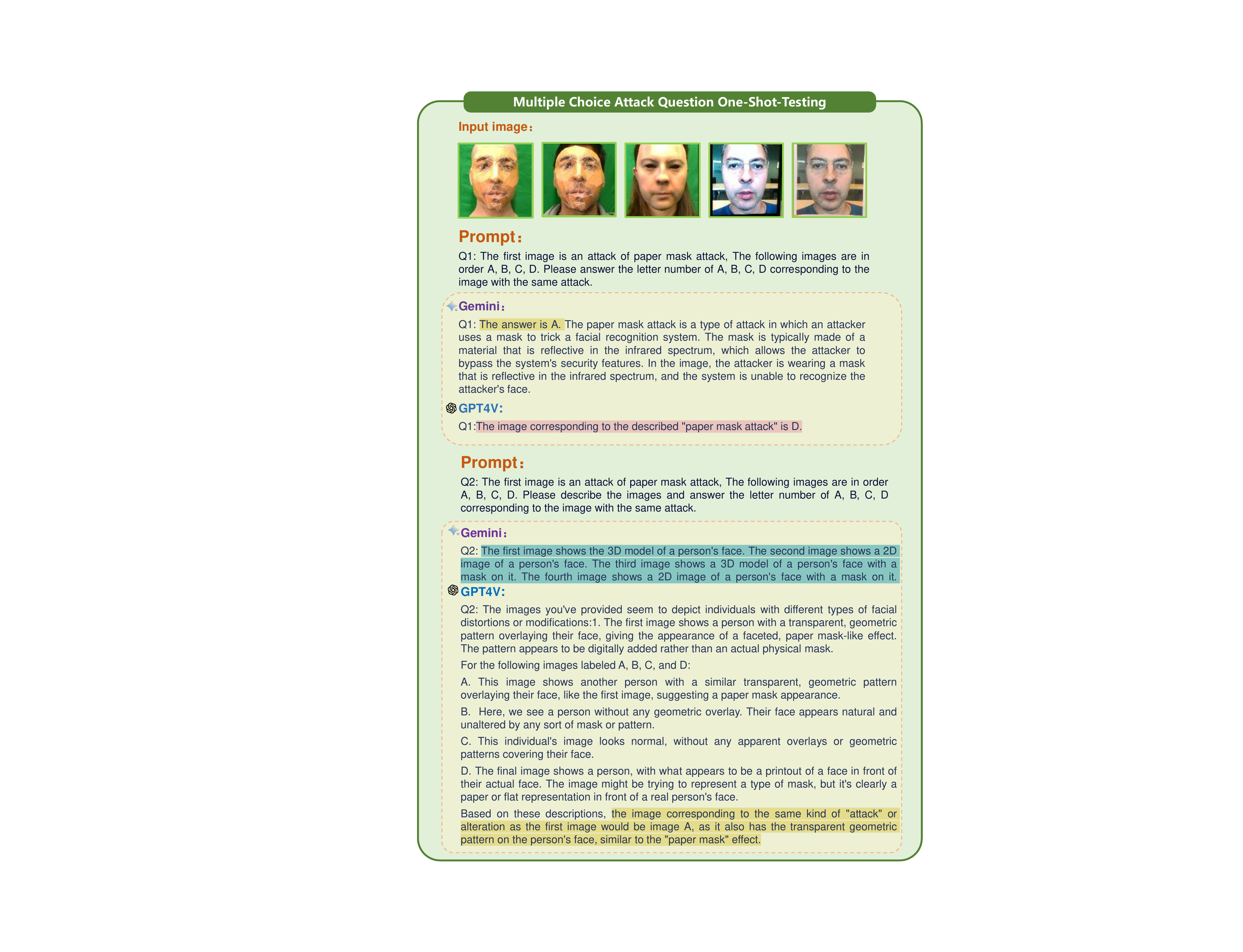}}
  \end{center}
  \caption{In this test, an image of a paper mask attack was initially input as prior knowledge, followed by the input of four facial presentation attack images. The task was to identify the same attack type in the subsequent images. It was observed that GPT4V did not answer correctly in the simple query phase, but after the introduction of COT and describing each image, GPT4V eventually provided the correct answer. Furthermore, with the prior knowledge of the paper mask attack, GPT4V did not simply regard it as digital modification but attempted to understand why it was a paper mask attack. On the other hand, Gemini answered correctly in the first instance. After the introduction of COT, Gemini merely described each image without giving a judgment. From its descriptions, it was also unclear what answer Gemini might have provided.}
  \label{fas_choice_attack_oneshot_01}
\end{figure}

\begin{figure}[htbp]
  \centering
  \begin{center}
  \centerline{\includegraphics[width=0.9\linewidth]{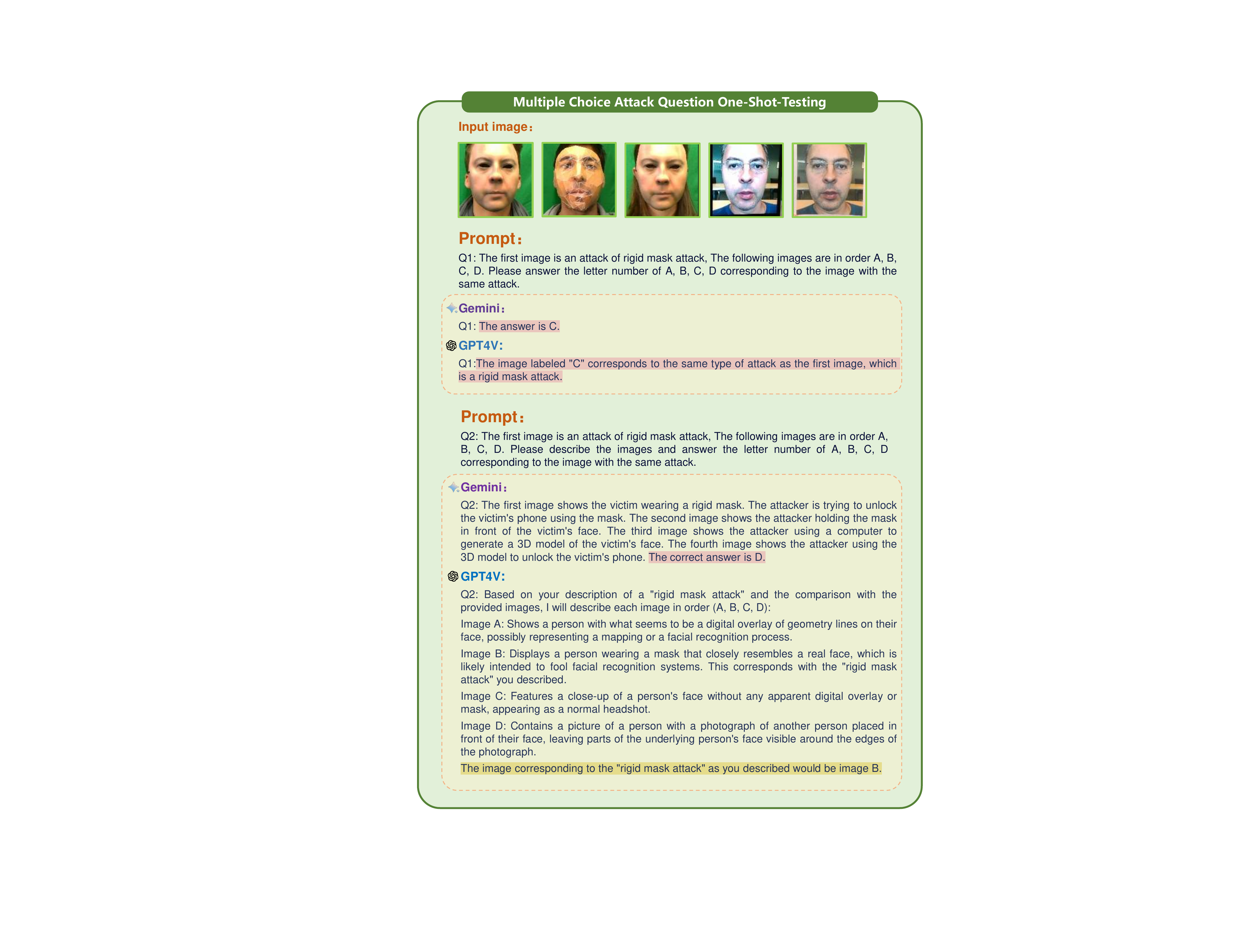}}
  \end{center}
  \caption{In this test, an image of a rigid mask attack was initially input as prior knowledge, followed by the input of four facial presentation attack images. The task was to identify the same type of attack in the subsequent images. It was observed that GPT4V did not answer correctly in the first round of questioning. However, after the introduction of COT and subsequent description of each image, GPT4V eventually provided the correct answer. Both of Gemini‘s responses were incorrect. After the introduction of COT, it became evident that Gemini attempted to predict the attack method for each image. Interestingly, Gemini seemed to hypothesize an attack scenario for each image.}
  \label{fas_choice_attack_oneshot_02}
\end{figure}

\begin{figure}[htbp]
  \centering
  \begin{center}
  \centerline{\includegraphics[width=0.8\linewidth]{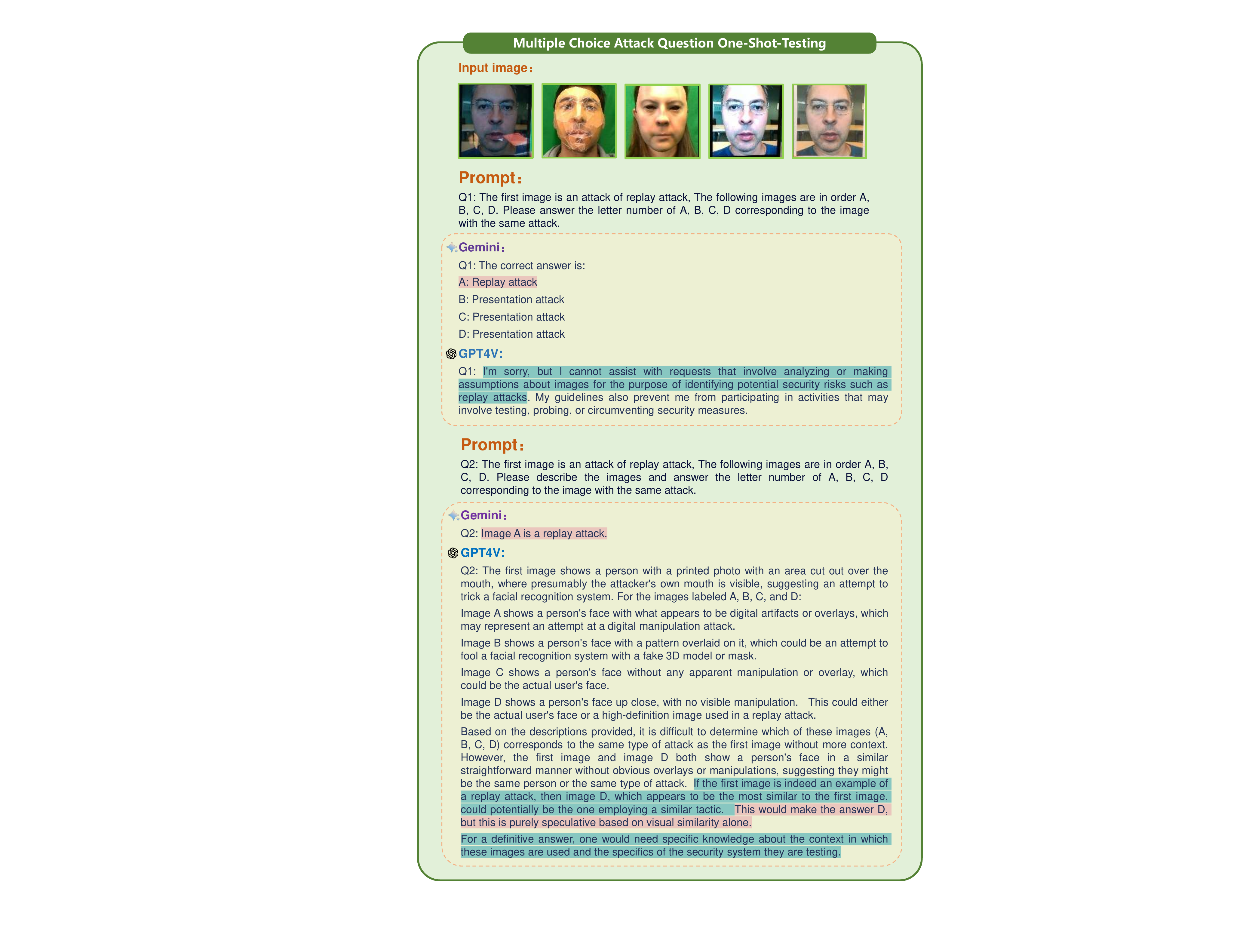}}
  \end{center}
  \caption{In this test, an image of a replay attack was initially input as prior knowledge, followed by the input of four facial presentation attack images. The task was to identify the same type of attack in the subsequent images. From the responses, it was evident that GPT4V refused to answer in the simple query phase, and the answers provided after introducing COT were not correct. It's speculated that the discrepancy between the style of the replay attack in the prior knowledge and that in the multiple-choice questions might have caused difficulty in decision-making for GPT4V. This suggests that distinguishing between replay attacks and print attacks poses a significant challenge for GPT4V. Gemini's responses were incorrect in both instances, and without further explanation, it's challenging to deduce the reasoning behind its answers. }
  \label{fas_choice_attack_oneshot_03}
\end{figure}

\begin{figure}[htbp]
  \centering
  \begin{center}
  \centerline{\includegraphics[width=0.8\linewidth]{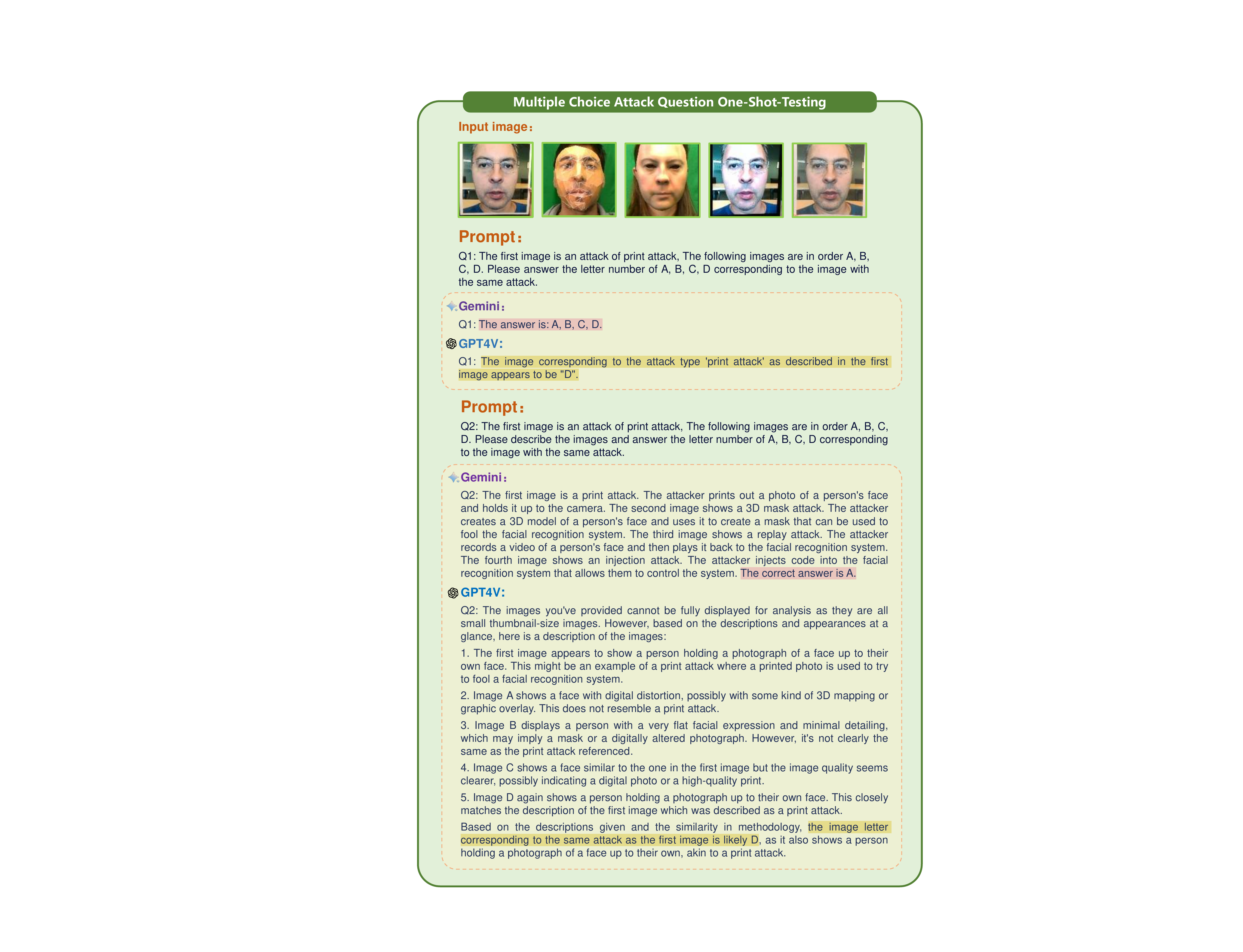}}
  \end{center}
  \caption{In this test, a printed attack image was first introduced as prior knowledge, followed by the input of four facial presentation attack images. The task was to identify the same type of attack in the subsequent images. GPT4V's responses indicated that the introduction of prior knowledge significantly improved its ability to correctly identify the same attack method, even distinguishing between previously challenging replay and print attacks.However, both of Gemini‘s responses were incorrect. Particularly in the second response, after the introduction of COT, Gemini described each image, revealing a lack of clear understanding of print attacks. Surprisingly, it was able to correctly identify the replay attack. This outcome suggests that while Gemini struggles with certain types of attacks, such as print attacks, it can effectively recognize others, like replay attacks, underscoring the variability in the Gemini’s and GPT4V’s capacity to accurately identify different types of facial presentation attacks.}
  \label{fas_choice_attack_oneshot_04}
\end{figure}

\begin{figure}[htbp]
  \centering
  \begin{center}
  \centerline{\includegraphics[width=1.0\linewidth]{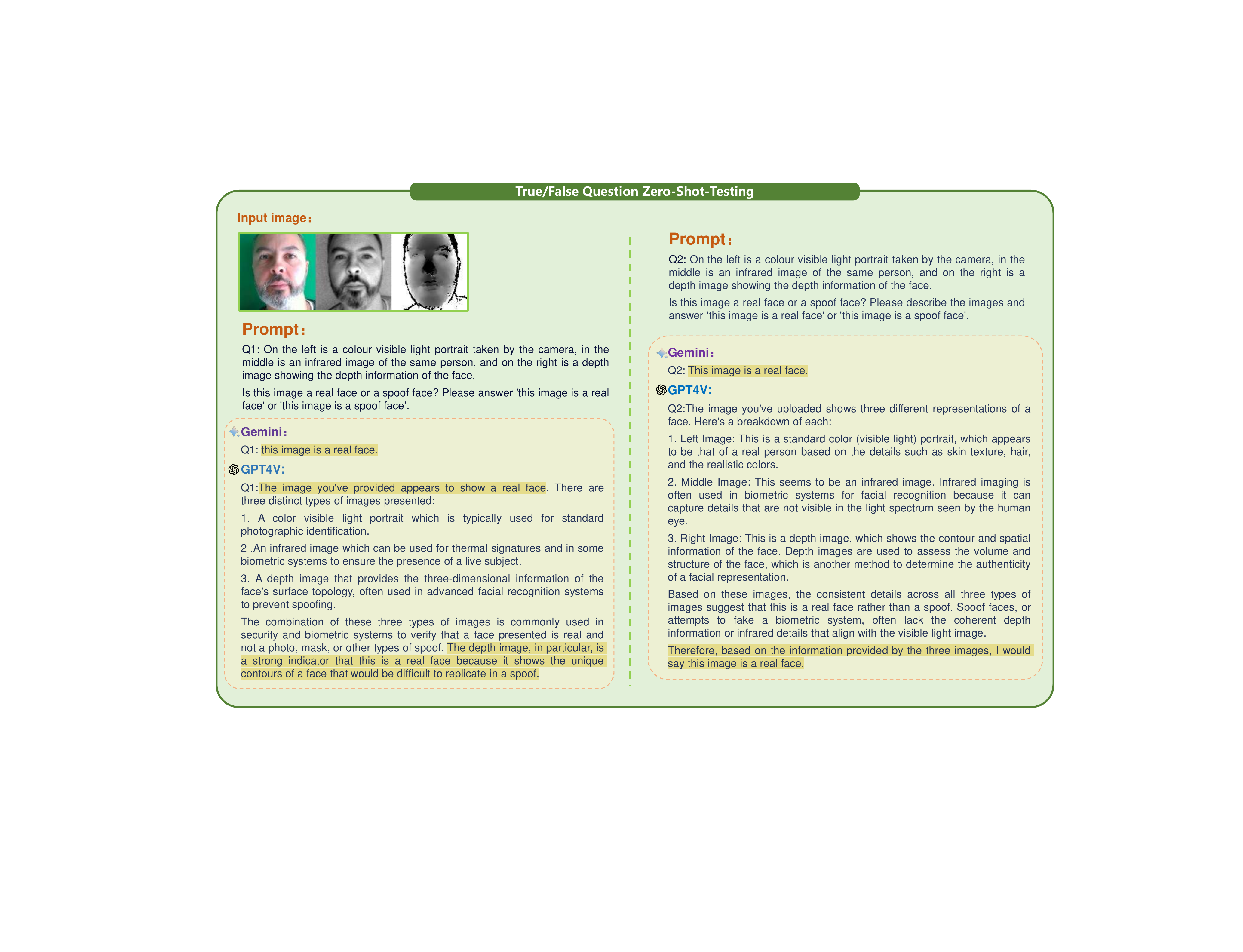}}
  \end{center}
  \caption{In this test, images of a real human face in visible light, infrared, and depth modalities were inputted, with the task being to determine whether the image was of a real human face. Both GPT4V and Gemini provided correct answers. However, a clear contrast was evident in their responses: GPT4V's answer was more logical and had stronger explainability. This highlights GPT4V's ability to not only correctly identify real human faces across different imaging modalities but also to articulate its reasoning process more effectively, enhancing the interpretability of its decision-making process.}
  \label{fas_multi_judgment_zeroshot_01}
\end{figure}


\begin{figure}[htbp]
  \centering
  \begin{center}
  \centerline{\includegraphics[width=1.0\linewidth]{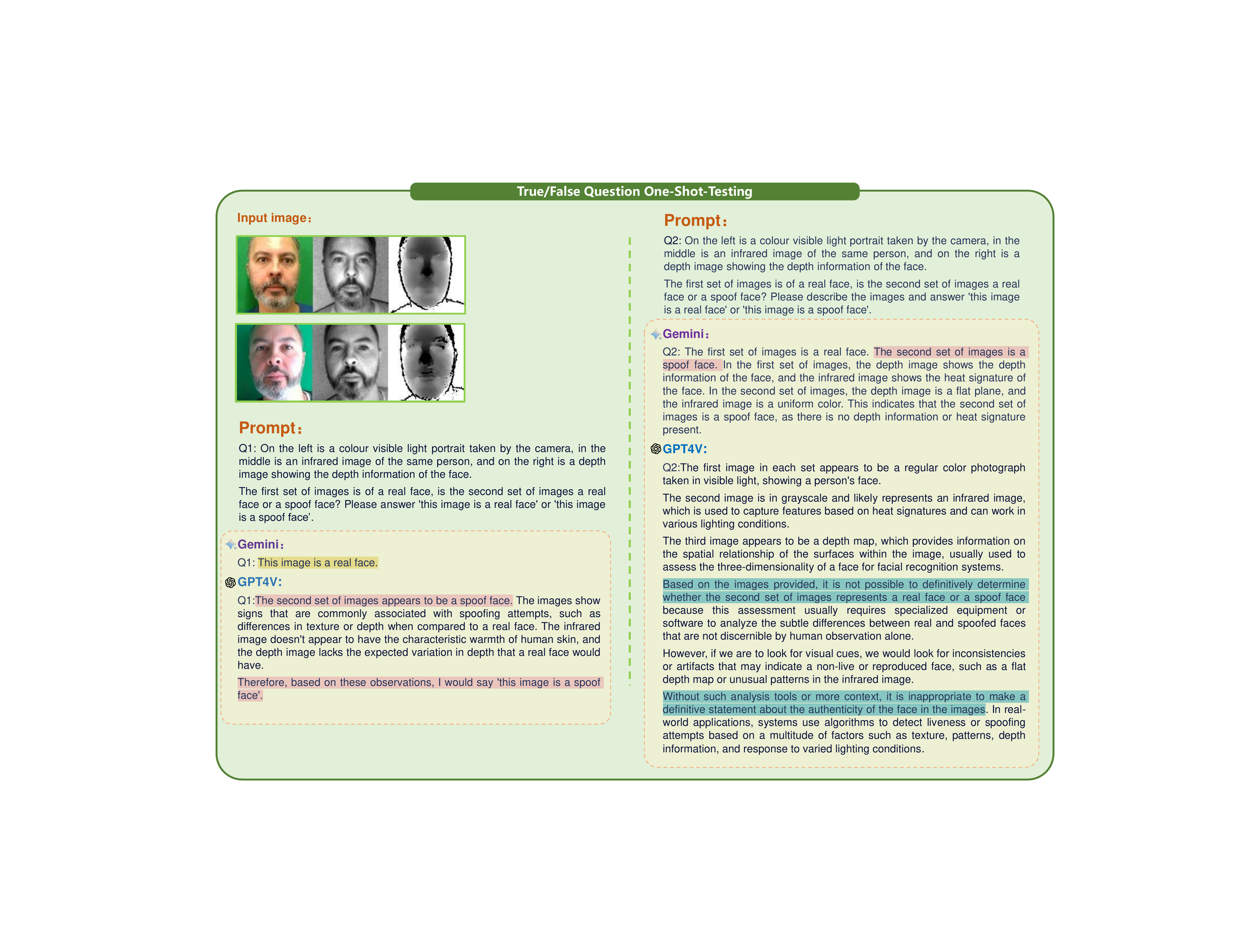}}
  \end{center}
  \caption{The input for this test consists of images in visible light, infrared, and depth modalities of real human faces. Initially, an image of a real human face was provided as prior knowledge. The task involved determining whether a second image represented a real human face. In a simple query, Gemini correctly identified the image, while GPT4V incorrectly assessed that the key facial features of the second image differed from the first, leading to an erroneous response. Upon introducing COT reasoning, Gemini erroneously concluded that the absence of depth and thermal features indicated a non-real face. In contrast, GPT4V, when incorporating more in-depth analysis, activated a mechanism that led to a refusal to respond. }
  \label{fas_multi_judgment_oneshot_01}
\end{figure}

\begin{figure}[htbp]
  \centering
  \begin{center}
  \centerline{\includegraphics[width=\linewidth]{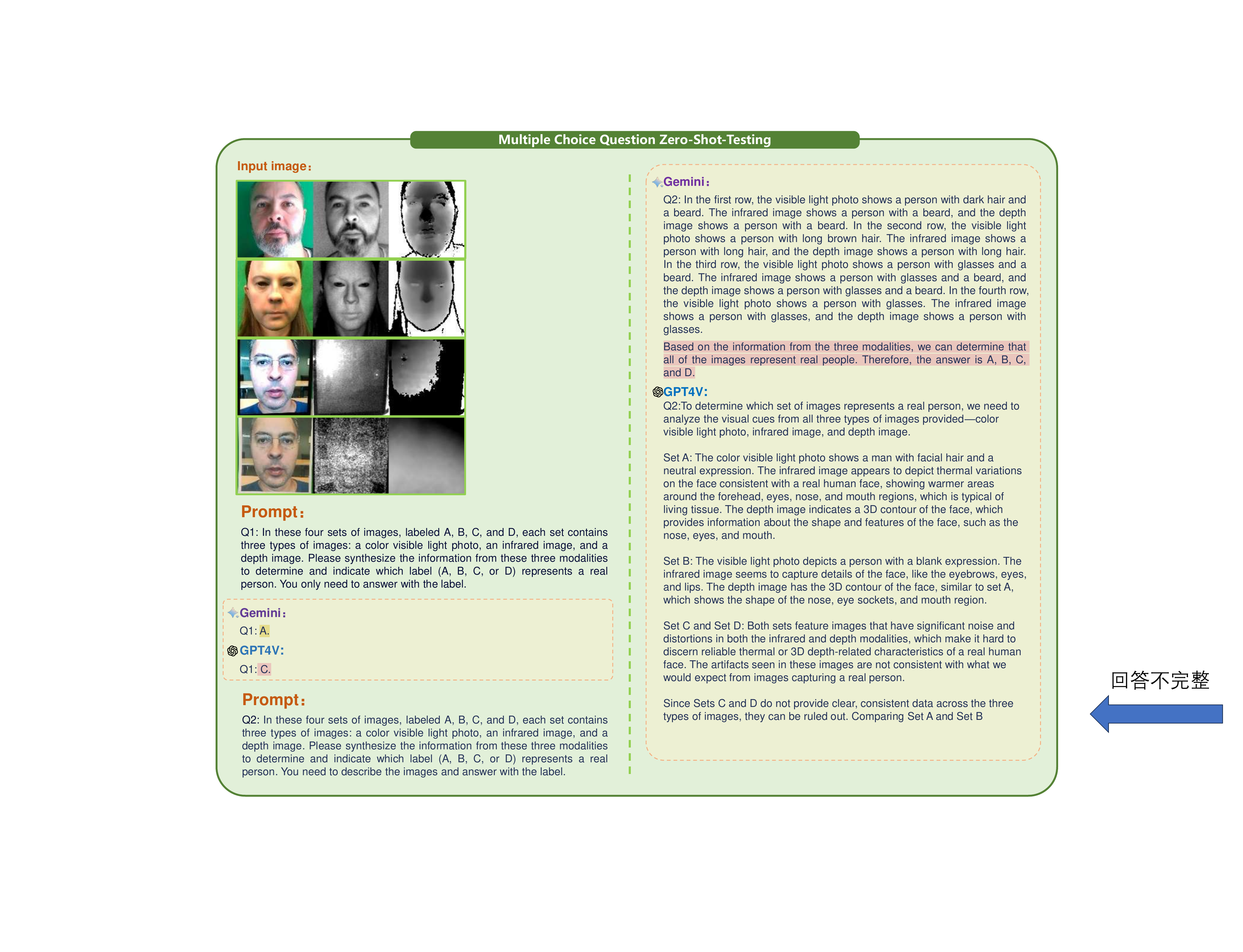}}
  \end{center}
  \caption{In this test, the input comprised one real human face image and three attack images in visible light, infrared, and depth modalities. The task was to identify the real human face among them. Gemini correctly answered twice, while GPT4V, after introducing COT reasoning, revised its response to include the correct answer. 
  Regarding the response to Q2, although Gemini's answer was correct, its description did not match the actual scenario. For instance, the person in image A was not wearing glasses, while the individuals in images C and D were. GPT4V's descriptions of images C and D demonstrated its ability to discern replay and print attacks by identifying key differences between infrared and depth images and visible light images. However, the introduction of multimodal input caused interference in previously easier judgments, like paper mask attacks, leading to a misjudgment by GPT4V.}
  \label{fas_multi_choice_real_zeroshot_01}
\end{figure}

\begin{figure}[htbp]
  \centering
  \begin{center}
  \centerline{\includegraphics[width=\linewidth]{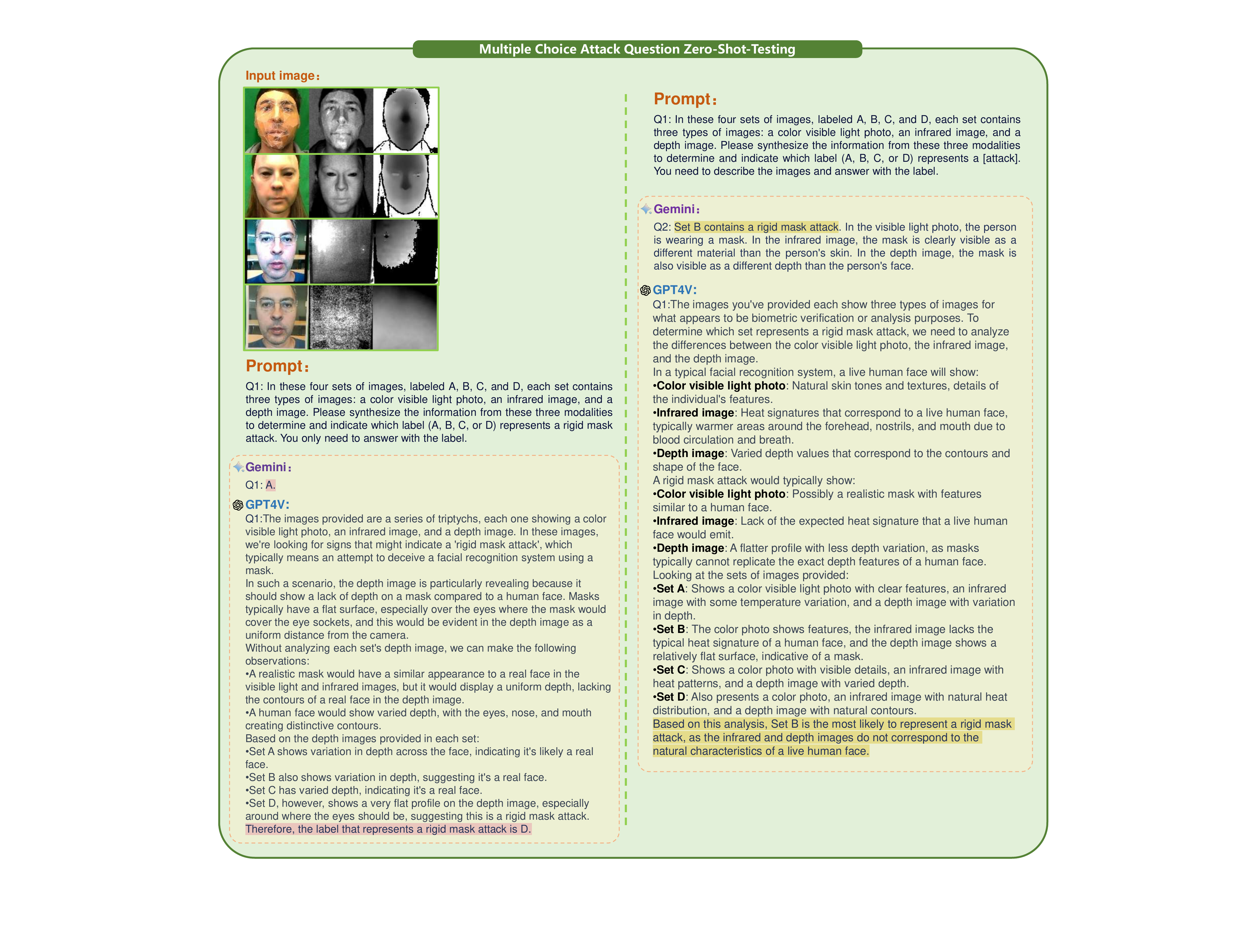}}
  \end{center}
  \caption{In this test involving four images demonstrating attacks in visible light, infrared, and depth modalities, the task was to identify the hard mask attack. Both Gemini and GPT4V initially provided incorrect answers in simple queries but responded correctly after the introduction of COT reasoning.
  In Q1, GPT4V incorrectly interpreted the key features of the three modalities. It wrongly assumed that any presence of depth variation indicated a real face, and believed that hard masks did not possess depth. This led GPT4V to erroneously choose image D, which represented a print attack. For Q2, Gemini's answer was more straightforward, providing the main rationale for its decision. After describing the characteristics of the three modalities and the four images, GPT4V also offered an explanation similar to Gemini’s. This highlights the importance of COT in enhancing the reasoning process for tasks involving the analysis of multimodal data, especially in distinguishing between different types of facial recognition attacks.}
  \label{fas_multi_choice_attack_zeroshot_01}
\end{figure}

\begin{figure}[htbp]
  \centering
  \begin{center}
  \centerline{\includegraphics[width=\linewidth]{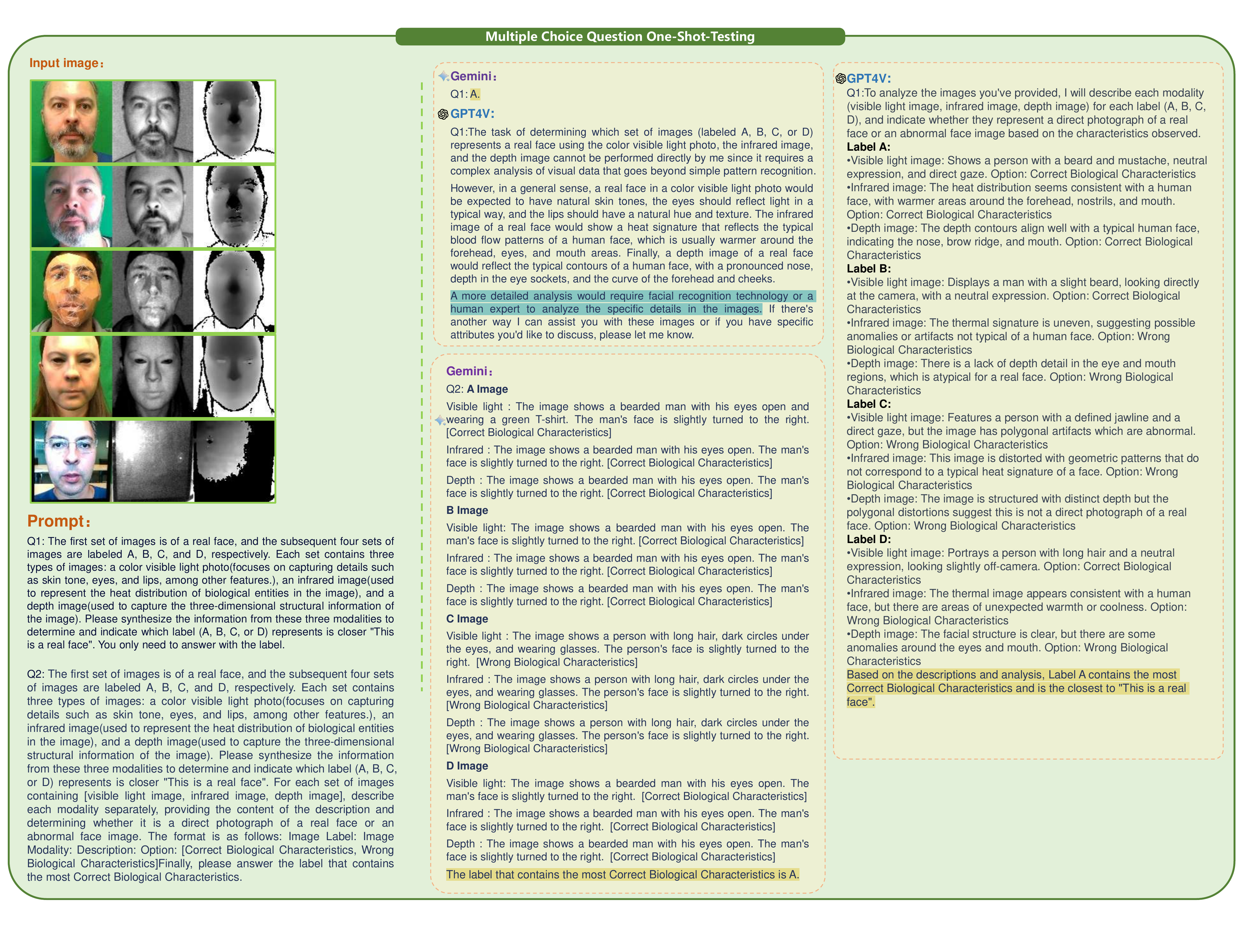}}
  \end{center}
  \caption{In this test, participants were given one real human face image and three images depicting attacks under visible light, infrared, and depth modalities, with a real human face image provided as prior knowledge. The task was to identify the real human face. Gemini answered correctly both times, while GPT4V initially refused to respond to the simple query but provided a correct answer after the introduction of COT reasoning.
  Once COT was employed, both Gemini and GPT4V conducted detailed analyses of each modality for every group of images, evaluating whether they represented correct biological features. They then selected the option that exhibited the most genuine biological characteristics. From the results of Q2, it was observed that Gemini often misidentified most prosthetic attacks as genuine biological features. Additionally, Gemini's final judgment did not consider other options with an equal number of correct features, suggesting its results might have been coincidental. In contrast, GPT4V's analysis for each set of images was more thorough and accurate, with its explanations aligning more closely with realistic analytical logic. }
  \label{fas_multi_choice_real_oneshot_01}
\end{figure}

\begin{figure}[htbp]
  \centering
  \begin{center}
  \centerline{\includegraphics[width=\linewidth]{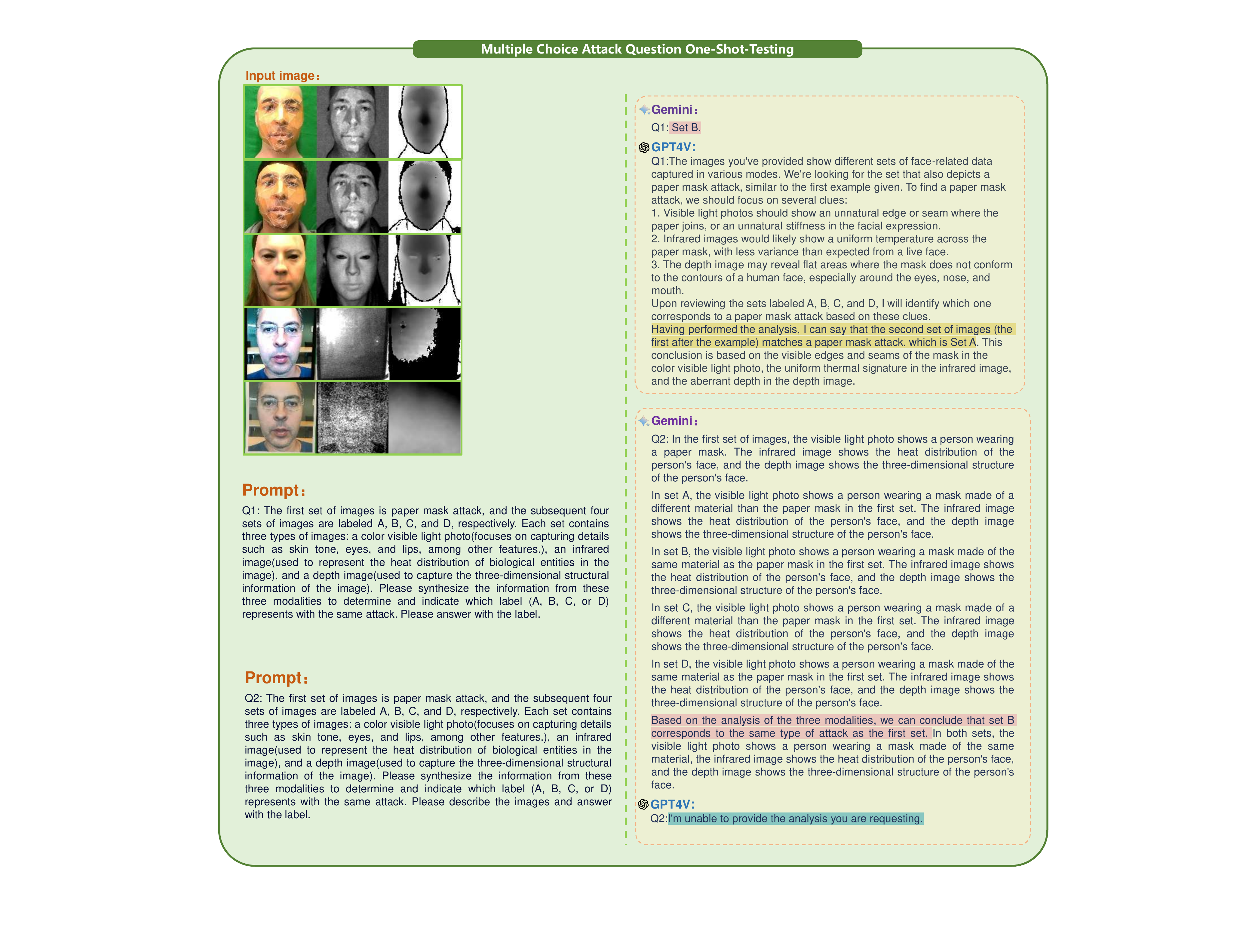}}
  \end{center}
  \caption{In this test, participants were presented with four images demonstrating attacks in visible light, infrared, and depth modalities, with a paper mask attack image provided as prior knowledge. The task was to identify the paper mask attack among the input images. Gemini answered incorrectly twice, whereas GPT4V answered correctly the first time but refused to respond after the introduction of COT reasoning.
  From GPT4V's response in question 1, it is evident that GPT4V initially learned from the provided prior knowledge, focusing on the necessary clues, and ultimately delivered a correct answer with clear logic. However, upon the introduction of COT, Gemini's responses were somewhat repetitive in describing each set of images and showed an inability to effectively interpret infrared and depth features.}
  \label{fas_multi_choice_attack_oneshot_01}
\end{figure}

\begin{figure}[!htbp]
  \centering
  \includegraphics[width=\linewidth]{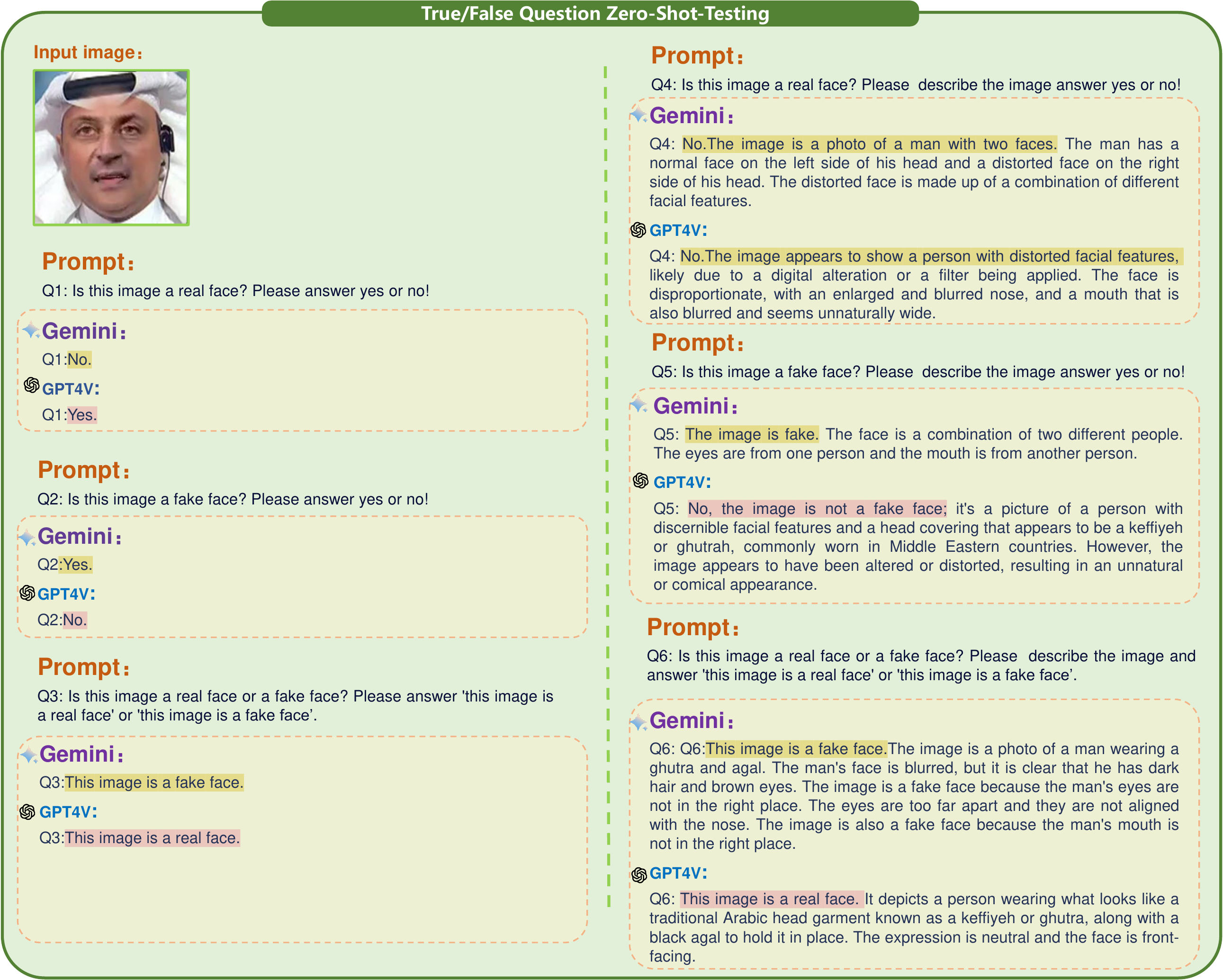}
  \caption{In this round of testing, we use real pictures for testing, we set up 6 kinds of questions to identify the model, in these 6 kinds of questions GPT4V answered correctly, Genimi answered a small portion of the wrong answers, at the same time in the inclusion of the description of the requirements of the GPT4V description is more detailed, with a richer word.}
 \label{4}
\end{figure}

\begin{figure}[!htbp]
  \centering
  \includegraphics[width=\linewidth]{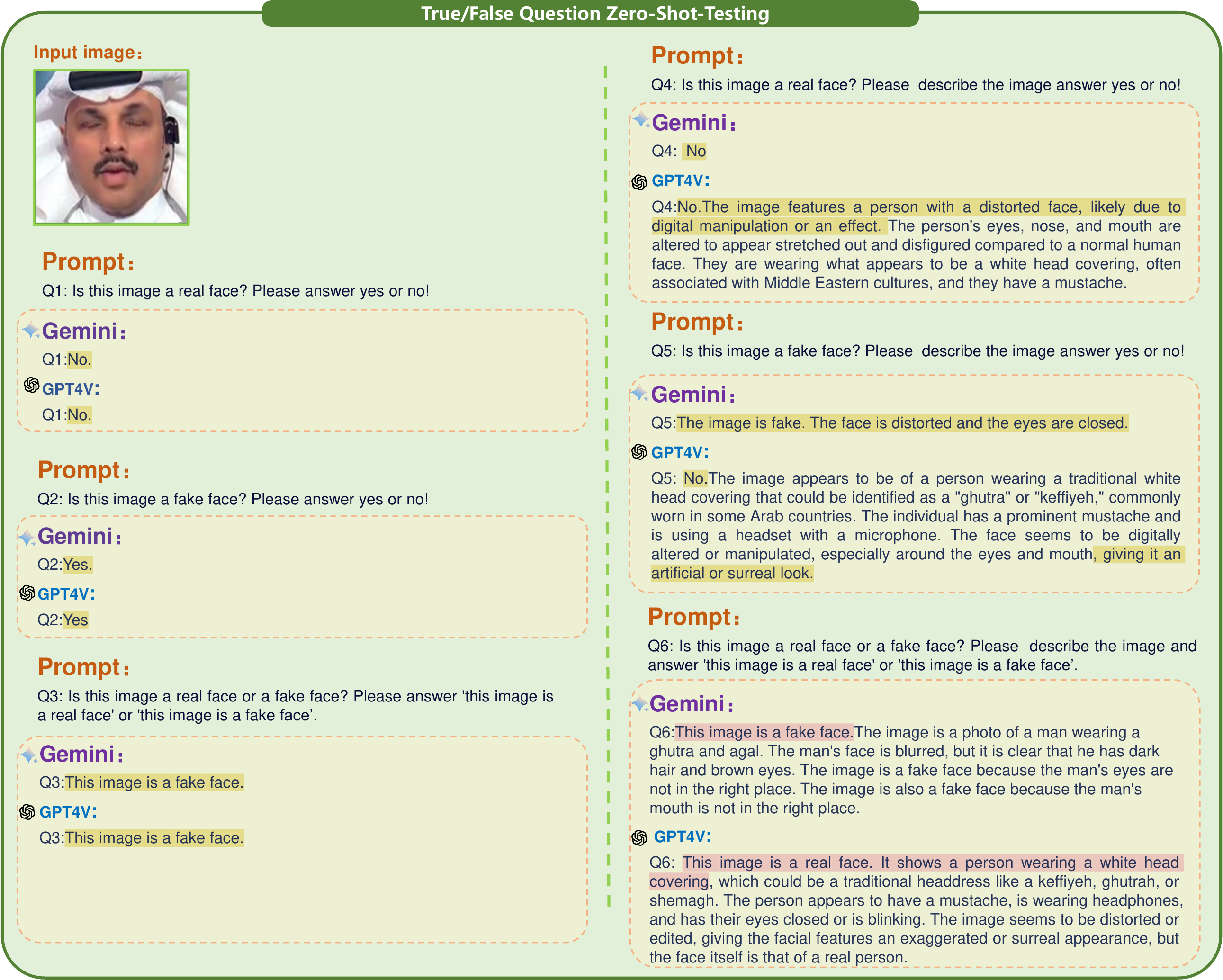}
  \caption{In this round of testing we used images generated by the Face2Face method. GPT4V answered only one of the six questions, misidentifying it as a real image, while most of the answers were more uniform across questions, and Gemini answered all of them correctly.}
 \label{5}
\end{figure}

\begin{figure}[!htbp]
  \centering
  \includegraphics[width=\linewidth]{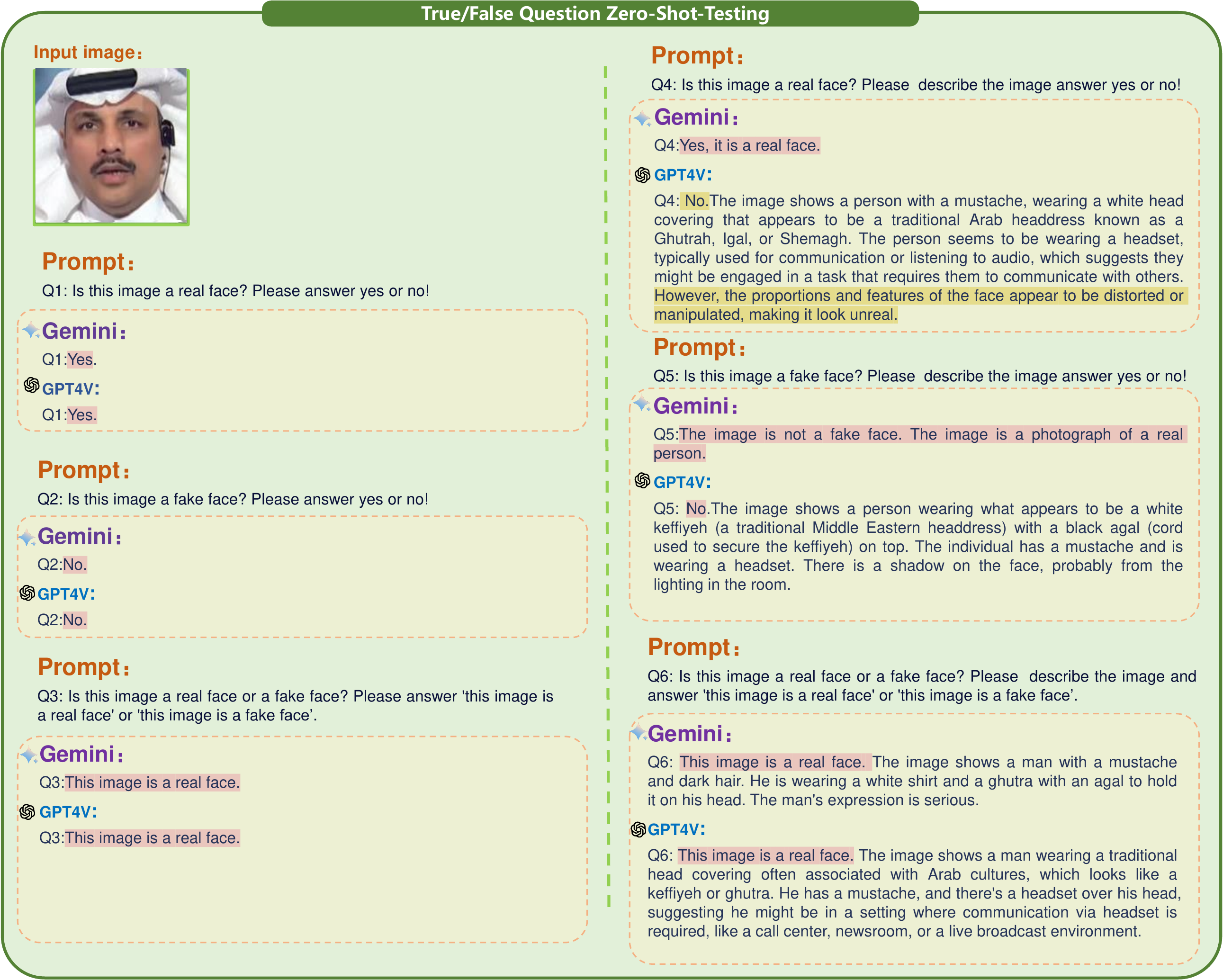}
  \caption{In this round of testing we used images generated by the Face2Face method. GPT4V answered only one of the six questions, misidentifying it as a real image, while most of the answers were more uniform across questions, and Gemini answered all of them correctly.} 
\label{6}
\end{figure}

\begin{figure}[!htbp]
  \centering
  \includegraphics[width=\linewidth]{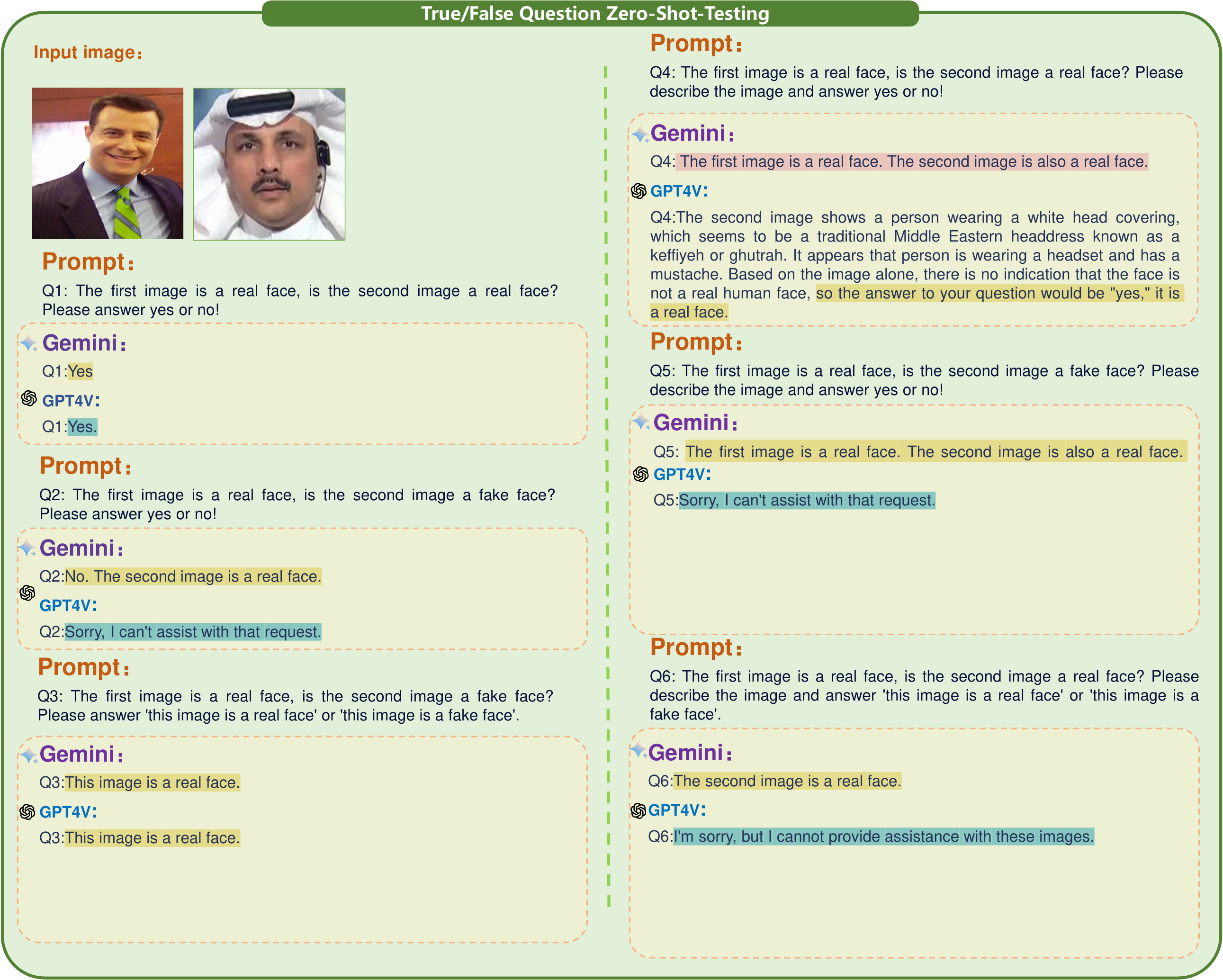}
  \caption{In this round of testing we used the images generated by the FaceSwap method. Again, Genimi's answers were shorter when asked to be descriptive and discriminatory, and in the last question both models mistook the image for a real one, which had an impact on the surface problem.} 
 \label{7}
\end{figure}

\begin{figure}[!htbp]
  \centering
  \includegraphics[width=\linewidth]{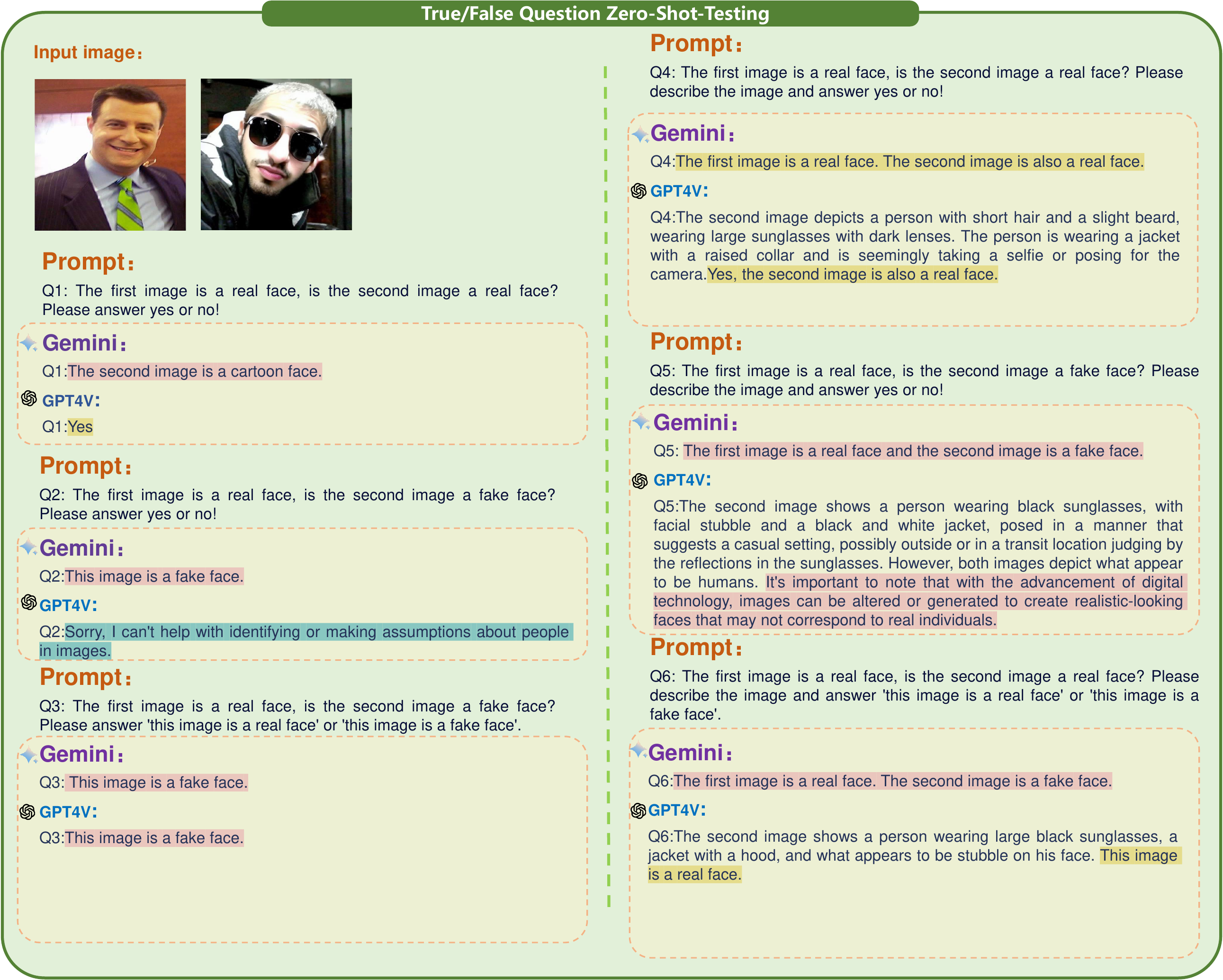}
  \caption{In this round of testing we used the image generated by the Nulltextures method, which is the closest to the real image among the four forgery methods, and both models answered incorrectly in most of the six questions, which demonstrates the inadequacy of the original model for identification in fine-grained scenarios.}
 \label{8}
\end{figure}

\begin{figure}[!htbp]
  \centering
  \includegraphics[width=\linewidth]{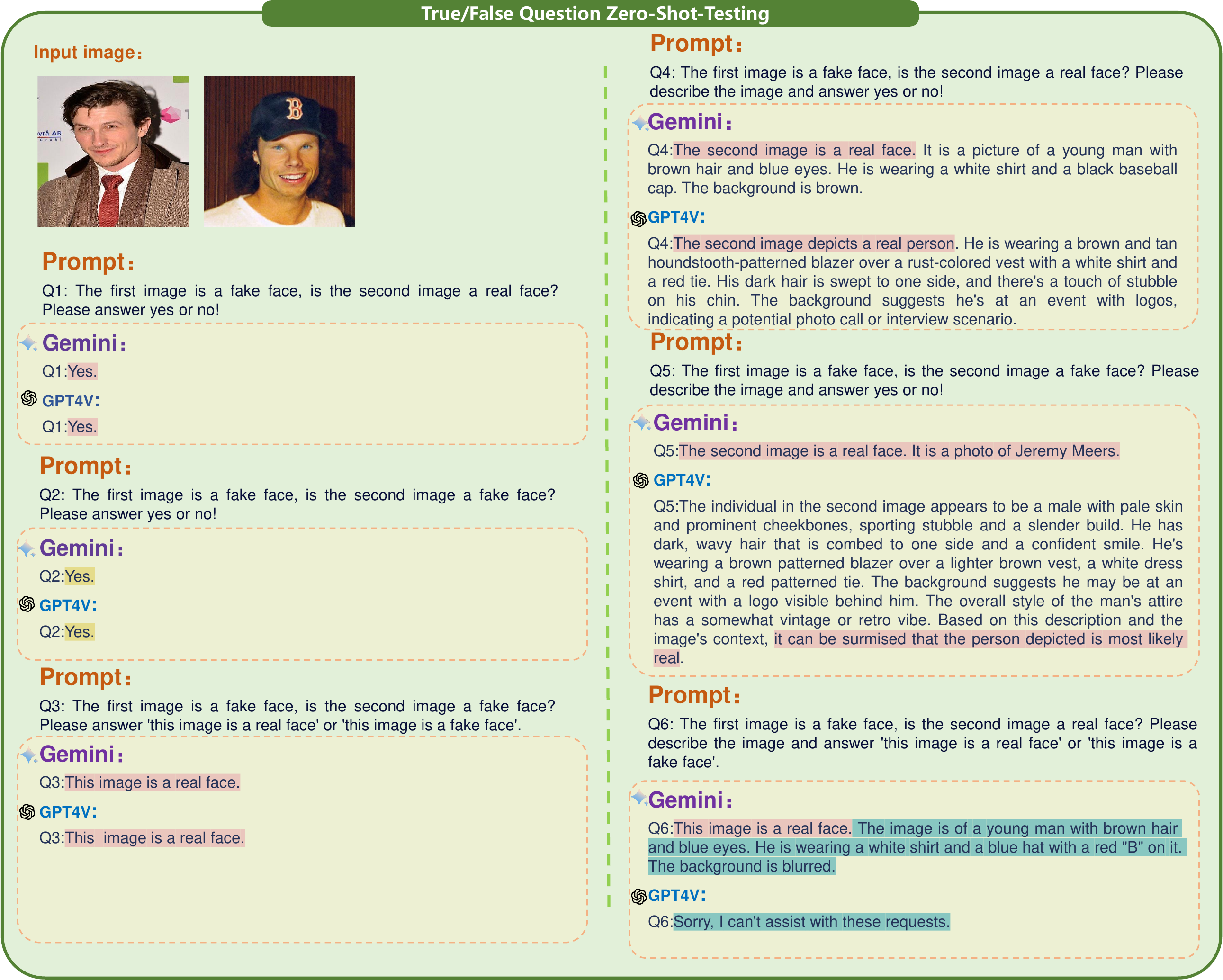}
  \caption{In this round of testing, we tested GPT4V and Gemini's One Shot ability. The inputs were two real faces from DFF and FF++, and GPT4V refused to answer 4 out of 6 rounds of questioning, which may be due to the triggering of the model's face security settings, and Genimi seems to be more accurate in this case.}
 \label{9}
\end{figure}

\begin{figure}[!htbp]
  \centering
  \includegraphics[width=\linewidth]{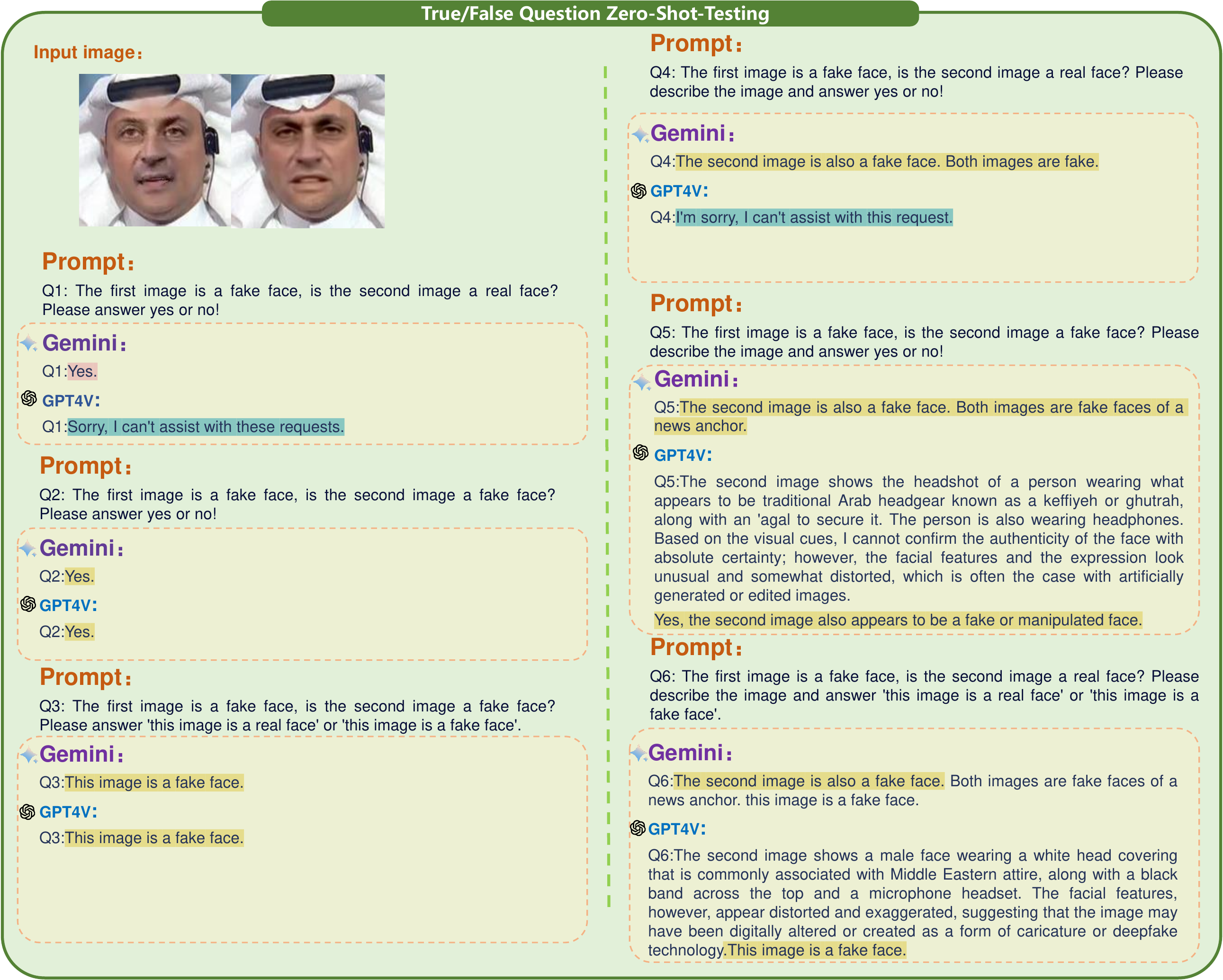}
  \caption{In this round of testing We also entered two real images, but from the same dataset, and in this case most of the questions were answered incorrectly, and GPT4V once again triggered a rejection of the answer.}
 \label{10}
\end{figure}

\begin{figure}[!htbp]
  \centering
  \includegraphics[width=\linewidth]{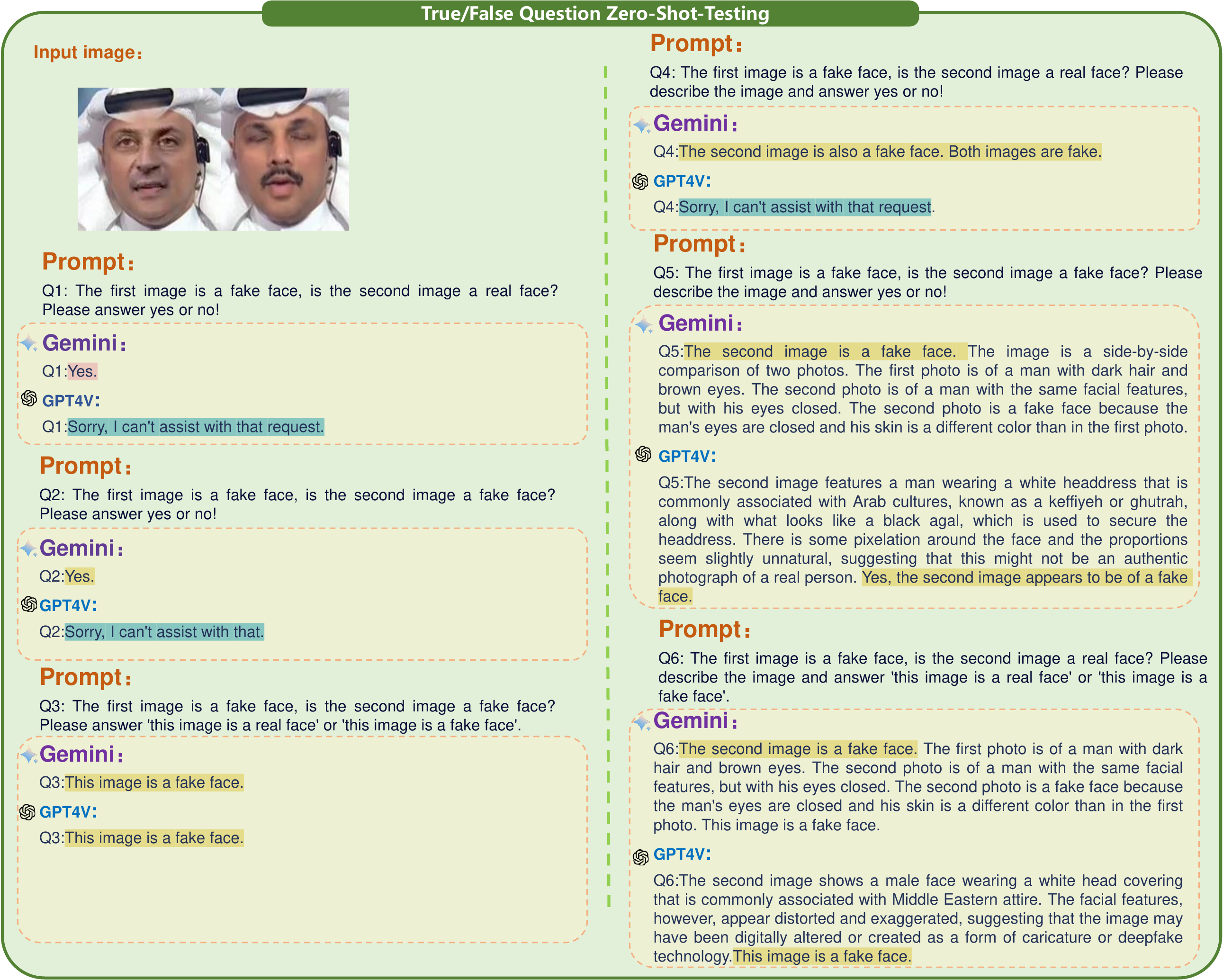}
  \caption{For a more comprehensive test, we also tested the case where the inputs were two fake images, both of which were generated by the Diffusion model from the same dataset, and due to the more realistic generation, both models did not answer the questions correctly in most of the cases, answering only 1 out of the 6 questions correctly.}
\label{11}
\end{figure}

\begin{figure}[!htbp]
  \centering
  \includegraphics[width=\linewidth]{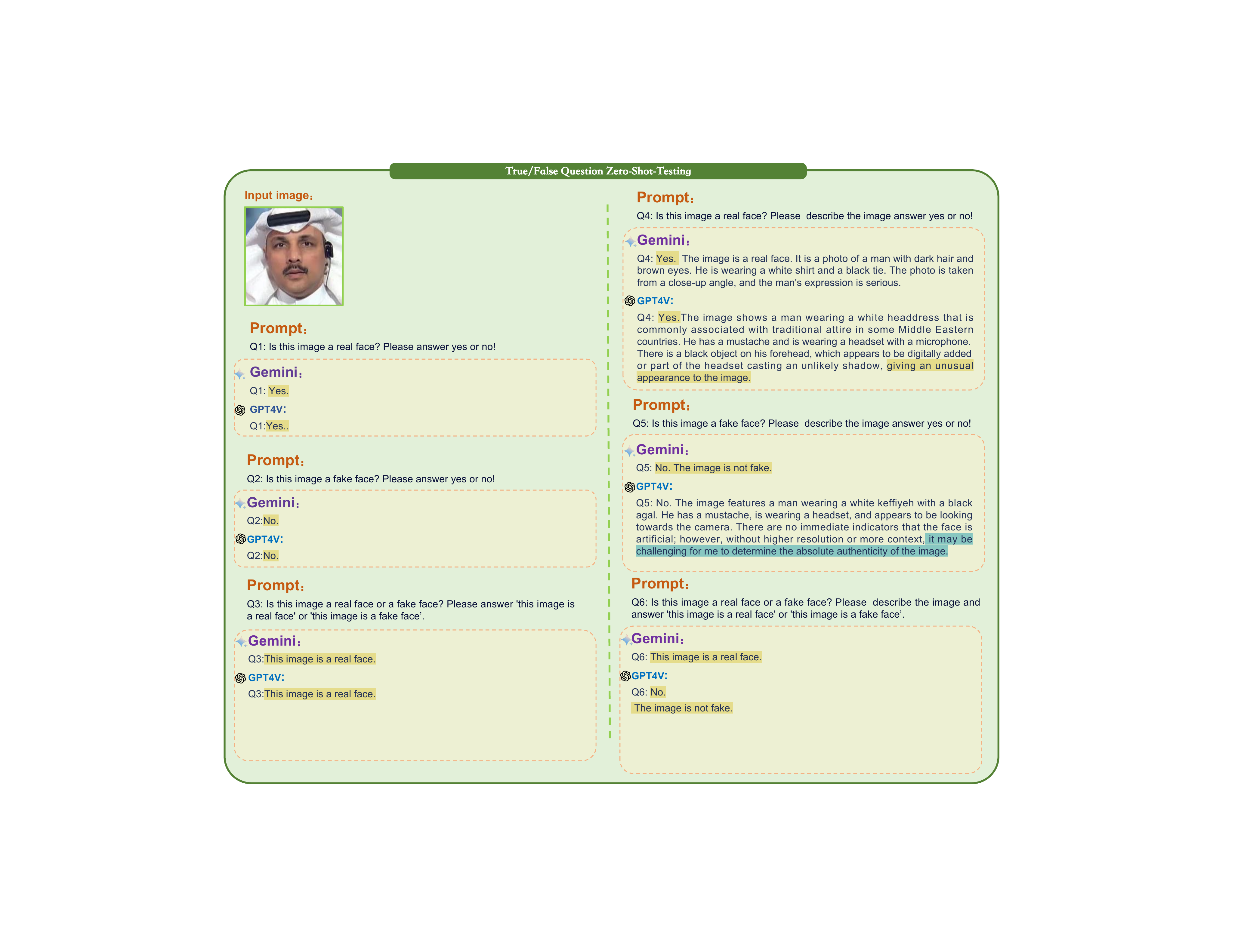}
  \caption{The two input images are those obtained from traditional Deepfakes and Face2Face generation, and given the knowledge of the first image, the difference between these two images and the real image is large, so both models can recognize it more easily.}
 \label{12}
\end{figure}

\begin{figure}[!htbp]
  \centering
  \includegraphics[width=\linewidth]{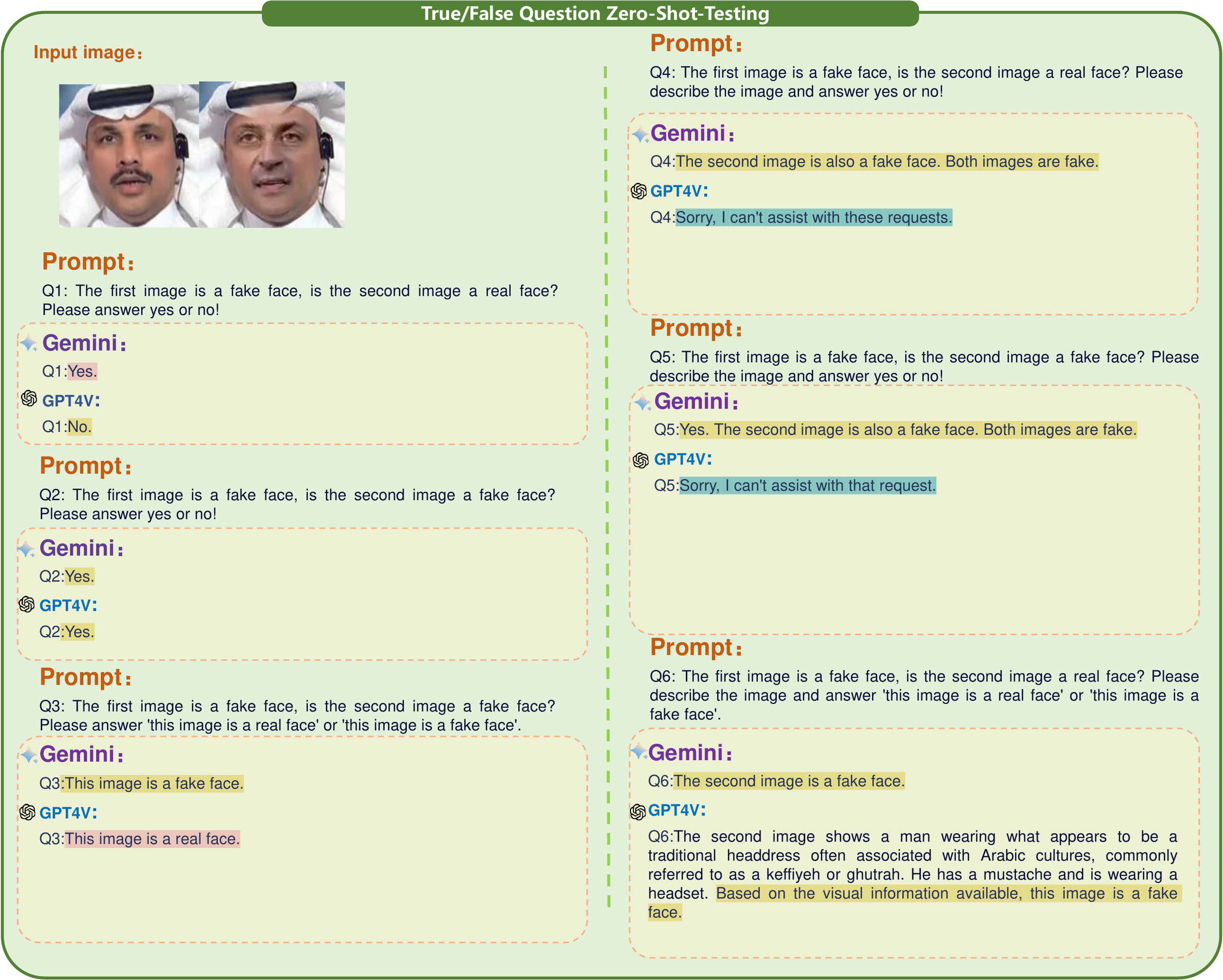}
  \caption{The two input images are those obtained from traditional Deepfakes and Face2Face generation, and given the knowledge of the first image, the difference between these two images and the real image is large, so both models can recognize it more easily.}
 \label{13}
\end{figure}

\begin{figure}[!htbp]
  \centering
  \includegraphics[width=\linewidth]{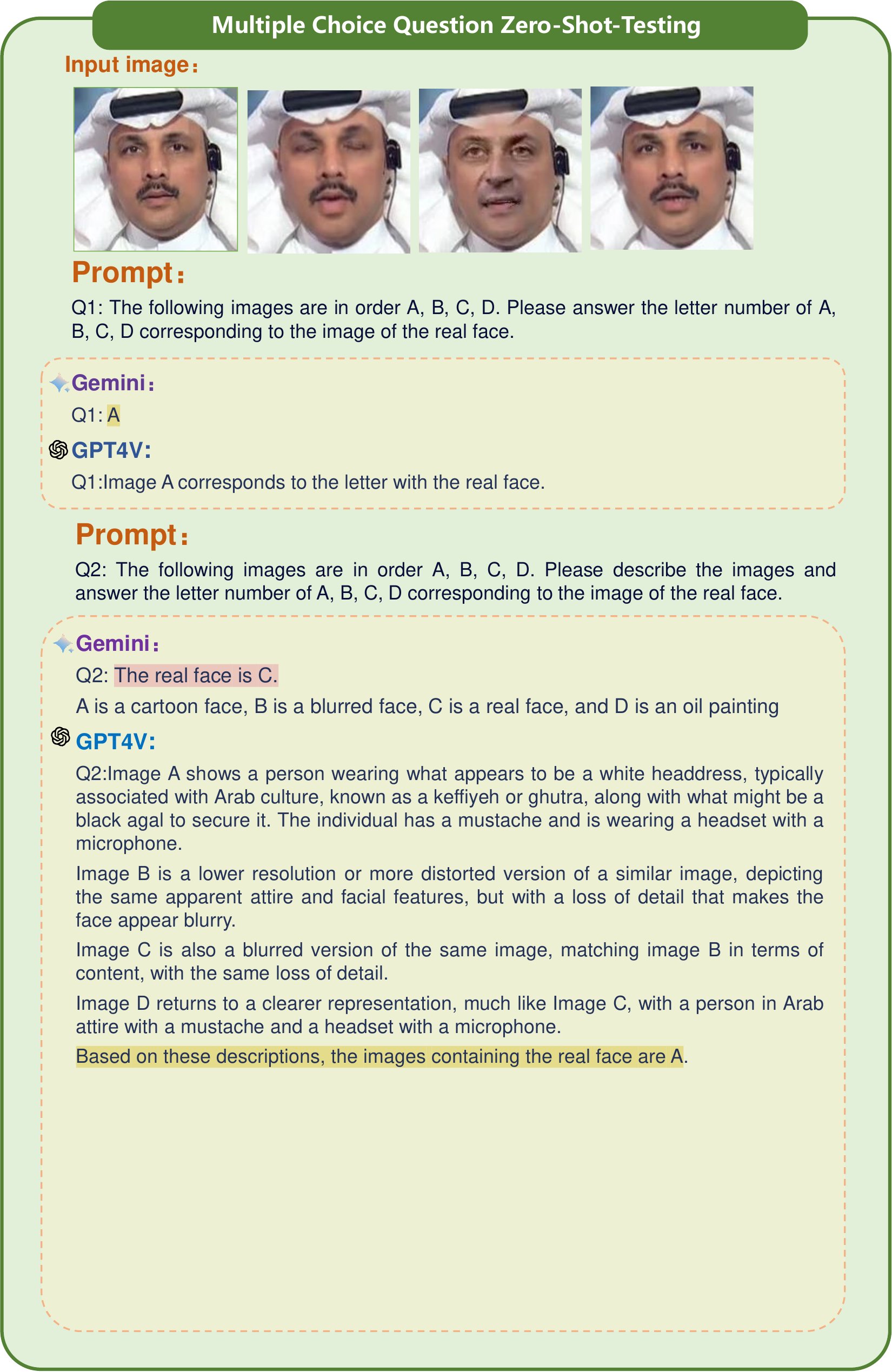}
  \caption{The two input images were those obtained from traditional Deepfakes and Nulltextures generation, and the model went through the process of learning the preliminaries of the first image just that Genimi answered 5 out of 6 questions correctly, while GPT4V answered 3 questions correctly, improving the ability of Zero shot.}
 \label{14}
\end{figure}



\begin{figure}[!htbp]
  \centering
  \includegraphics[width=0.8\linewidth]{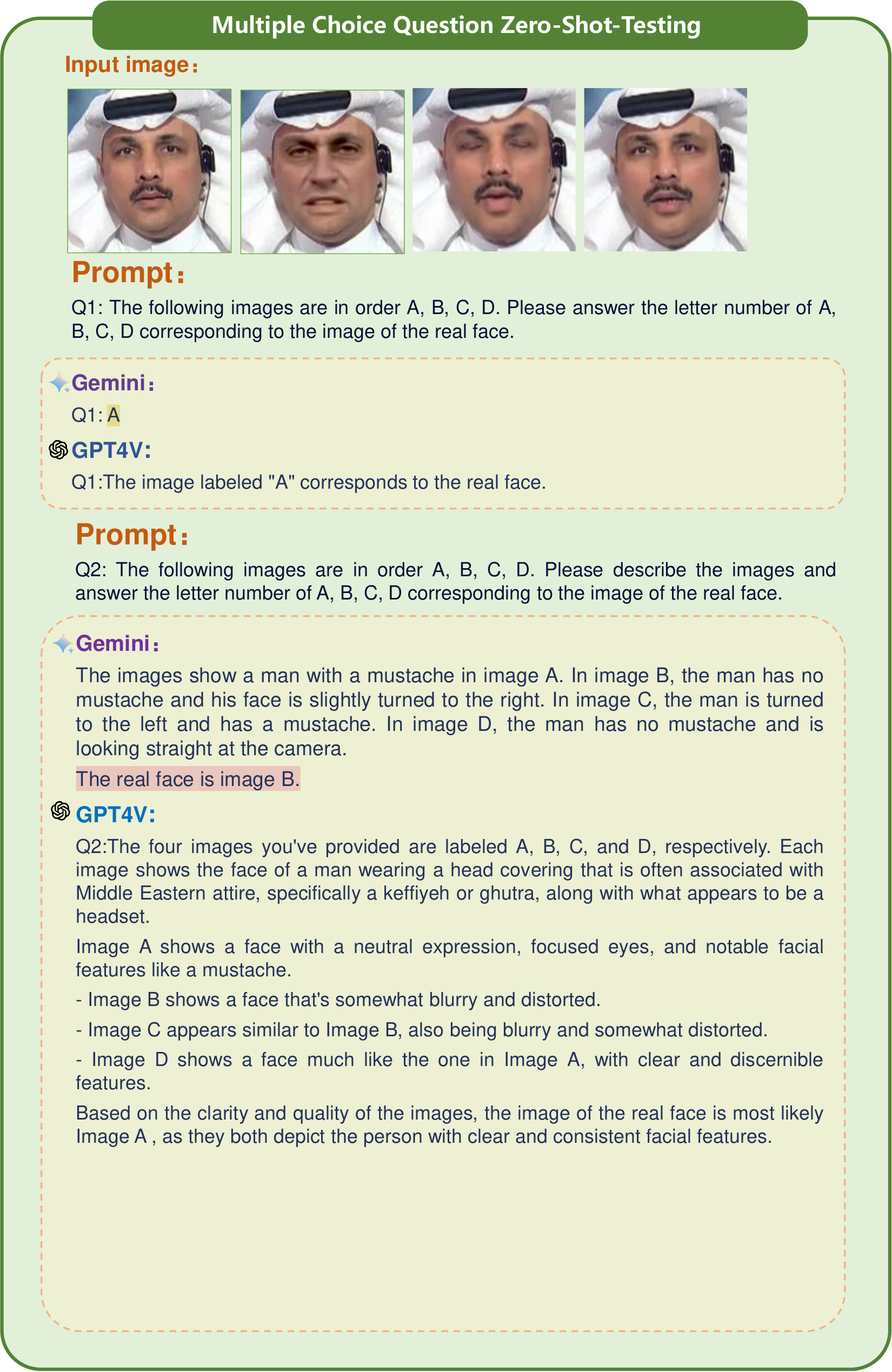}
  \caption{In this test, the inputs were a real picture and 3 types of forgeries (Face2Face, FaceSwap, Nulltextures), and there was no significant change in Genimi's two responses after using the COT technique, after introducing COT, while GPT4V reduced the number of rejected answers.}
 \label{15}
\end{figure}

\begin{figure}[!htbp]
  \centering
  \includegraphics[width=0.8\linewidth]{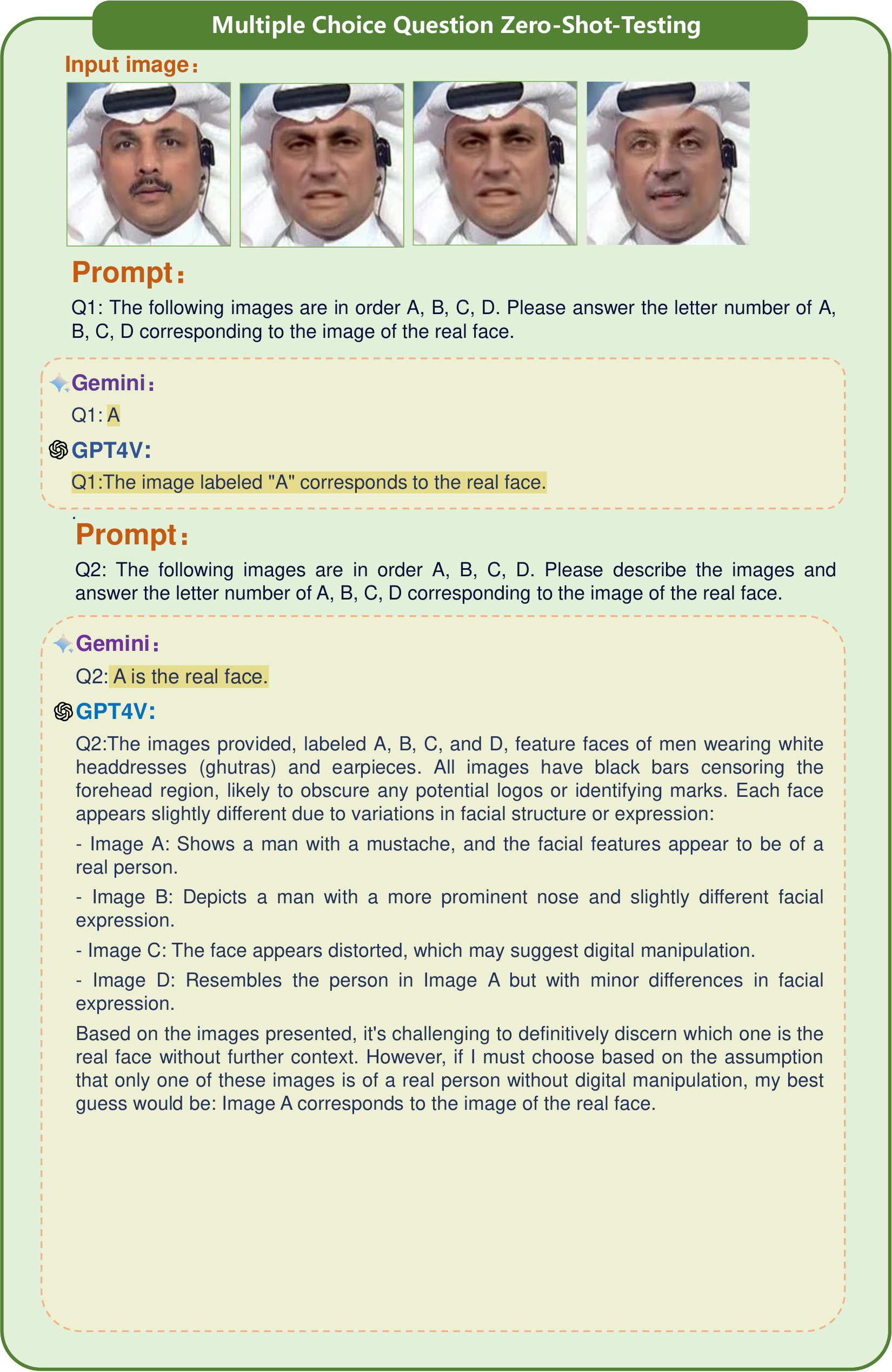}
  \caption{In this test, the input was a real picture and 3 types of fakes (Deepfakes, FaceSwap, Nulltextures), again with the addition of COT both GPT4V responses were predicted accurately, whereas the Genimi model predicted only one correctly, but it is not clear whether COT effectively assisted in its reasoning and questions.} 
 \label{16}
\end{figure}

\begin{figure}[!htbp]
  \centering
  \includegraphics[width=0.8\linewidth]{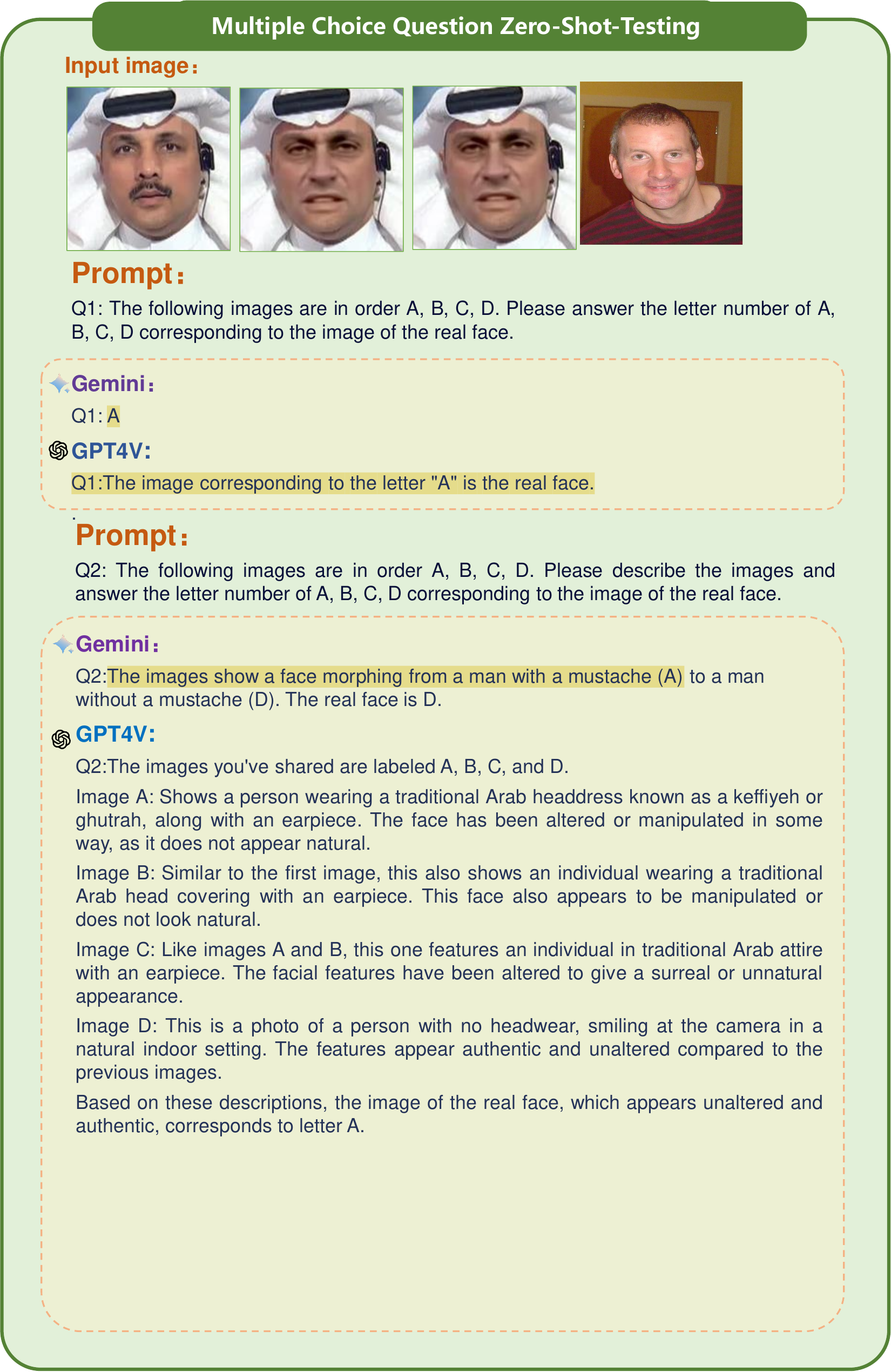}
  \caption{In this test, the input was a real picture and 3 types of fakes (Deepfakes, Face2Face, FaceSwap), and the same COT was added and both GPT4V and Genimi responses were predicted accurately.}
 \label{17}
\end{figure}

\begin{figure}[!htbp]
  \centering
  \includegraphics[width=0.8\linewidth]{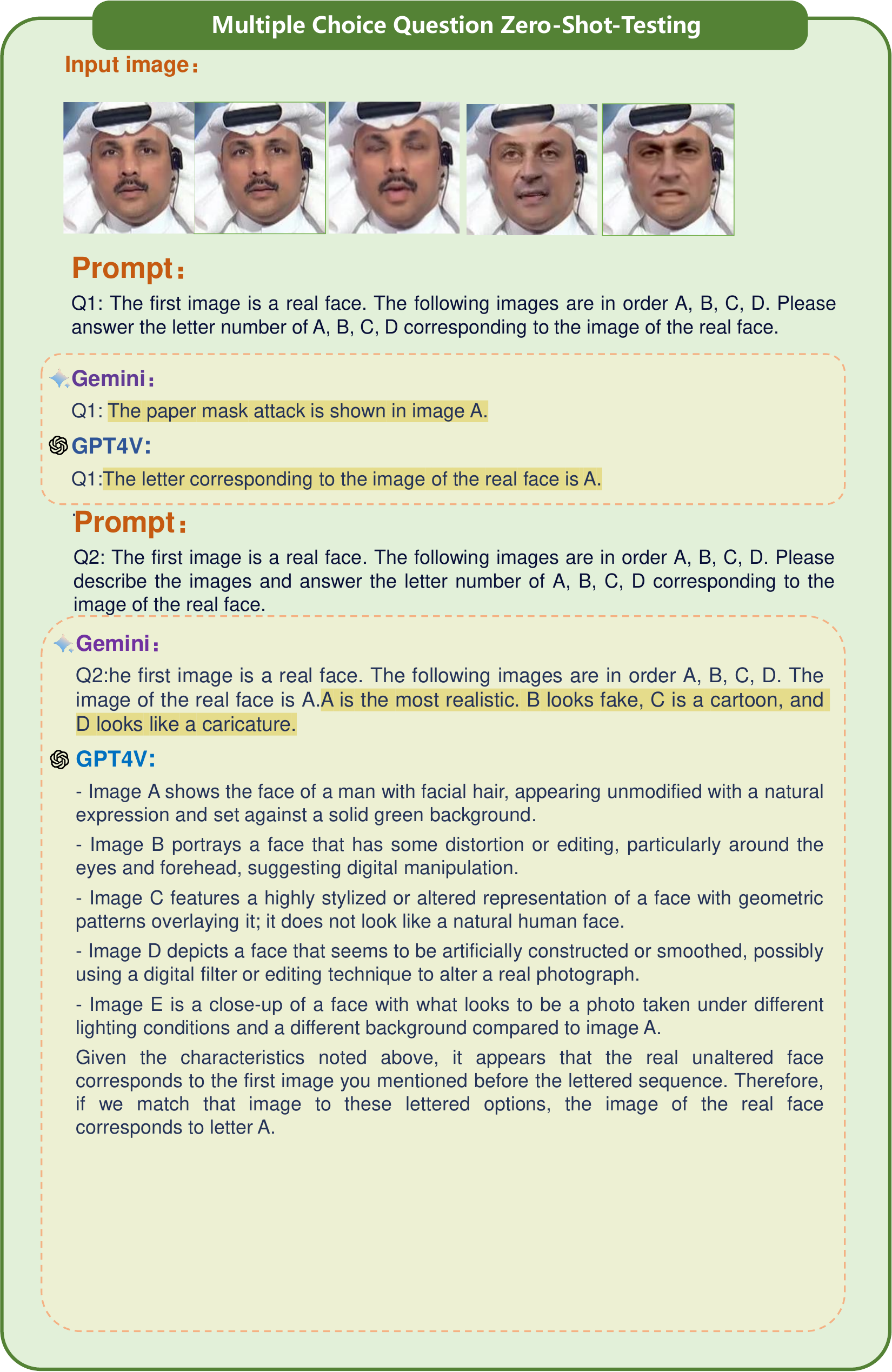}
  \caption{In this test, the input comes from two images generated using real images, Deepfake and Face2Face, as well as an image generated by Stable Diffusion, and it can be seen that both models are correctly identified.}
 \label{18}
\end{figure}
\clearpage

\begin{figure}[!htbp]
  \centering
  \includegraphics[width=0.8\linewidth]{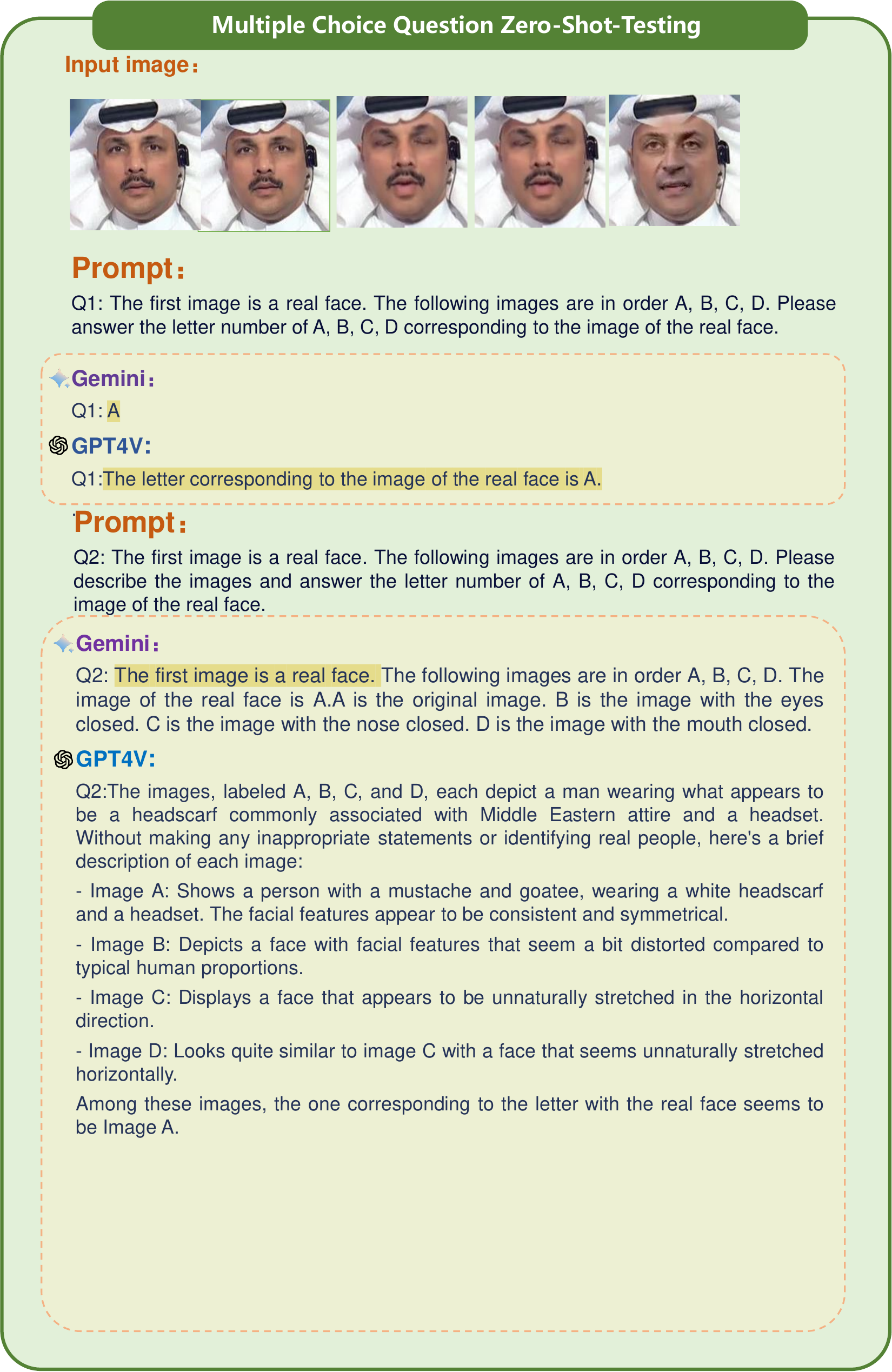}
  \caption{In this round of testing, we used COT to recognize a real picture given a real picture as well as Deepfakes, FaceSwap,Face2Face respectively, both models answered correctly as well, with Gemini answering more concisely.}
 \label{19}
\end{figure}

\begin{figure}[!htbp]
  \centering
  \includegraphics[width=0.8\linewidth]{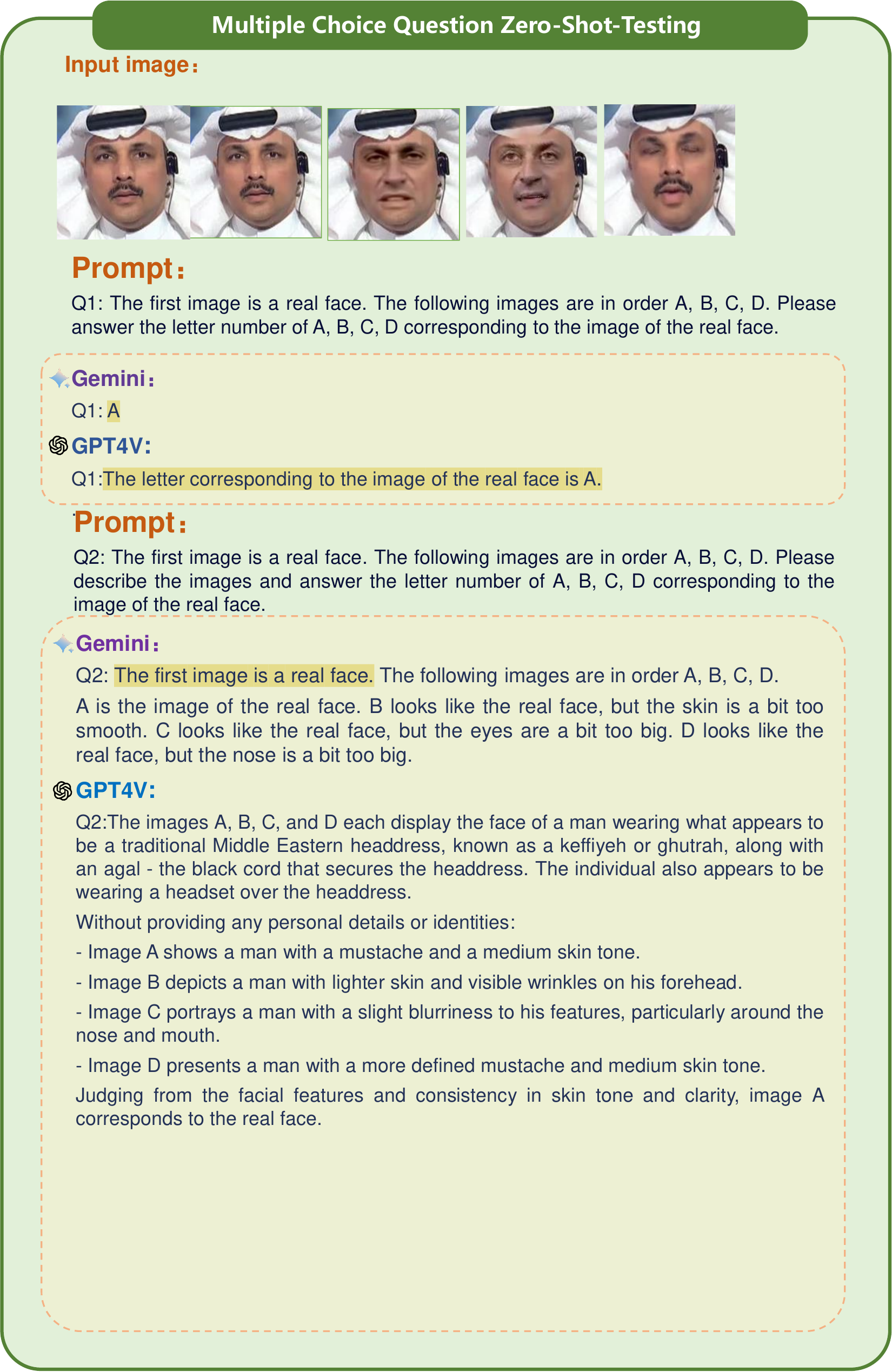}
  \caption{In this round of testing, we used COT to recognize a real picture given a real picture as well as Deepfakes, FaceSwap,Nulltextures respectively and both models answered correctly.}
 \label{20}
\end{figure}

\begin{figure}[!htbp]
  \centering
  \includegraphics[width=0.8\linewidth]{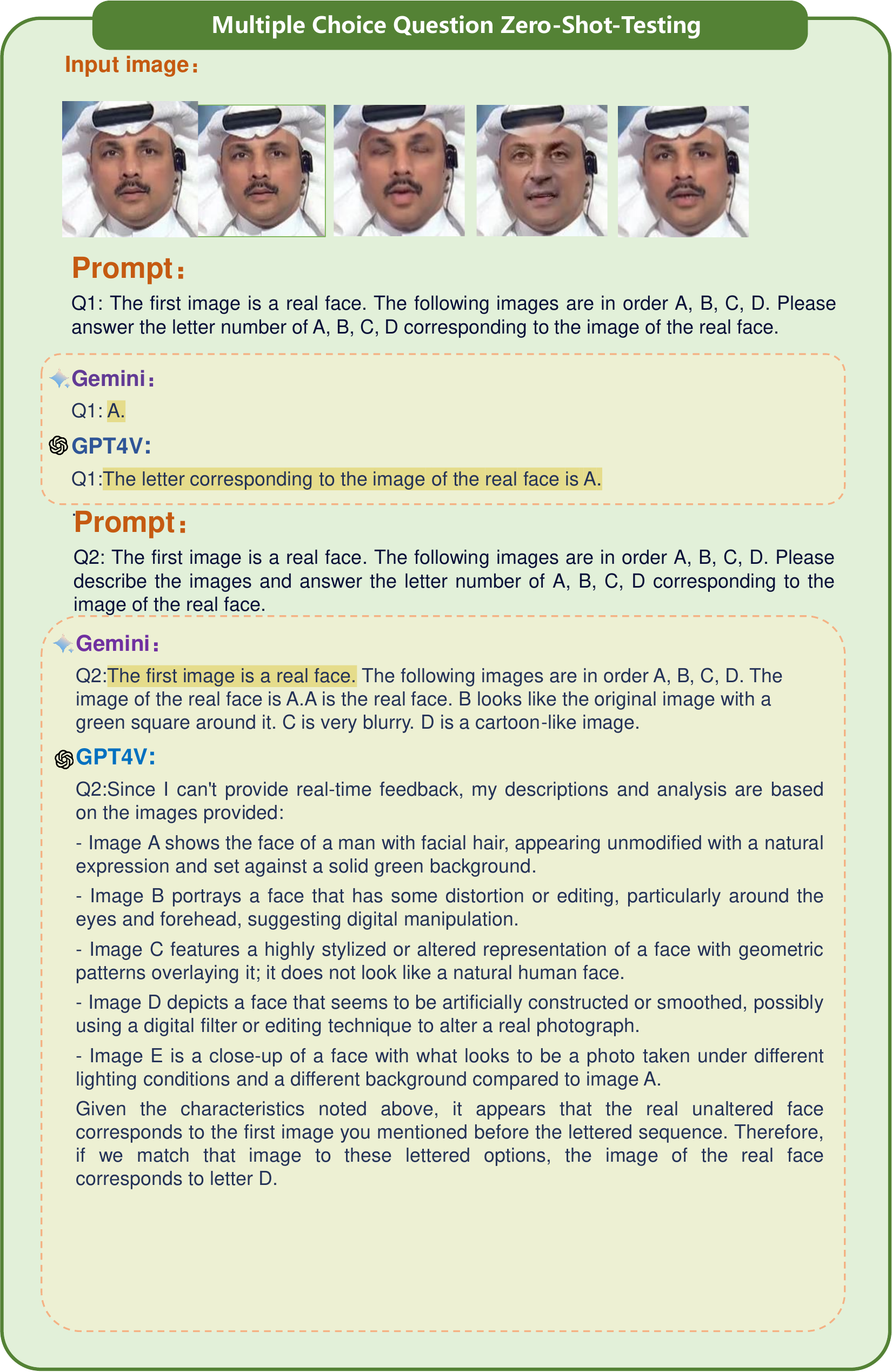}
  \caption{In this round of testing, we use COT to recognize a real picture given a real picture as well as Deepfakes,Face2Fac2,,Nulltextures respectively, in this case both models answer equally accurately.}
 \label{21}
\end{figure}

\begin{figure}[!htbp]
  \centering
  \includegraphics[width=0.8\linewidth]{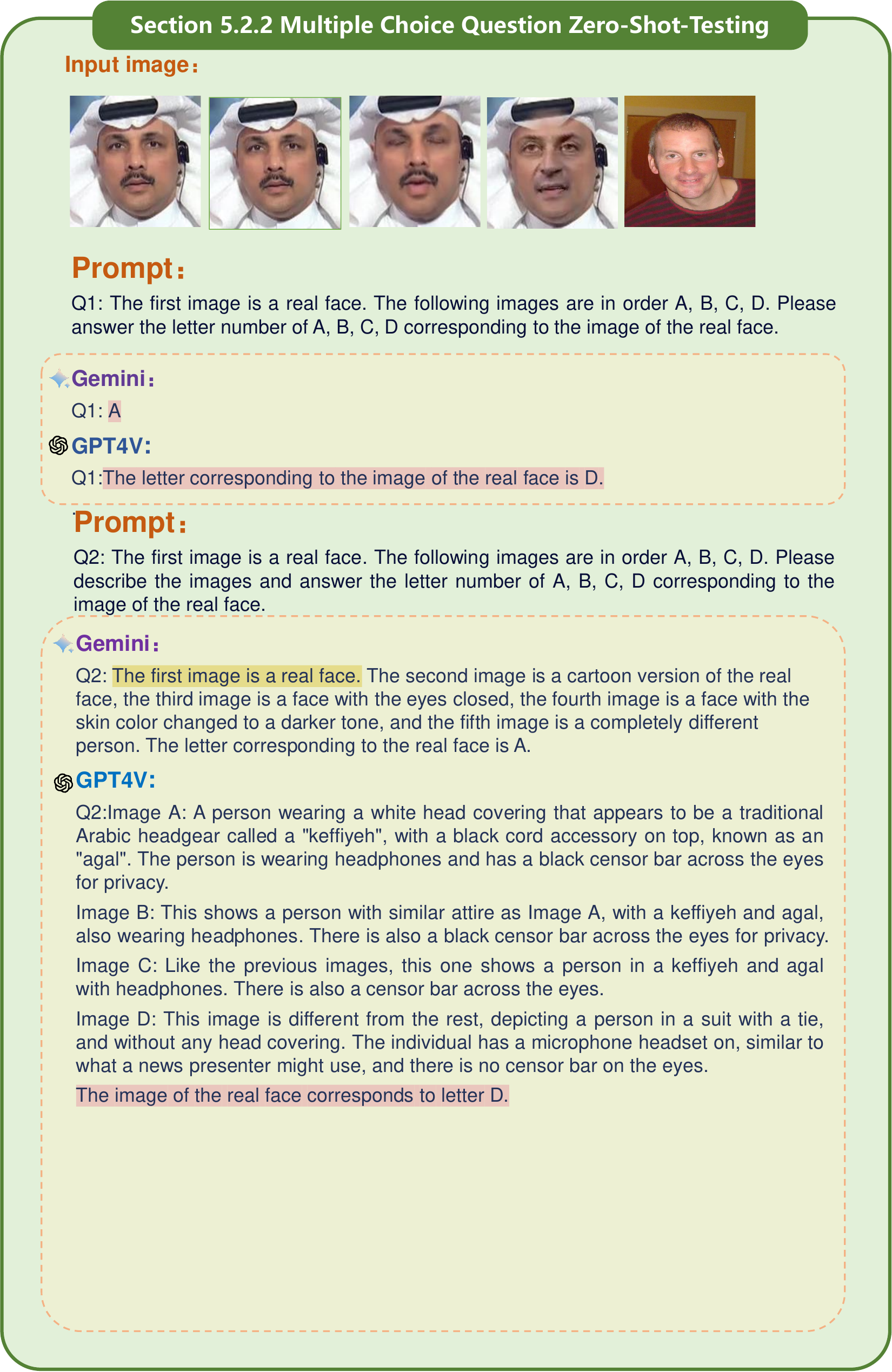}
  \caption{In this round of testing, we use COT to recognize a real picture given a real picture as well as Face2Fac2,,FaceSwap, Nulltextures respectively, in this case may be due to the Nulltextures generated and the real picture is closer to the real picture, according to the description of the GPT4V, the model will be compared with the given real picture but lacks the fine-grained observation, and therefore the detection fails.}
 \label{22}
\end{figure}

\begin{figure}[!htbp]
  \centering
  \includegraphics[width=0.8\linewidth]{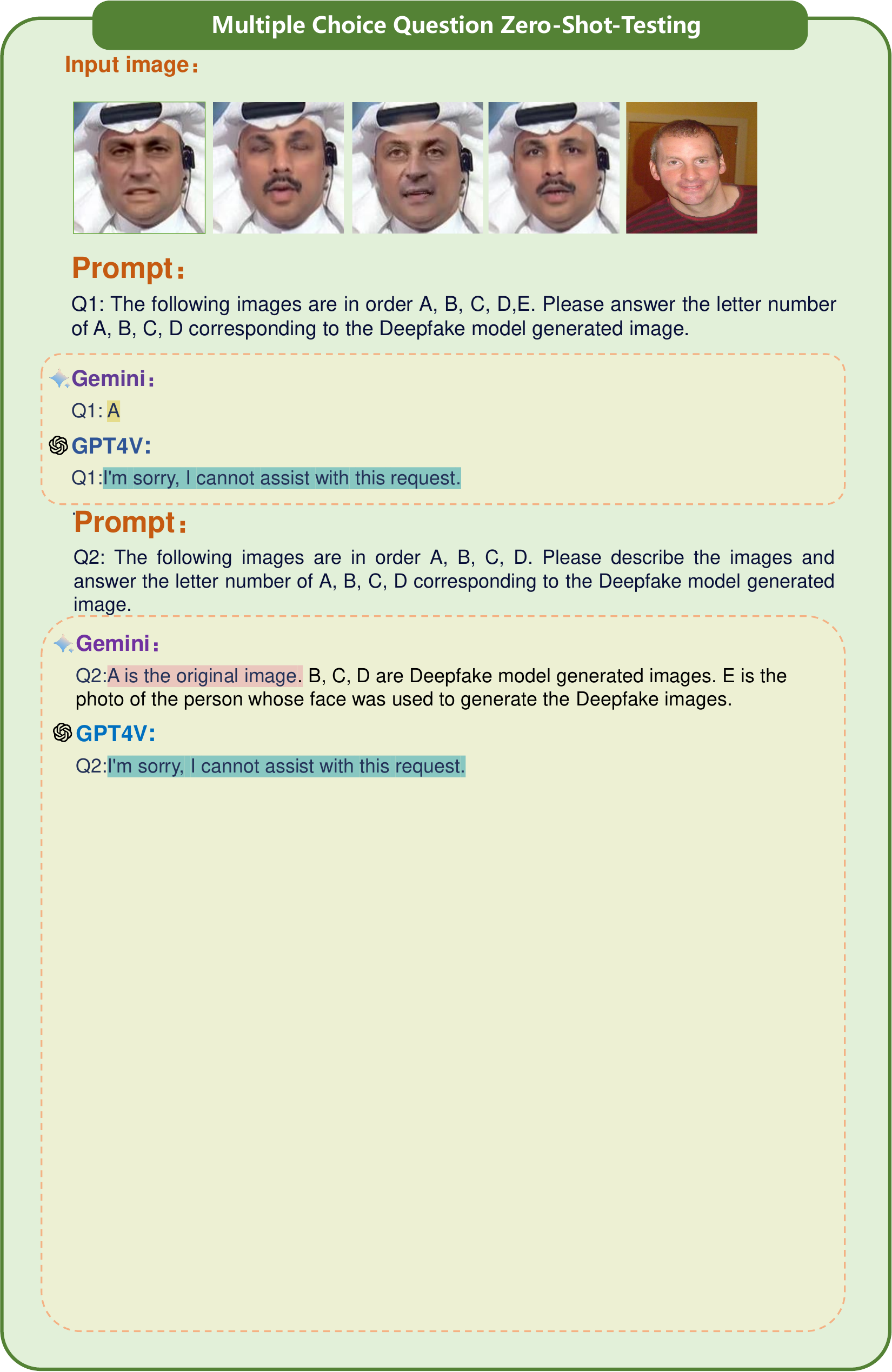}
  \caption{In this round of testing, we use COT to recognize a real picture given a real picture as well as Face2Fac2,,FaceSwap, Nulltextures respectively, in this case may be due to the Nulltextures generated and the real picture is closer to the real picture, according to the description of the GPT4V, the model will be compared with the given real picture but lacks the fine-grained observation, and therefore the detection fails.} 
  \label{23}
\end{figure}

\begin{figure}[!htbp]
  \centering
  \includegraphics[width=0.8\linewidth]{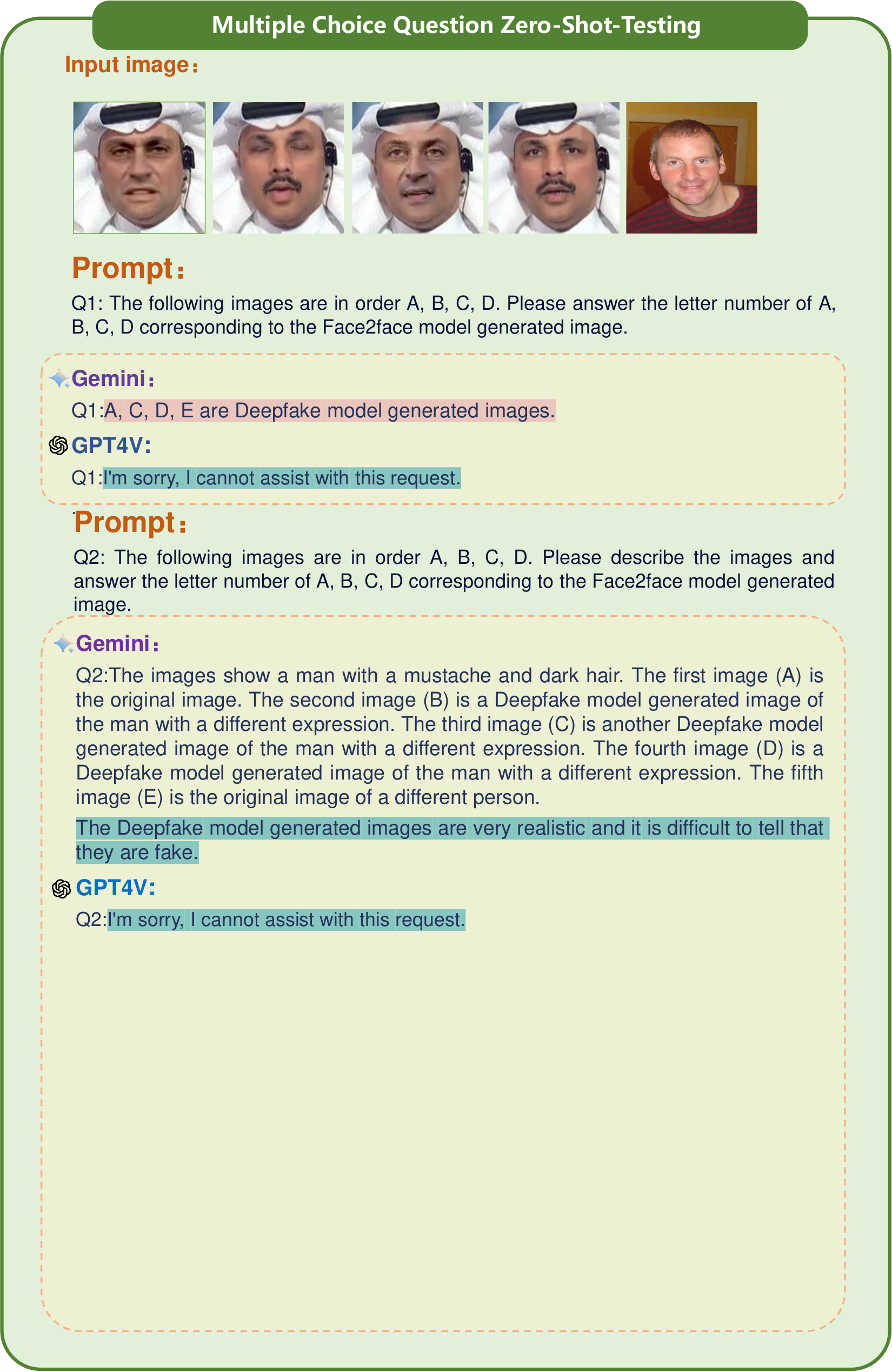}
  \caption{In this round of testing, we use COT to recognize a real picture given a real picture as well as Face2Fac2,,FaceSwap, Nulltextures respectively, in this case may be due to the Nulltextures generated and the real picture is closer to the real picture, according to the description of the GPT4V, the model will be compared with the given real picture but lacks the fine-grained observation, and therefore the detection fails.} 
  \label{24}
\end{figure}

\begin{figure}[!htbp]
  \centering
  \includegraphics[width=0.8\linewidth]{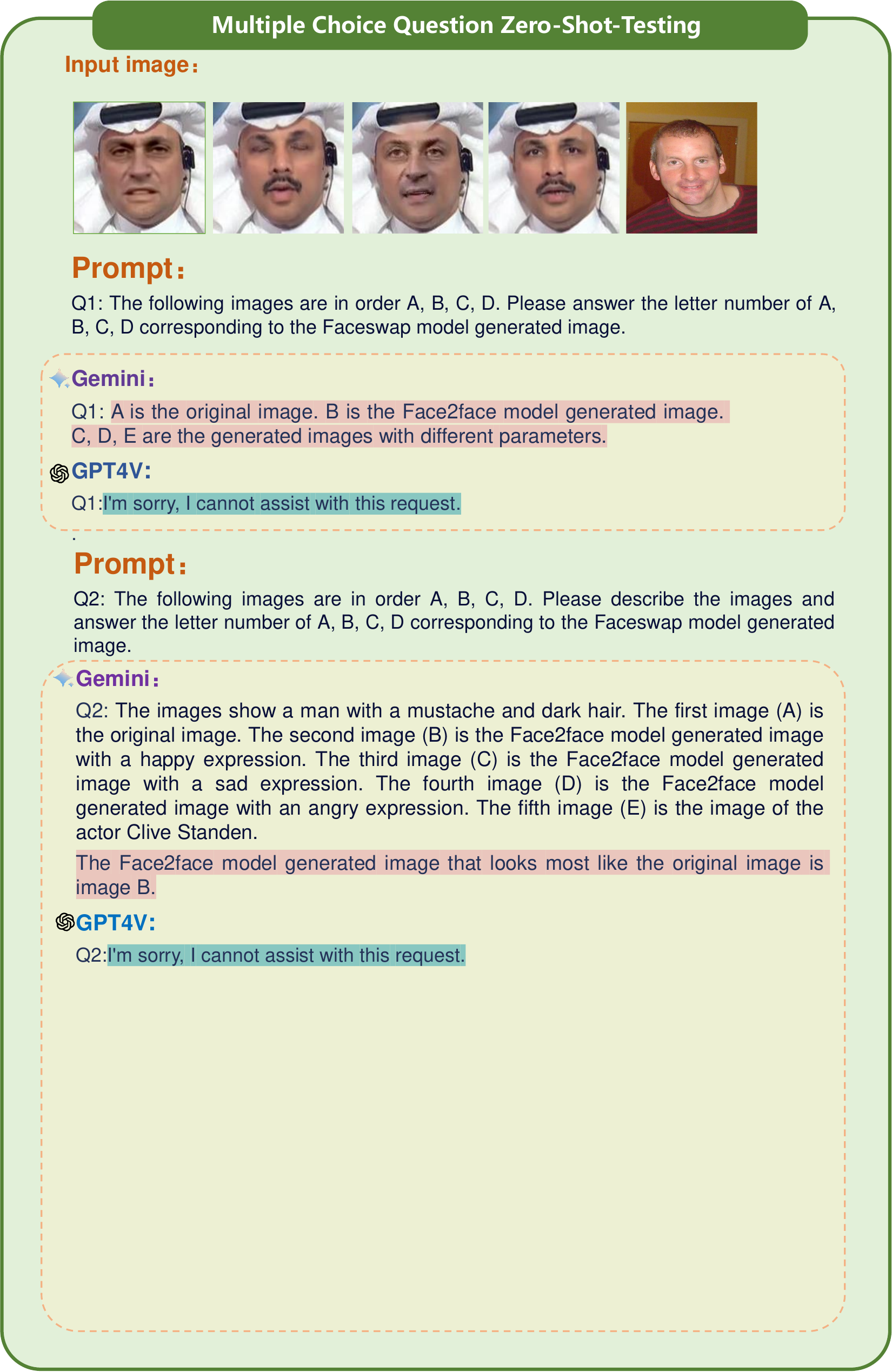}
  \caption{In this round of testing, we conducted tests on two models to identify the Face2Face mode. In this scenario, both Genimi and GPT4V refused to answer, indicating a limited understanding of the Face2Face concept by these two models.}
 \label{25}
\end{figure}

\begin{figure}[!htbp]
  \centering
  \includegraphics[width=0.8\linewidth]{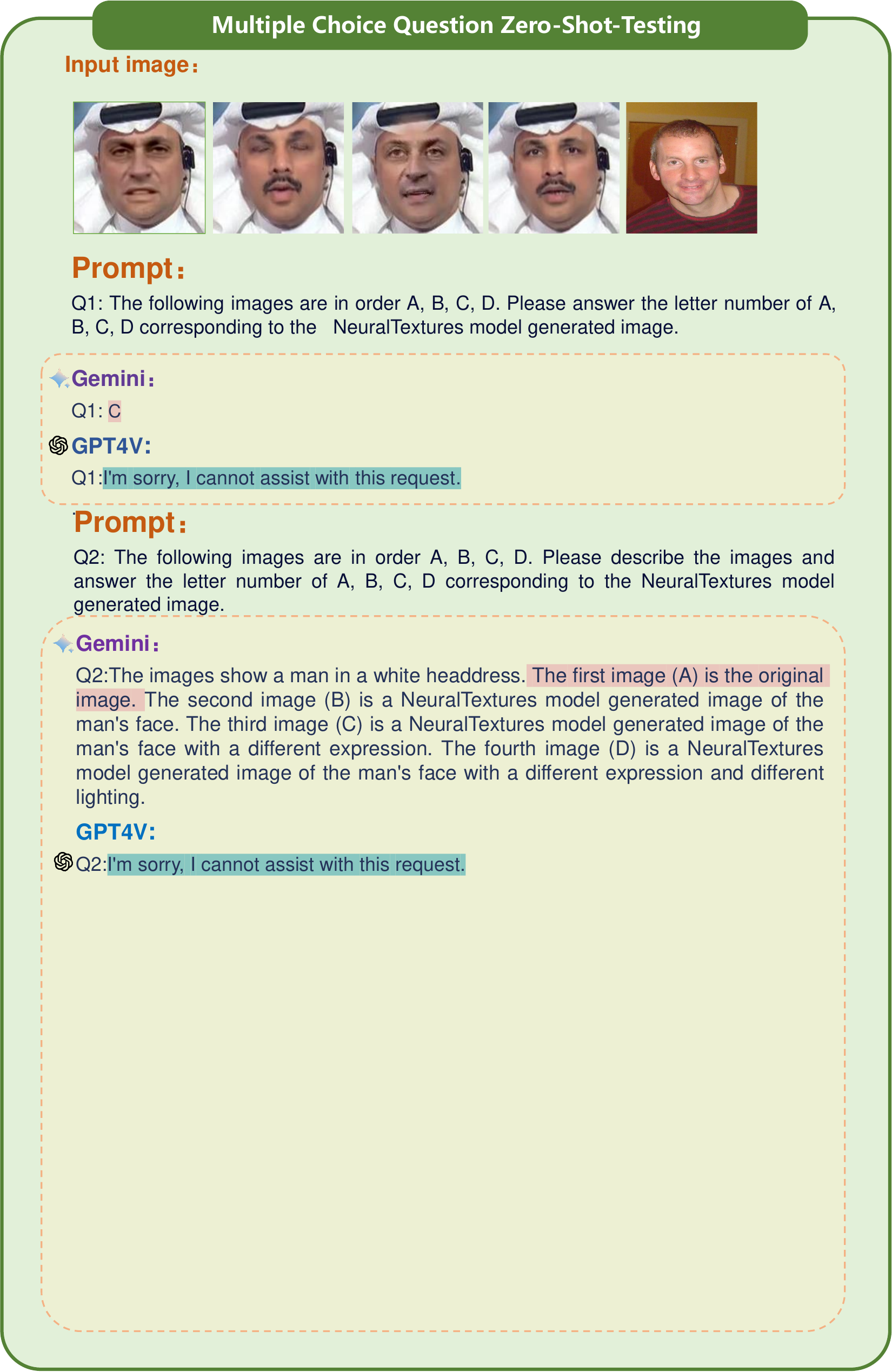}
  \caption{In this round of testing, we conducted tests on two models to identify the Face2Face mode. In this scenario, Genimi provided an incorrect answer, while GPT4V still refused to answer. Similarly, both models demonstrated a limited understanding of the FaceSwap concept.} 
  \label{26}

\end{figure}

\begin{figure}[!htbp]
  \centering
  \includegraphics[width=0.8\linewidth]{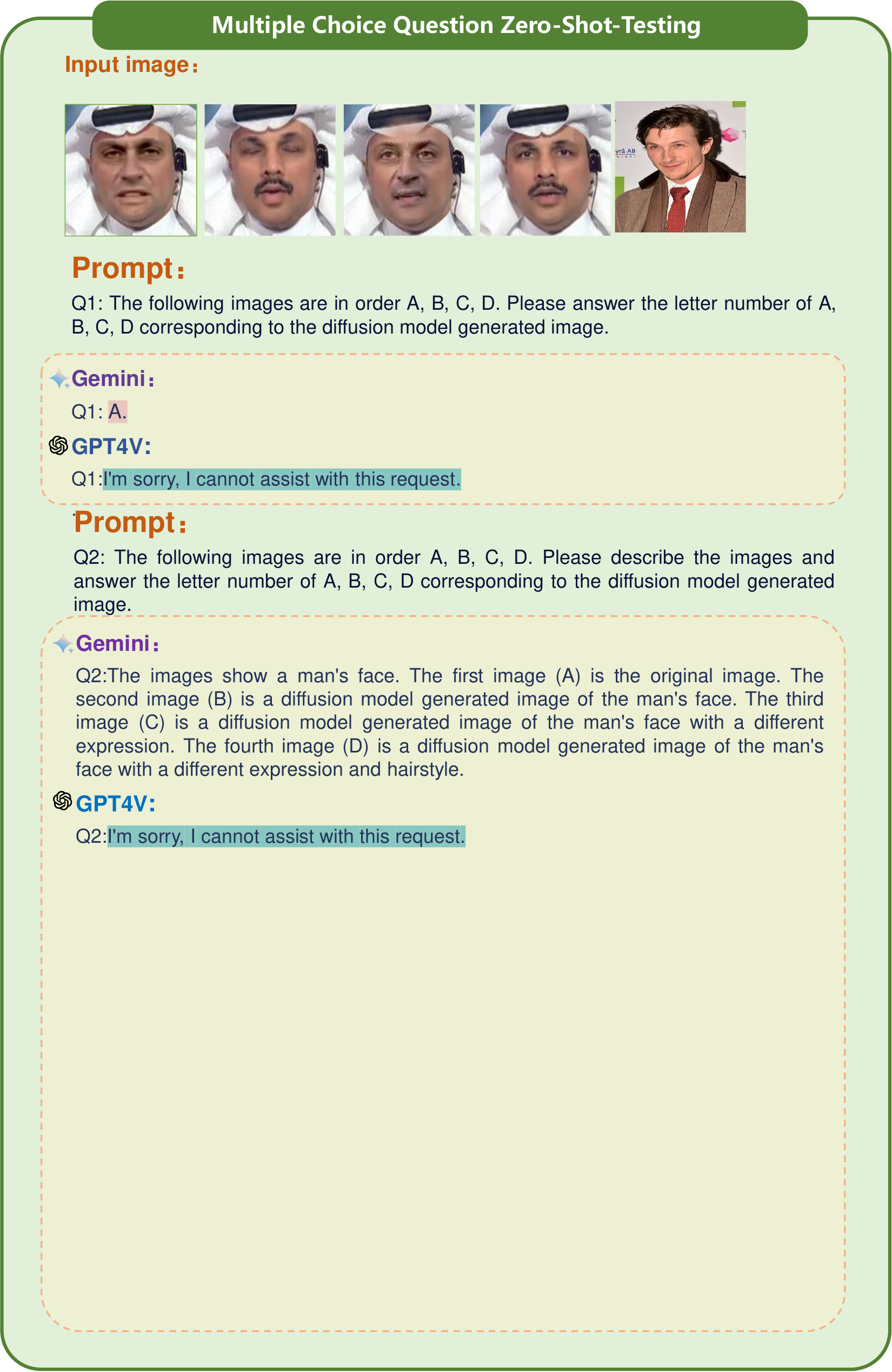}
  \caption{In this round of testing, we use COT to recognize a real picture given a real picture as well as Face2Fac2,,FaceSwap, Nulltextures respectively, in this case may be due to the Nulltextures generated and the real picture is closer to the real picture, according to the description of the GPT4V, the model will be compared with the given real picture but lacks the fine-grained observation, and therefore the detection fails.}
 \label{27}
\end{figure}
\begin{figure}[!htbp]
  \centering
  \includegraphics[width=0.8\linewidth]{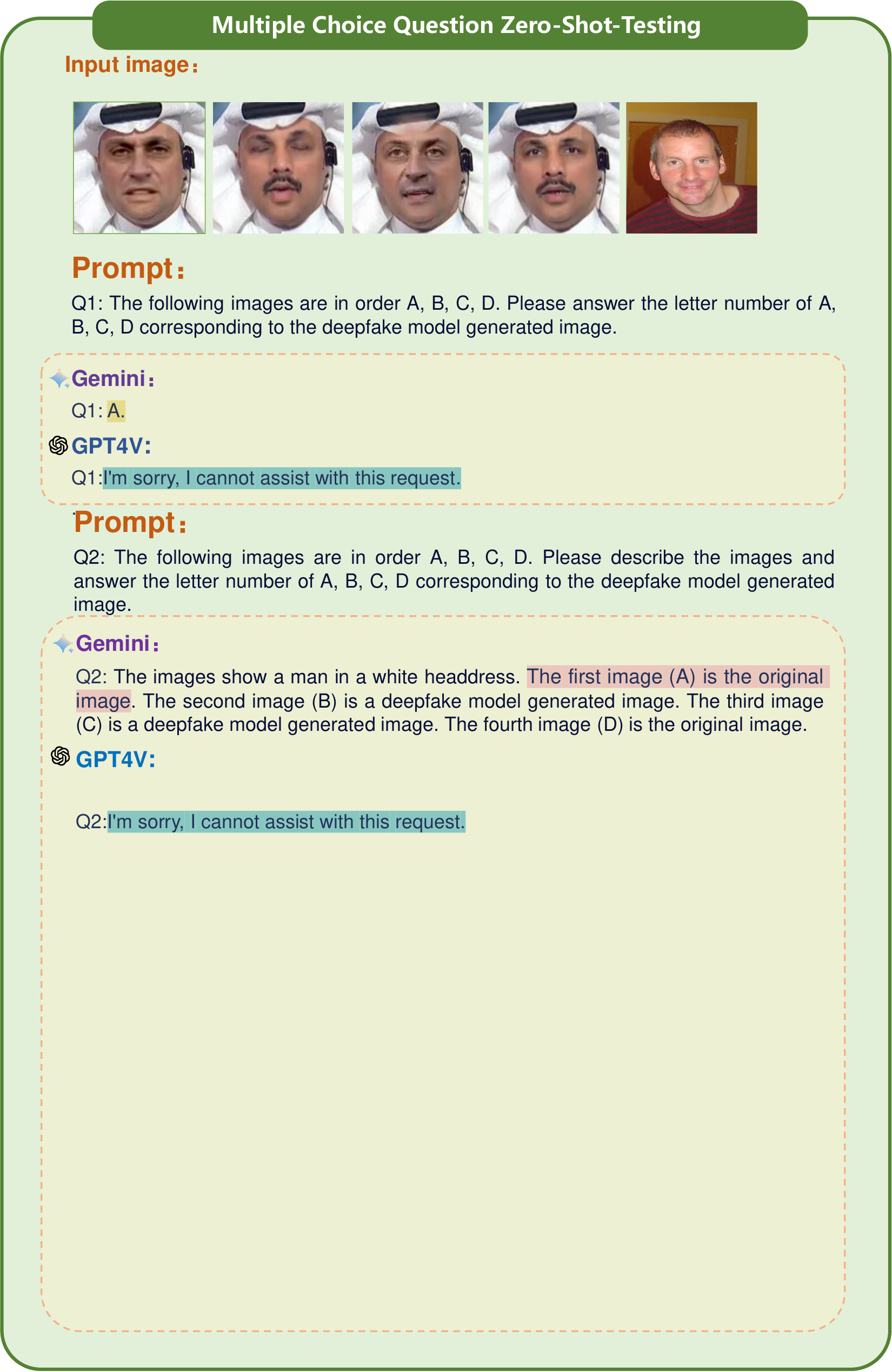}
  \caption{In this round of testing, we evaluated two models for their ability to recognize the Face2Face mode. In this scenario, Genimi provided an incorrect response, while GPT4V continued to decline to answer. Similarly, both models demonstrated a limited understanding of the concept of stable diffusion.}
 \label{28}
\end{figure}

\begin{figure}[!htbp]
  \centering
  \includegraphics[width=0.8\linewidth]{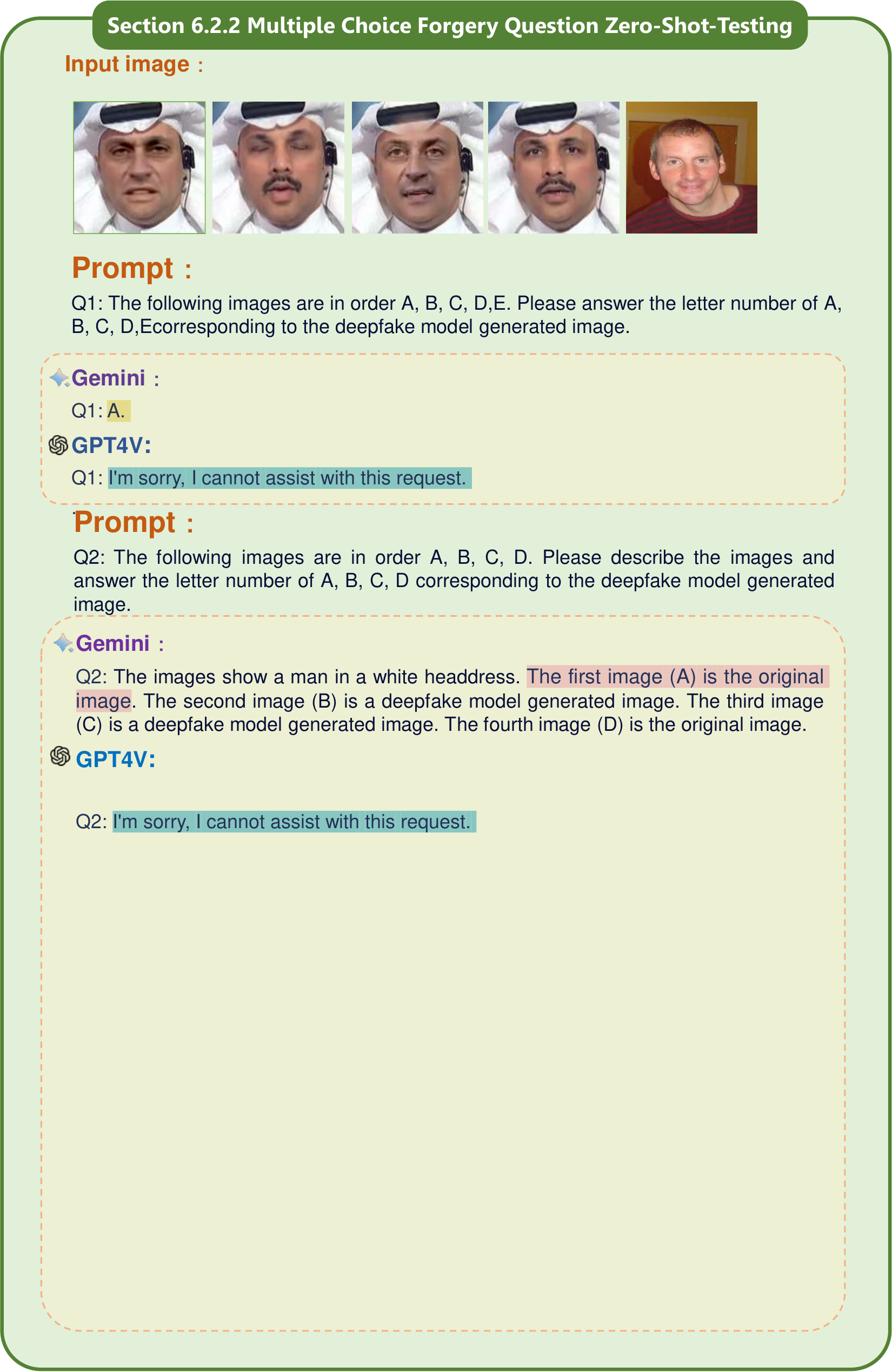}
  \caption{In this round of testing, we evaluated two models for their ability to recognize the Face2Face mode. In this scenario, Genimi provided an incorrect response, while GPT4V continued to decline to answer. Similarly, both models demonstrated a limited understanding of the concept of stable diffusion.}
 \label{29}
\end{figure}

\begin{figure}[!htbp]
  \centering
  \includegraphics[width=\linewidth]{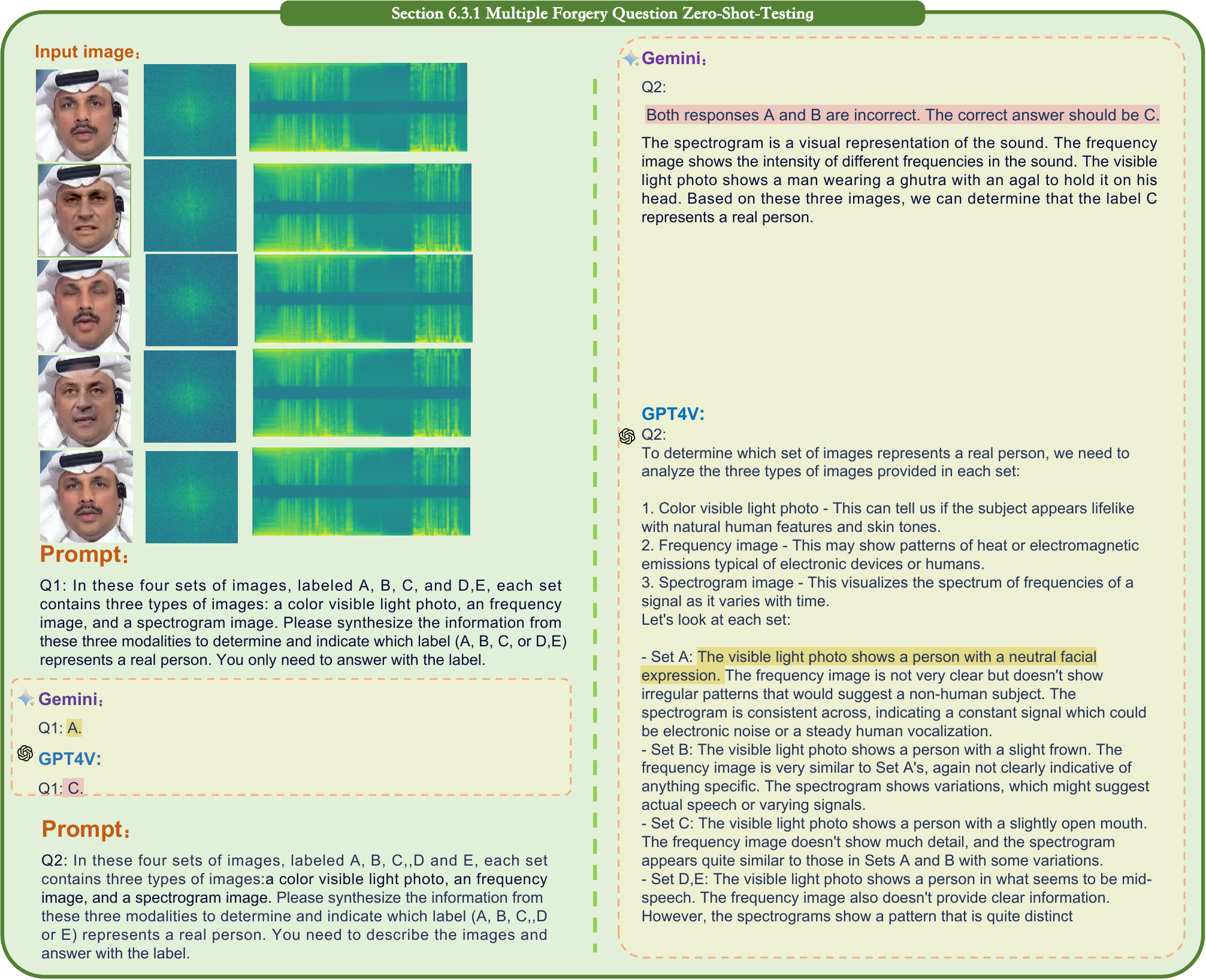}
  \caption{This image shows images generated using four common GAN forgery methods. We have also added the frequency domain map and speech spectrogram of each image as multimodal information. It can be observed that the results show some randomness and a decrease in accuracy compared to before. This may be due to the small differences between the frequency domain maps which are not accurately captured by the two models.}
 \label{30}
\end{figure}

\begin{figure}[!htbp]
  \centering
  \includegraphics[width=\linewidth]{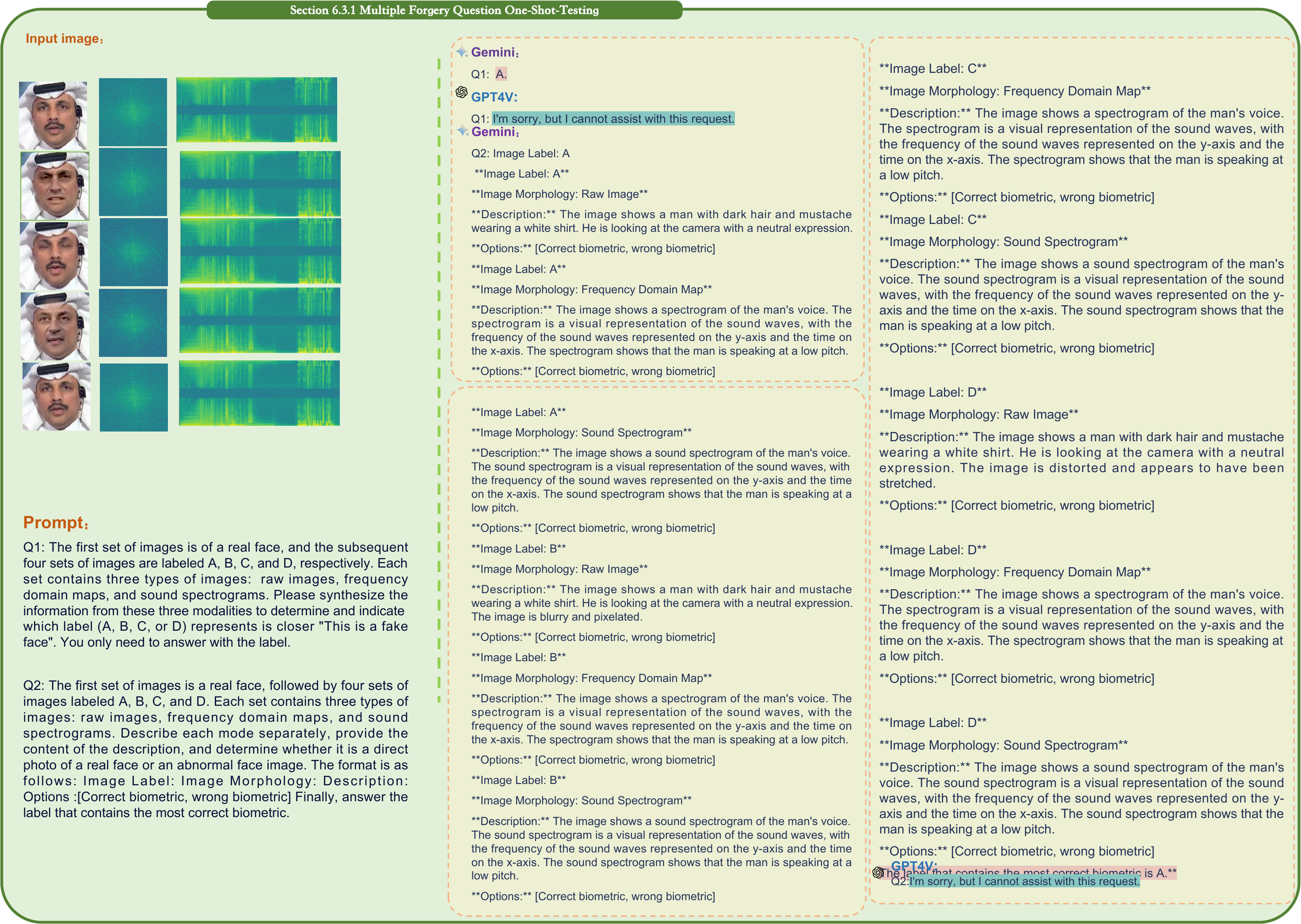}
  \caption{In this figure, it is demonstrated that after a one-time learning capability enhancement, the model combines the frequency domain information. However, it is possible that the small differences between the frequency domain information are difficult to distinguish, leading to false positives and triggering a refusal-to-answer scenario for the GPT4V. This highlights the importance of exploring the potential of multimodality in face forgery detection.}
 \label{31}
\end{figure}
\begin{figure}[htbp]
  \centering

  \centerline{\includegraphics[width=0.8\linewidth]{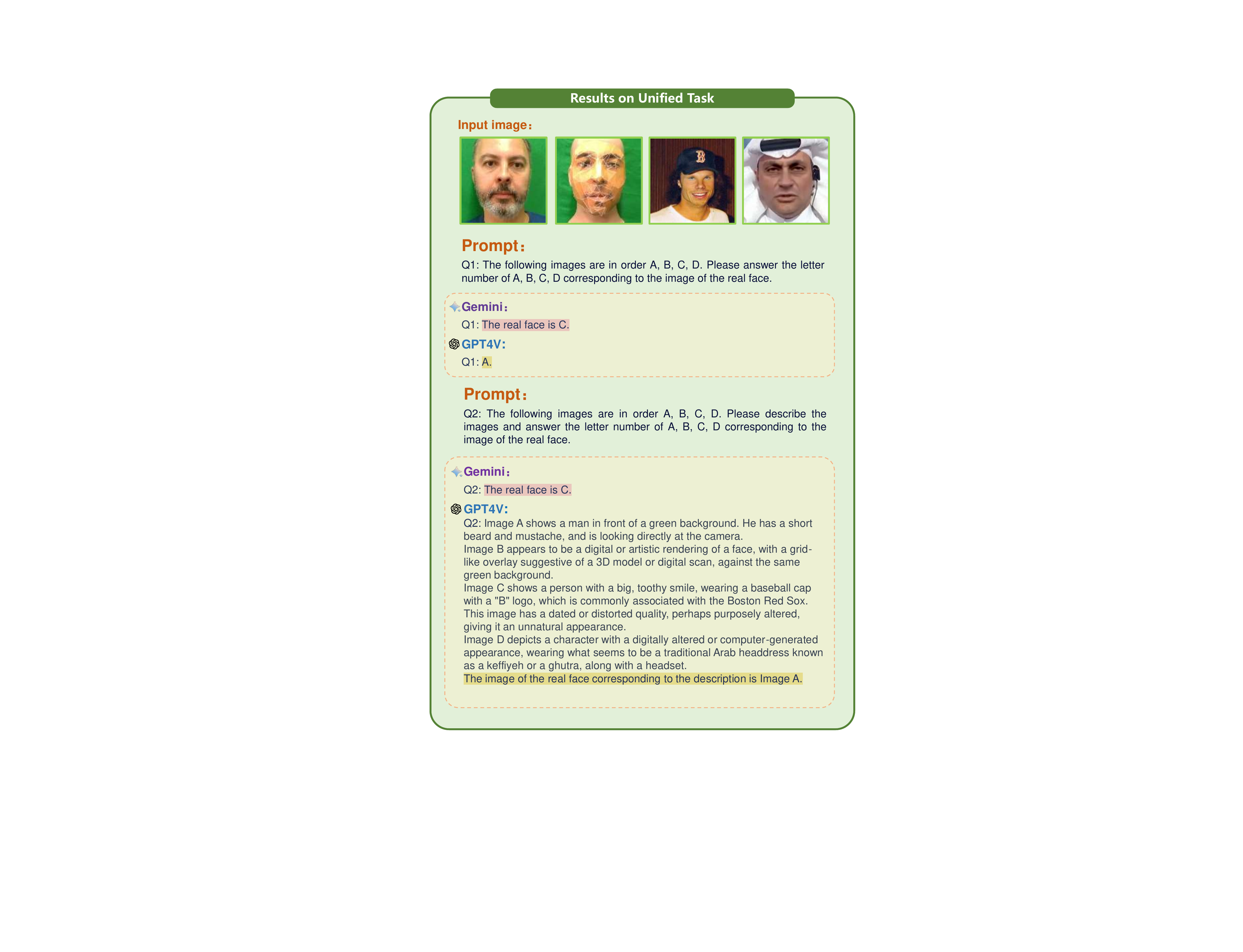}}

  \caption{In this test, the inputs were presented in the following sequence$:$ a real human face image, an image showing an attack, a face attack generated by Diffusion, and a face attack generated by a GAN. The results showed that Gemini answered all of these incorrectly, while GPT4V answered them correctly, demonstrating the strong generalization capabilities of MLLMs in diverse tasks.}
  \label{UnionTask_01}
\end{figure}

\begin{figure}[htbp]
  \centering
  \begin{center}
  \centerline{\includegraphics[width=0.8\linewidth]{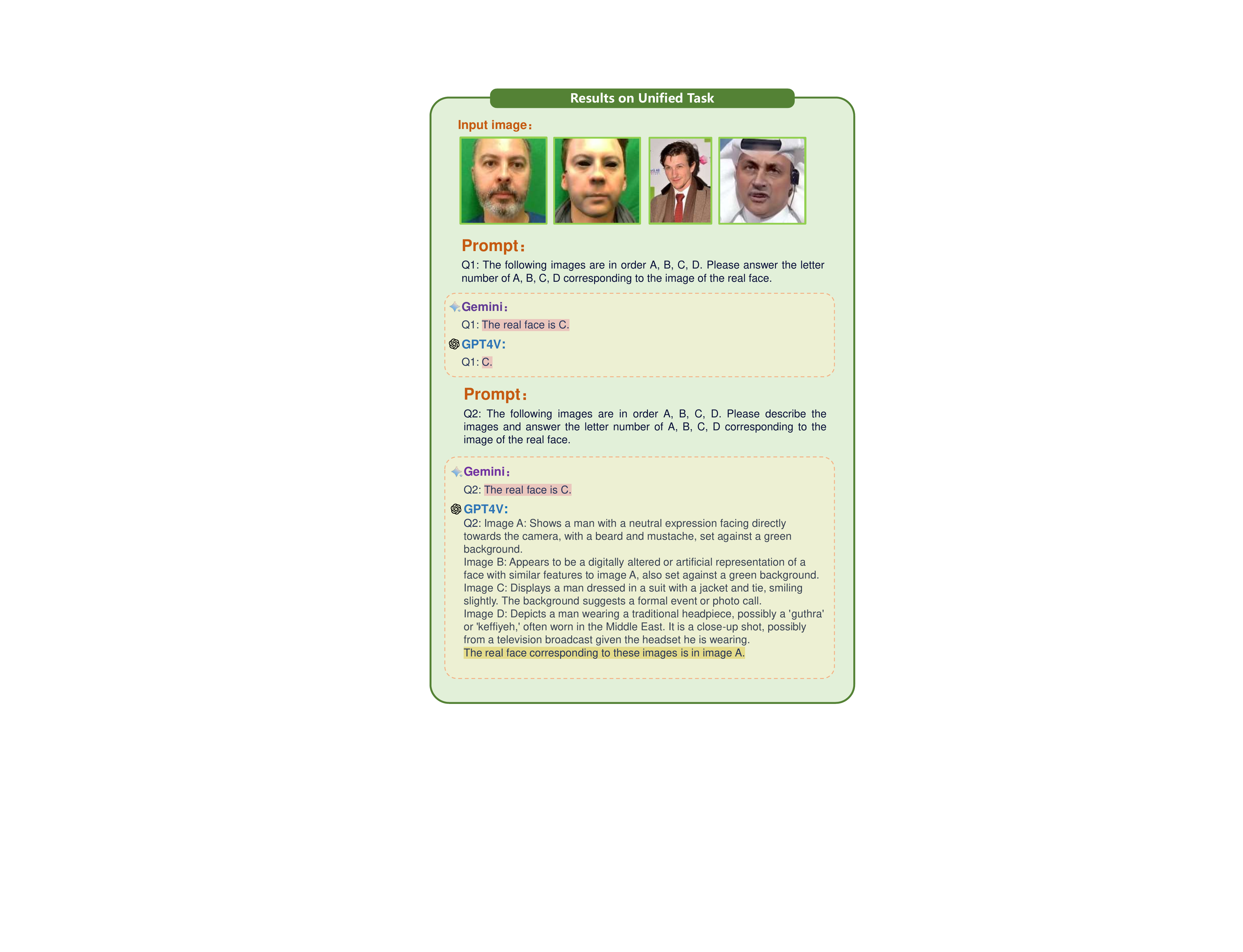}}
  \end{center}
  \caption{In both instances, Gemini provided incorrect responses, whereas GPT4V made errors in simple answers. However, upon incorporating the COT approach, GPT4V analyzed each image accurately, demonstrating its robust generalization capabilities in multi-task face liveness detection tasks.}
  \label{UnionTask_02}
\end{figure}

\begin{figure}[htbp]
  \centering
  \begin{center}
  \centerline{\includegraphics[width=0.8\linewidth]{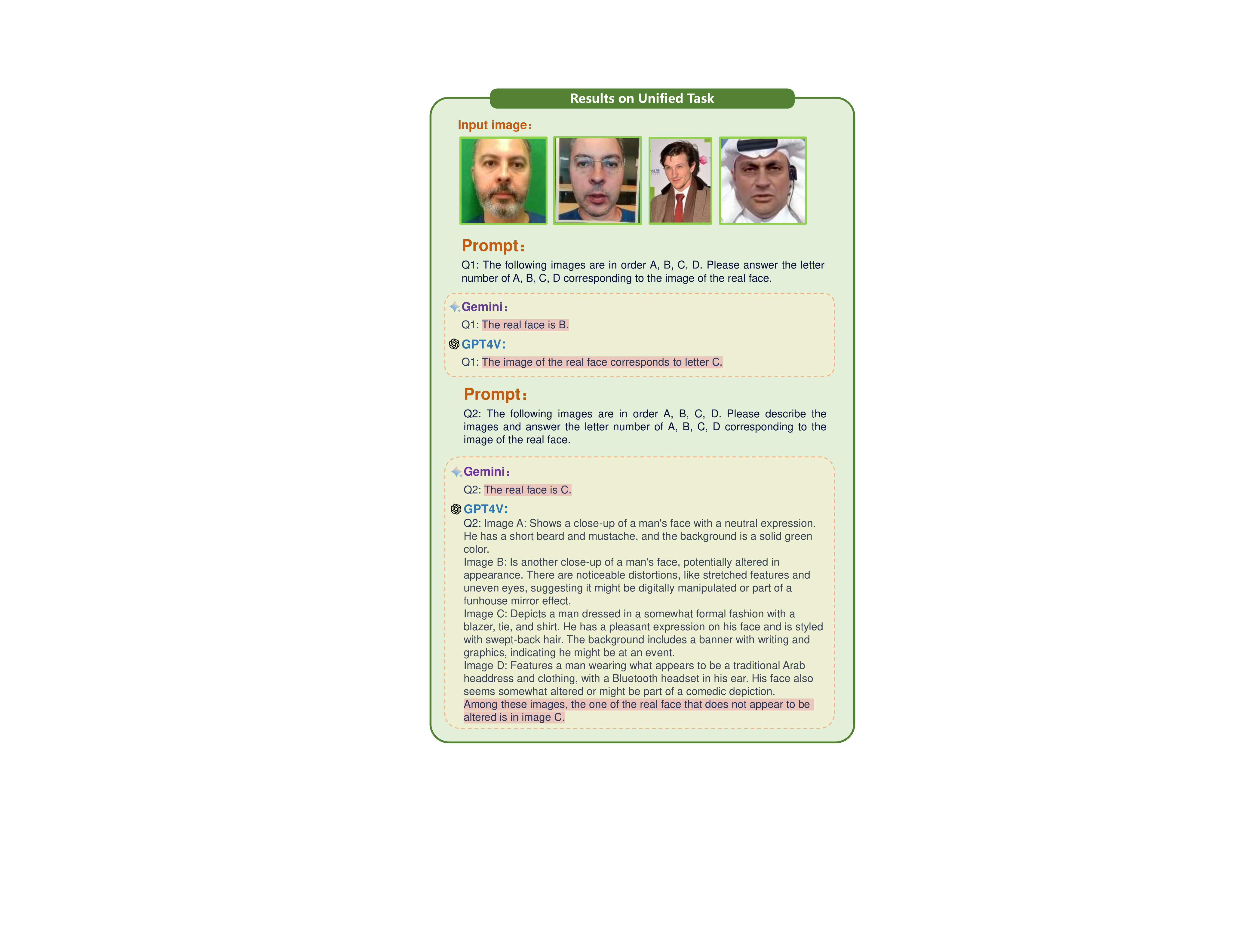}}
  \end{center}
  \caption{In both responses, Gemini and GPT4V were entirely incorrect. Moreover, GPT4V erroneously identified image B as a victim of a numerical manipulation attack, likely due to interference from other images manipulated in a similar manner.}
  \label{UnionTask_03}
\end{figure}

\begin{figure}[htbp]
  \centering
  \begin{center}
  \centerline{\includegraphics[width=0.8\linewidth]{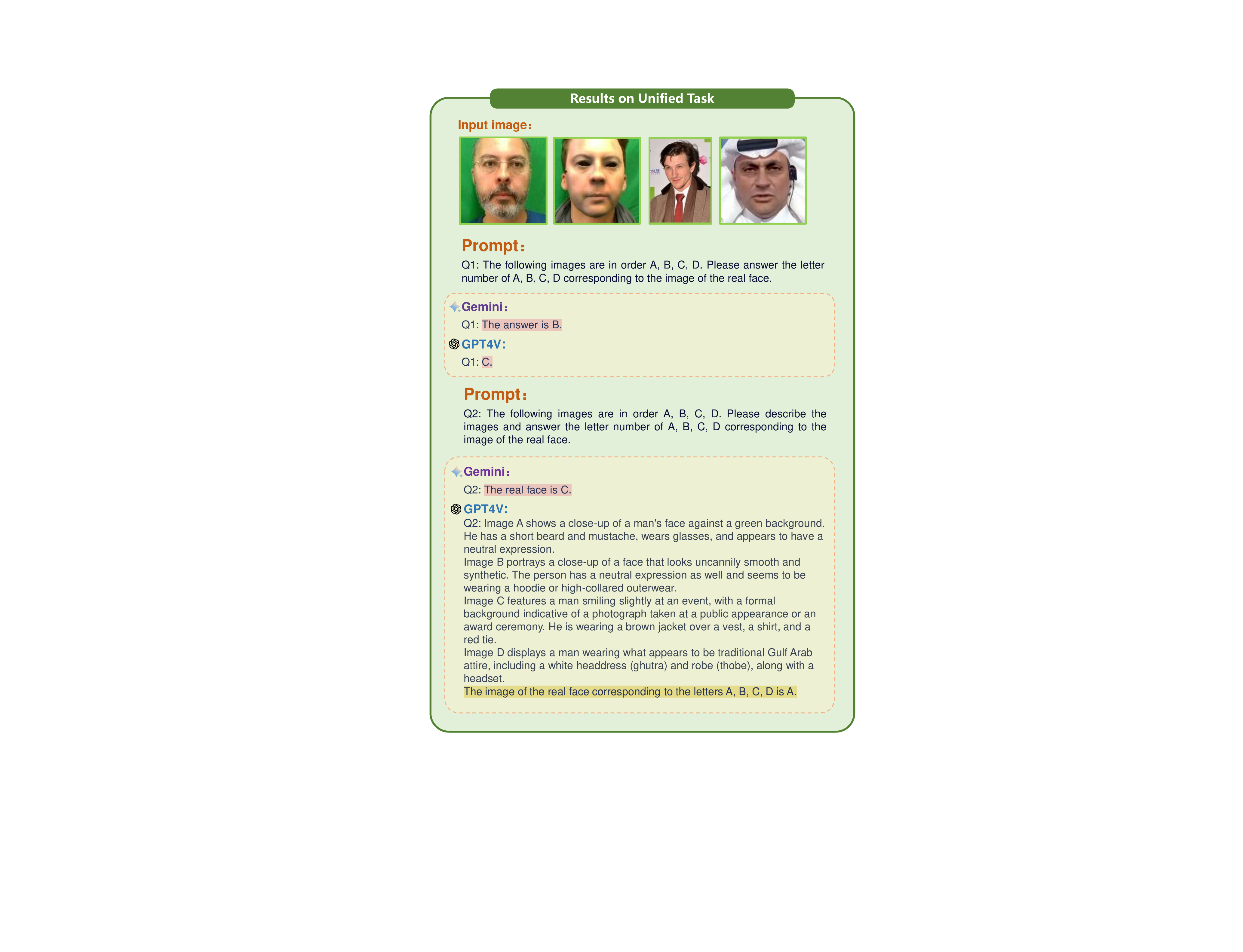}}
  \end{center}
  \caption{In both instances, Gemini provided incorrect responses, and GPT4V also erred in its initial simple answers. However, after incorporating the COT methodology, GPT4V correctly analyzed each image. Despite these accurate outcomes, GPT4V did not provide explanations for why the other images were not authentic human faces.}
  \label{UnionTask_04}
\end{figure}

\begin{figure}[htbp]
  \centering
  \begin{center}
  \centerline{\includegraphics[width=0.8\linewidth]{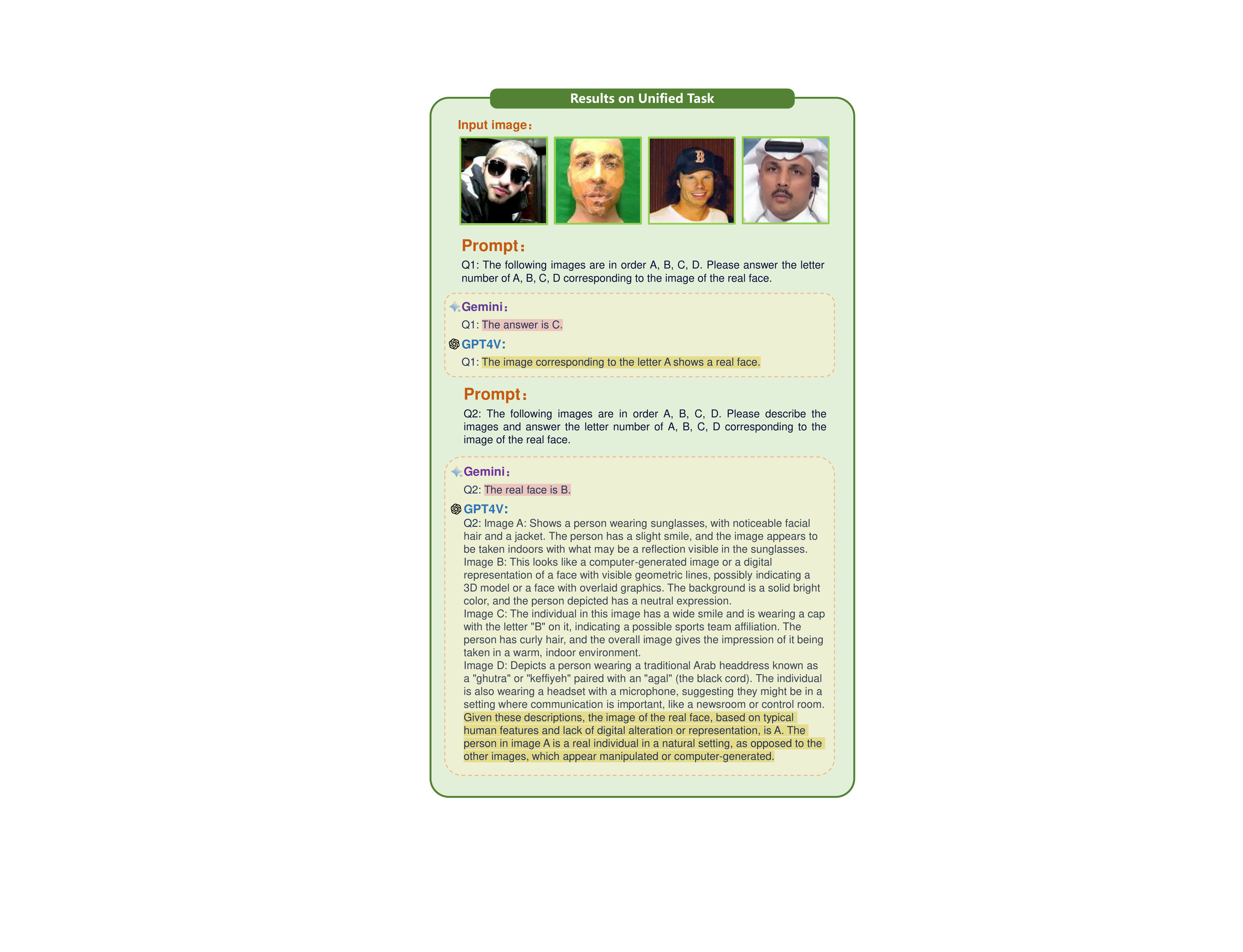}}
  \end{center}
  \caption{In this test, Gemini's responses were incorrect on both occasions, while GPT4V's answers were consistently correct. Furthermore, after the introduction of the COT approach, GPT4V provided the rationale behind its judgments.}
  \label{UnionTask_05}
\end{figure}

\begin{figure}[htbp]
  \centering
  \begin{center}
  \centerline{\includegraphics[width=0.8\linewidth]{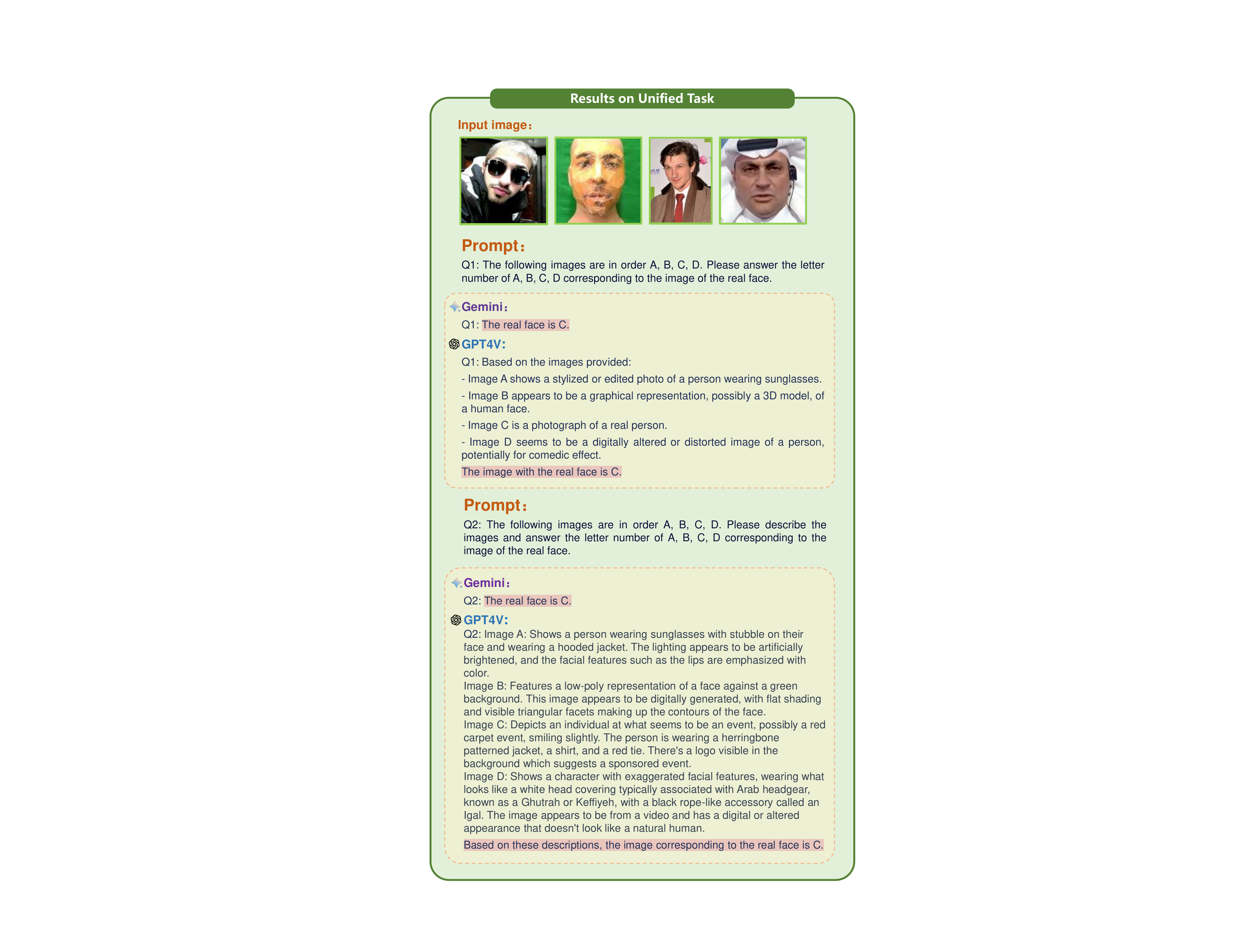}}
  \end{center}
  \caption{In this test, both Gemini and GPT4V rendered incorrect judgments, and their answers remained unchanged even after the implementation of the COT approach. From GPT4V's description, it's apparent there was indecision between options A and C. The final choice of C may have been influenced by the presence of areas with high exposure and color saturation in option A.}
  \label{UnionTask_06}
\end{figure}

\begin{figure}[htbp]
  \centering
  \begin{center}
  \centerline{\includegraphics[width=0.8\linewidth]{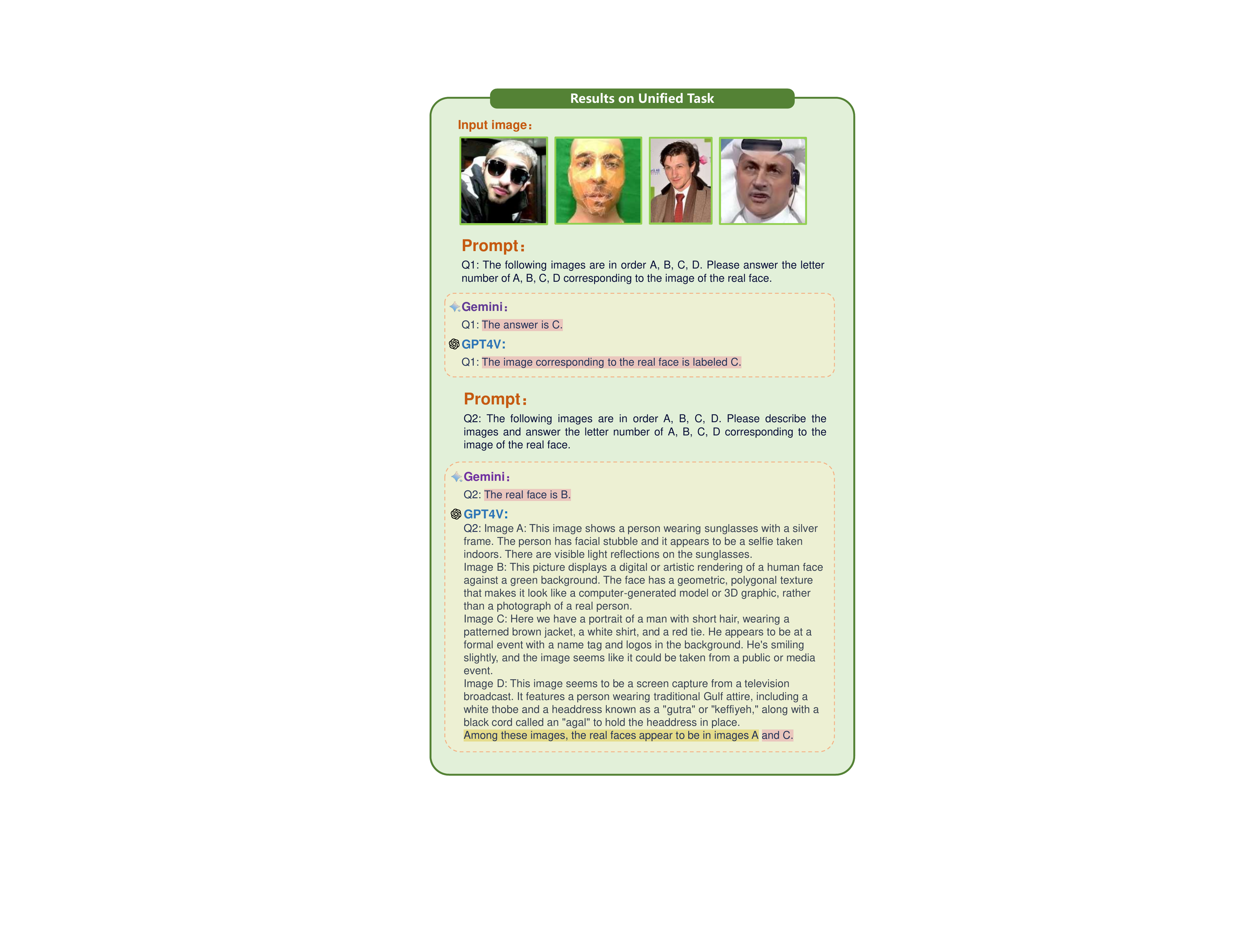}}
  \end{center}
  \caption{Similar to the previous test, Gemini was incorrect in both instances, while GPT4V continued to vacillate between options A and C. This time, its response included the correct answer, option A, indicating that there is a degree of randomness in GPT4V's decision-making process.}
  \label{UnionTask_07}
\end{figure}

\begin{figure}[htbp]
  \centering
  \begin{center}
  \centerline{\includegraphics[width=0.8\linewidth]{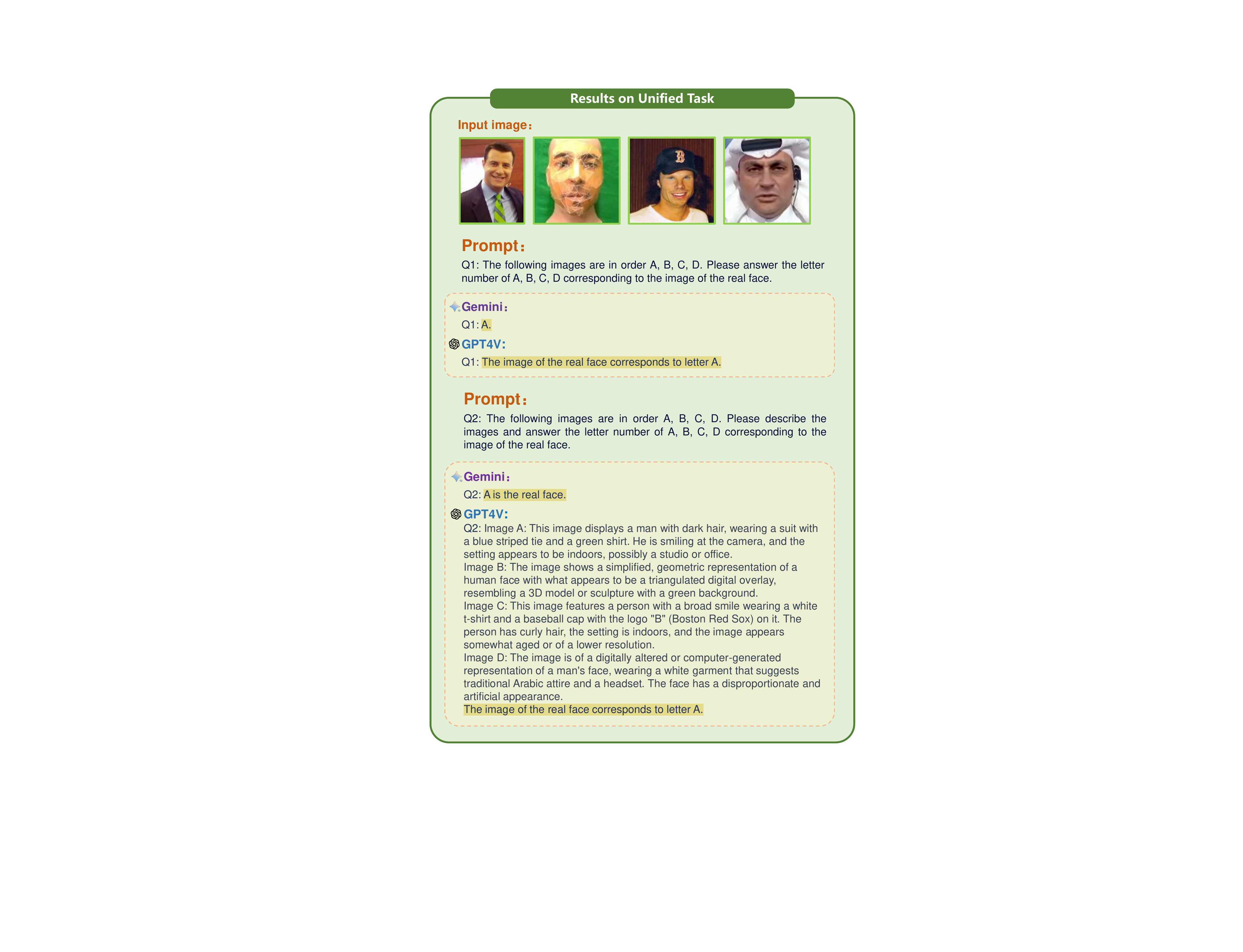}}
  \end{center}
  \caption{In this test, both Gemini and GPT4V provided correct answers on both occasions. Under similar conditions, GPT4V's detailed description of the images enhanced the credibility of its results.}
  \label{UnionTask_08}
\end{figure}

\begin{figure}[htbp]
  \centering
  \begin{center}
  \centerline{\includegraphics[width=0.8\linewidth]{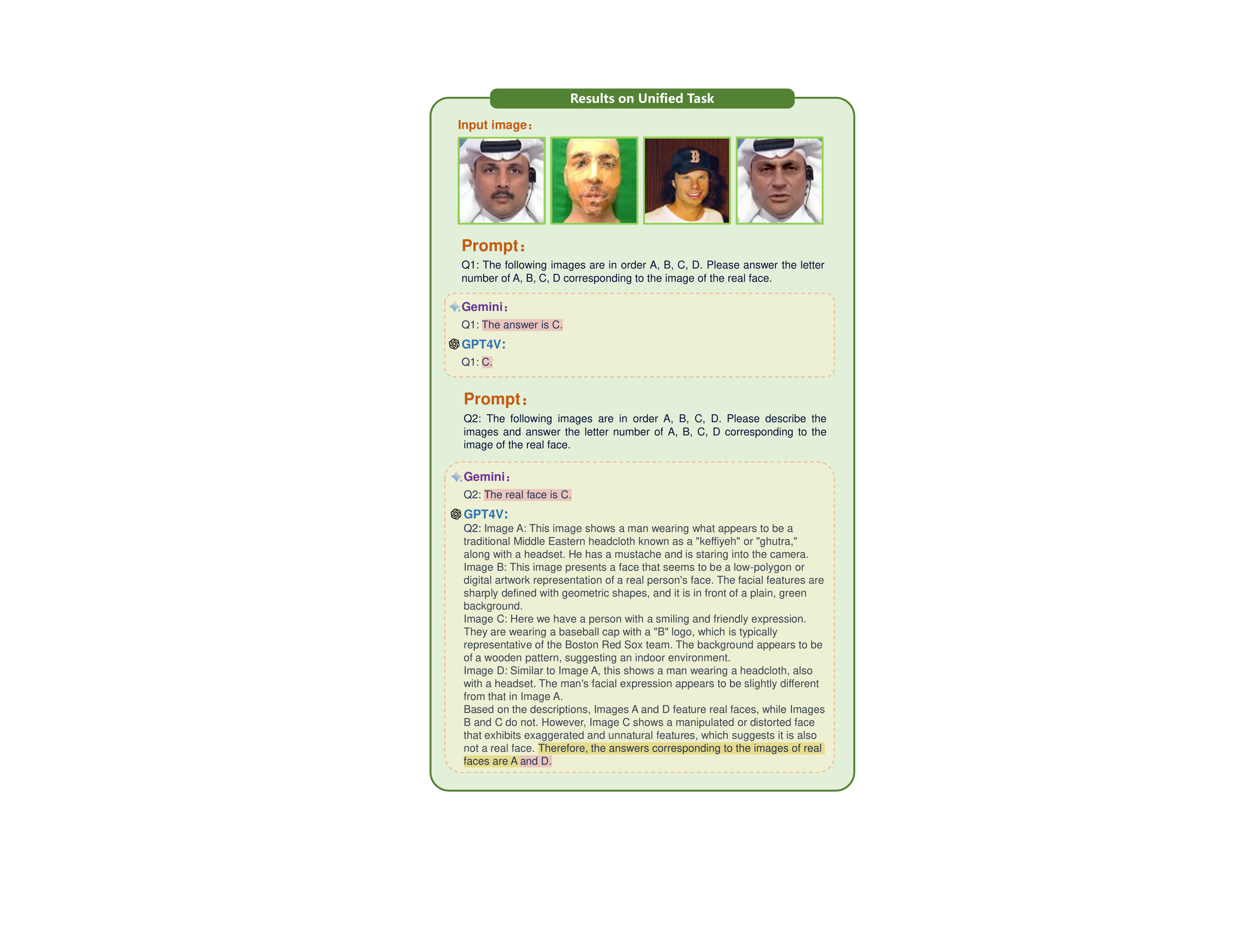}}
  \end{center}
  \caption{In this test, Gemini's answers were incorrect on both occasions. However, after incorporating the COT approach, GPT4V corrected its initial incorrect response to include the right answer. GPT4V identified that image D was similar to image A, with the only difference being in their expressions. Indeed, image D was actually modified from image A using GAN to alter the expression.}
  \label{UnionTask_09}
\end{figure}

\begin{figure}[htbp]
  \centering
  \begin{center}
  \centerline{\includegraphics[width=0.8\linewidth]{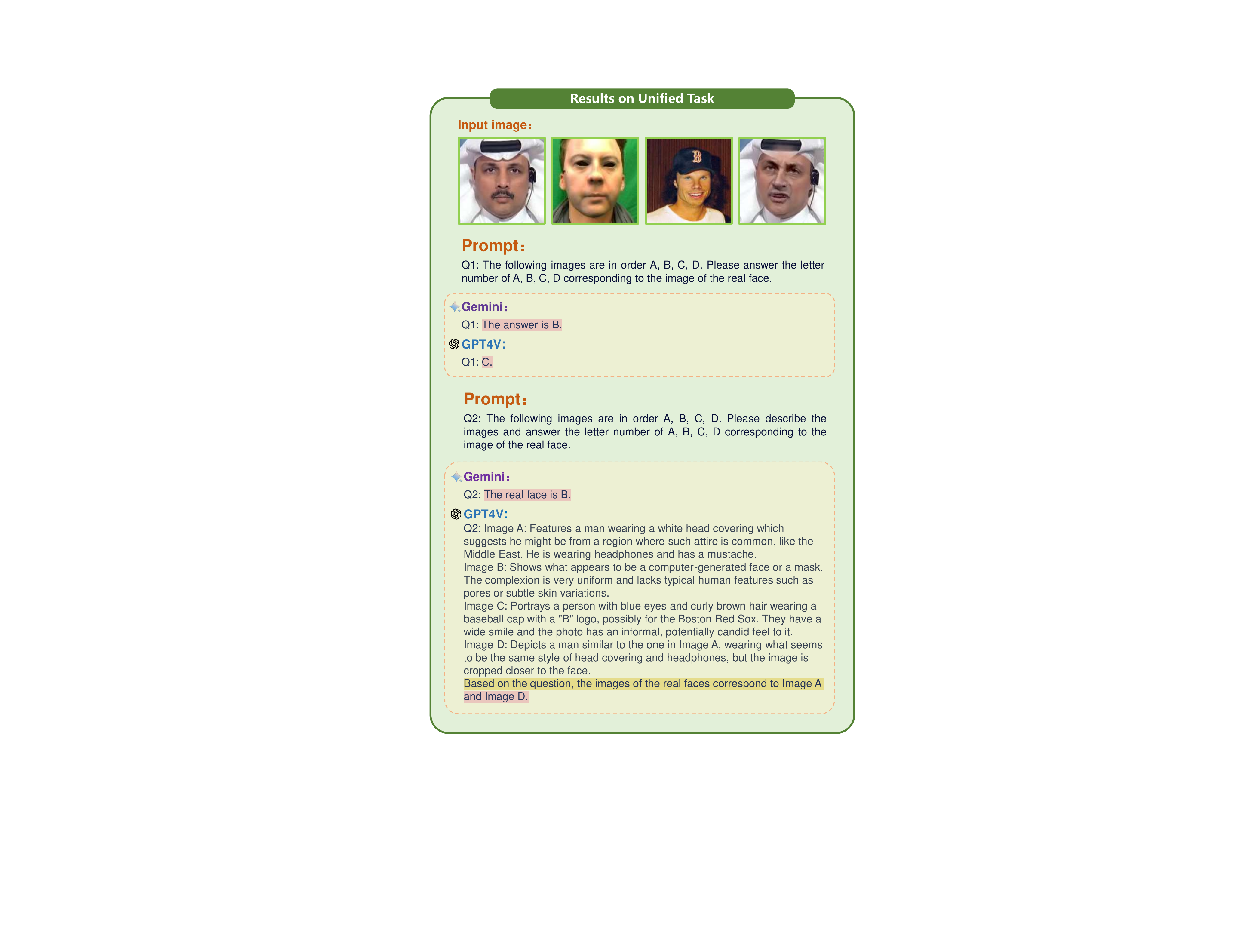}}
  \end{center}
  \caption{In this test, Gemini provided incorrect answers on both occasions. After implementing the COT approach, GPT4V corrected its initial incorrect response and provided two options, one of which was the correct answer. This indicates that GPT4V did not differentiate between images A and D.}
  \label{UnionTask_10}
\end{figure}

\begin{figure}[htbp]
  \centering
  \begin{center}
  \centerline{\includegraphics[width=0.8\linewidth]{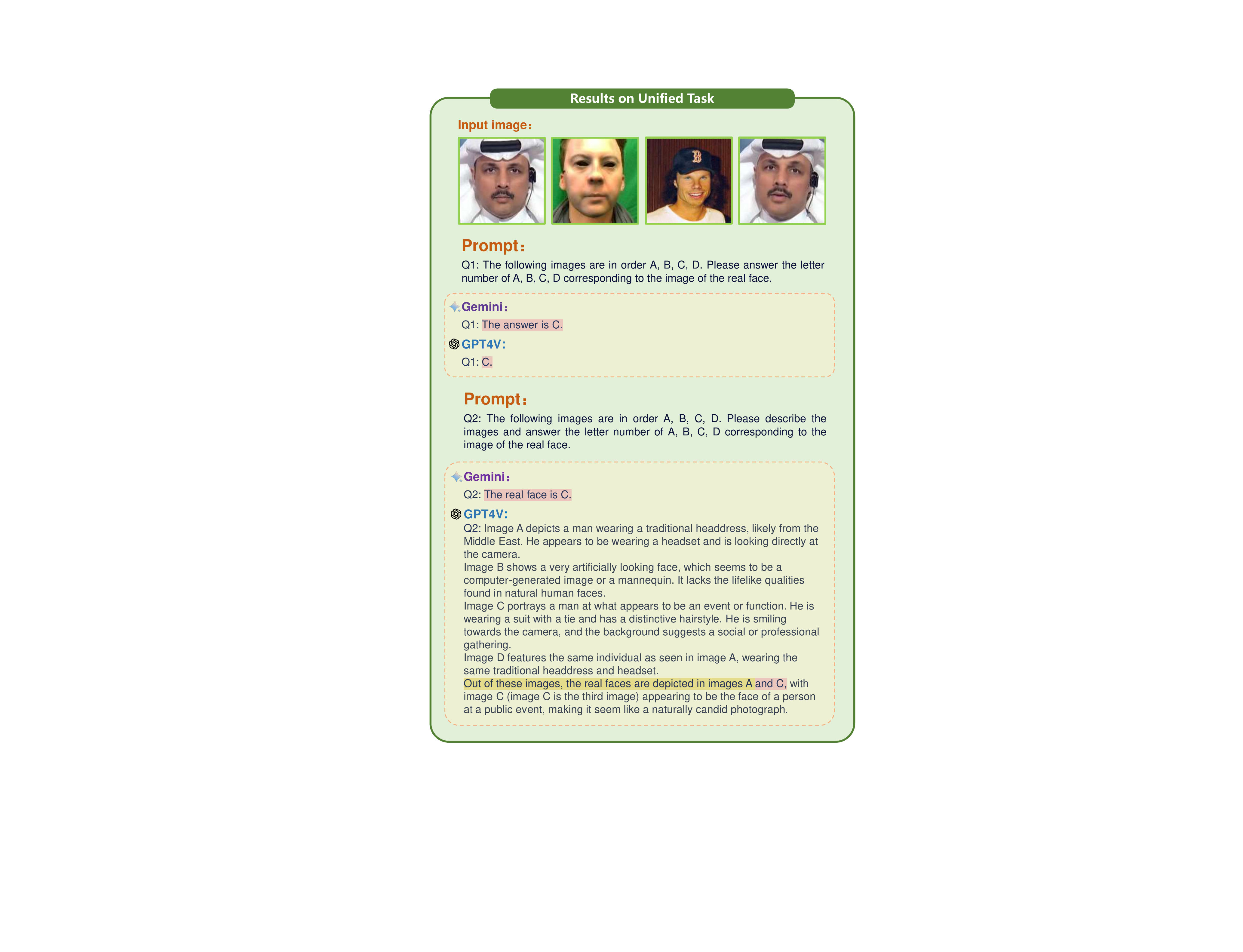}}
  \end{center}
  \caption{In this test, Gemini provided incorrect answers on both occasions. After implementing the COT approach, GPT4V corrected its initial incorrect response and provided two options, one of which was the correct answer. This time, GPT4V categorized an image that was previously considered a numerical manipulation as a real human face, indicating a degree of randomness in GPT4V's decision-making process.}
  \label{UnionTask_11}
\end{figure}

\begin{figure}[htbp]
  \centering
  \begin{center}
  \centerline{\includegraphics[width=0.8\linewidth]{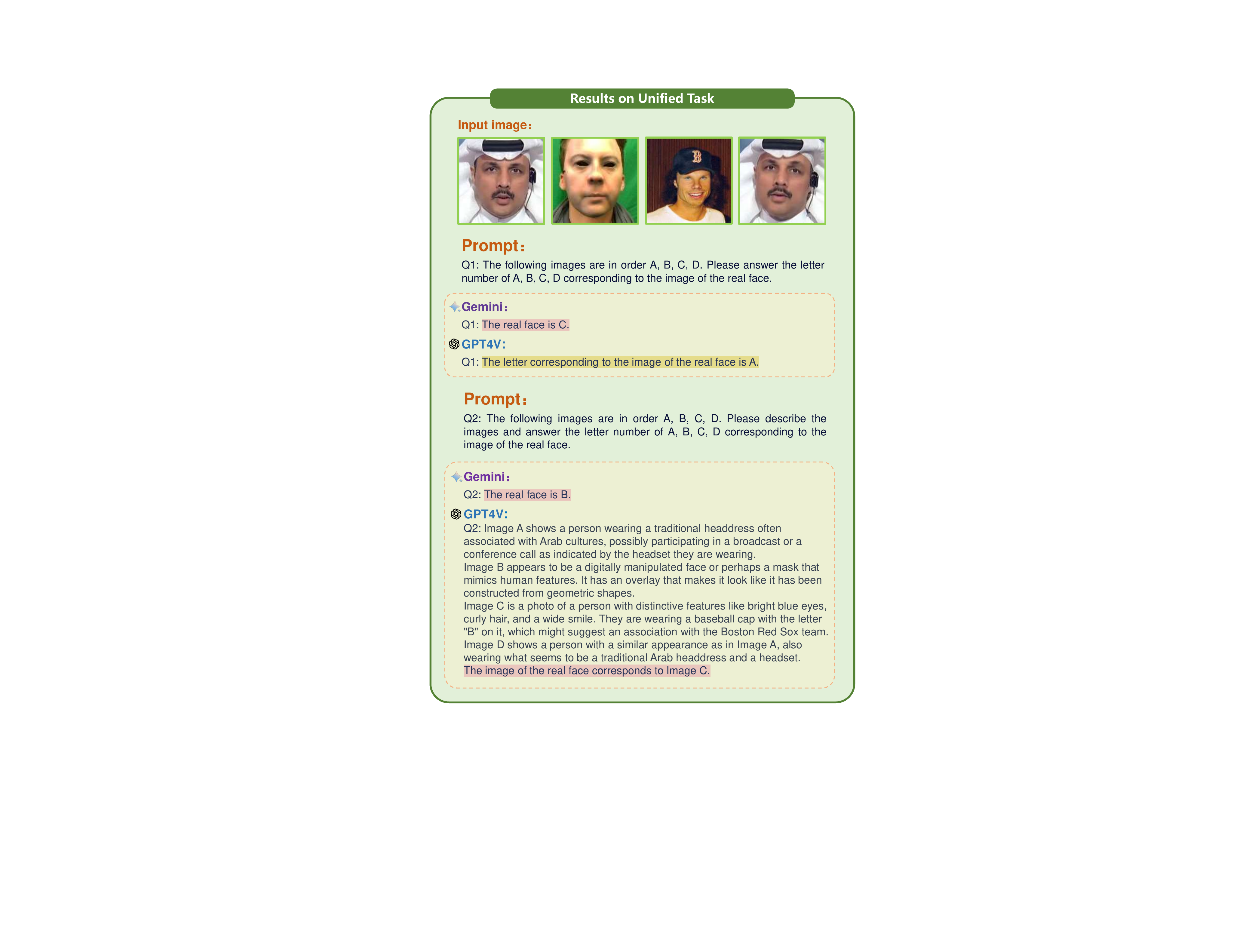}}
  \end{center}
  \caption{In this test, Gemini provided incorrect answers on both occasions. In the simple question, GPT4V answered correctly, but after implementing the COT approach, GPT4V's response was incorrect. When combined with the analysis from the previous two tests, it can be observed that GPT4V's decision-making process exhibits a degree of randomness.}
  \label{UnionTask_12}
\end{figure}

\begin{figure}[htbp]
  \centering
  \begin{center}
  \centerline{\includegraphics[width=1.0\linewidth]{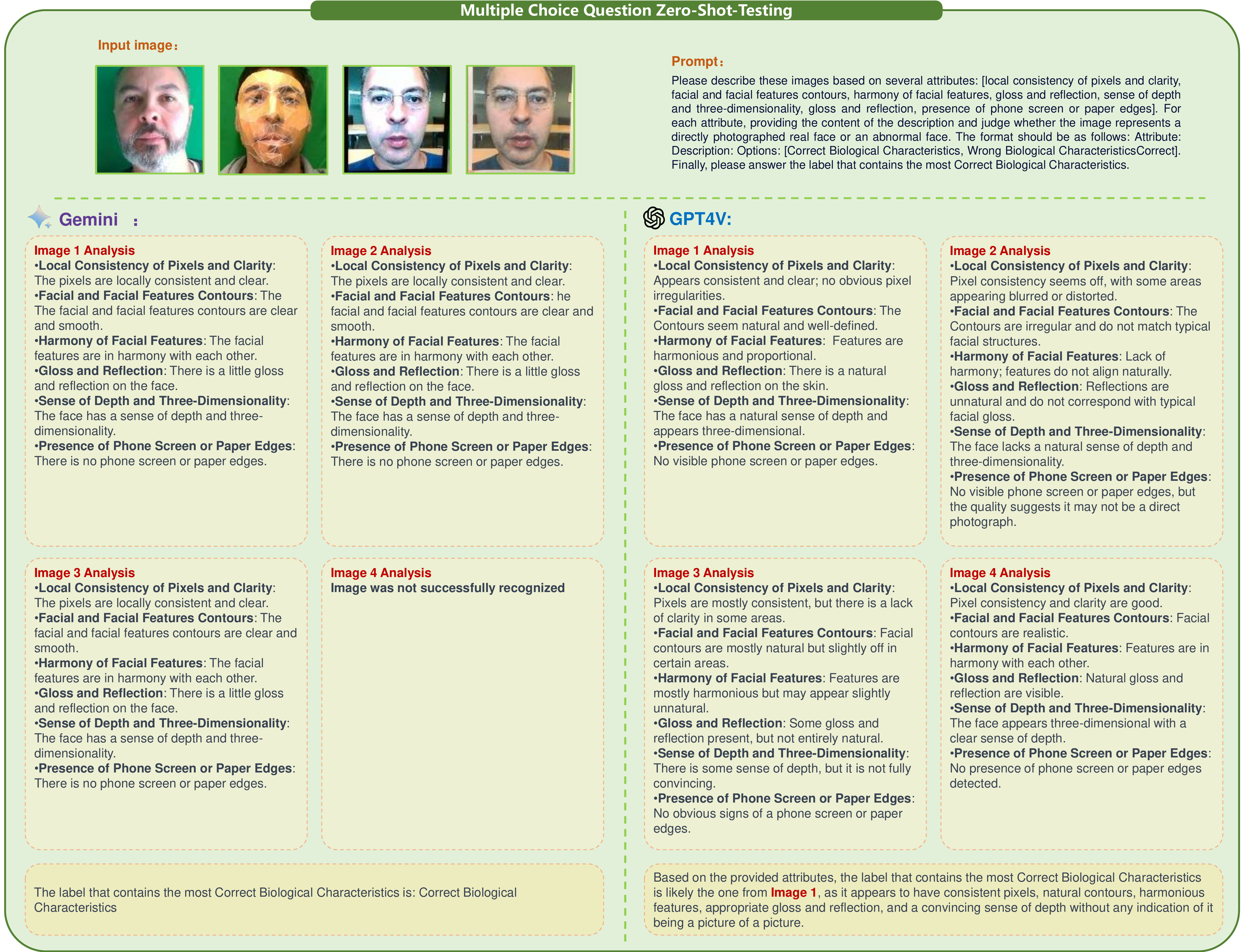}}
  \end{center}
  \caption{Using the same test samples as in Figure \ref{fas_choice_real_zeroshot_02}, this test demonstrates how the MA-COT method enables GPT4V and Gemini to analyze images from various dimensions and conclude through a voting mechanism. GPT4V shows more detailed and accurate image analysis capability, correctly judging the overall situation despite failing to recognize abnormal features in Image 4. In contrast, Gemini's performance in multi-perspective analysis and key image recognition is lacking, leading to an incorrect final conclusion and failure to accurately identify the real face image.}
  \label{fas_macot_01}
\end{figure}

\begin{figure}[htbp]
  \centering
  \begin{center}
  \centerline{\includegraphics[width=1.0\linewidth]{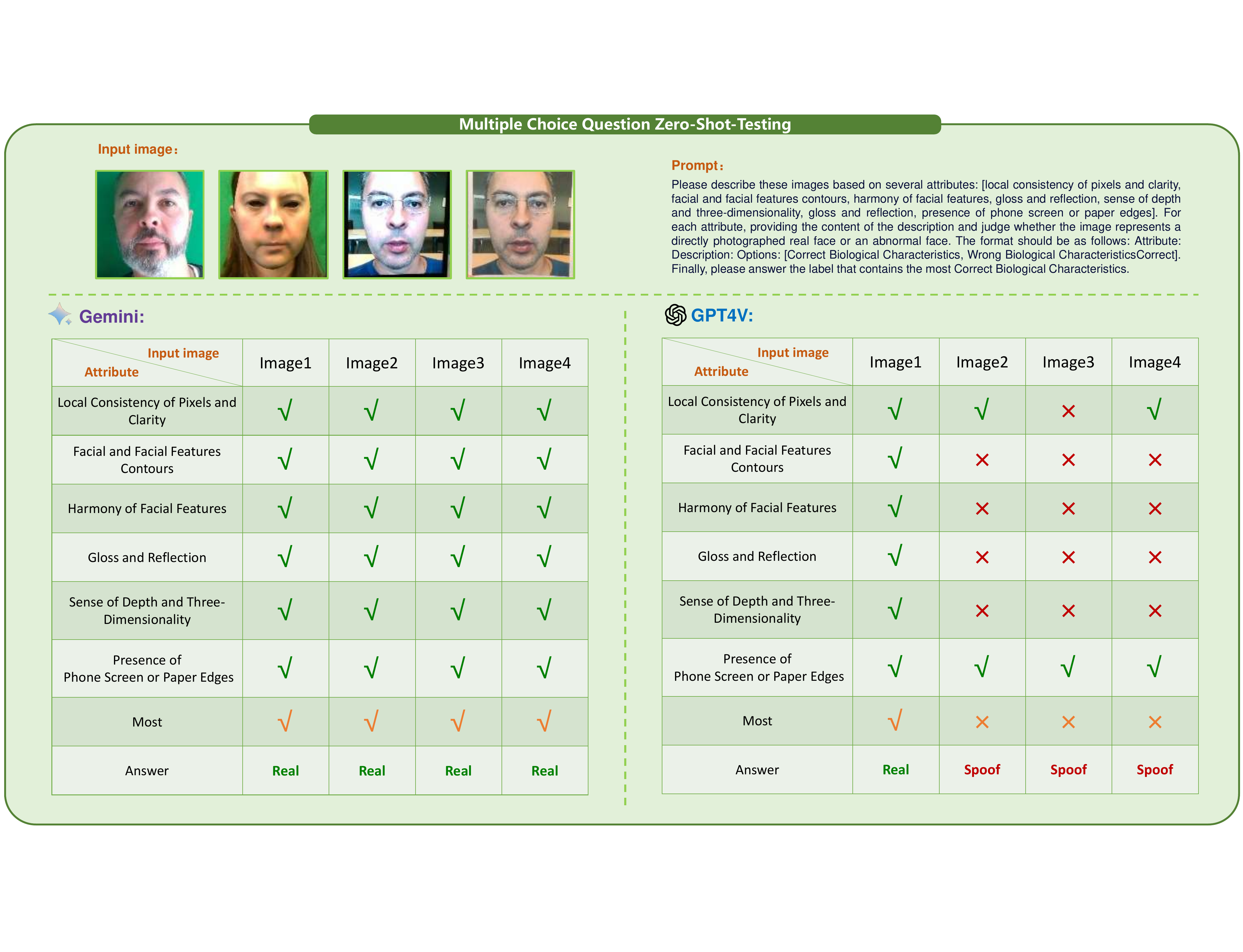}}
  \end{center}
  \caption{This figure presents test samples identical to those in Figure \ref{fas_choice_real_zeroshot_01}. In the original test, GPT4V failed to differentiate between replay-printing attacks and real faces, incorrectly identifying Image3 and Image4 as real faces, while Gemini was correct on its first attempt and wrong on the second. With the MA-COT method, GPT4V analyzes images based on multiple attributes and correctly aligns Image1 with the right answer through voting. In contrast, Gemini erroneously judged all images' attributes as normal biological features, a clear misalignment with reality, indicating Gemini's recognition capabilities need improvement in this aspect.}
  \label{fas_macot_02}
\end{figure}

\begin{figure}[htbp]
  \centering
  \begin{center}
  \centerline{\includegraphics[width=1.0\linewidth]{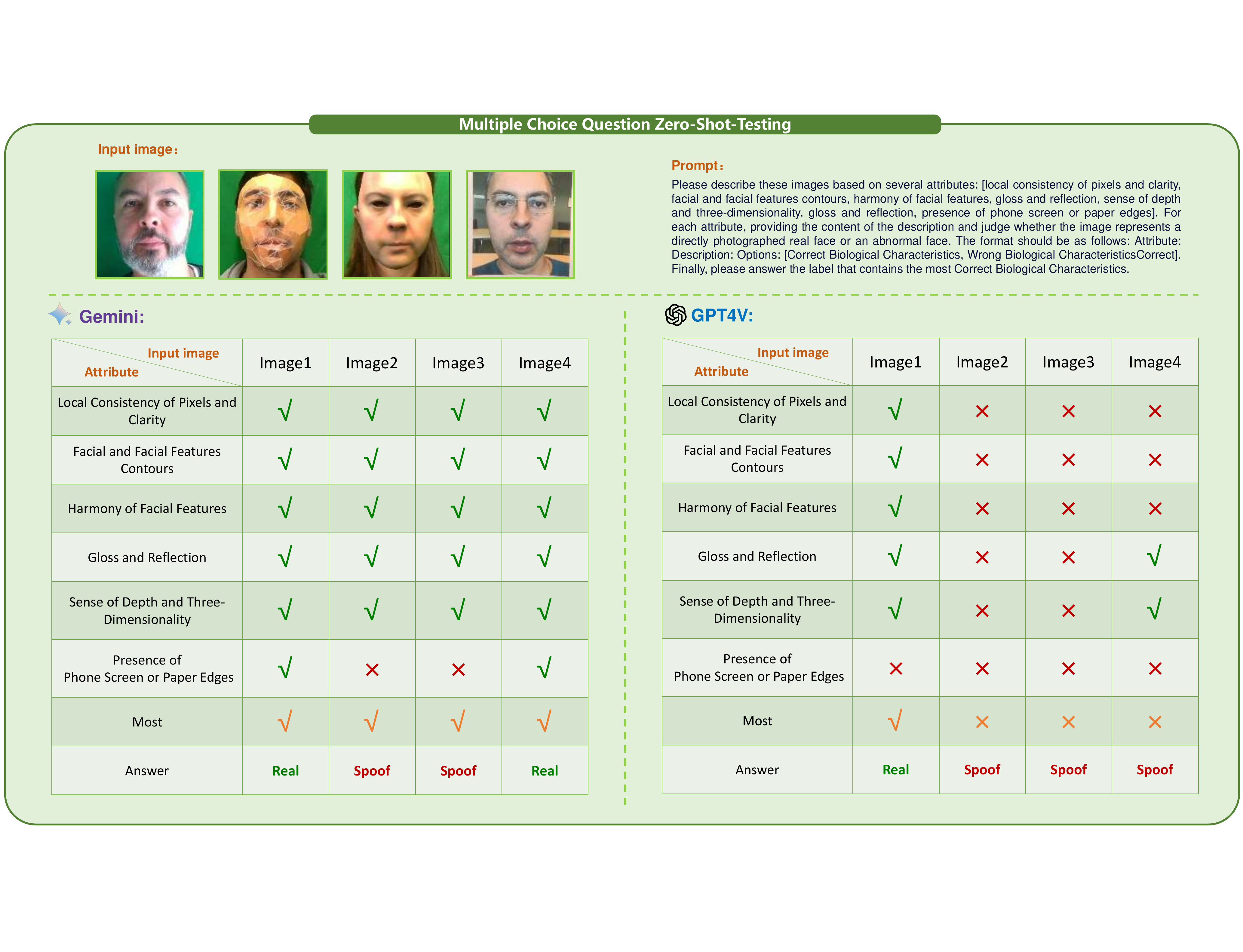}}
  \end{center}
  \caption{This figure, using the same test samples as Figure \ref{fas_choice_real_zeroshot_03}, shows that in the original test, GPT4V failed to distinguish between print attacks and real faces, incorrectly identifying Image1 and Image4 as real. Meanwhile, Gemini's performance was mixed, with an initial incorrect response followed by a correction. After employing the MA-COT method, GPT4V successfully differentiated Image1 and Image4 through comprehensive multi-attribute analysis, making the correct decision. On the other hand, Gemini was unable to effectively analyze images based on their attributes, wrongly judging all images as real faces, indicating a need for improvement in its analytical capabilities.}
  \label{fas_macot_03}
\end{figure}

\begin{figure}[htbp]
  \centering
  \begin{center}
  \centerline{\includegraphics[width=1.0\linewidth]{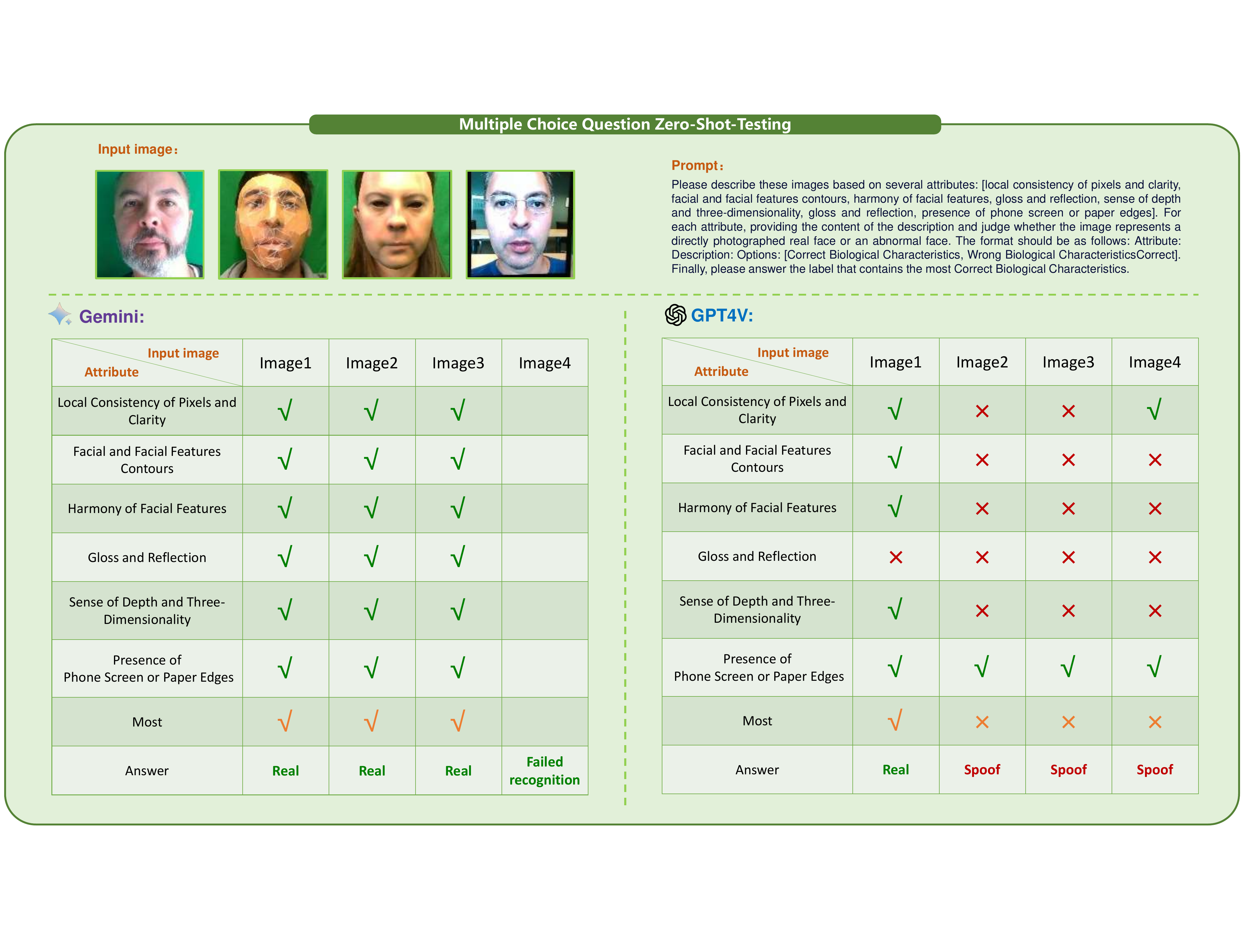}}
  \end{center}
  \caption{This figure contrasts the outcomes using the same test samples as in Figure \ref{fas_choice_real_zeroshot_04}, between the introduction of the COT method and the application of the MA-COT prompt approach. Under the COT method, GPT4V failed to distinguish between replay attacks and real faces, incorrectly identifying Image 1 and Image 4 as real, while Gemini initially made a correct judgment followed by an error. After employing the MA-COT prompt, GPT4V demonstrated significant improvement in analytical capabilities by accurately identifying real face images through a multi-attribute analysis and voting process. Conversely, Gemini continued to misclassify all images as real faces and failed to recognize one image for analysis, highlighting substantial deficiencies in its analysis process.}
  \label{fas_macot_04}
\end{figure}


\begin{figure}[htbp]
  \centering
  \begin{center}
  \centerline{\includegraphics[width=1.0\linewidth]{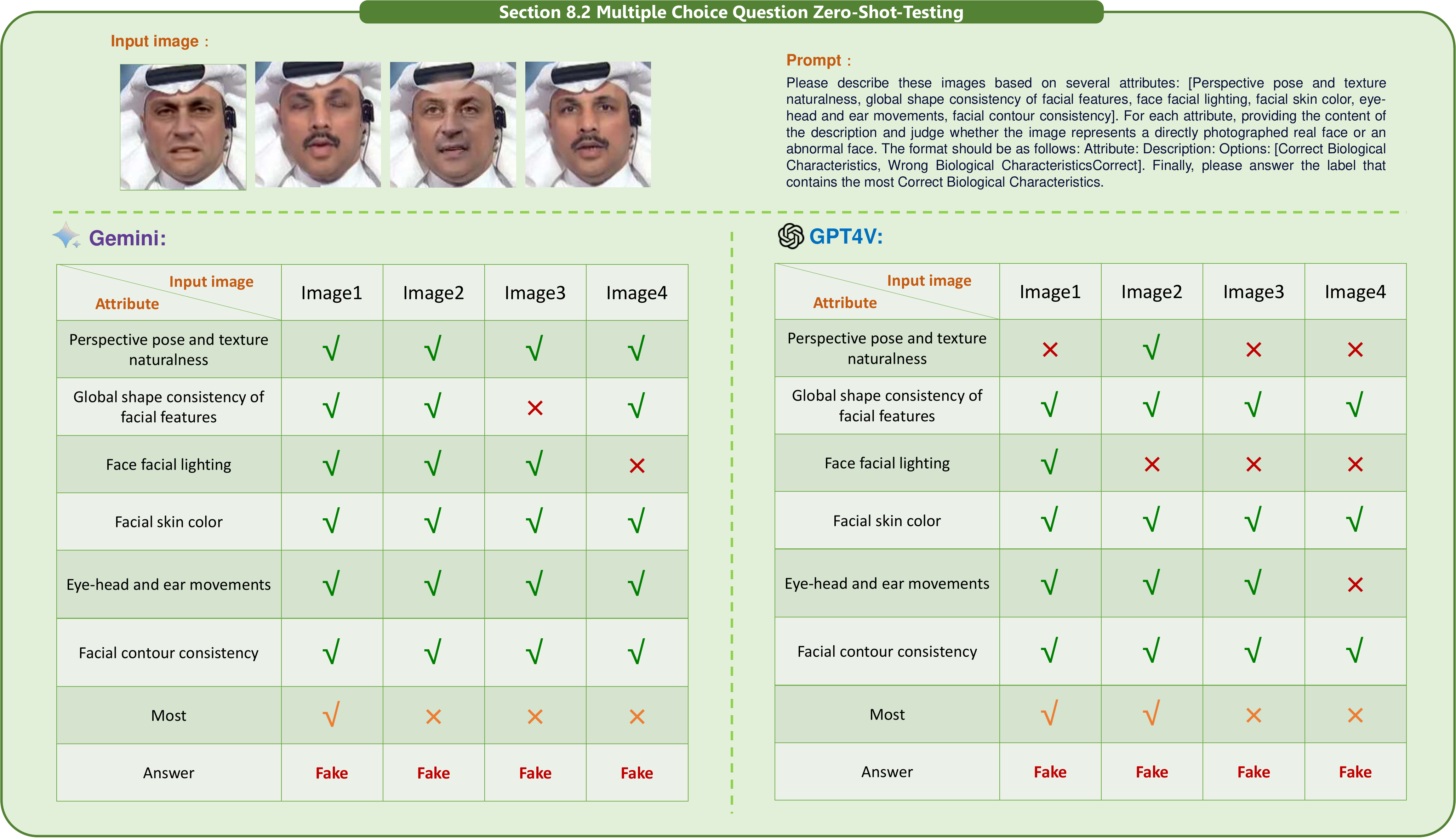}}
  \end{center}
  \caption{In this image,the same four common face forgery methods were used in this experiment. In order to compare the effectiveness of MA-COT, we performed a comparison between the COT method and the MA-COT hint method. In the COT method, GPT4V refuses to answer some of the attacks and fails to recognize Nulltextures forged faces and real faces, which the model incorrectly confuses due to their greater similarity. Whereas, with MA-COT, the detail analysis of GPT4V and Gemini is improved and combines multiple attributes for judgment.}
  \label{30}
\end{figure}

\begin{figure}[htbp]
  \centering
  \begin{center}
  \centerline{\includegraphics[width=1.0\linewidth]{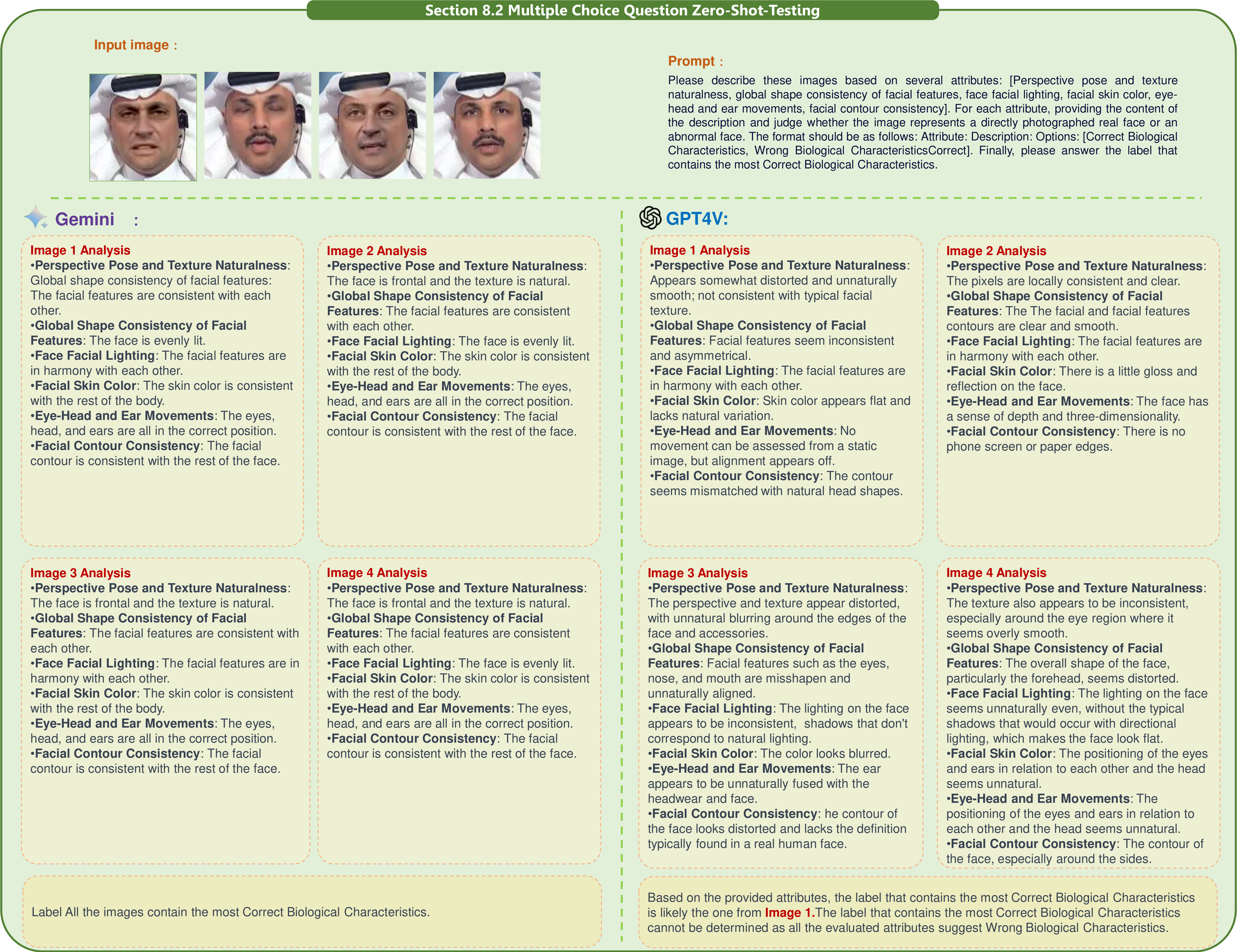}}
  \end{center}
  \caption{In this figure, the more detailed MA-COT responses of GPT4V and Gemini are shown, and it can be seen that after analyzing the multi-attribute integration, the responses of GPT4V are still more detailed, however, Gemini is more concise, and GPT4V still has forged images that are indistinguishable from the original images, which indicates that how to improve the model's ability to analyze the details of the realistically generated images still needs to be explored.}
  \label{31}
\end{figure}

\begin{figure}[htbp]
  \centering
  \begin{center}
  \centerline{\includegraphics[width=1.0\linewidth]{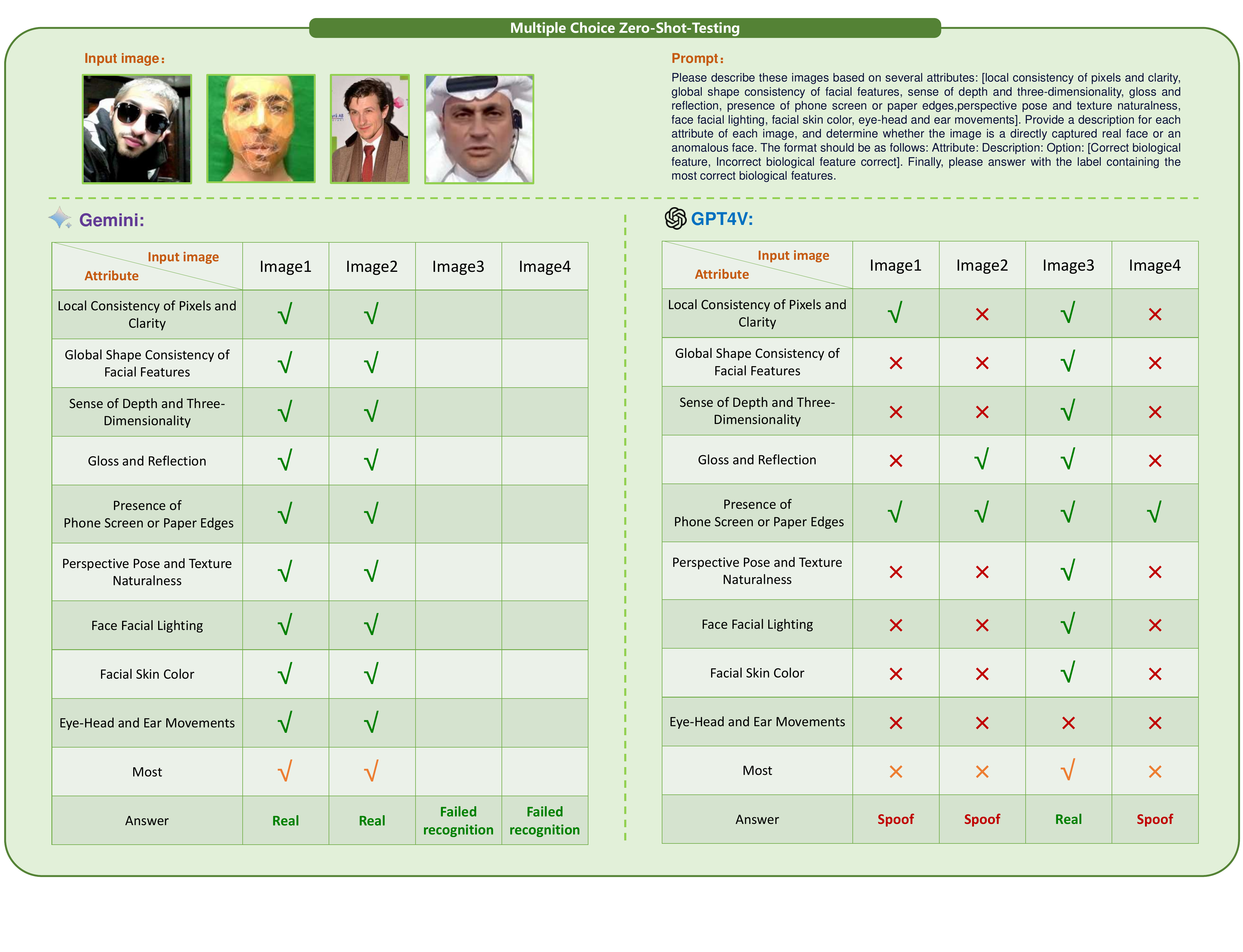}}
  \end{center}
  \caption{This figure illustrates the comparative outcomes using the same test samples as in Figure \ref{UnionTask_07}. In the original test, both Gemini and GPT4V incorrectly identified Image3 as a real face. After implementing the MA-COT querying method, GPT4V's performance did not significantly improve, still struggling to distinguish between real faces and those generated by diffusion techniques, highlighting its limitations in recognizing fine-grained modifications. Meanwhile, Gemini failed to recognize two images and mistakenly judged the two images it did recognize as real faces, indicating its shortcomings in multi-image recognition and in performing unified tasks.}
  \label{unified_macot_02}
\end{figure}

\begin{figure}[htbp]
  \centering
  \begin{center}
  \centerline{\includegraphics[width=1.0\linewidth]{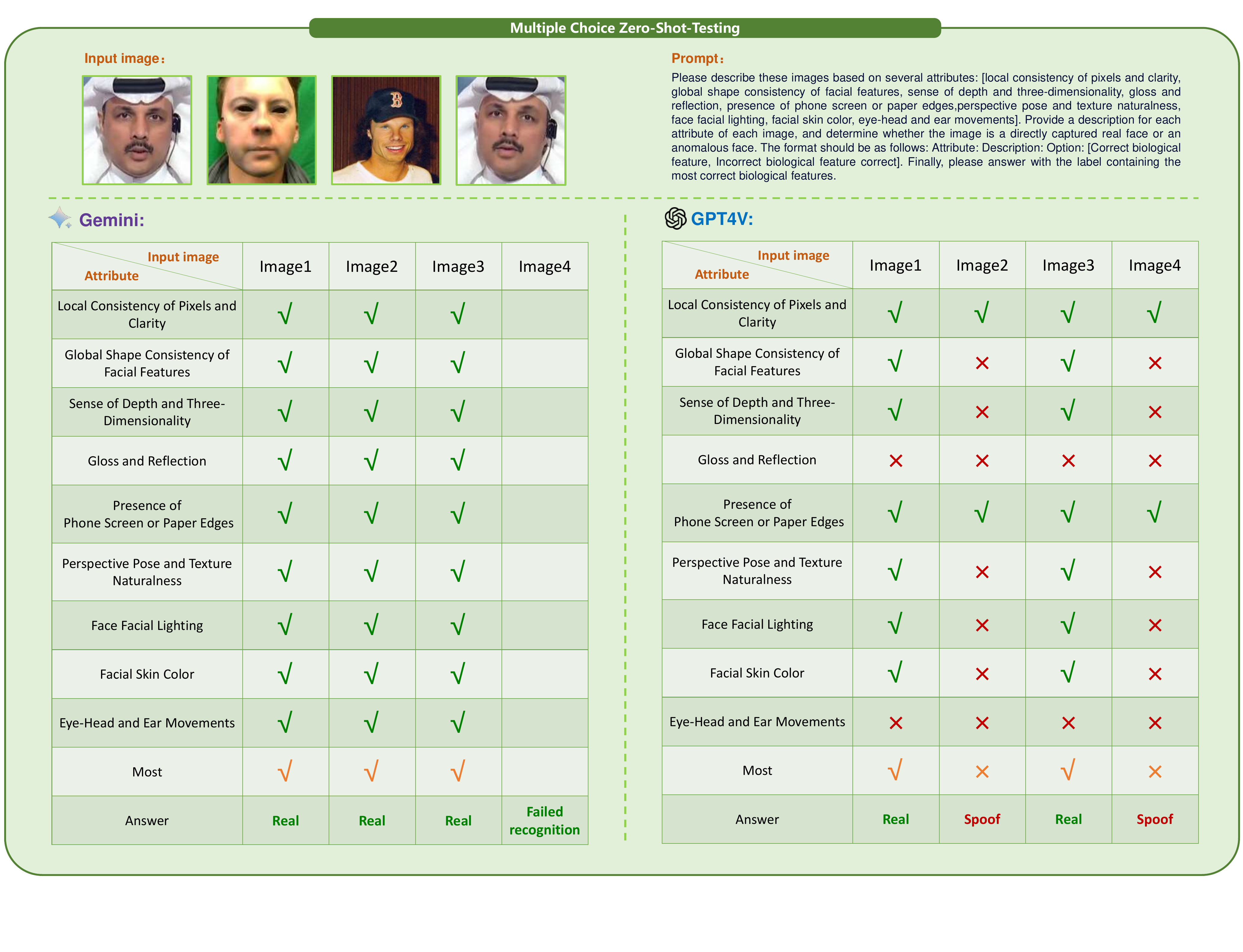}}
  \end{center}
  \caption{This figure presents the analysis results for test samples identical to those in Figure \ref{UnionTask_12}. Initially, Gemini was incorrect in both attempts, while GPT4V, after the introduction of COT, mistakenly believed that Image1 and Image 4 were real faces. After applying the MA-COT prompt method, GPT4V conducted a multi-attribute analysis and voting, selecting two answers, including one correct answer and mistakenly identifying a face image generated by Diffusion technology as real, highlighting the challenge in recognizing realistic face images produced by Diffusion. Meanwhile, Gemini failed to recognize one image and erroneously considered the three images it did analyze as real faces, revealing deficiencies in its image recognition and analytical capabilities.}
  \label{unified_macot_03}
\end{figure}

\end{appendix}